\documentclass[final=True,colorinlistoftodos,a4paper,12pt]{report}

\usepackage[a4paper,inner=3cm,outer=3cm, bottom=2cm, top=2cm]{geometry}
\usepackage[T1]{fontenc}
\usepackage[utf8]{inputenc}
\usepackage[english]{babel}
\usepackage[pagebackref]{hyperref}
\usepackage{bm,bbm} %
\usepackage{amsmath}
\usepackage{amsthm}
\usepackage{amssymb}
\usepackage{mathtools}

\usepackage[noabbrev,capitalize,nameinlink]{cleveref}
\usepackage{graphicx}
\usepackage{csquotes}

\usepackage{thmtools}
\usepackage[numbers]{natbib}
\usepackage{url}
\usepackage{nicefrac}
\usepackage[inline]{enumitem}
\usepackage{xspace}
\usepackage{xcolor}
\usepackage{algorithmic}
\usepackage{algorithm}

\usepackage{xpatch}
\usepackage{amartya_ltx}
\usepackage{fancyhdr}
\usepackage[obeyFinal]{todonotes}
\usepackage{booktabs}
\usepackage{amartya_ltx}

\let\oldnumberline\numberline%
\renewcommand{\numberline}[1][\negmedspace]{#1 \oldnumberline}
\makeatletter
\@for\thmt@envname:=\thmt@allenvs\do{ 
    \@xa\def\csname ll@\thmt@envname\endcsname{%
            \protect\numberline[\thmt@thmname]{\csname the\thmt@envname\endcsname}%
            \ifx\@empty\thmt@shortoptarg%
            \else
                \thmt@shortoptarg%
            \fi
        }
}
\makeatother

\usepackage{rotating}
\usepackage{float}
\usepackage{setspace}
\usepackage[bottom]{footmisc}
\usepackage{pifont}
\usepackage{multirow}
\usepackage{color}
\usepackage{xspace}
\usepackage{array}
\usepackage{wrapfig}
\usepackage{thm-restate}
\usepackage[labelformat=simple,font=small,]{subcaption}
\usepackage[font=small]{caption}

\usepackage[utf8]{inputenc} %
\usepackage{hyperref}       %
\usepackage{url}            %
\usepackage{amsfonts}       %
\usepackage{nicefrac}       %
\usepackage[compact,raggedright]{titlesec}

\titleformat{\chapter}[block]
  {\onehalfspacing\bfseries\raggedright}{\Huge\thechapter.}{1em}{\Huge}
\titlespacing*{\chapter}{0pt}{-19pt}{10pt}

\newcommand{\ifgsm}{Iter-FGSM\xspace}
\newcommand{\deepfool}{DeepFool\xspace}

\newcommand{\ill}{Iter-LL-FGSM\xspace}
\newcommand{\lr}{\(\mathsf{LR}\)-\(\mathrm{layer}\)\xspace}
\newcommand{\lrs}{\(\mathsf{LR}\)-\(\mathrm{layers}\)\xspace}

\newcommand{\prnk}[2]{\mathsf{\Pi}_{#1}^{\mathrm{rank}}(#2)}

\renewcommand{\th}{{\it th}}
\definecolor{DarkGreen}{rgb}{0, 0.4, 0}
\def\blankpage{%
      \clearpage%
      \thispagestyle{empty}%
      \addtocounter{page}{-1}%
      \null%
      \clearpage}

\newenvironment{dedication}
  {\clearpage           %
   \thispagestyle{empty}%
   \vspace*{\stretch{1}}%
   \itshape{}             %
   \raggedleft{}          %
  }
  {\par %
   \vspace{\stretch{3}} %
   \clearpage           %
  }

\usetikzlibrary{arrows,decorations.pathmorphing,backgrounds,fit,positioning,shapes.symbols,chains}
\tikzset{ node distance = 1cm, auto,font=\footnotesize,
tensors/.style={circle, rounded corners, draw=black, fill=black!10, inner sep=0.5pt, text width=1cm, text badly centered, minimum height=1.2cm,, font=\bfseries\footnotesize\sffamily},
temp_tensors/.style={circle, rounded corners, dashed, draw=black, fill=black!5, inner sep=0.5pt, text width=1cm, text badly centered, minimum height=1.2cm,, font=\bfseries\footnotesize\sffamily},
parameters/.style={align=center, text width=2cm, font=\bfseries\footnotesize\sffamily}}

\usepackage{glossaries}
\glsdisablehyper{}
\newglossaryentry{lip}
{
    name=lipschitz,
    description={Lipschitz constant}
}
\newglossaryentry{Lip}
{
    name=lipschitz,
    description={Lipschitz constant}
}

\newacronym{nn}{NN}{neural network}
\newacronym{sgd}{SGD}{stochastic gradient descent}
\newacronym{svd}{SVD}{Singular Value Decomposition}
\newacronym{sota}{SOTA}{state-of-the-art}
\newacronym{gan}{GAN}{Generative Adversarial Networks}
\newacronym{srn}{SRN}{Stable Rank Normalization}
\newacronym{srngan}{SRN-GAN}{Stable Rank Normalization GAN}
\newacronym{sngan}{SN-GAN}{Spectral Normalization GAN}
\newacronym{sn}{SN}{Spectral Normalization}

\usepackage{tikz}
\usetikzlibrary{arrows}
\usetikzlibrary{shapes.misc}
\usetikzlibrary{positioning}
\tikzset{cross/.style={cross out, draw, 
         minimum size=4,%
         inner sep=0pt, outer sep=0pt}}
     \definecolor{cadmiumgreen}{rgb}{0.0, 0.42, 0.24}

\graphicspath{{./figures/},{./figures/imagenet_high_inf/}}
\usepackage{pgfplots}
\pgfplotsset{compat=1.11}
\usepgfplotslibrary{fillbetween}
\usetikzlibrary{intersections}
\usetikzlibrary{patterns}
\usetikzlibrary{calc}

\tikzset{
    right angle quadrant/.code={
        \pgfmathsetmacro\quadranta{{1,1,-1,-1}[#1-1]}     %
        \pgfmathsetmacro\quadrantb{{1,-1,-1,1}[#1-1]}},
    right angle quadrant=1, %
    right angle length/.code={\def\rightanglelength{#1}},   %
    right angle length=2ex, %
    right angle symbol/.style n args={3}{
        insert path={
            let \p0 = (\((#1)!(#3)!(#2)\)) in     %
                let \p1 = (\((\p0)!\quadranta*\rightanglelength!(#3)\)), %
                \p2 = (\((\p0)!\quadrantb*\rightanglelength!(#2)\)) in %
                let \p3 = (\((\p1)+(\p2)-(\p0)\)) in  %
            (\p1) -- (\p3) -- (\p2)
        }
    }
}
\pgfdeclarelayer{bg}
\pgfsetlayers{bg,main}

\def\checkmark{\tikz\fill[scale=0.4](0,.35) -- (.25,0) -- (1,.7) -- (.25,.15) -- cycle;} 

\newcommand{\x}{\mathbf{x}}
\newcommand{\w}{\mathbf{w}}

\newcommand{\tikzmark}[1]{\tikz[overlay,remember picture] \node (#1) {};}

\newcommand\icu{\mu_{i\rightarrow}}

\title{Identifying and Exploiting Structures \\for Reliable Deep Learning}

\author{Amartya Sanyal}

\graphicspath{{./low_rank_folder/figs/},
{./low_rank_folder/images_adv/}, {./figs/}, {./stable_rank_folder/figs/},
{./stable_rank_folder/figs/st_pr_70_img/},
{./stable_rank_folder/figs/sn_pr_img/},
{./stable_rank_folder/figs/wgan_img/},
{./stable_rank_folder/figs/celeb_images_proper/},
{./stable_rank_folder/figs/sn_celeb_img_proper/},
{./adv_cause_folder/figures/},
{./adv_cause_folder/figures/imagenet_high_inf/},
{./calibration_main/figs}}

\begin{document}
\pagenumbering{gobble}
\begin{titlepage}
    \begin{center}
        \vspace*{1cm}
        
        \Huge
        \textbf{Identifying and Exploiting Structures \\for Reliable Deep Learning}

        \vspace{1.5cm}
        
        \begin{figure}[h!]
          \centering
          \def\svgwidth{0.35\columnwidth}
          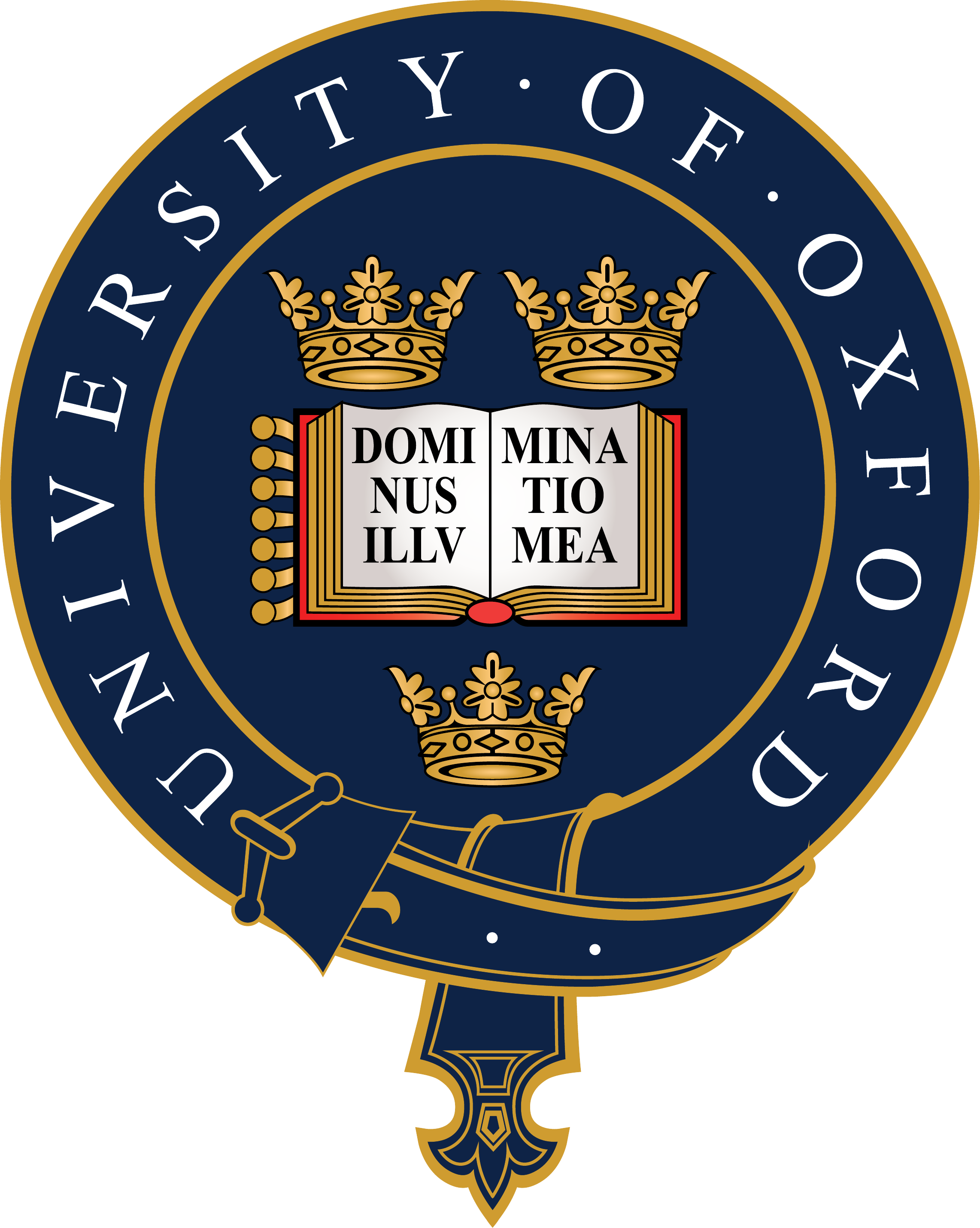
        \end{figure}
        \LARGE
        \vspace{1.5cm}
        \textbf{Amartya Sanyal\\}
        St Hugh's College
        \vspace{1.5cm}
        \vfill
        \normalsize
        A thesis submitted for the degree of\\
       {\em Doctor of Philosophy}\\
        \vspace{20pt}
        University of Oxford\\
        Trinity 2021.
        
    \end{center}
  \end{titlepage}
  \setcounter{tocdepth}{2}
  \setcounter{secnumdepth}{2}

\listoftodos
\begin{dedication}
  to the Weasleys~(even Percy)\\
  and Professor Shonku
\end{dedication}
\blankpage

\pagenumbering{roman}
\chapter*{Acknowledgements}%
\addcontentsline{toc}{chapter}{\numberline{}Acknowledgements\protect\vspace{-10pt}}%
My D.Phil would have been far less enjoyable without the many meetings,
supervision and otherwise, with my D.Phil advisors Varun Kanade and
Philip Torr. I am grateful to them for allowing me to pursue research questions
of my own choosing and putting up with me despite most of them failing. Varun's
ability to constantly see through the inconsequential details and isolate the
fundamental questions has been a constant surprise for me and something I hope I
have learnt from. I also thank Varun for all I have learnt from him about
computational learning theory and the many interesting chats about machine
learning which I believe has had a lasting impact on what questions I find
interesting. I thank Phil for constantly believing in me, spreading his
ever-present extremely infectious energy, eagerness to dive into new research
topics, seeing the bigger picture whenever I have proposed a new idea, and most
importantly always having time for me despite being a very busy person. I will
lie if I do not admit that I have always believed that he has a doppelganger at
his service. Combined together, I have had an unique opportunity to work in both
the theory and practice of machine learning.

Without the constant support and determination of my parents, this thesis~(and
most of my education) would have been impossible. For that and more, I am
forever indebted to them. I also owe it to everyone in my childhood including my
extended family, who has contributed to keep me interested in the mathematical
sciences. I should specifically thank Purushottam, Prateek, and all my teachers
at IITK who introduced me to machine learning in a way that I decided to pursue
it for my DPhil. I also thank my friends from IIT --- Bhuvesh, Asim, Nirbhay, Nihar, Sivasankar, and many more.

I thank Prof. Shimon Whiteson and Prof. Yarin Gal for examining my transfer and
my confirmation, respectively. I thank Prof. Pawan Kumar for taking the time to
be on my transfer, confirmation as well as my final thesis examination
committee. Their advice has been invaluable for guiding me at the respective
stages of my D.Phil. I also thank Prof. Pranjal Awasthi for agreeing to be on my
thesis committee. 

I owe a special thanks to Puneet for indulging all my crazy ideas over the last
three years, re-reading and re-writing my drafts, and being my longest and
closest research collaborator.  I have been very lucky to meet a number of
extraordinary researchers and collaborators on the way including Adri{\`a}
Gasc{\'o}n, Matt Kusner, Lorenzo Rosasco, Patrick Rebeschini, Edward
Grefenstette, Tim Rockt{\"a}schel, Stuart Golodetz, Tomas
Va{\u{s}}kevi{\u{c}}ius, and Mario Lezcano Casado. I hope I have learned a lot from each of them.

I thank David for keeping me constantly informed and interested in optimisation.
I have always enjoyed discussing with him our innumerable ideas. I thank Limor
for introducing me to causality and interpretability. Working on a paper with
David and Limor is on the very top of my todo list. TVG has been a wonderful
group to be a part of especially to be able work so closely with Jishnu, Viveka,
Tom, Pau, Harirat, and Yuge~(in chronological order). I have learnt a lot from
each of them and the diversity in their research expertise has made this a very
enriching experience.

St. Hugh's college~(along with its vast gardens) and the St. Hugh's MCR has been
the most wonderful place to live in and be a part of, especially during the
pandemic. I should specifically thank Alex for our many escapades and for always
saying a joke when I needed to hear it~(or not), Lizzie for being the best
neighbour and for pointing out how my cooked dishes are consistently brown, Josh
T for many things including the timely car rides to the McDonald's and KFC's,
Kat for being a constant reminder of the amazing powers of a cup of coffee,
Anwar, Florence, and Ed for being excellent reasons to go to pubs, socials, and
formals, Josh C for sharing my love for Bourdain, Sharan for his unlimited
energy and positivity, which often~(luckily) overflowed into me, Lili for making
dinners a social and delicious affair, Qian and Brandon for making my time at
Oxford worth remembering, Michelle and Yves for being the best running partners,
and Lukas and Arnaud for being the constant friends that they have been.

I also thank the Department of Computer Science at the University of Oxford for
providing me with a productive research atmosphere and workspace, as well as the
department of Engineering Science for allowing me to be their member. I cannot
thank Julie and Sarah enough for making all the admin work seem like a breeze.

This work was made possible by the generous support of the Turing doctoral
studentship from The Alan Turing Institute. I thank Sam, Georgia, Ben, and
everyone else at the Turing institute for making sure I had all the help I
needed. I also thank my colleagues at the Turing institute, especially the
Oxford cohort --- Florentine, Sanna, Prateek, Julien, Michael, Jessie,
Francesco, Charlie, and Ayman. 

Last but not the least I also want to thank everyone who made my life easier
during the COVID-19 pandemic --- everyone at  Brew, Nine2Nine, RnC, Vinny's
Cafe, Pepper's,  Ed Reid and everyone in Hugh's gardening team, the college
porters and many more. Finally, I also thank Bourdain, Hemingway and Salinger
for taking me away from my work when I most needed a break.

\newpage
\phantomsection
\addcontentsline{toc}{chapter}{\numberline{}Table of Contents\protect\vspace{-10pt}}%
\tableofcontents

\newpage
\phantomsection
\addcontentsline{toc}{chapter}{\numberline{}Abstract\protect\vspace{-10pt}}%
\chapter*{Abstract}%
Deep learning research has recently witnessed an impressively fast-paced
progress in a wide range of tasks including computer vision, natural language
processing, and reinforcement learning. The extraordinary performance of these
systems often gives the impression that they can be used to revolutionise our
lives for the better. However, as recent works point out, these systems suffer
from several issues that make them unreliable for use in the real world,
including vulnerability to adversarial attacks~(\citet{szegedy2013intriguing}),
tendency to memorise noise~(\citet{Zhang2016}), being over-confident on incorrect
predictions~(miscalibration)~(\citet{Guo2017}), and unsuitability for handling
private data~(\citet{G-BDL+:2016}). In this thesis, we look at each of these
issues in detail, investigate their causes, and propose computationally cheap
algorithms for mitigating them in practice. 

To do this, we identify structures in deep neural networks that can be exploited
to mitigate the above causes of unreliability of deep learning algorithms.
In~\Cref{chap:stable_rank_main}, we show that minimizing a property of matrices,
called stable rank, for individual weight matrix in a neural network
reduces the tendency of the network to memorise noise without sacrificing its
performance on noiseless data. 

In~\Cref{chap:causes_vul}, we prove that memorising label noise or doing
improper representation learning makes achieving adversarial robustness
impossible.~\Cref{chap:low_rank_main} shows that a low-rank prior on the
representation space of neural networks increases the robustness of neural
networks to adversarial perturbations without inducing any tradeoff with
accuracy in practice. 

In~\Cref{chap:focal_loss}, we highlight the use of focal loss, which weights
loss components from individual samples differentially by how well the neural
network classifies each of them, as an alternative loss function to cross-entropy for minimizing miscalibration in neural networks.

In~\Cref{chap:TAPAS}, we first define a new framework called \emph{Encrypted
Prediction As A Service}~(EPAAS) along with a set of computational and privacy
constraints. Then we propose the use of a Fully Homomorphic
Encryption~\citep{Gen:2009} scheme which can be used with a Binary
neural network~\citep{CHSEB:2016}, along with a set of algebraic and
computational tricks, to satisfy all our conditions for EPAAS while being computationally efficient.

\doublespacing
\newpage
\phantomsection
\addcontentsline{toc}{chapter}{\numberline{}Bibliographic Notes\protect\vspace{-10pt}}%
\chapter*{Bibliographic Notes}

The material in~\Cref{chap:stable_rank_main} is based
on~\citet{sanyal2020stable}, which is published as a Spotlight paper in
International Conference on Learning Representations~(ICLR), 2020 and is
co-authored with Dr. Puneet Dokania and my DPhil advisor Prof. Philip H.S. Torr.
The material in~\Cref{chap:causes_vul} is primarily based
on~\citet{Sanyal2020Benign}, which is published as a Spotlight paper in the
International Conference on Learning Representations~(ICLR)
2021.~\Cref{chap:low_rank_main} is based on the work in~\citet{Sanyal2020LR}.
Both of these works are co-authored with Dr. Puneet Dokania and my D.Phil
advisors Prof. Varun Kanade and Prof. Philip H.S. Torr. The material
in~\Cref{chap:focal_loss} is mostly based on the work
in~\citet{Mukhoti2020Focal}, which is published in the Advances in Neural
Information Processing Systems~(NeurIPS) 2020 and was also presented as a
Spotlight paper in the Workshop on Uncertainty and Robustness in Deep Learning,
held at the International Conference in Machine Learning~(ICML), 2020. The
material in~\Cref{chap:TAPAS} is based on~\citet{Sanyal2018a}, which was
published in the International Conference in Machine Learning~(ICML), 2018 and
is co-authored with Prof. Matt Kusner, Dr. Adri{\`a} Gasc{\'o}n and Prof. Varun
Kanade. I have also co-authored~\citet{Jorge2020Force}, which is published in
the International Conference on Learning Representations~(ICLR) 2021 but is not
included in this thesis.

~\Cref{sec:slt,sec:trml} are written solely by me but the original research is
not performed by me. The original sources of the material is cited in the
relevant parts and my contribution is in unifying it and presenting it
in the context of the original material presented in this thesis. \todo[color=green]{Mention
that I do have some contribution in Chap 3}I am the primary author and the sole
student author of the works presented
in~\Cref{chap:stable_rank_main,chap:causes_vul,chap:low_rank_main}.
The primary student authors of the work presented in~\Cref{chap:focal_loss} is
Jishnu Mukhoti and Viveka Kulharia. While, they have done most of the
experimental work, I have worked on the theoretical and conceptual part of this
work. I am also the sole student author of the work presented in~\Cref{chap:TAPAS}.

\newpage
\phantomsection
\addcontentsline{toc}{chapter}{\normalfont\numberline{}List of Theorems\protect\vspace{-10pt}}%
\listoftheorems[ignoreall,show={thm}]

\newpage
\phantomsection
\addcontentsline{toc}{chapter}{\normalfont\numberline{}List of Figures\protect\vspace{-10pt}}%
\listoffigures

\newpage
\phantomsection
\addcontentsline{toc}{chapter}{\normalfont\numberline{}List of Tables\protect\vspace{-10pt}}%
\listoftables

\newpage
\phantomsection
\addcontentsline{toc}{chapter}{\normalfont\numberline{}List of Algorithms\protect}%
\listofalgorithms

\clearpage

\newcolumntype{L}[1]{>{\raggedright\let\newline\\\arraybackslash\hspace{0pt}}m{#1}}
\newcolumntype{C}[1]{>{\centering\let\newline\\\arraybackslash\hspace{0pt}}m{#1}}
\newcolumntype{R}[1]{>{\raggedleft\let\newline\\\arraybackslash\hspace{0pt}}m{#1}}
\newcolumntype{H}{>{\setbox0=\hbox\bgroup}c<{\egroup}@{}} %

\cleardoublepage
\pagenumbering{arabic}
\chapter{Introduction}
\label{sec:intro}
    Machine learning research, and in particular deep learning research, has
    shown remarkable progress in recent years. On several tasks, machine
    learning algorithms have already surpassed human-level performance, albeit
    only on some specific measures of performance. For example, on a popular
    computer vision dataset {\em Imagenet},~\citet{he2015delving} obtained
    a~(top-5) classification error of \(4.9\%\) in 2015, surpassing the human
    benchmark of \(5.1\%\)~(reported in~\citet{imagenet}). Since then, there
    have been consistent improvements in classification accuracies on the
    dataset from an error of \(4.7\%\) in 2017~\citep{Chen2017}, to \(3.4\%\) in
    2019~\citep{Real2019}, and as low as \(1.2\%\) in 2020~\citep{Pham2020}. It
    is not just computer vision that has enjoyed such fast progress but also
    tasks like language modelling~\citep{Grave2017,Yang2018,Wang2019}, language
    translation~\citep{Luong2015,Wu2016,Gehring2017}, and  medical image
    segmentation~\citep{Ronneberger2015,Fan2020,Tomar2021}.

    The outcomes of these studies have not remained within the confines of
    research labs; they are now actively deployed in multiple sectors in the
    real world by governments and private companies alike. Automobile companies,
    including Tesla~\citep{tesla}, Waymo~\citep{waymo},
    GMCruise~\citep{gmcruise}, and
    ArgoAI~\footnote{https://www.theverge.com/2017/8/16/16155254/argo-ai-ford-self-driving-car-autonomous},
    are actively deploying deep neural networks in their autonomous driving
    systems. Companies like Pixsy~\citep{pixsy}, Tineye~\citep{tineye}, and
    Salesforce~\citep{salesforce} are among those who are using computer vision
    algorithms to search for images in large databases rapidly. In health care,
    companies like Infervision, Deepmind, and Microsoft Research are using deep
    learning algorithms to speed up, other-wise time-consuming, manual
    activities like detecting cancerous tissues in medical
    images~\footnote{https://techcrunch.com/2018/11/30/infervision-medical-imaging-280-hospitals/}\footnote{http://www.imperial.ac.uk/news/183293/research-collaboration-aims-improve-breast-cancer/}
    or segmenting medical scans during radiotherapy
    preparations~\footnote{https://www.cambridgeindependent.co.uk/news/addenbrooke-s-and-microsoft-research-create-world-first-ai-technology-to-speed-up-radiography-preparations-9147027/}.

    Online services like Tilde~\citep{tilde}, Bing~\citep{bing}, and
    Lengoo~\citep{lengoo} uses deep neural networks to offer language
    translation services. A prominent example of the use of deep learning by
    governments is the {\em EU Presidency Translator}~\footnote{https://www.eu2020.de/eu2020-en/presidency/uebersetzungstool/2361002}, developed by Tilde and
    used by Finland during its 2019 presidency of the EU council to allow
    delegates, and attendees to overcome language barriers and access
    multilingual information by automatically translating documents and local
    websites. Perhaps playing a more direct role in shaping our daily lives and
    world view are social media companies like Twitter and Facebook, who use
    deep learning algorithms to recommend social content and connections.

    This surge in deployment of deep learning algorithms is primarily due to its
    remarkable feats in performance-based metrics like the accuracy of
    prediction in classification tasks, {\em Intersection-Over-Union (IoU)} in
    image segmentation tasks, {\em BLEU score} in translation tasks, and {\em
    perplexity} in language modelling tasks. In turn, research in deep learning
    has also focused primarily on improving these performance-based metrics. A
    large portion of these improvements has come from designing novel
    regularisation methods and exploiting new inductive biases in deep learning
    algorithms. For example, regularisation techniques like Batch
    Normalization~\citep{Ioffe2015}, DropOut~\citep{Srivastava2014}, Label
    Smoothing~\citep{SzegedyVISW16}, and DropConnect~\citep{Merity2018} provided
    state-of-the-art accuracy in their respective tasks at the time of their
    development. Large performance increases have also come from incorporating
    the right inductive biases in learning algorithms. In deep learning,
    inductive biases have mainly been incorporated in the form of
    changes to the architecture of the neural network. Convolutional Layers~\citep{LBBH:1998} and Recurrent
    Layers~\citep{graves2013speech} provide domain-dependent inductive biases by
    introducing translation invariance and temporal invariance, respectively.
    The positional embeddings in transformer
    architectures~\citep{vaswani2017attention} allow the model to encode
    absolute and relative positions and positionally invariant relationships. In
    language models, using Byte Pair
    Encoding~\citep{Sennrich2015,VanMerrienboer2017} for words reflects the
    inductive bias that words composed of similar sub-words are related. All of
    these inductive biases have provided significant advances in the performance
    of machine learning algorithms. These techniques identify structures that
    can be helpful for the performance-based metric on the task at hand and then
    exploit it through a specially designed regulariser, data-augmentation, or
    architectural change.
    \section{Challenges with the reliability of deep learning}
    However, as recent works have pointed out, state-of-art deep neural
    networks, while providing impressive performance, suffer from a range
    of issues that hurt their reliability:\begin{itemize}
        \item~\textbf{Generalisation}:~\citet{Zhang2016} observed that neural
        networks, that achieve small test errors, are also equally capable of
        memorising a random labelling of the observed data. This is troubling as
        in deep learning, future performance on unseen data is usually estimated
        from performance on observed data. However, the experiments
        of~\citet{Zhang2016} indicate that such trust in a neural network
        performance on the observed data might be mis-placed. This calls for
        further research into certifying the generalisability of a learnt model
        on unseen data. 
        \item~\textbf{Robustness}:~\citet{szegedy2013intriguing} pointed out
        that deep neural networks are extremely vulnerable to imperceptible
        perturbations to input images. They show that, with minimal
        adversarially crafted distortions to input images, a network's error
        could be increased from \(2.1\%\) to \(100\%\) on a commonly used
        computer vision dataset called MNIST~\citep{LBBH:1998}. This
        vulnerability is referred to as adversarial vulnerability. Subsequent
        works have shown that this phenomenon is seen across a wide variety of
        deep learning models and datasets.
        
        \item~\textbf{Calibration}:~Deep neural networks, used for
        classification tasks, do not just output a class label but also a
        probability distribution over the possible classes. The probability
        value, associated with a class, is supposed to indicate the likelihood of
        that class being the correct output. When this property is satisfied by
        a machine learning model, the machine learning model is said to be
        calibrated. In 2005,~\citet{NiculescuMizil2005} showed that neural
        networks are well-calibrated compared to other prevalent classifiers at
        the time. However, in 2017,~\citet{Guo2017} showed that with increasing
        focus towards producing highly accurate classifiers, state-of-the-art
        neural networks are extremely miscalibrated and as a result, are over-confident on their incorrect predictions.
        \item~\textbf{Privacy}:~Apart from these issues, which relate to the
        quality of the prediction of a deep neural network, modern deep learning
        models have also been built without considering the privacy of the user.
        If these machine learning models are to be deployed in the real world to
        cater to the needs of a large number of people, the privacy of the users
        needs to be guaranteed. One such guarantee of privacy could be to allow
        users to use a machine learning model, operated by a private company, by
        sending over an encrypted version of their private data to the private
        company and receiving an encrypted result, computed by the machine
        learning model. However, this is computationally very expensive with
        modern deep learning architectures and thus unappealing for private
        companies who want their systems to have a high throughput, so that they
        can cater to a large number of requests.
    \end{itemize}
    
    In this thesis, we consider the above-mentioned issues with reliability in
    deep neural networks, identify metrics to measure them in practice and
    propose ways to mitigate them. To do this, we investigate specific causes
    for these vulnerabilities, pinpoint structures in the data and the model
    architecture, which can be exploited to mitigate the recognised cause, and
    design easy-to-use algorithms to implement to exploit these structures in
    practice and verify whether it improves on the metrics of reliability. Our
    approach is similar to what has been used to boost performance-based metrics
    of deep neural networks but instead of focusing on metrics like test
    accuracy, we look at metrics of reliability like avoiding memorisation,
    increasing adversarial robustness and calibration, and enabling privacy of
    data. In the next section, we briefly comment on how our contributions to each of these topics are organised in this thesis.

\section{Summary of contributions and organisation}
\label{sec:organization}
~\Cref{sec:slt} defines the main problem of learning from data and discusses the
importance of the various components of the learning problem. This chapter
provides the necessary background required to understand the rest of the thesis
and puts it in the context of the existing scientific research in the field of
machine learning.~\Cref{sec:trml} discusses the issues with the reliability of
deep learning methods, which are briefly mentioned in the previous section, in
greater detail along with a brief survey of the existing works in this field.

In~\Cref{chap:stable_rank_main}, we look at the problem of generalisation in
deep learning. Inspired by recent theoretical research~\citep{neyshabur2018a},
we identify a structure called stable rank~\citep{Rudelson2007}, measured for
individual weight matrices of a neural network, as being important for the
generalisation of deep neural networks. Then we propose an algorithm called
Stable Rank Normalization~(SRN), with theoretical guarantees, to control the
stable rank of neural networks. Finally, we devise principled experimental
methods to measure generalisation in practice and show, through experiments on a
wide range of datasets and architectures, that SRN indeed improves the
generalisation of deep neural networks. 

~\Cref{chap:causes_vul} identifies two specific reasons for the lack of
adversarial robustness in machine learning models. In particular, we show
theoretically and experimentally how overfitting label noise can give a false
sense of security in that it might not increase test error but can drastically
increase adversarial error. We identify the second cause to be improper
representation learning and show that using incorrect representations can get
low test error but can never get small adversarial error whereas using a
different representation can achieve both low test and adversarial error
simultaneously even without needing more data. Continuing from this,
in~\Cref{chap:low_rank_main}, we show that a low-rank prior on the
representation space of neural networks, if applied properly, can impart better
adversarial robustness for deep neural networks. We propose a computationally
efficient algorithm that scalably learns a neural network with low rank
representations without significant modifications to the architecture. Along
with a large boost to adversarial robustness, while maintaining test accuracy,
our experiments show that this has implications for compression of the learned
representations and the model as well.

We find that minimising the cross-entropy loss function minimises the difference
between the softmax distribution and the one-hot encoding of the labels for all
samples, irrespective of how well the model classifies individual samples.
This leads to a phenomenon referred to as {NLL overfitting}~\citep{Guo2017},
which is responsible for miscalibration in deep neural networks.
In~\Cref{chap:focal_loss}, we propose the use of  an alternative loss function,
popularly known as \textit{focal loss} \citep{Lin2017}, that tackles this by
weighting loss components generated from individual samples by how well the
model classifies each of them. We show, through experiments on a diverse set of
datasets and model architectures, that focal loss is much better than its
competitors at producing calibrated models without sacrificing test accuracy.

In~\Cref{chap:TAPAS}, we look at the problem of preserving the privacy of the
data while being able to do prediction on it using a machine learning model. To
this end, we define a framework called {\em Encrypted Prediction As A
Service~(EPAAS)} along with a set of privacy and computational requirements that
an EPAAS framework should satisfy. We find that a cryptography protocol called
{\em Fully Homomorphic Encryption~(FHE)}~\citep{Gen:2009} while being perfectly
suited to satisfy our conditions for EPAAS, is computationally very expensive
when applied to deep neural networks. We propose the use of binary neural
networks, which along with a set of algebraic and computational tricks, makes
the application of FHE on deep neural networks feasible. We show,
experimentally, that our approach suffers very little in terms of accuracy while
satisfying all the criterion of EPAAS. 

~\Cref{sec:math-prer-notat} lists some mathematical preliminaries including
inequalities and definitions that are used elsewhere in the thesis and is meant
to save time for the reader by not having to look up external
resources.~\Cref{sec:expr-settings} describes the datasets and neural network
architectures, along with details of training algorithms, that have been used in
the main chapters of the thesis. These are mostly details a deep learning
practitioner would be familiar with but have been provided for the sake of
completeness and
reproducibility.~\Cref{app:stable_rank,sec:appendix_adv_causes,app:low_rank,app:focal_loss}
contains formal proofs and additional figures and tables for the material
presented
in~\Cref{chap:stable_rank_main,chap:causes_vul,chap:low_rank_main,chap:focal_loss}
respectively.

\paragraph{Notations}
\label{sec:notation}

All vectors are denoted with lower case bold $\vec{v}, \vec{x}$ letters and all
matrices are represented by upper case bold $\vec{A}, \vec{B}$ letters. Some
commonly used vectors are $\vec{x}_i$ to represent the \(i^{\it th}\) example in
the dataset and, \(\vec{z}_l\) and \(\vec{a}_l\) to represent the pre-activation
and post-activation vectors of the \(l^{\it th}\) layer in a neural network,
respectively. Commonly used matrices include \(\vec{W}_l\) to represent the
weight matrix of the \(l^{\it th}\) layer and \(\vec{A}_l\) to represent the
matrix of activations for the \(i^{\it th}\) layer on the entire dataset. A
collection of objects is represented by upper case scripted $\cX,~\cY$ letters.
An example of a collection is a set  or a vector space. When we define a random variable to
take a value from a set, with a slight abuse of notation, we will also denote
that random variable with the same notation.

Any lemmas or theorems that are borrowed from existing literature are numbered
alphabetically~(eg.~\Cref{lem:ineq:frob_sing} and ~\Cref{thm:basic-gen-thm}) and any
result that is original to this thesis are numbered
numerically~(eg.~\Cref{lem:upper-bound-nn-lip} and ~\Cref{thm:srankOptimal}).
Unless otherwise stated, \(d\) represents the dimensionality of the input domain
of a machine learning model, \(k\) represents the number of classes in a
multi-class classification problem, \(\eta\) represents the uniform label noise
rate, \(L\) represents the depth of a neural network and \(W\) represents the
width i.e. the maximum width of any of its layers. The rest of the notations that are specific to each chapter are defined within the chapter itself.

\chapter{Preliminaries of Learning Theory}
\label{sec:slt}
\section{The Learning from Data Problem}
\label{sec:clt}

This chapter will first discuss the various components of the {\em Learning from
data} problem and then formally state the learning problem.
Let $\cX$ and $\cY$ be two random variables defined over the instance space and
the label space, respectively. With a slight abuse of notation, we use the same
notations \(\cX\) and \(\cY\) to refer to the instance space and the label
space, as well. For most problems in this thesis, $\cX$ will be a compact domain
in $\reals^d$ where $d$ is the dimension of the problem and $\cY$ will be an
unordered finite set. %
In an image classification problem, the space of vector representations of
natural images is an example of the instance space %
and the set of one-hot representations of the classes in a multi-class
classification problem is an example of the label space. A distribution $\cD$,
commonly referred to as the {\em data distribution}, is defined on the product
space $\cX\times\cY$. Importantly, this distribution is fixed but unknown to the
learner.~A finite sample
$\cS_m=\bc{\br{\vec{x}_1,y_1},\ldots,\br{\vec{x}_m,y_m}}$, referred to as a
{\em training dataset}, is created by sampling $m$ independently and identically
distributed~(i.i.d.) points from $\cD$. 

The next component of the learning problem is the  {\em hypothesis class} or the
{\em concept class }$\cH$ over the instance space $\cX$. The hypothesis class
over $\cX$ is defined as the space of functions  from $\cX$ to $\cY$ i.e.
$f:\cX\rightarrow\cY$. An element of this set is referred to as a hypothesis or
a concept. Examples of $\cH$ include all linear separators and all neural
networks that have a fixed architecture. For example, the class of multi-layer
perceptrons defined below is an example of a hypothesis class and one that we
will use throughout the rest of the thesis.

\begin{restatable}[Multi-Layer Perceptron]{defn}{mlpnet}\label{defn:net-mlp} A
    Multi-Layer Perceptron~(MLP) is a hierarchial model with a sequence of
    $L$ layers. The $i^{\it th}$ layer is parameterized by a
    matrix~$\vec{W}_i\in\reals^{\ell_{i-1}\times\ell_i}$ where \(\ell_i\)
    represents the width of the \(i^{\it th}\) layer and \(W=\max_{1\le L}
    \ell_i\) is the width of the MLP. The hypothesis class of MLPs
    \(\cM\) is 
    \[\cM: \bc{h_\theta~\vert 
    h_\theta\br{\vec{x}}=
    {\vec{W}_1\phi_{1}\br{\cdots\phi_{l-1}\br{\vec{W}_{L-1}\phi_L\br{\vec{W}_L\vec{x}}}}}, \theta=\bc{\vec{W}_1,\cdots,\vec{W}_L}}
    \]
      
    Each $\phi_i$ is an activation function and can all be identical or
    different depending on the network configuration. In practice, they are
    usually the same. When $\phi$ operates on a vector, it operates element-wise
    on each element of the vector.
  \end{restatable}

Finally, we also define the {\em loss function}
$\ell:\cY\times\cY\rightarrow\reals$, which is used to measure the goodness of
the functions in the hypothesis class. Intuitively, a loss function denotes the
price we pay when a function $f\in\cH$ sees $\vec{x}\in\cX$ and guesses the
output to be $f\br{\vec{x}}\in\cY$ when it is actually $y\in\cY$. The price paid
is small when $f\br{\vec{x}}$ and $y$ are close and large when they are
far. The goodness of the
function $f\in\cH$ on the whole distribution is measured with the notion of
Expected Risk.

\begin{restatable}[Expected risk]{defn}{emprisk}
    \label{defn:exp_loss}
    For a distribution $\cD$, loss function $\ell$, and a hypothesis $h\in\cH$,
    the expected risk is defined as \[\riskOne{h;\cD,\ell} =
    \int_{\vec{z}=\br{\vec{x},y}\sim\cD} \ell(h\br{\vec{x}},y)~dD(\vec{z})\]
    When we omit $\ell$ from the definition of the expected risk, we will refer to the expected error for the $0/1$ loss function $\ell_{0,1}\br{y,\hat{y}}=\bI\bc{y\neq \hat{y}}$
  \[  \risk{\cD}{f}=\riskOne{h;\cD,\ell_{0,1}}=\bP_{\br{\vec{x},y}\sim\cD}\bs{f\br{\vec{x}}\neq y},\]
  \end{restatable}

Now, we are ready to define the learning problem. Given a hypothesis  class
$\cH$, a loss function $\ell$, and a  training set $\cS_m$ consisting of $m$
i.i.d points sampled from an unknown but fixed {\em data distribution} $\cD$,
the objective of a learning algorithm $\cA$ is to return a hypothesis
$h=\cA\br{\cS_m}\in\cH$ that, with  high probability, minimizes the expected
risk $\riskOne{h;\cD,\ell}$. For simplicity, we will omit $\cD$ and $\ell$ from
the definition of expected risk when they will be clear from the context.

\subsection{PAC Learning}
We will provide the formal definition of the learning problem using the Probably
Approximate Correct~(PAC)~\citep{Valiant1984} learning framework. Let the
distribution $\cD_{\cX}$ be the restriction of the distribution $\cD$ on the
instance space $\cX$ such that there is a {\em target} concept class $\cC$ along
with a target concept $c\in\cC$, which labels the data drawn from $\cD_{\cX}$.
The objective of the learning problem is to  capture $c$ as closely as possible.
To highlight that the generation of the training dataset is a random process and
that depends on the target concept $c$, the training dataset is obtained as
follows. First, an instance $\vec{x}\in\cX$ is sampled from $\cD_{\cX}$ and is
then assigned a label by the target concept $c$. This process is referred to
as a call to the example oracle, denoted as $EX\br{c;D_{\cX}}$. Therefore,
drawing an $m$-sized dataset $\cS_m$ can be simulated with $m$ calls to
$EX\br{c;\cD_{\cX}}$.

\begin{defn}[PAC-Learnability] \label{defn:pac-learning}
    A concept class $\cC$ is said to be PAC-learnable using the hypothesis class $\cH$, if there exists a~(possibly randomized) algorithm $\cA$ such that the following holds true. For every $c\in\cC$, for every distribution $\cD_{\cX}$ over $\cX$, for every $0<\epsilon,\delta<1$, if $\cA$ is given access to $EX\br{c;\cD_{\cX}}$ and is given as input $\epsilon,\delta$, then $\cA$ makes $m$ calls to  $EX\br{c;\cD_{\cX}}$ and returns $h\in\cH$ such that with probability at least $1-\delta$, $\riskOne{h;\cD}\le\epsilon$. The probability is over the randomization in the calls to the example oracle and the internal randomization of $\cA$.
     
The number of calls made to $EX\br{c;\cD_{\cX}}$  is referred to as the sample complexity~(denoted by $m$)  and must be bounded by a polynomial in $\frac{1}{\epsilon},\frac{1}{\delta}$, and some parameters depending on the size of $c$ and the size of the instance space~$\cX$. Further, all hypotheses in $\cH$ must also be evaluable in polynomial time~(in the size of the input).
\end{defn}

Without diving too deep into the details of PAC Learning\footnote{For a more
detailed expose into this topic, please refer to the excellent lecture notes
in~\url{http://www.cs.ox.ac.uk/people/varun.kanade/teaching/CLT-HT-TT2021/lectures/CLT.pdf}},
we will discuss the importance of the various components of the PAC learning
framework. It is important to note at this stage that our main goal in this
thesis is to control not just the classification error, but also to improve
certain additional properties of reliability~(which will be discussed in detail
in~\Cref{sec:trml}) in machine learning. Thus, our discussion will diverge, at
places, from the original PAC learning framework even though the essential
components like sample complexity, error and confidence parameter and choice of
hypothesis class remain equally relevant. Below, we discuss some of the
important points.
 
\begin{enumerate}
    \item  The sample complexity for learning should be relatively small i.e.
    polynomial in problem parameters. When learning from real-world data, it is
    usually not possible to see more than a fixed number of samples, so it might
    not be practical to expect our learning algorithm to achieve arbitrarily low
    error or very high confidence in the low data regime. Thus, it is important
    to understand what error and confidence are achievable in the low data
    regime. 
    
    Further, when along with classification error the learning problem  also
    requires controlling an additional metric of reliability like adversarial
    vulnerability or mis-calibration, the sample complexity of the problem might
    increase. We will identify theoretically and experimentally when this
    increase in sample complexity happens and if necessary, design learning
    algorithms where this increase is small or absent.
    
    \item Choosing the hypothesis class $\cH$ that is used to learn the target
    concept $c$  is perhaps the most relevant component of PAC learning for this
    thesis. This is important for multiple reasons. First, the sample complexity
    of learning the same target concept can be  large or small depending on the
    choice of hypothesis class it is being learnt from. This is discussed
    further in~\Cref{sec:erm}.
    
    Second, we are looking to control not just the classification error but also
    other metrics of reliability. Depending on the choice of hypothesis class,
    it might be impossible to control both the test error and the metric of
    reliability simultaneously. Even in situations where they can both be
    controlled simultaneously, it might lead to an increase in sample
    complexity. We will investigate which combined choice of an additional
    metric of reliability and hypothesis class can pose such
    problems.%
    
\end{enumerate}

\subsection{Learning with noise}
\label{sec:lbl-noise-pac}
One of the important restrictions of the PAC learning framework, as defined
in~\Cref{defn:pac-learning}, is the assumption that the example oracle always
faithfully returns unblemished examples drawn from the target distribution and
labels them according to the target concept. Often, for the data given to
training algorithms, this is too strict a requirement as label noise is
ubiquitous in real-world data. Such noise can arise from multiple sources
including a malicious or careless data annotator, faulty communication equipment
transmitting the data, or faulty recording equipment recording the data. This
necessitates the design of a formal framework that can capture noise in the data
generation process. The noisy PAC learning framework of~\citet{Angluin1988} is
an example of such a framework. 

The important change from the noiseless framework is in the way data is
generated using the example oracle. Let $0\le\eta<\frac{1}{2}$ be the noise
rate. A simple framework to simulate the noise is through the Random
Classification Noise~(RCN) oracle denoted as $EX_{\eta}\br{c;\cD_{\cX}}$. To
generate an instance using this oracle, first, an instance is generated using the original example oracle $EX\br{c;\cD_{\cX}}$ and then with probability
$\eta$ the label is flipped to an arbitrary incorrect label.

PAC-Learnability with Random Classification Noise is defined exactly as PAC
Learnability but with the example oracle $EX(c;\cD_{\cX})$ replaced with the
noisy example oracle $EX_{\eta}\br{c;\cD_{\cX}}$ and $\frac{1}{1-2\eta}$ added to the set of parameters the sample complexity should be polynomial in.

\begin{defn}[PAC-Learnability with Random Classification Noise] \label{defn:noisy-pac-learning}
    A concept class $\cC$ is said to be PAC-learnable in the presence of random classification noise using the hypothesis class $\cH$, if there exists a~(possibly randomized) algorithm $\cA$ such that the following holds. For every $c\in\cC$, for every distribution $\cD_{\cX}$ over $\cX$, for every $0<\epsilon,\delta<1$, and for every $0\le\eta< \frac{1}{2}$, if $\cA$ is given access to $EX_{\eta}\br{c;\cD_{\cX}}$ and is given as input $\epsilon,\delta$ and $\eta$, then $\cA$ makes $m$ calls to  $EX_\eta\br{c;\cD_{\cX}}$ and returns $h\in\cH$ such that with probability at least $1-\delta$, $\riskOne{h;\cD}\le\epsilon$, where the probability is over the randomization in the calls to the example oracle and the internal randomization of $\cA$.
     
The number of calls made to $EX\br{c;\cD_{\cX}}$, referred to as the sample
complexity~(denoted with $m$), must be bounded by a polynomial in
$\frac{1}{\epsilon},\frac{1}{\delta},\frac{1}{1-2\eta}$ and some parameters
depending on the size of $c$ and size of the instance space~$\cX$. Further, all
hypotheses in $\cH$ must also be evaluable in polynomial time~(in the size of
the input).
\end{defn}

While the presence of Random Classification Noise~(RCN) is an added challenge,
several hypothesis classes that are PAC learnable in the original definition of
PAC learning~(c.f.~\Cref{defn:pac-learning}) have also been shown to be PAC
learnable with random classification noise. One of the only problems for which
no algorithm is known in the noisy setting of~\Cref{defn:noisy-pac-learning} but
is known to be learnable in the noiseless PAC  setting
of~\Cref{defn:pac-learning} is the parity problem~\citep{Kearns1998}. However,
even for problems that are learnable in the noisy setting, the learning
algorithm might be different from the noiseless setting.
In~\Cref{chap:stable_rank_main}, we discuss and provide experimental evidence
for how a learning algorithm can impose additional structure on the hypothesis
class to aid learning in the presence of random classification noise.

When the learning task must also satisfy additional metrics of
reliability like robustness and calibration, the presence of random classification noise can pose further
challenges. In~\Cref{chap:causes_vul}, we discuss this for the specific case of
adversarial robustness.

\section{Empirical Risk Minimisation}
\label{sec:erm}

The most common approach to solving the learning problem described above is the
Empirical Risk Minimisation~\citep{vapnik1992principles} approach. In this
approach, instead of minimizing the expected risk, which is harder as $\cD$ is
unknown, the learning algorithm instead minimises the empirical risk
$\empRisk{h, \ell}{N}$~(defined in~\Cref{defn:emp_loss}) on the observed
training dataset. Like the expected risk, we will ignore $\ell$ from the
definition of empirical risk when it will be clear from context\footnote{Please refer to~\cite{vapnik1998statistical} for a more detailed
description of statistical learning theory that discusses the properties of
minimizing the empirical risk.}.

\begin{defn}[Empirical risk]
  \label{defn:emp_loss}
  If the training set consists of $\{(\vec{x}_i, y_i)\}$ for $i \in \{1\cdots
  N\}$ then the empirical loss is defined as
  \[\empRisk{h}{N} = \frac{1}{N} \sum_{i=1}^N \ell(h\br{\vec{x}_i}, y_i)\]
\end{defn}

The theory of minimizing this empirical risk is usually tackled via the theory
of optimisation. The main purpose of the theory of optimisation is to find a
model $h$ and guarantee that for some small $\epsilon$, the empirical risk of
$h$ is close to the best possible empirical risk i.e.
\[\empRisk{h}{N} - \min_{h\in\cH}\empRisk{h}{N}\le \epsilon.\]

However, the  task of \emph{learning} requires us to perform well not only on
the seen dataset $\cS_N$ but also on unseen data from the entire distribution
$\cD$, which is measured using the expected risk from~\Cref{defn:exp_loss}.
The theory of generalisation guarantees with a high probability, that for any
hypothesis $h$ in the hypothesis class $\cH$, the empirical and the expected
risk are close i.e. 
\[|\empRisk{h}{N} - \riskOne{h}|\le \sup_{h\in\cH} |\empRisk{h}{N} - \riskOne{h}| = \zeta(N; \cH), \]
where $\zeta(N; \cH)$ decreases with increasing $N$.

The main focus of the theory is two-fold:
\begin{itemize}
\item To guarantee that with an increasing number  $N$ of observations, the difference of the two quantities goes to zero for
any model i.e. $\lim_{N\rightarrow\infty}\zeta\br{N,\cH}\rightarrow 0$.
\item To estimate the difference $\zeta\br{N;\cH}$, when
only a finite number $N$ of samples are available.
\end{itemize}

It turns out that the second property is more important for our purposes when
dealing with neural networks. Not only does it have more practical relevance,
but while the first property almost always holds~\footnote{The asymptomatic
behaviour of the empirical error functions is captured by the Glivenko–Cantelli
theorem as the number of i.i.d samples goes to infinity}, the second property is
mathematically more complicated especially for complex hypothesis classes like
neural networks. For example, if the empirical and expected risks are bounded
between $0$ and $1$, which is a very reasonable assumption, then any value of
$\zeta\br{N;\cH}$ that is greater than $1$ is a vacuous upper bound. The main goal
of generalisation theories is to get accurate estimates of $\zeta\br{N,\cH}$ for
moderate values of $N$ and some useful hypothesis class $\cH$. This is usually
done through theorems like~\Cref{thm:basic-gen-thm} referred to as the
\emph{basic theorem of generalisation}. However, this is not necessarily the
only way~(c.f.~\citet{bousquet2002stability}).
\todo[color=blue]{Convert \(h\) to \(\cH\) in the theorems}
\begin{thmL}[Basic Theorem of Generalisation]\label{thm:basic-gen-thm}
Let $\cH$ be a set of hypotheses from the space $\cX$ to $\bc{0,1}$ and
the rest of the terms be as defined in the previous sections. Then
$\forall \delta \in\br{0,1}$, with a probability of at least $1 - \delta,
~\forall h\in\cH$  
  \[\abs{\empRisk{h}{N} - \riskOne{h}} \le \zeta\br{N;\cH}=
  \tildeO{\dfrac{\sqrt{C(\cH)}}{N}}, \] where $C(\cH)$ denotes a measure of
  complexity for the hypothesis class \(\cH\)\footnote{For example, if \(\cH\)
  is a finite set of hypotheses, then the cardinality function is a valid instance of \(C\br{\cdot}\). We will look into more examples of the
  complexity function in~\Cref{sec:trml}.}. $\tildeO{\cdot}$ is a big-Oh notation
  that omits logarithmic terms.
\end{thmL}

Using this theorem, we can argue that empirical risk minimisation yields a hypothesis whose expected risk is close to the best possible expected risk as long as the complexity of the classifier is low. Let the hypothesis returned by the algorithm $\cA$ be $h_L\in\cH$. We know
that $\empRisk{h_L}{N}\le \min_{h\in\cH} \empRisk{h}{N} +
\epsilon$ for some $\epsilon$. Define
$h^* = \argmin_{h\in\cH}\riskOne{h}$.

 According to Theorem~\ref{thm:basic-gen-thm}, with probability
 at least $1 - \frac{\delta}{2}$ each of~\Cref{eq:use-gen-thm-1,eq:use-gen-thm-2} holds
 \begin{equation}\label{eq:use-gen-thm-1}
  \riskOne{h_L} -  \tildeO{\dfrac{\sqrt{C(\cH)}}{N}} \le \empRisk{h_L}{N} \le  \riskOne{h_L} +  \tildeO{\dfrac{\sqrt{C(\cH)}}{N}}
 \end{equation} and
  \begin{equation}\label{eq:use-gen-thm-2}
    \riskOne{h^*} -  \tildeO{\dfrac{\sqrt{C(\cH)}}{N}} \le \empRisk{h^*}{N} \le  \riskOne{h^*} +  \tildeO{\dfrac{\sqrt{C(\cH)}}{N}}.
  \end{equation}

 Thus,
 \begin{align*}
   \riskOne{h_L} &\le  \empRisk{h_L}{N} + \tildeO{\dfrac{\sqrt{C(\cH)}}{N}}\\
              &\le  \min_{h\in\cH} \empRisk{h_L}{N} + \epsilon +  \tildeO{\dfrac{\sqrt{C(\cH)}}{N}}\\
              &\le \empRisk{h^*}{N} + \epsilon + \tildeO{\dfrac{\sqrt{C(\cH)}}{N}}\\
              &\le  \riskOne{h^*} +  \tildeO{\dfrac{\sqrt{C(\cH)}}{N}} + \epsilon +  \tildeO{\dfrac{\sqrt{C(\cH)}}{N}}.
 \end{align*}
Therefore,  with probability at least  $\br{1 - \delta}$ 
\begin{equation}\label{eq:gen-erm}
  \riskOne{h_L} -  \riskOne{h^*} \le 2\tildeO{\dfrac{\sqrt{C(\cH)}}{N}}
+ \epsilon.
\end{equation} 
As discussed before,
it is the purpose of optimisation algorithms to minimise $\epsilon$.
In fact, neural networks usually obtain $\epsilon=0$ as has been
shown empirically in multiple papers including the seminal paper
of~\citet{Zhang2016}. So, the real difficulty is in coming up with
estimations of $C\br{\cH}$, that are practically useful and not
vacuous.

\subsection{ERM in the presence of Label Noise}

Label noise is ubiquitous  in real-world data. In fact, common datasets like
MNIST, CIFAR10, and CIFAR100 contain label noise in the training dataset as we
show in~\Cref{chap:causes_vul}. In~\Cref{defn:noisy-pac-learning}, we discussed
a variant of PAC learning which allows for noise in the data generation process
and we made a note that most problems that are learnable in the noiseless
setting are also learnable in the noisy setting. However, even in cases where
the problem is learnable in the noisy setting, it does not automatically imply
that the algorithm for learning in the noiseless setting is still adequate for
learning in the noisy setting. Therefore, understanding whether algorithms and
frameworks that are useful for learning in the noiseless setting can also be
used in the noisy setting and identifying the associated risks is important for deploying learning algorithms in the real world.

Empirical Risk Minimisation~(ERM) is the most commonly used framework for
learning in the noiseless setting and enjoys formal generalisation guarantees as
shown in~\Cref{eq:gen-erm}. However, the guarantees do not immediately apply to
learning in the presence of noise. To see this, note that the distribution, the
noisy example oracle samples from in~\Cref{defn:noisy-pac-learning}, is
different from the original distribution $\cD$. ~\citet{belkin18akernel} provide
more rigorous arguments for why under conventional statistical generalisation
theory, one would expect empirical risk minimisation on noisy data to cause
generalisation error to increase rapidly with increasing noise.  They show that
for a certain class of kernel classifiers, the RKHS norm of any classifier that
overfits noisy training data grows nearly exponentially with dataset
size\footnote{Note that for a constant noise rate the amount of noise increases
with increasing dataset size}. As most generalisation bounds for kernel
classifiers depend at most polynomially on the RKHS norm of the classifier, they
diverge to infinity as the dataset size increases. 

However,~\citet{belkin18akernel} observe empirically that the generalisation
performance of these classifiers does not deteriorate as quickly in practice as
the existing theory suggests.~They observe that as training loss tends towards
zero even in the presence of label noise, test error either remains stable or
decreases and then stabilises. Similar behaviour, albeit in the less rigorous
setting of over-parameterised deep neural networks, has been observed
in~\citet{Zhang2016} where deep neural networks were trained on CIFAR10 with
varying levels of random classification noise. The results reported in Figure
1(c) in~\citet{Zhang2016} show that when neural networks fit the noisy dataset,
even for large noise levels, test accuracy remains high on clean test-sets.

Some recent works look at characterizing learning problems where fitting noisy
labels does not cause the test error to blow up, a phenomenon referred to as
\emph{benign overfitting}.~\citet{Chatterji2020} show that if a linear
classifier fits noisy data then the test error will be close to the noise
rate~(i.e. have a decent test accuracy) only if the model is highly
over-parameterised. In Theorem 3.1, they show that, under some conditions, the
error rate is upper bounded by (label noise rate~($\eta$)) + (a term depending
on $e^{-\nicefrac{p}{C}}$), where $p$ is the dimensionality of the parameters
and $C$ is a constant. Thus for small $p$, the model that overfits the noisy
training set can have a very large test error whereas for large p, this error
will be close to the noise level.~\citet{Bartlett2020} look at this in the
setting of linear regression. In Theorem 4, they provide matching lower and
upper bounds for the test risk when the linear regression achieves zero
regression error. Their result shows that the dimensionality of the problem
determines whether the test risk will be good for a regressor that perfectly
fits a noisy training set. Thus, depending on the properties of the problem such
as its dimensionality, the perfectly fit model may get very bad or very good
test loss. 

These works provide support for the applicability of ERM in the presence of
noisy labels for large  deep neural networks and highly over-parameterised
models in general. However, even in the favourable scenario where ERM with noisy
labels yields a generalisable model, the model may suffer in
other notions of reliability that we care about.
In~\Cref{chap:causes_vul,chap:focal_loss}, we show how empirical risk
minimisation can provide good test error and yet suffer in adversarial
robustness and calibration.

\section{Regularisation}
\label{sec:regularisation}

To guarantee generalisation of models based on empirical error using the Basic
Theorem of Generalisation~(\Cref{thm:basic-gen-thm}), one solution is obviously
to just obtain a large amount of data~(i.e. large $N$). However, often this is
infeasible as data collection is expensive, time-consuming, and sometimes it is
impossible. 

On the other hand, if we have the right definitions of the complexity measure
$C$ in~\Cref{thm:basic-gen-thm}, we can prescribe the amount of data required to
guarantee generalisation. Similarly, if a limited amount of data is available,
then one can prescribe the maximum value \(r\in\reals_+\) of the complexity
measure $C$ for which the available data is sufficient to guarantee
generalisation. In that case, the search of hypothesis can be restricted to a
subset \(H^\prime\subseteq\cH\) that satisfies the prescribed constraint on
model complexity i.e. \(C\br{\cH^\prime}\le r\). It is this problem of
restricting the search of the hypothesis that regularisation tries to solve. 

Regularisation can be implemented through a regularisation function
$\Omega:\cH\rightarrow\reals_+$ that assigns a complexity value to every
hypothesis $h$ in $\cH$. One can think of $\Omega$ as a proxy of the complexity
function $C$ in~\Cref{thm:basic-gen-thm}. Note that schemes like early stopping
in gradient-based training, that do not directly use a regularisation function,
are also referred to as regularisation because they have been shown to control
some kind of complexity like a regularisation function~\citep{Yao07onearly}. In
the rest of this section, we will see different notions of optimisation with
regularisations i.e. different ways of imposing the constraints on the
identified structures in neural networks.

\subsection*{Ivanov regularisation}
The purpose of regularisation is to restrict the hypothesis space $\cH$ by
imposing further constraints.  For example, if $\cH$ were originally the class
of linear predictors in $\reals^2$, a valid form of regularisation would be to
consider only those linear predictors that are parallel to one of the coordinate
axes. This particular example is called $\ell_0$ regularisation. Effectively,
this reduces the complexity of $\cH$. In machine learning literature, such a
regularisation is referred to as Ivanov regularisation.~\Cref{eq:ivanov-erm}
defines an Ivanov-regularised ERM problem.

\begin{defn}[Ivanov regularisation]\label{eq:ivanov-erm} Let
  $\Omega:\cH\rightarrow\reals_+$ be the regularisation function and $r\ge 0$ be
  the maximum allowed value of the regularisation function. Then the regularised
  ERM problem is defined as obtaining a hypothesis $h_r^*$:
\begin{align*}
  h_r^* &= \min_{h\in\cH} \empRisk{h}{N}\\
  &\text{s.t.}~\Omega\br{h}\le r
\end{align*}
\end{defn}
In Neural networks, regularisations like spectral
normalization~\citep{miyato2018spectral} and Parseval networks~\citep{cisse17a}
can be thought of as an attempt to solve this problem\footnote{I have referred
to this as an attempt because they solve the problem of projection onto the set
of classifiers with correct spectral norm approximately}. Some prominent
instances are problems in low rank matrix approximation~\citep{PCA2002a}, matrix
recovery~\citep{candes2009exact},and sparse
recovery~\citep{blumensath2009iterative}. 
In~\Cref{chap:stable_rank_main}, we  devise a novel Ivanov regularisation called
Stable Rank Normalization~(SRN). Interestingly, we also obtain a closed-form
optimal solution to the projection problem which is not very common among
problems of this type which are solved only approximately.

In terms of generalisation theory, solving an Ivanov-regularised optimisation
problem can immediately provide meaningful guarantees on generalisation error.
If the capacity function $C$ is monotonically dependent on the regularisation
function $\Omega$, then constraining $\Omega$ provides an upper bound on the
capacity $C$ which can be used to compute the generalisation error. The problem
referred to above and in~\Cref{chap:stable_rank_main} is motivated by this idea.

\subsection*{Tikhonov regularisation}
The more commonly used regularisation in machine learning and particularly
gradient-based learning is  Tikhonov regularisation defined below in~\Cref{eq:tikhonov-erm}.

\begin{defn}[Tikhonov regularisation]\label{eq:tikhonov-erm} Let
  $\Omega:\cH\rightarrow\reals_+$ be the regularisation function and $\lambda\ge
  0$ be the regularisation coefficient. Then the Tikhonov regularised ERM
  problem is defined as obtaining a hypothesis $h_\lambda^*$ as follows
\begin{align*}
  h_\lambda^* &= \min_{h\in\cH} \empRisk{h}{N} + \lambda\Omega\br{h}
\end{align*}
\end{defn}

Regularisers like weight decay~\citep{KSH:2012}, $L_1$
regularisation~\citep{engelcke2017vote3deep,Sukhbaatar2019}, and spectral
regularisation~\citep{Yoshida2017} are examples of Tikhonov regularisation in
deep learning. While Ivanov and Tikhonov regularisations are equivalent for some
value of the coefficients, the main advantage of Tikhonov regularisation is that
it converts a constrained optimisation problem to an unconstrained optimisation
problem. Therefore, it is usually enough to apply gradient-based optimisation
methods. However, it must be noted that this is not ERM though it is sometimes
referred to as regularised Empirical Risk Minimisation. 

\subsection*{Morozov regularisation}
Another form of commonly studied regularisation is  residual learning
or also referred to as Morozov
regularisation~\citep{morozov2012methods}. The main objective here is
to minimise the regularisation function with constraints on the
original ERM loss. This is common in settings where the noise rate of
the problem is known and hence expecting a loss value lesser than the
noise rate would be unreasonable.

\begin{defn}[Morozov regularisation]\label{eq:morozov-erm} Let
  $\Omega:\cH\rightarrow\reals_+$ be the regularisation function and $\delta\ge
  0$ be the discrepancy parameter. Then the Morozov regularised ERM problem is
  defined as obtaining a hypothesis $h_\epsilon^*$ that satisfies
\begin{align*}
  h_\delta^* &= \min_{h\in\cH} \Omega\br{h}\\
  &s.t.~\empRisk{h}{N} \le \delta
\end{align*}
\end{defn}

\subsection{Data-dependent regularisation}

Common examples of regularisations usually comprise data-independent
regularisations. In the formulation
of~\Cref{eq:tikhonov-erm,eq:ivanov-erm,eq:morozov-erm}, the value of the
regulariser term $\Omega\br{h}$ does not depend on the data directly unlike the
empirical loss term $\empRisk{h}{N}$. Thus, the
penalisation~(\Cref{eq:tikhonov-erm,eq:morozov-erm}) or the
feasibility~(\Cref{eq:ivanov-erm}) of a particular hypothesis $h$  is the same
regardless of the properties of the dataset. In some sense, these regularisers
solve learning problems with little or no assumptions about the data
distribution.

However, there are cases when information about the data-generating distribution
needs to be accounted for. As a toy example, consider a high dimensional
instance space where the information in the data is captured in a small number
of principal input features and the remaining nuisance features contain no
information. The purpose of the regulariser, in this problem, is to find a low
complexity model that fits the observed dataset from this distribution. As it is
known that the unseen test data from this distribution will also not have any
weight in the nuisance features, two models that behave identically on the
principal components but differently on the nuisance components should not be
penalised differently by the regulariser. However, to do this the regulariser
needs to be aware of the data distribution to identify the principal
and nuisance components.

In practice, one common setting where data-dependent regularisers are applied is
semi-supervised learning problems where most data samples have no labels and
thus further information about the data distribution is required to relate
available information from the labelled data to predict labels for the
unlabelled data. Information regularisation~\citep{Corduneanu2012}  enforces
this by assuming that label predictions can be made by clustering points and
that each cluster is somewhat pure in its label. Their algorithm can be cast as
a modified data-dependent Tikhonov regularisation. Another prominent example of
data-dependent Tikhonov regularisation includes maximizing the entropy of the
predicted class-label distribution~\citep{Pereyra2017} to reduce the
confidence of the model on incorrect predictions.

In~\Cref{chap:low_rank_main}, we look at an optimisation problem with
data-dependent Ivanov regularisation. In particular, we look at constraining the
rank of the representation space of deep neural networks. The intuition is that
the model should  exploit as small a number of features as possible to do the
classification, as unnecessary features can be exploited by an adversary to
construct adversarial attacks. We solve the problem by converting it to an
equivalent data-dependent Tikhonov regularisation problem.
In~\Cref{chap:focal_loss}, we discuss Focal Loss which is a modified loss
function but that can be shown to be also acting as maximizing the entropy of
the predicted class probabilities, inherently doing a form of data-dependent
Tikhonov regularisation. Further, while the stable rank normalization presented
in~\Cref{chap:stable_rank_main} is an Ivanov regularisation approach, we also
discuss its impact on some data-dependent Lipschitz terms.

\chapter{Background for Reliable Deep Learning}
\label{sec:trml}
Algorithms to learn from data are now widely applied to make our lives easier.
Applications of these systems include medical diagnoses
\citep{K:2001,BSHLKPRK:2017}, fraud detection from personal finance data
\citep{GR:1994}, and online community detection from user data \citep{F:2010}.
They also governs our online presence by recommending what content we should see
on social networks, who we should connect with on professional networks, and
which words we should use to complete our emails. At the core, these are all
instances of algorithms that have been learnt from past data to maximise some
metric chosen by the service providers that deploy them. These metrics include
accuracy of classification~(eg. in medical diagnosis and fraud detection),
engagement~(eg. in social media and professional media recommendations) and/or
revenue.

However, as these algorithms are deployed at large scale it is becoming
increasingly clear that they suffer from some pathological issues. In
particular, deep learning, with its huge success both in terms of performance
and the fast pace at which it is being adopted by industry, poses a particularly
critical challenge as we are still unaware of the extent of its vulnerabilities.
In some cases, as the later chapters demonstrate, the pursuit of higher
performance under the original metrics further aggravates these vulnerabilities.
If such pursuit remains unchecked, it will lead to a decrease in trust in this
technology and, possibly, even have an adverse effect on the proper functioning
of society. \todo[color=green]{too dramatic}This chapter gives an overview of
four such issues that are particularly critical for deep neural networks. In
addition, it discusses past works in these fields and  how the rest of the
thesis fits into the context of these works.
\section{Generalisation in Neural Networks}
\label{sec:gen_nn}
A machine learning model that achieves a very small error on the observed training set but fails to perform when tested on unseen data is of little use. In fact, such a model might be dangerous if the user is unable to distinguish models that perform well on both the observed training set and unseen test data from models that perform well on training sets but fail on test data. The first kind of model, discussed extensively in~\Cref{sec:clt}, is said to have generalised whereas the second model is said to have failed at generalisation.

For neural networks, this is particularly worrisome. Consider the following
thought experiment with a data distribution over natural images equally
distributed in $k$ classes and two target concepts~(labelling functions): one of
which labels each image correctly and the other which labels each image
randomly, The two training datasets are constructed by first sampling $N$ points
independently from the distribution and then labelling the first dataset with
the {\em correct} target concept and the second dataset with the {\em random}
target concept. A sufficiently large and  properly trained neural network is
able to achieve zero training error via ERM for both of these training datasets
as seen from the experiments in~\citet{Zhang2016} and further
in~\Cref{chap:stable_rank_main}. Further, the first network would also
generalise to unseen data from that distribution labelled by the {\em correct}
target concept. On the other hand,  for the second concept, the target labelling
concept is a random labelling function. Hence, no model can achieve an expected
risk that is better than what can be achieved with random guesses. 
Thus, two neural networks with the same architecture achieve the same training
error on two different learning tasks but while one of the them generalises, the
other does not. This shows that given a neural network architecture and a
training dataset, it is not possible to certify, whether the neural network
trained on that dataset will generalise without any further information about
the data distribution. This tendency of neural networks to obtain zero training
error via ERM, irrespective of their test error, makes it ever more important to
have a rigorous generalisation theory for them. The  gold standard would be a
theory which, given a model architecture, learning algorithm, and a dataset, is
able to certify with a very high confidence whether the model generalises.

\subsection{Importance of structures for generalisation in neural networks}
\label{sec:struc-gn-nn}

Guaranteeing generalisation  for models obtained via ERM using results akin to~\Cref{eq:gen-erm} requires 
\begin{enumerate}
  \item an upper-bound on the true complexity of the model and 
  \item a dataset, whose size is proportionally large to match the upper-bound
  on model complexity. 
\end{enumerate}

If the proposed upper-bound on the model complexity is loose, the corresponding
number of samples required to theoretically guarantee generalisation
using~\Cref{thm:basic-gen-thm} will be large even if the true complexity of the
model is low and thus, in practice, generalisation will be observed with a
smaller dataset.

In the case of neural networks, it has been difficult to find accurate
definitions of the complexity measure $C$ in~\Cref{thm:basic-gen-thm}. In fact,
definitions of $C$ from most recent works prescribe a value of $N$ that is too
large  to justify the generalisation we see in real life with an amount of data
that is orders of magnitude less than the prescribed amount.~\citet{arora18b}
shows~(See Figure 3 in~\citet{arora18b}) that when recent measures of
complexity~\citep{bartlett2002rademacher,neyshabur2015norm,bartlett2017spectrally,neyshabur2018a,arora18b}
are applied to  VGG-19, the resultant prescribed sample complexities are many
orders of magnitude larger than the actual number of trainable parameters, let
alone the number of training examples.~\citet{Dziugaite2017} also show~(see
Appendix D in~\citet{Dziugaite2017}) that Rademacher complexity bounds computed
using the complexity bound of~\citet{neyshabur2015norm} are vacuous for deep
neural networks i.e. the bounds predict that the test error will be less than
one, which is trivially satisfied by the definition of test error.

While existing works in generalisation theory for neural networks seem to be
unable to prescribe the absolute amount of data~(even approximately) required
for generalisation in practice, it is natural to ask whether the {\em
structures}, identified by these works as being important for generalisation, do
in fact causally impact generalisation in practice\footnote{In this thesis, we
use the term {\em structure} to loosely refer to properties of both --- neural
networks and the data distribution eg. rank of the weight matrices, entropy of
the probability distribution over classes, rank of representations, and noise in
data.}. To address this, ~\citet{Jiang2020Fantastic} conduct a large-scale
empirical study to find out if there are causal relationships between recently
proposed complexity measures and generalisation in neural networks. They train
more than 2000 models on CIFAR10 in a controlled setup by systematically varying
important hyper-parameters, optimisation algorithms, and stopping criterion, and
investigate whether there is a strong correlation between generalisation and any
of the over 40 complexity measures that they study. Their results indicate that
a large number of them do indeed appear to be causally related to generalisation
even though the bounds themselves are vacuous i.e. the test error of different
learnt models maintain the same ordering as predicted by applying the
generalisation bounds on the the learnt models though the exact upper-bound on
the test error predicted by the generalisation bounds are greater than one,
which is true by the definition of test error. Despite the bounds being vacuous,
the causal relationship between the generalisation error and the complexity
bounds explored in their work suggests that regularising these complexity
measures might benefit generalisation in practice.

However, their study also suggests, somewhat counterintuitively and without a
causal explanation, that a large number of these complexity measures are
negatively correlated with generalisation. One of the limitations of their study
in identifying correlation between complexity measures and generalisation is
that they do not explicitly penalise the complexity measures in their
experiments. Without explicit penalties, the complexity measures of the learnt
models have higher magnitudes than what would be the case with said penalty. They compute the correlation between generalisation and complexity
measures by studying only the distribution on the complexity measures induced by
training without explicit penalties~(natural training). Thus, it is possible
that when the complexity measures lie in a range of larger magnitudes via
natural training, they do not show a strong correlation with generalisation but
lower magnitudes obtained through direct penalisation will show a strong
correlation. We show in ~\Cref{chap:stable_rank_main} that explicit penalisation
of the structures identified by these complexity measures indeed shows a greater
impact on the generalisation behaviour in practice. 

Thus our focus in this thesis is on studying the structures identified by these
complexity bounds and designing optimisation algorithms or regularisation
techniques to constrain them in practice in an efficient way. 
As the main purpose of this thesis is  not to propose new tighter generalisation
bounds, we will not delve deep into defining the complexity measure~$C$.
However, a general introduction to this topic is necessary to demonstrate the
context and significance of some of the future chapters,
especially~\cref{chap:low_rank_main,chap:stable_rank_main}. 

\subsection{Norm-based complexity measures}
\label{sec:common-gen}

Ultimately, the objective of designing complexity measures for neural networks
is to use them in some variant of~\Cref{thm:basic-gen-thm} and obtain
generalisation guarantees for these networks. One well-known instantiation of
the theorem involves Rademacher complexities~(\Cref{defn:radem-compl}).
Rademacher complexity~\citep{Koltchinskii2001} is a relatively modern notion of
complexity that is ~(data) distribution dependent and is defined for any class
of real-valued functions. The Rademacher complexity of a hypothesis class $\cH$
over sets, of size $N$, drawn i.i.d from a distribution $\cD$~($\cD$ is usually
omitted from the notation) is denoted by $\rad{\cH}{m}$.
\begin{defn}[Rademacher
  Complexity~\citep{Koltchinskii2001}]\label{defn:radem-compl} Let a sample $\cS
  = \bc{\vec{x}_1,\cdots,\vec{x}_m}$ be drawn i.i.d. from a data distribution
  $\cD_{\cX}$ over the instance space $\cX$ and consider a hypothesis class
  $\cH$, then the \emph{empirical rademacher complexity} of $\cH$ on the sample
  $\cS$ is defined as
  \begin{equation}
    \emprad{\cH}{m} = \bE\bs{\sup_{h\in\cH}\dfrac{1}{m}
  \sum_{i=1}^m\sigma_ih(\vec{x_i}) }
  \end{equation} where the expectation is over the
  rademacher variables $\sigma_i$, which are independent, uniform,
  $\bc{\pm 1}$ valued random variables
  
  The rademacher complexity is then defined as the expectation of the
  empirical rademacher complexity over the selection of the sample set
  using the underlying distribution.
  \begin{equation}
    \rad{\cH}{m} = \bE_{\cS\sim\cD_{\cX}^m}\bs{\emprad{\cH}{m}}
  \end{equation}
  \end{defn}

  Recall the second learning task in the thought experiment from
  ~\Cref{sec:gen_nn} where the target concept is a random labelling function.
  Clearly, this is a meaningless learning problem. The empirical rademacher
  complexity in~\Cref{defn:radem-compl} captures (in expectation over the
  labelling) how well the best hypothesis from the hypothesis class \(\cH\) can
  fit these random labels on the given dataset. In other words, it measures how
  well \(\cH\) can fit noise. Experiments in~\citet{Zhang2016}, showing that
  certain classes of deep neural networks like ResNet can memorise noise,
  suggests that these classes have a very large empirical rademacher complexity.

\begin{thmL}[Rademacher Complexity Generalisation
  Bound]\label{thm:gen-radem-error} Consider a hypothesis class $\cH$ of
  hypothesis $h:\reals^d\rightarrow\bs{0,1}$ and, $\riskOne{h}$ and
  $\empRisk{h}{N}$ are the expected and empirical risks as defined
  in~\Cref{defn:exp_loss,defn:emp_loss}. Then $\forall\delta\ge 0$, with
  probability $1-\delta$ ,the following holds\footnote{Proof can be found in~\url{http://www.cs.ox.ac.uk/people/varun.kanade/teaching/CLT-HT2018/lectures/lecture08.pdf}. \todo[color=green]{Find the proof}} $\forall h\in\cH$
  \[\riskOne{h}\le \empRisk{h}{N}+ 2\rad{\cH}{N} + \sqrt{\dfrac{\log{\br{\frac{1}{\delta}}}}{2N}}\]

\end{thmL}

To use~\Cref{thm:gen-radem-error} for neural networks, we need to compute the
  rademacher complexities of classes of neural networks. A series of
  papers~\citep{bartlett2002rademacher,neyshabur2015norm,golowich18a} has
  designed Rademacher complexity bounds for neural networks. These bounds
  primarily depend on the product of the Frobenius norm of the weight parameters
  of each layer $\norm{\vec{W}_i}_\mathrm{F}$ and the depth of the network $L$.
  As~\Cref{thm:simpl-radm-nn} shows, some of these
  bounds~(c.f.~\Cref{ineq:rad-exp-depth-frob-simple}) have an exponential
  dependence on network depth $L$ while others~\citep{golowich18a} avoid the
  exponential dependence but still have an indirect dependence on the network
  depth through the product of the norms. Note that the only term
  in~\Cref{thm:simpl-radm-nn} that contains learnable parameters and can be
  controlled through regularisation are the Frobenius norms
  $\norm{\vec{W}_i}_\mathrm{F}$ of each weight matrix. Thus, explicit
  regularisation of the norms  of each weight parameter in a neural network
  through regularisers like weight decay might help with generalisation.
  Experiments in~\citet{Zhang2016}~(c.f. Table 1,2 in~\citet{Zhang2016}) show
  that while this has a positive effect on generalisation, the effect is
  small. Thus, there is a need to identify further structures which have
  stronger impact on generalisation.
  \begin{thmL}[Simplified Rademacher Complexity of MLP~\citep{bartlett2002rademacher,neyshabur2015norm,golowich18a}]
    \label{thm:simpl-radm-nn} Consider a class of real-valued
    Multi-Layer Perceptrons~(~\Cref{defn:net-mlp}) $\cN$ with ReLU activation
    function and $L$ layers.
    The Rademacher complexity of $\cN$ scales as
    follows~\citep{neyshabur2015norm}\footnote{~\Cref{thm:simpl-radm-nn}
    uses Empirical Rademacher Complexity while the generalisation bound
    in~\Cref{thm:gen-radem-error} is with Rademacher complexities. However,
    one can be transferred to the other.
    See~\url{https://www.cs.ox.ac.uk/people/james.worrell/rademacher.pdf}
    for further information.}

    \begin{equation}  
      \emprad{\cN}{N} \le \frac{1}{\sqrt{N}}2^{L+1}
    \br{\prod_{j=1}^L\norm{\vec{W}_j}_F} \br{\max_{x\in\cX}\norm{x}}
    \label{ineq:rad-exp-depth-frob-simple} 
    \end{equation}
  where $\vec{W}_j$ is the weight parameter for the $j^{\it th}$ layer of the neural network and $\cX$ is the instance space.~\citet{golowich18a} further improved bounds of this form to
    \begin{equation} \emprad{\cN}{N} \le \frac{1}{\sqrt{N}}
    \br{\sqrt{2\log{\br{2}L}}+1} \br{\prod_{j=1}^L\norm{W_j}_F} \br{\max_{x\in\cX}\norm{x}}
    \label{ineq:rad-depth-indep-simple} \end{equation}
  \end{thmL}

\subsection{Rank and margin-based complexity measures}
\label{sec:marg-based-gener-main}

Despite its~(marginal) impact on generalisation in practice through explicit
penalisation, these bounds fail to explain some simple behaviours of neural
networks. Multi-layer perceptrons~(MLPs) and even convolutional neural
networks~(without special structures like skip connections) show the interesting
property of positive homogeneity. A function $f$ is $k$-positive homogenous if
there exists an integer $k$ such that for any positive constant $c$ and an
element $x$ in the domain of $f$, we have $f\br{cx}=c^k\br{x}$. It is easy to
verify that a $k$-layer MLP is $k$-positive homogenous in the space of
parameters. An important consequence of this fact is that, for a $k$-layer MLP,
multiplying all the weights of the network with a positive constant does not
alter the test error of the model even though it increases the norm of the
weights. Thus, using a suitably large positive number and the positive
homogeneity of MLPs, the complexity measures in~\Cref{thm:simpl-radm-nn} can be
increased to any arbitrarily large number without altering the generalisation
error of the model, which is a direct contradiction to what a generalisation
bound is supposed to achieve. This shows a simple failure case of norm-based
complexity measures like in~\Cref{thm:simpl-radm-nn}. Margin-based measures,
discussed below, can be used to overcome this particular failure case.

Before progressing further we will first define the concept of margin.  Consider
a hypothesis class $\cH$ and the learning task to be multi-class classification.
Every hypotheses $h\in\cH$ takes a vector $\vec{x}$ from $\cX\subset\reals^d$
and generates a probability distribution over the classes identified by the $k$
dimensions of $\cY\subseteq\Delta^k$, the $k$-dimensional simplex. The standard
strategy for prediction using this model is to output the index of
$h\br{\vec{x}}$ which has the largest magnitude i.e.
$\argmax_i{h\br{\vec{x}}\bs{i}}$. In~\cref{defn:margin,defn:margin-loss}, we
define the concept of margin and margin loss respectively.

\begin{restatable}[Margin]{defn}{margin}\label{defn:margin} The margin of a
  hypothesis $h$ at a data sample $\br{\vec{x},y}$ is defined as
  \[\gamma\br{h;\vec{x},y} = h\br{\vec{x}}\bs{y} - \max_{j\neq
  y}h\br{\vec{x}}\bs{j}\] where \(h\br{\vec{x}}\bs{j}\) indexes the \(j^{\it
  th}\) element of \(h\br{\vec{x}}\).
\end{restatable}
\begin{restatable}[Margin Loss]{defn}{marginloss}\label{defn:margin-loss}
  The margin loss is defined as
  \begin{equation}
    \ell_{\gamma^*}\br{h; \vec{x}, y} =
    \begin{cases}
      0 & \gamma\br{h;\vec{x},y}  \ge \gamma^*\\
      1 & \gamma\br{h;\vec{x},y} < \gamma^*
    \end{cases}
  \end{equation} where $\gamma$ is the margin defined in~\Cref{defn:margin}.
  We will use $\empRisk{h}{\gamma,N}$ and $ \cR_{\gamma}\br{h}$ to represent the
  empirical and the expected margin risk, respectively, of the classifier $h$.
  The normal classification risk is simply $\cR_{0}\br{h}$.
\end{restatable}

\paragraph{Spectrally normalized margin bounds}

 Consider an MLP $g\in\cN$ as defined in~\Cref{defn:net-mlp}. 
Let the network $g$ be parameterised by a sequence of weight matrices
$\vec{W}_1,\cdots,\vec{W}_L$ with $l_k$ neurons in the $k^{\it{th}}$ layer. The
width of the network $W$ is the maximum of $\{l_1,\ldots,l_L\}$.

\begin{defn}[Spectral Complexity
from~\citet{bartlett2017spectrally,neyshabur2018a}] \label{eq:spec_comp_nn} The
\emph{spectral complexity} $ R_g$ of such a network $g\in\cN$ %
is defined as\footnote{For simplicity, we assume 1-Lipschitz
Activation functions}
\begin{equation}\label{eq:defn-ins-spec-compl}
  R_{g} =   \br{\prod_{i=1}^L \norm{\vec{W_i}}_2}
  \br{\sum_{i=1}^L \frac{ \norm{\vec{W}_i^{\top}}_{2,1}^{2/3}}{\norm{\vec{W}_i}_2^{2/3}}}^{3/2}
\end{equation} where $\norm{\cdot}_{p,q}$ is the entry-wise $p,q$ norm and is  defined in~\Cref{eq:entry-wise-pq-norm} along with other basic linear algebra concepts.
\end{defn}
The following theorem provides a generalisation bound for neural
networks $g$ whose 
weight matrices $\bc{\vec{W}_1,\cdots,\vec{W}_L}$ have bounded
spectral complexity $R_g$\footnote{The authors~\citep{bartlett2017spectrally} also defines a collection of \emph{reference matrices} $(\vec{M}_1,\ldots,\vec{M}_L)$ with the same dimensions as
$\vec{W}_1,\ldots,\vec{W}_L$. This can be adjusted for various network
structures to get the best bound. Then, the spectral complexity is defined as \begin{equation}
  R_{g} =   \br{\prod_{i=1}^L \norm{\vec{W_i}}}
  \br{\sum_{i=1}^L \frac{ \norm{\vec{W}_i^{\top} -\vec{M}_i^\top}_{2,1}^{2/3}}{\norm{\vec{W}_i}_2^{2/3}}}^{3/2}.
\end{equation}}.

\begin{thmL}[Spectrally Normalized Margin Bounds from~\citet{bartlett2017spectrally}]
  \label{thm:spec-norm-marg-main}
  For any MLP $g\in\cN$ with width $W$ and whose weight matrices
  $\bc{\vec{W}_1,\cdots,\vec{W}_L}$ have bounded spectral complexity $R_g$, any
  $\bc{(x_1,y_1),\ldots,(x_N,y_N)}$ drawn i.i.d. from a distribution over
  $\reals^d\times\{1,\ldots,k\}$, and for any $\gamma > 0$ , the following holds
  with probability at least $1-\delta$:
  \begin{equation}
    \riskOne{g}
    \le
    \empRisk{g}{\gamma,N} + \tildeO{ \frac
    {\sqrt{\sum_{i=1}^N\norm{\vec{x}_i}^2} R_g}{\gamma N}    \log{W} +
    \sqrt{\frac{\log{\frac{1}{\delta}}}{N}} } 
  \end{equation}
\end{thmL}

Though the bound in Theorem~\ref{thm:spec-norm-marg-main} does not have a direct
exponential dependence on the depth of the network $L$,~\citet{neyshabur2018a}
points out that for any matrix $\vec{W}_i$,
\[\dfrac{\norm{\vec{W}_i^\top}_{2,1}}{\norm{\vec{W}_i}_2}\ge 1.\] Thus if
$\max_{i\le N}\norm{\vec{x}_i} = R_\cX$ and $R_g$ is the spectral
complexity as defined above, then the excess error term in~\Cref{thm:spec-norm-marg-main} can be written as
\begin{equation}\label{eq:bartlett-excess-error}
  \frac{\sqrt{\sum_{i=1}^N\norm{\vec{x}_i}^2} R_g}{N} =
\tildeO{\br{R_\cX\prod_{i=1}^L\norm{\vec{W}_i}_2}\sqrt{\dfrac{L^3}{N}}}.
\end{equation}
The expression in~\Cref{eq:bartlett-excess-error} still depends exponentially on
$L$ if the spectral norm of the individual matrices $\vec{W}_i$ is not bounded
below one. Even if the norms are bounded, the bound becomes trivial when $L\ge
N^{\frac{1}{3}}$. In contrast,~\citet{golowich18a} shows a bound, which under
suitable bounds on various norms, is independent of the size of the network.

~\citet{neyshabur2018a} develop a slightly different expression for spectral
complexity through a Pac-bayesian analysis. The original paper gives a
qualitative comparison between the two bounds. Their form of spectral complexity
can be written as follows where $\norm{\cdot}_\forb$ is the Frobenius norm and
$\norm{\cdot}_2$ is the $2$-operator norm~(see~\Cref{defn:induced_norm}).

\begin{equation}\label{eq:stable-spec-compl}
  R_g =   \br{\prod_{i=1}^L \norm{\vec{W_i}}_2^2
  \sum_{i=1}^L \frac{ \norm{\vec{W}_i^{\top}}_{\forb}^2}{\norm{\vec{W}_i}_2^{2}}}^{\frac{1}{2}}
\end{equation}

The term $\frac{\norm{\vec{W}_i^{\top}}_\forb^{2}}{\norm{\vec{W}_i}_2^{2}}$
in~\Cref{eq:stable-spec-compl} is also known as the stable rank of matrices.
In~\Cref{chap:stable_rank_main}, we develop an algorithm to directly control the
spectral complexity~(see~\Cref{eq:stable-spec-compl}) of neural
networks during training. Our experiments on a wide range of settings show that this indeed helps generalisation in practice.

\subsection{Complexity measures based on compression}
\label{sec:compression-compl}

Next, we discuss a slightly different idea for formulating generalisation bounds
of neural networks. The main idea in~\citet{arora18b} is that if the performance
of a \emph{large} network on the training set can be copied very closely by a
\emph{small} network, then one can use the low empirical risk of the
\emph{large} network to guarantee that the smaller network will also have a low
empirical risk. Then, standard complexity based generalisation bounds can
leverage the low complexity of the \emph{small} network to guarantee a better
generalisation guarantee for the small network. As the small network behaves
similar to the large network, this generalisation guarantee can then be
transferred to the large network. To be able to use results of this kind in
practice, we need to a) obtain a large network that gets small training error
and b) show that the large network can be closely emulated  by a smaller
network. We will refer to the copying of the large network by a small network as
{\em compressing} the network. The first is easy to obtain in practice as neural
networks often train to very small training error. For the
second,~\citet{arora18b} defines a set of sufficient properties for the neural
network to be compressible. We list these properties below and then evaluate
them in later chapters to show that  neural networks obtained via our
regularisations enjoy some of these desirable properties.

~\Cref{defn:gamma-s-compressible} defines $\br{\gamma,S}$-compressible
functions. A  function $h$ is $\br{\gamma,S}$-compressible with respect to a set
of functions $\cG$ if one of the functions in $G$ can simulate $h$ within an
error tolerance of $\gamma$. Then, assuming that $\cG$ is a set of
low-complexity functions and $h$ has a low empirical error on a fixed
dataset,~\Cref{thm:comp-bound-main} guarantees that the function from $\cG$ that
closely simulates $h$ will have a low expected risk. 

\begin{restatable}[($\gamma$,$S$)-compressible using helper string $s$~(see~\citet{arora18b})]{defn}{gammacompressible}
\label{defn:gamma-s-compressible} Let $\cA$ be the set of all possible
parameters and $G_{\mathcal{A},s} =\{g_{A,s}|A\in \mathcal{A}\}$ be a class of
classifiers indexed by trainable parameters $A$ and a fixed string $s$. Then,
for $\gamma>0$, a classifier $h$ is ($\gamma,S$)-compressible with respect to
$G_{\mathcal{A},s}$ using helper string $s$ if there exists $A\in \mathcal{A}$
such that for any $x\in S$ and all $y$.
\[\abs{h(\vec{x})[y] - g_{A,s}(\vec{x})[y]} \le \gamma.\]
\end{restatable}

\begin{restatable}{thmL}{generalisationcompressed}\label{thm:comp-bound-main}
  Suppose $G_{\mathcal{A},s} =\{g_{A,s}|A\in \mathcal{A}\}$ where $A$ is a set
  of $q$ parameters each of which can have at most $r$ discrete values and $s$
  is a helper string. Let $S$ be a training set with $N$ samples. If the trained
  classifier $f$ is $(\gamma_c,S)$-compressible via $G_{\mathcal{A},s}$ with
  helper string $s$, then there exists $A\in\mathcal{A}$ such that $\forall
  \gamma\ge2\gamma_c$ with probability at-least $1-\delta$ over the training
  set,
  \[L_0(g_A) - \hat{L}_\gamma(f)\le \sqrt{\frac{1}{2N}\br{q\log{r} +
  \log{\frac{1}{\delta}}}} = \tildeO{\sqrt{\frac{q\log r}{m}}}.\] 
\end{restatable}

The problem with Theorem~\ref{thm:comp-bound-main} is that it does not
provide a generalisation guarantee for the original classifier but only for the
compressed version. Below, we will identify some properties of neural networks
which will allow us to circumvent this problem. These properties have been
presented in~\citet{arora18b}.

\paragraph{Compression based bounds and data dependent properties}
\label{data-dependent-properties}

Let $\cS$ be a set of $N$ examples drawn i.i.d. from any probability
distribution on $\reals^d\times \bc{1\cdots k}$  and let $g$ be an MLP as
described above with $L$ layers.  In particular, let $\vec{W}_i$ be the weight
matrix of the linear transformation of the $i^{\it th}$ layer, let $\vec{z}_i$
be the pre-activation representation of the $i^{\it th}$ layer, and let $\phi$
be the activation function. For any two layers $i\le j$, denote by $M^{i,j}$ the
operator for composition of these layers and by $J^{i,j}_{\vec{x}}$ the Jacobian
of this operator at input $\vec{x}$. Then, the following data-dependent
properties of the network are defined for a given network. We empirically
evaluate these properties later in the thesis to understand the impact of our
regularisations on the behaviour of the network and how they affects properties
like noise-sensitivity and generalisation.
      
\begin{defn}[Layer Cushion]\label{defn:lyr-cushion}
  For any layer $i$,  the layer cushion is defined as the largest number $\mu_i$
  such that for any $x\in S$:
		\[\mu_{i}\norm{\vec{W}_i}_F \norm{\phi(\vec{z}_{i-1})} \leq
		\norm{\vec{W}_i\phi(\vec{z}_{i-1})}. \]
\end{defn}

\begin{defn}[Inter-Layer Cushion]
  For any two layers $i\leq j$, the inter-layer cushion  is defined as the
  largest number $\mu_{i,j}$ such that for any $x\in S$:
  \[ \mu_{i,j}\norm{\vec{J}^{i,j}_{\vec{z}_i}}_F \norm{\vec{z}_i} \leq
  \norm{\vec{J}^{i,j}_{\vec{z}_i}\vec{z}_i}. \] Furthermore,  define the minimal
  inter-layer cushion \(\icu\) as \(\min_{i\leq j\leq d} \mu_{i,j} =
  \min\{1/\sqrt{l_i},\min_{i< j\leq d} \mu_{i,j}\}\).
\end{defn}

\begin{defn}[Activation Contraction]
  The activation contraction is defined as the smallest number \(c\) such that
  for any layer $i$ and any $x\in S$,
  \[ \norm{\vec{z}_i} \leq c \norm{\phi(\vec{z}_i)}.\]
\end{defn}

\begin{thmL}[Bounds Based on
  Compression,~\citep{arora18b}]\label{thm:comp-class-gen} 
  
  For any MLP $g$ as defined above\footnote{The result also requires some
  additional assumption on a property known as Inter-Layer Smoothness~(c.f.
  Definition 7 in~\citet{arora18b})}, any probability $0<\delta\le 1$, and any
  margin $\gamma$, Algorithm 1 in~\citet{arora18b} generates $\widehat{g}$ such
  that with probability at least $1 - \delta$ over the training set and the
  randomness in creating $\widehat{g}$, the following holds
\[ \riskOne{\hat{g}} \le \empRisk{g}{\gamma,N} +
\tildeO{\sqrt{\dfrac{c^2L^2\cR_{\cX}^2
\sum_{i=1}^L \frac{1}{\mu_i^2\icu^2} }{\gamma^2N}}}, \] where
$c,\mu_i,\rho_d,$ and $\mu_{i\rightarrow}$ are defined above, and
$L$ is the number of layers. This $\widehat{g}$ can be thought of as the compressed version of $g$.
\end{thmL}

Through these properties, this bound captures more data-dependent
characteristics than the other generalisation bounds discussed above. The sample
complexity obtained using~\Cref{thm:comp-class-gen} for standard neural networks
is also more realistic than any of the other results discussed in the earlier
paragraphs. This suggests that we need to find structures in the data  and
training algorithm that play a role in determining the generalisability of
neural networks trained on real-world data as opposed to just considering
the architecture of the neural network. One way of doing this is to identify
what properties of the learned network reflect said properties of the data and
the algorithm. In~\Cref{chap:stable_rank_main}, we look at {\em Empirical
Lipschitzness} to measure the sensitivity of the network on the dataset.
In~\Cref{chap:low_rank_main}, we look at layer
cushion~(see~\Cref{defn:lyr-cushion}) to measure the noise-sensitivity of neural
networks, and in~\Cref{chap:focal_loss}, we look at the entropy of the predicted
distribution over target class labels to understand causes of mis-calibration of
neural networks. We find that these data-dependent properties of neural networks
are a better indicator of the network's behaviour than data-agnostic properties.

\section{Adversarial Robustness of Neural Networks}
\label{sec:robustness}
In a benign real world setting, where the probability distributions from which
the training and the testing data are sampled are identical, generalisation, as
discussed in~\Cref{sec:slt}, provides guarantees on the reliability of deployed
machine learning systems. However, machine learning algorithms are also used in
settings where the assumption that the data distribution remains stationary
between training and deployment does not hold. In such settings, a direct
use of generalisation theories, as discussed in the last section, does not
provide meaningful guarantees of the reliability of the machine learning system.

This can be particularly worrisome when a malicious adversary has the power to
change the data distribution for their own benefit. Adversarial
attacks~\citep{Biggio2018,szegedy2013intriguing} --- where an adversary
imperceptibly changes the data  to force the model into making a mistake, is one
example of such a setting. Recently, this has been shown to be extremely
relevant for neural networks~\citep{Dalvi2004,Biggio2018,szegedy2013intriguing,
goodfellow2014explaining,Carlini2017,Papernot2016,mosaavi2016}. This section
will discuss the threat model, different types of adversaries, ways to measure
vulnerability against these adversaries, and existing work in the literature
regarding how to protect from them.

\subsection{Characterising an adversary}
Before formally defining adversarial error, we need to first
characterise what it means for the adversary to a) {\em imperceptibly
change the data} and b) {\em be successful in forcing the model in
making a mistake}. 

\paragraph{Threat model} The first part is captured by defining a {\em threat
model} - which puts a constraint on what the adversary is allowed to do while
perturbing the data. There are different types of constraints the adversary can
enforce:
\begin{enumerate}
\item  Computational Constraint -- A computational constraint restricts the
number of operations the adversary is allowed to  execute while perturbing the
data point. For example, if an adversary is allowed to do only $k$ steps of
gradient ascent while constructing the perturbed data point, that is an example
of a computational constraint on the adversary.
\item Information Theoretic Constraint -- An information theoretic constraint
restricts the adversary to ensure that the information contained in the
perturbed data is not too different from that in the original data. For example,
an $\ell_p$ bounded adversary ensures that the $\ell_p$ norm of the induced
perturbation is small.
\item Knowledge Constraint -- The adversary can also be constrained by the
information it has access to while constructing the attack. Some adversaries
might have full knowledge of the model they are attacking whereas others may not
and have to construct their proxy of the model while crafting the attack.
Examples of this distinction can be seen between white box and black box attacks
respectively.
\end{enumerate}

Formally, the threat model of the adversary can be defined as a function
$\cA:\cX\rightarrow\cP^{\cX}$ where $\cP^{\cX}$ denotes the power set of $\cX$.
$\cA$ is a function that maps a point in $\cX$ to a set of points in $\cX$ that
can be feasibly obtained by the constrained adversary. For an $\ell_p$ bounded
adversary with radius $r$, $\cA\br{\vec{x}}$ is the set of all points within an
$\ell_p$ ball of radius $r$ around $\vec{x}$. \[\cA\br{\vec{x}} = \bc{\vec{z}:
\norm{\vec{z}-\vec{x}}_p\le r}\]
\begin{remark}
    For real application purposes, ``imperceptibly changing the data'' usually means to be imperceptible to a human. The definition of the threat model above (a function from a single example to a  set of examples) is powerful enough to represent this. However, it is difficult to provide a precise mathematical definition of such a threat model. Thus, research in adversarial robustness has stuck to easy-to-define and computationally tractable threat models. However, neither does this mean that more powerful threat models do not exist nor does it mean that this simple threat model is of no practical utility.
\end{remark}

The second component of characterizing an adversary is defining  what it means
to be successful in forcing the model to make a mistake. Generally, this
corresponds to forcing the value of the loss function of the attacked model to
increase on the perturbed data point as compared to the original data point.
Usually, adversarial robustness is discussed in the context of supervised
classification problems. Forcing the model to make a mistake in this context
corresponds to forcing the model to make a classification error on the
adversarially perturbed data point despite being correct on the original data
point. 

Combining these two components, the job of an adversary is to find a perturbed
data point within its threat model that maximally increases the loss value of
the perturbed data point. Given an adversary $\cA:\cX\rightarrow\cP^{\cX}$, a
learned classification model $h:\cX\rightarrow\cY$, a loss function
$\ell:\cY\times\cY\rightarrow\reals$, and an {\em original data point}
$\br{\vec{x}_d,y}$,  the {\em adversarially perturbed data point} $\vec{x}_a$ is
the solution of the following optimisation problem.

\begin{equation}\label{eq:gen_adv_att}
    \vec{x}_a = \max_{\vec{x}\in\cA\br{\vec{x}_d}}\ell\br{h\br{\vec{x}},y}
\end{equation}

Even for relatively simple threat models like $\ell_p$ bounded adversaries, the
maximisation problem can be difficult to solve exactly. Especially in neural
networks, the classifier $h$ is a non-convex function and thus the inner
maximisation problem is non-convex. Thus, attacks are usually created
by considering various approximations to the problem. The most common approach
to solve this is projected gradient ascent.

\begin{restatable}[$\ell_p$ Adversarial
    Error]{defn}{advrisk}\label{defn:adv_risk} For any distribution $\cD$
    defined over $\br{\vec{x},y}\in\reals^d\times\cY$, any classifier
    $h:\reals^d\rightarrow\cY$, and any $\gamma>0$, 
    the $\gamma$-\emph{adversarial} error is 
            \begin{equation} 
                \radv{\gamma}{h;\cD}=\bP_{\br{\vec{x},y} 
                \sim\cD}\bs{\exists \vec{z}\in\cB_{\gamma}
                \br{\vec{x}};h\br{\vec{z}}\neq y},
            \end{equation} 
    where $\cB_\gamma^p\br{\vec{x}}$ is the $\ell_p$ ball of radius $\gamma \ge
    0$ around $\vec{x}$ under the $\ell_p$ norm.
  \end{restatable}

In the adversarial robustness literature, the adversarial error is usually defined
by~\Cref{defn:adv_risk}. This particular definition measures the risk when the
adversary is only constrained by information theoretic constraints, more
specifically an $\ell_p$ norm ball. Most work has looked at $\ell_\infty$~\citep{goodfellow2014explaining,madry2018towards}
constraints though some have considered $\ell_2$~\citep{Carlini2017}, $\ell_1$~\citep{chen2018ead}, and
$\ell_0$~\citep{Carlini2017}\todo[color=green]{Cite three papers for the diff kinds} balls as well. However, when
reporting empirical results, computational constraints are also necessarily
included for practical reasons. Thus most practical evaluations of adversarial
robustness use a combination of computational, information theoretic, and
knowledge constraints.~\citet{Gluch2020} discuss a generalisation of the
knowledge constraint where the adversary is characterised by the number of
queries it is allowed to make to the attacked model while constructing the
attack. If this number is unbounded, then the adversary is called a white box
adversary and if the number is zero, the adversary is called a black box
adversary. Any finite non-zero number characterises the strength of the
adversary. 

We will now discuss some commonly used adversaries, which we will also use later in~\Cref{chap:causes_vul,chap:low_rank_main} and then discuss some
commonly used adversarial defence methods.

\subsection{Adversarial attacks}
\label{sec:adv-attack-bg}
One of the first instances of adversarial attacks with Neural Networks is in the
work of~\citet{goodfellow2014explaining} where the adversary is constrained both
computationally and information theoretically, in terms of $\ell_p$ bounds. When
the perturbed data point is constrained to lie within an $\ell_p$ norm of a
certain radius, we will refer to the radius as the perturbation budget of the
adversary. The adversary in~\citet{goodfellow2014explaining} is allowed to do
one step of gradient ascent while ensuring that the perturbed point is within an
$\ell_\infty$ perturbation budget of the original data point. Here, the
magnitude of the perturbation budget defines the strength of the adversary.
Known as Fast Sign Gradient Method~(FGSM), this adversary is characterised by
the perturbation budget $\epsilon$ and attacks a point $\vec{x}_d$ by generating
the adversarially perturbed $\vec{x}_a$ as\todo[color=green]{FGSM vs FGSM}

\begin{equation}\label{eq:fgsm_one_step}
    \vec{x}_a = \vec{x}_d + \epsilon\cdot\sgn{\nabla_x\ell\br{h\br{\vec{x}_d},
    y}}.
\end{equation}
This is in fact one step of projected gradient descent on the cross-entropy loss
function $\ell$ where the projection set is  an $\ell_\infty$ ball of radius
$\epsilon$ around $\vec{x}_d$. The other important characteristic of this attack
is that the sole aim of the adversary is to force $h$  to misclassify
$\vec{x}_a$ without any specific target for what $\vec{x}_a$ should be
misclassified to. 

\paragraph{Multi-step adversaries}

\Cref{eq:fgsm_one_step} and the FGSM adversary describe an attack threat model
where the adversary is constrained to only one step of projected gradient
ascent; a stronger version of this allows the adversary is allowed to take
\(T\ge 1\) steps of length $\alpha$~\citep{kurakin2016} while satisfying the
constraint of being within the \(\ell_\infty\) perturbation
budget.~\Cref{alg:fgsm_pgd} describes this process. In this case, the definition
of the threat model includes a computational constraint hyper-parameter $T$ to
control the number of allowed gradient ascent steps, a hyper-parameter $\alpha$
for the length of each gradient step, and an information theoretic parameter
$\epsilon$ for the perturbation budget. We will refer to this as the~\ifgsm
adversary. It is one of the most commonly used threat models as it allows
varying the computational and the information theoretic constraints
independently. It has been used in a series of works
including~\citet{madry2018towards} and~\citet{Zhang2019}.
\begin{algorithm}
    \caption{Multi-Step Fast Gradient Sign Method}
    \label{alg:fgsm_pgd}
    \begin{algorithmic}[1]
            \INPUT Original Data~($\vec{x}_d,y$), classification
            model~($h$), loss function~($\ell$), Threat model~($T, \epsilon,\alpha$) 
            \STATE
            $\vec{x}_a^0 \gets \vec{x}_d$ 
            \FOR{$t\in\bc{0,\cdots
            T-1}$} 
            \STATE $\vec{x}_a^{t+1} \gets
            \clipBig{\vec{x}_d}{\epsilon}{\vec{x}_a^t+\alpha\nabla_x\ell\br{h\br{\vec{x}_a^t},
            y}}$ \label{step:update_adv}
            \ENDFOR 
            \OUTPUT $\vec{x}_a^T$
        \end{algorithmic}
\end{algorithm}

The $\mathrm{Clip}$ operation in~\Cref{alg:fgsm_pgd} is specific to the
$\ell_\infty$ bounded adversary and image data where each pixel ranges between
$0$ and $255$ and is defined as 
\[\clipBig{x}{\epsilon}{z} = \mathrm{min}\br{255, x + \epsilon,
\mathrm{max}\br{0, x - \epsilon, z}}\] However, these attacks can also be
extended to norms other than the $\ell_\infty$ norm by changing the projection
operator to a different operator depending on the norm. For example,~\citet{szegedy2013intriguing} used the $\ell_2$ norm. 

\paragraph{Minimum-Norm unbounded attacks} 

\Cref{eq:min_norm_unb_att} describes a generic form of such attacks where
$\norm{\cdot}$ is a particular norm and $h\br{\vec{x}}\neq
h\br{\vec{x}+\vec{r}}$ can be cast into our generic adversarial example
formulation of~\Cref{eq:gen_adv_att} by considering the loss function $\ell$ to
be the classification error function $\ell_{01}\br{y,\hat{y}}=\bI\bc{y\neq
\hat{y}}$. Another form of attack where there is no specific computational or
information-theoretic constraints are minimum norm unbounded attacks.
\begin{align}\label{eq:min_norm_unb_att}
    \min_{\vec{r}\in\reals^d}&\norm{\vec{r}}\\
    \text{s.t.}&~h\br{\vec{x}}\neq h\br{\vec{x}+\vec{r}}.\nonumber
\end{align}

A specific instantiation of this with the $\ell_2$ norm is the DeepFool
adversary by~\citet{mosaavi2016}. However, measuring the success of these
unbounded attacks is  different from measuring the success of bounded attacks
as, by design, these unbounded attacks are more likely to succeed.
Thus~\citet{mosaavi2016} measure the ratio $\dfrac{\norm{r}}{\norm{\vec{x}}}$
for successful misclassifications, which indicates how far the attack had to
move the original data point for the classification algorithm to fail.

\paragraph{Targeted attacks} The attacks we saw so far are known as {\em
untargeted attacks}; the sole aim of the adversary is to make the target model
misclassify the data point without any constraint on what it should be
misclassified as. In a $k$-class multiclass classification problem, this means
that the adversary is satisfied if the model classifies the perturbed data point
into one of the $k-1$ incorrect classes. A {\em Targeted adversarial attack}
changes the data point imperceptibly with the aim that the classifier model
classifies the perturbed data point into a label of the adversary's choosing.

In addition to the arguments of~\Cref{alg:fgsm_pgd}, a targeted
adversarial attack adversary would also usually take a target label
$y_t$. The only change in the algorithm would be the update step,
where instead of increasing the loss value for the correct label, the
adversary would now decrease the loss value for the target label as
follows:
\[\vec{x}_a^{t+1} \gets
\clipBig{\vec{x}_d}{\epsilon}{\vec{x}_a^t-\alpha\nabla_x\ell\br{h\br{\vec{x}_a^t},
y_t}}.\]

\citet{kurakin2016} observed that for datasets with a large number of classes
and varying degrees of significance in the difference between classes,
untargeted adversarial attacks can result in {\em uninteresting}
misclassifications, such as mistaking one breed of sled dog for another breed of
sled dog. Thus, in order to create visually striking adversarial
mis-classifications, they developed a method which they refer to as {\em
Iteratively Least Likely Class Method}~(\ill). Their adversary forces the
classifier to classify the perturbed data point to a label that would have been
the least likely class for the original data point. For a given data point
$\vec{x}$ and a classifier $h:\cX\rightarrow\cY$, they choose the least likely
class as $y_{\mathrm{LL}}$ as 
\[y_{\mathrm{LL}} = \argmax_{y\in\cY}\ell\br{h\br{\vec{x}}, y}.\]

The algorithm for the Iteratively Least Likely Class Method is described
in~\Cref{alg:ill}.

\begin{algorithm}
    \caption{Iterative Least Likely Class Method}
    \label{alg:ill}
    \begin{algorithmic}[1]
            \INPUT Original Data~($\vec{x}_d,y$), classification
            model($h, \ell$), Threat model~($T, \epsilon,\alpha$) 
            \STATE
            $\vec{x}_a^0 \gets \vec{x}_d$ 
            \FOR{$t\in\bc{0,\cdots
            T-1}$} 
            \STATE $y_{\mathrm{LL}} = \argmax_{y\in\cY}\ell\br{h\br{\vec{x}}, y}$
            \STATE $\vec{x}_a^{t+1} \gets
            \clipBig{\vec{x}_d}{\epsilon}{\vec{x}_a^t-\alpha\nabla_x\ell\br{h\br{\vec{x}_a^t},
            y_{\mathrm{LL}}}}$ \label{step:update_adv}
            \ENDFOR 
            \OUTPUT $\vec{x}_a^T$
        \end{algorithmic}
\end{algorithm}

\paragraph{Unstability of fixed steps adversarial attacks}
\label{sec:fixed-number-steps}

Both~\ifgsm and~\ill run for a fixed number of steps \(T\). An adaptive version
of these attacks add the adversarial noise for a maximum of \(T\) steps but
stops early upon successful misclassification even before \(T\) steps. We show
empirical evidence that an attack that adds noise for a fixed number of
steps~\citep{kurakin2016adversarial,kurakin2016} to the input is significantly
weaker than one that stops on successful misclassification. While it would be
natural to expect that once a classifier has misclassified an example, adding
more adversarial perturbation will only preserve the
misclassification,~\Cref{fig:adv_fix_step} suggests that a misclassified example
can be possibly classified correctly upon further addition of noise.

\begin{figure}[h!]
  \begin{subfigure}[c]{0.4\linewidth} \centering
\def\svgwidth{0.99\columnwidth} 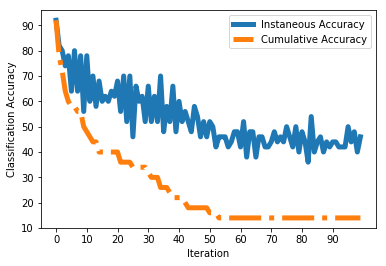
			 \caption{\ill \label{sfig:ll_inst_cum_adv}}
  \end{subfigure}\hfill
     \begin{subfigure}[c]{0.4\linewidth} \centering
\def\svgwidth{0.99\columnwidth}
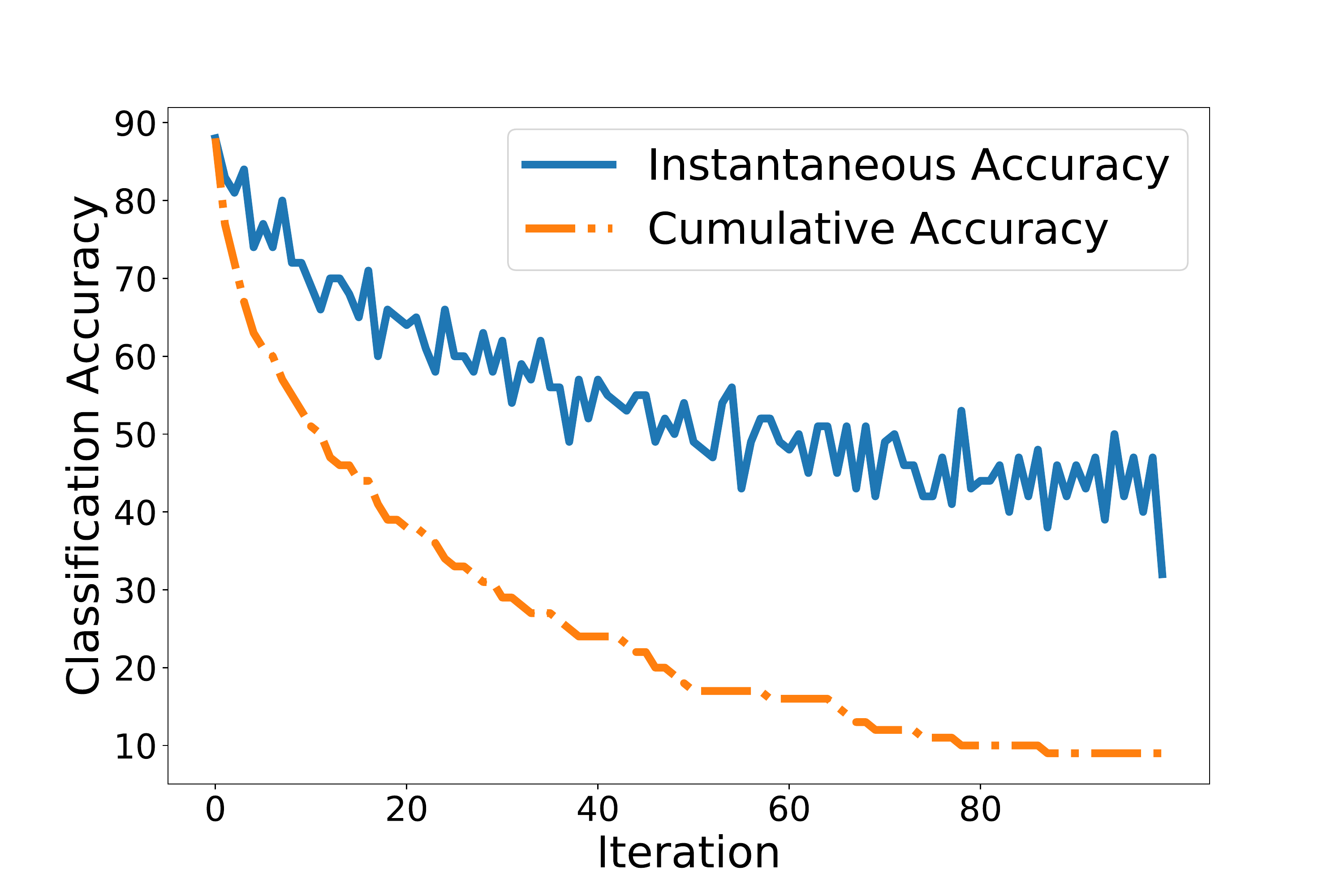
		  \caption{\ifgsm \label{sfig:fsgm_inst_cum_adv}}
        \end{subfigure}
         \caption[Instability of Adversarial Attacks]{An adversarial example that has
successfully fooled the classifier in a previous step can be
classified correctly upon adding more
perturbation. Figure~\ref{sfig:ll_inst_cum_adv} and
~\ref{sfig:fsgm_inst_cum_adv} refers to the two attack schemes - \ill
and \ifgsm respectively.}
         \label{fig:adv_fix_step}
       \end{figure}

Let $y_{a}(\vec{x}; k)$ be the label given to $\vx$ after adding
adversarial perturbation to $\vx$ for $k$ steps. We define
\emph{instantaneous accuracy} ($a_{\cI}(k)$) and \emph{cumulative
accuracy} ($a_{\cC}(k)$) as 
\begin{align}
    a_{\cI}(k) &= 1 -
\dfrac{1}{N}\sum_{i=1}^N \cI_{0,1}\bc{y_{a}(\vx; k) \neq y_{a}(\vx;
0)}\label{eq:instantaneous-acc}\\
a_{\cC}(k) &=1 - \dfrac{1}{N}\sum_{i=1}^N \max_{1\le j\le
k}\bc{ \cI_{0,1} \bc{y_{a}(\vx; j) \neq y_{a}(\vx; 0)}}.\label{eq:cumulative-acc}
\end{align}

In Figure~\ref{fig:adv_fix_step}, we see the \emph{instantaneous accuracy} and
the \emph{cumulative accuracy} for a ResNet model trained on the CIFAR10
dataset~(see~\Cref{sec:expr-settings} for a description of the model and the
dataset) where $\alpha=0.01,\epsilon=0.1$, and $T$ is plotted in the x-axis. The
cumulative accuracy is by definition a non-increasing sequence. However,
surprisingly the instantaneous accuracy is not monotonic and has a lower rate of
decrease than the cumulative accuracy. It also appears to stabilise at a value
much higher than the cumulative accuracy.

This is by no means an exhaustive list of proposed adversarial attacks but
merely a brief list of the main categories of these attacks. We discuss them
because we use them in later chapters to measure adversarial vulnerability.
Similarly, the defences proposed in the next section are not an exhaustive list
of all defence approaches that have been proposed against adversarial attacks.
\subsection{Adversarial defences}

There are various types of defences that have been proposed in the literature.
Some of them are based on the regularisation of the neural network~\citep{cisse17a},
some on data-augmentation~\citep{madry2018towards,Zhang2019}, and others on
theoretical guarantees~\citep{Lecuyer2019DP,pmlr-v97-cohen19c}.

\paragraph{Regularisation-based defences}

One of the main causes of adversarial vulnerability in neural networks is their
high sensitivity to input perturbations in certain directions. This means that
when the input is slightly perturbed in that direction there is an unexpectedly
large change in the output of the neural network. Sensitivity of the network is
commonly measured with various measures of lipschitzness\footnote{We discuss
lipschitzness in greater detail in~\Cref{sec:lipschitz}}.
Multiple works have discussed constraining these measures of lipschitzness of
neural networks in the context of adversarial vulnerability. The product of
operator norms of linear transformations of the individual layers of neural
networks is one way of measuring lipschitzness of neural
networks.~\citet{szegedy2013intriguing} discuss how linear and convolution
layers, whose operator norms are greater than one, magnify the adversarial
perturbation as the perturbation propagates through the layers. If the operator
norms are smaller than one then the transformations will attenuate the magnitude
of the perturbation as it propagates through the layers and protect against high
sensitivity in neural networks. To obtain this in practice,~\citet{cisse17a}
implement {\em Parseval tight frames}, which are extensions of orthogonal
matrices to non-square matrices. By constraining  the parameters of linear and
convolutional layers to remain in Parseval tight frames, they guarantee that the
lipschitzness of these layers will be smaller than one. While~\citet{cisse17a}
and~\citet{szegedy2013intriguing} use properties of individual weight matrices
to calculate the lipschitzness of the entire network, ~\citet{Tsuzuku2018} use
zeroth-order information of the network to approximate the
lipschitzness.~\citet{Tsuzuku2018} introduce lipschitz margin training, where
they directly regularise a differentiable approximation of the lipschitzness of
the network. Another way of approximating the lipschitzness is through first-order information obtained via the Jacobian of the network. Multiple
works~\citep{roth2019adversarially,Ross2018,Jakubovitz_2018_ECCV,Lyu2015} use
Jacobian regularisation to constrain the lipschitzness and thus the sensitivity
of the neural networks. All of these approaches based on regularising the
lipschitzness of neural networks provide empirical boosts to adversarial
robustness.

However, such regularisations pose a problem with regards to the faithful
measurement of the true robustness of neural networks. As pointed out
by~\citet{athalye2018obfuscated}, directly penalizing gradients around points
from the training data sets up gradient-based attacks to fail  without
regularising the loss surface outside the immediate vicinity of that point.
These approaches protect against the computation of successful gradient-based
adversarial attacks around the training points but the network still remains
vulnerable to other attacks, for example, attacks that do not use first-order
information at training points. This phenomenon is called "Obfuscated Gradients"
and~\citet{athalye2018obfuscated} show that multiple defences that exhibit this
phenomenon can be overcome with specially designed attacks.
\todo[color=green]{Check all paragraph heading are consistent}
\paragraph{Noisy training-based defence}
Another way of regularising a neural network is to add noise to the training
procedure.~\citet{bishop1995training} show that training with noise is,
sometimes, equivalent to a form of Tikhonov
regularisation.~\citet{Jin2015robust} show that adding stochastic noise to the
input and model parameters during training improves the adversarial robustness of
convolutional neural network~(CNN) models.~\citet{dhillon2018stochastic}
propose randomly pruning activations and their experiments indicate that this
strategy improves the robustness of networks without requiring further
fine-tuning.~\citet{Sankaranarayanan2018} perturb intermediate layer
activations and use this as a regulariser during training to impart robustness
to trained networks.

\paragraph{Defence based on smooth training}
In regularisation-based defences, we saw that regularising a network to
constrain the operator norm of each layer improves robustness of the network in
practice. The reason why the operator norm of different layers increases in
practice is a combination of positive homogeneity of neural networks and
exponential loss functions like cross-entropy with one-hot target labels.
Positive homogeneity means that multiplying the weights by a positive constant
does not alter the accuracy of the network but can change the logits in the
penultimate layer. A key property of exponential loss functions with one-hot
target labels is that the only way to reduce the loss to zero is by increasing
the magnitude of weights and consequently the logits to infinity. Thus, neural
network training is inherently biased towards magnifying the operator norms of
each layer of the network, and this leads to heightened sensitivity and worse
robustness of neural networks.

A natural defence to control this behaviour is to prevent the exponential loss
function from encouraging the magnification of the operator norms of the weight
matrices in a neural network. One way to do this is by replacing the one-hot
target vectors with smoothed vectors. Two common techniques in the literature to
achieve this are {\em defensive distillation}~\citep{Hinton2015distillation} and
{\em label smoothing}~\citep{SzegedyVISW16}. In defensive distillation, a neural
network is first trained on the classification problem, and then the one-hot
target label vectors in the training dataset are replaced with class
probability\todo[color=green]{replace data point with data instance everywhere} vectors
obtained from the trained network's prediction on the points from the
training dataset. Then a new network is trained on the modified dataset~(where
the label vectors are smoothed). First proposed
in~\citet{Hinton2015distillation} as a method for transferring knowledge from
larger networks to smaller networks,~\citet{papernot2016distillation} adapted
this strategy for imparting adversarial robustness and renamed it~{\em defensive
distillation}.

Another form of regularisation that directly prevents models from increasing the
operator norms of their layers is through {\em label
smoothing}~\citep{SzegedyVISW16}. It is a regularisation technique that
introduces noise for the labels. For a small constant $\epsilon$, in  a
$k$-class classification problem, label smoothing regularises a model by
replacing the hard $0$ and $1$ classification targets in the one-hot target
label vector with $\frac{\epsilon}{k-1}$ and $1-\epsilon$ respectively. Its
benefits were initially observed for calibration
in~\citet{muller2019does}.~\citet{Goibert2019smoothing} conducted a more
systematic study in the context of adversarial robustness and explored various
smoothing techniques especially those that are more suited for adversarial
robustness. 

\paragraph{Data-augmentation-based defences}

The most popular form of adversarial defence by far is {\em adversarial
training}~\citep{madry2018towards} and its more sophisticated
variants~\citep{Zhang2019}. Essentially, it exploits the fact that adversarial
examples are so ubiquitous in deep neural networks that constructing them is
relatively straightforward using gradient-based attacks as discussed in the
previous section. Adversarial training augments the training procedure by
replacing the original training point with an adversarial training point
constructed by an adversary. It can be viewed as minimizing an approximate
minimisation of the adversarial risk defined in~\Cref{defn:adv_risk}. 

However, it has been observed empirically that adversarial training causes a
perceived tradeoff between robustness and
accuracy~\citep{kurakin2016adversarial}.~\citet{Zhang2019} provide a
differentiable upper bound on the combined adversarial and natural test error
and term the algorithm for minimizing this upper bound TRADES. Using TRADES
instead of adversarial training, they show that this perceived trade-off can be
avoided to a certain extent. This is perhaps the most commonly used variant of
adversarial training in practice. 

\paragraph{Defences based on detection}

Another line of defence is based on detecting adversarial examples so that  the
classifier is not forced to predict a label for an example if a detector can
detect the example to be an adversarial example. ~\citet{Hendrycks2016detecting}
present three methods to detect adversarial examples which rely on properties
like identifying the subspaces in the image space exploited by the adversary and
abnormal softmax values for adversarial examples as compared to natural
examples. ~\citet{yang2018characterizing} use temporal dependency in audio
signals to detect adversarial examples specifically for Automated Speech
Recognition tasks. In a \(k\)-class classification
problem,~\citet{Yin2019detection} use \(k\) different detectors to do the
detection. In particular, if the model predicts the class to be $i$, the $i^{\it
th}$ detector is used to confirm that the image is not adversarially
perturbed.~\citet{akhtar2018defence} learn both a detector network  and a
Perturbation Rectification Network~(PRN) and use the PRN to rectify the
perturbation when a perturbation is detected by the detector network.

However, such detection methods usually suffer a great deal from incorrect and
weak evaluations. ~\citet{Carlini2017a} and~\citet{Tramer2020} show how a dozen
of these detection methods can be successfully bypassed by designing better
adversaries. This suggests that a lot of these properties that were thought to
be inherent to adversarial examples are in fact not inherent to the problem of
adversarial robustness but rather artefacts of the particular techniques used
for constructing the adversaries in their evaluation techniques. This calls for
developing defence methods with theoretical guarantees for robustness agnostic
to the particular technique used for generating adversarial attacks during
empirical evaluation.

\paragraph{Defence with guarantees}

The famous Goodhart's law~\citep{goodhart1984problems} states that 

\begin{displayquote}[Goodhart]
Any observed statistical regularity will tend to collapse once
pressure is placed upon it for control purposes.
\end{displayquote}

Arguably, the cat and dog race that has prevailed in the realm of adversarial
attacks and defences can be attributed to this law. As discussed, the
minimisation of the true adversarial risk is a difficult problem.  Thus most
defence techniques tend to minimise one particular approximation of the risk.
Future attacks that break the defence usually succeed by breaking that
assumption inherent to that approximation. 

However, recent work has looked at creating provable defences against
adversarial vulnerability. In terms of guarantees of robustness, the most
powerful of those come from the verification literature. Verification, in the
context of adversarial robustness, means to provide a theoretical guarantee that
if inputs to the neural network belong to a pre-defined set of inputs~(eg.
inputs within a norm ball), then the outputs of the network satisfy a desired
property~(eg. the logit corresponding to the correct label has a higher value
than all the other logits). Research from the formal verification community has
used Satisfiability Modulo Theory~(SMT) solvers to provide guarantees against
bounded norm perturbation
attacks~\citep{ehlers2017formal,huang2017safety,katz2017reluplex}. However, the
verification bounds from these approaches, while sufficient to guarantee
robustness, ends up guaranteeing robustness against only very weak adversaries.
Thus, they might be of little practical significance. 

To make these bounds more useful, a line of research has tried to alter the
training process of neural networks so that the bounds obtained from these
trained networks are more practically
relevant~\citep{Raghunathan2018verification,wong2018provable}. Another line of
research has proposed approaches that use {\em branch and bound} algorithms
~\citep{bunel2018verification,cheng2017maximum,Tjeng2017}  to provide stronger
bounds. However, all of these methods are usually hard to adapt for large
networks as the computational complexity of these approaches depends directly on
the SMT problem instances. Moreover, they are usually limited to problems with
piecewise linear activation functions like ReLU and maxpooling and it has proved
hard to adapt them to general architectures and activation functions unless
further approximations are made. Consequently, this is an active direction of
future research with a potentially significant impact on certifying the
robustness of deep neural networks.

Another approach to providing guaranteed robustness is through the approach of
PixelDP~({\em Pixel Differential Privacy})~\citep{Lecuyer2019DP,Li2018smoothing}
or randomised smoothing~\citep{pmlr-v97-cohen19c}. While verification
techniques, as discussed in the previous paragraph, prove the robustness of an
existing classifier, randomised smoothing techniques construct a new classifier
that is smooth within a {\em certified radius}; the {\em certified radius} of
the smoothed classifier depends on a hyper-parameter of the smoothing process
and properties of the original model.

Specifically, consider a classification problem from $\reals^d$ to the set of
labels $\cY$ and  a base classifier $f:\reals^d\rightarrow\cY$ that has been
trained to do the classification. The objective of randomised smoothing is to
use a smoothing distribution $\cN_S$ to obtain a smoothed classifier
$g:\reals^d\rightarrow\cY$ from $f$, such that when queried at
$\vec{x}\in\reals^d$, the smoothed classifier $g$ returns whichever class $f$ is
most likely to return when $\vec{x}$ is perturbed by noise drawn from the
smoothing distribution $\cN_S$. 

The larger the variance of the smoothing distribution, the greater the magnitude
of the radius that can be certified for the smoothed classifier. One of the key
questions of research in this direction is what kind of smoothing distribution
should be used for a particular kind of adversary.~\citet{Lecuyer2019DP}
and~\citet{pmlr-v97-cohen19c} used the gaussian distribution to certify against
an adversary with an $\ell_2$ perturbation bound.~\citet{yang2020randomized}
generalised this to other kinds of adversaries and proposed a generic technique
to find the best possible distribution for multiple $\ell_p$ perturbation
bounds.~\citet{Awasthi2020} further improved these approaches by leveraging
natural low-rank representations of data to provide improved guarantees.

It is easy to see that if the variance of the smoothing distribution is very
large, then the classifier may become a constant
predictor.~\citet{Mohapatra2020} shows that with increasing noise variance, the
decision regions shrink in size and the classifier becomes overwhelmingly likely
to always predict one specific class and ignore the others. Thus, there seems to
be a tradeoff between robustness and accuracy when randomised smoothing is used
as a technique to guarantee robustness. 
\subsection{Tradeoffs associated with robustness}\label{sec:robustness-tradeoffs}
Several works have suggested that robustness is inherently at odds with other
measures that are important in the learning problem or even unavoidable,
entirely. 

~\citet{fawzi18} constructs a distribution where data~\(\cX\) is generated from
a generative model \(\cX=g\br{\vec{r}}\), where \(\vec{r}\in\reals^d\) is
sampled from an isotropic gaussian and is, roughly, of euclidean norm
\(\sqrt{d}\). They show that due to isoperimetry of gaussian distribution, no
classifier is robust to adversarial perturbations of euclidean norm~\(\bigO{1}\)
irrespective of its (clean) expected error~(\Cref{defn:exp_loss}). When \(g\) is
an \(L\)-\gls{lip} function, this corresponds to perturbations  of at most
\(\bigO{L}\). ~\citet{tsipras2018robustness} show a slightly more optimistic
setting, where robust classification is possible but not simultaneously
achievable with low expected error. In particular, they construct a simple
setting where, for any classifier, as expected error approaches \(0\%\),
adversarial error~(\Cref{defn:adv_risk}) tends towards \(100\%\). They use a
distinction between {\em robust} and {\em non-robust} features, where the
classifier needs to use the non-robust features to attain a low expected error
at the cost of a large adversarial error. Similarly, using the robust features
leads to low adversarial error at the cost of large expected
error.~\citet{Zhang2019} show another simple setting where the bayes-optimal
classifier obtains a low expected error but a large adversarial error whereas
the optimal adversarial error is obtained by a constant classifier at the cost
of a large expected error. 

Interestingly, in both of these examples, neither additional data nor choosing a
different representation of data can help in making the classifiers more robust.
However, humans are considered to be robust as well as accurate classifiers at
least for natural images. Thus, any result that rejects the possibility of a
robust and accurate classifier entirely is highly unlikely to apply to
real-world data. ~\citet{schmidt2018adversarially} consider a setting where
learning with low adversarial error is possible without being at odds with
expected error but requires more data than learning a classifier with low
expected error. In their setting, learning a classifier with low expected error
requires just one sample whereas learning a classifier with low adversarial
error requires \(\Omega\br{\sqrt{d}}\) samples where \(d\) is the dimensionality
of the input space i.e. a polynomial separation in sample
complexity.~\citet{bpr18} show that this gap is tight, in the sense that if
non-robust learning~(low expected error) is possible with polynomial sample
complexity then robust learning is also possible with polynomial sample
complexity. While this might seem like an optimistic result, empirical evidence
suggests that this is not observed in practice.

~\citet{blpr18} and ~\citet{degwekar19a} showed~(under some cryptographic
assumptions) that there exist learning tasks where a computationally efficient
robust and accurate classifier exists, can be learnt from a small number of
samples, but the learning algorithm will necessarily be computationally
expensive. A computationally efficient classifier is defined as a classifier
that can compute the output, given the input in time polynomial in the size of
the input.~\citet{degwekar19a} further extend along this direction showing
settings where robust classification is possible but only via a computationally
expensive classifier. Their results show that, for certain distributions,
computationally simple hypothesis classes do not admit a robust classifier
whereas robust learning is possible using a different, more
complex~(computationally) hypothesis class. ~\citet{madry2018towards} provide
some empirical validation of the hypothesis in~\citet{degwekar19a} in the sense
that choosing wider and deeper neural networks have generally led to lower
adversarial error even when simpler models~(shallower or narrower neural networks)
models suffice for low expected risk.

~\citet{montasser19a} establish that there are hypothesis classes with finite
VC dimensions i.e. are \emph{properly} PAC-learnable but are only
\emph{improperly} robustly PAC learnable. This implies that to learn the problem
with small adversarial error, a different concept class is required whereas for
low expected risk, the original hypothesis class suffices.
Unlike~\citet{degwekar19a}, the difference in complexity between the two concept
classes here is not that of computational complexity but rather statistical~(or
sample) complexity.

All of these works indicate that for different distributions learning tasks
robustness might be at odds with different quantities --- accuracy, statistical
complexity, computational complexity of learning, and computational complexity
of the hypothesis class. However, real-world data distributions and learning
tasks might not possess the same hardness as the distributions and learning
tasks used in the construction of these examples. This presents a hope that
these tradeoffs, despite appearing in practice, are not inherent to the real
world learning problems but are only artefacts of the algorithms and hypothesis
classes we use in practice.

\todo[color=blue]{Another section on robustness tradeoffs}

\section{Calibration of Neural Networks}
\label{sec:prelim_calibration}
\paragraph{Calibration in multi-component systems}
Generalisability of machine learning models, as discussed in~\Cref{sec:gen_nn},
allows us to extrapolate our confidence in the performance of a machine learning
model from a fixed-size training set to {\em unseen} data. Adversarial
robustness of machine learning~(~\Cref{sec:robustness}) models  provides
confidence that the performance of the model will not be vulnerable to malicious
changes to the input data. Another component of the reliability of machine learning
models is {\em calibration}. When machine learning models are deployed in the
real world, they are often deployed to assist humans in certain tasks or they
are used in an ensemble with other machine learning models. 

For example, in medical science, machine learning models are used to combine
multiple medical signals to determine rapidly whether the patient might require
imminent critical care~\citep{Churpek2016,Kipnis2016,Brekke2019} or to diagnose
health conditions based on imaging data~\citep{Oktay2020,Ardila2019}. However,
these solutions are usually meant to assist a clinician as opposed to remove the
clinician from the decision-making loop. Thus, the clinician needs to have an
estimate of the confidence of the machine learning system on its prediction.
Such an estimate allows the clinician to reliably trust the machine learning
system's diagnosis or overrule its diagnosis with their own diagnosis.

In automated driving systems, machine learning is used for tasks like detecting
lane changes, labelling obstacles, steering, accelerating, and braking. Multiple
machine learning components, usually relying on different sensors, are used to
make these decisions. To reliably combine the predictions of these
different components, a measure of the confidence of the individual components
is essential. For example, when visibility is poor it is safer to trust a
component that does not depend on image data whereas when information on the
road is presented with signs and visual cues, it is safer to trust a
vision-based component. Therefore, if a system can calibrate its confidence on
its various components, their outputs can be combined effectively to provide a
reliable prediction.

In both of these situations, an estimation of the uncertainty of the decisions
made by the individual machine learning models is required for the safe
deployment of the entire system. One particular measure of this uncertainty is through the notion of
calibration. 
Deep neural networks are designed to output a vector in
a $\abs{\cY}$-dimensional simplex. This is usually done by first producing a
$k$-dimensional vector $\vec{z}\in\reals^k$ and then a softmax function
transforms $\vec{z}$ element-wise to $\mathrm{softmax}\br{\vec{z}}\in\Delta^k$.
\begin{equation}\label{eq:softmax-fn}
    \mathrm{softmax}\br{\vec{z}}_i = \dfrac{\exp{\br{\vec{z}_i}}}{\sum_{i=1}^k\exp{\br{\vec{z}_i}}}
\end{equation} 
This vector can be interpreted as a conditional likelihood distribution
$\bP\bs{\cY\vert \cX, h}$\footnote{This is a slight abuse of notation as we have
not defined $h$ to be a random variable.} over the label space $\cY$ given the
input $\cX$ and the model~(architecture-cum-parameters) $h$. A calibrated model
returns a {\em probability} vector  where the individual entries
$\bP\bs{\cY=y\vert \cX=\vec{x}, h}$ of the vector are truly indicative of the
actual likelihood of the class $y$ as a correct prediction for the  input
$\vec{x}$. In the rest of this section, we discuss the different notions of
calibrations discussed in the literature, different approaches for measuring
calibration of models in practice, and finally approaches to calibrate a model.

\subsection{Types of calibration}\label{sec:types-of-calibration} A strong
notion of calibration is discussed
in~\citet{Kull2019,Widmann2019,Vaicenavicius2019}, and~\citet{Kumar2019a}. Let
$h\br{\vec{x}}\bs{y}$ be the conditional probability that the model $h$
attributes to the $y^{\it th}$ class on seeing the input $\vec{x}$. Strong
calibration requires that for a fully calibrated model $h$, the following holds
for all $y\in\cY$ and all $p\in\bs{0,1}$:
\begin{equation}\label{eq:strong-calib}
    \bP\bs{\cY=y~\vert h\br{\vec{x}}\bs{y}=p } = p.
\end{equation} 
This implies that for any class $y$, for all examples to which the model assigns
a conditional likelihood of $p$ for belonging to that class, the model should be
correct on a $p$ fraction of those examples. This is an especially strong notion
of calibration as it accounts for all classes for every example even when it is
not the correct class for that example.

A relatively weaker notion of calibration, known as {\em weak calibration}, is
discussed in~\citet{Guo2017}. This notion only requires that the conditional
probability for the predicted class be calibrated as opposed to all the
classes being calibrated. This notion of calibration requires that for all
$p_{max}\in\bs{0,1}$, we have

\begin{align}\label{eq:weak-calibration}
    \bP\bs{\cY=\argmax_{y}h\br{\vec{x}}\bs{y} \biggl\vert \max_{y}h\br{\vec{x}}\bs{y} = p_{max} }= p_{max}.
\end{align}

~\citet{Kull2019} also consider another measure, known as {\em class-specific
calibration}, that is not as restrictive as strong calibration but is stronger
than weak calibration. Under this measure,  calibration is enforced only for a
specific class. Class-specific calibration is ensured for the class $c^*$ if
the following holds for all values $p_{c^*}\in\bs{0,1}$: 
\begin{equation}
    \bP\bs{\cY=c^*\biggl\vert h\br{\vec{x}}\bs{c^*}=p_{c^*}} = p_{c^*}.
\end{equation}

\subsection{Measuring miscalibration}\label{sec:measuring-calibration} The
previous section discusses three notions of calibration expressed as conditional
probabilities. However, the conditional probabilities  cannot be computed
accurately with a finite number of samples since the conditional likelihood is a
continuous random variable. The natural way of dealing with  issues like these
in practice is through discretisation of the continuous random variable.
Ideally, weak calibration, as defined in~\Cref{eq:weak-calibration}, would be
measured by grouping all examples $\vec{x}$ whose likelihood of the predicted
class is exactly $p$ and then measuring the absolute difference between the
accuracy of the model in that group and the value $p$. If the difference is zero
for all values of $p$, then the model is perfectly weakly-calibrated. However,
as mentioned before this is impractical as the value $p$ is continuous. So, the
problem is usually tackled through discretisation of the $\bs{0,1}$ interval for
values of $p$.

\paragraph{Expected calibration error} One of the most commonly used metrics to
measure calibration in practice is the Expected Calibration
Error~(ECE)~\citep{Naeini2015}. The measure is characterised by an integer
hyper-parameter $M>1$ that is used for discretisation. The $\bs{0,1}$ interval
is divided into $M$ equal-width bins~$\bc{B_1,\ldots,B_M}$ where
$B_i=\bs{\frac{i-1}{M},\frac{i}{M}}$. Then, all predictions on the test-set are
categorised into one of these bins depending on the conditional likelihood~(i.e.
the softmax value) of the predicted class. For example, if an example is
predicted to be in class $c$ with probability $p$, that example is placed in the
bin $B_i$ such that $\frac{i-1}{M}\le p< \frac{i}{M}$. Then, the confidence
$C_i$ and accuracy $A_i$ of bin $B_i$ are computed as the average
conditional likelihoods and the average accuracy respectively of the examples in
that bin. ECE is measured as 
\begin{equation}
    \textrm{ECE}\br{h} = \sum_{i=1}^M \frac{\abs{B_i}}{N}\abs{A_i - C_i},
\end{equation}
where $\abs{B_i}$ represents the number of examples in the bin $B_i$ and $N$ is
the total number of examples in the test-set. 

\paragraph{Maximum Calibration Error} While ECE can be thought of as the $L_1$
calibration error, another popular measure of calibration error is $L_\infty$
calibration error, otherwise known as the Maximum Calibration
Error~(MCE)~\citep{Naeini2015}. To compute MCE, the bins $B_i$, their
corresponding accuracies $A_i$, and confidences $C_i$ are computed as above.
Then, MCE is computed as 
\begin{equation}
    \textrm{MCE}\br{h} = \max_i\abs{A_i - C_i}.
\end{equation}

\paragraph{Modifications to the calibration errors} Other versions, including
the $L_2$ version~\citep{Kumar2019a}, have also been used in literature to
measure the calibration error. A consequence of discretising $p$ uniformly is
that each bin can end up with a wildly varying number of samples. For neural
networks, most examples end up in a bin with a high value of $p$. Thus, the  low
confidence bins consist of very few samples and  this can heavily impact the
measures of both ECE and MCE. To overcome this, both~\citet{Nixon2019} and our
work~(discussed in~\Cref{chap:focal_loss}) propose the use of adaptive binning
strategies~(Adaptive ECE or AdaECE) where the bins are created to ensure that
every bin is equally populated.

\paragraph{Classwise expected calibration error} Another drawback of the
calibration errors is that they are designed to only measure weak calibration as
they account for the confidence of only the predicted class. Stronger
definitions of calibration require that all the classes be
calibrated.~\citet{Nixon2019} propose a new metric called the Static Calibration
Error~(SCE) to overcome this. We will refer to this as classwise ECE instead. To
compute this metric, $M$ separate bins indexed as $B_{i,j}$ are created  for
each of the $K$ classes. The bin $B_{i,j}$ represents the $i^{\it th}$ bin for
the $j^{\it th}$ class. Once all the examples have been binned into their
respective bins, the accuracies $A_{i,j}$ and confidences $C_{i,j}$ are computed
for each bin. Then, classwise ECE is measured as follows
\begin{equation}
    \textrm{Classwise-ECE}\br{h} = \frac{1}{K}\sum_{i=1}^M\sum_{j=1}^K\dfrac{\abs{B_{i,j}}}{N}\abs{A_{i,j}-C_{i,j}}
\end{equation}

\paragraph{Reliability plots} These metrics of calibration error summarise the
error into a single statistic without offering an insight into which of the bins
were more mis-calibrated and which were less. A useful visual tool for this is
the reliability plot introduced in~\citet{NiculescuMizil2005}. The reliability
plot is a bar plot where the x-axis measures the confidence and the y-axis
measures the accuracy. Each bar in the reliability plot represents a bin and is
put on the X-axis in increasing order of the average confidence of that bin. The
height of the bar denotes the average accuracy of the examples in that bin. For
a fully calibrated model, the height of each bar should be the same as its
X-axis label. If the height is more than its X-axis label, then the model is
said to be under-confident and if the height is less than the X-axis label, then
the model is said to be over-confident.

\paragraph{Other metrics of calibration} While the  ECE, its variants, and
reliability plots have remained the most widely used metrics for
mis-calibration, several other loss functions are also used to measure
mis-calibration of models. The most popular of these are the Negative Loss
Likelihood and the Brier Score~\citep{Brier1950verification}. While some works
have looked at minimizing these measures directly to boost both accuracy and
calibration, those techniques have also led to an apparent trade-off between
accuracy and calibration. It is easy to see that a model can be extremely
inaccurate while being fully calibrated and sometimes, this solution is
preferred over the more desirable one of being nearly accurate with very low
calibration error~\citep{Guo2017}.

\subsection{Approaches for calibrating deep neural networks}
\label{sec:approaches-calibration} 
Multiple approaches have been proposed in the literature for calibrating a
neural network in practice. They can be broadly categorised into approaches for
post-hoc calibration, and approaches for calibration during training.

\paragraph{Post-hoc calibration} The basic idea for post-hoc calibration is to
learn a function that maps the uncalibrated output of the machine learning model
into calibrated likelihoods. Different approaches for post-hoc calibration vary
in the nature of this function. Without any restriction on the nature of the
function, the function itself can overfit to the training data and not
generalise to new data. Thus various strategies of post-hoc calibration
restrict this function to different classes of parametric and non-parametric
functions.

\paragraph{Platt scaling}~\citet{Platt1999} proposed a post-hoc calibration
technique through scaling the output of the model via a sigmoid function
parameterised by two learnable scalars $a$ and $b$. In this approach, the output
of the machine learning model $h\br{\vec{x}}$ is replaced by
$\dfrac{1}{1+\exp{\br{ah\br{\vec{x}}+b}}}$. Their initial approach was applied
for SVMs and was motivated by the empirical observation that a sigmoid function
captured the relationship between SVM scores and empirical conditional
likelihoods for many commonly used datasets .

~\citet{Guo2017} modified Platt scaling for use in neural networks by removing
the bias term in the affine transformation and extending it to multi-class
classification. They termed this approach {\em temperature scaling}, which
operates on the pre-softmax vector~(also referred to as the logit vector) by
replacing the original softmax function~\Cref{eq:softmax-fn}
with~\Cref{eq:TS-softmax-fn}
\begin{equation}\label{eq:TS-softmax-fn}
    \mathrm{softmax}_{\mathrm{TS}}\br{\vec{z}; T}_i = \dfrac{\exp{\br{\nicefrac{\vec{z}_i}{T}}}}{\sum_{i=1}^K \exp{\br{\nicefrac{\vec{z}_i}{T}}}}
\end{equation}
This approach, however, suffers from multiple drawbacks. For example, while it
scales the logits to reduce the network's confidence in incorrect predictions,
it also reduces the network's confidence in predictions that are
correct~\citep{Kumar2018}. Moreover, it has been observed that temperature
scaling does not calibrate a model under data distribution
shift~\citep{Ovadia2019}. Despite these drawbacks, temperature scaling is
perhaps the most commonly used method for post-hoc calibration. This is due to
its simplicity and impressive performance on a wide range of neural network
architectures~\citep{Guo2017}.

However, when the exact nature of the mapping function is unknown it makes
little sense to use a sigmoid function with a fixed structure. For this
purpose,~\citet{Zadrozny2001} proposed a non-parametric technique referred to
as binning. In this technique, the outputs of the function on the training set
are sorted in decreasing order of magnitude  of the conditional likelihood and
then placed into bins of equal size. When a test example is evaluated, it is
first placed into one of the bins depending on the output of the model on that
example. Then the output of the model on that test example is replaced with the
average accuracy of training examples in that bin.

Another approach that is mid-way between a parametric model and a fully
non-parametric method is {\em isotonic regression}~\citep{Nueesch1991}, which
is a form of non-parametric regression. In isotonic regression, the learned
function is chosen from the class of all non-decreasing~(or isotonic) functions.
The underlying intuition for why this is suitable for calibration is that even
though the base machine learning classifier might not output the correct
conditional likelihoods, it should still rank the classes properly. In that
case,  the correct mapping from the space of model outputs to the true
conditional likelihood is an isotonic function.

Several other post-hoc calibration techniques have been proposed in recent
 literature. These include further modifications to temperature scaling like
 Bin-wise Temperature Scaling~(BTS)~\citep{Ji2019}, learning sample-wise
 temperature parameters~\citep{Ding2020}, and Dirichlet
 Calibration~\citep{Kull2019}. 
 
 \paragraph{Calibrating during training} While post-hoc calibration techniques
 are easy to implement and can be used to calibrate a model without interfering
 in its training process, there are multiple issues associated with these
 techniques. First, post-hoc methods require a considerably large validation set
 to tune themselves on. Second, there is an extra computational overhead
 associated with training the calibration technique once the training of the
 base model is complete. Methods that calibrate the model during training
 overcome these issues by producing a calibrated model at the end of training
 without requiring an extra calibration phase of training. One popular way of
 calibrating during training is by training a network using modified loss
 functions. 
 
 The brier Score, introduced by~\citet{Brier1950verification}, is a loss
 function that measures the accuracy of probabilistic predictions. It can
 intuitively be thought of as a squared error between the predicted likelihood
 vector and the one-hot target. Let the $K$-dimensional one-hot representation
 of the label $y$ be $y^{01}$. Then, the brier score of a model $h$ on the
 example $\br{\vec{x},y}$ is
 \begin{equation}\label{eq:brier-score-defn}
     BS\br{h;\vec{x},y} = \sum_{i=1}^N\sum_{j=1}^K\br{h\br{\vec{x}_i}\bs{j} - y_i^{01}\bs{j}}^2.
 \end{equation}
 The Brier score can be algebraically decomposed into two components associated
 with accuracy and calibration. Thus, minimizing Brier loss provides a
 simultaneously accurate and calibrated model as we observe
 in~\Cref{chap:focal_loss}. However, we also observe that while minimizing the
 Brier score provides better calibration error than NLL, it still trades off
 calibration error for test error.

We could try to minimise the calibration errors we saw in the previous sections
 directly, but most of the errors like ECE and MCE are non-differentiable
 metrics.~\citet{Kumar2018} propose a differentiable proxy for the calibration
 error, called {\em Maximum Mean Calibration Error~(MMCE)}, which they add as an
 extra regulariser during training in addition to the negative log-likelihood
 loss.

 Several other approaches have been proposed for calibration during
 training.~\citet{muller2019does} adapt label smoothing for calibration for
 reasons similar to those discussed in the context of adversarial robustness.
 Similarly,~\citet{Thulasidasan2019} show that using mixup~\citep{Zhang2018a}
 training also leads to a drop in calibration error. 
 
 In~\Cref{chap:focal_loss}, we propose the use of focal loss~\citep{Lin2017} as
 a loss function that can simultaneously provide high accuracy and low
 calibration errors. We show that minimizing focal loss is approximately
 equivalent to minimizing a regularised Bregman divergence where the
 regularisation component helps with calibration and the Bregman divergence is
 associated with the usual negative log-likelihood classification loss. Our
 experiments show that our method performs better than the other methods
 mentioned in this section. Combining focal loss with temperature scaling
 further boosts its performance. We use the various metrics discussed in this
 section to report the performance.

 \paragraph{Model ensembles and other approaches}
 An approach that combines post-hoc calibration and calibration during training
 are model ensembles. Model ensembles combine the predictions of multiple
 models, trained on different subsets of the data, and they have long been known
 to improve generalisation~\citep{Hansen1990} of machine learning models. The
 diversity of the models in an ensemble have been known to help in
 generalisation and recent works~\citep{Sinha2020diversity,Kim2018attention}
 have developed approaches to increase this diversity in
 practice.~\citet{Raftery2005} and ~\citet{Stickland2020} have shown that
 ensemble diversity also helps with improved model
 calibration.~\citet{Zhong2013ensemble} uses ensembles of SVMs, logistic
 regressors, and boosted decision trees for improved calibration.

 In the case of neural networks, various works
 including~\citet{Lakshminarayan2017} and~\citet{Ovadia2019} have shown that
 ensemble of neural networks can also help with calibration in deep learning.
 Calibration of neural networks is closely related to estimating predictive
 uncertainty, which is something bayesian neural networks~(eg. Variational
 inference or MCMC techniques) excel at. However, bayesian neural networks are
 computationally intensive and hard to implement. In contrast, {\em deep
 ensembles} proposed by~\citet{Lakshminarayan2017} is a simple approach that
 scales well and uses a combination of ensembles and proper scoring
 functions.~\citet{Ashukha2020Pitfalls} provides empirical evidence that deep
 ensembles outperform some bayesian neural networks approaches in regards to
 calibration.~\citet{Wilson2020} develops an approach called MultiSWAG which
 combines the benefits of bayesian deep learning and deep ensembles to propose
 an approach that is not computationally more expensive than deep ensembles~(at
 train time) but provides better calibration than deep ensembles. However, their
 approach is indeed more expensive to use at test time.~\citet{Wenzel2020HowGI}
 provides empirical evidence that the posterior predictive induced by the Bayes
 posterior yields systematically worse predictions compared to point estimates
 obtained from SGD. Interestingly, we have not found papers that propose
 ensemble techniques and bayesian neural networks for calibration to also report
 results on calibration metrics like ECE and MCE, as we discussed above.
 Nevertheless, all of these approaches are complementary to the usage of a
 different loss function, which is the approach we take in this thesis.
\section{Privacy in deep learning}
\label{sec:privacy}
So far we have discussed notions of reliability in machine learning viz.
generalisation, robustness, and calibration  that deal directly with
guaranteeing certain properties in the learned machine learning models. In this
section, we will look at a different issue that arises when machine learning
models are deployed as a service~(MLaaS). With rapidly evolving machine learning
algorithms for a wide range of tasks, growing computational capabilities to
deploy these algorithms in practice, and ever-growing access to the internet and
online services, MLaaS is poised to become, if it is not already, a ubiquitous
component of our lives. However, along with its numerous benefits, this also
poses major potential issues arising from inadequate security protocols to
protect the data that is fed into these machine learning algorithms either
during training, during inference, or both.

While privacy issues arising from amassing and storing data for training a
machine learning model are important, in this thesis we will not focus on that
aspect of privacy in machine learning. Instead, we will focus on an important
recent paradigm called \emph{prediction as a service}, whereby a service
provider with expertise and resources can make predictions on data provided by
the clients. However, this approach requires trust between the service provider
and client; there are several instances where clients may be unwilling or unable
to provide data to service providers due to privacy concerns. Examples include
assisting in medical diagnoses \citep{K:2001,BSHLKPRK:2017}, detecting fraud
from personal finance data \citep{GR:1994}, and detecting online communities
from user data \citep{F:2010}. The ability of a service provider to predict on
private data without being able to see the actual data can alleviate concerns
of data leakage.  We will look at a paradigm where a service provider has
trained a machine learning model and customers can use that model by sending
their data to the service provider. 

\subsection{Techniques for Privacy-Preserving Inference}
Consider a medical diagnosis company that provides a service where customers can
upload chest radiographs and the company's algorithms diagnose whether the
patient is suffering from a bone fracture. If the algorithms are accurate and
reliable under the notions of reliability that we have seen in the previous
sections, this can be a very valuable service with the ability to rapidly
diagnose multiple patients without human intervention. However, without
sufficient trust between the company and the customer, the customer might be
disinclined to upload their chest radiographs to the service. For example, a
dishonest company can use the images to estimate their general lung health and
sell that information to insurance companies or predict whether the customer
smokes and sell that information to tobacco companies. Both of these outcomes
can have very damaging financial and health consequences for the customer. To
prevent these possibilities, the machine learning algorithms should protect the
privacy of the user while making accurate predictions. The broad field concerned
with this is commonly referred to as {\em Privacy-Preserving Inference}. The
following paragraphs discuss three generic techniques for tackling this
problem.

\paragraph{Trusted Execution Environments} Broadly, the problems discussed above
arise due to the execution of a machine learning algorithm on private data in an
untrusted environment. Trusted Execution Environments~(TEE)~like Intel
SGX~\citep{McKeen2016}, ARM TrustZone~\citep{Alves2004}, and
Sanctum~\citep{Costan2016} present a solution to this problem by providing a
secure environment to run  the code. TEEs are a secure area within the main
processor of a computer. They are an isolated environment that runs in parallel
with the operating system  and guarantees that code and data loaded inside is
protected with respect to integrity and confidentiality. TEEs use hardware and
software protections to guarantee that any code and data stored in that secure
enclave can be accessed only by code in that enclave irrespective of privileges
in the software stack. Therefore, even the OS or the hypervisor cannot access
the information stored in the enclave.

However, there are significant computational and security challenges associated
with this approach. First, the secure enclave is usually extremely limited in
memory and computation. Large models cannot be loaded on it due to memory
limitations and computations using large model require significant paging, which
introduces computational overheads. Second, these enclaves do not have a large
number of parallel threads, which further slows down the computation. Finally, the
enclave is maintained by a {\em trusted} third party. If the third party is
dishonest, the third party can use its admin privileges on the system to mount
side-channel attacks on the enclave~(see Section 6 in~\citet{Hunt2018}).

\paragraph{Multi-Party Computation~(MPC)} Another way of approaching the problem
is by designing a protocol that itself emulates the secure third party, which
collects the data from multiple parties, evaluates the function, and returns the
result to all~(or a specific set of) parties. This allows two or more parties to
evaluate a function without disclosing their data to one another or anyone else.
Garbled Circuits~\citep{Yao1986} and the GMW protocol~\citep{Goldreich1987} are
examples of protocols that aim to solve this problem. They look at the following
scenario. There are $n$ players each having a piece of secret information
$\bc{x_1, \ldots, x_n}$ and there is a boolean function $g$ that needs to be
evaluated on these $n$ inputs. However, the function needs to be evaluated
without any of the players learning anything more from the process than they
would have learned from just observing the output $g\br{x_1,\ldots, x_n}$.

While this seems to provide the required security guarantees, there are multiple
issues associated with this technique especially if the protocol is used in the
Prediction As A Service framework. First, the protocol assumes a level of honesty
among the players which might not hold in the real world. In particular, it
requires that the players stick to the given protocol. Ideally, we would like a
protocol that protects against a malicious adversary. Second, MPC requires
multiple rounds of communication between the players and it requires all the
parties to execute some components of the computation. In the Prediction As A
Service framework, customers might have low capacity devices and be unable to
execute complex computations on their device, and depending on their internet
access, multiple rounds of communication can be prohibitively expensive or
time-consuming. Ideally, we would like a protocol where the customer operates
once on the data at the beginning and sends it to the service provider and then
receives the output. Third, all parties in the protocol need to have access to
the function being evaluated. Thus, if one o the players is the service
provider and the other is the user, it requires the user to have access to
the~(possibly encrypted) machine learning model on their private device so that
they can  do some of the evaluation on their private device. This is often
undesirable as service providers would be unwilling to share their models with
customers. In addition, the service providers would have to share a new model
every time they update their model.

\paragraph{Homomorphic Encryption}  While TEEs and MPC schemes provide some
level of privacy and reliability in using  Prediction As A Service, these
approaches protect against weaker threat models than we would like. TEEs do not
provide cryptographic privacy guarantees in the execution environment, thus
their computations are vulnerable to side-channel attacks and rely on the third
party being honest. MPC schemes also rely on partial honesty among the parties
and require multiple rounds of communication between the service provider and
the user. Further, MPCs require the function that is going to be evaluated to
be shared between the different users. Ideally, we would like to develop a
protocol that can overcome all of these disadvantages.

One particular way to achieve this is if the data provided by the customer is
cryptographically hidden from the machine learning model while still enabling
the model to make accurate, albeit encrypted, predictions. This protection is
exactly what "Encrypted Prediction As A Service~(EPAAS)" defined by us
in~\Cref{chap:TAPAS} provides. The basic service required by EPAAS is that the
service provider has access to a learned machine learning model in plaintext.
The customer encrypts the private personal data  and sends the encrypted data to
the service provider, along with the public key but not the private key. The
service provider computes an encrypted prediction on the received data using
their model and sends the encrypted prediction back to the customer, who then
decrypts it. The framework of {\em Fully Homomorphic Encryption (FHE)} is ideal
for this paradigm. {\em Homomorphic encryption}, first proposed
by~\citet{RAD:1978}, is an encryption methodology that allows certain operations
on it without decrypting it first. ~\citet{Gen:2009} proposed the first FHE 
scheme that allows performing arbitrarily
many operations on the encrypted data.  Since then several other schemes have
been proposed~\citep{GHS:2012,GSW:2013,BV:2014a,DM:2015,CGGI:2016}. 

\subsection{Challenges in using Homomorphic Encryption}
While EPAAS using homomorphic encryption schemes offer powerful privacy
protections, the major challenge associated with using homomorphic encryption is
its computational inefficiency. Without significant changes to the machine
learning model and improved algorithmic tools, homomorphic encryption does not
scale to modern deep neural networks.

Indeed, already there have been several recent works trying to accelerate
predictions of machine learning models on fully homomorphically encrypted data.
In general, the approach has been to approximate all or parts of a machine
learning model to accommodate the restrictions of an FHE framework. Often,
certain kind of FHE schemes
is preferred because they allow for ``batched'' parallel encrypted computations,
called SIMD operations~\citep{SV:2014}. This technique is exemplified by the
CryptoNets model %
\citep{G-BDL+:2016}. While these models allow for high-throughput (via SIMD),
they are not particularly suited for the Prediction As A Service framework for
individual users, as single predictions are slow. Further, because they employ a
leveled homomorphic encryption scheme, they are unable to perform many nested
multiplications, a requirement for state-of-the-art deep learning models~\citep{HZRS:2016,HZWV:2017}.

In~\Cref{chap:TAPAS}, we will look at a novel solution that demonstrates how
existing work on Binary Neural Networks (BNNs) \citep{KS:2015,CHSEB:2016} can be
adapted to produce efficient and highly accurate predictions on encrypted data.
We show that a recent %
FHE encryption scheme~\citep{CGGI:2016} which only supports operations on binary
data can be leveraged to compute all of the operations of BNNs. To do so, we
develop specialised circuits for fully-connected, convolutional, and batch
normalization layers~\citep{IS:2015}. Additionally, we design tricks to sparsify
encrypted computation that reduces computation time even further. We lay down
some important computational and privacy criteria that need to be satisfied by
an EPAAS framework and we discuss why most recent approaches fail them. Then
we discuss various types of Homomorphic Encryption schemes and why our
particular choice of encryption scheme is suitable for our method.
\clearpage
\chapter{Improving Generalization via Stable Rank Normalization}
\label{chap:stable_rank_main}

In~\Cref{sec:clt,sec:gen_nn}, we discussed the importance of complexity measures
and regularisation techniques for certifying generalisation in machine learning
models. This chapter looks at how one particular
complexity measure derived from the study of generalisation in neural networks
can be explicitly penalised, as a regulariser, to improve generalisation in
practice.
We leverage recent results on the generalisation of deep networks to yield a
practical low-cost method to normalize the weights within a network using a
scheme, we call \gls{srn}.
\begin{remark}[Normalization And Regularisation]
Both {\em Normalization} and {\em regularisation} refer to techniques for
controlling the complexity of neural networks. The main difference between the
two is that while {\em regularisation} refers to a Tikhonov style regularisation
and softly penalises the complexity, {\em normalization} usually implements an
Ivanov style regularisation and guarantees that a hard constraint on the
complexity of the network is satisfied.

In addition, normalization techniques for neural networks also ensure that the
desired property is satisfied throughout training. Hence, by changing the
loss landscape~\citep{Santurkar2018}, normalization techniques also have
a significant impact on optimisation. Some commonly used normalization
techniques are Batch normalization~\citep{Ioffe2015} that controls the mean and
variance of the activations in one batch of examples, Spectral
Normalization~\citep{miyato2018spectral} that controls the spectral norms of
individual layers, and Weight  Normalization~\citep{Salimans2016} that controls
the euclidean norms of the weight parameters.
\end{remark}

 The motivation for \Gls{srn} comes from the generalisation bound for \glspl{nn}
given by~\citet{neyshabur2018a} and \citet{bartlett2017spectrally} and discussed
previously in~\Cref{thm:spec-norm-marg-main}~(in particular its variant
discussed in~\Cref{eq:stable-spec-compl}). Recall that the excess error due to
~\Cref{eq:stable-spec-compl} scales as~$
\bigO{\sqrt{\prod_{i}^L\norm{\vec{W}_i}_2^2\sum_{i=1}^L \srank{\vec{W}_i}}} $
where $L$ and $\norm{\vec{W}}_2$ represents the number of layers and the
spectral norm of the $i$-th linear layer $\vec{W}_i$, respectively. It depends
on two parameter-dependent quantities: 
\begin{enumerate}
\item The product of scale-dependent spectral norms of individual layers
$\prod_{i}^L\norm{\vec{W}}_2$, also referred to as the \Gls{lip} constant
upper-bound and 
\item The sum of scale-independent {\em stable ranks} of each layer \(\sum_{i=1}^L\srank{\vec{W}_i}\). Stable rank
is a softer version of the rank operator and is defined as the squared ratio of
the Frobenius norm to the spectral norm.
\end{enumerate} 

We refer to the spectral norm as a scale-dependent term as, like all norms, the
spectral norm of a matrix is {\em absolutely homogenous} and increases
proportionally with the scaling of the matrix. On the other hand, stable rank is
a ratio of two norms, and thus upon multiplying the matrix with a scalar, the
scaling of the two individual norms cancels each other. Like the rank of a
matrix, stable rank is thus insensitive to the scaling of a matrix.

It is common to regularise a  complexity bound obtained from generalisation
bounds to observe better empirical generalisation in practice.
Some examples, relevant for neural networks, include weight decay for $L_2$
regularisation, Path-SGD~\citep{Neyshabur2015,Zheng2018} for Path-norm
regularisation, and margin-jacobian regularisation~\citep{wei2019} for
controlling a data-dependent generalisation bound proposed in~\citet{wei2019}.
However, the empirical impact of simultaneously controlling both, the spectral
norm and the stable rank, on the generalisation behaviour of \gls{nn}s has not
been explored yet possibly because of the difficulties associated with
optimizing stable rank. This is precisely the goal of this chapter. Based on
extensive experiments across a wide variety of \gls{nn} architectures, we
observe that controlling them simultaneously indeed improves the generalisation
behaviour of \gls{nn}s. We also observe improved training of Generative
Adversarial Networks~(\glspl{gan})~\cite{goodfellow2014generative} with our
technique. 

To this end, we propose \acrfull{srn} which allows us to simultaneously control
the \Gls{lip} constant and the stable rank of a linear operator. Note that the
widely used \gls{sn}~\citep{miyato2018spectral} allows explicit control over the
\Gls{lip} constant, however, as we discuss later in this chapter, it is neither
the optimal solution to the spectral normalization problem and neither does it
have any impact on the stable rank. Unlike \gls{sn},~the \gls{srn} solution, to
controlling the stable rank, is optimal and unique despite being a non-convex
problem. It is one of those rare cases where an optimal solution to a provably
non-convex problem can be easily computed. Computationally, the proposed
\gls{srn} technique for \gls{nn}s is no more complicated than \gls{sn}, just
requiring the computation of the largest singular value, which can be done
efficiently using the power iteration method~\citep{Mises1929}.

\section{Main contributions}
Before going into a detailed discussion of the problem, in this section, we
briefly state the main contributions of this chapter. We divide the list of
contributions into two sections- a) theoretical and algorithmic contributions
and b) Experimental Results. 

\paragraph{Theoretical and Algorithmic Contributions}
\begin{enumerate}
	\item We propose Stable Rank Normalization~(\gls{srn})--- a novel normalization scheme for simultaneously controlling the \Gls{lip} constant and the stable rank of a linear operator.
	\item We also present a unique and optimal solution to the provably
	non-convex stable rank normalization problem.
	\item Finally, we devise and efficient and  easy algorithm to implement
	\gls{srn} for \gls{nn}s.
\end{enumerate}

\paragraph{Experimental Results} Although \gls{srn} is in principle applicable
to any  problem involving a sequence of affine transformations, considering
recent interests, we show its effectiveness when applied to the linear layers of
deep neural networks. We also experiment with \gls{gan}s and show that
\gls{srn} prefers learning discriminators with low empirical \Gls{lip} while
providing improved Inception, FID and Neural Divergence
scores~\citep{gulrajani2018towards}. 

\begin{enumerate}
\item We perform extensive experiments on a wide variety of \gls{nn}
architectures (DenseNet, WideResNet, ResNet, Alexnet, and VGG) for the analyses
and show that \gls{srn} improves classification accuracy on a wide variety of
architectures.
\item While providing the best classification accuracy (compared against
standard training, vanilla training, and \gls{sn} training), neural networks
trained with~\Gls{srn} shows remarkably less memorisation, even on settings that
are known to be hard to generalise in.
\item Further, networks trained with \Gls{srn} show much smaller sample
complexity measured using  complexity measures proposed in recent works.
\item Applying SRN to Generative Adversarial
Networks~(GANs)~\citep{goodfellow2014generative} improves the  performance of
GANs measured via various metrics and makes them more resilient to memorisation.
\end{enumerate}

We also note that although \gls{sn} is widely used for training \gls{gan}s, its
effect on the generalisation behaviour over a wide variety of multi-class
classification neural networks has not yet been explored. To the best of our
knowledge, we are the first to do so.

\section{Definitions and preliminaries}
\label{sec:background}
\paragraph{Neural Networks} Consider a neural network $f_{\theta}: \reals^d
\rightarrow \reals^k$ parameterised by $\theta \in \reals^p$, each layer of
which consists of a linear mapping followed by a non-linear\footnote{\eg ReLU,
tanh, sigmoid, and maxout.} activation function. Recall that it is defined as
follows in~\Cref{defn:net-mlp}.
\mlpnet*

For classification tasks, given a dataset with input-output pairs denoted as
$(\vec{x} \in \reals^d, \vec{y} \in \{0,1\}^k; \sum_j y_j = 1)$ \footnote{$y_j$
is the $j$-th element of vector $\vec{y}$. Only one class is assigned as the
ground-truth label to each $\vec{x}$.}, the parameter vector $\theta$ is learned
using back-propagation to optimise the classification loss ({\em e.g.},
cross-entropy). 
\paragraph{\gls{svd}} Singular Value Decomposition~(SVD) is a factorisation of a
matrix that generalises the eigendecomposition of square matrices to any rectangular
matrix. Given $\vec{W} \in \reals^{s \times r}$ with rank $k \leq \min (s,r)$,
we denote $\{\sigma_i\}_{i=1}^k$, $\{\vec{u}_i\}_{i=1}^k$, and
$\{\vec{v}_i\}_{i=1}^k$ as its singular values, left singular vectors, and right
singular vectors, respectively. Throughout this thesis, a set of singular values
is assumed to be sorted $\sigma_1 \geq \cdots \geq \sigma_k$.
$\sigma_i(\vec{W})$ denotes the $i$-th singular value of the matrix $\vec{W}$.
Using singular values, the matrix 2-norm $\norm{\vec{W}}_2$ and the Frobenius
norm $\norm{\vec{W}}_\forb$ can be computed as $\sigma_1$ and $\sqrt{\sum_i
\sigma_i^2}$, respectively. We discuss this further, along with other linear
algebra basics, in~\Cref{sec:math-prer-notat}.
\subsection{Stable Rank} Below we provide the formal definition and
some properties of stable rank. %
\begin{defn}[Stable Rank]
\label{def:stableRank}
The Stable Rank~\citep{Rudelson2007} of an arbitrary matrix $\vec{W}$ is
defined as \( \srank{\vec{W}} =
\frac{\norm{\vec{W}}_\forb^2}{\norm{\vec{W}}_2^2} = \frac{\sum_{i=1}^k
  \sigma_i^2(\vec{W})}{\sigma_1^2(\vec{W})} \), where $k$ is the rank of
the matrix. Stable rank is
\begin{enumerate}
\item  a soft version of the rank operator and, unlike rank, is less sensitive to small perturbations. %
\item almost always differentiable as both Frobenius and Spectral norms are almost always differentiable.
\item upper-bounded by the rank of the matrix: \( \srank{\vec{W}} =
\frac{\sum_{i=1}^k \sigma_i^2(\vec{W})}{\sigma_1^2(\vec{W})} \le
\frac{\sum_{i=1}^k \sigma_1^2(\vec{W})}{\sigma_1^2(\vec{W})}  = k\).
\item invariant to scaling, i.e. for any $\eta \in \reals\setminus\bc{0}$ we
have that \( \srank{\vec{W}} = \srank{\frac{\vec{W}}{\eta}} \).
\end{enumerate}
\end{defn}

\subsection{\Gls{lip} Constant}\label{sec:lipschitz} Lipschitzness of a function
is an indication of the smoothness of the function. The \Gls{lip} constant of a
function is a quantification of the sensitivity of the output of the function
with respect to the change in the input. Thus, functions with small lipschitz
constants are smoother than functions with large lipschitz constants. This
section first describes the concepts of global~\gls{lip}, local \gls{lip}, and
empirical lipschitz constants and then discusses why the product of spectral
norms is an upper bound on the local lipschitzness of neural networks.

\paragraph{Global and Local Lipschitzness}
 A function $f: \reals^d \mapsto \reals^k$ is {\em globally L-\Gls{lip}
continuous} if there exists \( L \in \reals_+\) such that \(\norm{f(\vec{x}_i) -
f(\vec{x}_j)}_q \leq L \norm{\vec{x}_i - \vec{x}_j}_p\) for all \( (\vec{x}_i,
\vec{x}_j) \in \reals^d \), where $\norm{\cdot}_p$ and $\norm{\cdot}_q$
represents the $\ell_p$ and $\ell_q$ norms in the input and the output metric spaces, respectively.
The global \Gls{lip} constant $L_g$ is:
\begin{align}
\label{eq:lipGlobal}
L_g = \max_{\substack{\vec{x}_i, \vec{x}_j \in \reals^d\\ \vec{x}_i\neq\vec{x}_j}} \frac{\norm{f(\vec{x}_i) - f(\vec{x}_j)}_q}{\norm{\vec{x}_i - \vec{x}_j}_p}.
\end{align}
The above definition of the \Gls{lip} constant accounts for all  pairs of inputs
in the domain $\reals^d\times\reals^d$. It is thus said to be the {\em global
\gls{lip} constant}. One can define the local \Gls{lip} constant based on the
sensitivity of $f$ in the vicinity of a given point $\vec{x}$. 

For a given $\vec{x}$ and for an arbitrarily small $\delta>0$, the local
\Gls{lip} constant is computed in an open ball of radius $\delta$ centred at
$\vec{x}$. Let $\vec{h} \in \reals^d$ with $\norm{\vec{h}}_p < \delta$, then,
similar to $L_g$, the {\em local \gls{lip} constant} of $f$ at $\vec{x}$,
$L_l(\vec{x})$, is greater than or equal to 
\begin{equation}
  \sup_{\vec{h} \neq 0,
\norm{\vec{h}}_p < \delta}~\frac{\norm{f(\vec{x} + \vec{h}) -
f(\vec{x})}_q}{\norm{\vec{h}}_p}
\end{equation}

Assuming $f$ to be Fr\'echet differentiable, as $\vec{h}$ tends to $0$, we can
use the first-order approximation on $f$: $f(\vec{x} + \vec{h}) - f(\vec{x})
\approx J_f(\vec{x}) \vec{h}$, where \(J_f\br{\vec{x}}=\frac{\partial
f\br{\vec{z}}}{\partial \vec{z}} \vert_{\vec{x}} \in \reals^{k \times d}\) is
the jacobian of $f$ at \(\vec{x}\). Then the local lipschitz constant of $f$ at
\(\vec{z}\) is the matrix (operator) norm of the Jacobian $J_f(\vec{x})$.

\begin{align}
\label{eq:lipLocal}
 L_l(\vec{x}) \stackrel{(a)}{=} \lim_{\delta\rightarrow 0}\sup_{\substack{\vec{h} \neq 0 \\ \norm{\vec{h}}_p < \delta}} \frac{\norm{J_f(\vec{x}) \vec{h}}_q}{\norm{\vec{h}}_p} \stackrel{(b)}{=}  \sup_{ \substack{\vec{h} \neq 0 \\ \vec{h} \in \reals^m}} \frac{\norm{J_f(\vec{x}) \vec{h}}_q}{\norm{\vec{h}}_p} = \norm{J_f(\vec{x})}_{p,q}^{\mathrm{op}}.
\end{align}
Here, (a) is by definition of local lipschitzness and (b) is due to the property
of  norms that for any non-negative scalar~$c$, $\norm{c\vec{x}} =
c\norm{\vec{x}}$. Note that \(\norm{J_f(\vec{x})}_{p,q}^{\mathrm{op}}\) is the \(p,q\) matrix operator norm, defined in~\cref{defn:p_q-induced_norm}, of the Jacobian matrix and is different from the entry-wise \(p,q\) norm, which is for example used in~\cref{eq:defn-ins-spec-compl}. A function is said to be \emph{locally \Gls{lip}} with
\emph{local Lipschitz constant} $L_l$ if, for all $\vec{x} \in \reals^d$, the
function is  \emph{$L_l$ locally-\Gls{lip}} at $\vec{x}$. %
 Thus, \begin{equation}\label{eq:localLipschitz}
  L_l  =\sup_{\vec{x}\in\reals^d}{L_l\br{\vec{x}}}
 \end{equation} 
 
Notice that the \Gls{lip} constant (global or local) greatly depends on the
chosen norms. When $p = q = 2$, the upper bound on the local \Gls{lip} constant
at $\vec{x}$ boils down to the 2-matrix norm (maximum singular value) of the
Jacobian $J_f(\vec{x})$~(see last equality of~\cref{eq:lipLocal}). 

\paragraph{The local \Gls{lip} upper-bound for Neural Networks}
\label{sec:LipNNUB}
Here we show that the local lipschitz constant for neural networks can be upper bounded by the product of spectral norms of individual weight matrices. Interestingly this upper-bound is data-agnostic.
\begin{restatable}[Local Lipschitz upper bound for NNs]{lem}{locallipNN}
  \label{lem:upper-bound-nn-lip}
  For a neural network $f$ belonging to the class of multi-layered perceptrons, as defined in~\Cref{defn:net-mlp}, the local lipschitzness at a point $\vec{x}\in\reals^d$ can be upper-bounded as 
  \begin{align}
    \label{eq:lipboundNN1}
    L_l(\vec{x}) = \leq \norm{\vec{W}_1}_{p,q}^{\mathrm{op}}
  \cdots \norm{\vec{W}_L}_{p,q}^{\mathrm{op}}
\end{align}
Further, as the upper-bound is data-agnostic, the local lipschitz constant is also equal to this upper-bound i.e. \[L_l = L_l\br{\vec{x}}\]

Proof in~\Cref{sec:lipsch-proof}.
\end{restatable}

Next we discuss more optimistic~(or empirical) estimates of $L_l$ and $L_g$, its
link with generalisation and then in~\cref{sec:experiments-srn}, we show
empirically the effect of SRN on empirical lipschitzness and generalisation.

\paragraph{Empirical Lipschitz constants}
It is clear from~\Cref{lem:upper-bound-nn-lip} that the \Gls{lip} constant upper
bound (\(\prod_{i}^L\norm{\vec{W}_i}_2\)), along with being scale-dependent, is
also {\em data-independent} and hence, provides a pessimistic estimate of the
behaviour of a  model on a particular task or dataset. We call this pessimistic
as the behaviour of the model on the entire input domain is not always relevant
especially when data lies in a more restricted portion of the domain.
Considering this, a relatively optimistic estimate of the model's behaviour
would be an {\em empirical} estimate of the \Gls{lip} constant ($L_e$) on a
task-specific dataset. The global and local $L_e$ are simply the equivalent of
the global and local lipschitz constant defined in~\cref{eq:lipGlobal}
and~\cref{eq:localLipschitz} respectively  with the maximisation done on just
the support of the data distribution as opposed to the whole of \(\reals^d\).
Note that local $L_e$ is just the norm of the Jacobian at a given point. For
completeness, we provide the relationship between the global and the local $L_e$
in~\Cref{prop:empiricalLocalLip}.

\begin{restatable}[Relating Empirical and Global Lipschitzness]{lem}{lipschitznesslocalglobal}
\label{prop:empiricalLocalLip}
Let $f: \reals^d \mapsto \reals$ be a Fr\'echet differentiable function,
$\data$ the dataset, and $\textit{Conv}\;(\vec{x}_i, \vec{x}_j)$ denotes the
convex combination of a pair of samples $\vec{x}_i$ and $\vec{x}_j$, then
$\forall p,q \in [1, \infty]$ such that $\frac{1}{p}+\frac{1}{q} = 1$
\begin{align}
\max_{\vec{x}_i, \vec{x}_j \in \data}\frac{\abs{f(\vec{x}_i) - f(\vec{x}_j)}}{\norm{\vec{x}_i - \vec{x}_j}_p} \; \leq \max_{\substack{{\vec{x}_i, \vec{x}_j \in \data} \\ \vec{x} \in \textit{Conv}\;(\vec{x}_i, \vec{x}_j) }} \norm{J_f(\vec{x})}_q \nonumber
\end{align}
Proof in~\Cref{sec:lipsch-proof}.
\end{restatable}

As discussed before, the local lipschitz constant upper bound
in~\Cref{eq:lipboundNN1} is independent of $\vec{x}$. This is one of the
main reasons why we consider the empirical \Gls{lip} to better reflect the true
behaviour of the function as the \gls{nn} is never exposed to the entire domain
$\reals^d$ but only a small subset dependent on the data distribution.
The other reason why this upper bound is a bad estimate is that the
inequality in Eq~\eqref{eq:lipboundNN1} is tight only when the partial
derivatives are aligned, implying, $ \norm{\frac{\partial \vec{z}_\ell}{\partial
  \vec{z}_{\ell-1}}  \frac{\partial \vec{z}_{\ell+1}}{\partial
  \vec{z}_{\ell}}}_2 =  \norm{\frac{\partial \vec{z}_\ell}{\partial
  \vec{z}_{\ell-1}}}_2  \norm{\frac{\partial \vec{z}_{\ell+1}}{\partial
  \vec{z}_{\ell}}}_2~\quad \forall l -  2\le \ell\le l$. This problem has been referred to as the problem of
mis-alignment and is similar to  quantities like layer cushion in~\citet{arora18b}.

\section{Stable Rank Normalization}
\label{sec:stable_rank_alg}

This section describes a technique, we call Stable Rank
Normalization~(\gls{srn}), to control the stable rank of linear operators. A big
challenge in stable rank normalization comes from the fact that stable rank is
scale-invariant~(refer to~\cref{def:stableRank}), thus, any normalization scheme
that modifies \(\vec{W} = \sum_i \sigma_i \vec{u}_i \vec{v}_i^{\top}\) to
$\widehat{\vec{W}} = \sum_i \frac{\sigma_i}{\eta} \vec{u}_i
\vec{v}_i^{\top}$~(for any \(\eta>0\) will not affect on the stable rank.
Examples of such schemes are \gls{sn}~\citep{miyato2018spectral} where $\eta =
\sigma_1$, and Frobenius normalization where $\eta = \norm{\vec{W}}_\forb$. But
first, we look at some conceptual motivation for controlling stable rank.

\subsection{Impact of stable rank on noise-sensitivity}
\label{sec:whyStable}

\paragraph{Stable rank controls the noise-sensitivity}
As shown by~\citet{arora18b}, one of the critical properties of generalisable \gls{nn}s 
is low noise sensitivity--- the ability of a network to
preferentially carry over the true signal in the data. For a given  
noise distribution $\cN$, it can be quantified as 
\begin{equation}\label{eq:stable-rank-noise-sensitivity}
  \Phi_{f_{\theta},\cN} = \max_{\vec{x} \in \data}\Phi_{f_{\theta},\cN}\br{\vec{x}}, \quad \textit{where} \quad \Phi_{f_{\theta},\cN}\br{\vec{x}} :=
  \bE_{\eta\sim\cN} \bs{\dfrac{\norm{f_\theta\br{\vec{x} +
        \eta\norm{\vec{x}}} -
      f_\theta\br{\vec{x}}}^2}{\norm{f_\theta\br{\vec{x}}}^2}} 
\end{equation}

  \begin{figure}
  \centering
  \def\svgwidth{0.2\columnwidth}
  \resizebox{0.3\linewidth}{!}{
  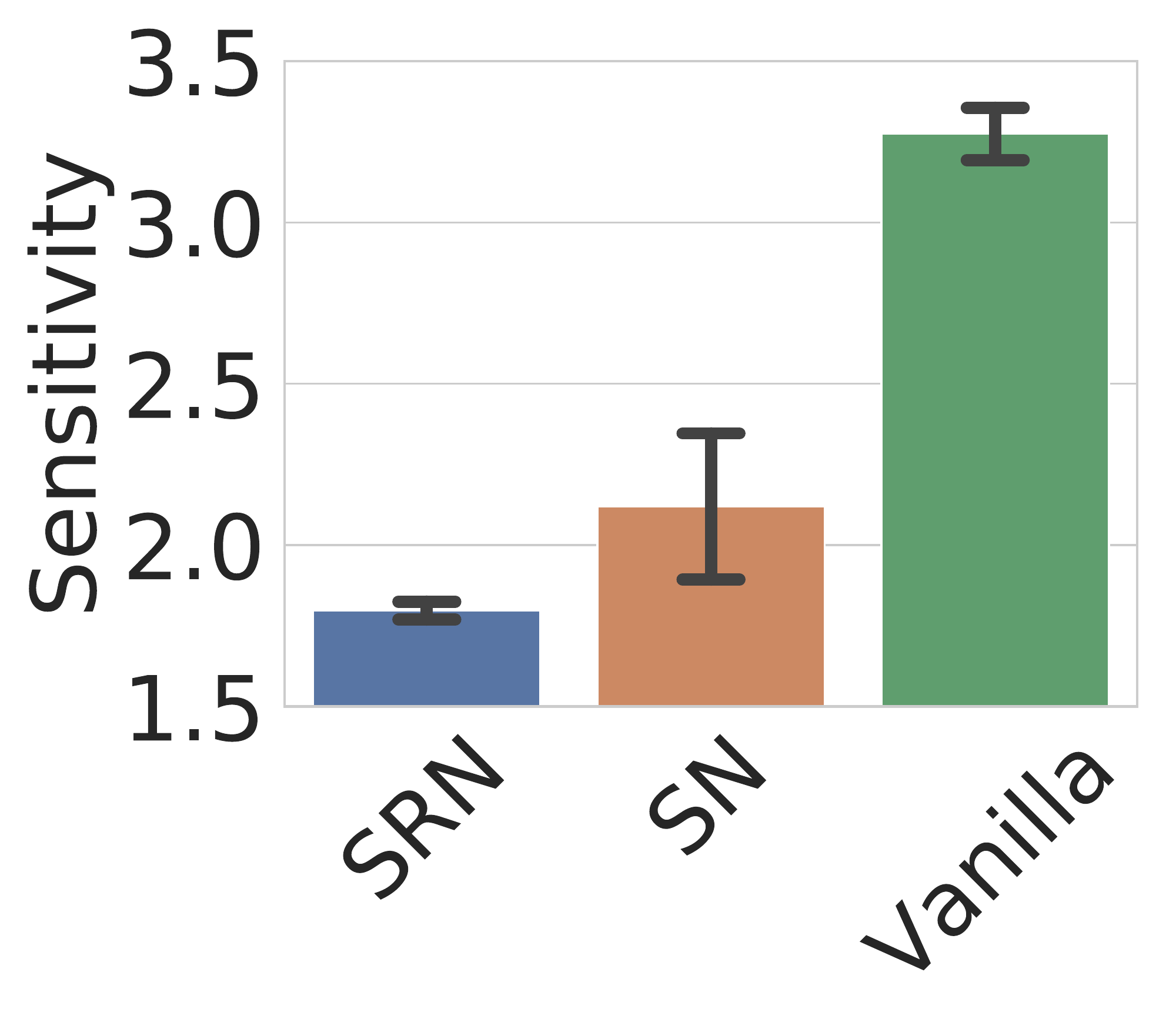
  } 
  \caption[Noise Sensitivity of SRN, SN, and Vanilla NN]{Noise Sensitivity (lower the better). Test accuracy: SRN ($73.1\%$), SN ($71.5\%$), and Vanilla ($72.4\%$).
  }\label{fig:noise_sensit}
  \end{figure}
      For a linear mapping with parameters $\vec{W}$ and the noise distribution
being normal-$\cN\br{0,\vec{I}}$, it can be shown that $\Phi_{f_\vec{W},\cN} \ge
\srank{\vec{W}}$~(see Proposition 3.1 in~\citet{arora18b}). Thus, decreasing the
stable rank decreases this lower bound on noise sensitivity.
In~\cref{fig:noise_sensit}, we show $\Phi_{f_\theta, \cN}$ of a ResNet110
trained on CIFAR100. Note that although the \Gls{lip} upper
bound~(\Cref{lem:upper-bound-nn-lip}) of \gls{srn} and \gls{sn} are the same,
\gls{srn} (algorithmic details in~\cref{sec:stable_rank_alg}) is much less
sensitive to noise than SN. 
This can be possibly explained by~\Gls{srn}'s impact on the
empirical~\gls{lip} constant, which we discuss below.

\paragraph{Stable rank impacts empirical \gls{lip} constant}
Empirically,~\citet{novak2018sensitivity} provided results showing how local
empirical lipschitz $L_e$ (in the vicinity of train data) is correlated with the
generalisation error of neural networks. This observation is further supported
by the theoretical works of~\citet{wei2019,nagarajan2018deterministic},
and~\citet{arora18b} whereby variants of $L_e$ are used to derive generalisation
bounds for neural networks. Thus, a tool that favours low $L_e$ is likely to
provide better generalisation behaviour in practice. A low local empirical
lipschitz also decreases the noise-sensitivity of neural
networks~(c.f.~\Cref{eq:stable-rank-noise-sensitivity}).

To this end, we first consider a simple two layer linear neural network and show
that low rank transformations favour low $L_e$. A linear neural network is a
neural network with linear activation functions. Since direct minimisation of
rank for \gls{nn}s is non-trivial, the expectation is that learning weight
matrices with low stable rank (softer version of rank) might induce similar
behaviour. We experimentally validate this hypothesis by showing that, as we
decrease the stable rank, the empirical \Gls{lip} decreases. This shows
\gls{srn} indeed prefers learning transformations with low empirical \Gls{lip}
constant.

Let $f(\vec{x}) = \vec{W}_2 \vec{W}_1 \vec{x}$ be a two-layer linear neural
network with weights $\vec{W}_1$ and $\vec{W}_2$. The Jacobian, in this case, is
independent of $\vec{x}$. Thus, the local \Gls{lip} constant is the same for all
$\vec{x} \in \reals^d$, implying, local $L_e = L_l(\vec{x}) = L_l =
\norm{\vec{W}_2 \vec{W}_1} \leq \norm{\vec{W}_2} \norm{\vec{W}_1}$. 
Note, in the case of the 2-matrix norm, reducing the rank of \(\vec{W_1}\) and
\(\vec{W}_2\) does not affect the upper bound
\(\norm{\vec{W}_1}_2\norm{\vec{W}_2}_2\). However, we discuss below that rank
reduction influences the global $L_e$.

Let $\vec{x}_i$ and $\vec{x}_j$ be a random pair from the dataset $\data$ and
$\Delta \vec{x} \neq \bf0$ be the difference  $\vec{x}_i - \vec{x}_j$. Then, the
global $L_e$ is \( \max_{\{\vec{x}_i, \vec{x}_j\} \in \data}
\frac{\norm{\vec{W}_2 \vec{W}_1 \Delta \vec{x}}}{\norm{\Delta \vec{x}}} \).
Let $k_1$ and $k_2$ be the ranks, and $\sigma_1 \geq \cdots \geq \sigma_{k_1}$
and $\lambda_1 \geq \cdots \geq \lambda_{k_2}$ be the singular values of the
matrices $\vec{W}_1$ and $\vec{W}_2$, respectively. Let $P_i = \vec{u}_i
\bar{\vec{u}}_i^\top$ be the orthogonal projection matrix corresponding to
$\vec{u}_i$ and $\bar{\vec{u}}_i$, the left and the right singular vectors of
$\vec{W}_1$. Similarly, we define $Q_i$ for $\vec{W}_2$ corresponding to
$\vec{v}_i$ and $\bar{\vec{v}}_i$. Then, the product of the two matrices can be
decomposed into $\vec{W}_2 \vec{W}_1  = \sum_{i=1}^{k_2} \sum_{j = 1}^{k_1}
\lambda_i \sigma_j Q_i P_j $. The largest singular value of the product is equal
to its maximum possible value $\lambda_1 \sigma_1$ if and only if
 $\Delta \vec{x} = \bar{\vec{u}}_1 \norm{\Delta \vec{x}}$ and $\vec{u}_1 =
\bar{\vec{v}}_1$ , i.e. a perfect alignment of the top most singular vectors of
\(\vec{W}_1\) and \(\vec{W}_2\) occurs, which is highly unlikely. In practice,
not just the maximum singular values, unlike the case with the \Gls{lip}
upper-bound in~\Cref{lem:upper-bound-nn-lip}, rather the combination of the
projection matrices and the singular values play a crucial role in providing an
estimate of global $L_e$. 
Thus, zeroing out certain singular values, which is equivalent to minimizing the
rank (or stable rank), reduces $L_e$. For example, assigning $\sigma_j = 0$,
which in effect reduces the rank of $\vec{W}_1$ by one, nullifies the
contribution of all projections of \(\vec{W}_2\) on $P_j$. Thus, all $k_2$
projections $\sigma_j (\sum_{i=1}^{k_2} \lambda_i Q_i) P_j$ that would have
propagated the input via $P_j$ are blocked. This influences $\norm{\vec{W}_2
\vec{W}_2 \Delta  \vec{x}}$ and hence, the global $L_e$. In a more general
setting, if $k_i$ is the rank of the $i^{\it th}$ linear layer, then, each singular
value of the $j$-th layer can influence a maximum of $\prod_{i=1}^{j-1} k_i
\prod_{i=j+1}^{l} k_i$ many paths through which an input can propagate. Thus,
transformations with low rank (stable) reduce the global $L_e$. Similar
arguments can be drawn for local $L_e$ in the case of \gls{nn} with
non-linearity.

\subsection{Solution to the Stable Rank Normalization problem}
\label{sec:optimalSrank}

In this section, we will define and solve the stable rank normalization problem.

\paragraph{The \gls{srn} problem statement} Given a matrix $\vec{W}
\in \reals^{m \times n}$ with rank $p$ and a spectral partitioning index
$k$ ($0 \leq k < p$), we formulate the SRN problem as
\begin{align}
\label{eq:srankProblem}
& \argmin_{\widehat{\vec{W}}_k\in\reals^{m\times n}} \norm{\vec{W} - \widehat{\vec{W}}_k}_{\forb}^2 \quad \textit{s.t.} \quad \underbrace{\srank{\widehat{\vec{W}}_k} = r}_{\textrm{stable rank constraint}}, \; \underbrace{\lambda_i = \sigma_i, \forall i \in \{1, \cdots, k\}}_{\textrm{spectrum preservation constraints}}. 
\end{align}

where, $1 \leq r < \srank{\vec{W}}$ is the desired stable rank, $\lambda_i$s and
$\sigma_i$s are the singular values of $\widehat{\vec{W}}_k$ and $\vec{W}$,
respectively. The partitioning index $k$ is the {\em singular value (or the
spectrum) preservation constraint}. It gives us the flexibility to obtain
$\widehat{\vec{W}}_k$ such that its top $k$ singular values are the same
as that of the original matrix. Note that the problem statement is more general
in the sense that putting $k=0$ removes the spectrum preservation constraint
entirely. 

\paragraph{The solution to \gls{srn}} The optimal unique solution to the above
problem is provided in~\Cref{thm:srankOptimal} and proved
in~\cref{sec:srankProof}. At $k=0$, the problem~\eqref{eq:srankProblem} is
non-convex, otherwise convex.

\begin{restatable}[Solution to the Stable Rank Problem]{thm}{optimalthm}
\label{thm:srankOptimal}
Given a real matrix $\vec{W}\in\reals^{m\times n}$ with rank $p$, a target
spectrum (or singular value) preservation index $k$ $(0\le k < p)$, and a target
stable rank $r$ $(1 \leq r < \srank{\vec{W}})$, the optimal solution
$\widehat{\vec{W}}_k$ to the optimisation problem in~\cref{eq:srankProblem} is
$\widehat{\vec{W}}_k = \gamma_1\vec{S}_{1} + \gamma_{2} \vec{S}_{2}$, where
$\vec{S}_1 = \sum_{i=1}^{\mathrm{max}\br{1,k}} \sigma_i \vec{u}_i
\vec{v}_i^\top$, $\vec{S}_{2} = \vec{W} - \vec{S}_{1}$.
$\{\sigma_i\}_{i=1}^k$, $\{\vec{u}_i\}_{i=1}^k$ and $\{\vec{v}_i\}_{i=1}^k$ are
the top $k$ singular values and vectors of $\vec{W}$, and, depending on $k$,
$\gamma_1$ and $\gamma_2$ are defined below. For simplicity, we first define
$\gamma = \frac{\sqrt{r \sigma_1^2 -
\norm{\vec{S}_1}_\forb^2}}{\norm{\vec{S}_2}_\forb}$, then
\begin{enumerate}
\item[a)]If $k=0$ (no spectrum preservation), the problem is non-convex, the
  optimal solution to which is obtained for $\gamma_2 = \dfrac{\gamma + r -
  1}{r}$ and $\gamma_1 = \dfrac{\gamma_2}{\gamma}$, when $r>1$. If $r=1$, then
  $\gamma_2 = 0$ and $\gamma_1 = 1$. In particular as
  $\norm{\vec{S}_1}_{\forb}^2 = \sigma_1^2$, we get \( \gamma = \frac{\sqrt{r-1}
  \sigma_1}{\norm{\vec{S}_2}_\forb}\).
\item[b)] If $k\ge 1$, the problem is convex. If $r\ge
  \frac{\norm{\vec{S}_1}_\forb^2}{\sigma_1^2}$ the optimal solution is
  obtained for $\gamma_1 = 1$,  and $\gamma_2=\gamma$ and if not, the
  problem is not feasible.
\item [c)] Also, $\norm{\widehat{\vec{W}}_k - \vec{W}}_\forb$
monotonically increases with $k$ for $k \geq 1$.
\end{enumerate}

Proof in~\Cref{sec:srankProof}
\end{restatable}
Intuitively,~\Cref{thm:srankOptimal} partitions the given matrix into two parts,
depending on $k$, and then scales them differently to obtain the
optimal solution. The value of the partitioning index $k$ is a design choice. If
there is no particular preference for $k$, then $k=0$ provides the most optimal
solution in terms of closeness to the original matrix. We provide a simple
example to demonstrate this. Given $\vec{W} = \vec{I}_3$ (rank =
$\srank{\vec{W}}$ = 3), the objective is to project it to a new matrix with
stable rank of $2$. Consider the following three solutions to the problem.
$\widehat{\vec{W}}_1$ is obtained using the standard rank minimisation
(Eckart-Young-Mirsky~\citep{Eckart1936}) while $\widehat{\vec{W}}_2$ and
$\widehat{\vec{W}}_3$ are the solutions of~\Cref{thm:srankOptimal} with $k=1$
and $k=0$, respectively. 
\begin{align}\small
\label{eq:srnExample}
\widehat{\vec{W}}_1=
  \begin{bmatrix}
    1 & 0 & 0\\
    0 & 1 & 0\\
    0 & 0 & 0
  \end{bmatrix}
  , \;
  \widehat{\vec{W}}_2=
  \begin{bmatrix}
    1 & 0 & 0\\
    0 & \frac{1}{\sqrt{2}} & 0\\
    0 & 0 & \frac{1}{\sqrt{2}}
  \end{bmatrix}
  , \;
  \widehat{\vec{W}}_3=
  \begin{bmatrix}
    \frac{\sqrt{2}+1}{2} & 0 & 0\\
    0 & \frac{\sqrt{2}+1}{2\sqrt{2}} & 0\\
    0 & 0 & \frac{\sqrt{2}+1}{2\sqrt{2}}
  \end{bmatrix}
\end{align}

It is easy to verify that the stable rank of all the above solutions
is $2$. However, the Frobenius distance (lower the better) of these
solutions from the original matrix follows the order 
\(
\norm{\vec{W} - \widehat{\vec{W}}_1}_{\forb} > \norm{\vec{W} - \widehat{\vec{W}}_2}_{\forb} > \norm{\vec{W} - \widehat{\vec{W}}_3}_{\forb}
\). Thus,~\Cref{thm:srankOptimal} with $k=0$ provides the closest solution followed by $k=1$ and then the hard rank solution $\widehat{\vec{W}}_1$.
As evident from the example, the solution to \gls{srn}, instead of completely
removing a particular singular value like \(\widehat{\vec{W}_1}\), scales them
(depending on $k$) such that the new matrix has the desired stable rank. Note
that for $\widehat{\vec{W}}_1$ and $\widehat{\vec{W}}_2$ (true for any
$k\geq1$), the spectral norm of the original and the normalized matrices are the
same, implying, $\gamma_1 = 1$. However, for $k=0$, the spectral norm of the
optimal solution is greater than that of the original matrix. It is easy to
verify from~\Cref{thm:srankOptimal} that as $k$ increases, $\gamma_2$ decreases.
Thus, the amount of scaling required for the second partition $\vec{S}_2$ is
more aggressive. In all situations, the following inequality holds: $\gamma_2
\leq 1 \leq \gamma_1$.

\paragraph{Optimal Spectral Normalization}
\label{sec:spectralNormOptimal}
The widely used spectral normalization~\citep{miyato2018spectral} where the given matrix $\vec{W} \in \reals^{m \times n}$ is divided by the maximum singular value is an approximation to the optimal solution of the spectral normalization problem defined as
\begin{align}
\label{eq:spectralNormProb}
\argmin_{\widehat{\vec{W}}} & \norm{\vec{W} - \widehat{\vec{W}}}_{\forb}^2  \\
\textit{s.t.} \quad & \sigma(\widehat{\vec{W}}) \leq s, \nonumber
\end{align}
where $\sigma(\widehat{\vec{W}})$ denotes the maximum singular value and $s>0$ is a hyperparameter. The optimal solution to this problem is shown in~\cref{alg:spectralNormOptimal}.
\begin{algorithm}[t]
\caption{Spectral Normalization}
\label{alg:spectralNormOptimal}
\begin{algorithmic}[1]
\INPUT $\vec{W} \in \reals^{m \times n}$, $s$ 
\STATE $\vec{W}_1 \gets \mathbf{0}$, $p \gets \min(m,n)$
\FOR {$k \in \{1, \cdots, p\}$}
\STATE $\{\vec{u}_k, \vec{v}_k, \sigma_k\} \gets SVD(\vec{W}, k)$
\COMMENT{perform power method to get $k$-th singular value}
\IF{$\sigma_k \geq s$}
\STATE $\vec{W}_1 \gets \vec{W}_1 + s \; \vec{u}_k \vec{v}_k^{\top}$
\STATE $\vec{W} \gets \vec{W} - \sigma_k \; \vec{u}_k \vec{v}_k^{\top}$
\ELSE
\STATE break
\COMMENT{exit for loop}
\ENDIF
\ENDFOR
\\
\OUTPUT $\vec{W} \gets \vec{W}_1 + \vec{W}$
\end{algorithmic}
\end{algorithm}
\noindent
Here, we provide the proof of optimality of~\cref{alg:spectralNormOptimal} for
the sake of completeness. 

Let $\mathrm{SVD}\br{\vec{W}} =
\vec{U}\vec{\Sigma}\vec{V}^\top$ and let us assume that $\vec{Z} =
\vec{S}\Lambda\vec{T}^\top$ is a solution to the
problem~\ref{eq:spectralNormProb}. Trivially, $\vec{X} =
\vec{U}\Lambda\vec{V}^\top$ also satisfies $\sigma\br{\vec{X}}\le s$. Now,
$\norm{\vec{W}-\vec{X}}_{\forb}^2 = \norm{\vec{U}\br{\Sigma -
\Lambda}\vec{V}^\top}_{\forb}^2  = \norm{\br{\Sigma - \Lambda}}_{\forb}^2 \leq
\norm{\vec{W}-\vec{Z}}_{\forb}^2$, where the last inequality directly comes
from~\Cref{lem:ineq:frob_sing}. Thus the singular vectors of the optimal
solution must be the same as that of $\vec{W}$. This boils down to solving the
following problem
  \begin{equation}
\label{eq:spectralNormProb_vec}
\argmin_{\vec{\Lambda}\in\reals^{\mathrm{min}\br{m,n}}_{+}}  \norm{\vec{\Lambda} - \vec{\Sigma}}_{\forb}^2 \; \textit{s.t.} \;  \vec{\Lambda}\bs{i} \leq s\enskip \forall i\in\bc{0, {\mathrm{min}\br{m,n}}}.
  \end{equation} 

  Here, without loss of generality, we abuse notations by considering
  $\vec{\Lambda}$ and $\vec{\Sigma}$ to represent the diagonal vectors of the
  original diagonal matrices $\vec{\Lambda}$ and $\vec{\Sigma}$, and
  $\vec{\Lambda\bs{i}}$ as its $i$-th index. It is trivial to see that the
  optimal solution with minimum Frobenius norm is achieved when 
  \[\vec{\Lambda}\bs{i} = \begin{cases} 
    \vec{\Sigma\bs{i}}, & \text{if} \; \; \vec{\Sigma}\bs{i} \le s \\
    s,  & \text{otherwise}.
   \end{cases} \] 
   This is exactly what~\cref{alg:spectralNormOptimal} implements.

\subsection{Algorithm for Stable Rank Normalization~(SRN)}
\label{sec:alg_srn}
We provide a
general procedure in~\cref{alg:stablerankNorm} to solve the stable
rank normalization problem for $k \geq 1$ (the solution for $k=0$ is
straightforward
from~\Cref{thm:srankOptimal}).~\cref{claim:stableRankAlgo} provides
the properties of the algorithm. The algorithm is constructed so that
the prior knowledge of the rank of the matrix is not necessary.
\begin{claim} 
\label{claim:stableRankAlgo}
Given a matrix $\vec{W}$, the desired stable rank $r$, and the
partitioning index $k\ge 1$,~\cref{alg:stablerankNorm} requires
computing the top $l$ $(l \leq k)$ singular values and vectors of
$\vec{W}$. It returns $\widehat{\vec{W}}_l$ and the scalar $l$ such
that $\srank{\widehat{\vec{W}}_l} = r$, and the top $l$ singular
values of $\vec{W}$ and $\widehat{\vec{W}}_l$ are the same. If $l=k$,
then the solution provided is the optimal solution to the
problem~\eqref{eq:srankProblem} with all the constraints satisfied,
otherwise, it returns the largest $l$ up to which the spectrum is
preserved. %
\end{claim}

\begin{minipage}[t]{0.45\linewidth}
  \begin{algorithm}[H]
    \centering
    \caption{Stable Rank Normalization}
    \label{alg:stablerankNorm}
    \begin{algorithmic}[1]\footnotesize
      \INPUT $\vec{W} \in \reals^{m \times n}$, $r$, $k \geq 1$
      \STATE $\vec{S}_1 \gets \mathbf{0}$, $\beta  \gets \norm{\vec{W}}_\forb^2$, $\eta \gets 0$, $l \gets 0$
      \FOR {$i \in \{1, \cdots, k\}$}
      \STATE $\{\vec{u}_i, \vec{v}_i, \sigma_i\} \gets SVD(\vec{W}, i)$\\
      \COMMENT{Power method to get $i$-th singular value}
      \IF{$r \geq \br{\sigma_i^2 + \eta}/\sigma_1^2$}
      \label{eq:stableIf}
      \STATE $\vec{S}_1 \gets \vec{S}_1 + \sigma_i \vec{u}_i \vec{v}_i^{\top}$
      \label{eq:greedyStep}
      \STATE $\eta \gets  \eta + \sigma_i^2, \beta \gets \beta - \sigma_i^2$
      \STATE $l \gets  l+1$
      \ELSE
      \STATE break
      \ENDIF
      \ENDFOR
      \STATE $\eta \gets r\sigma_1^2 - \eta$ \\
      \OUTPUT $\widehat{\vec{W}}_l \gets \vec{S}_1 +\sqrt{\frac{ \eta}{\beta}} (\vec{W} - \vec{S}_1)$, $l$
      \label{eq:returnStep}
    \end{algorithmic}
  \end{algorithm}
\end{minipage}\;\;\;\;\begin{minipage}[t]{0.45\linewidth}
  \begin{algorithm}[H]
    \centering
    \caption[Stable Rank Normalization for a Linear Layer]{\label{algo:spectralStable} SRN for a Linear Layer in NN}
    \label{alg:final}
    \begin{algorithmic}[1]\footnotesize
      \INPUT $\vec{W} \in \reals^{m \times n}$, $r$, learning rate $\alpha$, mini-batch dataset $\mathcal{D}$
      \STATE Initialize $\vec{u} \in \reals^m$ with a random vector.%
      \STATE $\vec{v} \gets \frac{\vec{W}^\top \vec{u}}{\norm{\vec{W}^\top \vec{u}}}$, $\vec{u} \gets \frac{\vec{W}^\top \vec{v}}{\norm{\vec{W}^\top \vec{v}}}$\\
      \COMMENT{Perform power iteration}
      \STATE  $\sigma(\vec{W})= \vec{u}^{\top} \vec{W} \vec{v}$
      \STATE  $\vec{W}_f = \vec{W}/ \sigma(\vec{W})$
      \COMMENT{Spectral Normalization}
      \STATE $\widehat{\vec{W}} = \vec{W}_f - \vec{u} \vec{v}^{\top}$
      \IF {$\norm{\widehat{\vec{W}}}_{\forb} \le \sqrt{r-1}$}
      \STATE \OUTPUT $\vec{W}_f$ 
      \ENDIF
      \STATE $\vec{W}_f = \vec{u} \vec{v}^{\top} + \widehat{\vec{W}} \frac{ \sqrt{r- 1} }{ \norm{\widehat{\vec{W}}}_{\forb} }$
      \COMMENT {Stable Rank Normalization}
      \STATE \OUTPUT $\vec{W} \leftarrow \vec{W} - \alpha \nabla_{\vec{W}} L(\vec{W}_f, \mathcal{D})$
    \end{algorithmic}  
  \end{algorithm}
\end{minipage}
\paragraph{Combining Stable Rank and Spectral Normalization for \gls{nn}s}

Following the arguments provided in~\cref{sec:intro,sec:whyStable},
for better generalisability, we propose to normalize {\em both} the
stable rank and the spectral norm of each linear layer of a \gls{nn}
simultaneously. To do so, we first perform approximate
\gls{sn}~\citep{miyato2018spectral}, and then perform optimal
\gls{srn} (using \cref{alg:stablerankNorm}). We use $k=1$ to ensure
that the first singular value (which is now normalized) is
preserved.~\cref{algo:spectralStable} provides a simplified procedure
for the same for a given linear layer of a \gls{nn}. Note that the
computational cost of this algorithm is {\em exactly the same as that
of~\gls{sn}}, which is to compute the top singular value using the
power iteration method.

\paragraph{Implementation details of Stable Rank Normalization}
The SRN algorithm~(\Cref{alg:final}) is applied on neural networks in exactly
the same way as Spectral Normalization~(SN) is applied
in~\citet{miyato2018spectral}. In particular, every iteration of gradient
descent~(or its variant) conducts one forward pass followed by a backward pass
to compute gradients, and then the parameters are updated according to the
update step of the optimisation algorithm. Finally, each weight matrix is
projected onto the space of matrices with constrained stable rank by
applying~\Cref{alg:final} on each weight matrix.

\begin{remark}
  While applying~\Cref{alg:final} on fully connected layers is straightforward,
for convolution layers, we apply SRN on a matrix corresponding to the linear
transformation of the convolution operation, similar
to~\citet{miyato2018spectral}. For a convolution layer with weight
\(W\in\reals^{c_i\times c_o\times h\times w}\) where  \(c_i\) and \(c_o\) are
the number of input and output filters respectively, and \(h\times w\) is the
dimension of individual filters, we flatten the layer into a \(\br{c_i, c_o h
w}\)-dimensional matrix and apply~\Cref{alg:final} on this matrix. While this is
not an exact representation of the linear transformation of a convolutional
layer, this heuristic  helps in the computational speed of the operation. We
also note that this framework is not accurate for residual networks. We apply
SRN on the learnable layers inside the residual block, however the actual
transformation of a residual block also includes the identity map applied by the
skip connection. It is not straightforward whether this can be represented by a linear transformation as there are also non-linear activations inside the residual block, which the skip connection bypasses. However, despite these shortcomings, our method works well in practice as the next section shows.
\end{remark}

\section{Experiments on a discriminative setup}
\label{sec:experiments-srn}
\paragraph{Dataset and architectures} For classification, we perform experiments
on ResNet-110~\citep{HZRS:2016}, WideResNet-28-10~\citep{Zagoruyko2016},
DenseNet-100~\citep{Huang2017}, VGG-19~\citep{simonyan2014very}, and
AlexNet~\citep{krizhevsky2009learning} using the CIFAR100 and
CIFAR10~\citep{krizhevsky2009learning} datasets. We train them using standard
training recipes with SGD, using a learning rate of $0.1$~(except AlexNet where
we use a learning rate of $0.01$), and a momentum of $0.9$ with a batch size of
$128$ (further details in Appendix~\ref{sec:expr-settings}). In addition to
training for a fixed number of epochs, we also present results where the
training accuracy~(as opposed to the number of iterations) is used as a stopping
criterion to show that our regulariser performs well with a range of stopping
criteria.

For GAN experiments, we use CIFAR100, CIFAR10, and
CelebA~\citep{liu2015faceattributes} datasets. We show results on both,
conditional and unconditional GANs. Please refer to~\cref{sec:expr-settings} for
further details about the training setup.

\paragraph{Choosing stable rank} Given a matrix $\vec{W} \in \reals^{m \times
  n}$, the desired stable rank $r$ is controlled using a single
hyperparameter $c$ as $r = c \; \min(m,n)$, where $c \in (0, 1]$.
For simplicity, we use the same $c$ for all the linear layers. 
Note that if $c=1$, or for a given $c$, if $\srank{\vec{W}} \leq r$,
then SRN boils down to SN. For classification, we choose $c = \{0.3, 0.5\}$, and compare SRN against standard
training (Vanilla) and training with \gls{sn}. For GAN experiments, we choose $c = \{0.1, 0.3, 0.5, 0.7, 0.9\}$, and
compare SRN-GAN against SN-GAN~\citep{miyato2018spectral}, WGAN-GP~\citep{Gulrajani2017}, and orthonormal regularisation GAN
(Ortho-GAN)~\citep{Brock2016}.

\begin{figure}[t]
  \centering\small
  \begin{subfigure}[!t]{0.185\linewidth}
    \def\svgwidth{0.98\linewidth}
    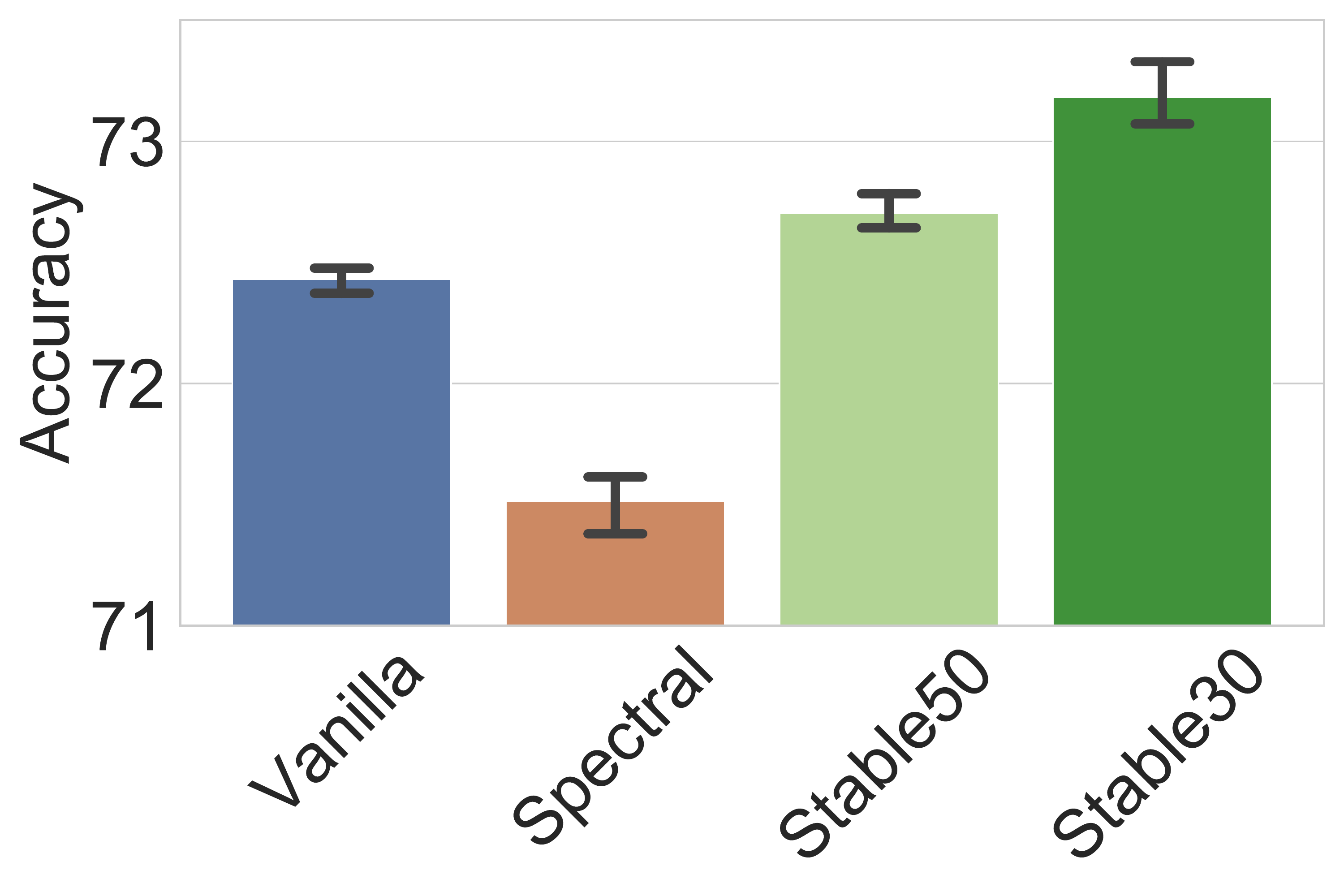
    \subcaption{ ResNet110}
  \end{subfigure}
  \begin{subfigure}[!t]{0.21\linewidth}
    \def\svgwidth{0.98\linewidth}
    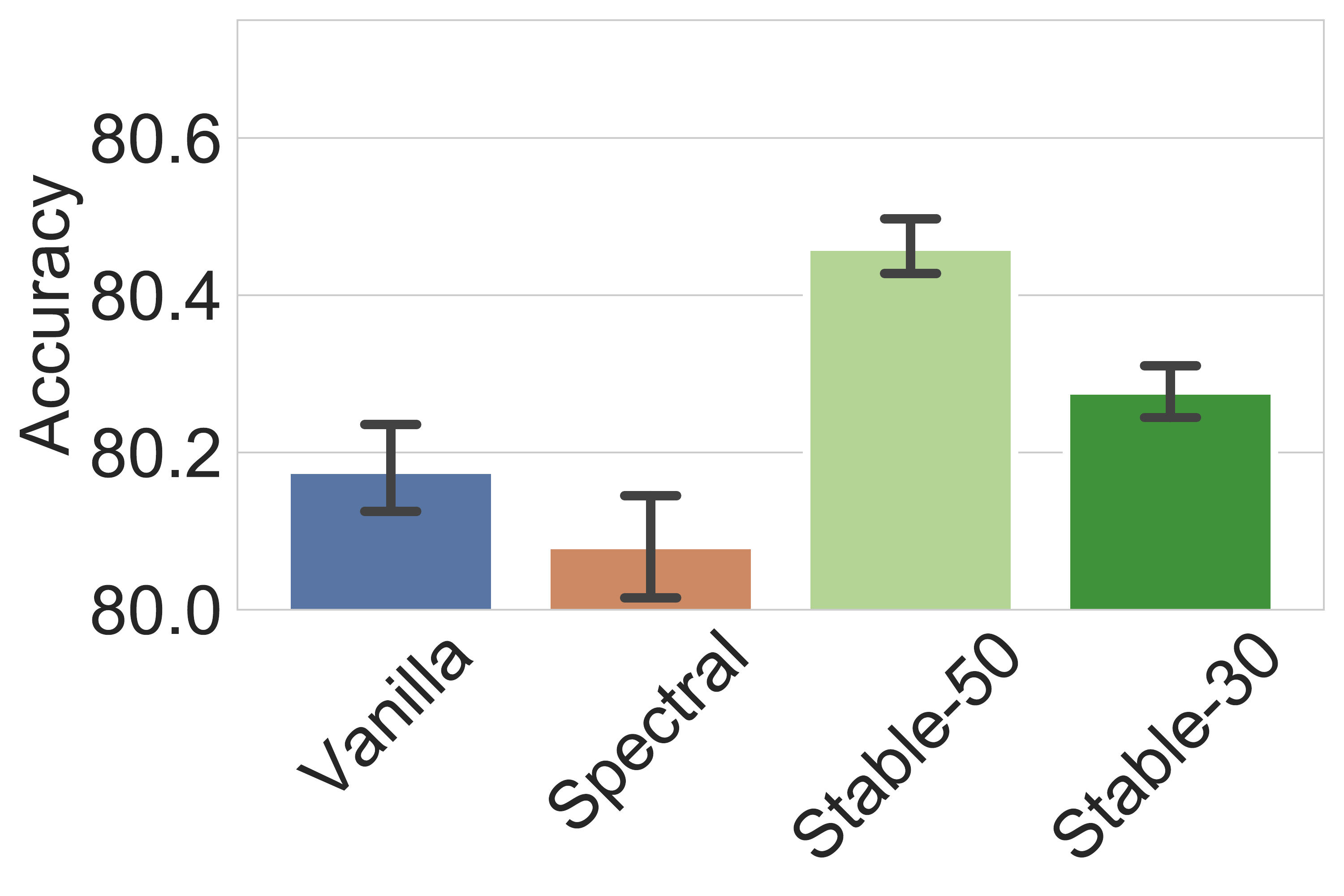
    \subcaption{ WideResNet28}
  \end{subfigure}
   \begin{subfigure}[!t]{0.185\linewidth}
    \def\svgwidth{0.98\linewidth}
    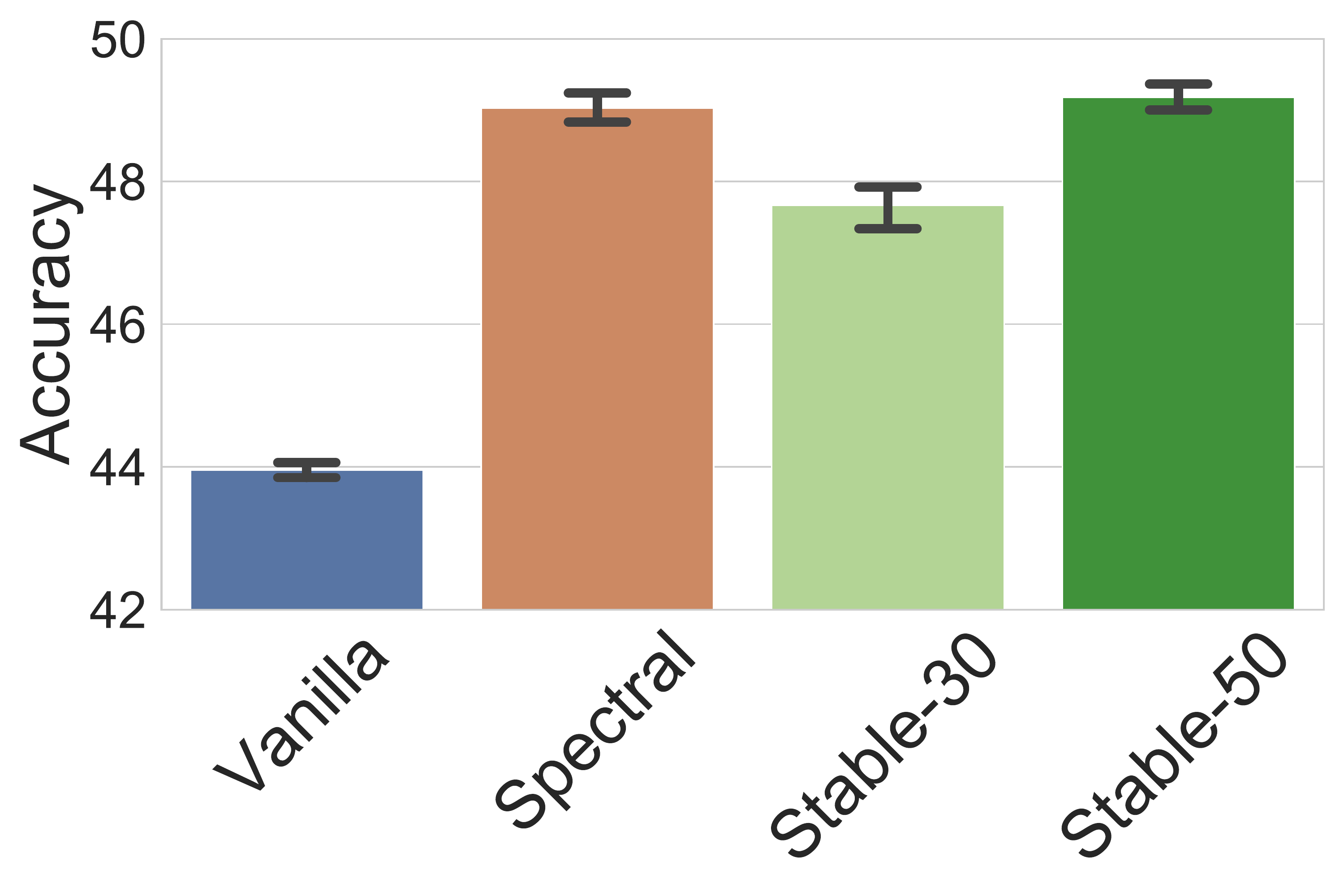
    \subcaption{ Alexnet}
  \end{subfigure}
  \begin{subfigure}[!t]{0.19\linewidth}
    \def\svgwidth{0.99\linewidth}
    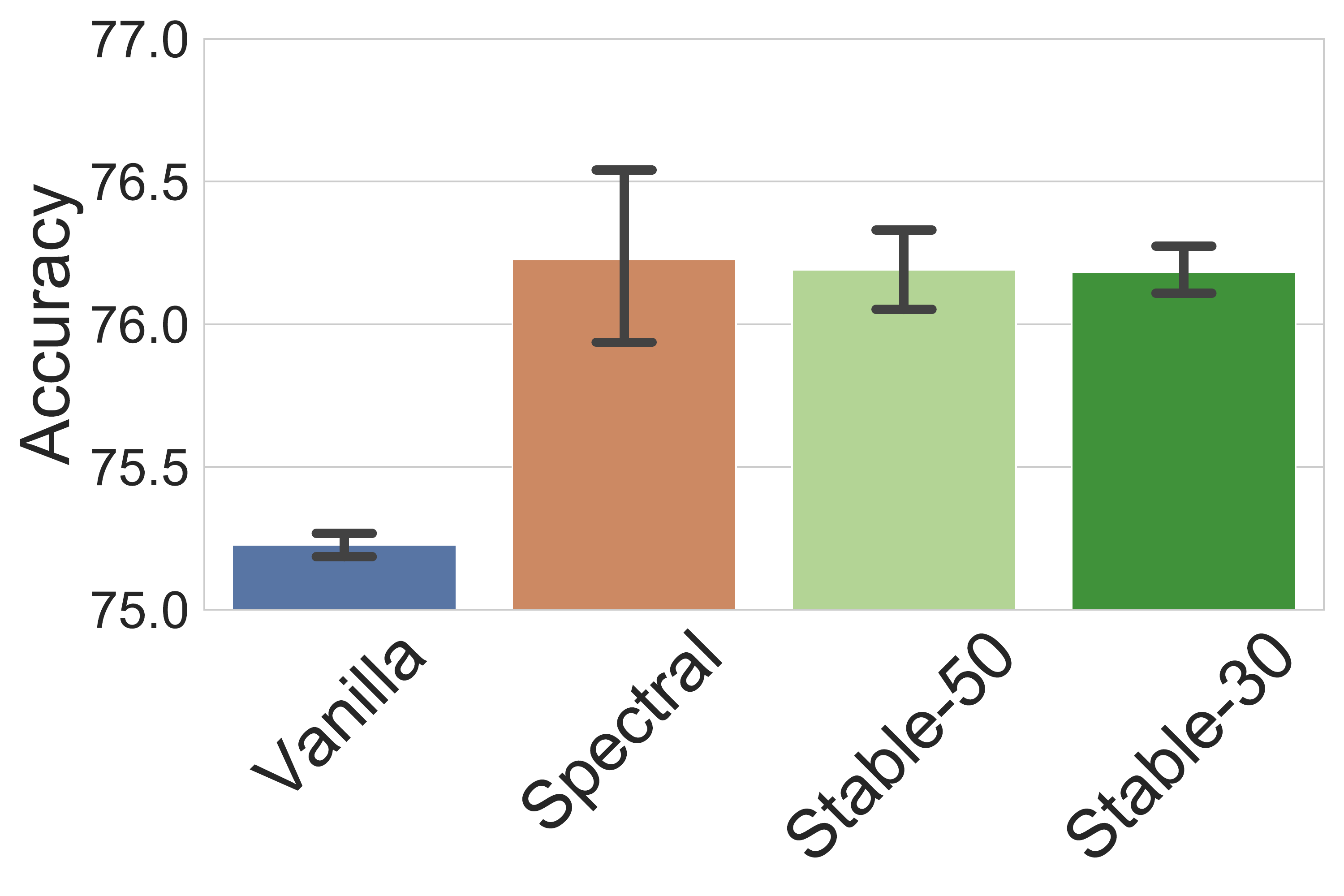
    \subcaption{ Densenet-100}
  \end{subfigure}
   \begin{subfigure}[!t]{0.185\linewidth}
    \def\svgwidth{0.98\linewidth}
    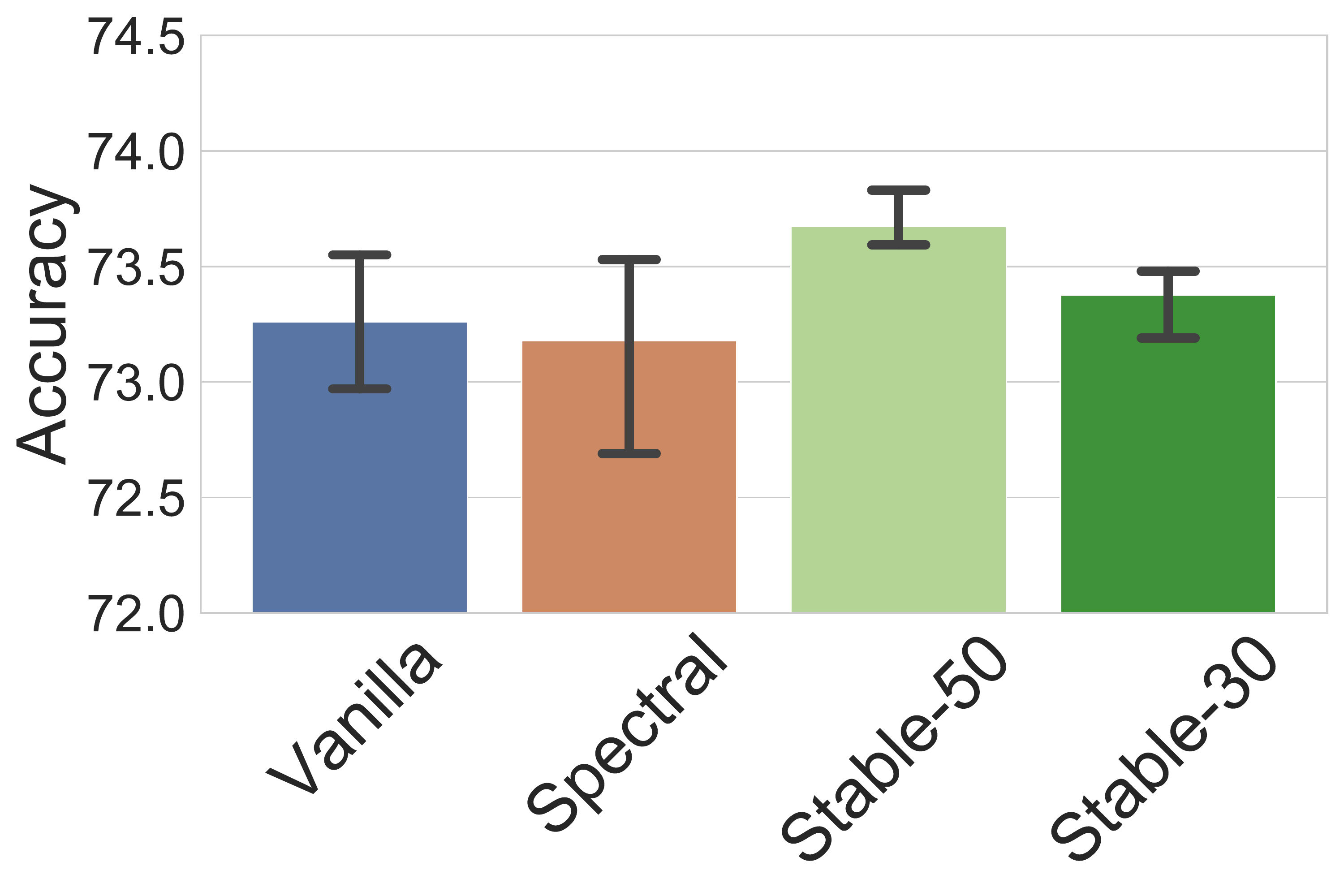
    \subcaption{ VGG-19}
  \end{subfigure}
  \caption[Test accuracies on CIFAR100 with fixed number of training epochs]{Test accuracies on CIFAR100 for clean data using the number of epochs as stopping criterion. Higher is better.}
  \label{fig:test-acc}
\end{figure}

\begin{figure}[t]
  \centering\small
  \begin{subfigure}[!t]{0.24\linewidth}
    \def\svgwidth{0.98\linewidth}
    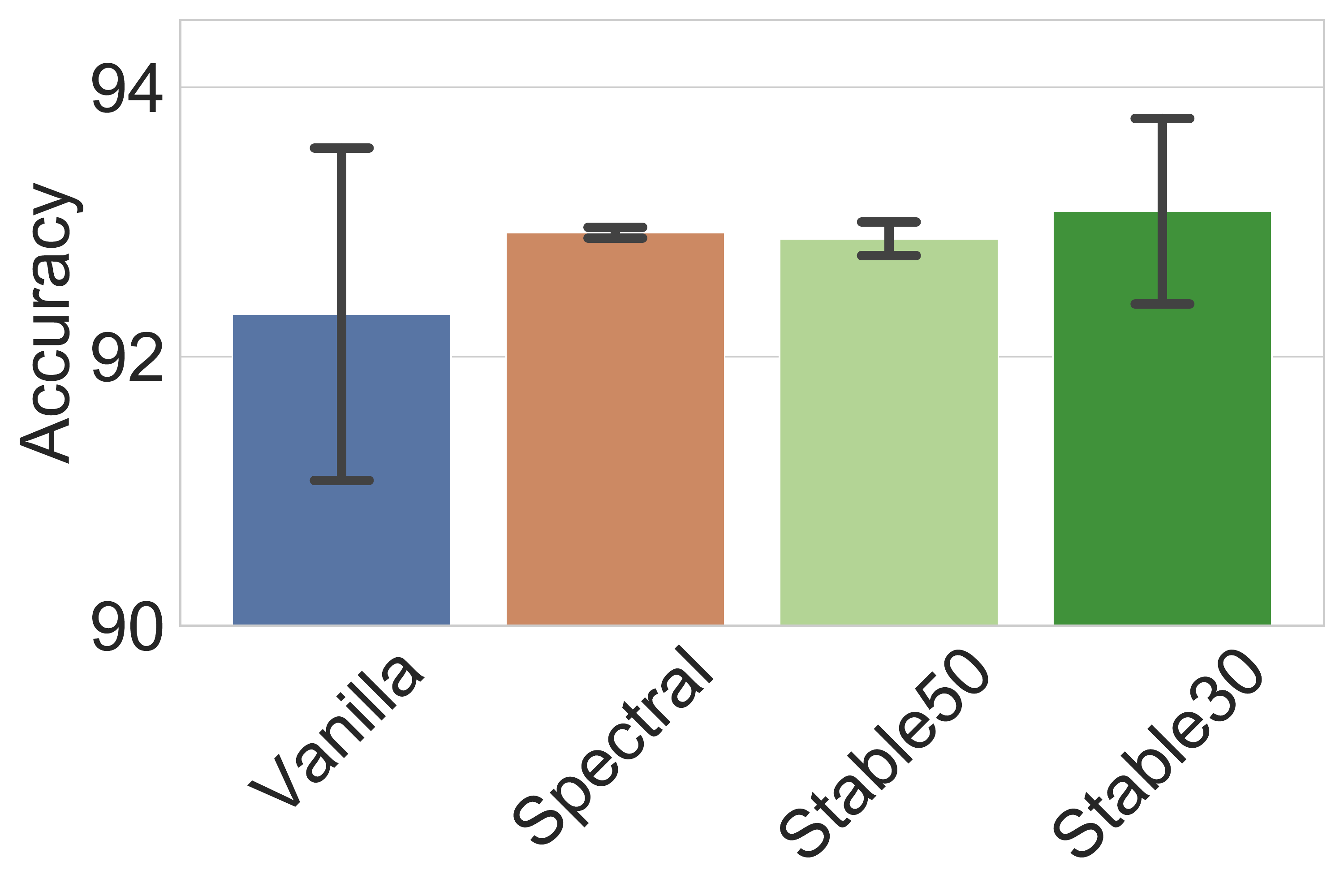
    \subcaption{ResNet110}
  \end{subfigure}
  \begin{subfigure}[!t]{0.24\linewidth}
    \def\svgwidth{0.98\linewidth}
    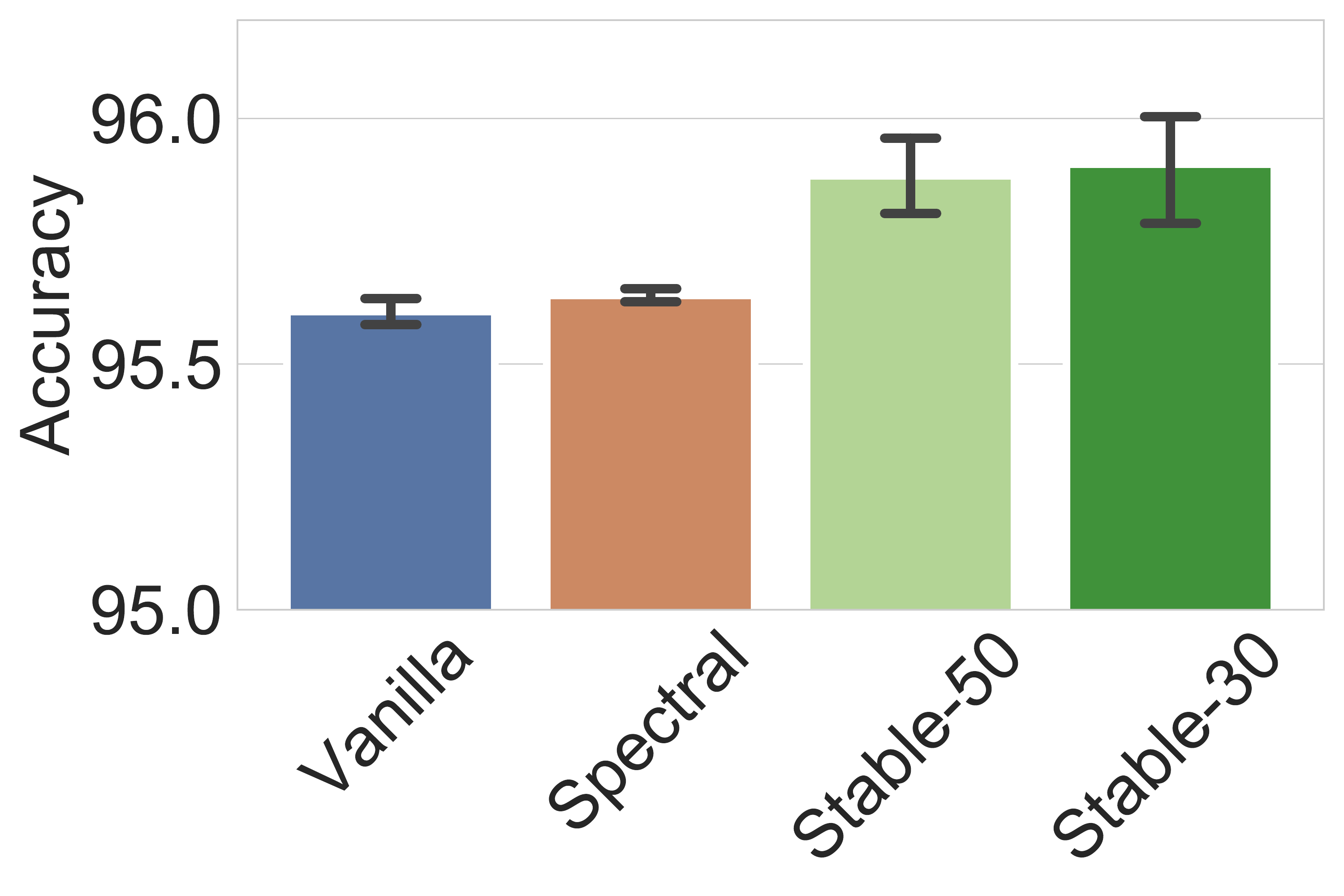
    \subcaption{WideResNet28}
  \end{subfigure}
   \begin{subfigure}[!t]{0.24\linewidth}
    \def\svgwidth{0.98\linewidth}
    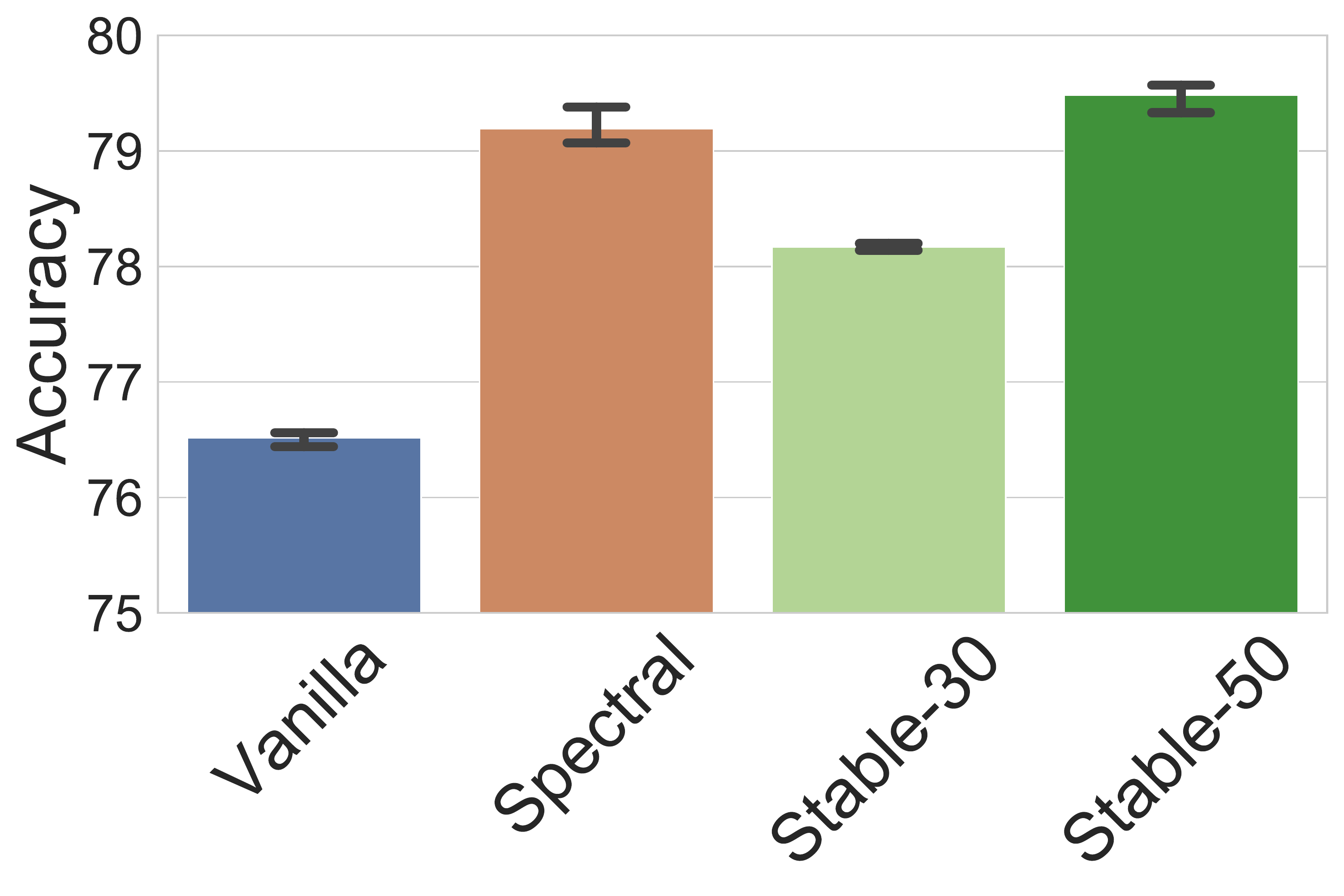
    \subcaption{ Alexnet}
  \end{subfigure}
  \begin{subfigure}[!t]{0.24\linewidth}
    \def\svgwidth{0.98\linewidth}
    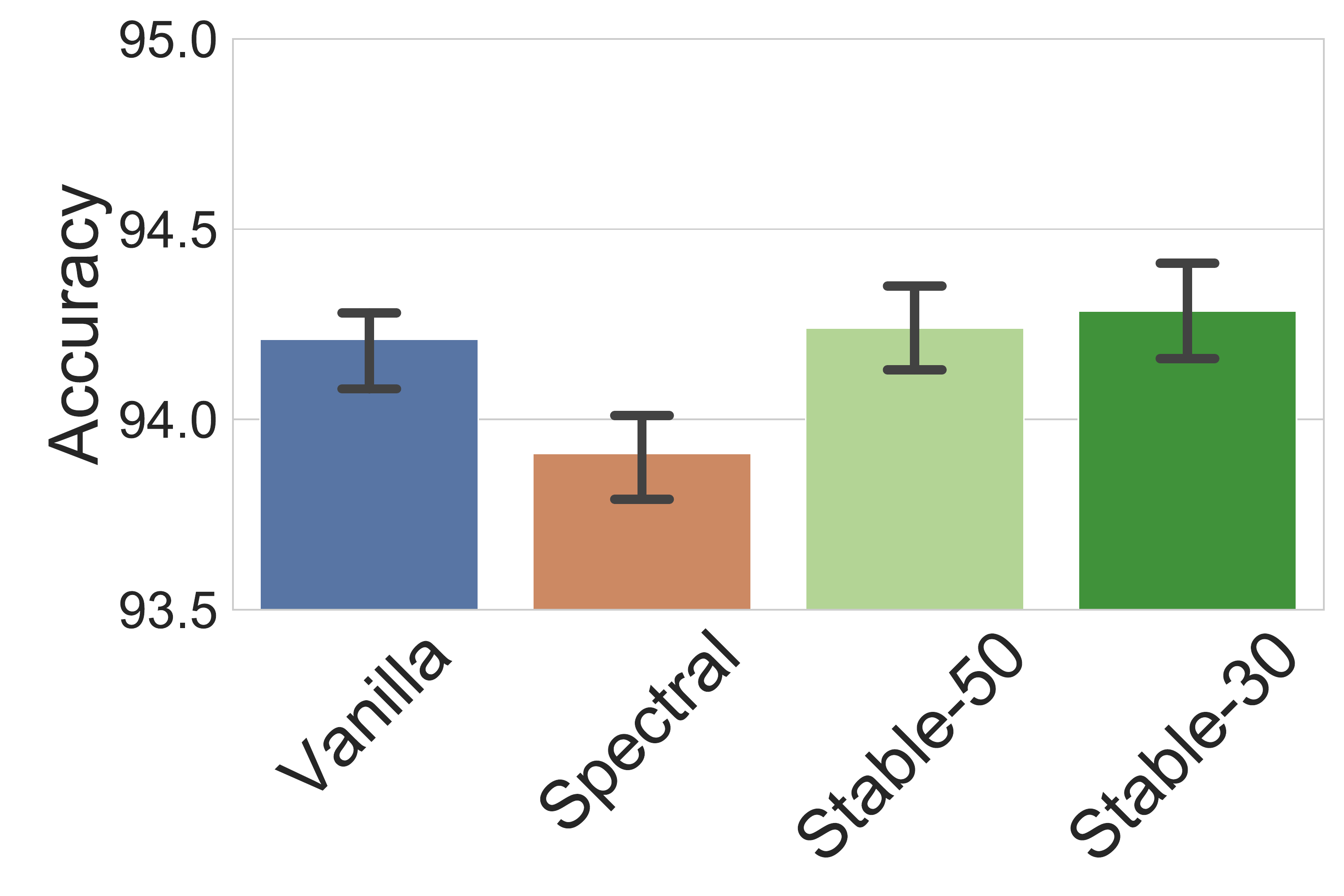
    \subcaption{ Densenet-100}
  \end{subfigure}
  \caption[Test error on CIFAR10 with fixed number of training epochs]{Test accuracies on CIFAR10 for clean data using the number of epochs as stopping criterion. Higher is better.}
  \label{fig:test-acc-epoch-c10}
\end{figure}

\subsection{Classification experiments}\label{sec:gen_exp_results} We perform
each experiment $5$ times using a new random seed each time and report the mean and the $75\%$ confidence interval for the test error. We plot the test accuracy
on CIFAR100 in~\cref{fig:test-acc} and CIFAR10
in~\cref{fig:test-acc-epoch-c10}.

These experiments show that the test accuracy of \gls{srn} on a wide variety
\gls{nn}s is always higher than the Vanilla and SN (except for SRN-50  on
Alexnet where \gls{srn} and \gls{sn} are almost equal). However, \gls{sn}
performs slightly worse than Vanilla for WideResNet-28 and ResNet110. The fact
that \gls{srn} also involves \gls{sn}, combined with the above observation,
indicates that even though \gls{sn} reduced the learning capability of these
networks, normalizing stable rank must have improved it significantly in order
for \gls{srn} to outperform Vanilla. For example, in the case of ResNet110,
\gls{sn} is $71.5\%$ accurate whereas \gls{srn} provides an accuracy of
$73.2\%$. In addition to this, we also note that even though \gls{sn} is used
extensively for the training of GANs, it is not a popular choice when it comes
to training standard \gls{nn}s for classification. We suspect that this is
because of the decrease in performance, we observe here. Hence, as our
experiments indicate that~\Gls{srn} overcomes this,~\Gls{srn} could be a
potentially important regulariser in the classification setting.

\begin{figure}
  \centering\small
  \begin{subfigure}[!t]{0.185\linewidth}
    \def\svgwidth{0.98\linewidth}
    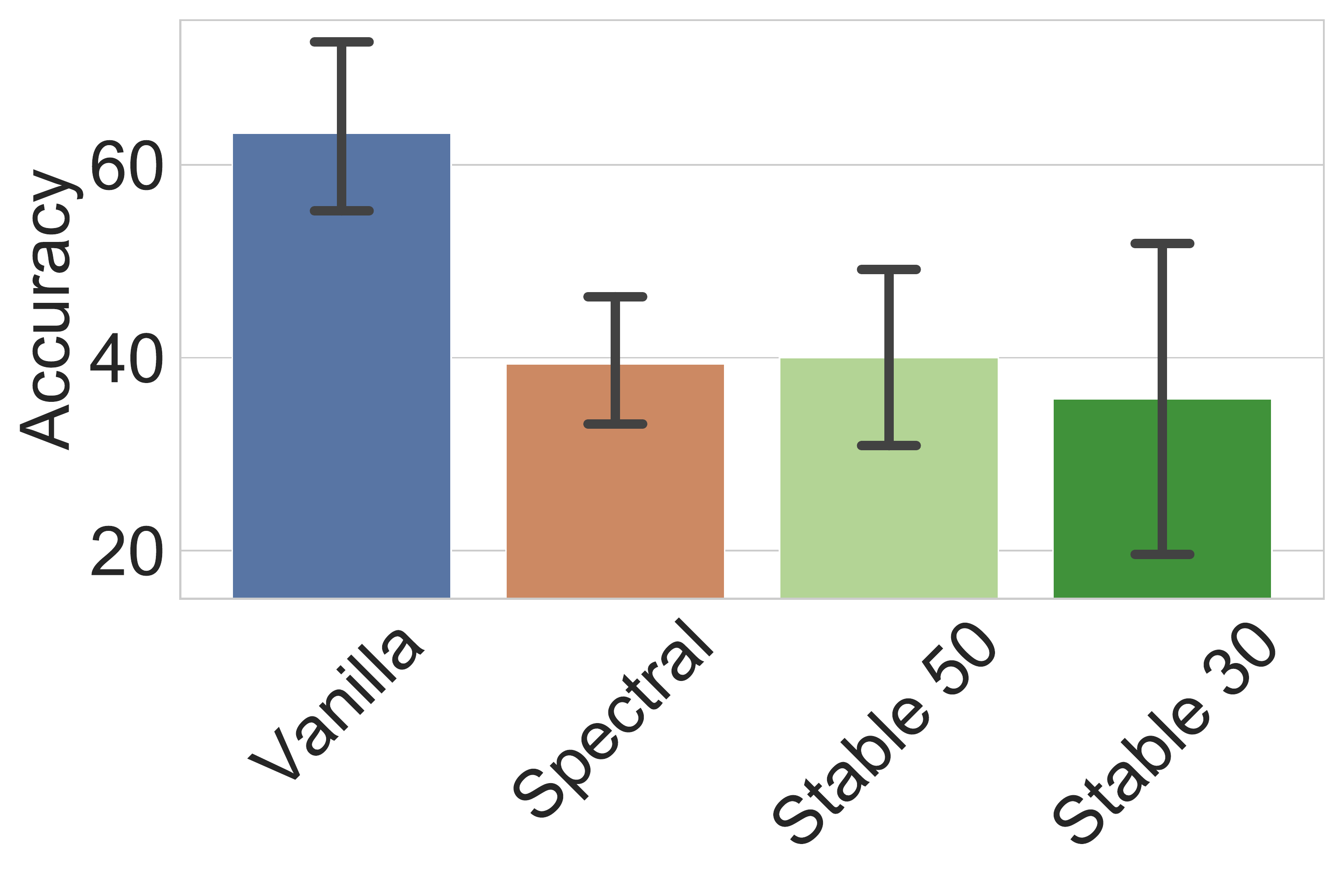
    \subcaption{ResNet110}
  \end{subfigure}
     \begin{subfigure}[!t]{0.21\linewidth}
    \def\svgwidth{0.99\linewidth}
    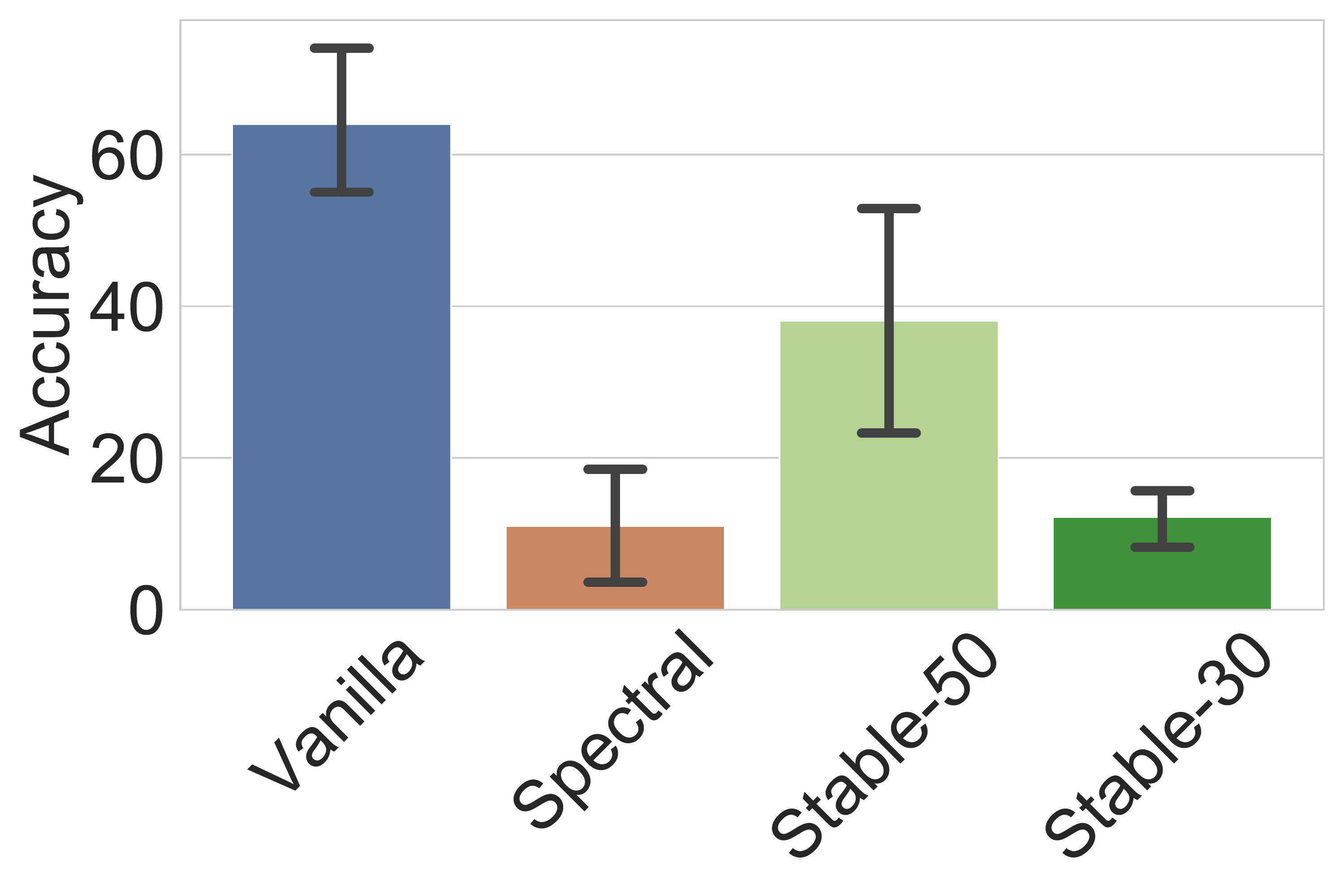
    \subcaption{WideResNet28}
  \end{subfigure}
   \begin{subfigure}[!t]{0.185\linewidth}
    \def\svgwidth{0.98\linewidth}
    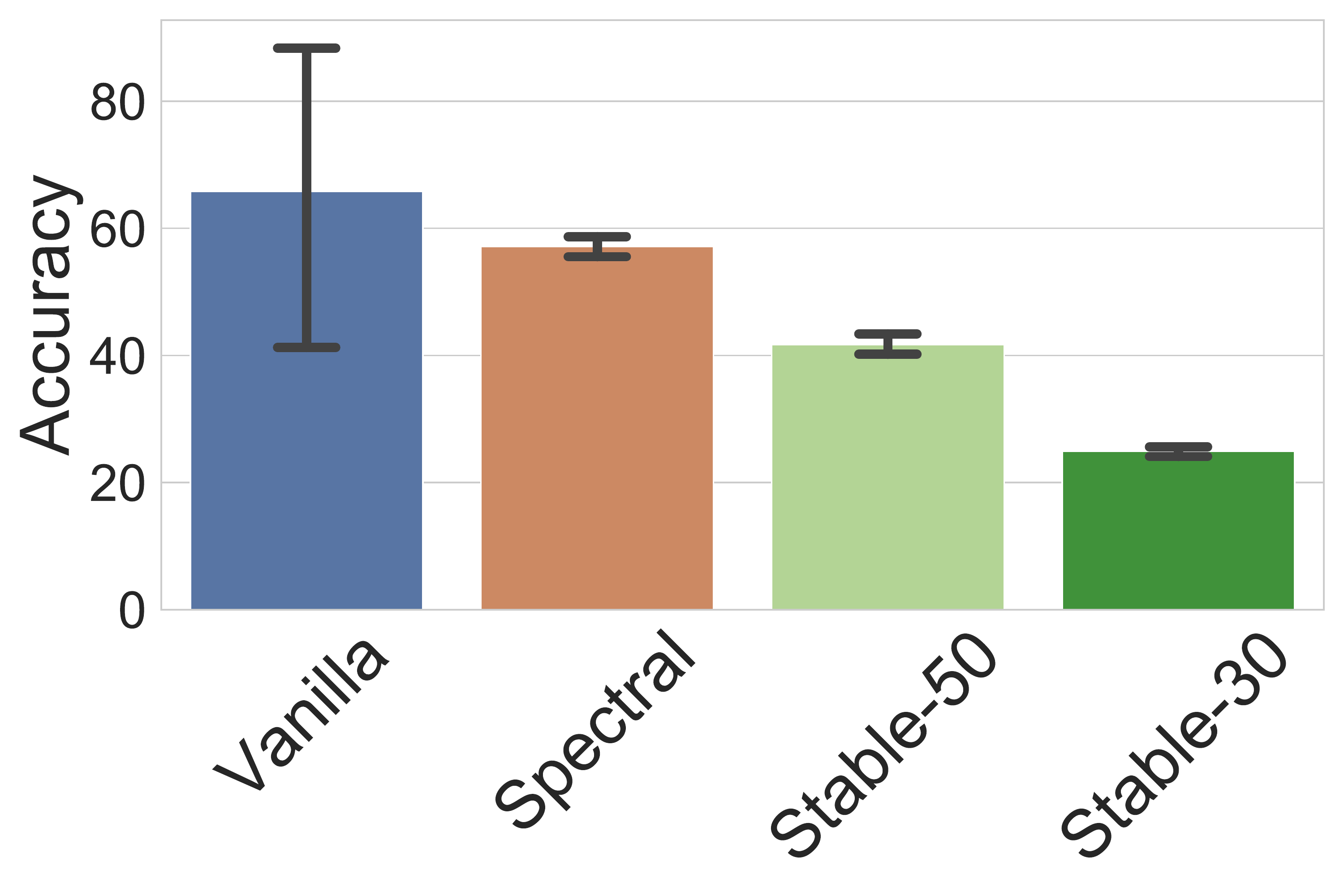
    \subcaption{Alexnet}
  \end{subfigure}
   \begin{subfigure}[!t]{0.19\linewidth}
    \def\svgwidth{0.99\linewidth}
    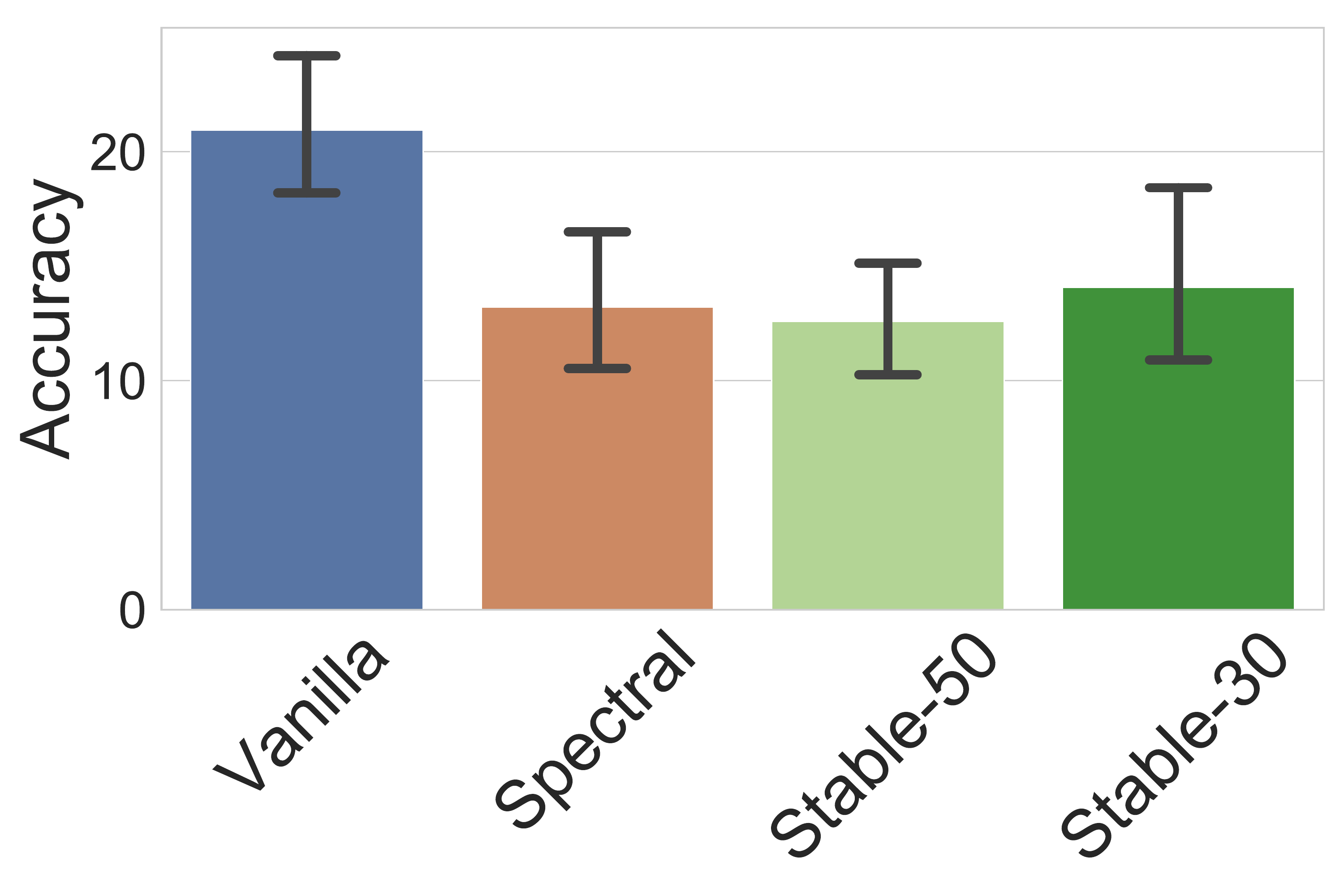
    \subcaption{Densenet-100}
  \end{subfigure}
   \begin{subfigure}[!t]{0.185\linewidth}
    \def\svgwidth{0.98\linewidth}
    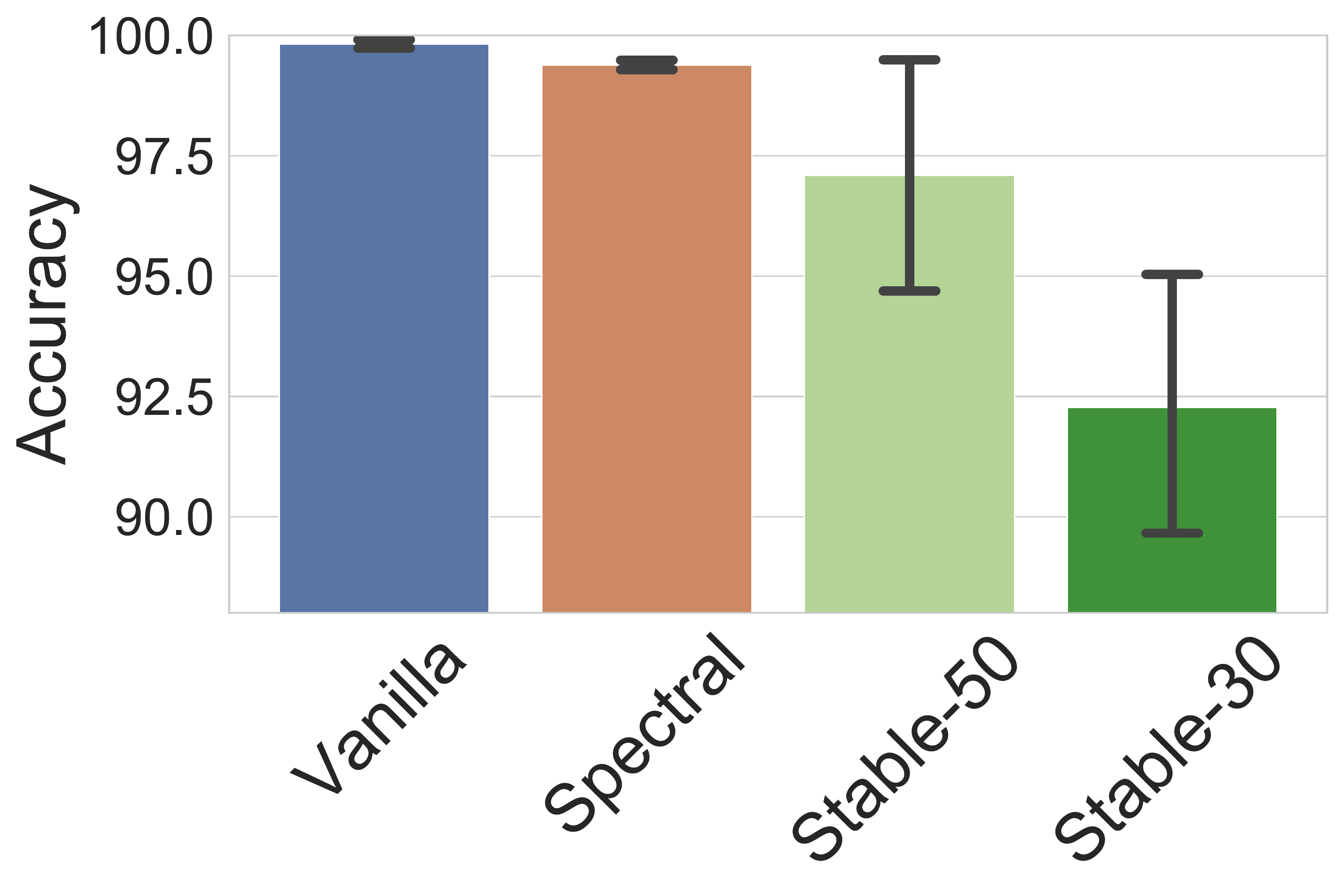
    \subcaption{VGG-19}
  \end{subfigure}
  \caption[Shattering Experiments on CIFAR100]{Train accuracies on CIFAR100 for shattering experiment. Lower indicates less memorisation, thus, better.}
  \label{fig:rand-acc}
\end{figure}
\begin{figure}[t]
  \centering
  \begin{subfigure}[t]{0.33\linewidth}
    \def\svgwidth{0.98\linewidth}
    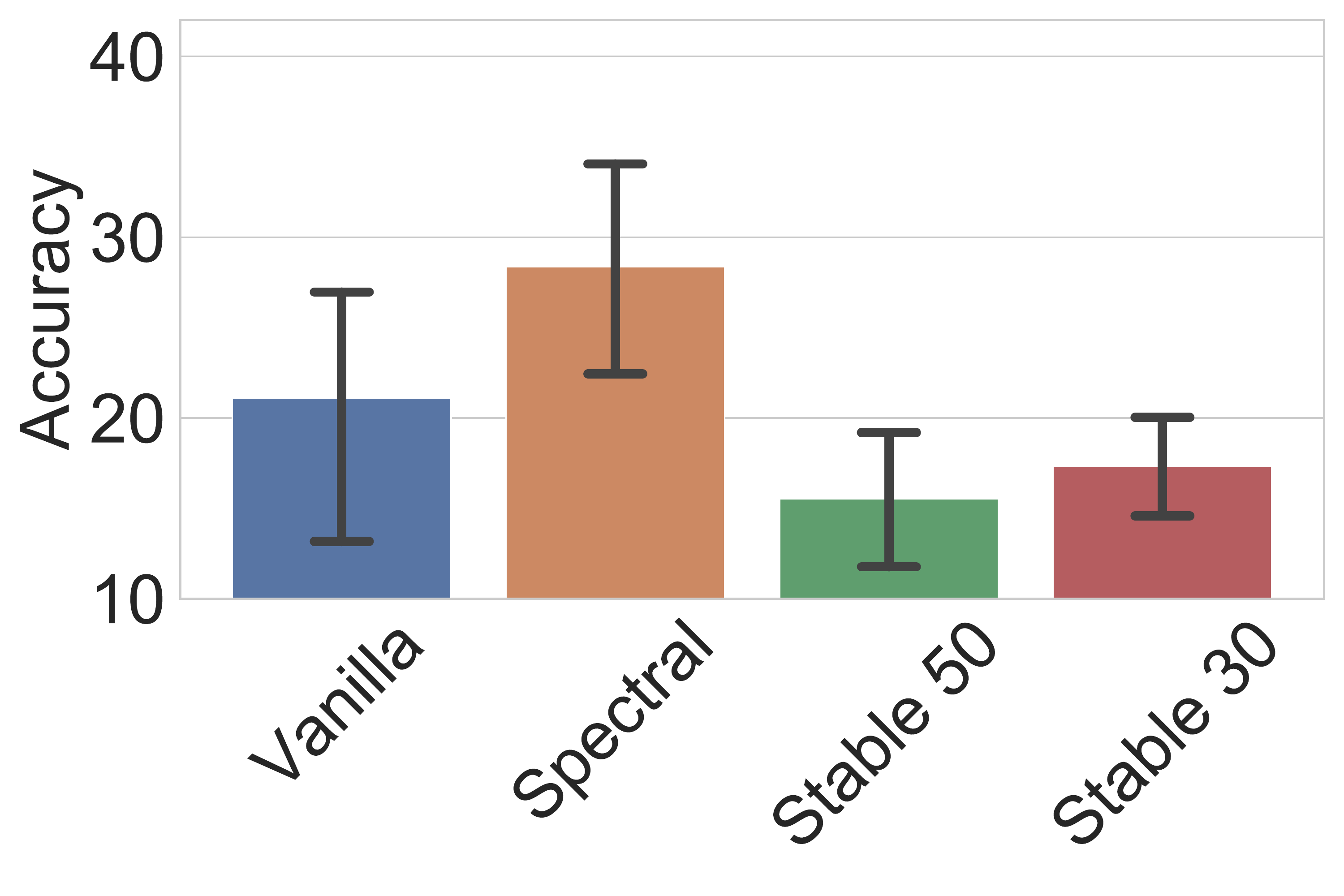
    \caption{ResNet110}\label{fig:r100-rand-train-c10}
  \end{subfigure}
  \begin{subfigure}[t]{0.33\linewidth}
    \def\svgwidth{0.98\linewidth}
    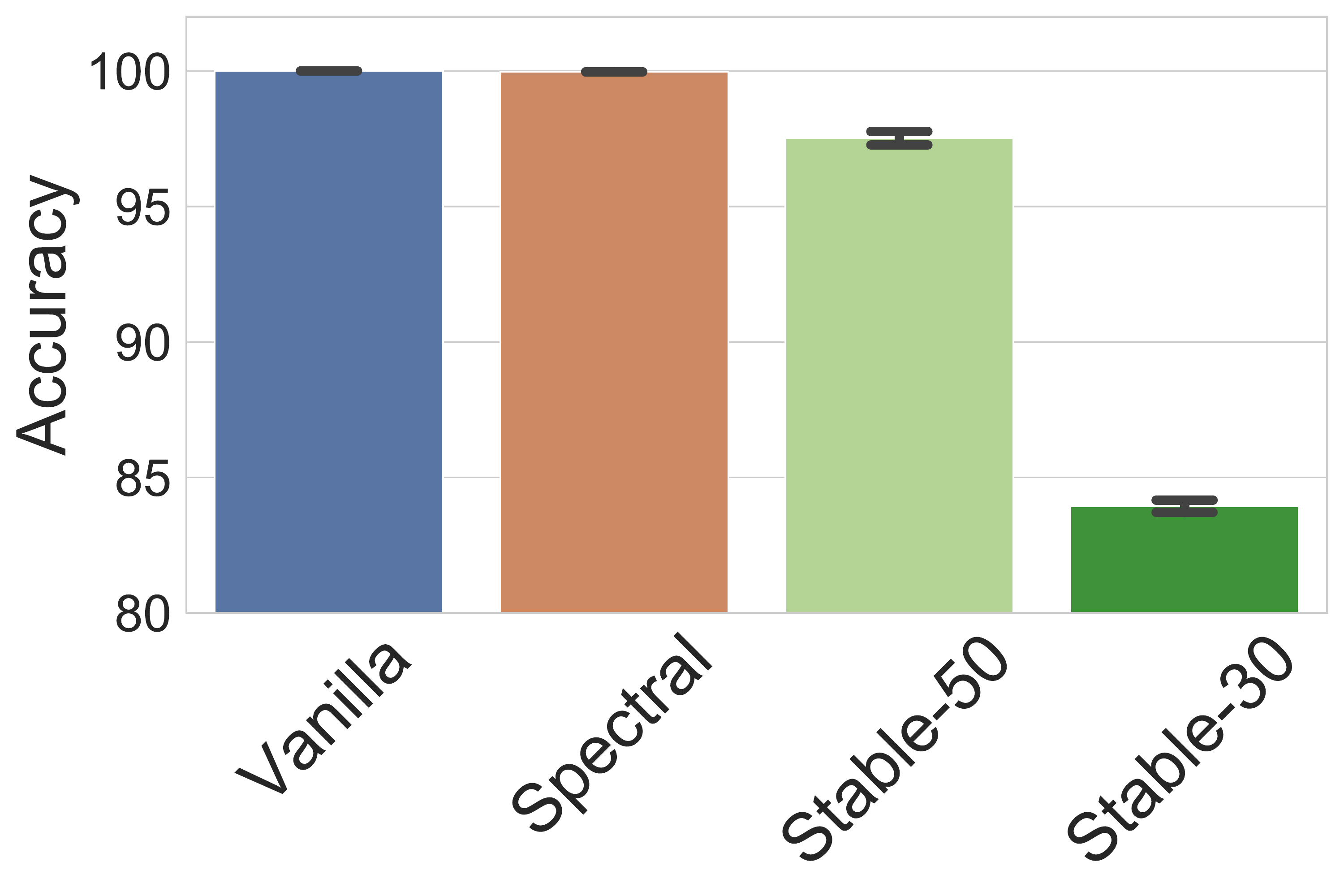
    \caption{Alexnet}\label{fig:alex-rand-train-c10}
  \end{subfigure}
  \caption[Shattering experiments on CIFAR10]{Train accuracies on CIFAR10 for shattering experiment. Lower indicates less memorisation, thus, better.}
  \label{fig:cifar-10-gen}
\end{figure}

\subsection{Shattering experiments}
Our previous set of experiments established that \gls{srn} provides improved
classification accuracies on various \gls{nn}s. Here we study the generalisation
behaviour of these models. Quantifying generalisation behaviour is non-trivial
and there is no clear answer to it. However, we utilise recent efforts that
explore the theoretical understanding of generalisation to study generalisation
of SRN in practice.

To inspect the generalisation behaviour in \gls{nn}s we begin with the
shattering experiment~\citep{Zhang2016}. It is a test of whether the network can
fit the training data well but not a label-randomised version of it (each image
of the dataset is assigned a random label). As $P\bs{y\vert \vec{x}}$ is
essentially uninformative because it is a uniformly random distribution, there
is no correlation between the labels and the data points. Thus, the best
possible test accuracy on this task~(the test labels would also be randomised)
is approximately $1\%$ on CIFAR100~(\(10\%\) in CIFAR10). A high training
accuracy --- which indicates a high generalisation gap (difference between train
and test accuracy) can be achieved only by memorising the train data. In these
experiments, the training of all models of one architecture are stopped after
the same number of epochs, which is double the number of epochs the model is
trained for on the clean datasets. ~\cref{fig:rand-acc} shows that \gls{srn}
reduces memorisation on random labels~(thus, reduces the estimate of the
Rademacher complexity~\citep{Zhang2016}) on CIFAR100.
In~\cref{fig:cifar-10-gen}, we plot the training accuracy on CIFAR10 and observe
a similar result. Importantly, the same model architectures and training
strategy were able to achieve high test accuracy for the clean dataset in the
previous section.

\subsection*{Training accuracy as stopping criterion}

We used the number of training epochs as our stopping criterion in the previous
experiments. To show that the results hold consistently across different
stopping criteria, we use a different stopping criterion here. In particular, we
use the training accuracy as a stopping criterion here. For ResNet110,
WideResNet-28, Densenet-100, and VGG-19 we use a training accuracy of $99\%$ as
a stopping criterion and report the test accuracy when that training accuracy is
achieved for the first time. For Alexnet, as SRN-30 never achieves a training
accuracy higher than $55\%$, we use $55\%$ as the stopping criterion and plot
the test accuracies in~\cref{fig:test-acc-stop} for CIFAR100. Our results show
that SRN-30 and SRN-50 outperform SN and vanilla consistently. In
Figure~\ref{fig:test-acc-stop-c10}, we show similar results for CIFAR10. As
AlexNet can achieve higher accuracy for CIFAR10, we use $85\%$ as the
stopping criterion for AlexNet on CIFAR10.

\begin{figure}[t]
  \centering\small
  \begin{subfigure}[!t]{0.185\linewidth}
    \def\svgwidth{0.99\linewidth}
    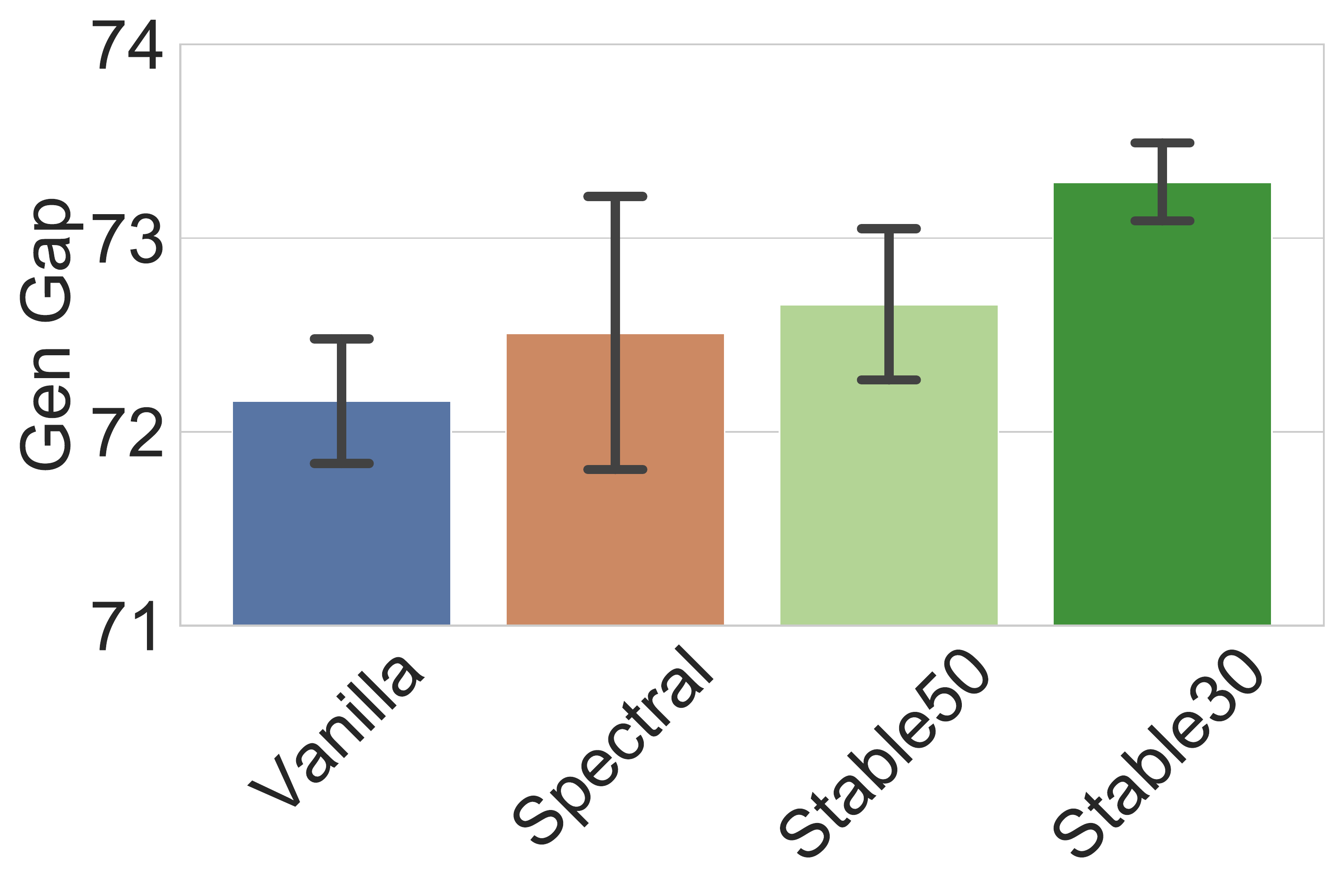
    \subcaption{ResNet110}
  \end{subfigure}
  \begin{subfigure}[!t]{0.21\linewidth}
    \def\svgwidth{0.99\linewidth}
    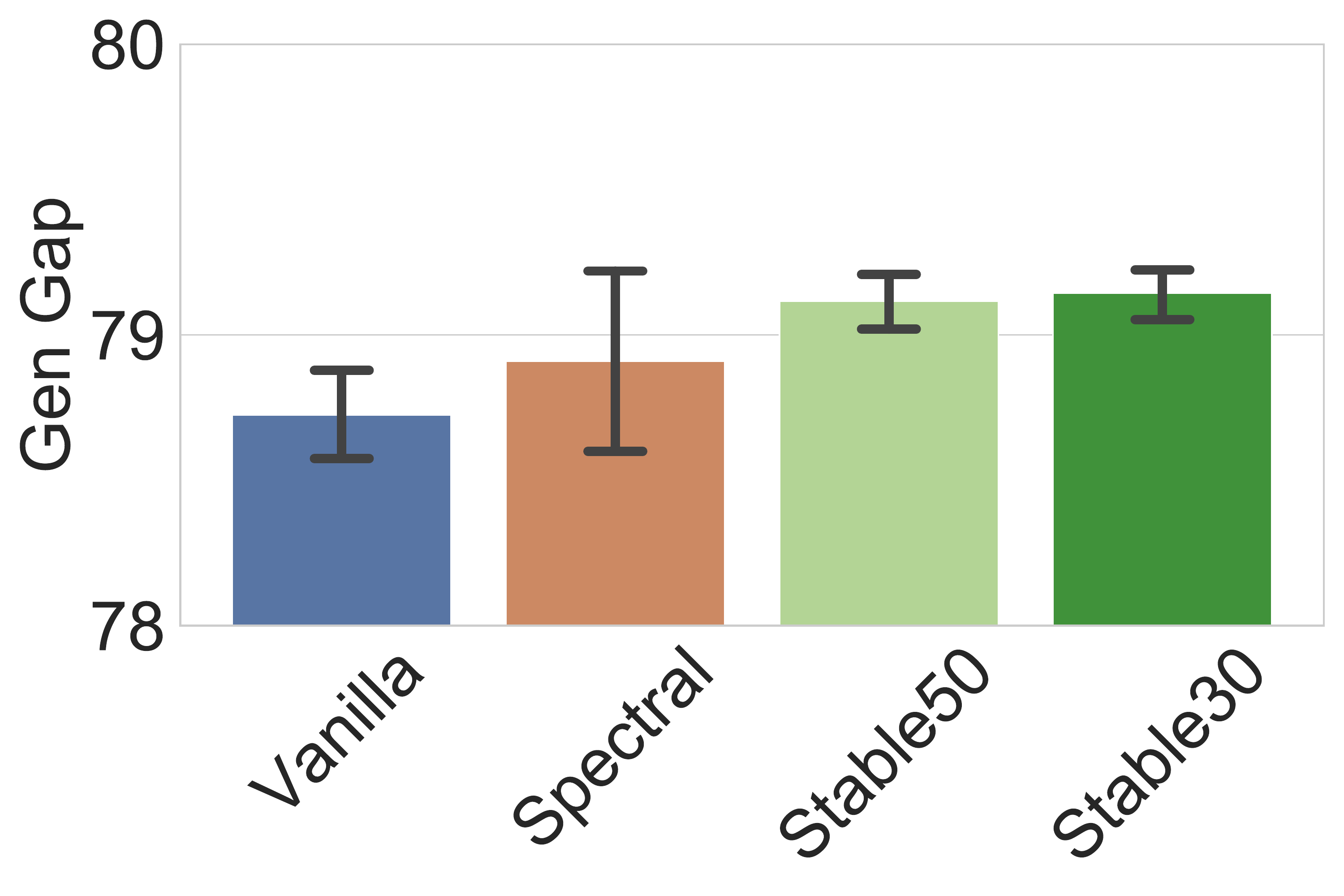
    \subcaption{WideResNet28}
  \end{subfigure}
   \begin{subfigure}[!t]{0.185\linewidth}
    \def\svgwidth{0.98\linewidth}
    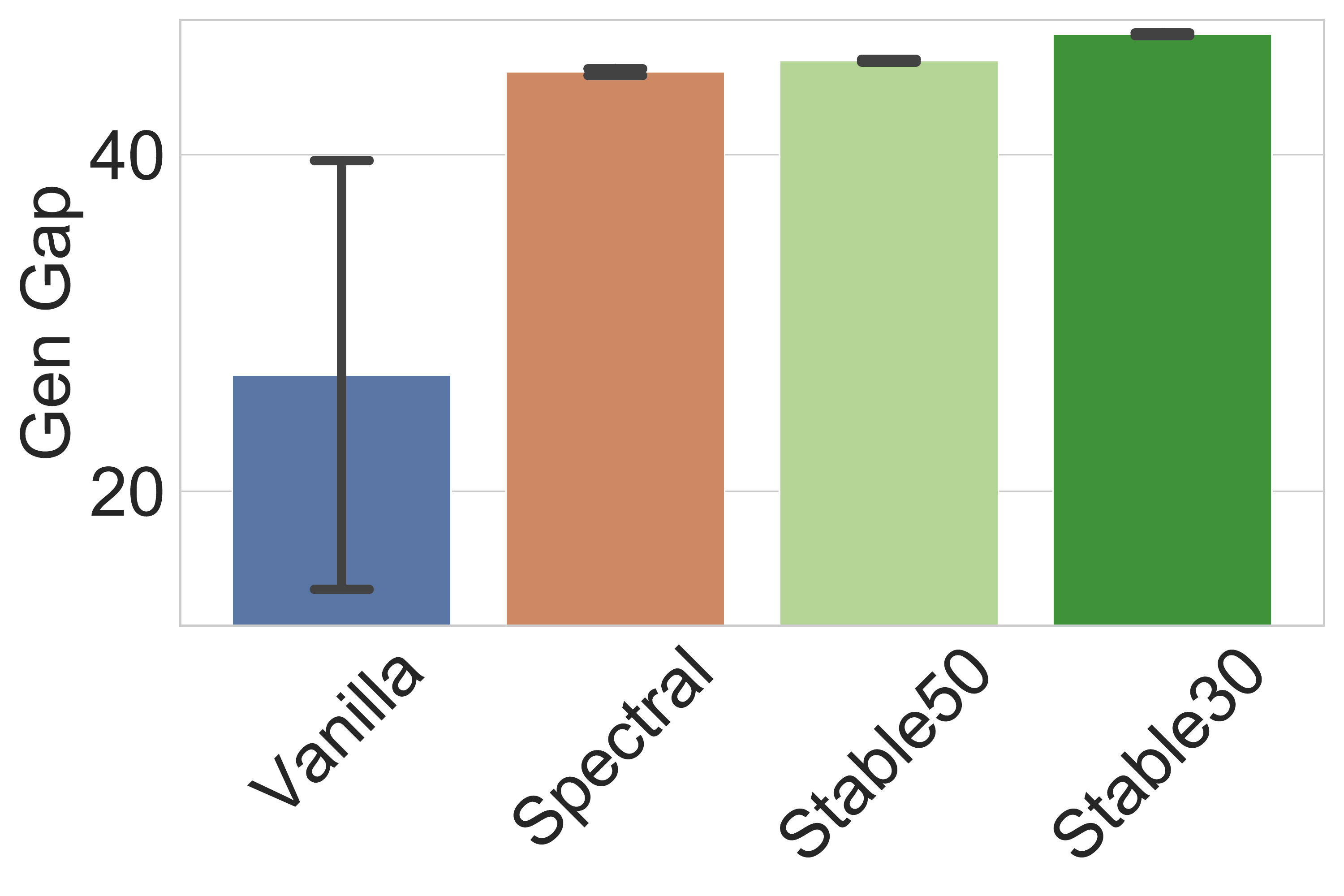
    \subcaption{Alexnet}
  \end{subfigure}
  \begin{subfigure}[!t]{0.19\linewidth}
    \def\svgwidth{0.99\linewidth}
    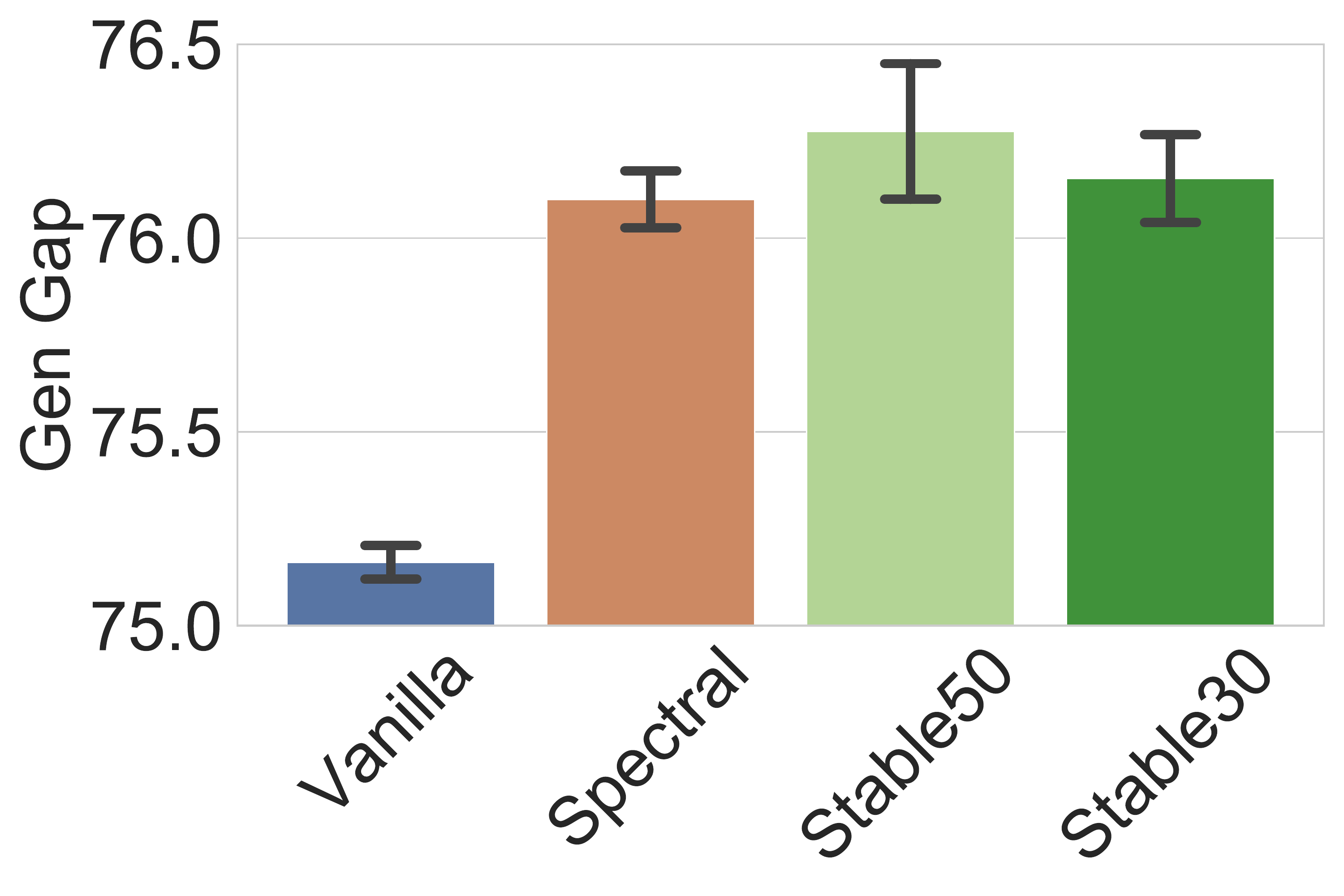
    \subcaption{Densenet-100}
  \end{subfigure}
   \begin{subfigure}[!t]{0.19\linewidth}
    \def\svgwidth{0.99\linewidth}
    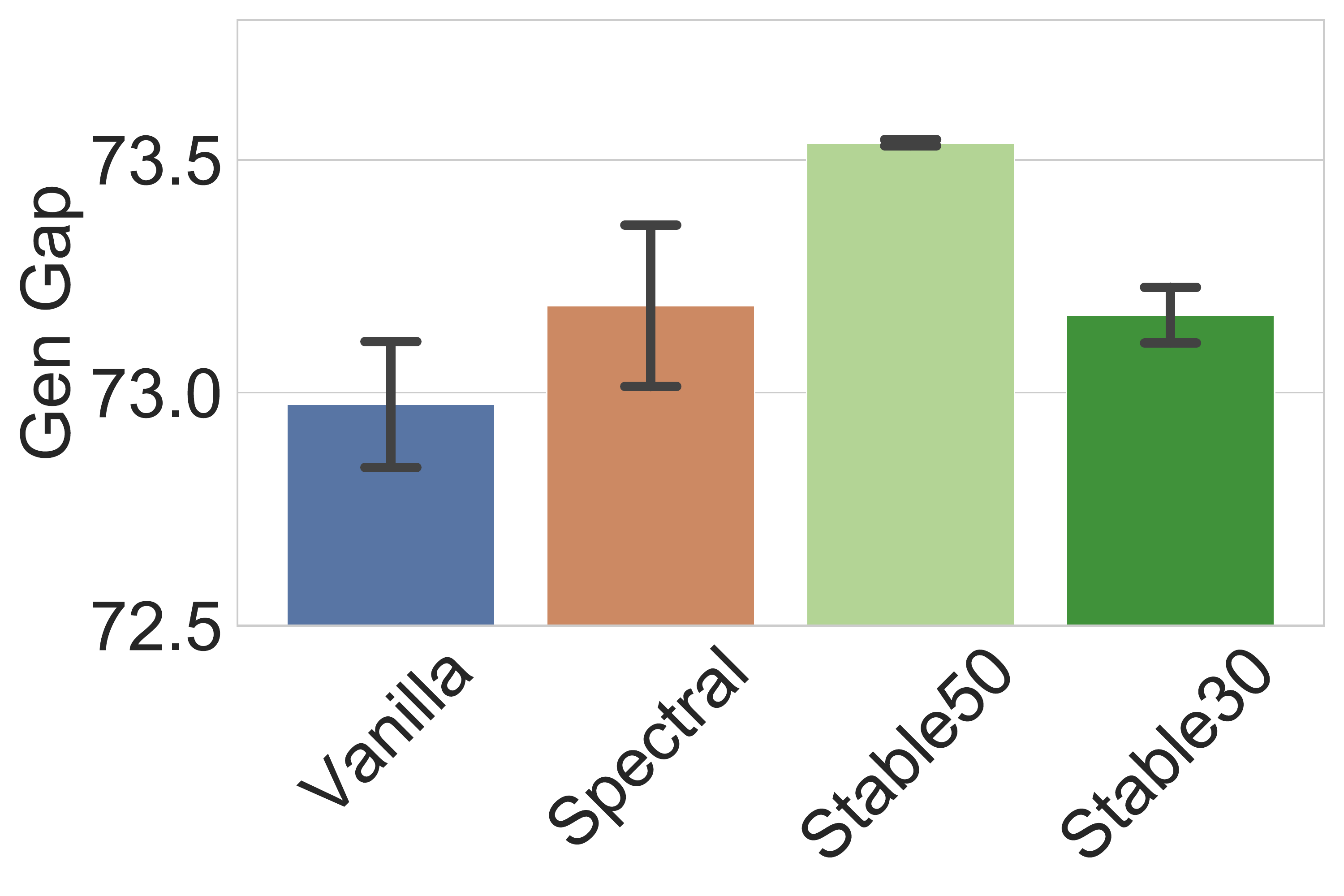
    \subcaption{ VGG-19}
  \end{subfigure}
  \caption[Test error on CIFAR100 with early stopping.]{Test accuracies on CIFAR100 for clean data using a stopping
    criterion based on train accuracy. Higher is better.}
  \label{fig:test-acc-stop}
\end{figure}

\begin{figure}[t]
  \centering\small
  \begin{subfigure}[!t]{0.24\linewidth}
    \def\svgwidth{0.98\linewidth}
    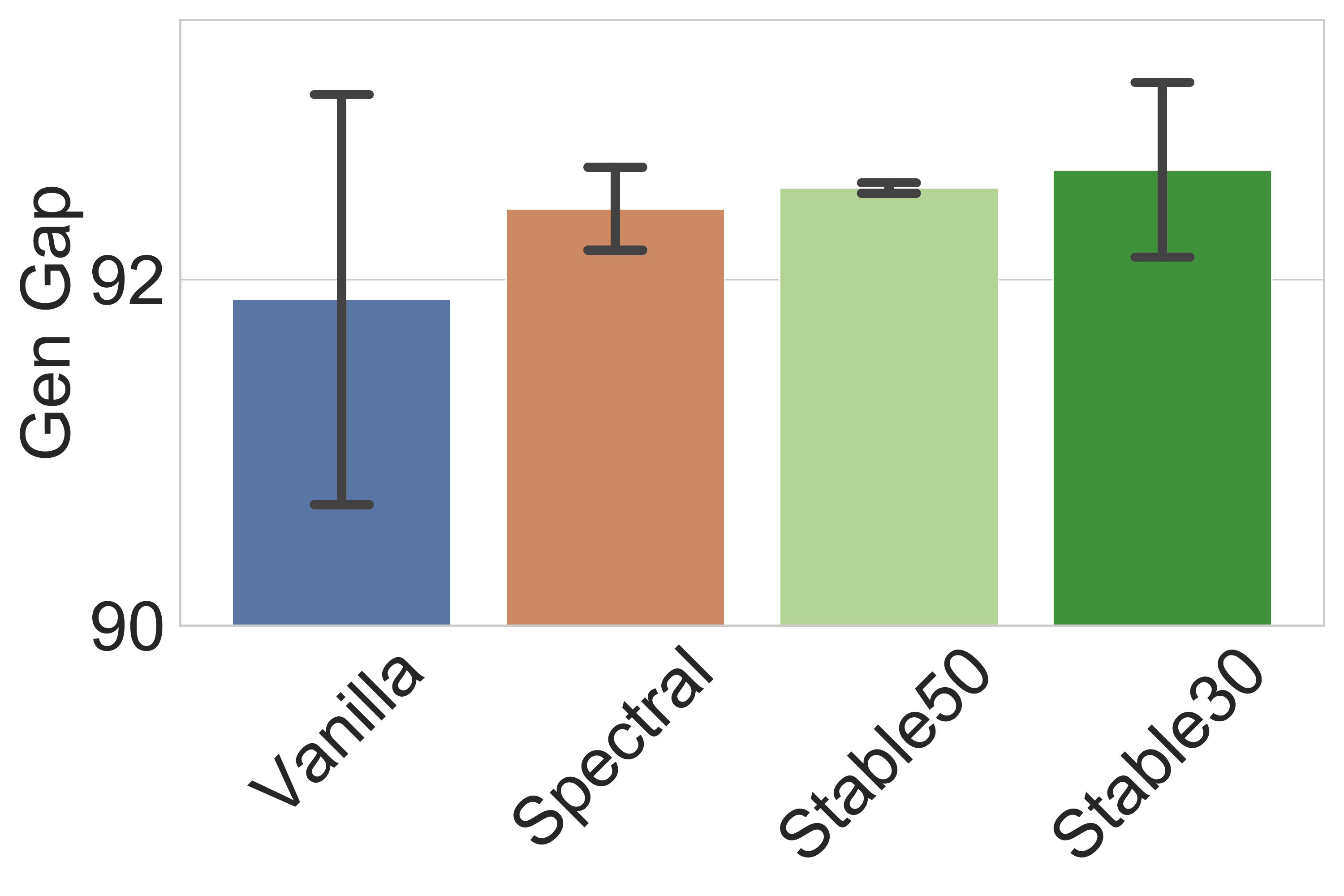
    \subcaption{ResNet110}
  \end{subfigure}
  \begin{subfigure}[!t]{0.24\linewidth}
    \def\svgwidth{0.98\linewidth}
    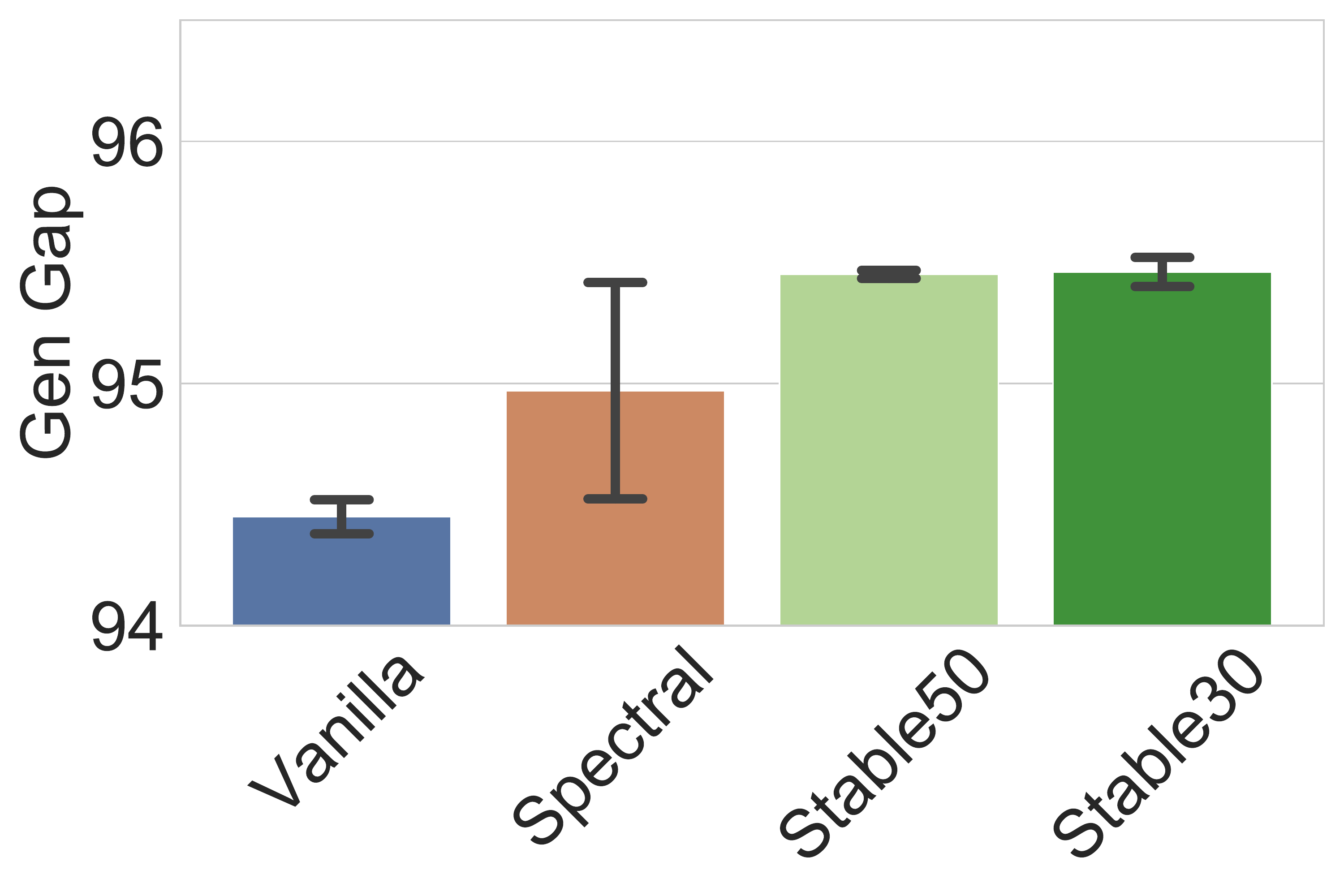
    \subcaption{WideResNet-28}
  \end{subfigure}
   \begin{subfigure}[!t]{0.24\linewidth}
    \def\svgwidth{0.98\linewidth}
    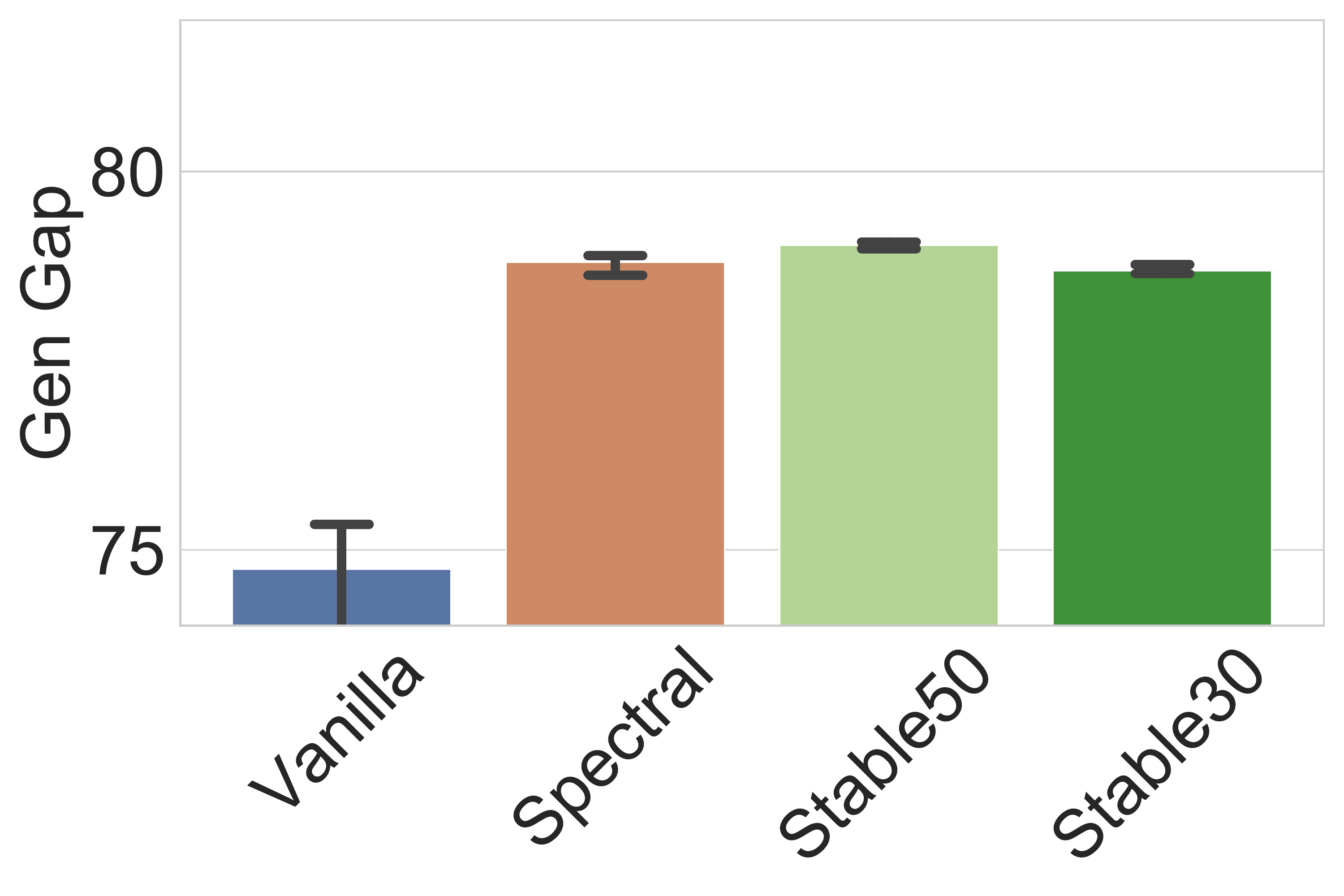
    \subcaption{ Alexnet}
  \end{subfigure}
  \begin{subfigure}[!t]{0.24\linewidth}
    \def\svgwidth{0.98\linewidth}
    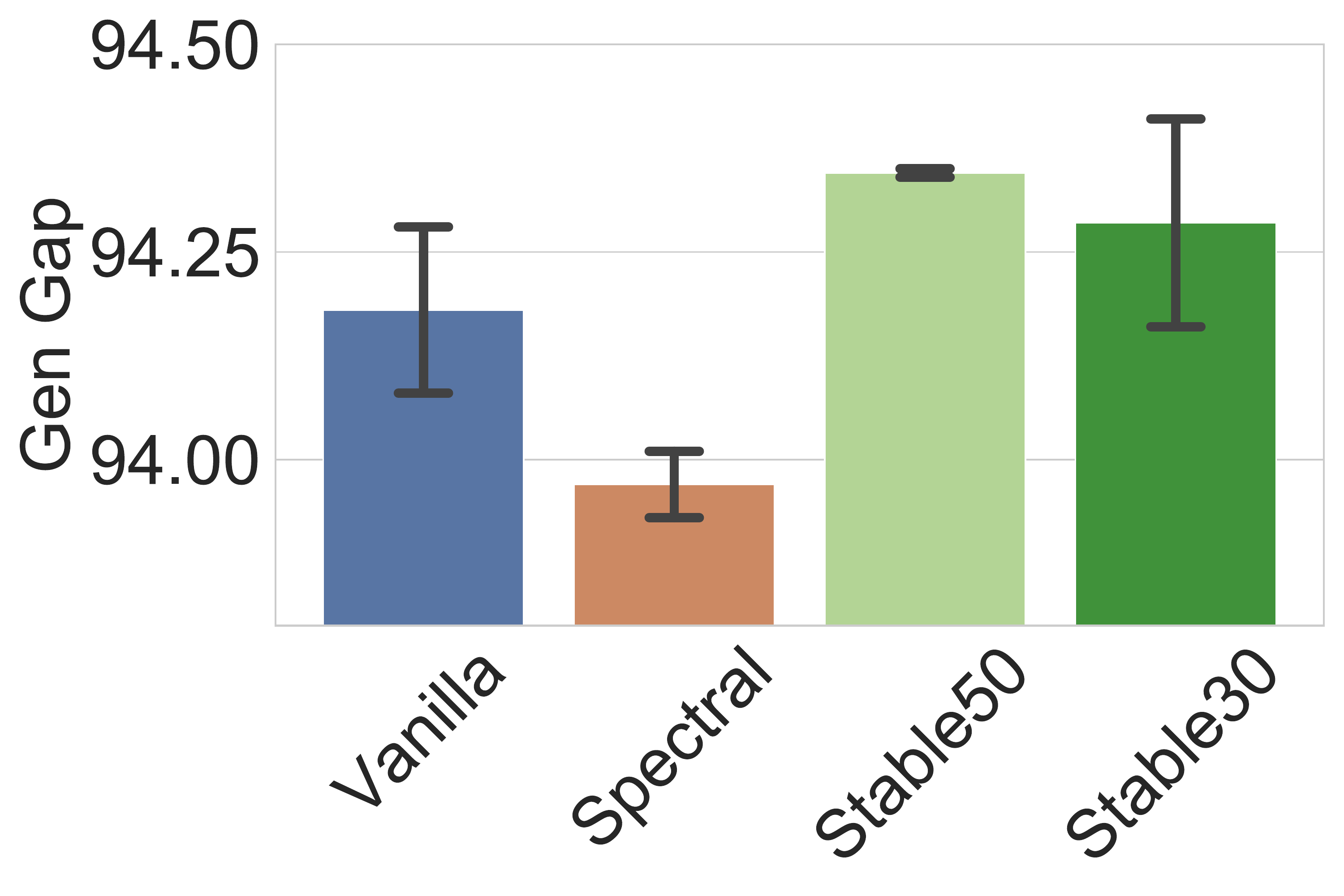
    \subcaption{ Densenet-100}
  \end{subfigure}
  \caption[Test Error on CIFAR10 with early stopping]{Test accuracies on CIFAR10 for clean data using a stopping
    criterion based on train accuracy. Higher is better.}
  \label{fig:test-acc-stop-c10}
\end{figure}

\subsection*{Non-generalisable settings}

So far, we have looked at learning with hyper-parameters like learning rate and
weight decay set to values commonly used in training of deep neural networks.
However, these hyper-parameters have been adjusted through years of practice.
This section looks specifically at  settings that are known to be highly
non-generalisable --- {\em low learning rate and without weight decay}.

First, we look at a WideResNet-28-10 trained with SRN, SN, and vanilla methods
but with a low learning rate of $0.01$ and a small weight decay of $5\times
10^{-4}$ on randomly labelled CIFAR100 for $50$ epochs. The results, shown
in~\cref{tab:low_lr_wrn}, supports our hypothesis that SRN is more robust to
random noise than SN or vanilla methods even in this highly-non generalisable
setup.

\begin{table}[!htb]
  \centering\small
  \begin{tabular}{||c|c|c||}\hline
    Stable-30&Spectral&Vanilla\\\hline
    $\mathbf{29.04}$& $17.24 $ & $1.22$ \\\hline
  \end{tabular}
  \caption[Shattering experiment with low-learning rate]{Training error for WideResNet-28-10 on CIFAR100 with randomised
    labels, low lr$=0.01$, and with weight decay.~(Higher
    is better.)}
  \label{tab:low_lr_wrn}
\end{table}

Secondly, we see whether weight decay provides a further edge to generalisation
when used in combination with SRN and SN. As shown in~\cref{tab:low_lr_rnet},
\gls{srn} consistently achieves lower generalisation error~(by achieving a low
train error) both in the presence and the absence of weight
decay.~\Cref{tab:clean-c100-lowlr} shows it also achieves a higher clean test
accuracy for this setting.  Thus~\Gls{srn} generalises better even with a low
learning rate and is further benefitted by regularisers like weight decay.

\begin{table}[!htb]
  \centering
  \begin{tabular}{l@{\quad}c@{\quad}c@{\quad}c@{\quad}c@{\quad}}\toprule
    &SRN-50&SRN-30&Spectral (SN) &Vanilla\\\midrule
    WD & $\mathbf{12.02 \pm 1.77}$ &$\mathbf{11.87 \pm 0.57}$ & $\mathbf{11.13 \pm 2.56}$ & $\mathbf{10.56
    \pm 2.32}$ \\
    w/o WD & $17.71 \pm 2.30$ &$19.04 \pm 4.53$ & $17.22 \pm 1.94$ & $13.49
    \pm 1.93$ \\\bottomrule
  \end{tabular}
  \caption[Shattering experiment with non-generalisable settings on CIFAR100]{{\bf Highly non-generalisable setting}. Training error
    for ResNet-110 on CIFAR100 with randomised labels, low lr$=0.01$,
    and with and without weight decay.~(Higher is better.)}
  \label{tab:low_lr_rnet}
\end{table}

\begin{table}[!htb]
  \centering
  \begin{tabular}{ccccc}\toprule
    & Vanilla & Spectral & Stable-50 & Stable-30 \\\midrule
    W/o WD & $69.2 \pm 0.5$ & $69\pm 0.1$ & $69.1 \pm 0.85$ & $69.3 \pm 0.4$ \\
    With WD & $\mathbf{70.4 \pm 0.3}$ & $\mathbf{71.35 \pm 0.25}$ & $\mathbf{70.6 \pm 0.1}$ & $\mathbf{70.6 \pm 0.1}$ \\\bottomrule
  \end{tabular}
  \caption[Clean test accuracy  for non-generalisable settings on CIFAR10.]{Clean test accuracy on CIFAR10. The learning configuration
    corresponds to the non-generalisable settings with a high learning
    rate.}
    \label{tab:clean-c100-lowlr}
\end{table}

\subsection{Empirical evaluation of generalisation behaviour}
When all the factors in training~(eg. architecture, dataset, optimiser, among
others) are fixed, and the only variability is in the normalization~(i.e. in SRN
vs SN vs Vanilla), the generalisation error can be written as $
\abs{\mathrm{Train\ Err} - \mathrm{Test\ Err}} \le
\tildeO{\sqrt{\nicefrac{C_{\mathrm{alg}}}{m}}}$ where $\tildeO{\cdot}$ ignores
the logarithmic terms, $m$ is the number of samples in the dataset, and
$C_{\mathrm{alg}}$ denotes a measure of {\em sample complexity} for a given
algorithm i.e. SRN vs SN vs Vanilla. The lower the value of $C_{\mathrm{alg}}$,
the better is the generalisation. This is reminiscent of the basic theorem of
generalisation previously discussed in~\Cref{thm:basic-gen-thm}. In this
section, we measure $C_{\mathrm{alg}}$ as a proxy for measuring generalisation.

First, we restate some of the concepts we had previously discussed
in~\Cref{sec:marg-based-gener-main}. The margin of a network, defined
in~\Cref{defn:margin} at a data point measures the gap in the confidence of the
network between the correct label and the other labels. In the rest of the
section, we will treat $\gamma$ as a random variable depending on the random
variables $\cX$ and $\cY$.
\margin*
We also restate the expressions for $C_{\mathrm{alg}}$ that we used previously in~\Cref{thm:spec-norm-marg-main} and~\Cref{eq:stable-spec-compl}. In addition, we define a new complexity measure from~\citet{wei2019}.

\begin{itemize}
  \item \textbf{Spec-Fro:}
    $\dfrac{\prod_{i=1}^L\norm{\vec{W}_i}_2^2\sum_{i=1}^L\srank{\vec{W}_i}}{\gamma^2}$~\citep{neyshabur2018a}.
    The two quantities used to normalize the margin~($\gamma$) are the product
    of spectral norm i.e. $\prod_{i=1}^L\norm{\vec{W}_i}_2^2$~(or worst case
    lipschitzness) and the sum of stable rank {\em i.e.},
    $\sum_{i=1}^L\srank{\vec{W}_i}$~(or an approximate parameter count like rank
    of a matrix).
  \item \textbf{Spec-L1:} $\dfrac{\prod_{i=1}^L\norm{\vec{W}_i}_2^2
      \br{\sum_{i=1}^L\frac{\norm{\vec{W}_i}_{2,1}^{\nicefrac{2}{3}}}{\norm{\vec{W}_i}_2^{\nicefrac{2}{3}}}}^3}{\gamma^2}$,
  where $\norm{.}_{2,1}$ is the matrix 2-1 norm. As showed
  by~\citet{bartlett2017spectrally}, Spec-L1 is the spectrally
  normalized margin, and unlike just the margin, is a good indicator of the
  generalisation properties of a network. 
  \item \textbf{Jac-Norm:}
    $\sum_{i=1}^L\dfrac{\norm{\vec{h}_i}_2\norm{\vec{J}_i}_2}{\gamma}$~\citep{wei2019},
    where  $\vec{h}_i$ is the $i^{\it th}$ hidden layer and $\vec{J}_i
    =\frac{\partial \gamma}{\partial h_i}$ {\em i.e.}, the Jacobian of the
    margin with respect to the $i^{\it th}$ hidden layer (thus, a vector). Note,
    Jac-Norm depends on the norm of the Jacobian (local empirical \Gls{lip}) and
    the norm of the hidden layers, which are additional data-dependent terms
    compared to Spec-Fro and Spec-L1.
  \end{itemize}

\begin{figure}[t]
  \centering
  \begin{subfigure}[t]{1.0\linewidth}
    \centering
  \begin{subfigure}[t]{0.32\linewidth}
    \def\svgwidth{0.98\linewidth} 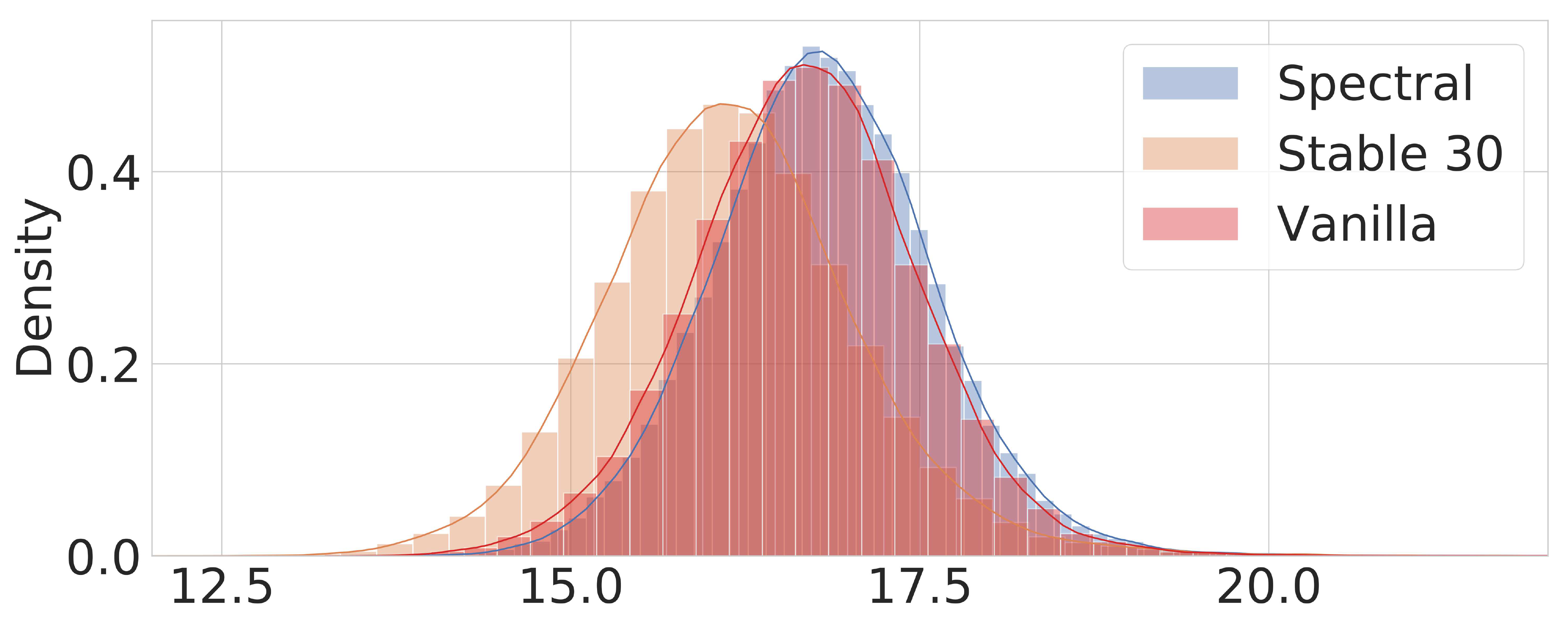
    \subcaption{R110-Jac-Norm}\label{fig:r110-jac-comp}
  \end{subfigure}\begin{subfigure}[t]{0.32\linewidth}
    \def\svgwidth{0.98\linewidth} 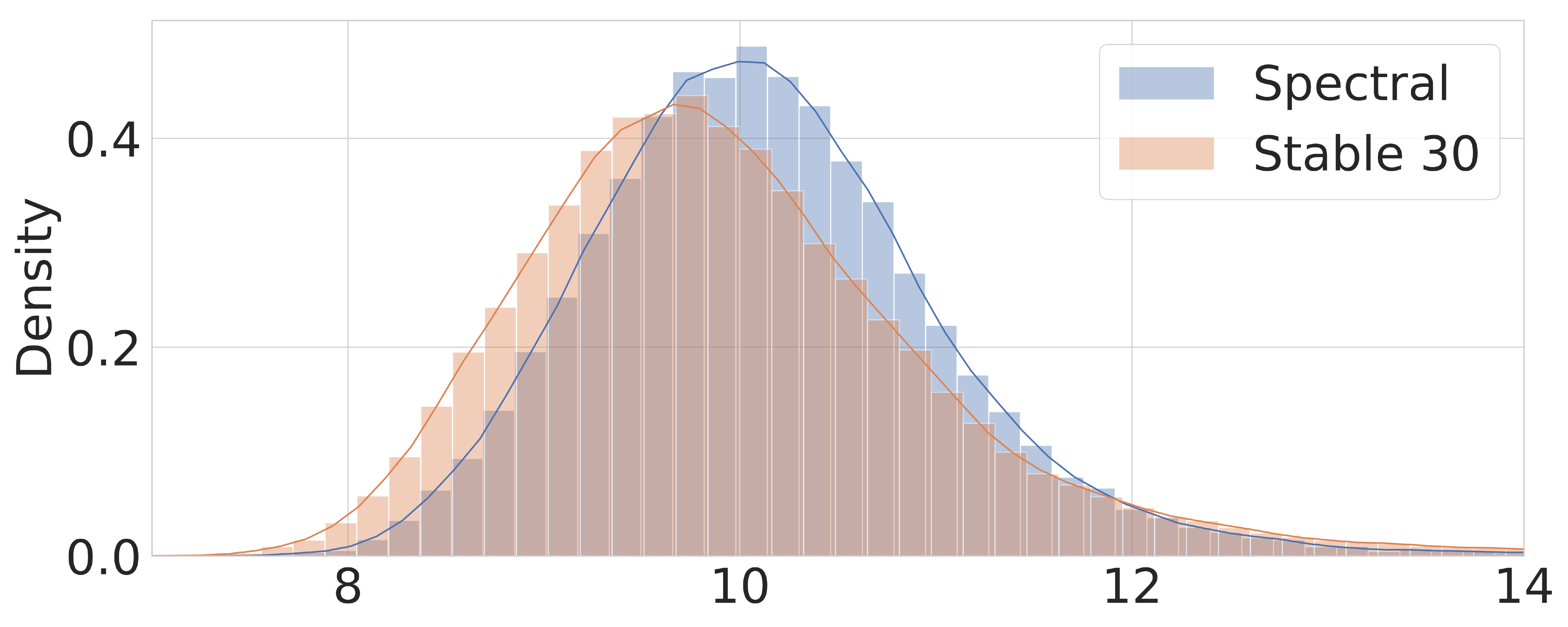
    \subcaption{R110-Spec-$L_1$}\label{fig:r110-spec-l1-comp}
  \end{subfigure}
  \begin{subfigure}[t]{0.32\linewidth}
    \def\svgwidth{0.98\textwidth}
    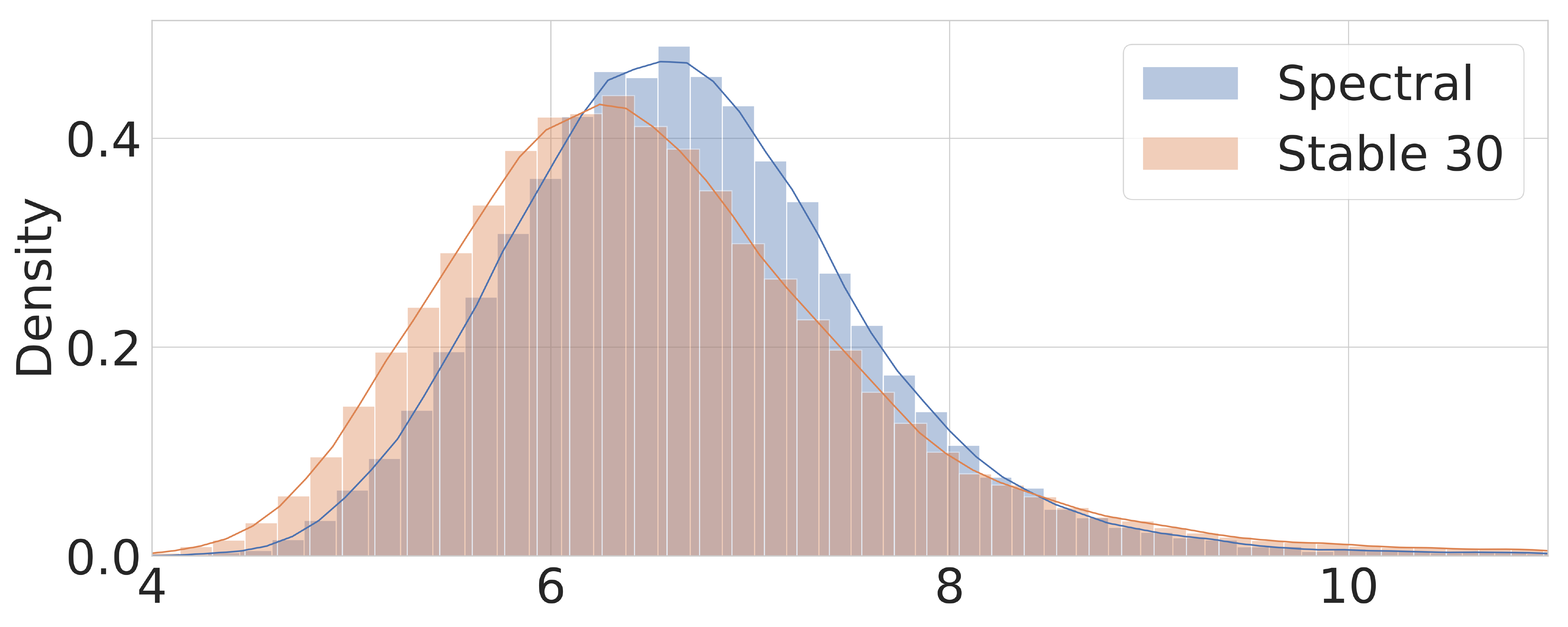
    \subcaption{R110-Spec-Fro}\label{fig:r110-spec-fro-comp}
  \end{subfigure}
\end{subfigure}
\begin{subfigure}[t]{1.0\linewidth}
    \centering
  \begin{subfigure}[t]{0.32\linewidth}
    \def\svgwidth{0.98\linewidth} 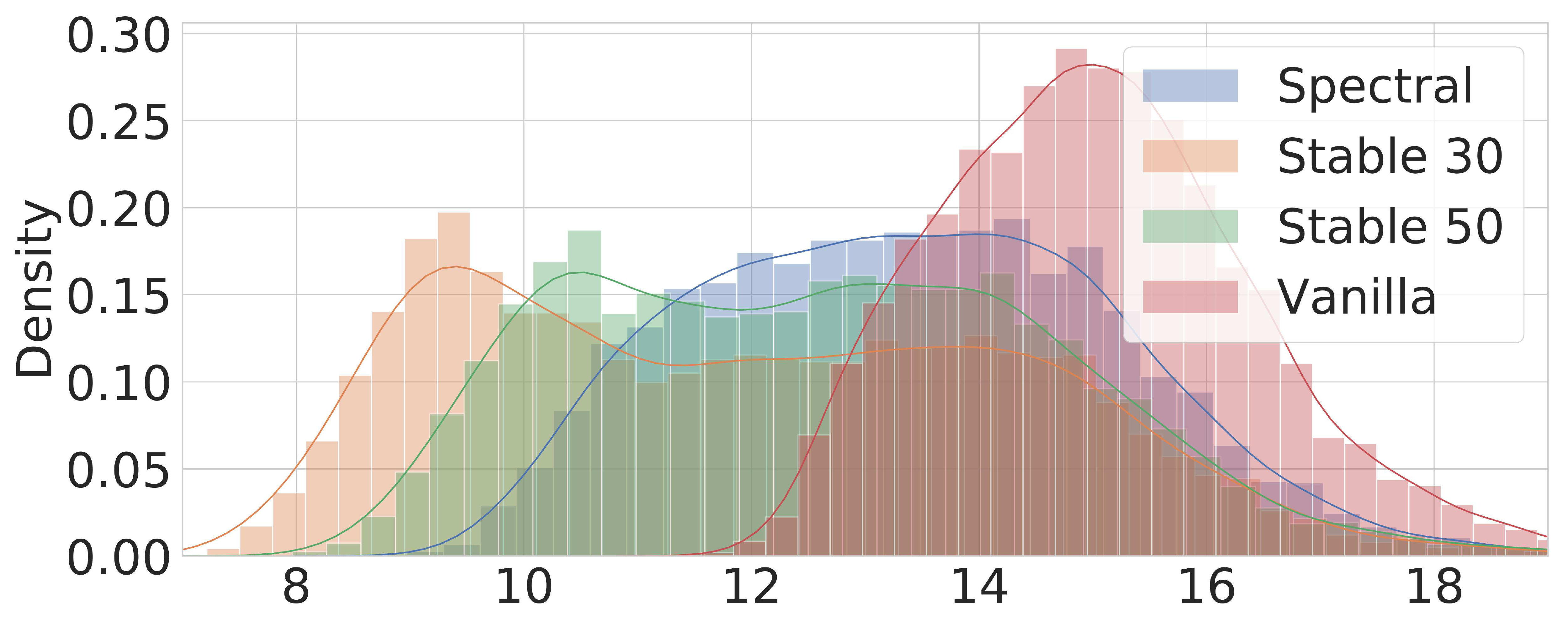
    \subcaption{WRN-Jac-Norm}\label{fig:wrn-jac-comp}
  \end{subfigure}
  \begin{subfigure}[t]{0.32\linewidth}
    \def\svgwidth{0.98\linewidth} 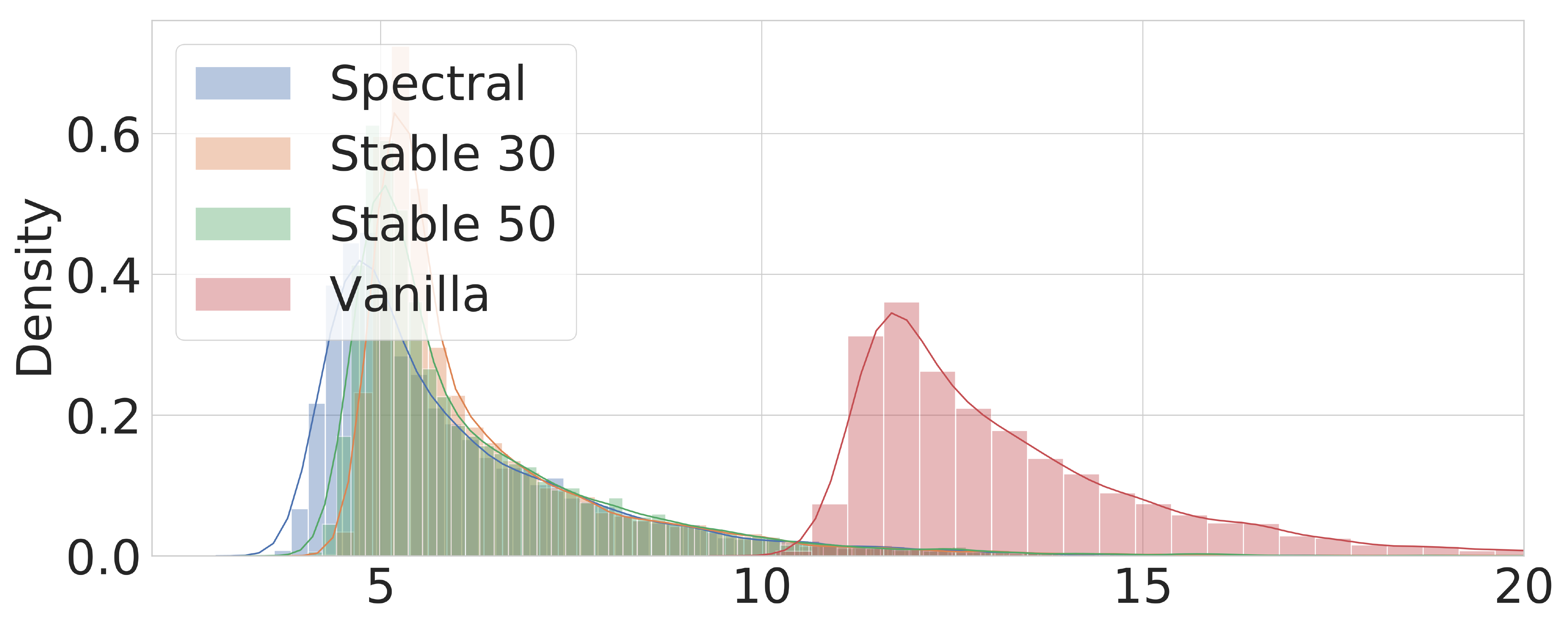
    \subcaption{WRN-Spec-$L_1$}\label{fig:wrn-spec-l1-comp}
  \end{subfigure}
  \begin{subfigure}[t]{0.32\linewidth}
    \def\svgwidth{0.98\textwidth}
    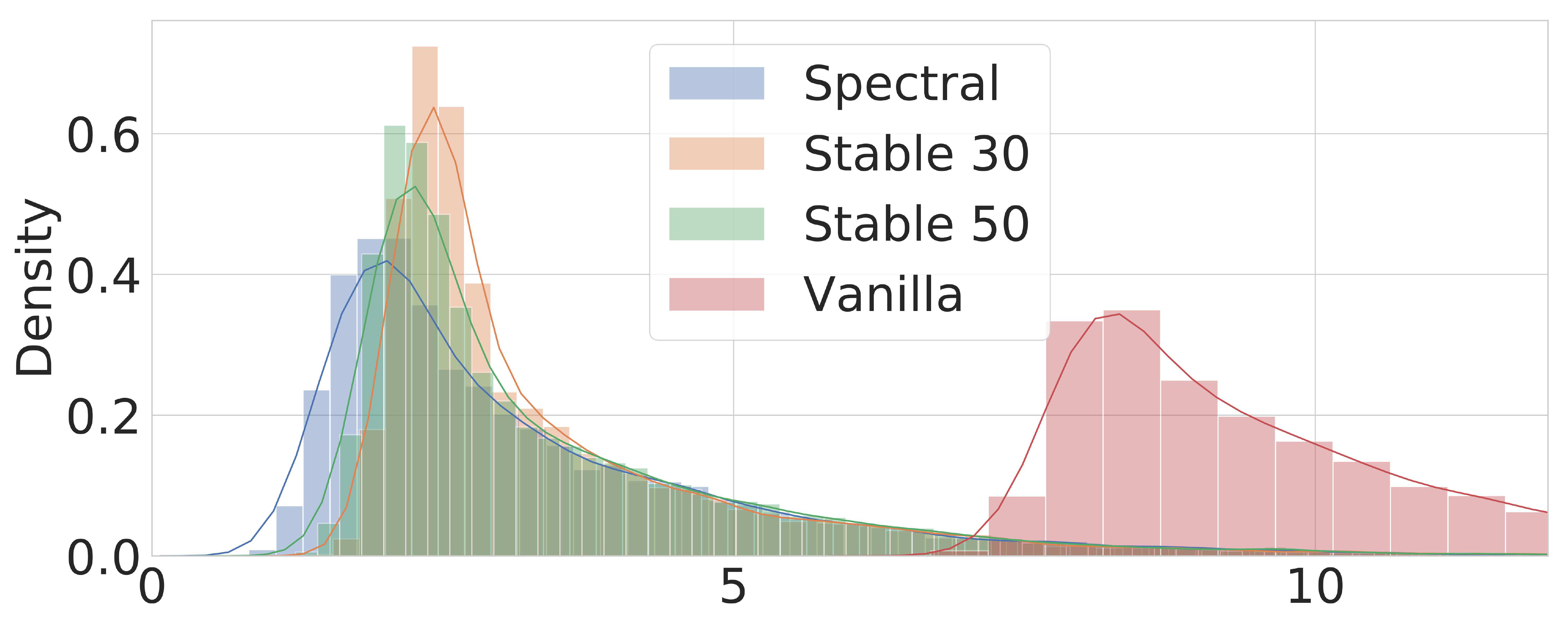
    \subcaption{WRN-Spec-Fro}\label{fig:wrn-spec-fro-comp}
  \end{subfigure}
\end{subfigure}
\begin{subfigure}[t]{1.0\linewidth}
    \centering
  \begin{subfigure}[t]{0.32\linewidth}
    \def\svgwidth{0.98\linewidth} 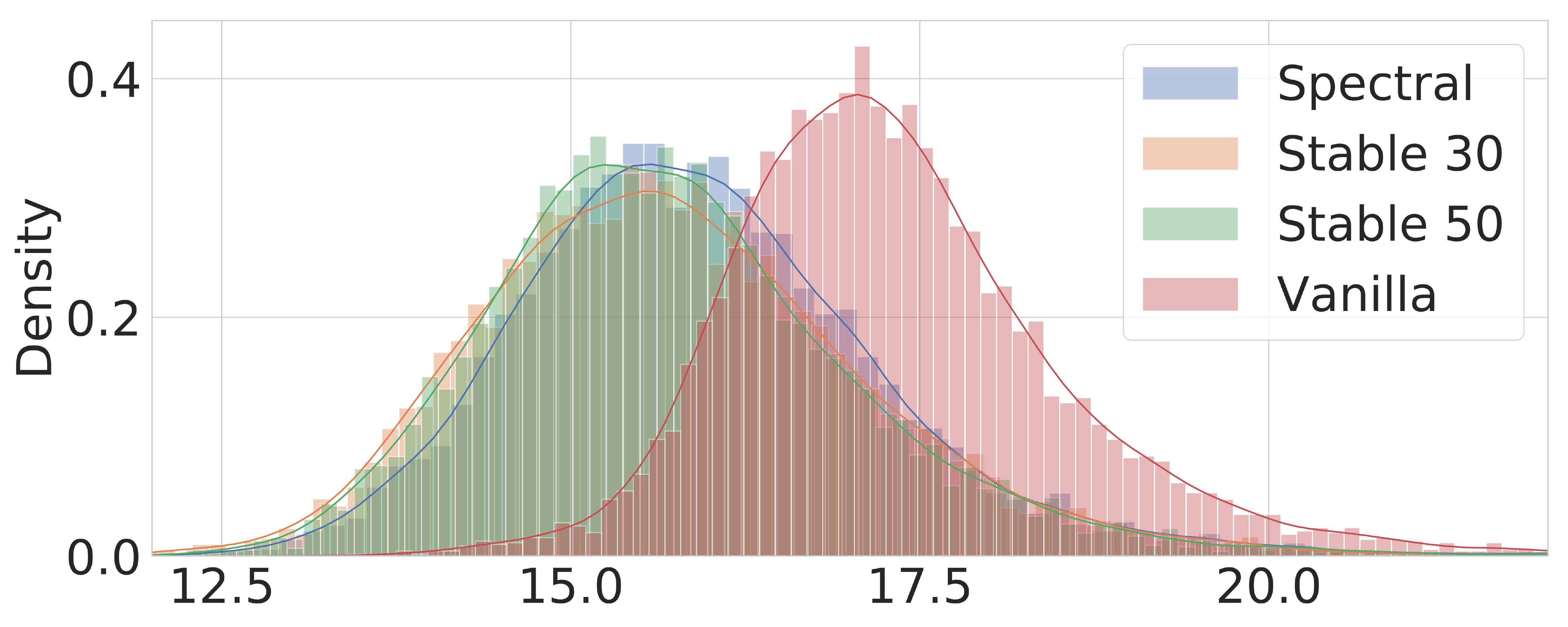
    \subcaption{D100-Jac-Norm}\label{fig:d100-jac-comp}
  \end{subfigure}
  \begin{subfigure}[t]{0.32\linewidth}
    \def\svgwidth{0.98\linewidth} 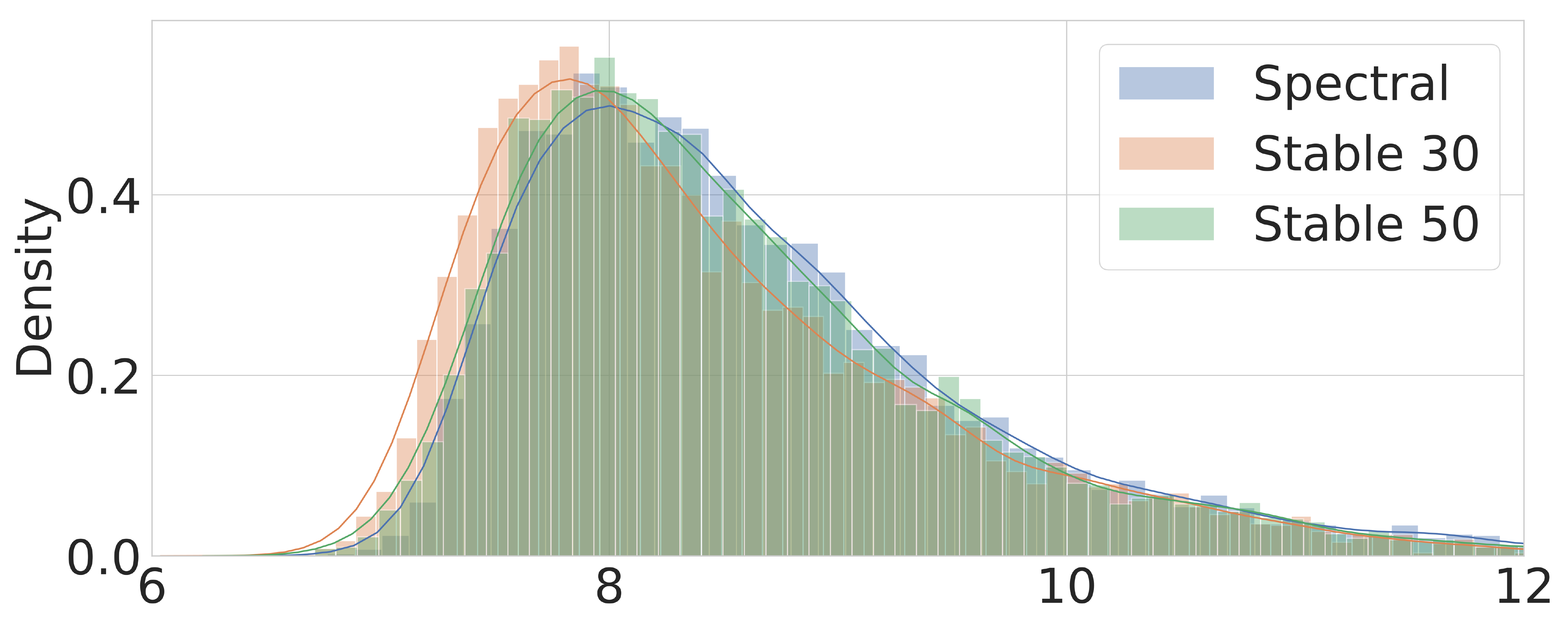
    \subcaption{D100-Spec-$L_1$}\label{fig:d100-spec-l1-comp}
  \end{subfigure}
  \begin{subfigure}[t]{0.32\linewidth}
    \def\svgwidth{0.98\textwidth}
    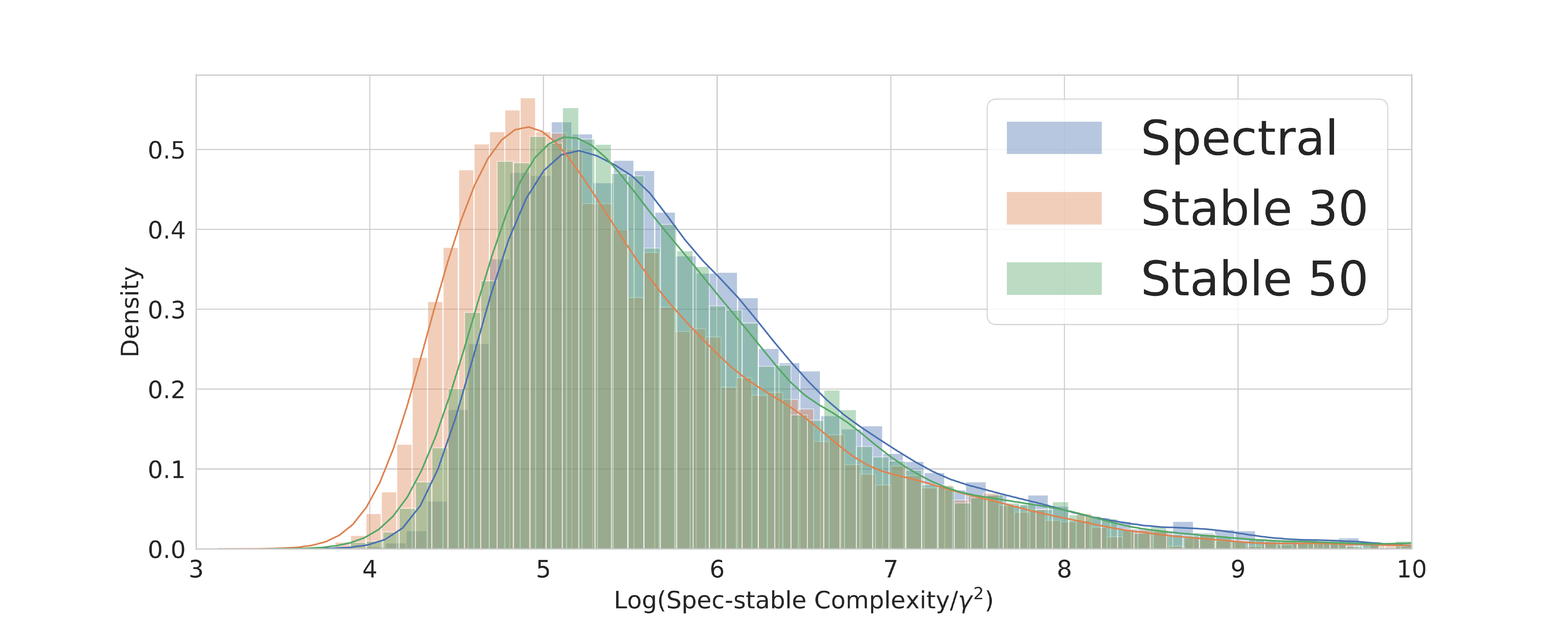
    \subcaption{D100-Spec-Fro}\label{fig:d100-spec-fro-comp}
  \end{subfigure}
\end{subfigure}
  \caption[Empirical comparison of generalisation bounds]{($\log$) Sample complexity ($C_{\mathrm{alg}}$) of
    ~ResNet-110~(\cref{fig:r110-jac-comp,fig:r110-spec-l1-comp,fig:r110-spec-fro-comp}),
    WideResNet-28-10~(\cref{fig:wrn-jac-comp,fig:wrn-spec-l1-comp,fig:wrn-spec-fro-comp}),
    and Densenet-100~(\cref{fig:d100-jac-comp,fig:d100-spec-l1-comp,fig:d100-spec-fro-comp})
quantified using the three measures discussed in this chapter. Left
is better. Vanilla is omitted
from~\cref{fig:r110-spec-l1-comp,fig:r110-spec-fro-comp,fig:d100-spec-l1-comp,fig:d100-spec-fro-comp}
as it is too far to the right. Also, in situations where SRN-50 and SN performed the same, we removed the histogram to avoid clutter.}
  \label{fig:compl}
\end{figure}

We evaluate the above-mentioned sample complexity measures on $10,000$ points
from the dataset and plot the distribution of its $\log$ using a histogram shown
in~\cref{fig:compl}. The more to the left the histogram is, the smaller is the
value of $C_{\mathrm{alg}}$ and thus the better is the generalisation capacity
of the network.

For better clarity, we report the $90$ percentile mark for each of these
histograms in ~\cref{tab:perc-compl}. As the plots and the table shows, both
\gls{srn} and \gls{sn} produces a much smaller complexity measure than a vanilla
network and in 7 out of the 9 cases, SRN is better than SN. The difference
between SRN and SN is much more significant in the case of Jac-Norm. As the
Jac-Norm depends on the empirical lipschitzness, it provides the empirical
validation of our arguments in~\Cref{sec:stable_rank_alg}.
%Additional details about these experiments are in~\cref{sec:values-compl}.
\begin{table}[!htb]\footnotesize
  \centering\renewcommand{\arraystretch}{1.1}
  \begin{tabular}[t]{C{3cm}C{3cm}c@{\quad}c@{\quad}c@{\quad}}\toprule
    Model & Algorithm & Jac-Norm & Spec-$L_1$ & Spec-Fro\\\hline
     \multirow{3}{*}{ResNet-110} & Vanilla & 17.7& $\infty$&$\infty$\\
         & Spectral (SN) & 17.8&10.8 &7.4\\
         & SRN-30 & {\bf 17.2}& {\bf 10.7} & {\bf 7.2}\\\cmidrule{2-5}
     \multirow{4}{*}{WideResNet-28-10} & Vanilla & 16.2& 14.60&11.18\\
         & Spectral (SN) & 16.13& 7.23&4.5\\
         & SRN-50 & 15.8& 7.3&4.5\\
         & SRN-30 & {\bf 15.7}& {\bf 7.20}&{\bf 4.4}\\\cmidrule{2-5}
     \multirow{4}{*}{Densenet-100} & Vanilla &19.2 &$\infty$ &$\infty$\\
         & Spectral (SN) &17.8 & 12.2&9.4\\
         & SRN-50 &17.6 &12 &9.2\\
         & SRN-30 & {\bf 17.7}&{\bf 11.8} &{\bf 9.0}\\\hline
  \end{tabular}
  \caption[Values of 90 percentile of $\log$ complexity measures
  from~\cref{fig:compl}]{Values of 90 percentile of $\log$ complexity measures
    from~\cref{fig:compl}. Here $\infty$ refers to the situations
    where the product of spectral norm blows up. This is the case in deep networks like ResNet-110 and Densenet-100 where the absence of spectral normalization (Vanilla)
    allows the product of spectral norm to grow arbitrarily large with an increasing number of layers. Lower is better.}
  \label{tab:perc-compl}
\end{table}
{\em The above experiments indicate that SRN, while providing enough capacity for the standard classification task, is remarkably less prone to memorisation and provides improved generalisation.}

%%% Local Variables:
%%% mode: latex
%%% TeX-master: "main"
%%% End:

  \section{Experiments on a generative modelling setup~
(SRN-GAN)}\label{sec:srn-exp}
\todo[color=blue]{Remove extra eLhist in App, stability of discriminator loss}
In GANs, there is a natural tension between the {\em capacity} and the {\em
generalisability} of the discriminator. The capacity ensures that if the
generated distribution and the data distribution are different, the
discriminator can distinguish them. At the same time, the discriminator has to
be generalisable, implying, the hypothesis class of discriminators should be
simple enough to ensure that the discriminator cannot memorise the dataset.
Based on these arguments, we use SRN in the discriminator of GAN which we call
SRN-GAN, and compare it against SN-GAN, WGAN-GP, and orthonormal regularisation
based GAN (Ortho-GAN). 

\paragraph{GAN objective functions}
In the case of conditional GANs~\citep{Mirza2014}, we use the conditional batch
normalization~\citep{dumoulin2017learned} to condition the generator and the
projection discriminator~\citep{Miyato2018} to condition the discriminator. The
dimension of the latent variable for the generator is set to $128$ and is
sampled from a zero mean and unit variance gaussian distribution.  For training
the model, we use the hinge loss version of the adversarial
loss~\citep{Lim2017,Tran2017} in all experiments except the experiments with
WGAN-GP. The hinge loss version is chosen  as it has  been shown to consistently
give better performance~\citep{Zhang2018, miyato2018spectral}. For training the
WGAN-GP model, we use the original loss function as described
in~\citet{Gulrajani2017}.

\paragraph{Constructing the empirical~\gls{Lip} histogram} Along with providing
results using evaluation metrics such as Inception score
(IS)~\citep{salimans2016improved} , FID~\citep{heusel2017gans}, and Neural
divergence score (ND)~\citep{gulrajani2018towards}, we use histograms of the
empirical \Gls{lip} constant, {\em referred to as eLhist} from now onwards, for
the purpose of analyses. For a given trained GAN (unconditional), we create
$2,000$ pairs of samples, where each pair $(\vec{x}_i, \vec{x}_j)$ consists of
$\vec{x}_i$ (randomly sampled from the `real' dataset) and $\vec{x}_j$ (randomly
sampled from the generator). Each pair is then passed through the discriminator
to compute the empirical lipschitzness based on the pair
$\nicefrac{\norm{f(\vec{x}_i) - f(\vec{x}_j)}_2}{\norm{\vec{x}_i -
\vec{x}_j}_2}$, which we then use to create the histogram. %
In the conditional setting, we  sample a class from a discrete
uniform distribution and then follow the same
approach as described for the unconditional setting.
Henceforth, we will use the word {\em histogram} as synonymous to
{\em the histogram of the empirical \Gls{lip} constant}.

\footnotetext[1]{Results are taken from ~\citet{miyato2018spectral}.
The rest of the results in the tables are generated by us.}

\begin{table}\centering
  \begin{tabular}{c@{\quad}cccc@{}}
    \toprule
    & Algorithm &  Inception Score & FID & Intra-FID\\
    \midrule
    \multirow{5}{*}{\rotatebox[origin=c]{90}{\footnotesize Uncond.}}
    & Orthonormal\footnotemark[1]  & $7.92\pm .04$ &$23.8$&-\\ 
    & WGAN-GP & $7.86\pm .07$ &$21.7$& - \\
    & SN-GAN\footnotemark[1] & $8.22\pm .04$ &$20.67$& -\\
    & SRN-70-GAN & $\mathbf{8.53}\pm 0.04$& $\mathbf{19.83}$&- \\
    & SRN-50-GAN & $8.33\pm 0.06$& $\mathbf{19.57}$&- \\
    \cmidrule{2-5}
    \multirow{3}{*}{\rotatebox[origin=c]{90}{\footnotesize Cond.}}
    & SN-GAN & $8.71\pm .04$ &$16.04$\footnote{This is different from
    what is reported in the original paper.}& $26.24$\\ 
    &SRN-$70$-GAN & $\mathbf{8.93}\pm 0.12$& $\mathbf{15.92}$ &
    $\mathbf{24.01}$\\
    &SRN-$50$-GAN & $ 8.76\pm 0.09$& $16.89$& $27.3$\\
   \bottomrule
  \end{tabular}
  \caption{Inception and FID score on CIFAR10.}
  \label{tbl:comp_uncond_model1}
\end{table}

\subsection{Effect of SRN on Inception Score, FID, and Neural Divergence}

Inception Score~(IS) of SRN-GANs, SN-GAN, WGAN-GP, and GANs with Orthonormal
regularisations on CIFAR10 and CIFAR100 are reported
in~\cref{tbl:comp_uncond_model1,tab:inc_fid_cifar100} respectively. For
measuring the inception score, we generate $50,000$ samples, as is recommended
in~\citet{salimans2016improved}. For measuring FID, we use the same setting
as~\citet{miyato2018spectral} where we sample $10,000$ data points from the
training set and compare its statistics with that of $5,000$ generated samples.
In addition, we use a recent evaluation metric called Neural divergence
score~\citet{gulrajani2018towards} which is more robust to memorisation. The
exact set-up for the same is discussed below. In the case of conditional image
generation, we also measure Intra-FID~\citep{miyato2018spectral}, which is the
mean of the FID of the generator, when it is conditioned over different classes.
Let $\mathrm{FID}(\mathcal{G}, c)$ be the FID of the generator $\mathcal{G}$
when it is conditioned on the class $c \in \mathcal{C}$ (where $\mathcal{C}$ is
the set of classes), then, $\mathrm{Intra~FID}(\mathcal{G}) =
\frac{1}{|\mathcal{C}|}\mathrm{FID}(\mathcal{G}, c)$

For CIFAR10, we report results on both the conditional and the unconditional
settings and for CIFAR100, we report on the unconditional setting. The results
show that SRN-GAN consistently provides a better FID score and an extremely
competitive inception score on  both: CIFAR10~(both conditional and
unconditional setting) and CIFAR100~(unconditional setting).

In Table~\ref{tab:nn_dis_cifar100}, we compare the Neural Divergence~(ND) loss
on CIFAR10 and CelebA datasets. Note that ND has been looked at as a metric {\em
more robust to memorisation} than FID and IS in recent
works~\citep{gulrajani2018towards,arora2017gans}. The exact setting for
computing ND is discussed in~\cref{sec:expr-settings}. We essentially report the
loss incurred by a \textit{fresh} classifier trained to discriminate the
generator distribution and the data distribution. Thus higher the loss, the
better are the generated images. As evident from~\Cref{tab:nn_dis_cifar100},
SRN-GAN provides better ND scores on both datasets.

\begin{table}[t]
  \centering
  \begin{tabular}{@{}c@{}cc@{}}
    \toprule
   Model&IS & FID\\
   \midrule
    SN-GAN&$\mathbf{9.04}$&$23.2$\\
    SRN-GAN (Our)&$8.85$&$\mathbf{19.55}$\\
    \bottomrule
  \end{tabular}
  \caption{Inception and FID score on  CIFAR100.}
  \label{tab:inc_fid_cifar100}
\end{table}

\begin{table}[!htb]
  \centering
  \begin{tabular}{@{}c@{}cc@{}}
  \toprule
   Model&CIFAR10 & CelebA\\
   \midrule
    SN-GAN&$10.69$&$0.36$\\
    SRN-GAN (Our)&$\mathbf{11.97}$&$\mathbf{0.64}$\\
    \bottomrule
  \end{tabular}
  \caption[Neural Discriminator Loss on CelebA and CIFAR10]{Neural Discriminator Loss~(Higher the better).}
  \label{tab:nn_dis_cifar100}
\end{table}

\subsection{Effect of SRN on eLhist}

We construct  eLhist for unconditional GANs on CIFAR10
in~\cref{fig:lip_rank_stable}. As the plot shows, lowering the value of $c$
(aggressive reduction in the stable rank) shifts the histogram towards zero,
implying a lower empirical \Gls{lip} constant. This validates the arguments
provided in~\cref{sec:whyStable}. Lowering $c$ also improves the  inception
score. However, an extreme reduction in the stable rank ($c=0.1$) dramatically
collapses the histogram to zero and also drops the inception score
significantly. This is because, at $c=0.1$, the capacity of the discriminator is
reduced to the point that it is not able to learn to differentiate between the
real and the fake samples anymore.

\begin{figure}[!htb]
  \centering
\begin{subfigure}[c]{0.495\linewidth}
  \centering
  \def\svgwidth{0.99\columnwidth}
  \resizebox{0.95\textwidth}{!}{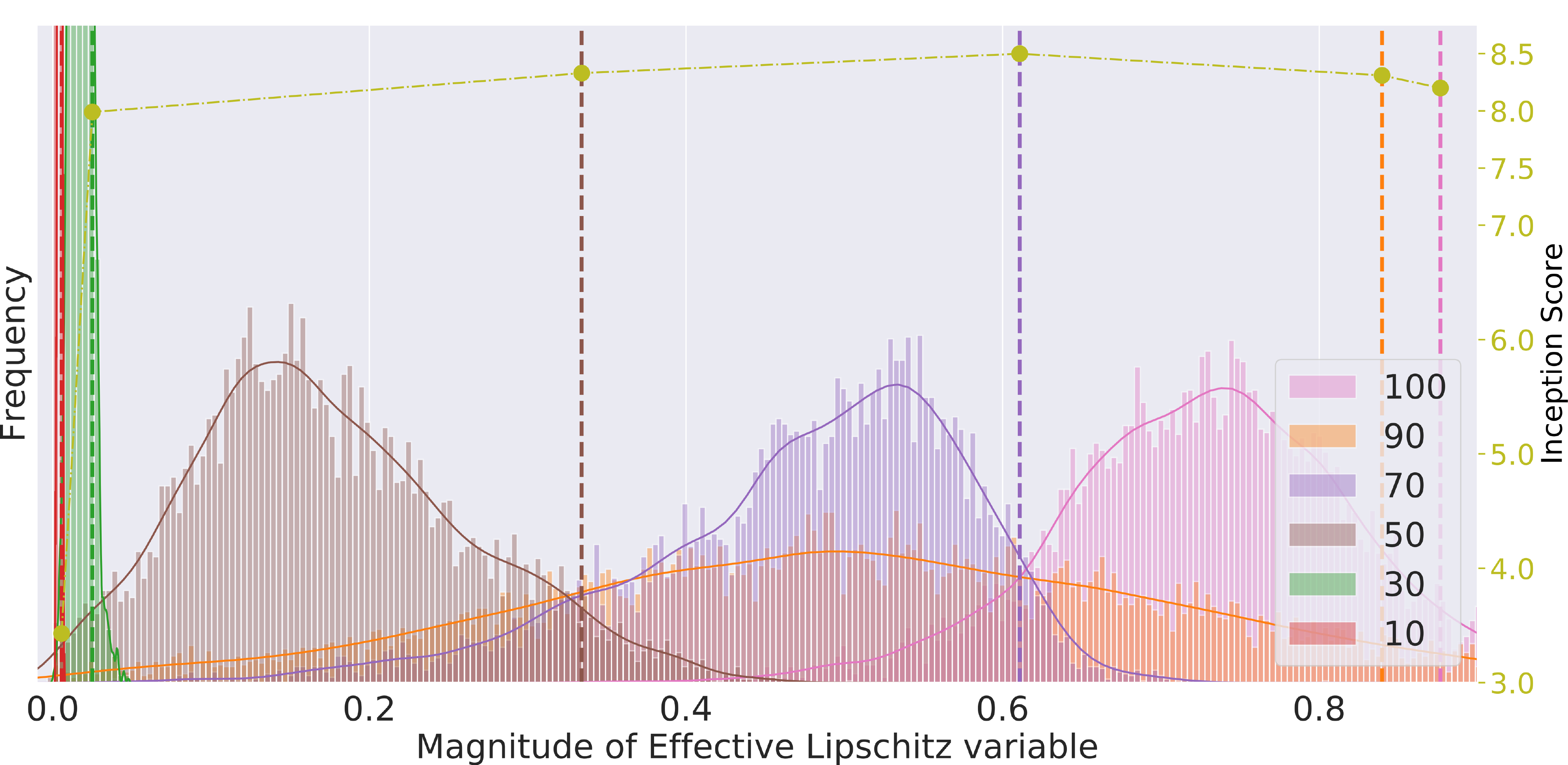} \caption{Unconditional GAN} \label{fig:lip_rank_stable}
\end{subfigure}
\begin{subfigure}[c]{0.495\linewidth}
  \centering
  \def\svgwidth{0.99\columnwidth}
  \resizebox{0.95\textwidth}{!}{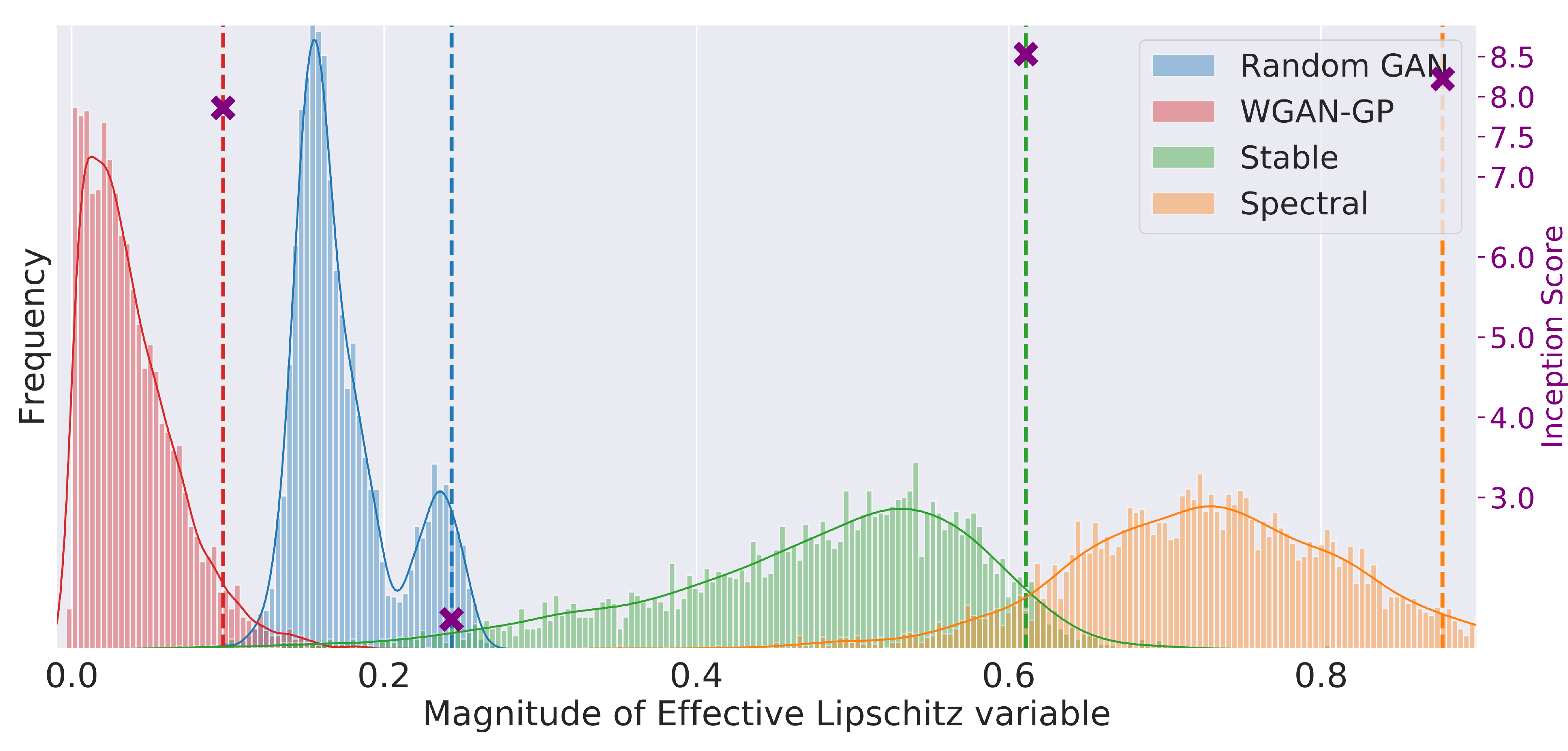}\caption{Unconditional GAN}  \label{fig:lip_rank_stable_uncond}
\end{subfigure}
\begin{subfigure}[c]{0.72\linewidth}
  \centering
  \def\svgwidth{0.99\columnwidth}
  \resizebox{0.99\textwidth}{!}{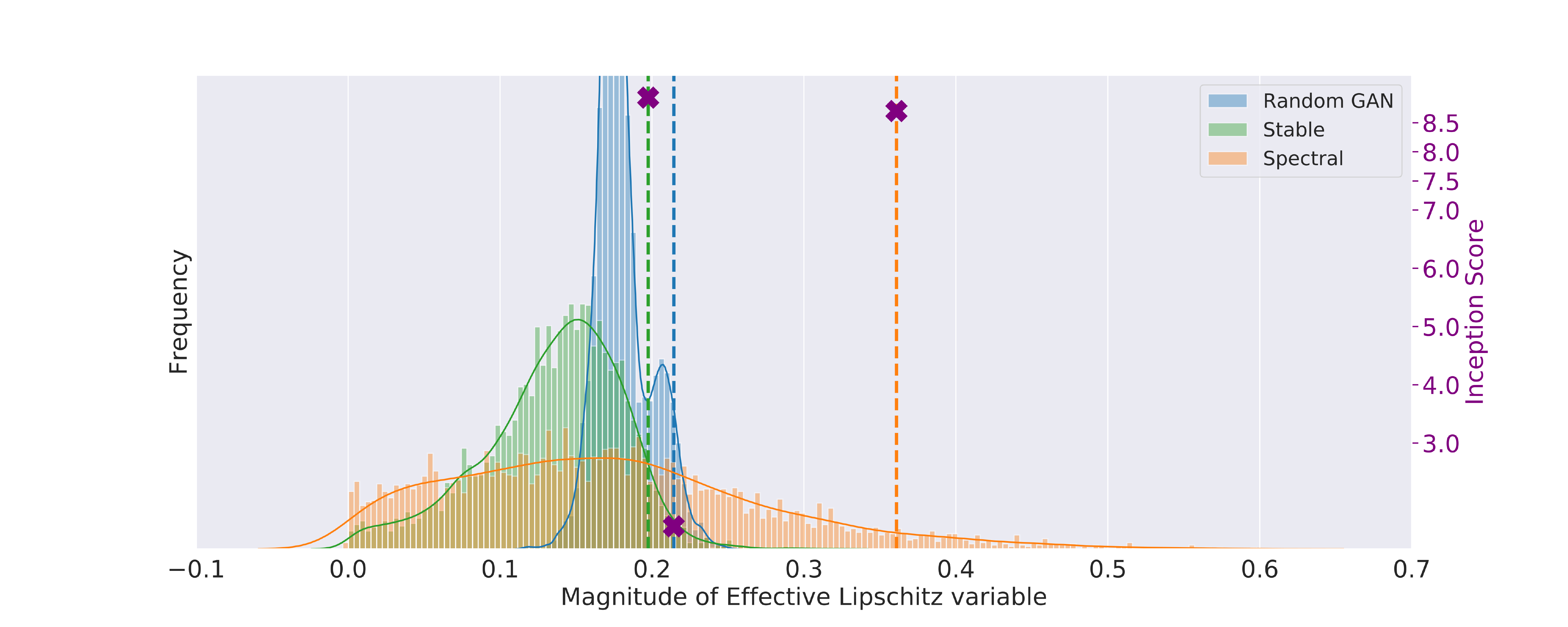}
  \caption{Conditional GAN with projection discriminator}\label{fig:lip_rank_stable_cond}
\end{subfigure}
\caption[eLhist of unconditional and conditional GANs on CIFAR10.]{{\bf eLhist}
  for unconditional and conditional GANs on CIFAR10. Dashed vertical lines
  represent  95{\it th} percentile. Solid circles and crosses represent the {\em
  inception score} for each histogram. Figure~\ref{fig:lip_rank_stable} shows
  SRN-GAN for different stable rank constraints (\eg $90$ implies $c=0.9$).
  Figure~\ref{fig:lip_rank_stable_uncond} compares various approaches for
  unconditional GANs.~\Cref{fig:lip_rank_stable_cond} compares various
  approaches for conditional GAN setting. Random-GAN represents random
  initialisation (no training). For SRN-GAN, we use $c=0.7$.}
\end{figure}

In addition, in~\cref{fig:lip_rank_stable_uncond}, we provide eLhist for
comparing different approaches. Random-GAN is a randomly initialised GAN with
the same architecture as SRN-GAN. As expected, Random-GAN has a low empirical
\Gls{lip} constant and an extremely poor inception score. Unsurprisingly,
WGAN-GP has a lower empirical~\gls{lip} constant than Random-GAN, due to its
explicit constraint on the \Gls{lip} constant, while providing a higher
inception score. On the other hand, SRN-GAN, by virtue of its softer constraints
on the \Gls{lip} constant, trades off a higher \Gls{lip} constant than WGAN-GP
for a better inception score---highlighting the flexibility provided by SRN. We
show results on the conditional GAN setup using the projection discriminator
from ~\citet{Miyato2018} in~\cref{fig:lip_rank_stable_cond}.

\begin{figure}[!h]
  \begin{subfigure}[t]{0.49\linewidth}
    \centering
  \def\svgwidth{0.99\columnwidth}
  \resizebox{0.95\textwidth}{!}{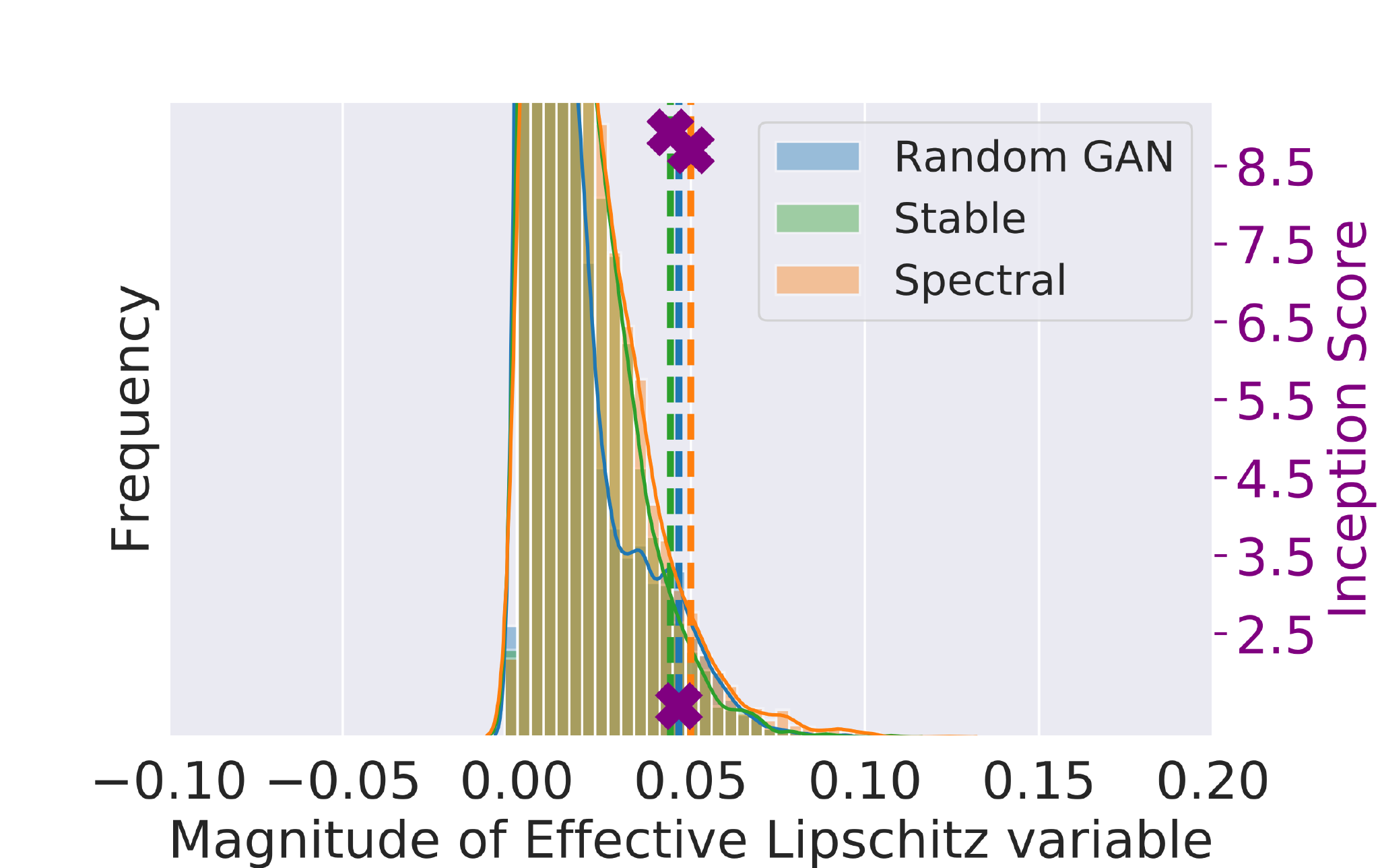}
  
  \caption{Conditional GAN  with projection discriminator.}  \label{fig:lip_rank_stable_cond_only_fake}
\end{subfigure}\hfill
\begin{subfigure}[t]{0.49\linewidth}
  \centering
  \def\svgwidth{0.99\columnwidth}
  \resizebox{0.95\textwidth}{!}{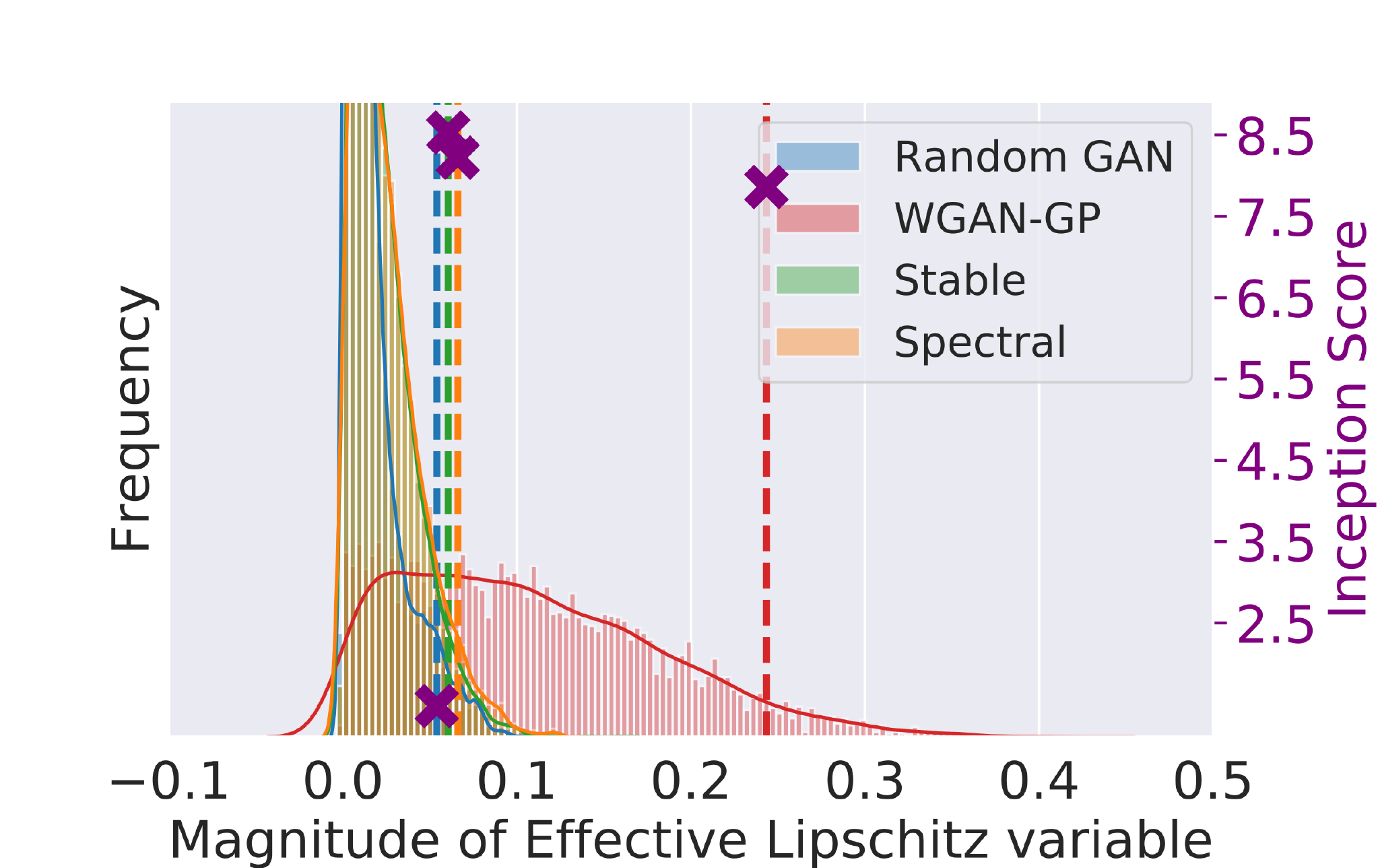}
  
  \caption{Unconditional GAN setting.} \label{fig:lip_rank_stable_uncond_only_fake}
\end{subfigure}
  \begin{subfigure}[!h]{0.49\linewidth}
    \centering
    \def\svgwidth{0.99\columnwidth}
    \resizebox{0.95\textwidth}{!}{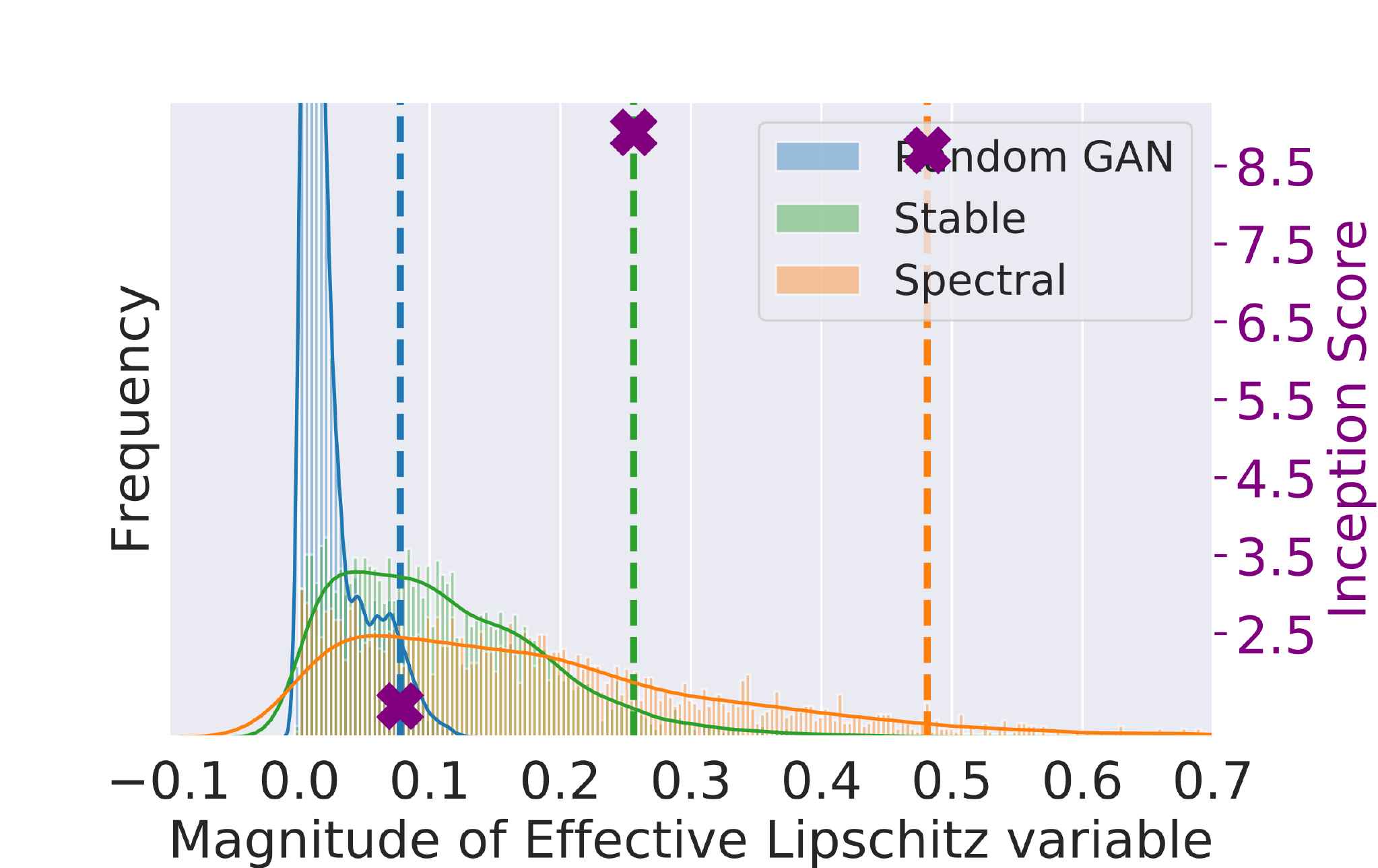}
    
    \caption{Conditional GAN with projection discriminator}  \label{fig:lip_rank_stable_cond_only_real}
  \end{subfigure}
  \begin{subfigure}[!h]{0.49\linewidth}
    \centering
    \def\svgwidth{0.99\columnwidth}
    \resizebox{0.95\textwidth}{!}{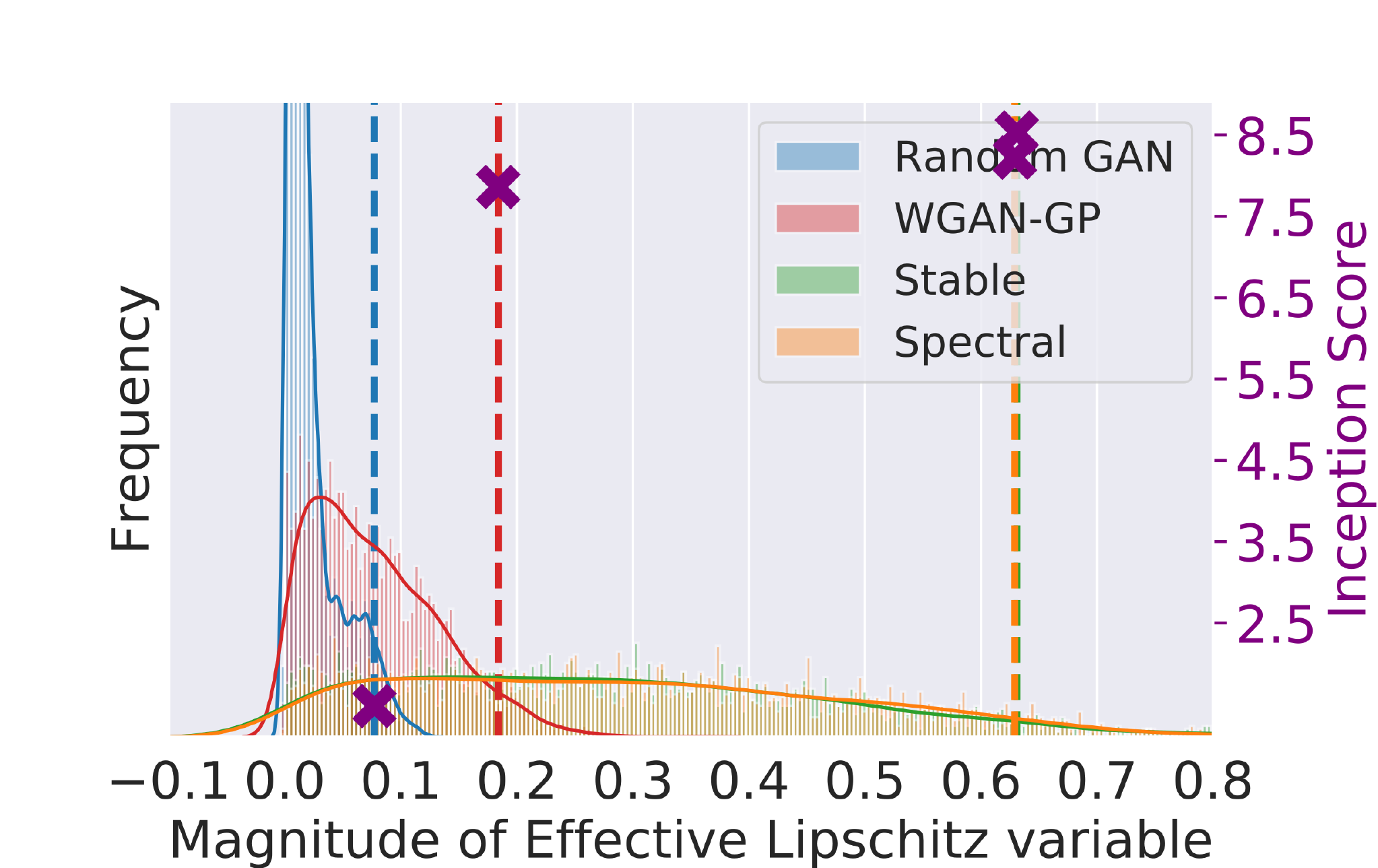}
     \caption{Unconditional GAN setting.}  \label{fig:lip_rank_stable_uncond_only_real_uncond}
  \end{subfigure}
  \caption[eLhist with pairs of reals/generated samples]{
  ~\Cref{fig:lip_rank_stable_cond_only_fake,fig:lip_rank_stable_uncond_only_fake} plots the {\bf  eLhist}  of the
  discriminator for pairs of samples drawn from the generator on CIFAR10.~\Cref{fig:lip_rank_stable_cond_only_real,fig:lip_rank_stable_uncond_only_real_uncond} plots the {\bf  eLhist}  of the
  discriminator for pairs of samples drawn from the real
  distribution on CIFAR10.} 
  \end{figure}

For the purpose of
analysis,~\cref{fig:lip_rank_stable_uncond_only_fake,fig:lip_rank_stable_uncond_only_real_uncond}
shows eLhist for pairs where each sample either comes from the true data or the
generator, and we observe a similar trend. To verify the same results hold in
the conditional setup, we show comparisons for~\gls{gan}s with projection
discriminator~\citep{Miyato2018}
in~\cref{fig:lip_rank_stable_cond_only_fake,fig:lip_rank_stable_cond_only_real},
and also observe a similar trend.

\chapter{Impact of Label Noise and Representation Learning on Adversarial Robustness}
\label{chap:causes_vul}
Despite being successful in a wide range of tasks e.g. in computer vision,
~\citep{KSH:2012,Zagoruyko2016,he2015delving,ren2015faster} and natural language
processing~\citep{graves2013speech,vaswani2017attention}, machine learning
algorithms have been shown to be highly vulnerable to small adversarial
perturbations that are otherwise imperceptible to the human
eye~\citep{Dalvi2004,Biggio2018,szegedy2013intriguing,
goodfellow2014explaining,Carlini2017,Papernot2016,mosaavi2016}. This
vulnerability poses serious security concerns when these models are deployed in
real-world tasks (cf.~\citep{Papernot2017,Schoenherr2018,Hendrycks2019,
li2019adversarial}). A large body of research has  been devoted to crafting
defences to protect neural networks from adversarial
attacks~(e.g.~\citep{goodfellow2014explaining,
Papernot2015,tramer2018ensemble,madry2018towards,Zhang2019}). However, as we
discussed in~\Cref{sec:robustness} these defences have usually been broken by
future attacks~\citep{athalye2018obfuscated, Tramer2020}. This arms race between
attacks and defences suggests that creating a truly robust model would require a
deeper understanding of the source of this vulnerability. 

Our goal in this chapter is not to propose new defences, but to provide better
answers to the question: what causes the adversarial vulnerability? In doing so,
we also try to understand how existing methods designed to achieve adversarial
robustness overcome some of these hurdles pointed out by our work.
In~\Cref{chap:stable_rank_main}, we discussed the memorisation of label noise in
the context of generalisation. In this chapter, we will look at the impact of
memorisation on adversarial vulnerability.

We identify two sources of adversarial vulnerability that, to the best of our
knowledge, have not been properly studied before: a) memorisation of label
noise, and b) \emph{improper} representation learning. 

\section{Theoretical result on the impact of memorising label noise}
\label{sec:overfit-theoretical-setting}

\begin{figure}[!htb]
	\centering
  	\begin{subfigure}[b]{0.4\linewidth}
		\def\svgwidth{0.99\linewidth}
    	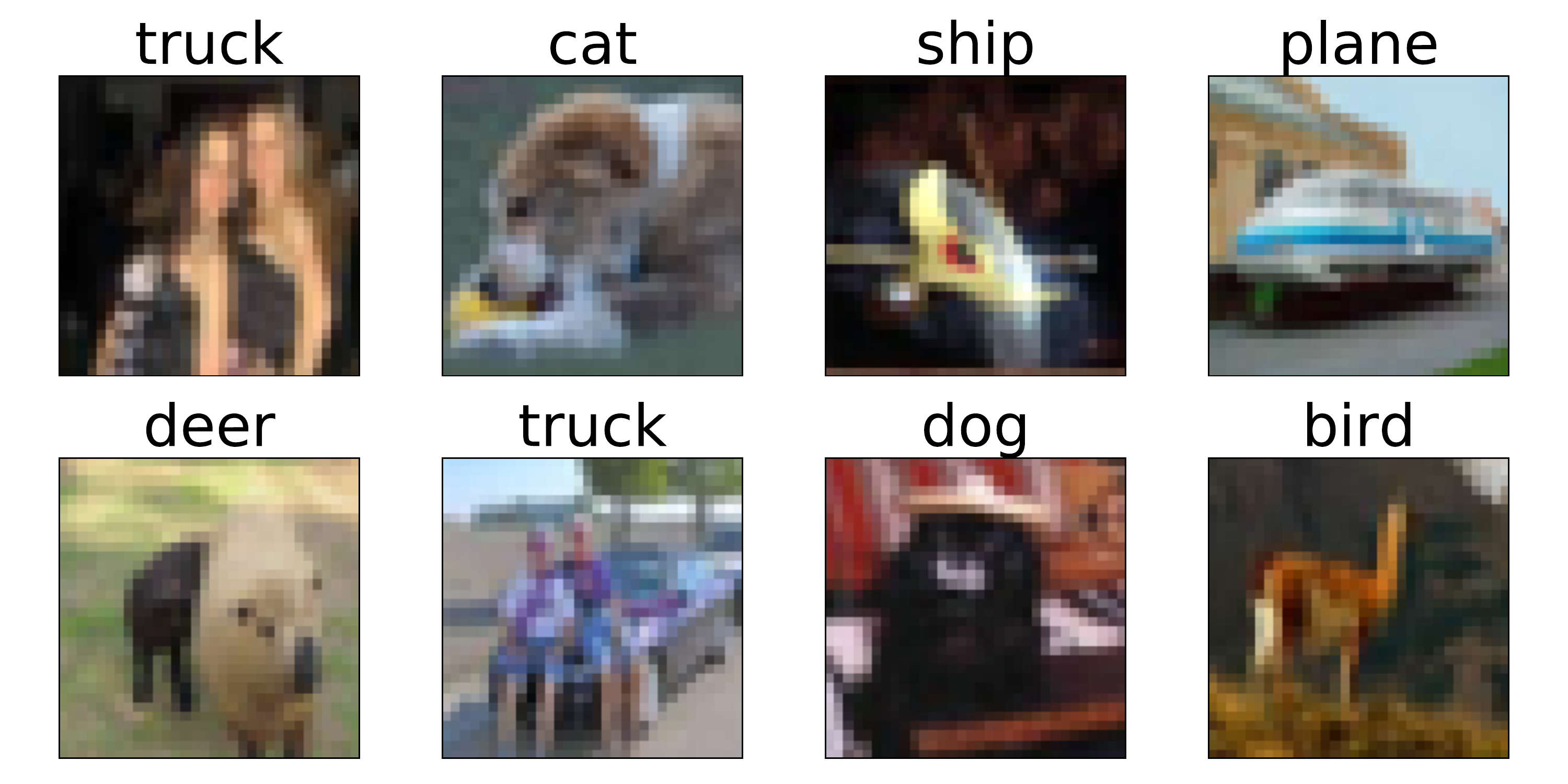
    	\caption*{CIFAR10}
  	\end{subfigure}
  	\begin{subfigure}[b]{0.4\linewidth}
   	\def\svgwidth{0.99\linewidth}
    	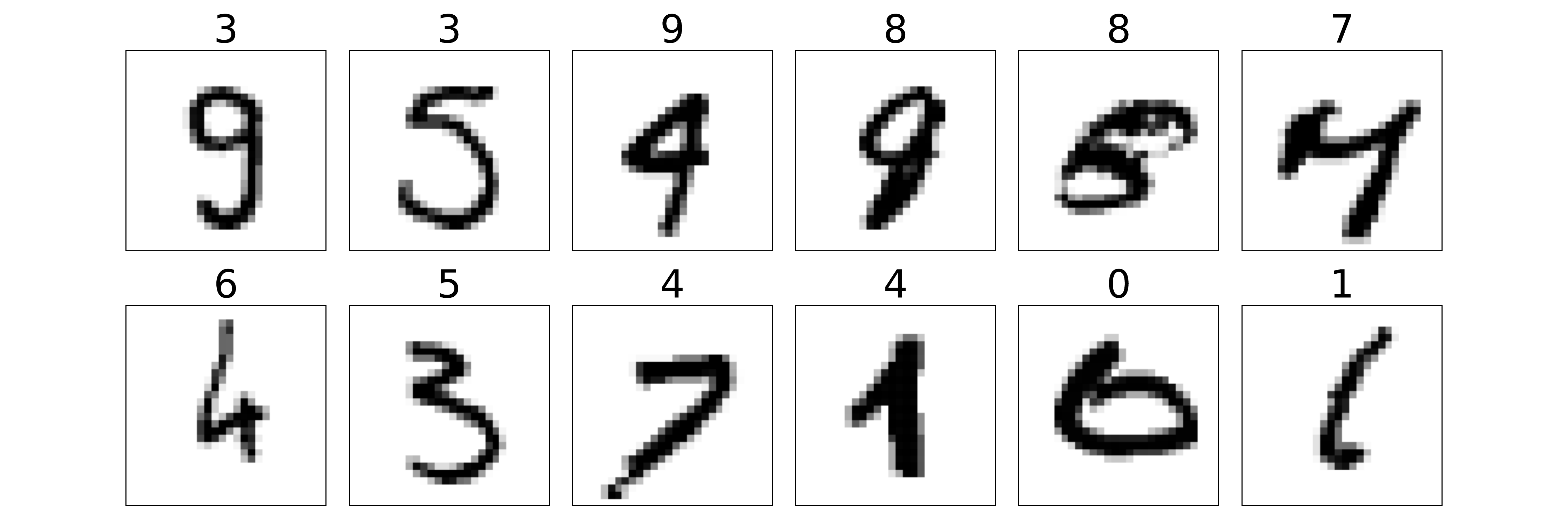
    	\caption*{MNIST}
  	\end{subfigure}
	 \caption[Label noise in real CIFAR10 and MNIST]{\label{fig:mislabelled_ds} Label noise in CIFAR10 and MNIST. The text above the image indicates the training set label.}
     \end{figure}

Starting with the celebrated work of~\citet{Zhang2016} it has been
observed that neural networks trained with SGD are capable of
memorising large amounts of label noise. Recent theoretical
work~(e.g.~\citep{Liang2018,belkin18akernel,
Belkin2018,Hastie2019,Belkin2019,belkin19a,
Bartlett2020,Muthukumar2020,Chatterji2020}) has also sought to explain
why fitting training data perfectly  does not lead to a large drop in
test accuracy, as the classical notion of overfitting might suggest.
This is commonly referred to as \emph{memorisation} or
\emph{interpolation}. We show through simple theoretical models, as
well as experiments on standard datasets, that there are scenarios
where label noise causes significant adversarial vulnerability, even
when high natural~(test) accuracy can be achieved. Surprisingly, we
find that label noise is not at all uncommon in datasets such as MNIST
and CIFAR-10 (see~\cref{fig:mislabelled_ds}).

We develop a simple theoretical framework to demonstrate how
overfitting, even very minimal, label noise causes significant
adversarial vulnerability. First, we recall the definitions of expected error, which we refer to as {\em Natural error} in this chapter, and adversarial error.

\begin{defn}[Natural Error]\label{defn:gen_risk} 
  For any distribution $\cD$
  defined over $\br{\vec{x},y}\in\reals^d\times\bc{0,1}$ and any binary
  classifier $f:\reals^d\rightarrow\bc{0,1}$, the \emph{natural} error is 
		  \begin{equation} 
			  \risk{\cD}{f}=\bP_{\br{\vec{x},y}\sim\cD}\bs{f\br{\vec{x}}\neq y},
		  \end{equation} 
    \end{defn}

\advrisk*

The following result provides a sufficient condition under which even a
small amount of label noise causes any classifier that fits the training data
perfectly to have significant adversarial error. Informally,~\Cref{thm:inf-label}
states that if the data distribution has significant probability mass in a
union of (a relatively small number of, and possibly overlapping) balls, each of which has roughly the
same probability mass (see~\Cref{eq:balls_density}), then even a small
amount of label noise renders this entire region vulnerable to adversarial
attacks to classifiers that fit the training data perfectly.

\begin{restatable}[Overfitting Label Noise Causes Adversarial Vulnerability]{thm}{infectedballs}\label{thm:inf-label}
  Let $c$ be the target classifier, and let $\cD$ be a distribution over
  $\br{\vec{x},y}$, such that $y=c\br{\vec{x}}$ in its support. Using the
  notation $\bP_\cD[A]$ to denote $\bP_{(\vec{x}, y) \sim \cD}[ \vec{x} \in A]$
  for any measurable subset $A \subseteq \reals^d$, suppose that there
  exists $c_1 \geq c_2 > 0$, $\rho>0$, and a finite set $\zeta \subset
  \reals^d$ satisfying
  \begin{equation}
    \label{eq:balls_density}
	 \bP_\cD\bs{\bigcup_{\vec{s}\in\zeta}\cB_\rho^p\br{\vec{s}}}\ge c_1\quad\text{and}\quad\forall \vec{s}\in\zeta,~\bP_\cD\bs{\cB_{\rho}^p\br{\vec{s}}}\ge\frac{c_2}{\abs{\zeta}}
  \end{equation}
  where $\cB_\rho^p\br{\vec{s}}$ represents an $\ell_p$-ball of radius $\rho$
  around $\vec{s}$. Further, suppose that each of these balls contains points
  from a single class i.e. for all $\vec{s}\in\zeta$, for all
  $\vec{x},\vec{z}\in\cB_{\rho}^p\br{\vec{s}}: c\br{\vec{x}}=c\br{\vec{z}}$.

  Let $\cS_m$ be a dataset of $m$ i.i.d. samples  drawn from
  $\cD$, which subsequently has each label flipped independently with probability $\eta$. 
  For any classifier $f$ that \emph{perfectly} fits the training data $\cS_m$
  i.e. $\forall~\vec{x},y\in\cS_m, f\br{\vec{x}}=y$, 
  $\forall \delta>0$ and $m\ge\frac{\abs{\zeta}}{\eta
    c_2}\log\br{\frac{\abs{\zeta}}{\delta}}$, with probability at least $1-\delta$, $\radv{2\rho}{f;\cD}\ge c_1$.
\end{restatable}

The goal is to find a relatively small set $\zeta$ that satisfies the condition
as this will mean that even for modest sample sizes, the trained models have
significant adversarial error. We remark that it is easy to construct concrete
instantiations of problems that satisfy the conditions of the theorem, e.g. when
each class is represented by a spherical (truncated) Gaussian with radius $\rho$
and the classes are well-separated, the distribution
satisfies~\cref{eq:balls_density}.\todo[color=red]{Prove this for gaussian dist}
The main idea of the proof is that there is sufficient probability mass for
points that are within distance $2\rho$ of a training datum that was
mislabelled. We note that the generality of the result, namely that \emph{any}
classifier~(including neural networks) that fits the training data must be
vulnerable irrespective of its structure, requires a result
like~\Cref{thm:inf-label}. For instance, one could construct the classifier $h$,
where $h(\vec{x}) = c(\vec{x})$, if $(\vec{x}, b) \not\in \cS_m$ for $b = 0, 1$,
and $h(\vec{x}) = y$ if $(\vec{x}, y) \in \cS_m$. Note that the classifier $h$
agrees with the target $c$ on \emph{every} point of $\reals^d$ except the
mislabelled training examples, and as a result, these examples are the only
source of vulnerability. Also note that if the classifier $h$ were smoother in
the vicinity of the memorised misclassified point, then the vulnerability of the
classifier would actually increase. This is surprising as normally smoother
classifiers are thought to be less vulnerable to adversarial attacks.

\begin{proof}[Proof of~\Cref{thm:inf-label}]
  From~\cref{eq:balls_density}, for any $\zeta$ and $s\in\zeta$,
  \[\bP_{\br{\vec{x},y}\sim\cD}\bs{\vec{x}\in\cB_{\rho}\br{s}}\ge
    \frac{c_2}{\abs{\zeta}}\]
  As the sampling of the point and the injection of label noise are
  independent events,
  \[\bP_{\br{\vec{x},y}\sim\cD}\bs{\vec{x}\in\cB_{\rho}\br{s}\wedge \vec{x}~\text{gets mislabelled}}\ge
    \frac{c_2\eta}{\abs{\zeta}}\]
  Thus,
  \begin{align*}
    \bP_{\cS_m\sim\cD^m}\bs{\exists\br{\vec{x},y}\in\cS_m:
    \vec{x}\in\cB_{\rho}\br{s}\wedge \vec{x}~\text{is
    mislabelled}}&\ge 
                   1 - \br{1-\frac{c_2\eta}{\abs{\zeta}}}^m\\
                 &\ge 1 - \exp\br{\frac{-c_2\eta m}{\abs{\zeta}}}&&\textrm{(By~\Cref{lem:exp_ineq})}\\
  \end{align*}
  Substituting $m\ge\frac{\abs{\zeta}}{\eta
    c_2}\log\br{\frac{\abs{\zeta}}{\delta}}$ and applying the union
  bound~(\Cref{ineq:union-bound}) over all $s\in\zeta$, we get
  \begin{equation}\label{proof:inf-label-1}
    \bP_{\cS_m\sim\cD^m}\bs{\forall s\in\zeta,~\exists\br{\vec{x},y}\in\cS_m:
    \vec{x}\in\cB_{\rho}\br{s}\wedge \vec{x}~\text{is
    mislabelled}}\ge 
1 - \delta \end{equation}
As for all
$\vec{s}\in\reals^d$ and $\forall\vec{x},\vec{z},\in\cB_\rho^p\br{\vec{s}},~\norm{\vec{x}-\vec{z}}_p\le2\rho$, we
have that
\begin{align*}
  \radv{2\rho}{f;\cD}&=\bP_{\cS_m\sim\cD^m}\bs{\bP_{\br{\vec{x},y}\sim\cD}\bs{\exists\vec{z}\in\cB_{2\rho}\br{\vec{x}}~\wedge y\neq f\br{\vec{z}}}}\\
                    &\ge\bP_{\cS_m\sim\cD^n}\bs{\bP_{\br{\vec{x},y}\sim\cD}\bs{\vec{x}\in\bigcup_{s\in\zeta}\cB_{\rho}^p\br{s}\wedge\bc{\exists\vec{z}\in\cB_{2\rho}\br{\vec{x}}: y\neq f\br{\vec{z}}}}}\\
                    &\stackrel{(1)}{=}\bP_{\cS_m\sim\cD^n}\bs{\bP_{\br{\vec{x},y}\sim\cD}\bs{\exists\vec{s}\in\zeta:\vec{x}\in\cB_{\rho}^p\br{s}\wedge\bc{\exists\vec{z}\in\cB_{2\rho}\br{\vec{x}}: y\neq f\br{\vec{z}}}}}\\
                     &\ge\bP_{\cS_m\sim\cD^m}\bs{\bP_{\br{\vec{x},y}\sim\cD}\bs{\exists\vec{s}\in\zeta:\vec{x}\in\cB_{\rho}^p\br{s}\wedge\bc{\exists\vec{z}\in\cB_{\rho}\br{\vec{s}}: y\neq f\br{\vec{z}}}}}\\
                     &\stackrel{(2)}{=}\bP_{\cS_m\sim\cD^m}\bs{\bP_{\br{\vec{x},y}\sim\cD}\bs{\exists\vec{s}\in\zeta:\vec{x}\in\cB_{\rho}^p\br{s}\wedge\bc{\exists\vec{z}\in\cB_{\rho}\br{\vec{s}}: c\br{\vec{z}}\neq f\br{\vec{z}}}}}\\
                     &\stackrel{(3)}{=}\bP_{\br{\vec{x},y}\sim\cD}\bs{\vec{x}\in
                       \bigcup_{s\in\zeta}\cB_\rho^p\br{s}}
                       \quad\text{w.p. at least}~~1-\delta\\
                     &\ge c_1  \quad\text{w.p.}~~1-\delta 
\end{align*}
where $c$ is the true concept for the distribution $\cD$. 

Equality (1) follows from the fact that the event $\vec{x}\in
\bigcup_{s\in\zeta}\cB_{\rho}^p\br{s}$ is equivalent to the event that there
exists $i\in\bc{1,..,\abs{\zeta}}$ such that $\vec{x}\in \cB_{\rho}^p\br{s_i}$.
Equality (2) follows from the assumptions that each of the balls around
$\vec{s}\in\zeta$ is pure in their labels. Equality (3) follows from
\ref{proof:inf-label-1} by using the fact that $\vec{x}$ is guaranteed to
exist in the ball around $\vec{s}$ and be mislabelled with a probability of at
least $1-\delta$. To see how, replace the $\vec{x}$ in
Assumption~\ref{proof:inf-label-1} with $\vec{z}$. The last inequality follows
from the first Assumption in~\ref{eq:balls_density}.
\end{proof}
 
There are a few things to note about~\Cref{thm:inf-label}. \begin{enumerate}
  \item   First, the
lower bound on adversarial error applies to any classifier $f$ that
fits the training data $\cS_m$ perfectly and is agnostic to the type
of model $f$ is. 
\item   Second, for a given $c_1$, there may be multiple $\zeta$s that satisfy
the bounds in~\cref{eq:balls_density} and the adversarial risk holds for all of
them. Thus, the smaller the value of $\abs{\zeta}$ the smaller the size of the
training data it needs to fit  which can be obtained by simpler classifiers. 
\item   Third, if the
distribution of the data  is such that it is concentrated around some
points then for a fixed $c_1,c_2$, a smaller value of $\rho$ would be
required to satisfy~\cref{eq:balls_density} and thus a weaker
adversary~(smaller perturbation budget~$2\rho$) can cause a much
larger adversarial error.\todo[color=blue]{Comment that mixture of gaussian in a bounded domain has a grater vulnerability than gaussian}
\end{enumerate}
\begin{remark} We remark that an interesting future direction of research would be to understand the relation between \(\zeta, c_1, c_2,\) and \(\rho\) for various common distributions and identify which distributions are more vulnerable to adversarial attacks than others. Further, it would be interesting to investigate whether the lower bound on adversarial risk can be further increased by removing the hypothesis-agnostic nature of~\Cref{thm:inf-label}. This would allow us to understand what kind of distributions and machine learning models are fundamentally more vulnerable to adversarial attacks due to memorisation of label noise.
\end{remark}
In practice, classifiers exhibit much greater vulnerability than
purely arising from the presence of memorised noisy data. However, experiments
in~\Cref{sec:exp-overfit-mislbl}~shows how label noise causes
vulnerability in a toy MNIST model, the full MNIST and CIFAR10 for a
variety of architectures. This indicates that while removing label noise is not sufficient to show adversarial robustness for interpolating classifiers, it is a necessary condition.
\section{Experimental results on the impact of memorising label noise}
\label{sec:exp-overfit-mislbl}

This section will look at empirical results on synthetic data, inspired
by the theory and on the standard datasets: MNIST~\citep{LBBH:1998} and
CIFAR10~\citep{krizhevsky2009learning}, that shows the impact of memorising label noise on adversarial vulnerability.
\subsection{Memorisation of label noise hurts adversarial accuracy}
\paragraph{Experiments on toy-MNIST}
We design a simple binary classification problem, \emph{toy-MNIST}, and show
that when fitting a complex classifier on a training dataset with label noise,
adversarial vulnerability increases with the amount of label noise and that this
vulnerability is caused by the label noise. The problem is constructed by
selecting two images from MNIST: one ``0'' and one ``1''. Each training/test
example is generated by selecting one of these images and adding i.i.d. Gaussian
noise sampled from $\cN\br{0,\sigma^2}$ for some \(\sigma>0\). We create a
training dataset of $4000$ samples by sampling uniformly from either class.
Finally, $\eta$ fraction of the training data is chosen randomly and its labels
are flipped. 
We train a neural network with four fully connected layers followed by a softmax
layer and minimise the cross-entropy loss using an SGD optimiser until the
training error becomes zero. Then, we attack this network with a ~\emph{strong}
$\ell_\infty$ PGD adversary~(discussed
in~\Cref{sec:adv-attack-bg})~\citep{madry2018towards} with
$\epsilon=\frac{64}{255}$ for $400$ steps with a step size of $0.01$. 

\begin{figure}[t]
    \begin{subfigure}[t]{0.49\linewidth}
      \begin{subfigure}[t]{0.32\linewidth}
      \centering
      \def\svgwidth{0.99\columnwidth}
      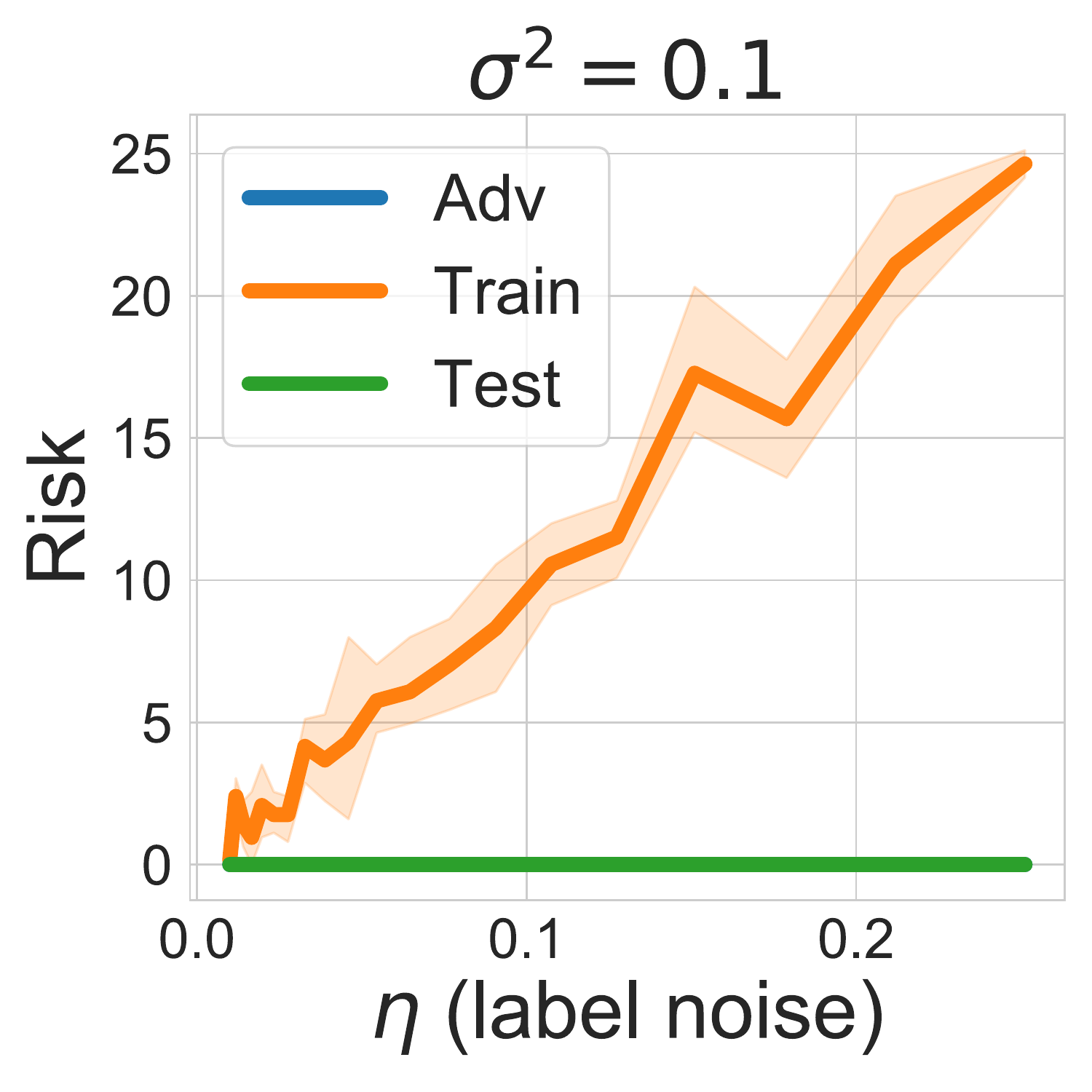
      \end{subfigure}
      \begin{subfigure}[t]{0.32\linewidth}
        \def\svgwidth{0.99\columnwidth}
        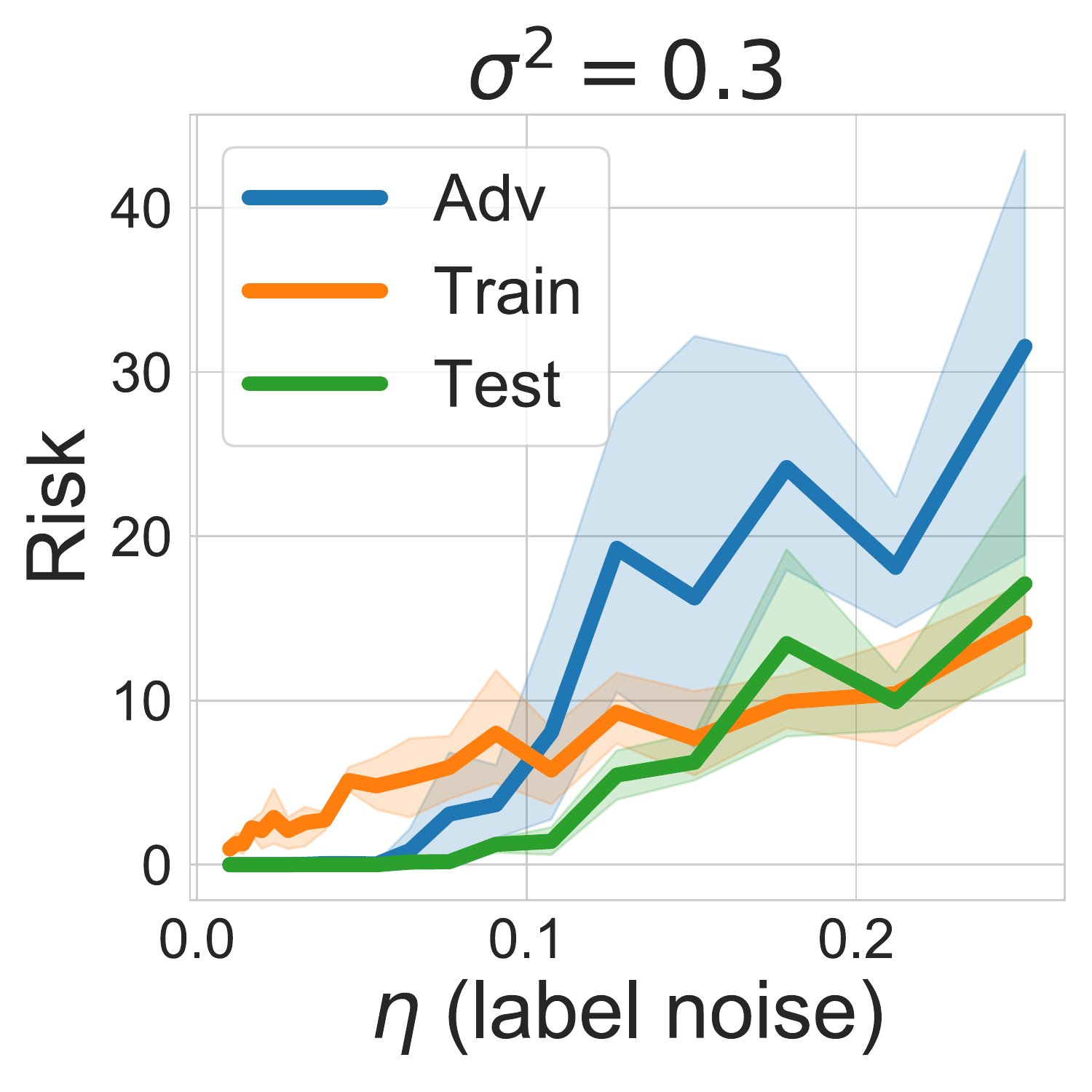
      \end{subfigure}
      \begin{subfigure}[t]{0.32\linewidth}
        \def\svgwidth{0.99\columnwidth}
        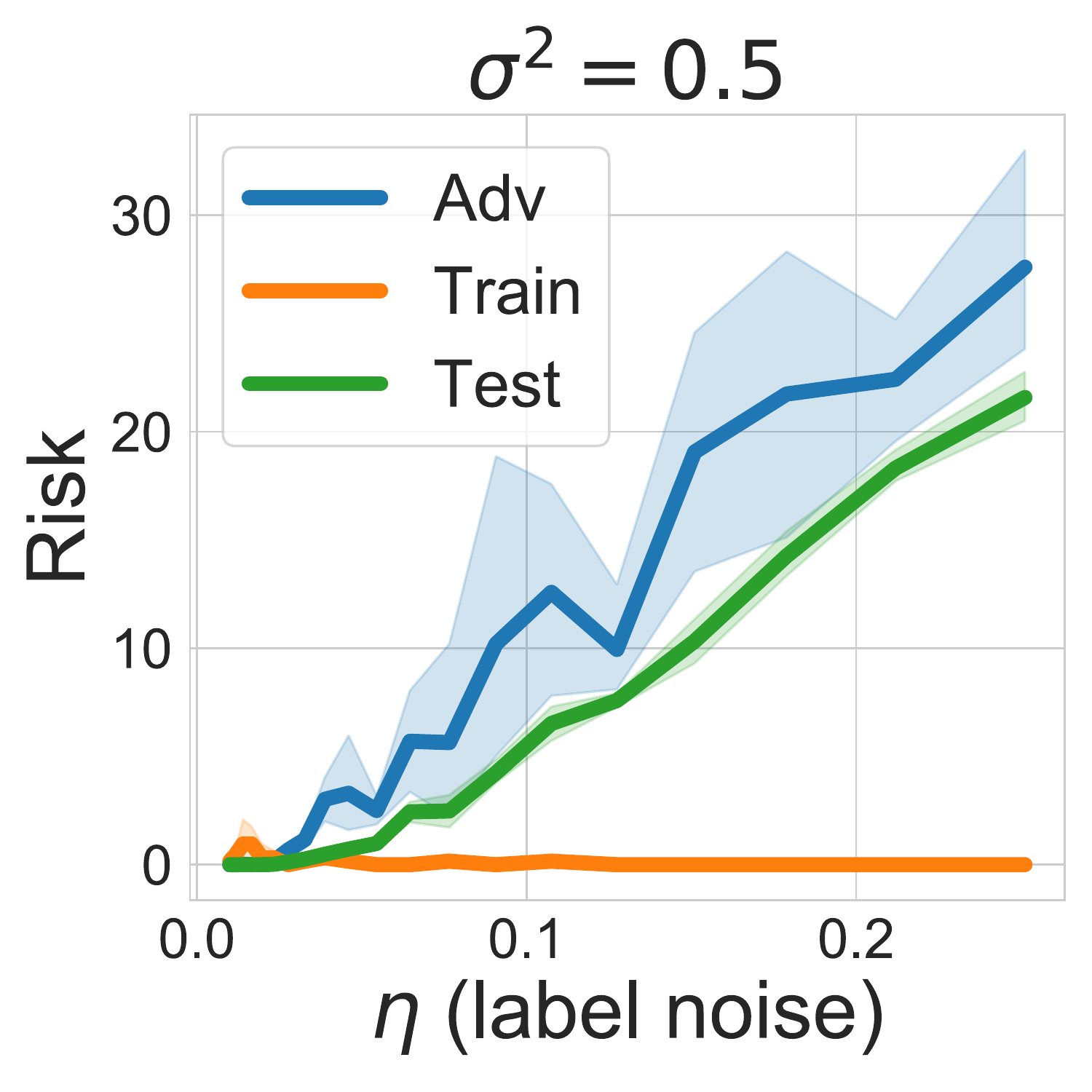
      \end{subfigure}
    \caption{Toy-MNIST , $\epsilon=\frac{64}{255}$}
    \label{fig:risk_vs_noise}
    \end{subfigure}\hfill
    \begin{subfigure}[t]{0.49\linewidth}
      \begin{subfigure}[t]{0.32\linewidth}
        \centering \def\svgwidth{0.99\columnwidth}
        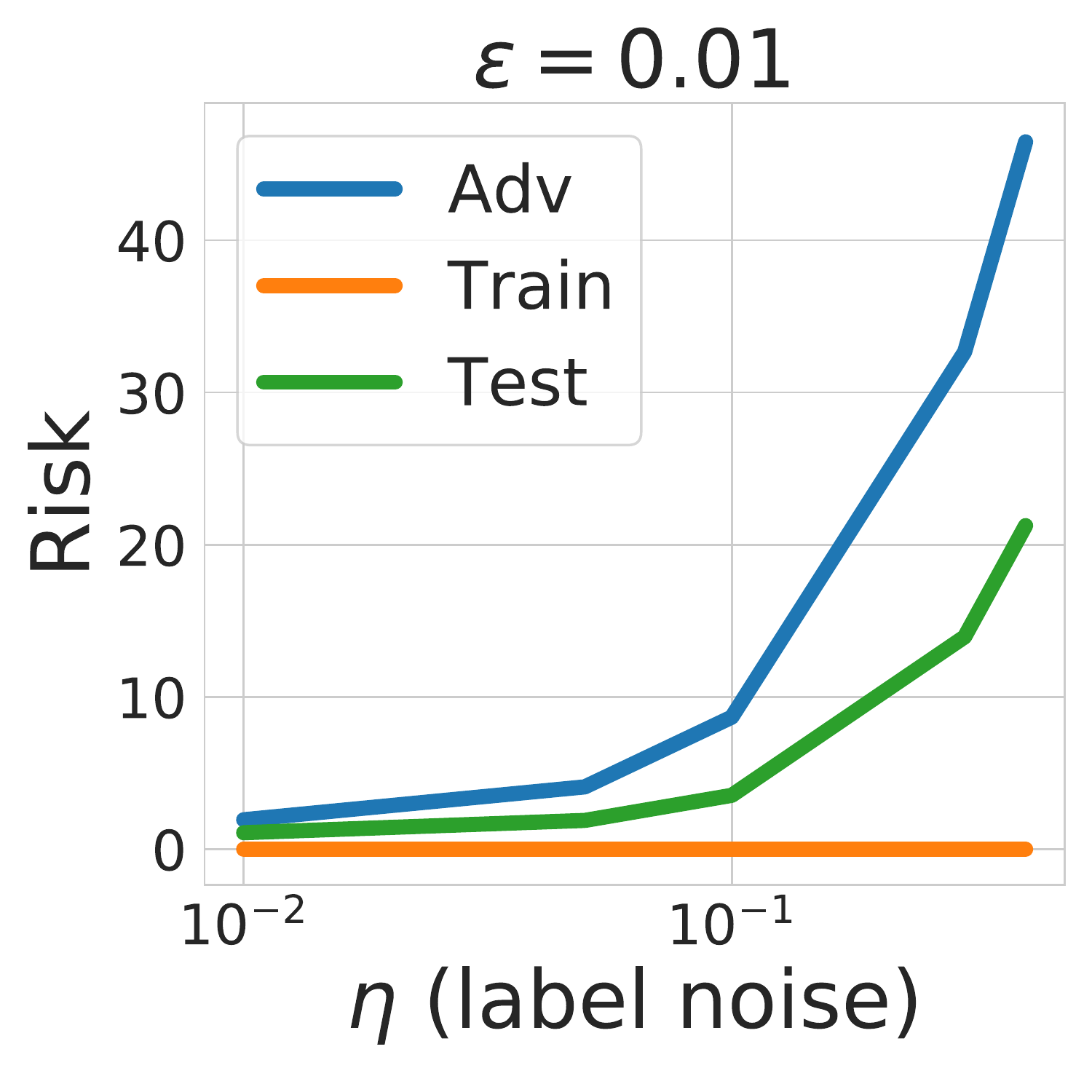
      \end{subfigure}
      \begin{subfigure}[t]{0.32\linewidth}
        \centering \def\svgwidth{0.99\columnwidth}
        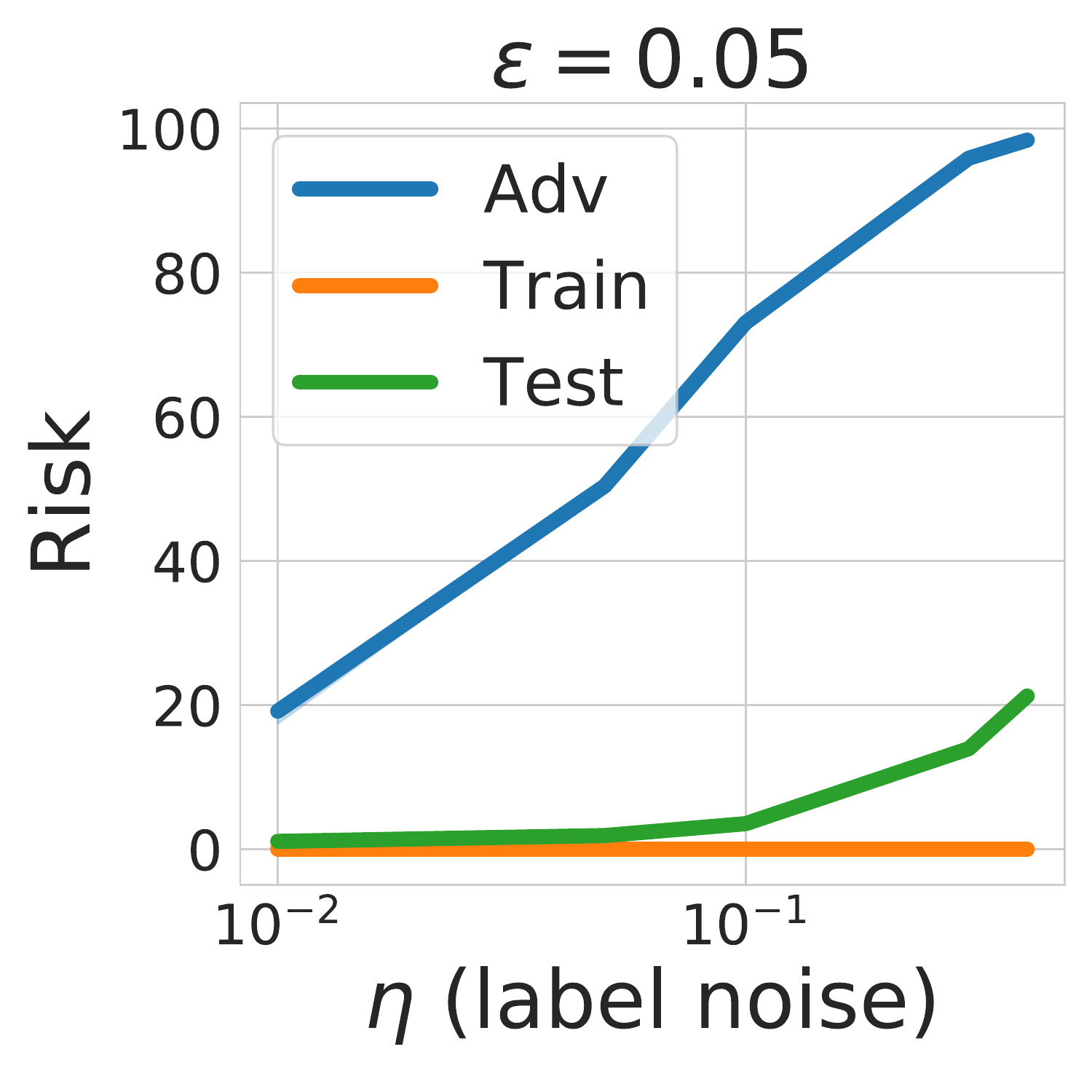
      \end{subfigure}
       \begin{subfigure}[t]{0.32\linewidth}
        \centering \def\svgwidth{0.99\columnwidth}
        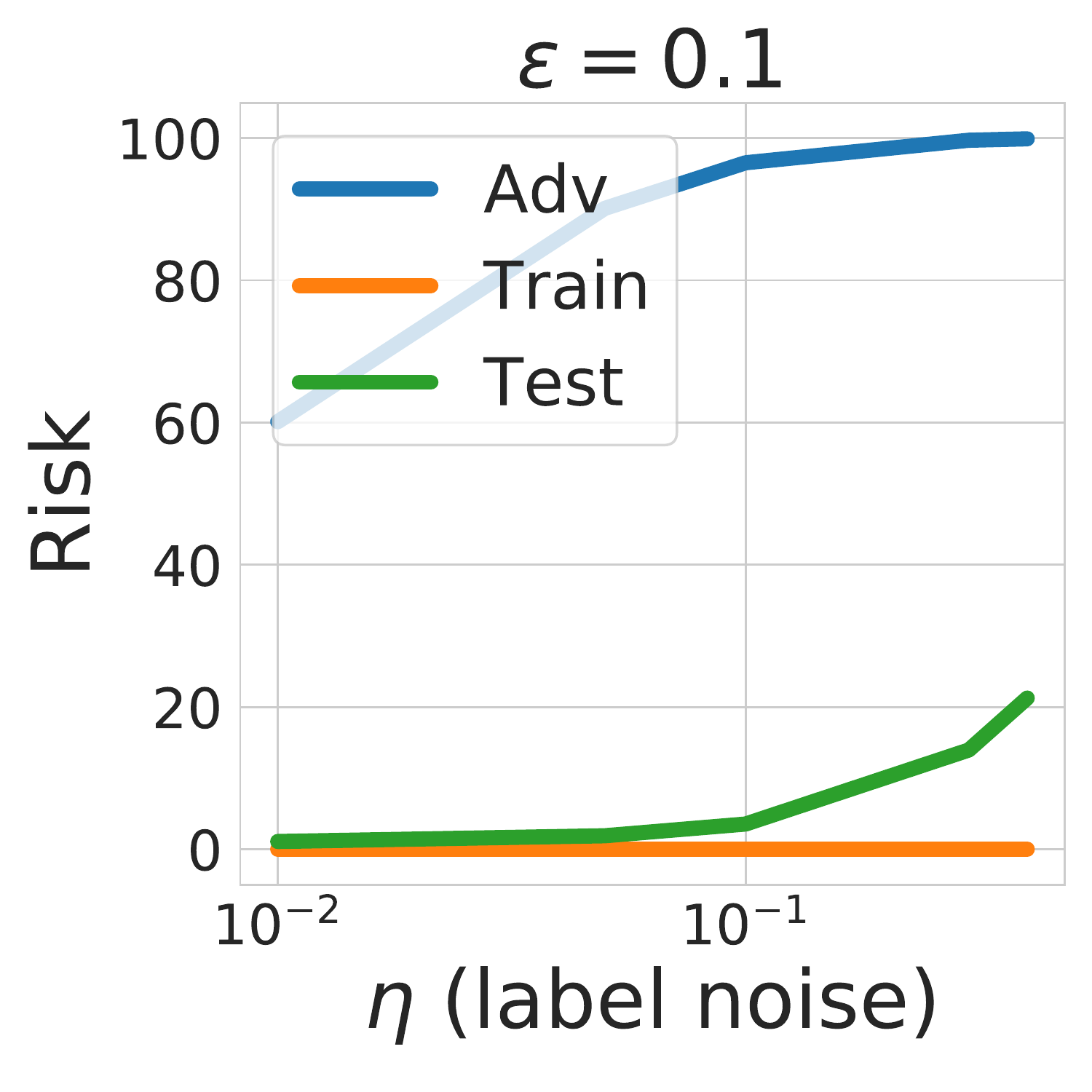
      \end{subfigure}
      \caption{Full-MNIST}
      \label{fig:mnist_lbl_noise_adv}
    \end{subfigure}\vspace{10pt}
      \begin{subfigure}[t]{0.99\linewidth}
      \begin{subfigure}[t]{0.32\linewidth}
        \centering \def\svgwidth{0.99\columnwidth}
        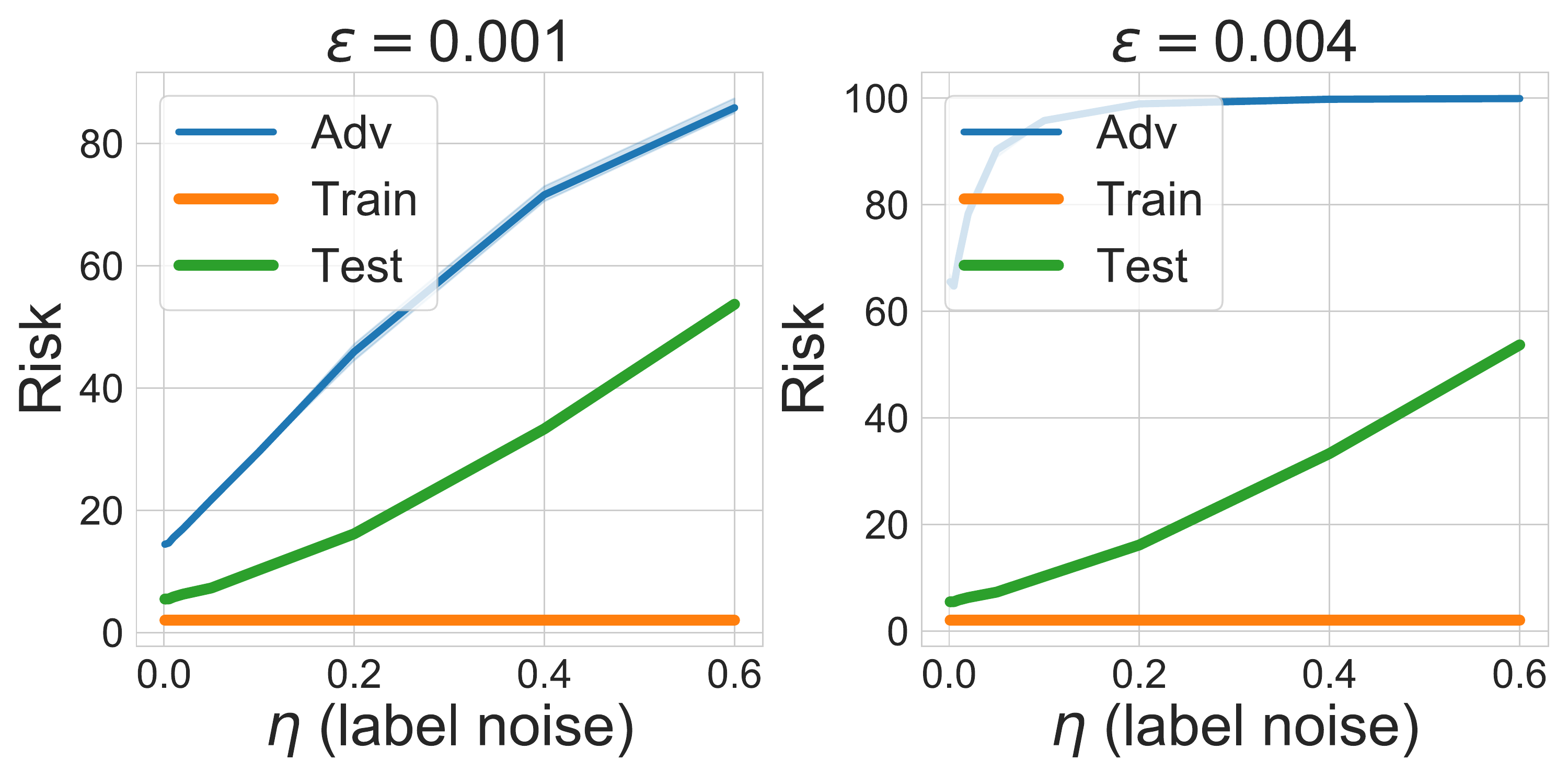
        \caption{ResNet18~(CIFAR10)}
      \label{fig:risk_vs_noise_cifar10_r18}
      \end{subfigure}
      \begin{subfigure}[t]{0.33\linewidth}
        \centering \def\svgwidth{0.99\columnwidth}
        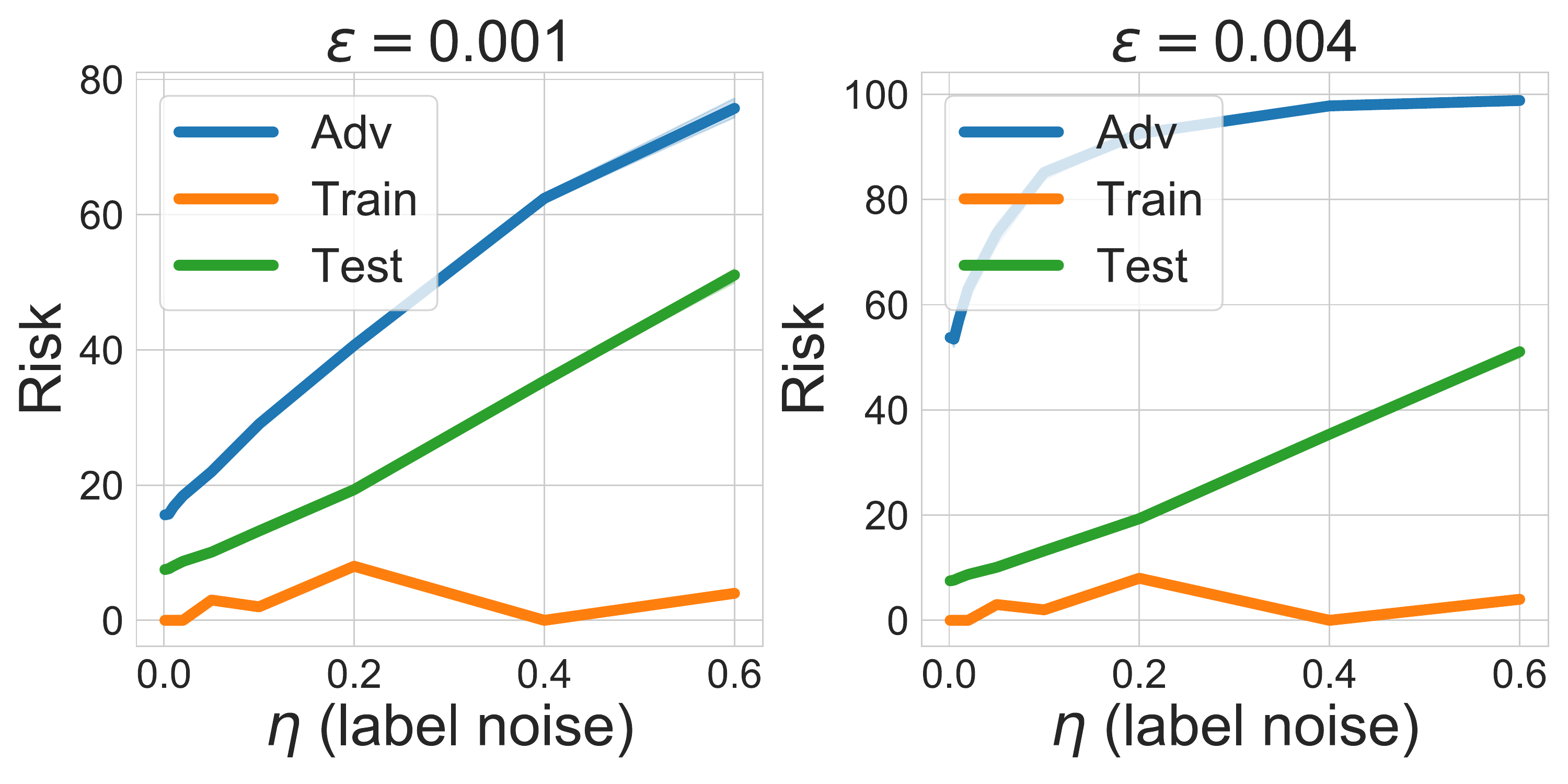
        \caption{DenseNet121~(CIFAR10)}
      \label{fig:risk_vs_noise_cifar10_d121}
      \end{subfigure}
      \begin{subfigure}[t]{0.32\linewidth}
        \centering \def\svgwidth{0.99\columnwidth}
        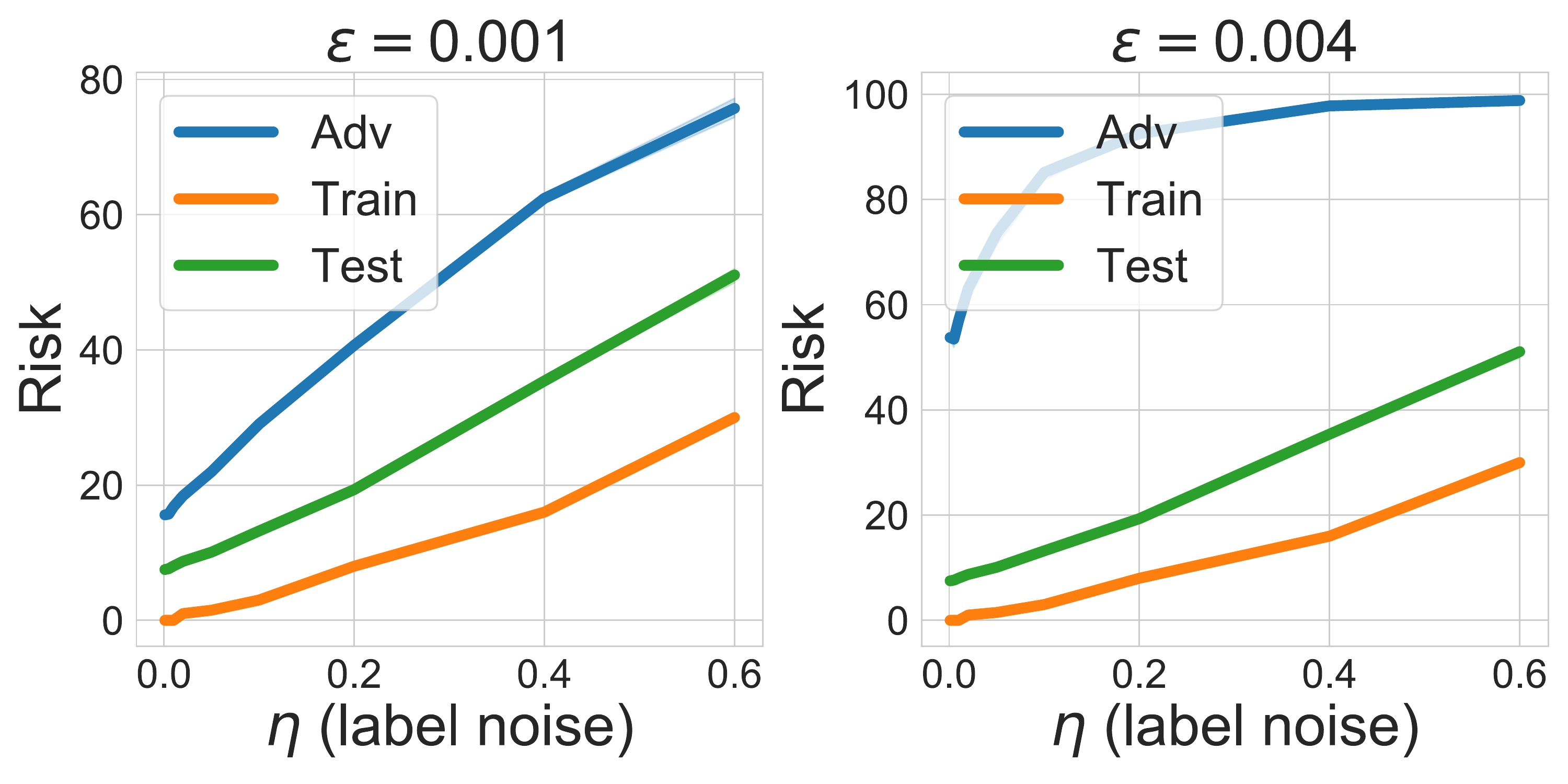
        \caption{VGG19~(CIFAR10)}
        \label{fig:risk_vs_noise_cifar10_vgg}
      \end{subfigure}
    \end{subfigure}
    \caption[Adversarial error increases with label noise]{Adversarial Error
    increases with increasing label noise~$\eta$.  The shaded region indicates a
    $95\%$ confidence interval. The absence of a shaded region indicates that it
    is invisible due to low variance.}
    
  \end{figure}
  
In~\Cref{fig:risk_vs_noise}, we plot the adversarial error, natural test error
and training error as the amount of label noise $(\eta)$ varies, for three
different values of sample variance~($\sigma^2$). For low values of
$\sigma^2$~($\sigma^2=0.1$), the training data from each class are all
concentrated around the same point; as a result, these models are unable to
memorise the label noise and the training error is high. In this case,
over-fitting label noise is impossible and the test error, as well as the
adversarial error, is low. This does not contradict~\Cref{thm:inf-label}, which
requires zero training error on the mislabelled dataset. However, as $\sigma^2$
increases to~$\sigma^2=0.5$, the neural network is flexible enough to use the
``noise component'' to extract features\todo[color=red]{Can we prove it or show evidence in support of this ?} that allow it to memorise label noise
and fit the training data perfectly. This brings the training error down to
zero while causing the test error to increase, and the adversarial error even
more so. This is in line with Theorem~\ref{thm:inf-label}. \begin{remark} The
case when $\sigma^2=0.3$ is particularly interesting; when the label noise is
low and the training error is high, there is no overfitting and the test error
and the adversarial error is zero. When the network starts memorising label
noise~(i.e. train error gets lesser than label noise), test error remains
very low  but adversarial error increases rapidly. \end{remark}

\paragraph{Experiments on the full MNIST and CIFAR10 dataset}
We perform a similar experiment on the full MNIST dataset trained on a 4-layered
Convolutional Neural Network. The model architecture consists of four
convolutional layers, followed by two fully connected layers. The first four
convolutional layers have $32,64, 128$, and  $256$ output filters with kernels of width $3,4,3,$ and $3$ respectively. The two fully connected
layers have a width of $1024$. The network is optimised with SGD with a batch
size of $128$ and an initial learning rate of $0.1$ for a total of $60$ epochs.
The learning rate is decreased to $0.01$ after $50$ epochs.  For varying values
of $\eta$, we assign a uniformly randomly label to a randomly chosen $\eta$
fraction of the training data.  We compute the natural test accuracy and the
adversarial test accuracy on a clean test-set with no label noise for when the
network is attacked with an $\ell_\infty$ bounded PGD adversary for varying
perturbation budget $\epsilon$, with a step size of $0.01$ and for $20$ steps
and plot the results in~\Cref{fig:mnist_lbl_noise_adv}.  We repeat the same
experiment for CIFAR10 with a DenseNet121~\citep{HZWV:2017},
ResNet18~\citep{HZRS:2016}, and VGG19~\citep{simonyan2014very}  to test the
phenomenon across multiple state-of-the-art architectures and plot the results
in~\Cref{fig:risk_vs_noise_cifar10_r18,fig:risk_vs_noise_cifar10_d121,fig:risk_vs_noise_cifar10_vgg}.
Please refer to~\Cref{sec:expr-settings} for more details on the architectures
and datasets. The results on both datasets show that the effect of over-fitting
label noise on adversarial error is even more clearly  visible here; for the
same PGD adversary, the adversarial error jumps upwards sharply with increasing
label noise, while the growth of natural test error is much slower. This
confirms the hypothesis that benign overfitting may not be so benign when it
comes to adversarial error.

\subsection{Representations of label noise and adversarial examples}

For the toy-MNIST problem, we plot a 2-d projection~(using PCA) of the learned
representations~(activations before the last layer) at various stages of
training in~\Cref{fig:represen}. We remark that the simplicity of the data model
ensures that even a 1-d PCA projection suffices to perfectly separate the
classes when there is no label noise; however, the representations learned by a
neural network in the presence of noise may be very different! We highlight two
key observations: 
\begin{figure}[t]
  \begin{subfigure}[t]{0.15\linewidth}
    \centering \def\svgwidth{0.99\columnwidth}
    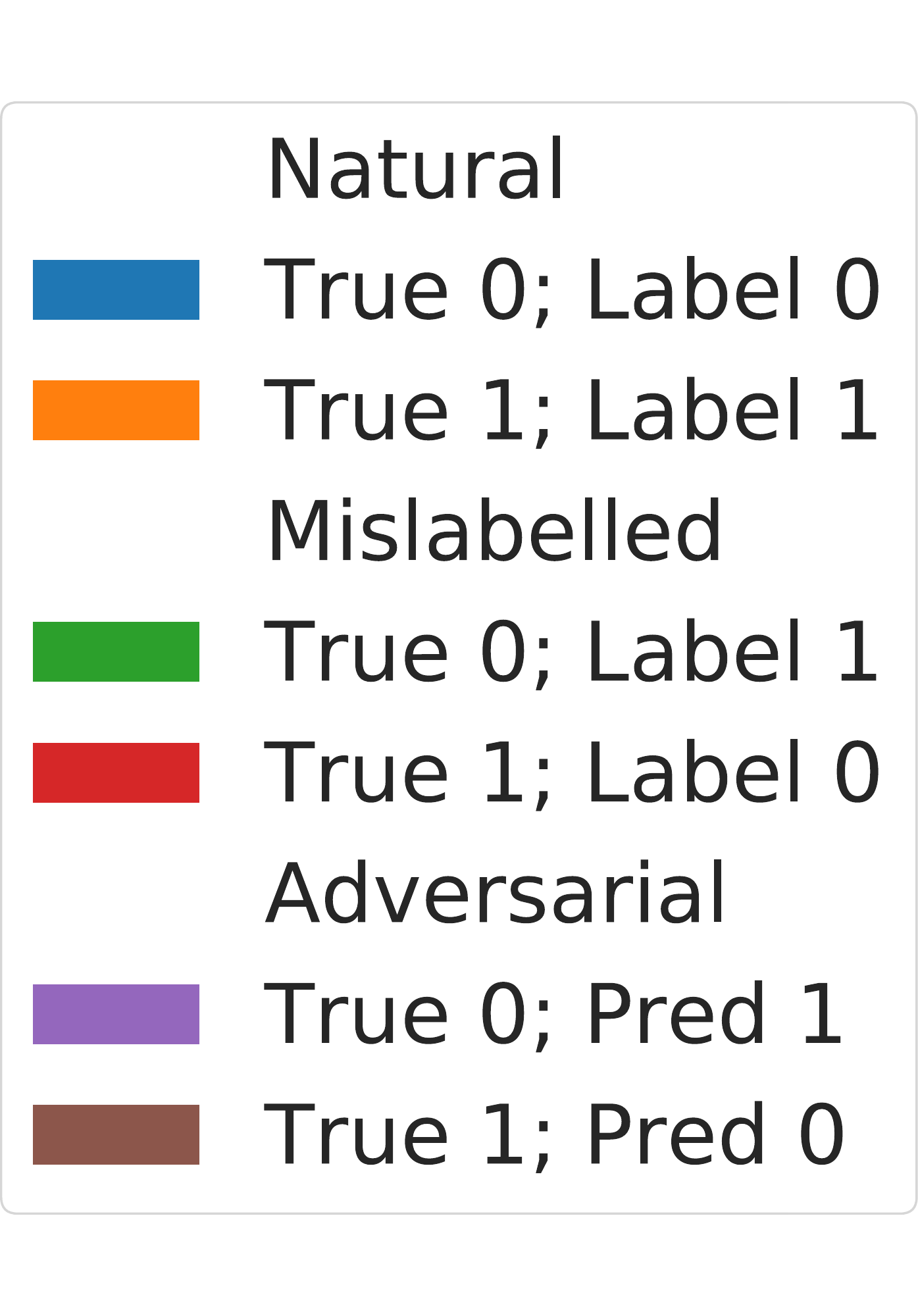
  \end{subfigure}\hfill
                 \begin{subfigure}[t]{0.8\linewidth}
    \centering \def\svgwidth{0.99\columnwidth}
    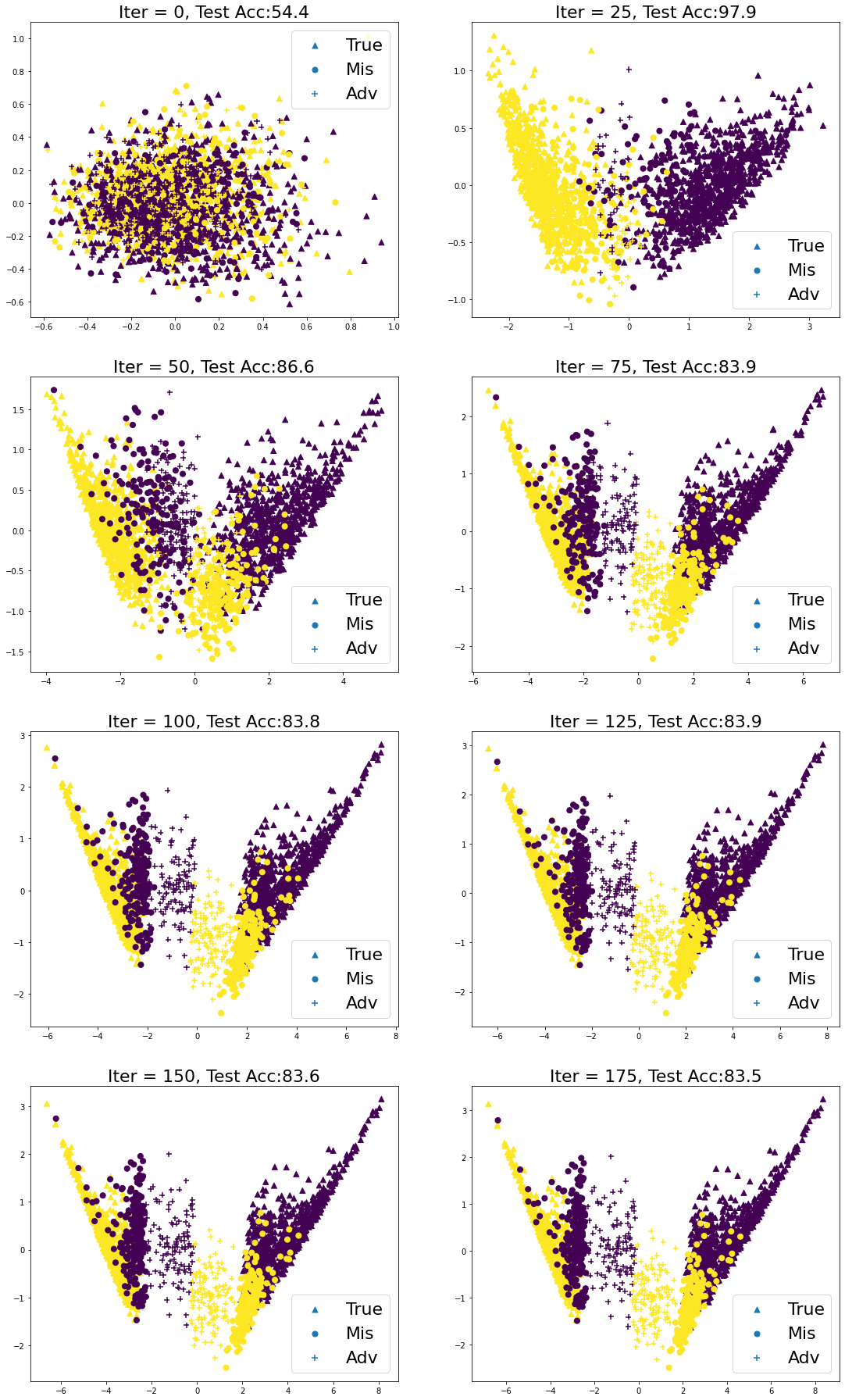
  \end{subfigure}
  \caption[Representation of mislabelled and adversarial examples]{Two dimensional PCA projections of the original correctly
    labelled~(blue and orange), original mislabelled~(green and red),
    and adversarial examples~(purple and brown) at different stages of
    training. The correct label for~\emph{True 0}~(blue),~\emph{Noisy 0}~(green),~\emph{Adv
    0}~(purple +) are the same i.e. 0 and similar for the other class.}
  \label{fig:represen}
\end{figure}

\begin{enumerate}
  \item   The bulk of adversarial examples~(``$+$''-es) are
concentrated around the mislabeled training data~(``$\circ$''-es) of the
opposite class. For example, the purple $+$-es~(Adversarially
perturbed: True: 0, Pred:1 ) are very close to the green
$\circ$-es~(Mislabelled: True:0, Pred: 1). This provides empirical
validation for the  hypothesis that if
there is a mislabeled data point in the vicinity that has been fit by the
model, an adversarial example is created by moving towards that data point
as predicted by~\Cref{thm:inf-label}. 
\item   The mislabeled training data take longer to be fit by the classifier.
For example, by iteration 20, the network learns a fairly good representation
and classification boundary that correctly fits the clean training data (but not
the noisy training data). At this stage, the number of adversarial examples is
much lower as compared to Iteration 160, by which point the network has
completely fit the noisy training data. Thus early stopping helps in avoiding
\emph{memorising} the label noise, and consequently also reduces adversarial
vulnerability. Early stopping has indeed been used as a defence in quite a few
recent papers in the context of adversarial
robustness~\citep{Wong2020Fast,hendrycks2019pretraining}, as well as learning in
the presence of label-noise~\citep{Li2019}. Our work shows \emph{why} early
stopping may reduce adversarial vulnerability by avoiding fitting noisy training
data.  
\end{enumerate}

\subsection{Robust training avoids memorisation of rare examples}
\label{sec:robust-avoid-mem}

Robust training methods like AT~\citep{madry2018towards} and
TRADES~\citep{Zhang2019} are commonly used data-augmentation based techniques to
increase the adversarial robustness of deep neural networks. However, it has been
pointed out that this comes at a cost to clean
accuracy~\citep{Raghunathan2019,tsipras2018robustness}. When trained with these
methods, both the training and test accuracy (on clean data) for commonly used
deep learning models drops with increasing strength of the PGD adversary used in
the adversarial training~(see~\Cref{tab:accs-robust-models}). In this section, we
provide  evidence to show that robust training avoids memorisation of label
noise and this also results in the drop of clean train and test accuracy. Before
going further, we will first describe how we measure memorisation and related
concepts. We borrow these two concepts from~\citet{Zhang2020} who measure the
label memorisation phenomenon using two related measures: {\em memorisation} and
{\em influence}. 

\begin{table}[t]
  \centering
  \begin{tabular}{|ccc|}\toprule
    $\epsilon$&Train-Acc.~($\%$)&Test-Acc~($\%$)\\\midrule
    0.0&99.98&95.25\\
    0.25&97.23&92.77\\
    1.0&86.03&81.62\\\bottomrule
  \end{tabular}
   \caption[Robust Training decreases clean train and test
    accuracies]{Train and Test Accuracies on Clean Dataset for
    ResNet-50 models trained using $\ell_2$ adversaries}
  \label{tab:accs-robust-models}
\end{table}

\paragraph{Memorisation or Self-Influence:} Self influence of an example
$\br{\vec{x}_i,y_i}$ for a dataset $\cS$ and learning algorithm
$\cA$~(including model, optimiser etc) can be defined as how unlikely it is for
the model learnt by $\cA$ to be correct on $\br{\vec{x}_i,y_i}$ if the training
dataset $\cS$ does not contain $\br{\vec{x}_i,y_i}$ compared to if $\cS$
contains $\br{\vec{x}_i,y_i}$. It can be formalised as follows which is
borrowed from Eq~(1) in~\citet{Zhang2020}

Memorisation by $\cA$ on example $(x_i,y_i) \in \cS$ is measured as
\[{\mathrm{mem}(\cA,\cS,i) := \bP_{h\sim \cA(\cS)}[h(x_i) = y_i] -
\bP_{h\sim \cA(\cS^{\setminus i})}[h(x_i) = y_i]}\]

where $\cS^{\setminus i}$ denotes the dataset $\cS$ with $(x_i,y_i)$ removed,
$h\sim \cA(\cS)$ denotes the model $h$ obtained by training using
algorithm $\cA$~(which includes the model architecture) on the dataset $\cS$ and
the probability is taken over the randomisation inherent in the training
algorithm $\cA$.

\paragraph{Influence of a training example on a test example:} Given a training
example $\br{\vec{x}_i, y_i}$, a test example $\br{\vec{x}'_j,y'_j}$, a training
dataset $\cS$ and a learning algorithm $\cA$, the influence of  $\br{\vec{x}_i,
y_i}$ on $\br{\vec{x}'_j,y'_j}$ measures the  probability that
$\br{\vec{x}'_j,y'_j}$ would be classified correctly if the training set $\cS$
does not contain  $\br{\vec{x}_i, y_i}$ compared to if it does. This can be
defined as follows which is borrowed from Eq 2 in~\citet{Zhang2020}. Using a
similar notation as memorisation, the influence of $(x_i, y_i)$ on $(x'_j,y'_j)$
for the learning algorithm $\cA$ with training dataset $\cS$ can be measured as

\[\mathrm{infl}(\cA,\cS,(x_i,y_i),(x'_j,y'_j)) := \bP_{h\sim
\cA(\cS)}[h(x'_j) = y'_j] - 
\bP_{h\sim \cA(\cS^{\setminus i})}[h(x'_j) = y'_j]\]

\paragraph{Robust training ignores label noise}
~\Cref{fig:mislabelled_ds} shows that label noise is not uncommon in standard
datasets like MNIST and CIFAR10. In fact, upon closely monitoring the
misclassified training set examples for both~\AT and TRADES, we found that
neither AT nor TRADES predicts correctly on the training set labels for any of
the examples identified in~\Cref{fig:mislabelled_ds}, all examples that have a
wrong label in the training set, whereas natural  training does. Thus, in line
with~\Cref{thm:inf-label}, robust training methods ignore fitting noisy labels.

We also observe this in a synthetic experiment on the full MNIST dataset where
we assigned random labels to 15\% of the dataset. A naturally trained CNN model
achieved $100\%$ train accuracy on this dataset whereas an adversarially trained
model~(standard setting with $\epsilon=0.3$ for $30$ steps) misclassified $997$
examples in the training set after the same training regime. Out of these $997$
samples, $994$ examples belonged to that \(15\%\) of the examples that were
mislabelled in the dataset.

\paragraph{Robust training ignores rare examples}
Certain examples in the training set belong to rare sub-populations~(eg. a
special kind of cat) and this sub-population is sufficiently distinct from the
rest of the examples of that class in the training dataset~(other cats in the
dataset). Next, we show that though ignoring rare samples possibly helps in
adversarial robustness, it hurts the natural test accuracy. We hypothesise
that one of the effects of robust training is to not \emph{memorise rare
examples}, which would otherwise be memorised by a naturally trained model.
As~\citet{Feldman2019} points out,\emph{ if these sub-populations are very
infrequent in the training dataset, they are indistinguishable from data points
with label noise with the difference being that examples from that
sub-population are also present in the test-set}. Natural training by
\emph{memorising} those rare training examples reduces the test error on the
corresponding test examples. Robust training, by not memorising these rare
samples~(and label noise), achieves better robustness but sacrifices the test
accuracy on the test examples corresponding to those  training points.
\paragraph{Experiments on MNIST, CIFAR10, and ImageNet} We visually demonstrate
this effect in~\Cref{fig:adv-train-test-mis-class} with examples from CIFAR10,
MNIST, and ImageNet and then provide more statistical evidence using the notions
of memorisation score and influence~\citep{Zhang2020}
in~\Cref{fig:infl-cifar10-mem}. Each pair of images contains a
misclassified~(by robustly trained models) test image and the misclassified
training image ``responsible'' for it.
Importantly both of
these images were correctly classified by a naturally trained
model. Visually, it is evident that the training images are  
extremely similar to the corresponding test image.  Inspecting the
rest of the training set, they are also very different from other
images in the training set. We can thus refer to these as rare
sub-populations. 

We found the images in~\Cref{fig:adv-train-test-mis-class} by manually searching
for each test  image, the training image that is misclassified and is visually
close to it. Our search space was shortened with the help of the influence
scores of each training image on the test image. We searched in the set of
top-$10$ most influential misclassified train images for each misclassified
test image. The model used for~\cref{fig:adv-train-test-mis-class} is a) an \AT
ResNet50 model for  CIFAR10 with $\ell_2$-adversary with an $\epsilon=0.25$, b)
a model trained with TRADES for MNIST with $\lambda=\frac{1}{6}$ and
$\epsilon=0.3$, and c) and \AT ResNet50 model for Imagenet with $\ell_2$
adversary with $\epsilon=3.0$. ~\citet{Zhang2020}  provided us the with the
memorisation scores for each image in CIFAR10 as well as the influence score of
each training image on each test image for each class in CIFAR-10. High
Influence pairs of Imagenet were obtained
from~\url{https://pluskid.github.io/influence-memorization/}. This was used to
obtain the figures for the Imagenet dataset
in~\Cref{fig:adv-train-test-mis-class}. 

\begin{figure}[t]
    \centering
   \begin{subfigure}[b]{0.53\linewidth}
   \begin{subfigure}[t]{0.99\linewidth}
      \includegraphics[width=0.24\linewidth]{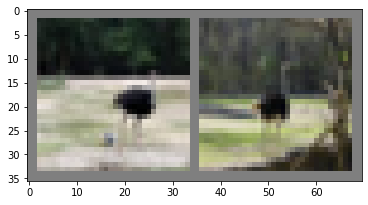}
      \includegraphics[width=0.24\linewidth]{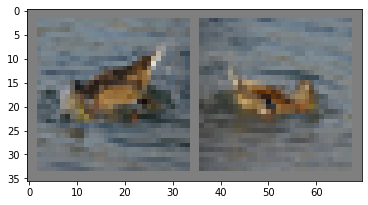}
      \includegraphics[width=0.24\linewidth]{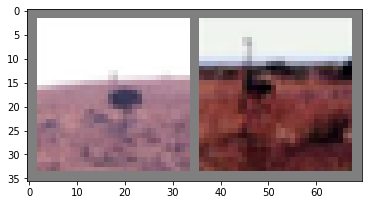}
      \includegraphics[width=0.24\linewidth]{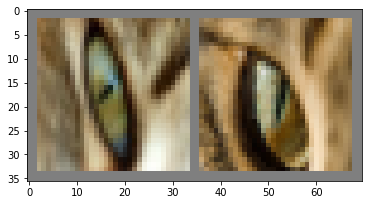}
    \end{subfigure}
    \begin{subfigure}[t]{0.99\linewidth}
      \includegraphics[width=0.24\linewidth]{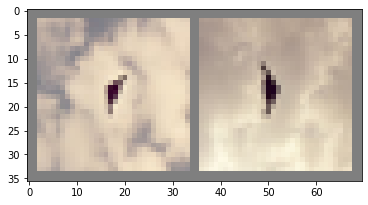}
      \includegraphics[width=0.24\linewidth]{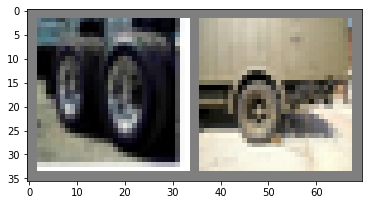}
      \includegraphics[width=0.24\linewidth]{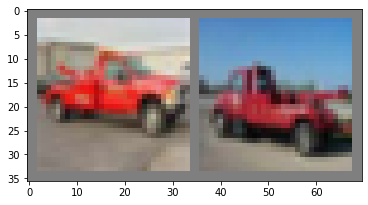}
      \includegraphics[width=0.24\linewidth]{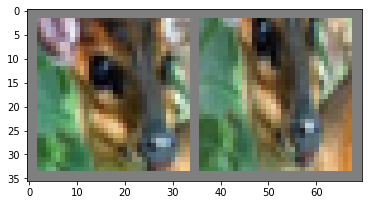}
    \end{subfigure}
    \begin{subfigure}[t]{0.99\linewidth}
      \includegraphics[width=0.24\linewidth]{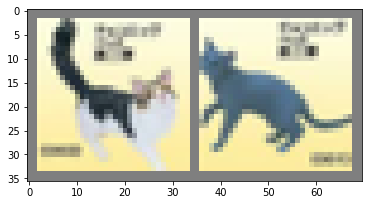}
      \includegraphics[width=0.24\linewidth]{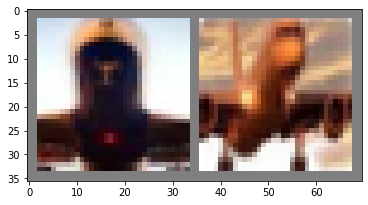}
      \includegraphics[width=0.24\linewidth]{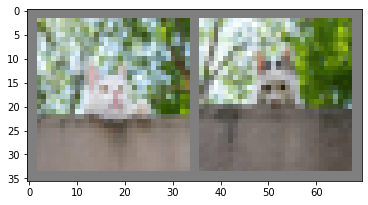}
  \includegraphics[width=0.24\linewidth]{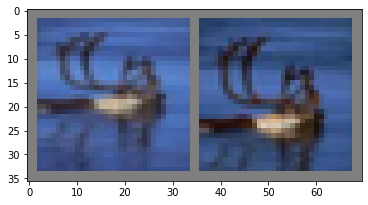}
  \end{subfigure}
  \caption*{CIFAR10}
  \end{subfigure}
  \begin{subfigure}[b]{0.45\linewidth}
    \centering \def\svgwidth{0.99\linewidth}
    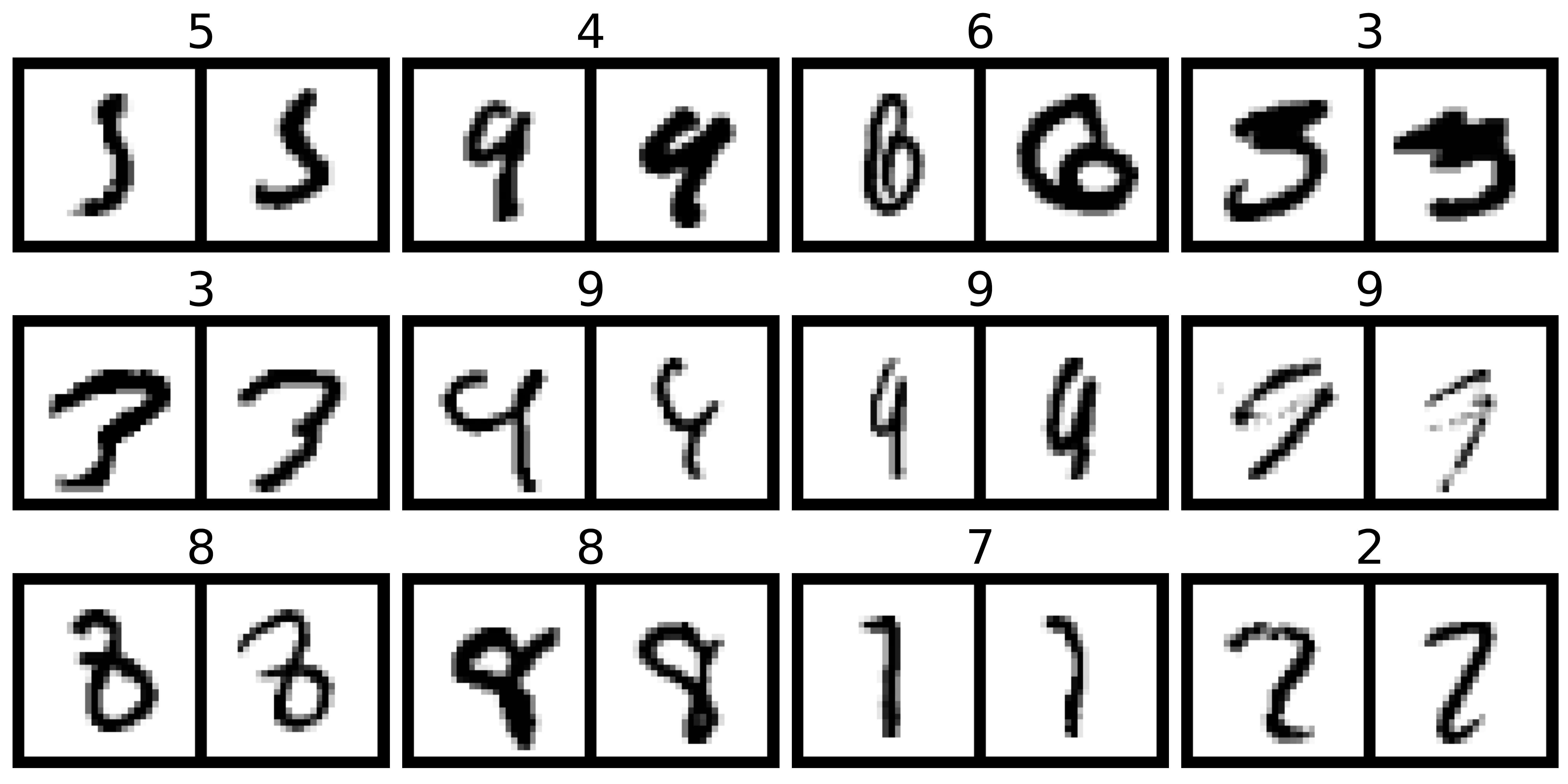
    \caption*{MNIST}
  \end{subfigure}
  
    \begin{subfigure}[b]{0.99\linewidth}
      \begin{subfigure}[t]{0.16\linewidth}
        \centering \def\svgwidth{0.99\columnwidth}
        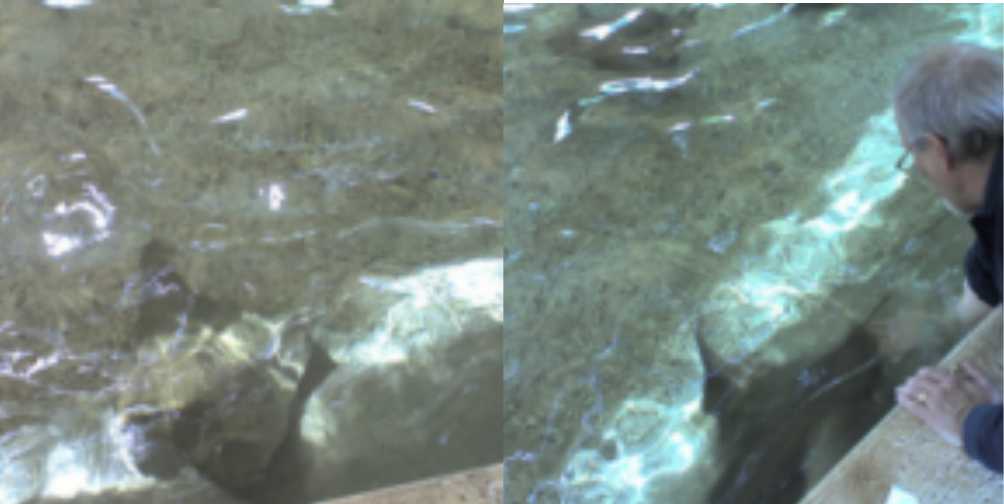
      \end{subfigure}
      \begin{subfigure}[t]{0.16\linewidth}
        \centering \def\svgwidth{0.99\columnwidth}
        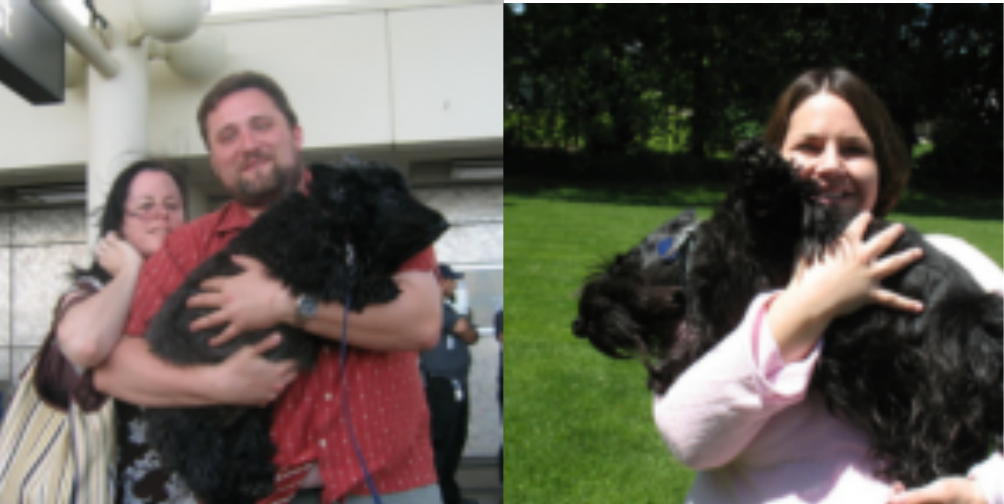
      \end{subfigure}
      \begin{subfigure}[t]{0.16\linewidth}
        \centering \def\svgwidth{0.99\columnwidth}
        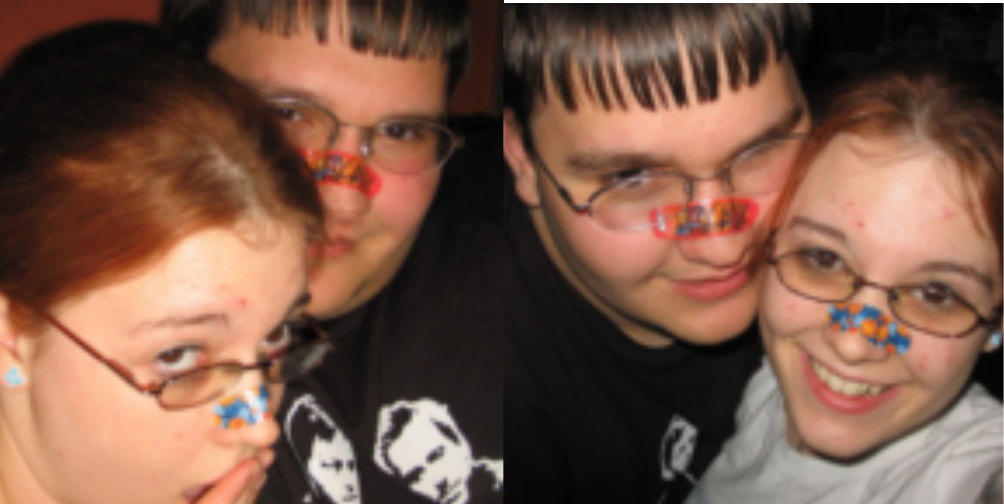
      \end{subfigure}
      \begin{subfigure}[t]{0.16\linewidth}
        \centering \def\svgwidth{0.99\columnwidth}
        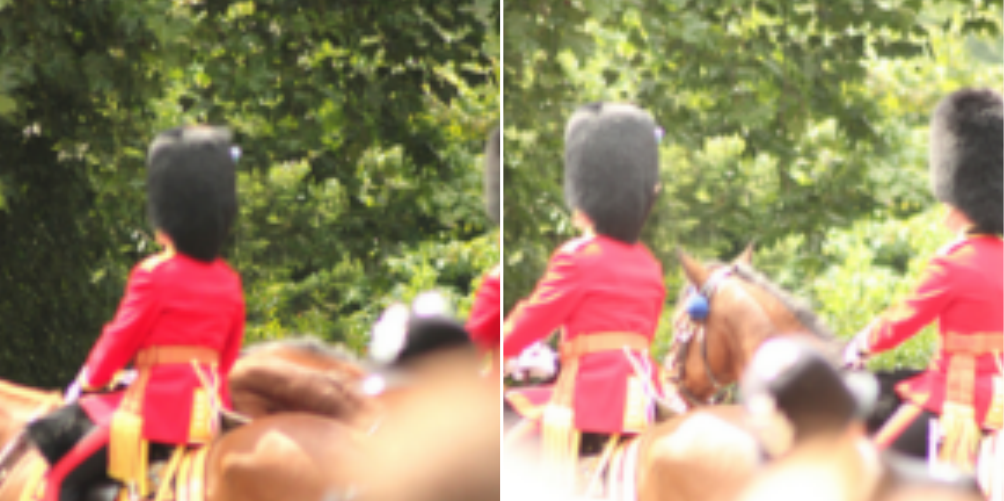
      \end{subfigure}
      \begin{subfigure}[t]{0.16\linewidth}
        \centering \def\svgwidth{0.99\columnwidth}
        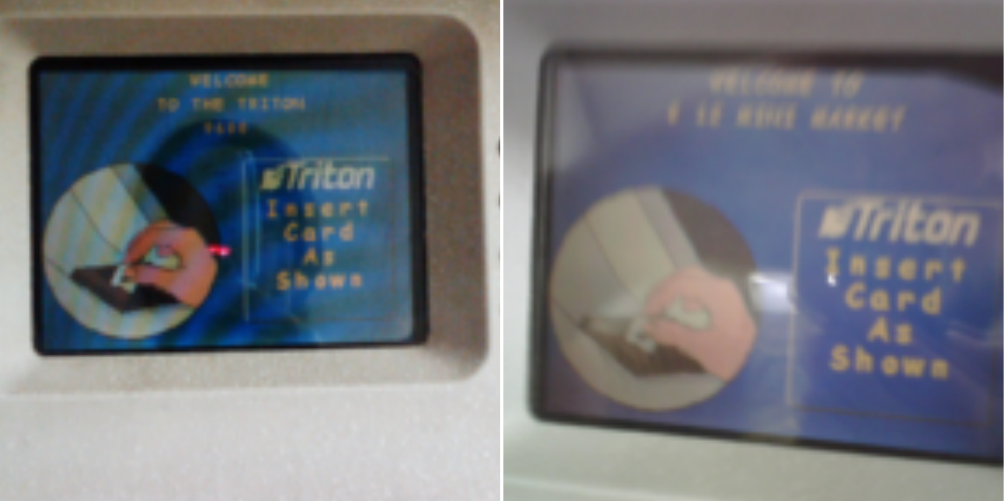
      \end{subfigure}
      \begin{subfigure}[t]{0.16\linewidth}
        \centering \def\svgwidth{0.99\columnwidth}
        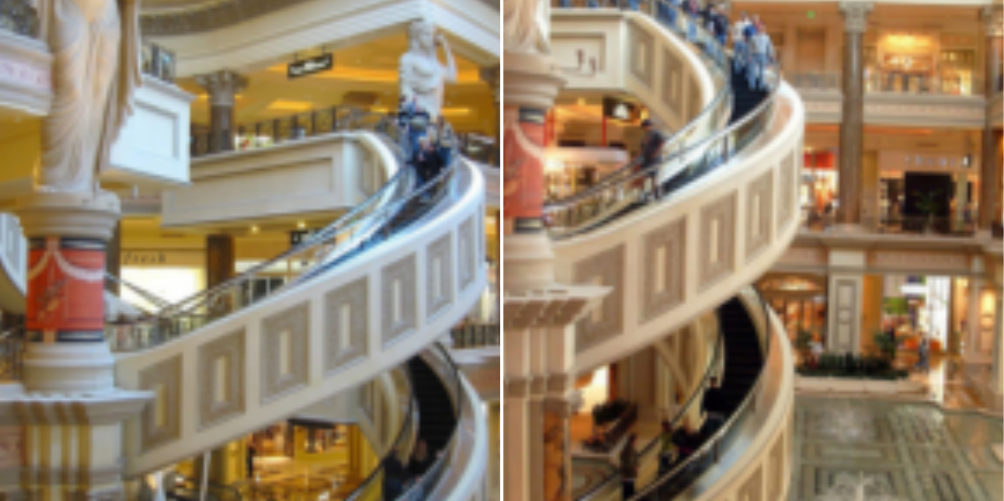
      \end{subfigure}
    \end{subfigure}
    \begin{subfigure}[b]{0.99\linewidth}
      \begin{subfigure}[t]{0.16\linewidth}
        \centering \def\svgwidth{0.99\columnwidth}
        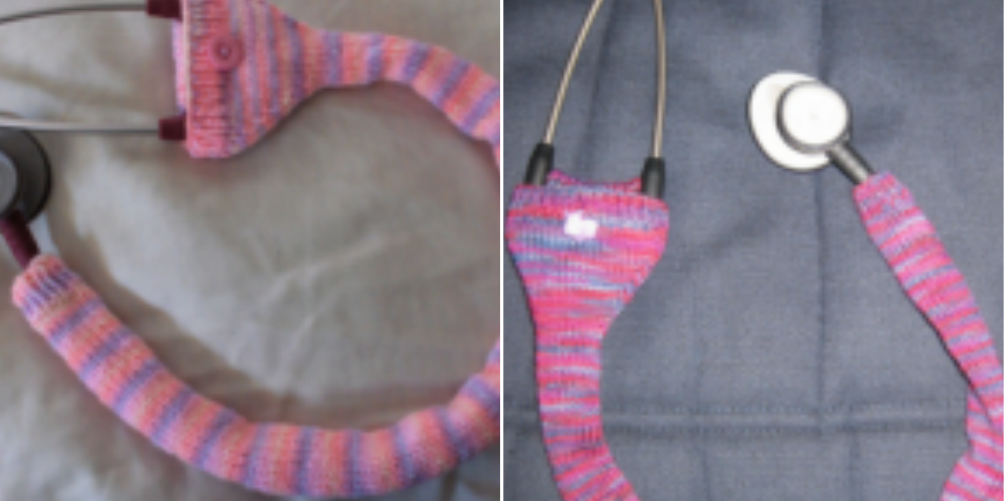
      \end{subfigure}
      \begin{subfigure}[t]{0.16\linewidth}
        \centering \def\svgwidth{0.99\columnwidth}
        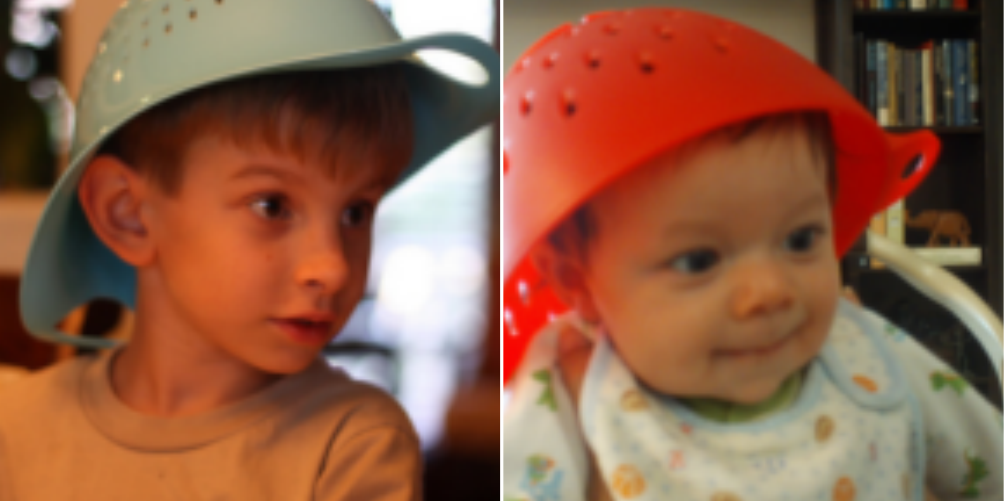
      \end{subfigure}
      \begin{subfigure}[t]{0.16\linewidth}
        \centering \def\svgwidth{0.99\columnwidth}
        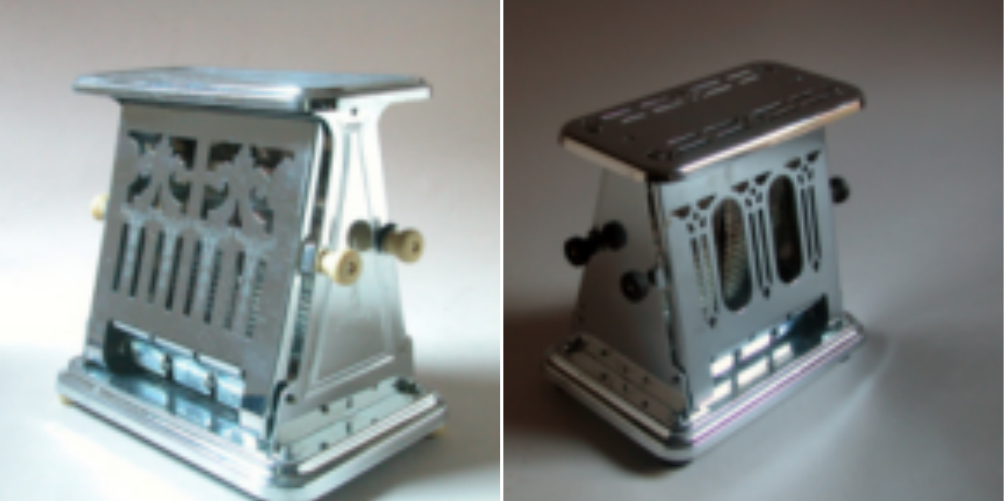
      \end{subfigure}
      \begin{subfigure}[t]{0.16\linewidth}
        \centering \def\svgwidth{0.99\columnwidth}
        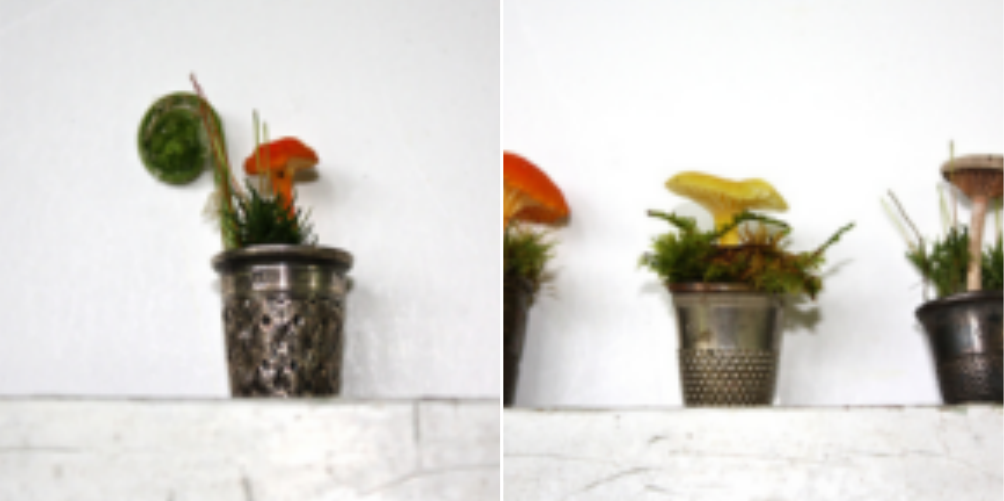
      \end{subfigure}
      \begin{subfigure}[t]{0.16\linewidth}
        \centering \def\svgwidth{0.99\columnwidth}
        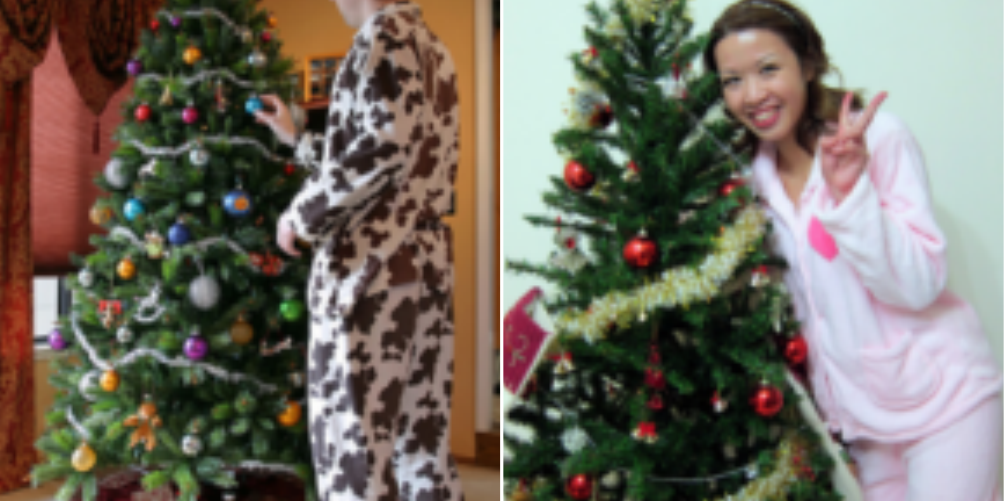
      \end{subfigure}
      \begin{subfigure}[t]{0.16\linewidth}
        \centering \def\svgwidth{0.99\columnwidth}
        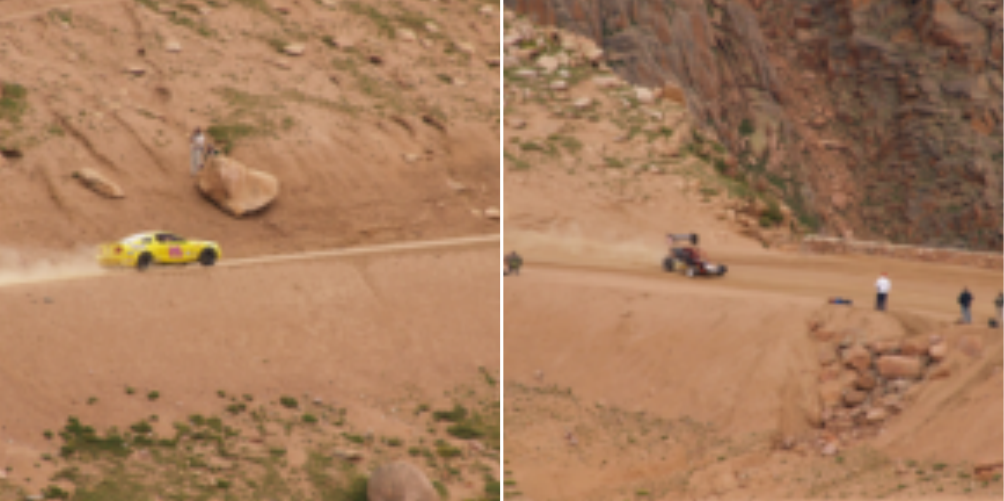
      \end{subfigure}
      \caption*{ImageNet}
    \end{subfigure}
  \caption[Adversarial Training Ignores High Influence Train-test Pairs]{Each pair is a  training~(left)
      and  test~(right) image misclassified by the adversarially
      trained model. They were both correctly classified
      by the naturally-trained model.}
    \label{fig:adv-train-test-mis-class}
  \end{figure}

  A precise notion of
  measuring if a sample is~\emph{rare} is through the concept of
  self-influence or memorisation. Self-influence for a~\emph{rare
  example}, that is unlike other examples of that class, will be high as
  the rest of the dataset will \emph{not} provide relevant information
  that will help the model to  
  correctly predict on that particular example.
  In~\cref{fig:cifar_self_influence}, we show that the self-influence of
  training samples that were misclassified by adversarially trained models but
  correctly classified by a naturally trained model is higher  compared to the
  distribution of self-influence on the entire train dataset. In other words, it
  means that the  self-influence  of the training examples misclassified by the
  robustly trained models is larger than the average self-influence of ~(all)
  examples belonging to that class. This supports our hypothesis that
  adversarial training excludes fitting these rare~(or ones that need to be
  memorised) samples.

The notion that certain test examples were not classified correctly due to a particular training example not being classified correctly is measured by the
\emph{influence} a training image has on the test image~(c.f. definition 3
in~\citet{Zhang2020}). We obtained the influence of each training image on each
test image for that class from~\citet{Zhang2020} and the training images in
~\Cref{fig:adv-train-test-mis-class} has a disproportionately higher influence
on the corresponding test image compared to influences of other train-test image
pairs in CIFAR10.  

In ~\Cref{fig:infl-cifar10_tr_te_adv}, we show that the influence of training
images are higher on test images that are misclassified by adversarially
trained models as compared to an average test image from the dataset. In other
words, this means that adversarially trained models misclassify test examples
that are being heavily influenced by some particular training example. As we saw
in~\Cref{fig:cifar_self_influence},  AT models do not memorise atypical train
examples; consequently, they misclassify test examples that are heavily
influenced by those atypical train examples~(visualised
in~\Cref{fig:adv-train-test-mis-class}).  This confirms our hypothesis that the
loss in test accuracy of robustly trained models is due to test images that
are~\emph{rare} and thus have a particularly high influence from a training
image.

\begin{figure}[t]
    \begin{subfigure}[b]{0.7\linewidth}
    \begin{subfigure}[t]{0.99\linewidth}
      \begin{subfigure}[t]{0.19\linewidth}
        \centering \def\svgwidth{0.99\linewidth}
        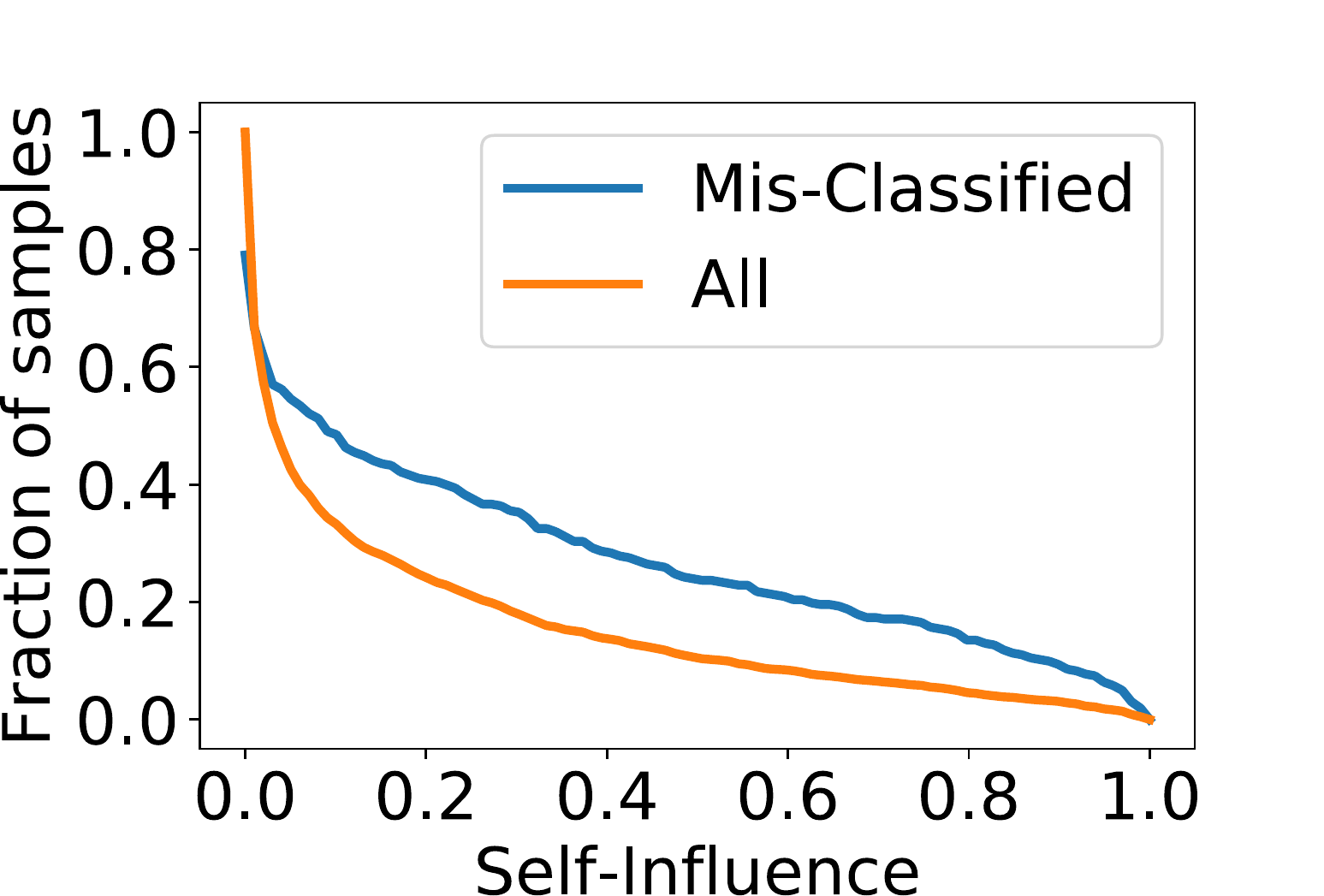
        \caption*{PLANES}    
      \end{subfigure}
      \begin{subfigure}[t]{0.19\linewidth}
        \centering \def\svgwidth{0.99\linewidth}
        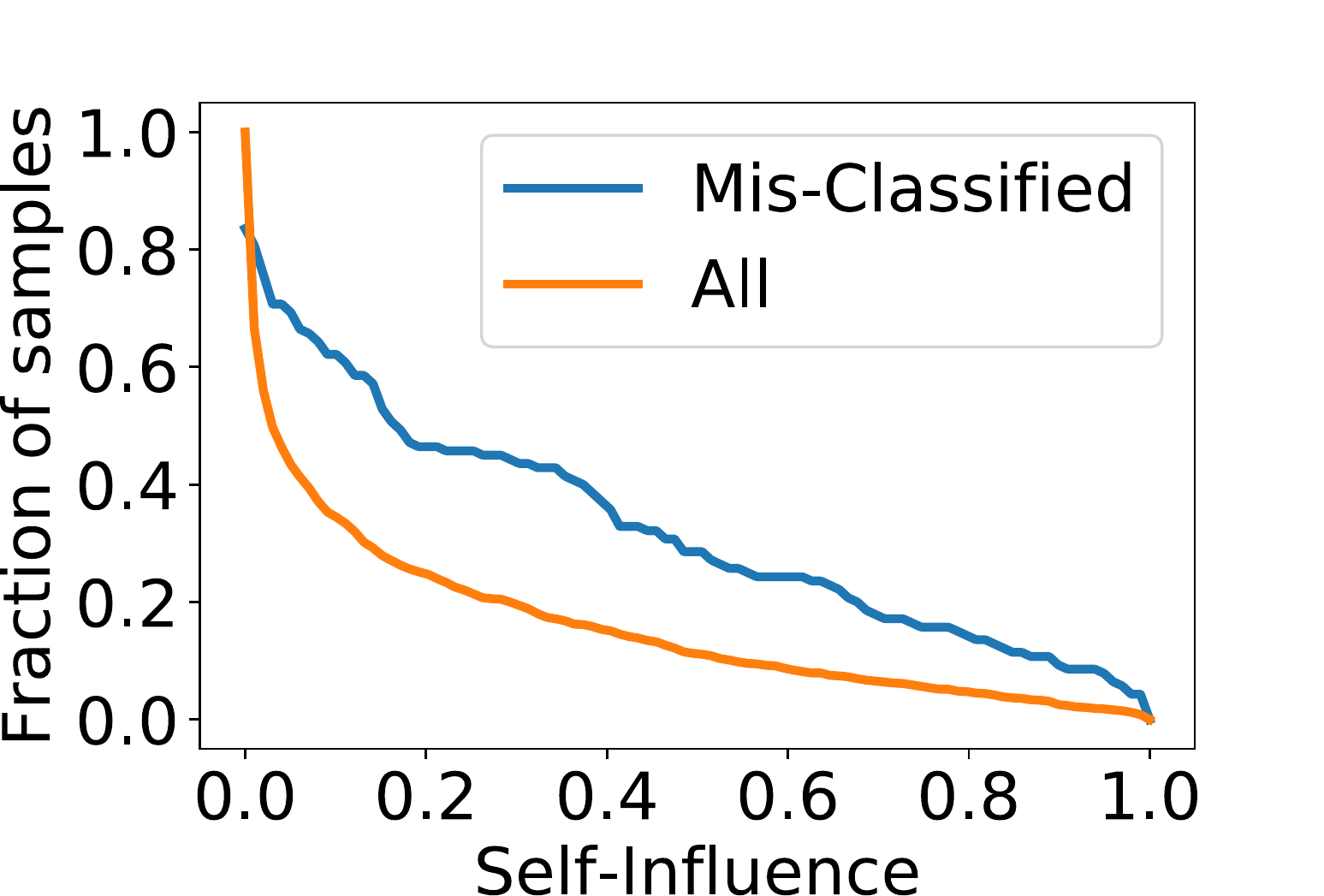
        \caption*{CAR}     
      \end{subfigure}
      \begin{subfigure}[t]{0.19\linewidth}
        \centering \def\svgwidth{0.99\linewidth}
        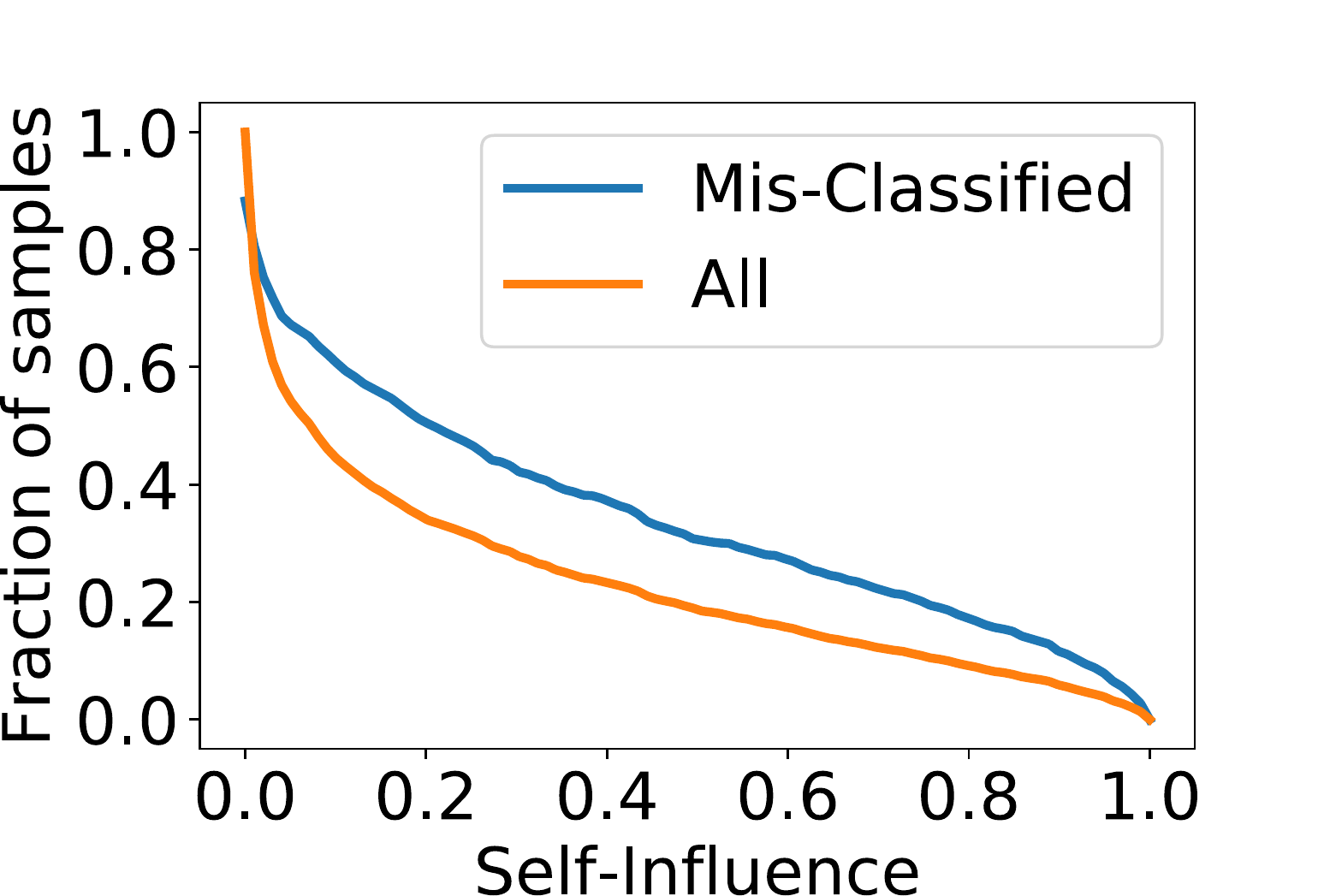
        \caption*{BIRD}     
      \end{subfigure}
      \begin{subfigure}[t]{0.19\linewidth}
        \centering \def\svgwidth{0.99\linewidth}
        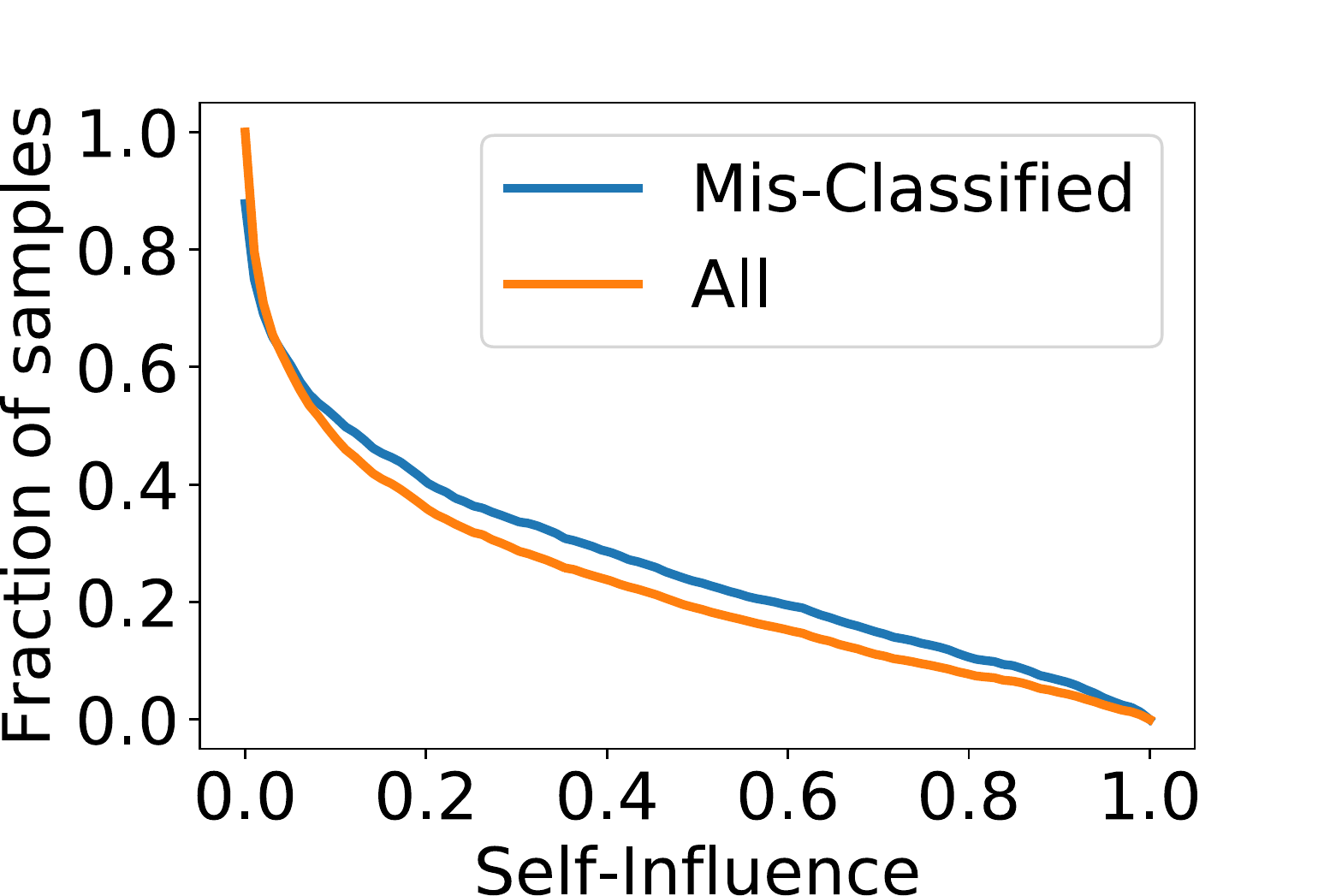
        \caption*{CAT}    
      \end{subfigure}
      \begin{subfigure}[t]{0.19\linewidth}
        \centering \def\svgwidth{0.99\linewidth}
        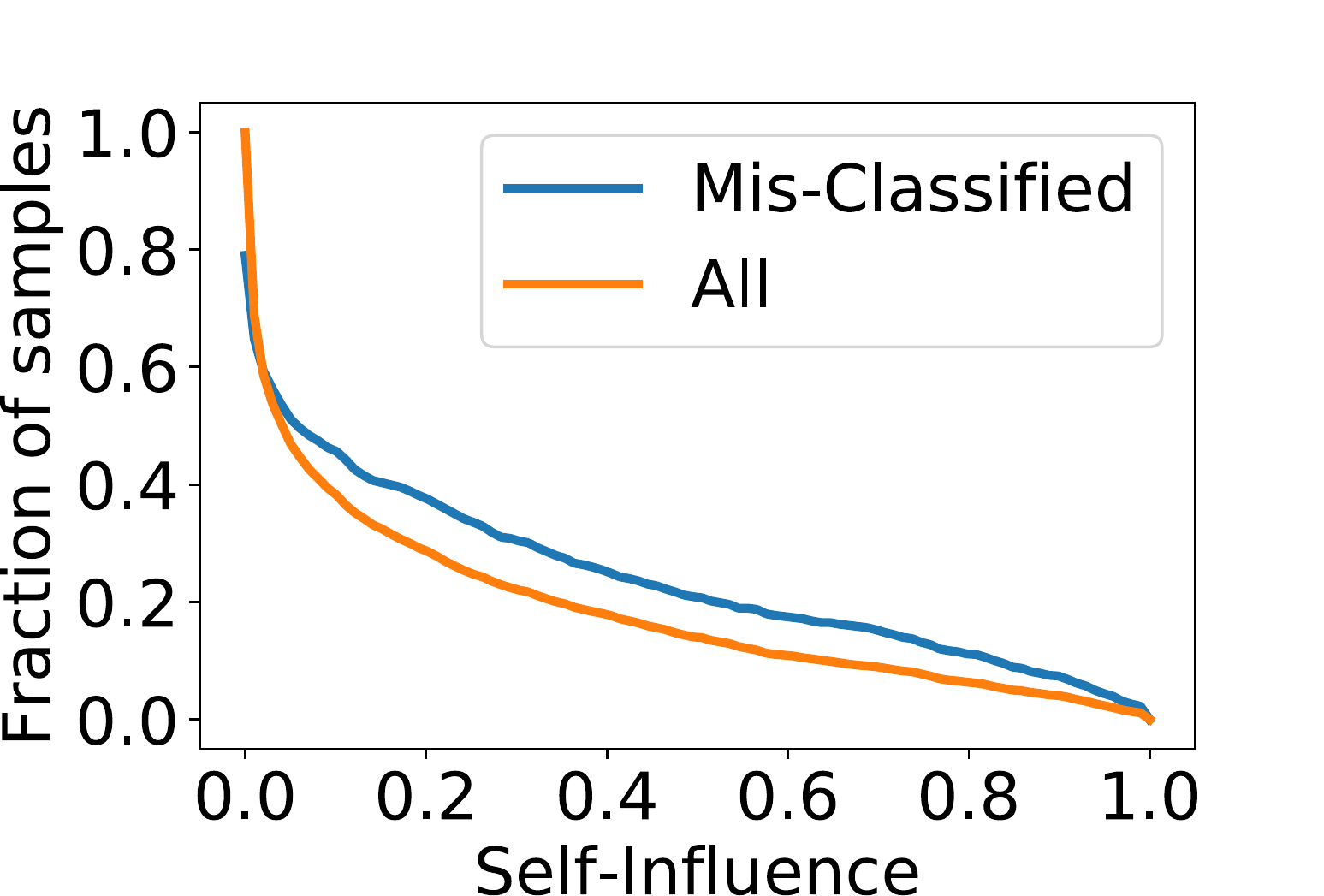
        \caption*{DEER}     
      \end{subfigure}
    \end{subfigure}
      \begin{subfigure}[t]{1.0\linewidth}
      \begin{subfigure}[t]{0.19\linewidth}
        \centering \def\svgwidth{0.99\linewidth}
        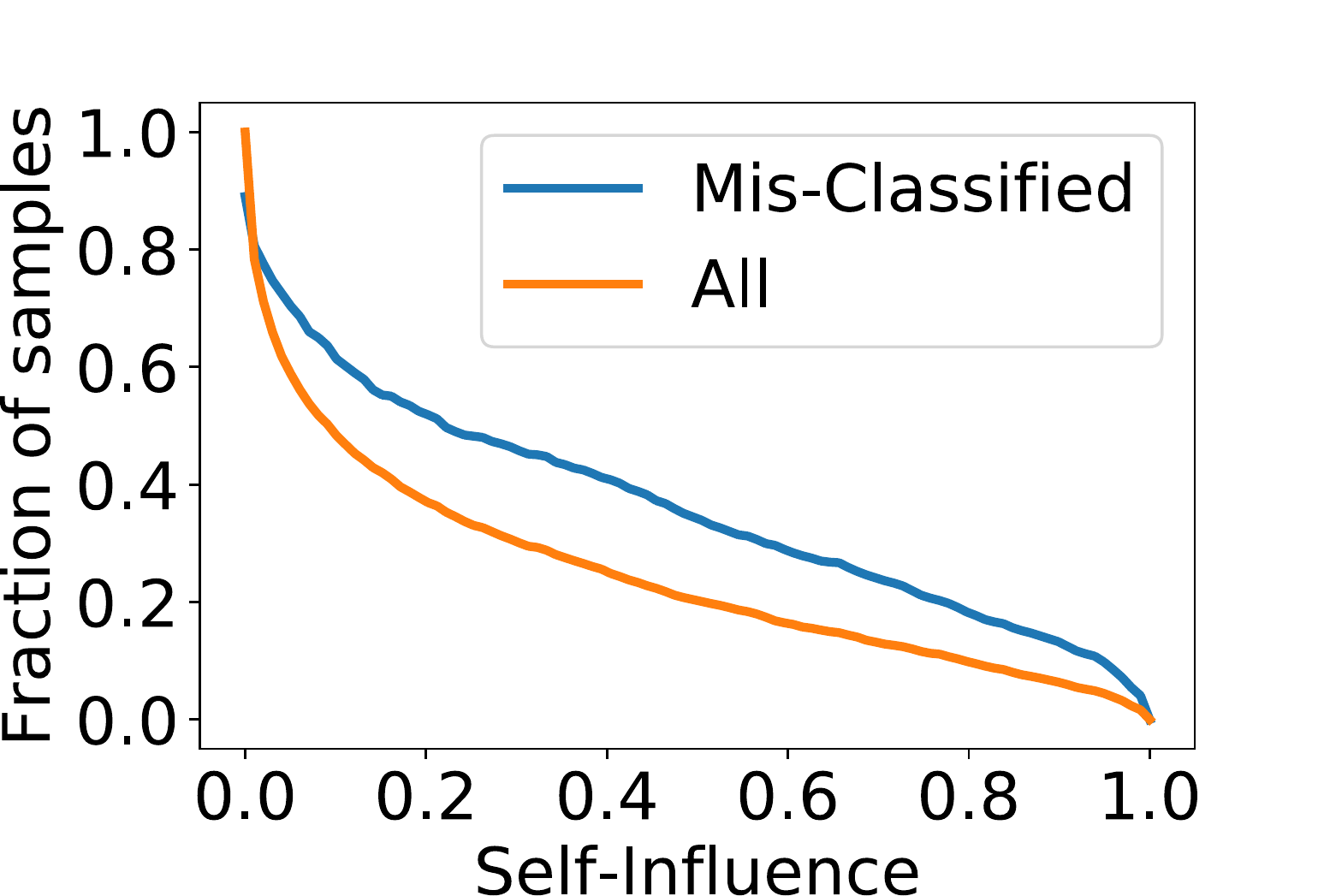
        \caption*{DOG}      
      \end{subfigure}
      \begin{subfigure}[t]{0.19\linewidth}
        \centering \def\svgwidth{0.99\linewidth}
        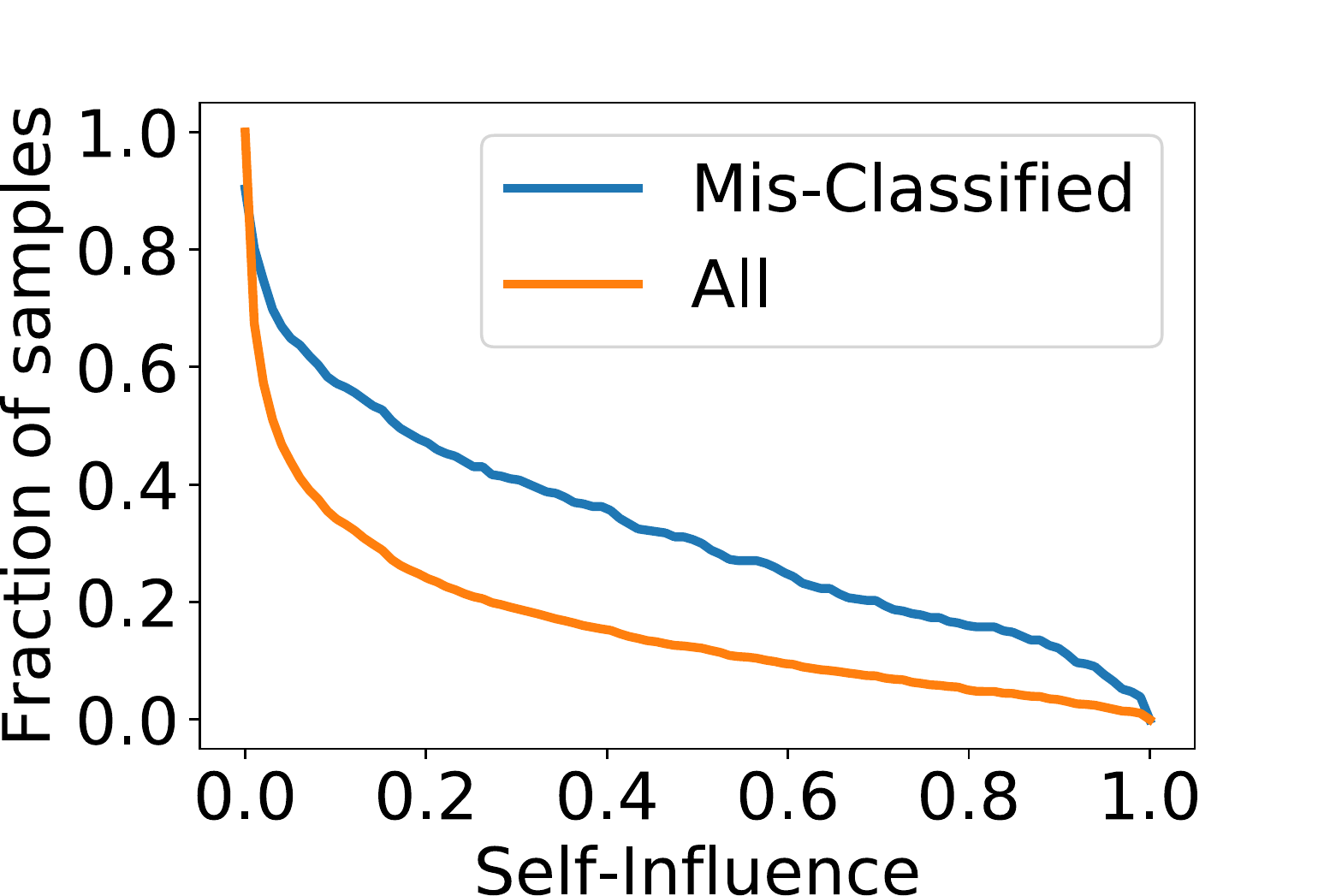
        \caption*{FROG}      
      \end{subfigure}
      \begin{subfigure}[t]{0.19\linewidth}
        \centering \def\svgwidth{0.99\linewidth}
        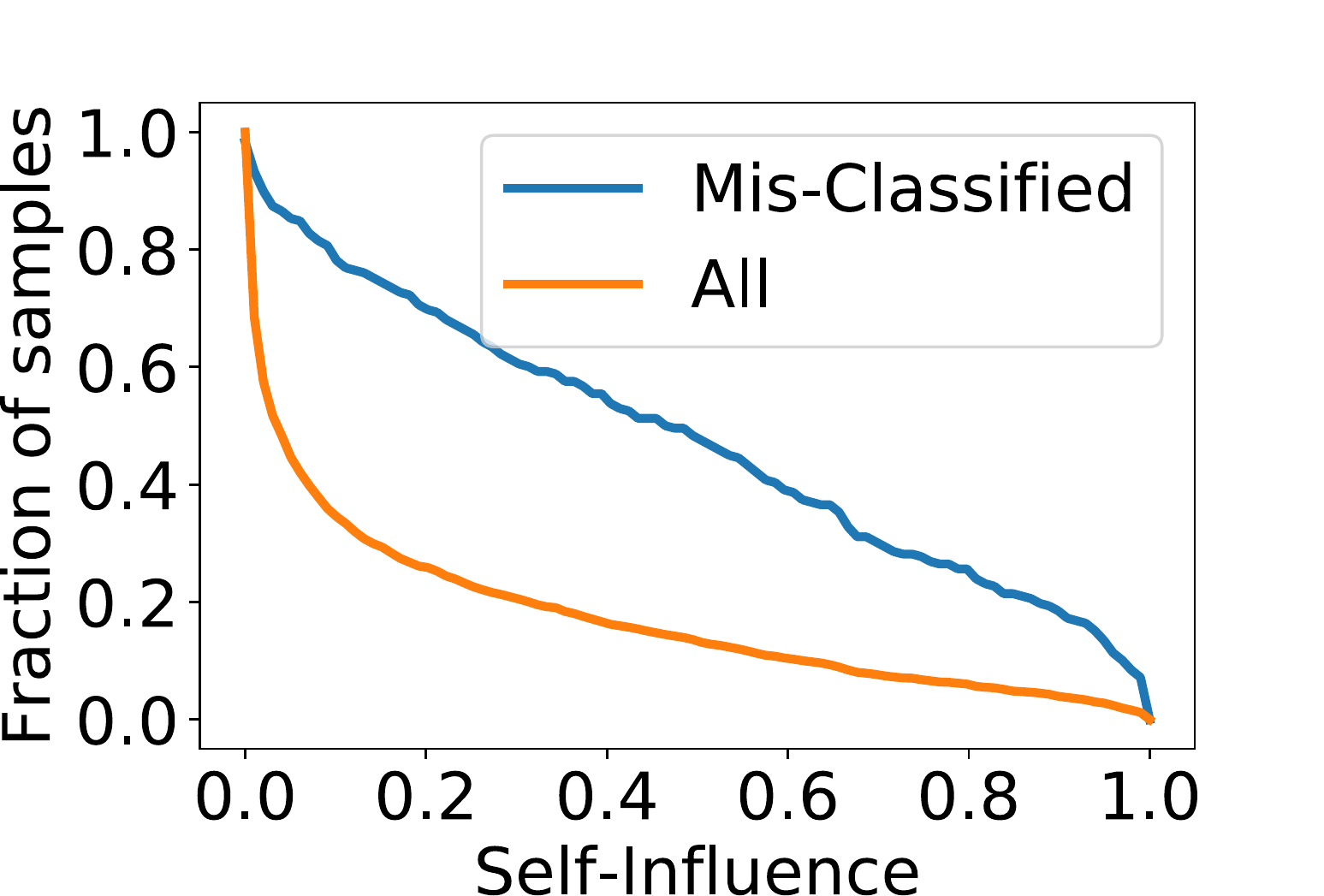
        \caption*{HORSE}     
      \end{subfigure}
      \begin{subfigure}[t]{0.19\linewidth}
        \centering \def\svgwidth{0.99\linewidth}
        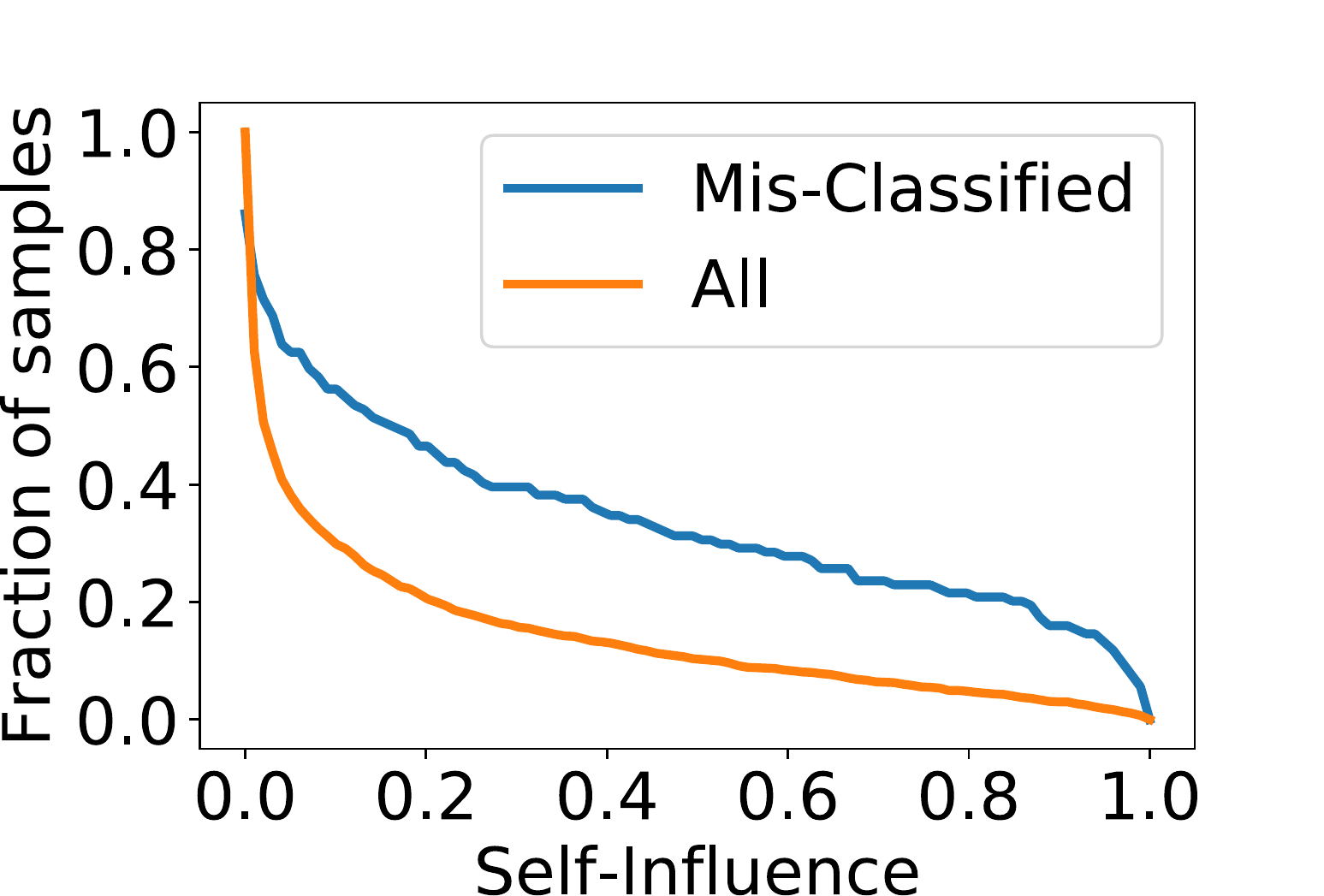
        \caption*{SHIP}
      \end{subfigure}
      \begin{subfigure}[t]{0.19\linewidth}
        \centering \def\svgwidth{0.99\linewidth}
        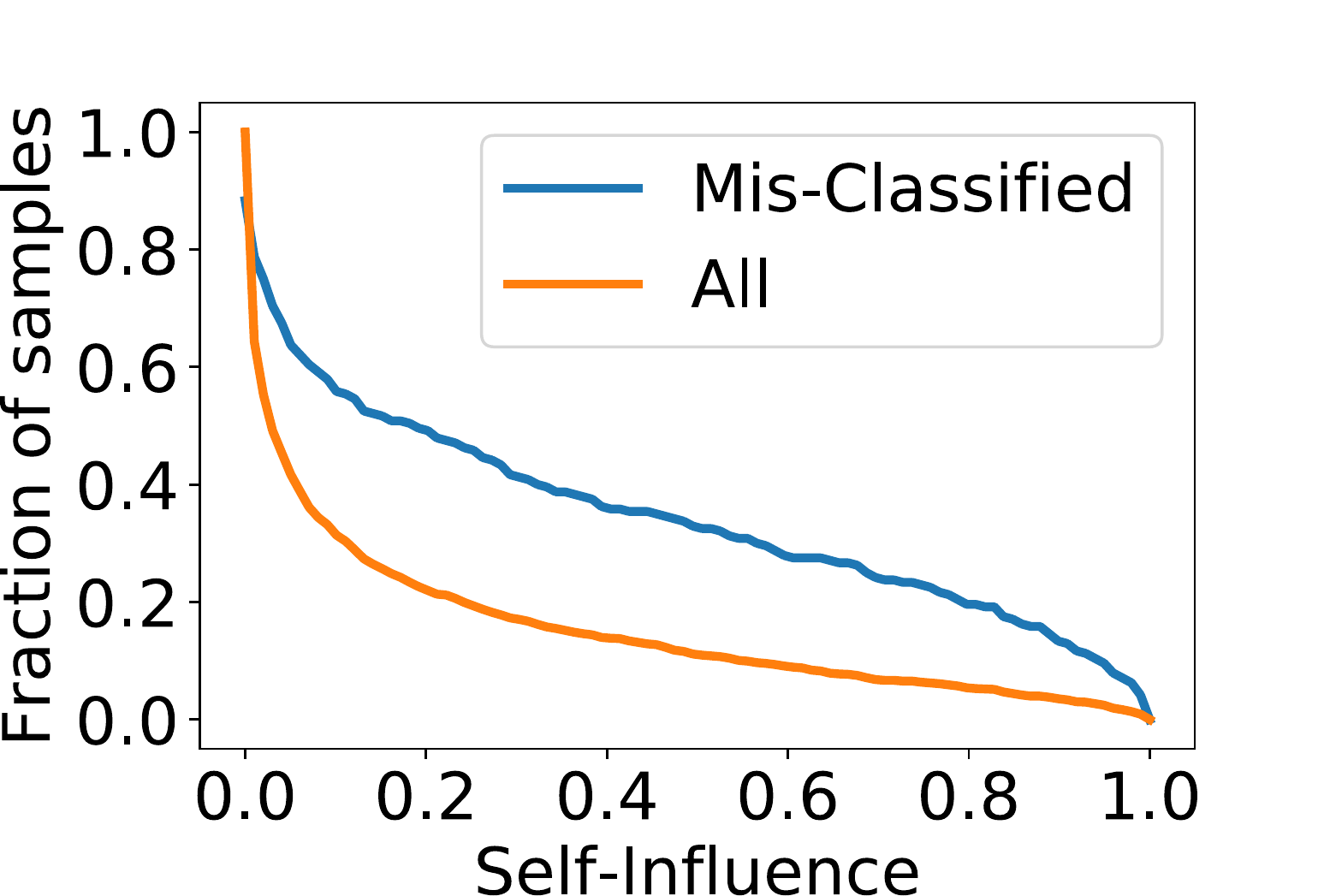
        \caption*{TRUCK}
      \end{subfigure}
    \end{subfigure}
    \caption{Fraction of train points that have a
      self-influence greater than $s$ is plotted versus $s$. }
  \label{fig:cifar_self_influence}
  \end{subfigure}\hfill
  \begin{subfigure}[b]{0.29\linewidth}
    \centering \def\svgwidth{0.99\linewidth}
    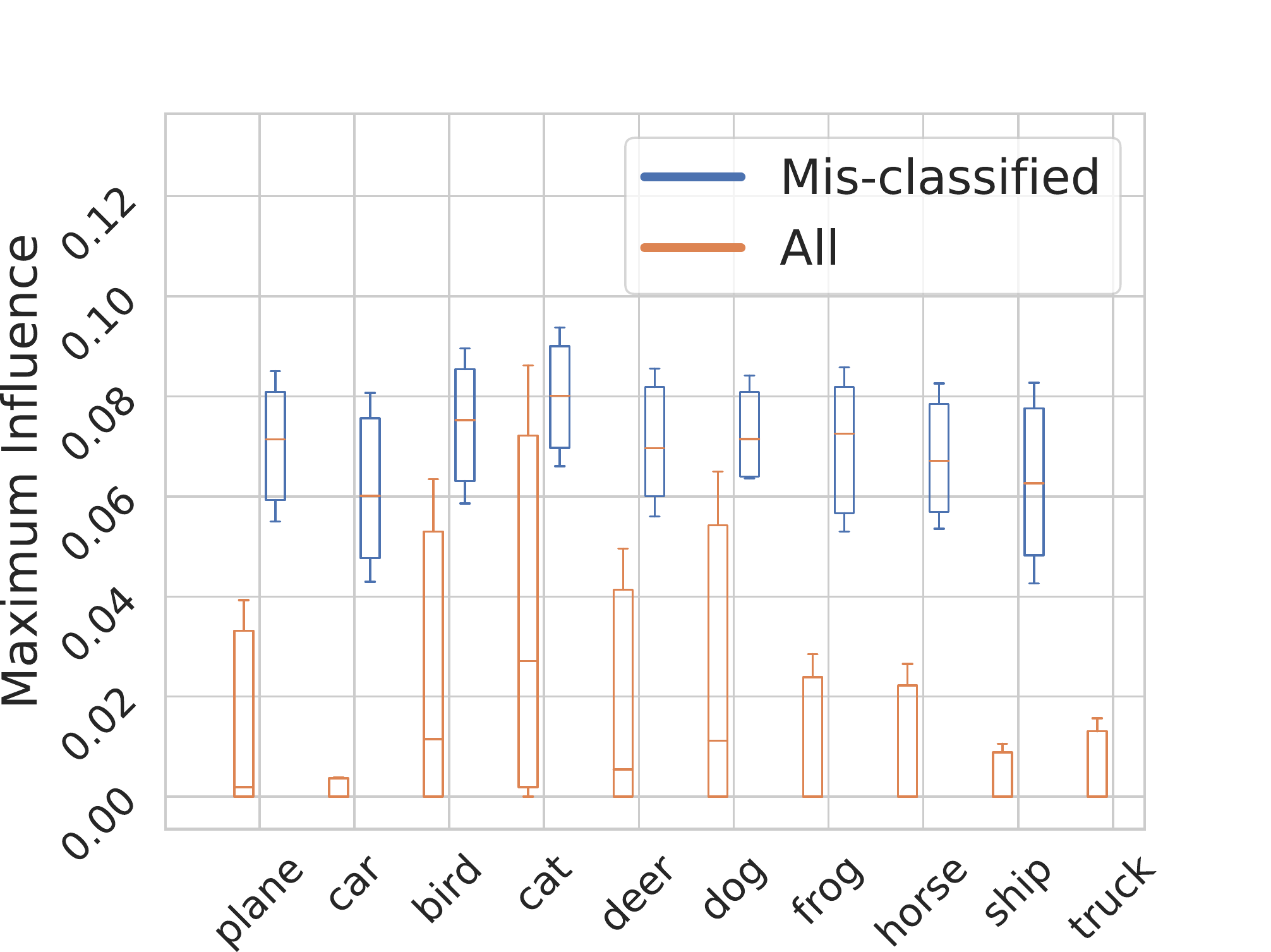
    \caption{Distribution of the influence of training point  on 
    all test points compared to the distribution of influence on test
    points mis-classified by adversarially trained points. }
    \label{fig:infl-cifar10_tr_te_adv}
  \end{subfigure}
  \caption[Adversarial
  training ignores high training points with high self-influence]{The blue represents the points misclassified by an
  adversarially trained model. The orange represents the distribution
  for all points in the dataset~(of the concerned class for CIFAR10).
  }
  \label{fig:infl-cifar10-mem}
  \end{figure}
  
\begin{remark}\todo[color=green]{Comment how any method that does not memorise will suffer from this}
We remark that this behaviour of \AT and TRADES seemingly induces a tradeoff
between natural test error and adversarial error in the presence of label noise.
However, this is under the assumption that the learning algorithm cannot
distinguish between examples with label noise and atypical examples. Our
experiments indicate that \AT and TRADES is unable to distinguish between them.
However, if an algorithm can distinguish between them then this
specific kind of tradeoff should not arise although there might still be
tradeoffs for other reasons.

This drop in test accuracy due to not memorising rare or atypical samples is not
specific to \AT and TRADES but seen in other learning algorithms, that avoid
memorisation, as well. In particular, this has been observed in differentially
private training~\citep{bagdasaryan19} and sparse training~\citep{hooker2019compressed} as well.\todo[color=green]{COmplete
this citation}
\end{remark}
\section{Theoretical results on the impact of representation learning}
\label{sec:theoretical-setting}
A large portion of the success of modern machine learning methods has
largely been made possible by incorporating proper inductive biases,
and that has helped achieve better generalisation and accelerate
optimisation. 
 In this section, we will
look at the harmful impacts of choosing the incorrect inductive bias.
In particular, we look at \emph{improper} representation learning by
way of incorrect inductive biases as a source of adversarial
vulnerability.

 Recent works~\citep{tsipras2018robustness,Zhang2019} have argued that the
trade-off between robustness and accuracy might be unavoidable. However, their
setting involves a distribution that is not robustly separable  by any
classifier. In such a situation there is indeed a trade-off between robustness
and accuracy. In this section, we focus on settings where robust classifiers
exist, which is a more realistic scenario for real-world data. At least for
vision, one may well argue that ``humans'' are robust classifiers, and as a
result, we would expect that classes are well-separated at least in some
representation space. In fact, \citet{Yang2020} show that classes are  already
well-separated in the input space. In such situations, there is no need for
robustness to be at odds with accuracy. A more plausible scenario which we
posit, and provide theoretical evidence in support of
in~\Cref{thm:repre-par-inter}, is that depending on the choice of
representations, the trade-off may exist or can be avoided. Recent empirical
work~\citep{Mao2020} including our work in~\Cref{chap:low_rank_main} has also
established that modifying the training objective to favour certain inductive
bias in the learned representations can automatically lead to improved
robustness.

On a related note, it has been suggested in recent works that
adversarially robust learning may require more ``complex'' 
decision boundaries, and as a result may require more data%
~\citep{Shah2020,schmidt2018adversarially,
pmlr-v97-yin19b,madry2018towards}. 
However, the question of
decision boundaries in neural networks is subtle as 
the network learns a \emph{feature representation} as well as a decision
 boundary on top of it. 
We develop  concrete theoretical examples
in~\Cref{thm:repre-par-inter,thm:parity_robust_repre_all} to establish
that choosing one feature representation over another may lead to
\emph{visually} more complex decision boundaries
on the input space, though these are not necessarily more complex in terms of
statistical learning theoretic concepts such as VC dimension.

\subsection{Representation learning without label noise}
\label{sec:repr-learn-no-lbl-noise}
The choice of inductive biases incorporated in a model affects
representations and introduces desirable and possibly even undesirable
(cf.~\citep{Liu2018}) invariances; for example, training convolutional
networks are invariant to (some) translations, while training fully
connected networks are invariant to permutations of input features.
This means that fully connected networks can learn even if the pixels
of each training image in the training set are permuted with a fixed
permutation~\citep{Zhang2016}. This invariance is worrying as it means
that such a network can effectively classify a matrix~(or tensor) that
is visually nothing like a real image into an image category.

In this section, we present a result to show that there exists a data
distribution where proper representation is necessary  for small adversarial
error as well as small test error whereas another representation can provide low
test error but necessarily have large adversarial error. Interestingly, the
representation that can achieve small adversarial error can look visually more
complex due to the larger number of  distinct linear regions in its decision
boundary. However, statistically, it will have a smaller VC dimension than its
counterpart. We first present the theorems with a proof sketch for ease of
understanding and the more detailed proofs in~\Cref{sec:proof-22}.
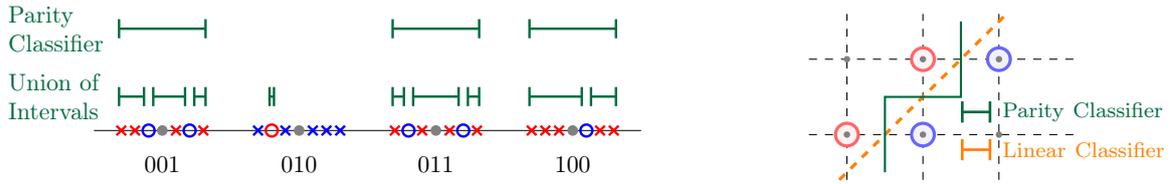
\begin{figure}[t]
  \begin{subfigure}[t]{0.67\linewidth}
    \scalebox{0.9}{\begin{tikzpicture}
      \draw (1.,0) -- (9,0); \filldraw [gray] (2,0) circle (2pt);
      \filldraw [gray] (4,0) circle (2pt); \filldraw [gray] (6,0)
      circle (2pt); \filldraw [gray] (8,0) circle (2pt);

      \draw[|-|, cadmiumgreen, line width=1pt] (1.35,0.5) --
      (1.75,0.5); \draw[|-|, cadmiumgreen, line width=1pt] (1.85,0.5)
      -- (2.35,0.5); \draw[|-|, cadmiumgreen, line width=1pt]
      (2.45,0.5) -- (2.65,0.5);

      \draw[|-|, cadmiumgreen, line width=1pt] (1.35,1.5) --
      (2.65,1.5);

      \draw (1.4,0 ) node[cross,red,line width=1pt] {}; \draw (1.6,0 )
      node[cross,red,line width=1pt] {}; \draw (2.2,0 )
      node[cross,red,line width=1pt] {}; \draw (2.6,0 )
      node[cross,red,line width=1pt] {};
      \draw[blue, line width=1pt] (1.8,0 ) circle (2.5pt); \draw[blue,
      line width=1pt] (2.4,0 ) circle (2.5pt); \node at (2,-0.5)
      {001};

      \draw[|-|, cadmiumgreen, line width=1pt] (3.55,0.5) --
      (3.65,0.5);

      \draw (2+1.4,0 ) node[cross,blue,line width=1pt] {}; \draw
      (2+1.8,0 ) node[cross,blue,line width=1pt] {}; \draw (2+2.2,0 )
      node[cross,blue,line width=1pt] {};
      \draw (2+2.4,0 ) node[cross,blue,line width=1pt] {}; \draw
      (2+2.6,0 ) node[cross,blue,line width=1pt] {}; \draw[red, line
      width=1pt] (2+1.6,0 ) circle (2.5pt); \node at (4,-0.5) {010};

      \draw[|-|, cadmiumgreen, line width=1pt] (2+2+1.35,1.5) --
      (2+2+2.65,1.5);

      \draw[|-|, cadmiumgreen, line width=1pt] (2+2+1.35,0.5) --
      (2+2+1.55,0.5); \draw[|-|, cadmiumgreen, line width=1pt]
      (2+2+1.65,0.5) -- (2+2+2.35,0.5); \draw[|-|, cadmiumgreen, line
      width=1pt] (2+2+2.45,0.5) -- (2+2+2.65,0.5);

      \draw (2+2+1.8,0 ) node[cross,red,line width=1pt] {}; \draw
      (2+2+1.4,0 ) node[cross,red,line width=1pt] {}; \draw (2+2+2.2,0
      ) node[cross,red,line width=1pt] {};
      \draw (2+2+2.6,0 ) node[cross,red,line width=1pt] {};
      \draw[blue, line width=1pt] (2+2+1.6,0 ) circle (2.5pt);
      \draw[blue, line width=1pt] (2+2+2.4,0 ) circle (2.5pt); \node
      at (6,-0.5) {011};

      \draw[|-|, cadmiumgreen, line width=1pt] (2+2+2+1.35,1.5) --
      (2+2+2+2.65,1.5);

      \draw[|-|, cadmiumgreen, line width=1pt] (2+2+2+1.35,0.5) --
      (2+2+2+2.15,0.5); \draw[|-|, cadmiumgreen, line width=1pt]
      (2+2+2+2.25,0.5) -- (2+2+2+2.65,0.5);

      \draw (2+2+2+1.4,0 ) node[cross,red,line width=1pt] {}; \draw
      (2+2+2+1.6,0 ) node[cross,red,line width=1pt] {}; \draw
      (2+2+2+1.8,0 ) node[cross,red,line width=1pt] {}; \draw
      (2+2+2+2.4,0 ) node[cross,red,line width=1pt] {}; \draw
      (2+2+2+2.6,0 ) node[cross,red,line width=1pt] {}; \draw[blue,
      line width=1pt] (2+2+2+2.2,0 ) circle (2.5pt); \node at (8,-0.5)
      { 100};

      \node[text width=1.5cm,color=cadmiumgreen,line width=2pt] at
      (0.5,0.5) {Union of Intervals}; \node[text
      width=1.5cm,color=cadmiumgreen,line width=2pt] at (0.5,1.5)
      {Parity Classifier}; \end{tikzpicture}}\caption{Both Parity and
      Union of Interval classifier predict {\color{red} red} if
      inside any {\color{cadmiumgreen}~green} interval and
      {\color{blue} blue} if outside all intervals. The $\times$-es
      are correctly labelled and the $\circ$-es are mislabeled
      points. Reference integer points on the line labelled in
      \emph{binary}.}\label{fig:thm-3} \end{subfigure}\hfill
  \begin{subfigure}[t]{0.29\linewidth}
    \begin{tikzpicture}
      \draw[thin, dashed] (-0.5,0) -- (3,0);
      \draw[thin, dashed] (-0.5,1) -- (3,1);
      \draw[thin, dashed] (2,-0.6) -- (2,1.6);
      \draw[thin, dashed] (1,-0.6) -- (1,1.6); 
      \draw[thin, dashed] (0,-0.6) -- (0,1.6);

      \filldraw[color=red!60, fill=red!5, very thick](0,0) circle
      (.15); \filldraw[color=red!60, fill=red!5, very thick](1,1)
      circle (.15); \filldraw[color=blue!60, fill=blue!5, very
      thick](1,0) circle (.15); \filldraw[color=blue!60, fill=blue!5,
      very thick](2,1) circle (.15); \draw[orange, dashed, very thick,
      name  path=plane] (-0.1, -0.6) -- (1.5, 1) -- (2.1, 1.6);
      \draw[cadmiumgreen, thick] (0.5, -0.5) -- (0.5, 0.5) -- (1.5,
      0.5) -- (1.5, 1.5);
     \draw[|-|, cadmiumgreen, line width=1pt]
      (1.5,0.3) -- (1.9,0.3);
      \draw[|-|, orange, line width=1pt]
     (1.5,-0.2) -- (1.9,-0.2);

       \filldraw [gray] (0,0) circle (1pt); \filldraw [gray] (0,1)
      circle (1pt); \filldraw [gray] (1,0) circle (1pt); \filldraw
      [gray] (1,1) circle (1pt); \filldraw [gray] (2,0) circle (1pt);
      \filldraw [gray] (2,1) circle (1pt); \node[text
      width=2.3cm,color=cadmiumgreen] at (3.2,0.3) {\scriptsize Parity
      Classifier}; \node[text width=2.3cm,color=orange] at (3.2,-0.2)
      {\scriptsize Linear Classifier};
    \end{tikzpicture}
   \caption{Robust generalisation needs more complex boundaries}
   \label{fig:complex_simple}
  \end{subfigure}\caption[Illustration
    of~\Cref{thm:parity_robust_repre_all,thm:repre-par-inter}]{Visualisation
    of the distribution and classifiers used in the Proof
    of~\Cref{thm:parity_robust_repre_all,thm:repre-par-inter}.~The
    {\color{red}Red} and {\color{blue}Blue} indicate the two classes.}
    
\end{figure}

\begin{restatable}[Representation Learning for Adversarial Robustness]{thm}{parityrobustrepre}
	\label{thm:parity_robust_repre_all} For some universal constant $c$,
	and any $0 < \gamma_0 < 1/\sqrt{2}$, there exists a family of
	distributions $\cD$ defined on $\cX\times\bc{0,1}$ where
	$\cX\subseteq\reals^2$ such that for all distributions $\cP\in\cD$,
	and denoting by $\cS_m
	=\bc{\br{\vec{x}_1,y_1},\cdots,(\vec{x}_m,y_m)}$ a sample of size
	$m$ drawn i.i.d. from $\cP$, 
  \begin{enumerate}[itemsep=-0.3em,leftmargin=*]
	  \item[(i)] For any $m \geq 0$, $\cS_m$ is linearly separable i.e.,
		  $\forall(\vec{x}_i, y_i) \in \cS_m$, there exist
		  $\vec{w}\in\reals^2, w_0\in\reals$ s.t.
		  $y_i\br{\vec{w}^\top\vec{x}_i+w_0}\ge 0$. Furthermore, for every
		  $\gamma > \gamma_0$, any linear separator $f$ that perfectly
		  fits the training data $\cS_m$ has $\radv{\gamma}{f; \cP} \geq
		  0.0005$, even though $\risk{\cP}{f} \rightarrow 0$ as $m
		  \rightarrow \infty $.
	  \item[(ii)] There exists a function class $\cH$ such that for some
		  $m \in O(\log(\delta^{-1}))$, any $h \in \cH$ that perfectly
		  fits the $\cS_m$, satisfies with probability at least $1 -
		  \delta$, $\risk{\cP}{h} = 0$ and $\radv{\gamma}{h; \cP} = 0$,
		  for any $\gamma \in [0, \gamma_0 + 1/8]$.
  \end{enumerate}
\end{restatable}

A complete proof of this result appears in~\Cref{sec:proof-22}, but
first, we provide a sketch of the key idea here.%
~The distributions in family $\cD$ will be supported on balls of radius at most
$1/\sqrt{2}$ on the integer lattice in $\reals^2$. The \emph{true} class label
for any point $\vec{x}$ is provided by the parity of $a + b$, where $(a, b)$ is
the lattice point closest to $\vec{x}$. However, the distributions in $\cD$ are
chosen to be such that there is also a linear classifier that can separate these
classes, e.g. a distribution only supported on balls centred at the points $(a,
a)$ and $(a, a+1)$ for some integer $a$~(See~\Cref{fig:complex_simple}).
\emph{Visually} learning the classification problem using the parity of $a + b$
results in a seemingly more complex decision boundary, a point that has been
made earlier regarding the need for more complex boundaries to achieve
adversarial robustness~\citep{degwekar19a,Shah2020}. However, it is worth noting
that this complexity is not rooted in any \emph{statistical theory}, e.g. the VC
dimension of the classes considered in Theorem~\ref{thm:parity_robust_repre_all}
is essentially the same (even lower for $\cH$ by $1$). This \emph{visual}
complexity arises purely because the linear classifier looks at a geometric
representation of the data whereas the parity classifier looks at the binary
representation of the sum of the nearest integer of the coordinates. In the case
of neural networks, recent works~\citep{kamath2020invariance} have indeed
provided empirical results to support that excessive invariance (eg. rotation
invariance) increases adversarial error.

\subsection{Representation learning with label noise}
\label{sec:repr-theorey-lbl-noise}

In this section, we show how  the choice of representation is
important in the presence of label noise to learn an adversarially
robust classifier. Informally, we show that if the \emph{correct}
representation is used, then in the presence of label noise, it will
be impossible to fit the training data perfectly, but the 
classifier that best fits the training data 
will have good test accuracy and adversarial accuracy. However, using an
``incorrect'' representation, we show that it is possible to find a classifier
that has zero training error, has good test accuracy, but has a high
\emph{adversarial error}. 
We posit this as a (partial) explanation of why classifiers trained on real data
(with label noise) have good test accuracy, while still being vulnerable to
adversarial attacks.  

\begin{restatable}[Representation Learning in the Presence of Noise]{thm}{robustpossibleful}~\label{thm:repre-par-inter}
	For any $n\in\bZ_+$, there exists a family of distributions $\cD^n$
	over $\reals \times \{0, 1\}$ and function classes $\cC,\cH$, such
	that for any $\cP$ from  
	$\cD^n$, and for any $0 < \gamma < 1/4$, and $\eta \in (0, 1/2)$ if $\cS_m =
        \{(\vec{x}_i, y_i)\}_{i=1}^m$ denotes a sample of size $m$
        drawn from $\cP$ where
        \[m=\bigO{\mathrm{max}\bc{
      n\log{\frac{n}{\delta}} 
   \br{\frac{\br{1-\eta}}{\br{1-2\eta}^2}+1},
   \frac{n}{\eta\gamma^2} 
   \log\br{\frac{n}{\gamma\delta}}}}\]
and if $\cS_{m, \eta}$ denotes the sample where each
label is flipped independently with probability $\eta$. 
  \begin{enumerate}[labelsep=-0.3em,leftmargin=*]
	  \item[(i)]~~the classifier $c \in \cC$ that minimises the training
	  error on $\cS_{m, \eta}$, has $\risk{\cP}{c} = 0$ and
	  $\radv{\gamma}{c; \cP} = 0$ for $0 \leq \gamma < 1/4$.  
	  \item[(ii)]~~there exist $h \in \cH$, $h$ has zero training error on
	  $\cS_{m, \eta}$, and $\risk{\cP}{h} = 0$. However, for any $\gamma > 0$, and
	  any $h \in \cH$ with zero training error on $\cS_{m, \eta}$,
	  $\radv{\gamma}{h; \cP} \geq 0.1$.  
  \end{enumerate}
  Furthermore, the required $c ,h \in \cC,\cH$  can be
  computed in $\bigO{\poly{n},\poly{\frac{1}{\frac{1}{2}-\eta}}, 
  \poly{\frac{1}{\delta}}}$.
\end{restatable}

We sketch the proof here and present the complete proof in~\Cref{sec:proof-22};
as in~\Cref{thm:parity_robust_repre_all}, we will make use of parity functions,
though the key point is the representations used. Let $\cX = [0, N]$, where $N =
2^n$, we consider distributions that are supported on intervals $(i - 1/4, i +
1/4)$ for $i \in \{1, \ldots, N-1 \}$~(See~\Cref{fig:thm-3}), but any such
distribution will only have a small number, $O(n)$, intervals on which it is
supported. The \emph{true} class label is given by a function that depends on
the parity of some hidden subsets $S$ of bits in the bit-representation of the
closest integer $i$,  e.g. as in~\Cref{fig:thm-3} if $S = \{0, 2\}$, then only
the least significant and the third least significant bit of $i$ are examined
and the class label is $1$ if an odd number  of them are $1$ and $0$ otherwise.
Despite the noise, the \emph{correct} label on any interval can be guessed by
using the majority vote and as a result, the correct parity learnt using
Gaussian elimination. (This corresponds to class $\cC$ in
~\Cref{thm:repre-par-inter}.) On the other hand, it is also possible to learn the
function as a union of intervals, i.e. find intervals, $I_1, I_2, \ldots, I_k$
such that any point that lies in one of these intervals is given the label $1$
and any other point is given the label $0$. By choosing intervals carefully, it
is possible to fit \emph{all the training data}, including noisy examples, but
yet not compromise on \emph{test accuracy} (Fig.~\ref{fig:thm-3}). Such a
classifier, however, will be vulnerable to adversarial examples by applying
Theorem~\ref{thm:inf-label}.  A classifier such as a union of intervals ($\cH$ in
Theorem~\ref{thm:repre-par-inter})%
is translation-invariant, whereas the parity classifier is not.  This suggests
that using classifiers, such as neural networks, that are designed to have too
many built-in invariances might hurt its robustness accuracy. 
In~\Cref{sec:robust-train-creat}, we present further experimental
evidence that neural networks trained with SGD learn more
linear-like~(simpler) decision boundaries than is necessary for
obtaining adversarial robustness.

\section{Experimental results on the impact of representation learning}
This section discusses the importance of representation learning for adversarial
robustness in the context of neural networks. In particular, we show that neural
networks are more linear-like than is required for them to be adversarially
robust. Then, we suggest a way to learn richer representations and show that
this helps with adversarial robustness.

\subsection{Complexity of decision boundaries}
\label{sec:robust-train-creat}
When neural networks are trained using SGD like  algorithms they create
classifiers whose decisions boundaries are geometrically much simpler than they
need to be for being adversarially robust. A few recent
studies~\citep{Shah2020,schmidt2018adversarially} have discussed that robustness
might require more complex classifiers.
In~\Cref{thm:parity_robust_repre_all,thm:repre-par-inter} we discussed this
theoretically and also why this might not violate the traditional wisdom of
Occam's Razor. In particular, complex decision boundaries does not necessarily
mean more complex classifiers in statistical notions of complexity like the VC
dimension. In this section, we show through a simple experiment how the decision
boundaries of neural networks are not ``complex'' enough to provide large enough
margins and are thus adversarially much more vulnerable than is possible.

\begin{figure}[!htb]
  \begin{subfigure}[t]{0.24\linewidth}
    \centering \def\svgwidth{0.99\linewidth}
    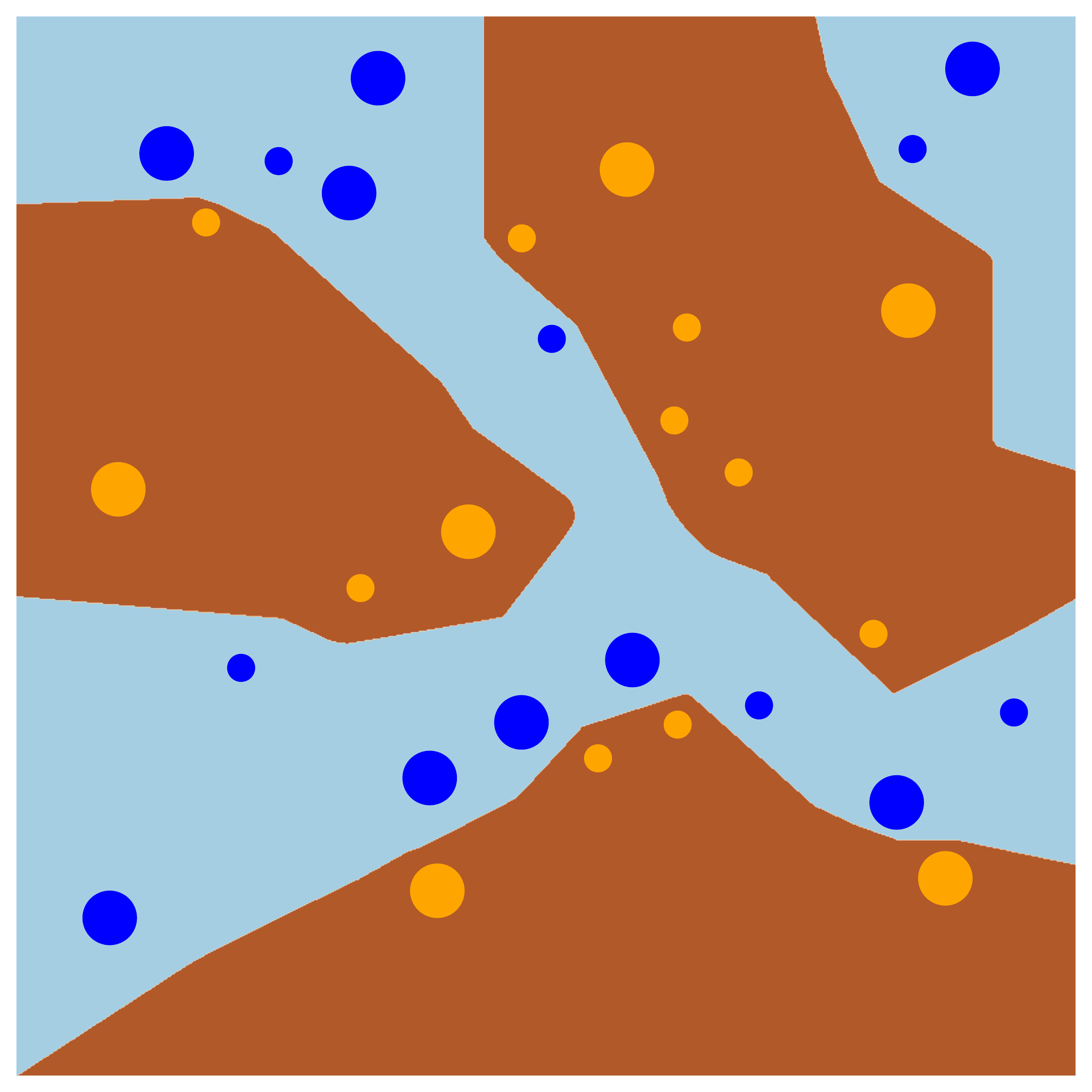
    \caption{Shallow NN}
    \label{fig:dec_shallow}
  \end{subfigure}
  \begin{subfigure}[t]{0.24\linewidth}
    \centering \def\svgwidth{0.99\linewidth}
    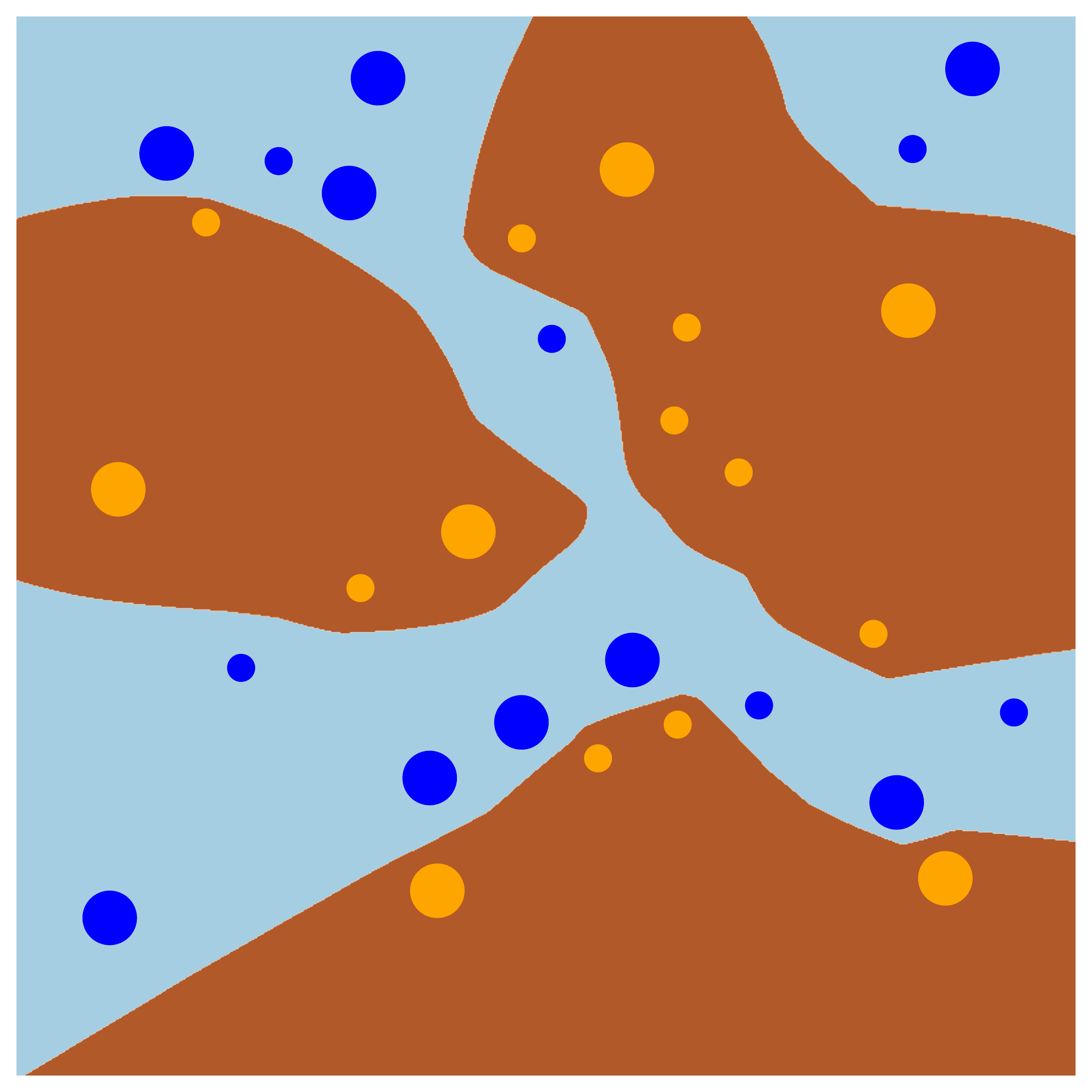
    \caption{Shallow-Wide NN}
    \label{fig:dec_shallow_wide}
  \end{subfigure}
  \begin{subfigure}[t]{0.24\linewidth}
    \centering \def\svgwidth{0.99\linewidth}
    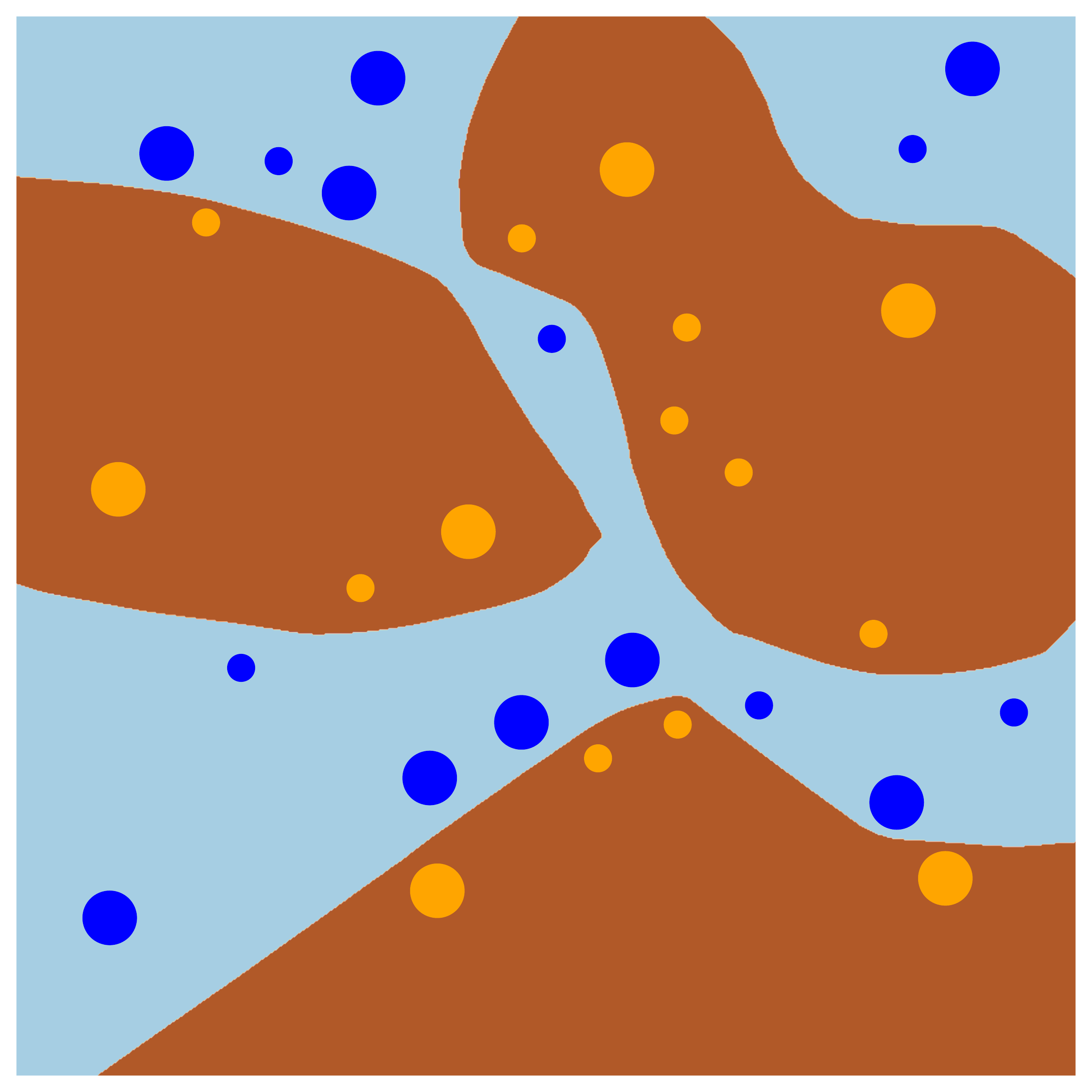
    \caption{Deep NN}
    \label{fig:dec_deep}
  \end{subfigure}
  \begin{subfigure}[t]{0.24\linewidth}
    \centering \def\svgwidth{0.99\linewidth}
    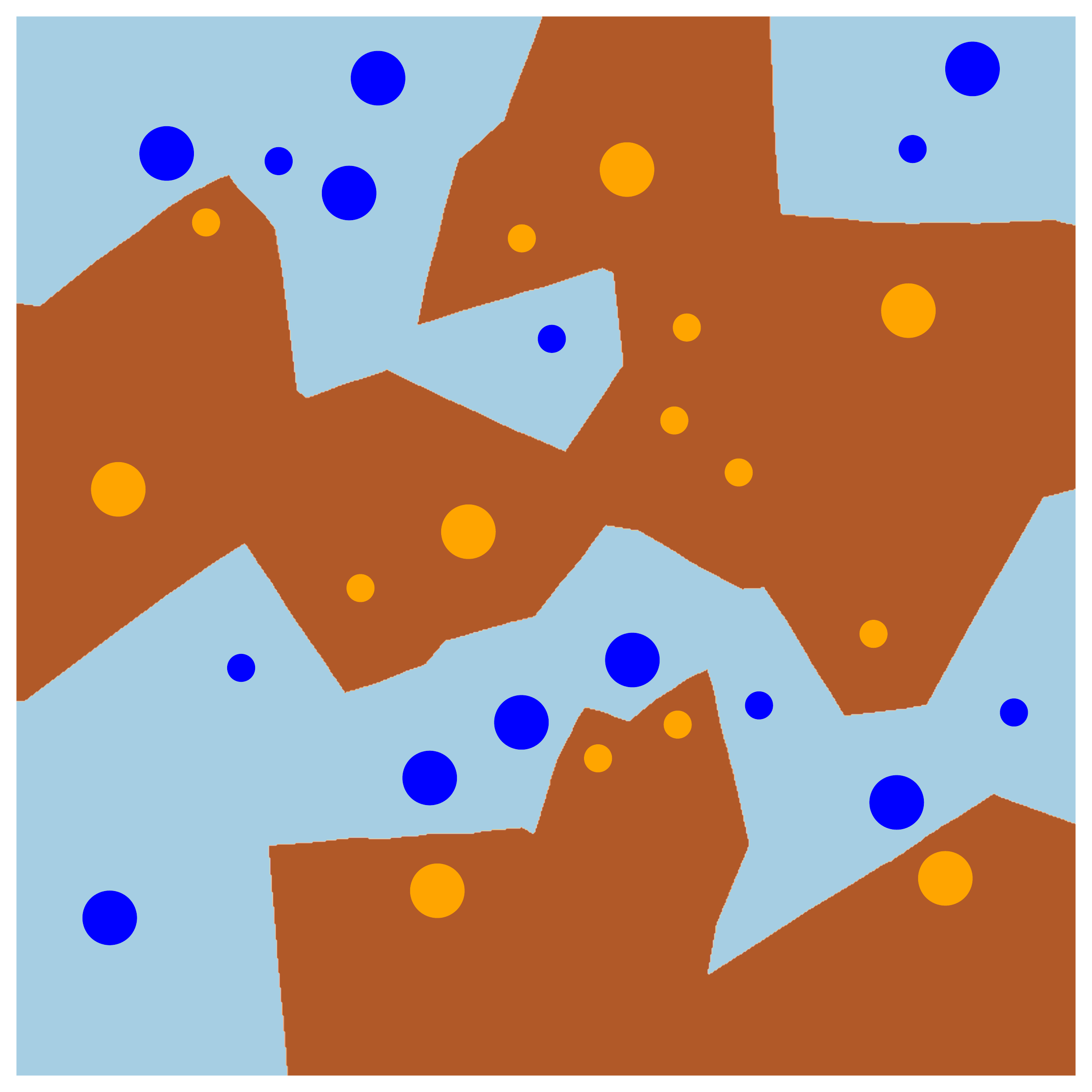
    \caption{Large Margin}
    \label{fig:dec_knn}
  \end{subfigure}
  \caption[Simplicity of Neural Network decision boundaries]{Decision boundaries of neural networks are much simpler
    than they should be.}\label{fig:dec_reg_app_fig}
 \end{figure} 
We train three different neural networks with ReLU activations, a shallow
network~(Shallow NN) with 2 layers and 100 neurons in each layer, a shallow
network with 2 layers and 1000 neurons in each layer~(Shallow-Wide NN), and a
deep network with 4 layers and 100 neurons in each layer. We train them for 200
epochs on a binary classification problem as constructed
in~\Cref{fig:dec_reg_app_fig}.  The distribution is supported on blobs and the
colour of each blob represents its label. On the right side, we have the
decision boundary of a large margin classifier, which is represented by a
1-nearest neighbour algorithm. 

From~\Cref{fig:dec_reg_app_fig}, it is evident that the decision boundaries of
neural networks trained with standard optimisers have far \emph{simpler}
decision boundaries than is needed to be robust~(eg. the 1- nearest neighbour is
much more robust than the neural networks.). In particular, the distinction
between neural networks and the large margin classifier can be noticed clearly
in the decision boundary between the blue and the orange balls in the top left
part of the images in~\Cref{fig:dec_reg_app_fig}. In an effort to have a less
jagged decision boundary for neural networks, the boundary passes very close to
the data~(the blue and orange balls) for the neural networks than it does for
the large margin classifier. This bias towards simplicity for NNs trained using
SGD, we hypothesise, is partly responsible for the increased vulnerability of neural networks.

\subsection{Accounting for sub-populations leads to better
  robustness}
\label{sec:fine-coarse}
One way to evaluate whether more meaningful representations lead to better
robust accuracy is to use training data with more fine-grained labels (e.g.
subclasses of a class); for example, one would expect that if different breeds
of dogs are labelled differently the network will learn features that are
relevant to that extra information. We show using synthetic data,
CIFAR100~\citep{krizhevsky2009learning}, and Restricted
Imagenet~\citep{tsipras2018robustness} that training on fine-grained labels does
increase robust accuracy. 

\begin{figure}[t]%
  \begin{subfigure}[b]{0.35\linewidth}
   \begin{subfigure}[t]{0.48\linewidth}
    \centering \def\svgwidth{0.99\linewidth}
    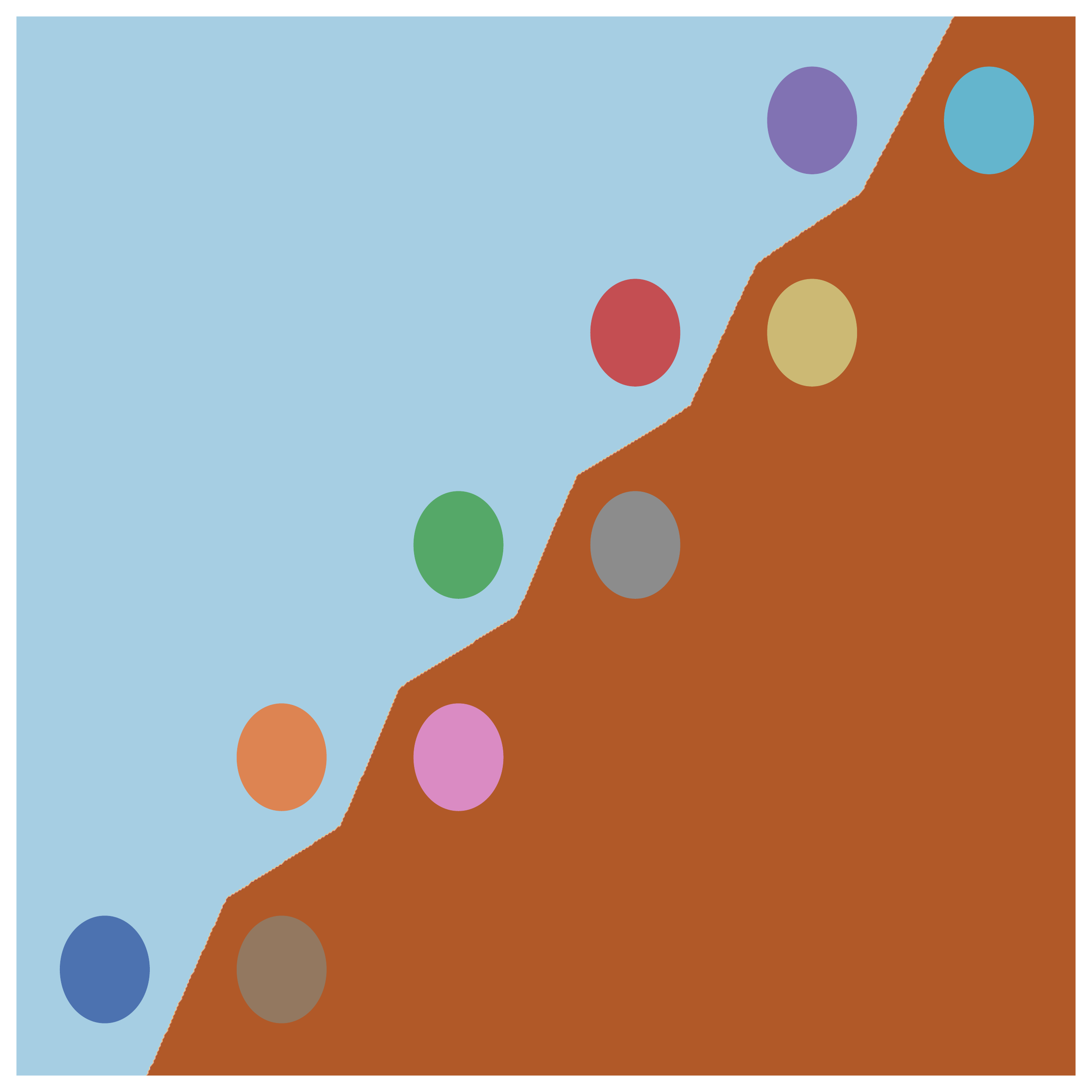
    \caption{MULTICLASS}
    \label{fig:thm_mc}
  \end{subfigure}\hfill
  \begin{subfigure}[t]{0.48\linewidth}
    \centering \def\svgwidth{0.99\linewidth}
    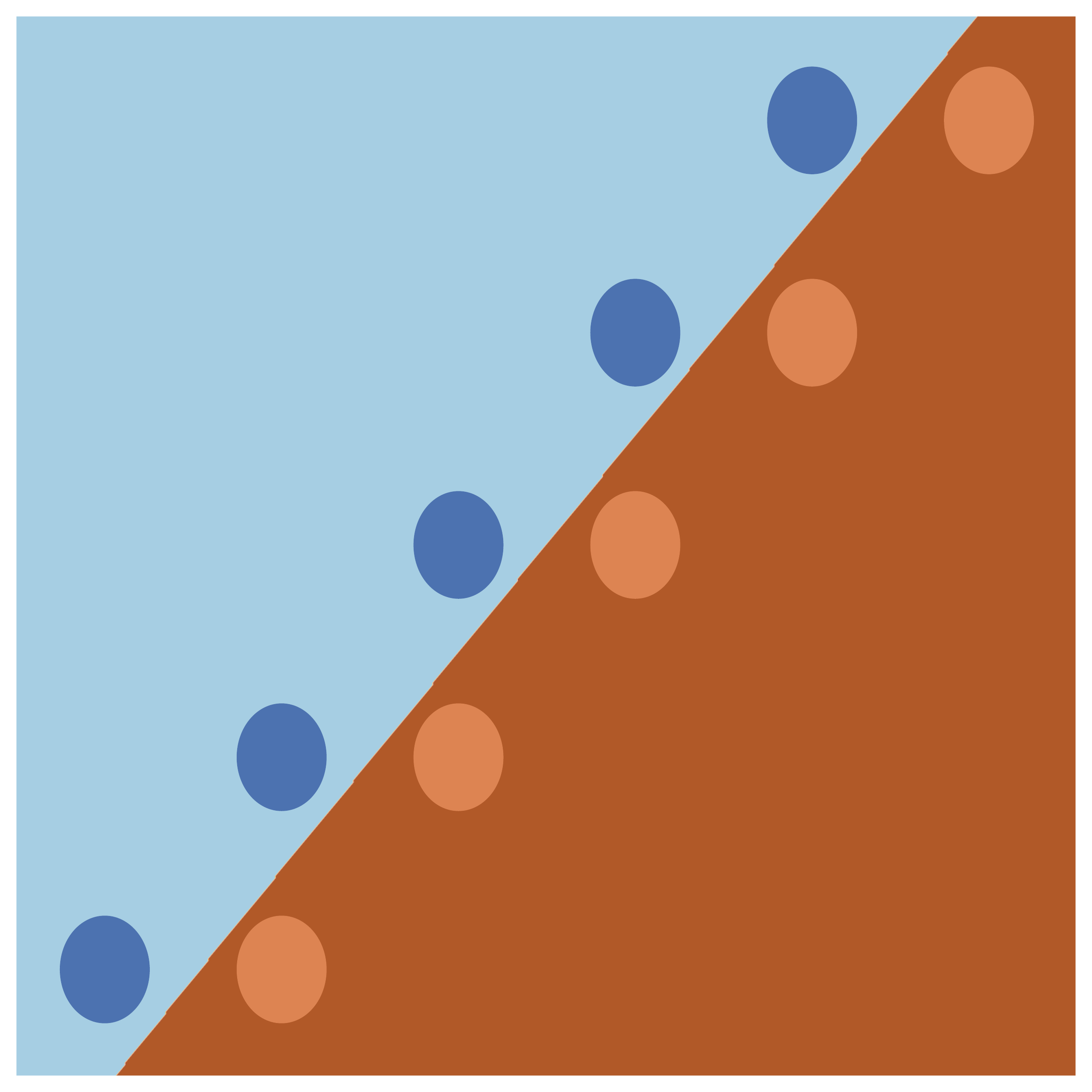
    \caption{NATURAL}
    \label{fig:thm_nat}
  \end{subfigure}
\end{subfigure}\hfill
    \begin{subfigure}[b]{0.60\linewidth}
  \begin{subfigure}[t]{0.49\linewidth}
    \centering \def\svgwidth{0.99\linewidth}
    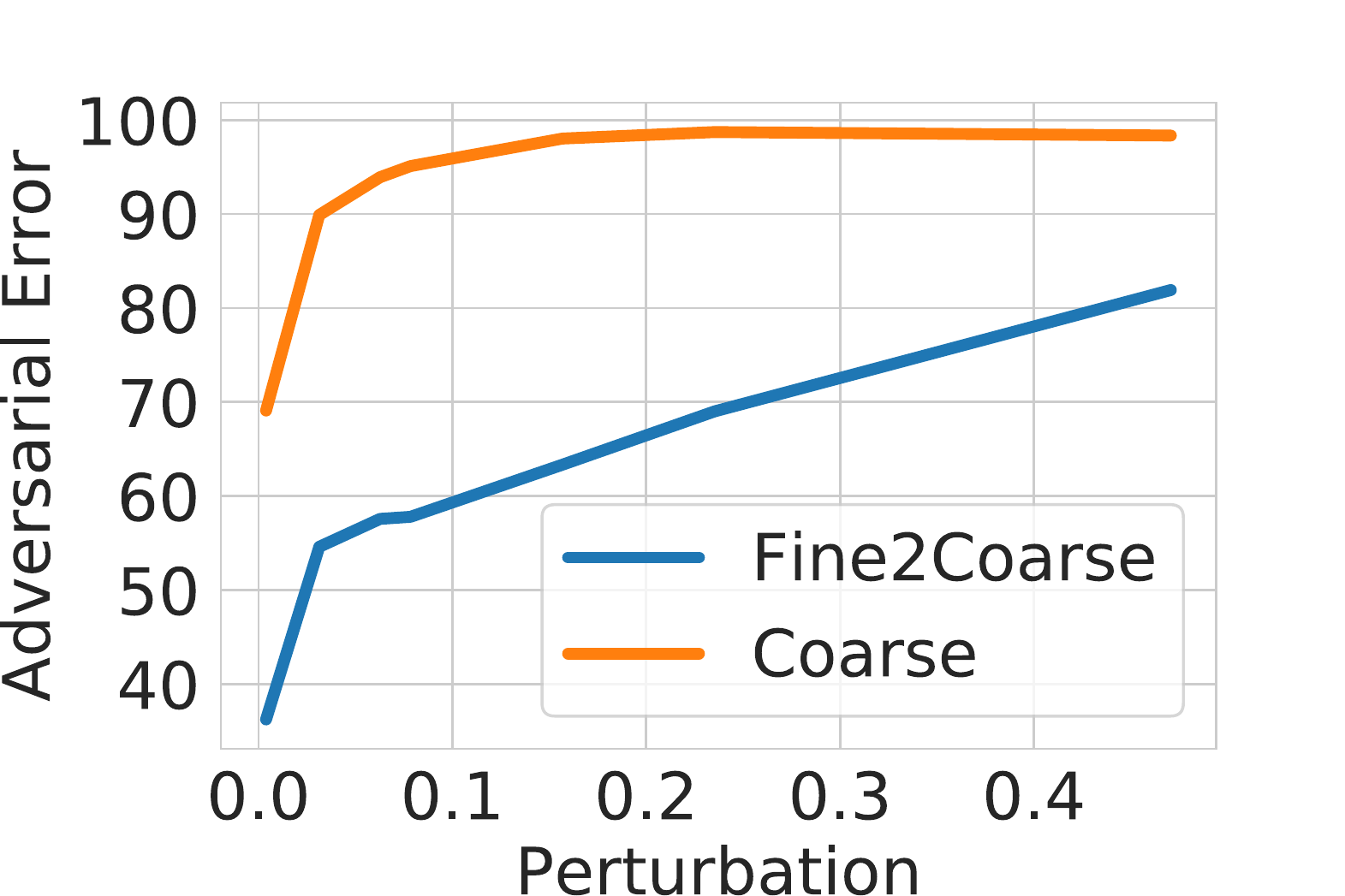
    \caption{ CIFAR-100. %
    }
    \label{fig:fine2coarse}
  \end{subfigure}
  \begin{subfigure}[t]{0.49\linewidth}
    \centering \def\svgwidth{0.99\linewidth}
    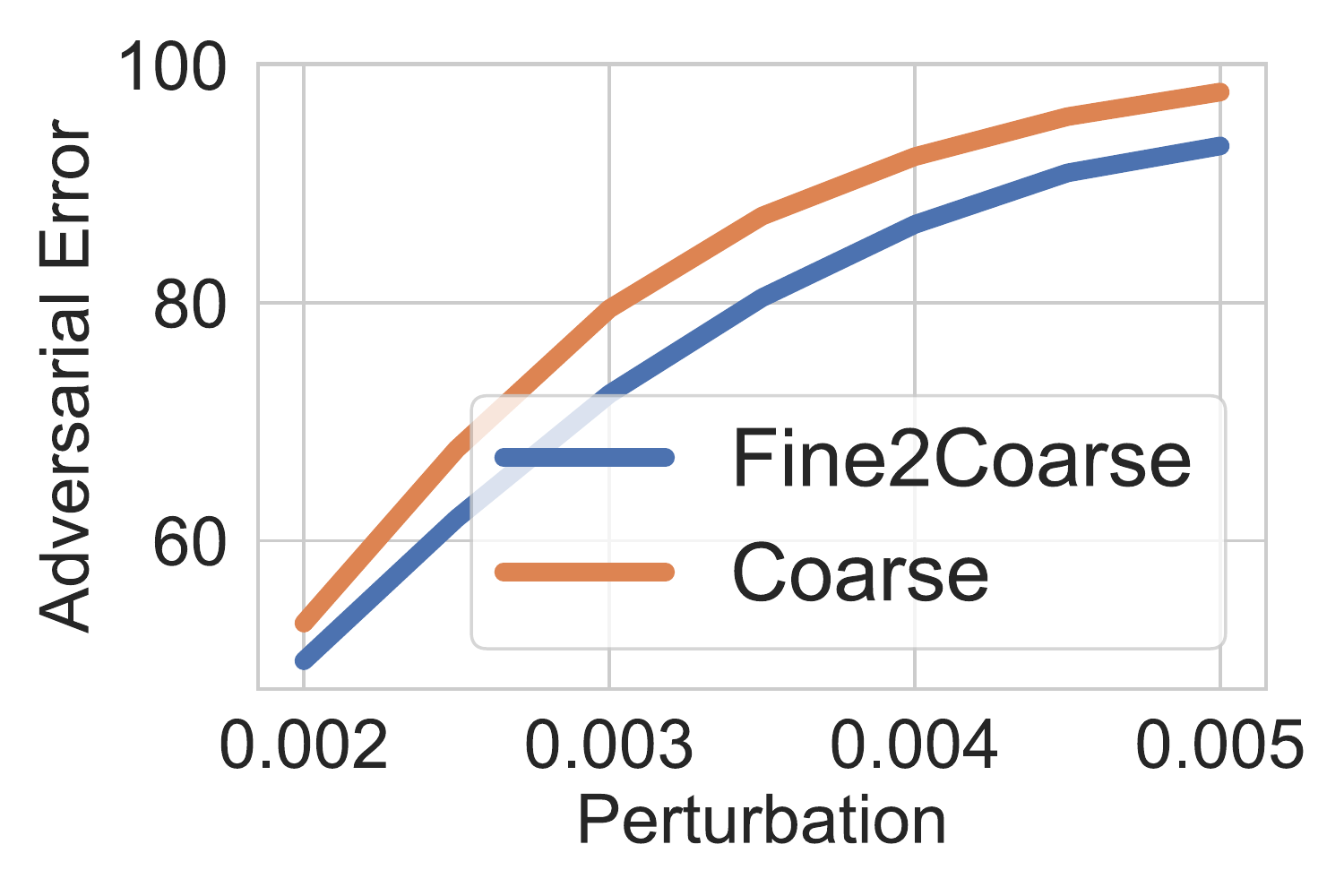
    \caption{Restricted Imagenet %
    }
    \label{fig:fine2coarse-img}
  \end{subfigure}
\end{subfigure}
  \caption[Learning better representation provides better robustness]{Assigning
    a separate class to each sub-population within the original class
    during training increases robustness by learning more meaningful
    representations.}
\end{figure}
We hypothesise that learning more meaningful representations by accounting for
  fine-grained sub-populations within each class may lead to better robustness.
  We use the theoretical setup presented in~\Cref{fig:complex_simple} to conduct
  our synthetic data experiment. 
  
  Recall that the data distributions for the binary learning problem
  in~\Cref{fig:complex_simple} is supported on balls of radius at most
  $1/\sqrt{2}$ on the integer lattice in $\reals^2$. The \emph{true} class label
  for any point $\vec{x}$ is provided by the parity of $a + b$, where $(a, b)$
  is the lattice point closest to $\vec{x}$. As we noted in the proof
  of~\Cref{thm:parity_robust_repre_all}, the distribution is separable by a {\em
  simple} linear classifier by a small margin. However, if each of the circles
  belonged to a separate class then  the decision boundary would have to be
  necessarily more complex as it needs to, now, separate the balls that were
  previously within the same class. We test this hypothesis with two
  experiments. First, we test it on the distribution defined
  in~\Cref{thm:parity_robust_repre_all} where for each ball with label $1$, we
  assign it a different label~(say $\alpha_1,\ldots, \alpha_{k}$) and similarly
  for balls with label $0$, we assign it a different
  label~($\beta_1,\ldots, \beta_k$). Now, we solve a multi-class classification
  problem for $2k$ classes with a deep neural network and then later aggregate
  the results by reporting all $\alpha_i$s as $1$ and all $\beta_i$s as $0$. The
  resulting decision boundary is drawn in~\Cref{fig:thm_mc} along with the
  decision boundary for natural training in~\Cref{fig:thm_nat}. Clearly, the
  margin for the multi-class model (and thus robustness) is greater than the
  naturally trained model.

Second, we also repeat the experiment with CIFAR-100 and Restricted
Imagenet~\citep{tsipras2018robustness} in~\Cref{fig:fine2coarse}
and~\Cref{fig:fine2coarse-img} respectively.~(see~\cref{sec:expr-settings} for
details on the datasets). For CIFAR-100, we train a ResNet50~\citep{HZRS:2016}
on the fine labels of CIFAR100 and then aggregate the fine labels corresponding
to a coarse label by summing up the logits of the fine classes corresponding to
each coarse class. For restricted imagenet, we use the fine-coarse division
mentioned in~\Cref{tab:fine-grained-classs}. We call this model the
\emph{Fine2Coarse} model and compare the adversarial risk of this network to a
ResNet-50 trained directly on the coarse labels. Note that the model is
end-to-end differentiable as the only addition is a layer to aggregate the
logits corresponding to the fine classes of each coarse class. Thus
PGD adversarial attacks can be applied off the
shelf.~\Cref{fig:fine2coarse,fig:fine2coarse-img} show that for all
perturbation budgets, \emph{Fine2Coarse} has a smaller adversarial risk than the
naturally trained model.

\subsection{Discussion and other relevant works} A related result by~\citet{montasser19a} shows that certain
hypothesis classes are only \emph{improperly} robustly PAC learnable despite
having a finite VC dimension. Thus, learning the problem with small adversarial
error requires using  a different class of models~(or representations) whereas,
for small natural test risk, the original model class~(or representation) can be
used~(it is properly learnable due to its finite VC dimension). In particular,
the examples from~\citet{montasser19a} that uses improper learning to learn a
robust classifier has  a much higher sample complexity. In our example, learning
algorithms for both the hypotheses classes that we use have polynomial sample
complexity. Another point of distinction is that~\Cref{thm:repre-par-inter} uses
a training set induced with random classification noise and hypothesis class
$\cH$ obtains zero training on this noisy training set whereas the examples
in~\citet{montasser19a} do not have any label noise.

\citet{Hanin2019} have shown that though the number of possible linear regions
that can be created by a deep ReLU network is exponential in depth, in practice
for networks trained with SGD this tends to grow only linearly thus creating
much simpler decision boundaries than is possible  due to sheer expressivity of
deep networks. Experiments on the data models from our theoretical settings show
that adversarial training indeed produces more ``complex'' decision boundaries

\citet{Jacobsen2019} have discussed that excessive
invariance in neural networks might increase adversarial
error. However, they argue that excessive invariance
can allow sufficient changes in the semantically important
features without changing the network's prediction. They
describe this as Invariance based adversarial examples as
opposed to perturbation based adversarial examples. We show
that excessive ~(incorrect) invariance might also result in
perturbation based adversarial examples.

Another contemporary work~\citep{Geirhos2020} discusses a phenomenon they refer
to as~\emph{Shortcut Learning} where deep learning models perform very well on
standard tasks like reducing classification error but fail to perform in more
difficult real-world situations. We discuss this in the context of models that
have small test error but large adversarial error and provide theoretical and
empirical to discuss why one of the reasons for this is sub-optimal
representation learning.

\chapter{Improving Adversarial Robustness  via low-rank representations}
\label{chap:low_rank_main}

Dimensionality reduction methods are some of the oldest techniques in
machine learning that extract a small number of factors from a dataset
that explains almost all of its variance; these factors contain most of
the {\em discriminative} power useful in classification or regression
tasks and are known to increase {\em robustness}, i.e. these methods
typically have a denoising effect. Perhaps, the most popular and
widely used among them are PCA (see e.g.~\citep{PCA2002a}) and
CCA~\citep{hotelling1935canonical}.

An intriguing aspect of deep neural networks has been their ability to learn
representations directly from the raw data that are useful in several tasks,
including ones for which they were not specifically trained, usually known
as ~\textit{representation
learning}~\citep{zeiler2014,Sermanet2014,donahue2013,KSH:2012,graves2013speech,
he2015delving,vaswani2017attention,ren2015faster}.
Essentially, for most models trained in a supervised fashion, the vector of
activations in the penultimate layer is a \emph{learned} representation of the
raw input. The remarkable success of Deep Neural Networks~(DNNs) is primarily
attributed to the discriminative quality of this representation space. However,
despite their impressive performance, DNNs are known to be brittle to input
perturbations as we discussed in the previous
chapter~\citep{szegedy2013intriguing,goodfellow2014explaining,Dalvi2004,
Biggio2018,Carlini2017,Papernot2016,mosaavi2016}. In this chapter, we study the
importance of proper inductive biases in the representation space for
adversarial robustness. In particular, we look at whether reducing the intrinsic
dimensionality of the representation space helps with adversarial robustness.

\section{Dimensionality of representations and adversarial robustness}
The vulnerability of deep neural networks towards adversarial perturbations
raises concerns regarding the robustness of the factors captured by these
learned representations. On the other hand, as mentioned earlier, dimensionality
reduction techniques capture factors that are, while being discriminative,
robust to input perturbations. This motivates the thesis behind this
chapter---{\em if we enforce DNNs to learn representations that lie in a
low-dimensional subspace (for the entire dataset), we might be able to obtain
more robust classifiers while preserving their discriminative power.}

\paragraph{Principal and un-principal components of representation space}
\label{sec:princ-unprin-comp}
To get further insights into why restricting the dimensionality of
representation space is helpful for adversarial robustness, we perform a simple
experiment that indicates that adversarial attacks exploit  ``un-principal"
components, i.e. components corresponding to the smallest eigenvalues of the
covariance matrix of the data. We first train a six-layer neural network with
four convolutional and two fully connected layers on the MNIST dataset to
convergence and attack it with a PGD adversary~(see~\Cref{sec:robustness}) to
obtain adversarial examples. Then, we project the adversarial examples on the
{\em top} $k$ and the {\em bottom} $k$ PCA components of the full MNIST dataset,
for all $1\le k\le 784$, to obtain $k$ sets each of projections of the
adversarial examples on the top and bottom $k$ components, respectively. We will
refer to these \(2k\) sets of projections as the {\em principal} and {\em
un-principal} components of the adversarial examples, respectively. Then for
varying $k$, we train $k$ separate linear classifiers to predict the adversarial
labels using just the $k$ principal components of the adversarial examples and
measure its training accuracy. Similarly, we train $k$ separate classifiers to
predict the adversarial label\footnote{An adversarial label is the incorrect
label the model predicts for the adversarial example.} using just the $k$
un-principal components. We plot these training accuracies in blue
in~\Cref{fig:mnist_adv_all_principal} (starting from the principal $k$
components in the left figure, and un-principal $k$ components in the right
figure). Training accuracy for the $k$ components is a measure of how
\emph{easy} it is to fit the data with just those $k$ components and is thus, a
measure of the amount of discriminatory information contained in them. A similar
procedure is done for a randomly sub-sampled set~(of the same size as the number
of adversarial examples) of~(clean) training images and their training accuracy
on the clean labels is plotted in orange.

~\Cref{fig:mnist_adv_all_principal} shows that for the $k$ principal components,
the training accuracy increases much faster on the clean images for a relatively
small number of principal components. For the \(k\) un-principal components,
the training accuracy increases faster for the adversarial examples. For
\emph{clean examples}, almost all the \emph{discriminatory information} lies in
the principal components. For adversarial examples, a significant amount of the
discriminatory information for predicting the adversarial labels are contained
in the un-principal components. This suggests that we should find
low-dimensional representations that retain the discriminatory information
needed for high test accuracy while removing ``noisy'' components that could be
exploited by an adversary. While for simple data, PCA can be used to do this,
when using deep neural networks, a method to induce low-rankness in some hidden
layers (learned representations) is needed, which is what this chapter provides.

\begin{figure}\centering
  \begin{subfigure}[t]{0.8\linewidth}
    \begin{subfigure}[t]{0.49\linewidth}
     \centering \def\svgwidth{0.99\linewidth}
      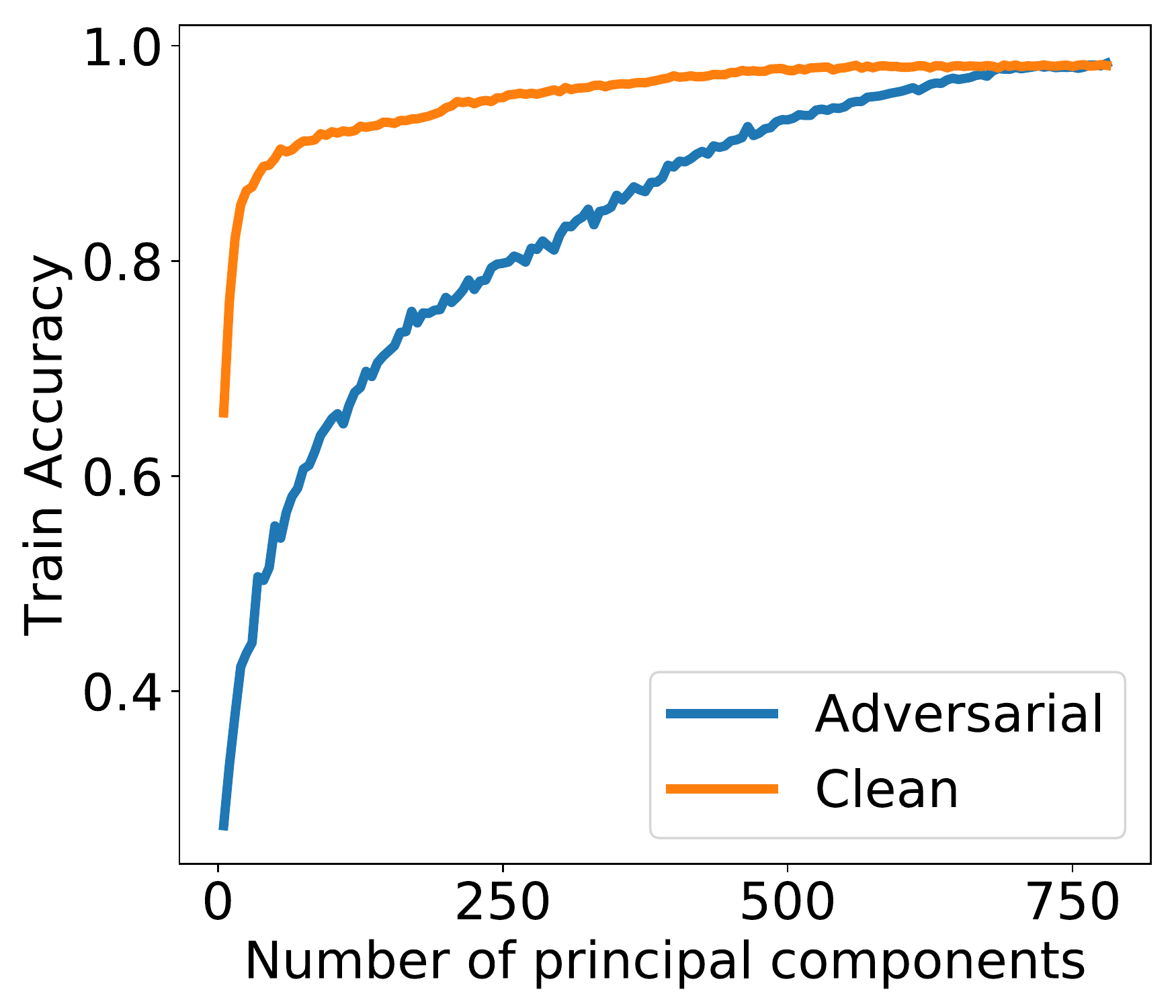
      \caption*{Top k~(Larger Variance)}
      \label{fig:mnist_adv_un_principal}
    \end{subfigure}
    \begin{subfigure}[t]{0.49\linewidth}
      \centering \def\svgwidth{0.99\linewidth}
      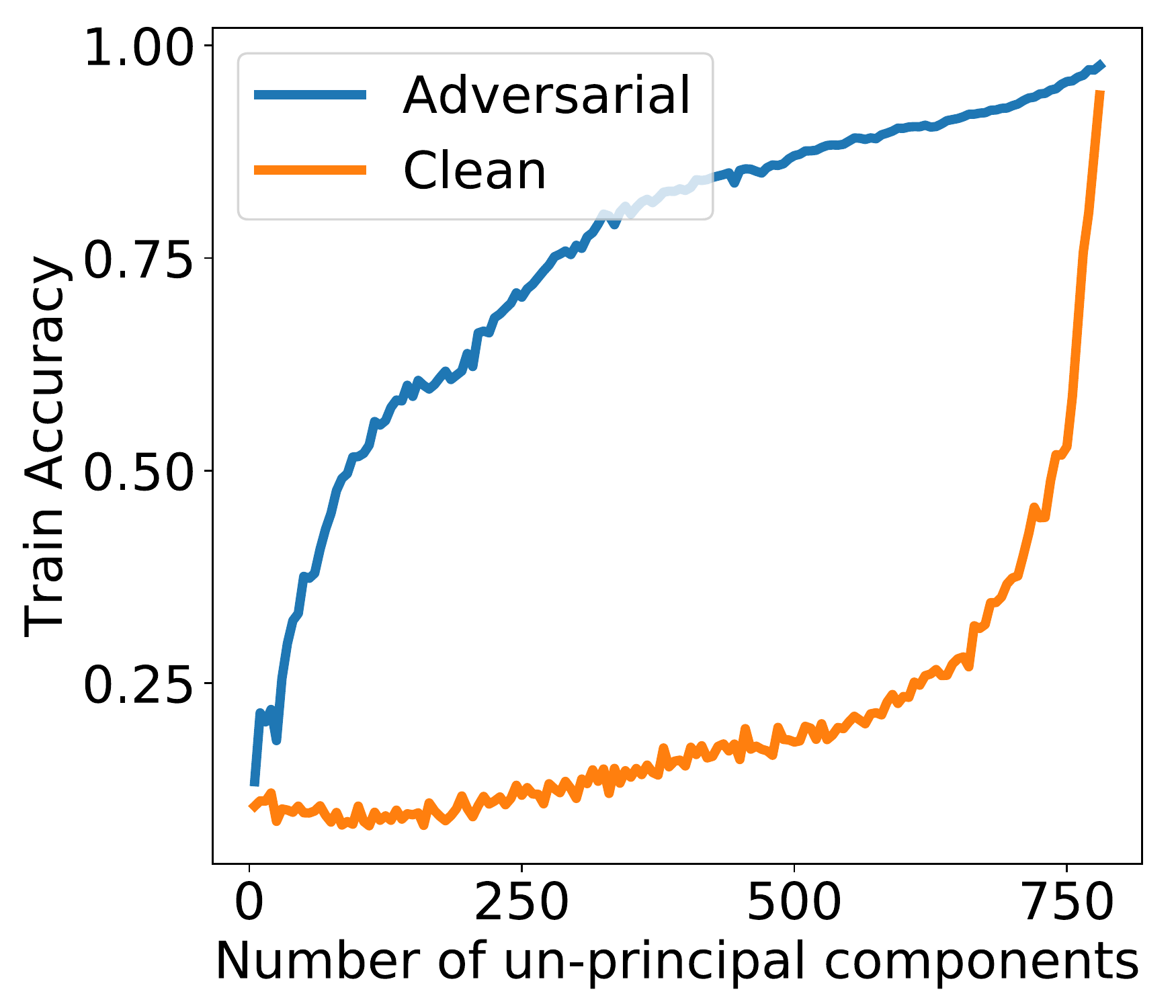
      \caption*{Bottom k~(Lesser variance)}
      \label{fig:mnist_adv_un_principal}
    \end{subfigure}
  \end{subfigure}
    \caption[Adversarial Noise exploits directions of lesser
      variance]{ Adversarial Noise manipulates the principal
      components with lesser
      variance.}
    \label{fig:mnist_adv_all_principal}
\end{figure}
\paragraph{Challenges to obtain low-dimensional
representations}\label{sec:challenges-low-dim} Ideally, to encourage learning
low-dimensional representations, we would like to insert a \emph{dimensionality
reduction} ``module'' in deep neural networks and develop an end-to-end training
method that simultaneously does supervised training and dimensionality
reduction. At first glance, using SVD to project representations onto a
low-dimensional subspace seems viable. However, this approach encounters
multiple challenges.
\begin{enumerate}
  \item The SVD algorithm needs to operate on a matrix one of whose dimensions
  is as large as the number of examples in the dataset. As the number of
  examples is usually very large, executing the SVD algorithm can be
  computationally prohibitive in practice.
  \item The representations themselves are not learnable parameters but just
parametric functions of the input data; as a consequence, they change after every
parameter update. It is not straightforward how to preserve the low rank of the representations, after doing the SVD, from one iteration to the next. 
\end{enumerate}

A workaround could be to design architectures with bottlenecks similar to
auto-encoders~\citep{Hinton2006}. The fact that most of the state-of-the-art
networks do not have such bottlenecks limit their usability. Further, as we discuss in~\Cref{sec:alt_algs}, it also doesn't improve adversarial robustness in practice. We discuss this and other alternate algorithms in~\Cref{sec:alt_algs}.

\section{Main contributions}
This algorithm proposed in this chapter provides the benefits of dimensionality reduction by inserting a
\emph{virtual layer} (not used at prediction time) and augmenting the loss
function to induce low-rank representations. 

\paragraph{Algorithmic Contributions}
We propose a novel low-rank regulariser (LR) to control the intrinsic
dimensionality of the representations that
\begin{enumerate}
  \item does not put any restriction on the network
architecture,%
\footnote{It puts no \emph{direct} restriction, though of course, any
extra regularisation will produce an inductive bias.}
\item is end-to-end trainable, and 
\item  is efficient in that it allows
mini-batch training. 
\end{enumerate}
LR explicitly {\em enforces} representations to lie in a linear subspace with
low {\em intrinsic} dimension and is guaranteed to provide low-rank
representations for the entire dataset even when trained using mini-batches. As
LR is a virtual layer~(discussed in later sections), it can be applied to any
intermediate representations of DNNs.

\paragraph{Experimental Contributions}
Experimentally, apart from successfully reducing the dimensionality of learned
representations, neural networks trained with LR turn out to be significantly
more robust to input perturbations, both adversarial and random, while providing
modest improvements over the natural~(unperturbed) test accuracy. This is of
particular interest as it provides empirical evidence that adding well-thought
priors over factors influencing the representation space (e.g. low-rank prior
over representations) might further improve the robustness of DNNs, without
encountering a trade-off\todo[color=green]{Discuss the different tradeoffs in
Chap 3} with, be it
computational~\citep{goodfellow2014explaining,madry2018towards},
statistical~\citep{schmidt2018adversarially} or a loss in
accuracy~\citep{tsipras2018robustness}. See~\Cref{sec:robustness-tradeoffs} for
a discussion on the various tradeoffs associated with this.  This is in line
with recent works, including~\Cref{chap:causes_vul} of this thesis, suggesting
that a ``correct'' representation may avoid the perceived trade-off between
robustness and accuracy~(see~\citep{madry2018towards,Zhang2019,montasser19a}). 

Lastly, because of the low dimensionality, we can compress
representations by a significant factor without losing its discriminative
power. Thus, discriminative features of the data can be stored using very
little memory. For example, we show in one of our experiments that, even with a
$5$-dimensional embedding~(400x compression), a model with LR loses only $6\%$
in accuracy.

\section{Deep Low-Rank Representations Layer~(LR Layer)}
\label{sec:low_rank}
\label{sec:what-low-rank}
Consider $f: \mathbb{R}^d \mapsto \mathbb{R}^k$ to be a feed-forward MLP that
maps $d$ dimensional input $\vx$ to a $k$ dimensional output $\vy$. We can
decompose this into two sub-networks, one consisting of the layers before the
$\ell^{\it th}$ layer and one after i.e.  $f(\vx) =
f^{+}_\ell\br{f^{-}_\ell(\vx; \phi) ; \theta}$, where $f^{-}_\ell (.;\phi)$,
parameterised by $\phi$, represents the part of the network up to layer $\ell$
and, $f^{+}_\ell(.;\theta)$ represents the part of the network thereafter. With
this notation, the $m$ dimensional representation (or the activations) after
any layer $\ell$ can simply be written as $\va = f^{-}_\ell(\vx; \phi) \in
\mathbb{R}^m$. In what follows, we first formalise the low-rank representation
problem and then propose our approach to solve it approximately and efficiently.

\subsection{Formulating the LR Layer problem} Let $\vec{X} = \{\vec{x}_i\}_{i=1}^N$
and $\vec{Y} = \{{\bf y}_i\}_{i=1}^N$ be the set of inputs and target outputs of
a given training dataset. By slight abuse of notation, we define $\vec{A}_\ell =
f^{-}_\ell(\vec{X}; \phi) =\bs{\vec{a}_1,\cdots,\vec{a}_N}^\top\in \mathbb{R}^{N
\times m}$ to be the activation matrix of the entire dataset, so that
$\vec{a}_i$ is the activation vector of the $i$-th sample. Note that for most
practical purposes $N\gg m$. In this setting, the problem of learning low-rank
representations can be formulated as a constrained optimisation problem as
follows:
\begin{align}
  \label{eq:opt_prob}
	\min_{\theta, \phi}\mathcal{L}(\vec{X}, \vec{Y}; \theta, 
  \phi),~\text{s.t.}~~&\rank{\vec{A}_\ell} = r,\end{align} 
where $\mathcal{L}(.)$ is the loss function and $r < m$ is the desired rank of
the representations at layer $\ell$. The rank $r$ is a hyperparameter (though
empirically not a sensitive one as observed in our experiments). Throughout this
section, we discuss the problem of imposing low-rank constraints over a single
intermediate layer, however, it can trivially be extended to any number of
layers. Note that both the loss and the constraint set of the above objective
function are non-convex. One approach to optimising this would be to perform an
alternate minimisation style algorithm~(eg.~\citep{Lloyd1982,Dempster1977}),
first over the loss (gradient descent) and then projecting onto the non-convex
set to satisfy the rank constraint.

Since $N \gg m$, ensuring $\rank{\vec{A}_\ell} = r$ would  be practically
infeasible as it would require performing SVD in every iteration at a cost
$\mathcal{O}(N^2m)$. A feasible, but incorrect, approach would be to do this on
mini-batches, instead of the entire dataset. Projecting each mini-batch onto the
space of rank $r$ matrices does not guarantee that the activation matrix of the
entire dataset will be of rank $r$, as each of these mini-batches can lie in
very different subspaces. As a simple example, consider the set of coordinate
vectors which can be looked at as rank one matrices corresponding to activations
in batches of size one. However, when these coordinate vectors are stacked
together to form a matrix they create the identity matrix, which is a full rank
matrix.

Computational issues aside, another crucial problem stems
from the fact that the activation matrix $\vec{A}_\ell = f^{-}_\ell(.;
\phi)$  is itself parameterised by $\phi$ and thus $\phi$ needs to be
updated in a way such that the generated $\vec{A}_\ell$ is low
rank. It is not immediately clear how to use the low-rank projection
of $\vec{A}_\ell$ to achieve this.
One might suggest to first fully train the network and then obtain low-rank
projections of the activations. Our experiments show that this procedure does
not provide the two main benefits we are looking for: preserving accuracy under
compression and robustness.

\paragraph{Low-Rank regulariser:} We now describe our regulariser that
encourages learning low-rank activations, and, if optimised properly, guarantees that the rank of the activation matrix (of any
size) will be  bounded by $r$. %
The primary ingredient of our approach is the introduction of an auxiliary
parameter~$\vec{W}\in\reals^{m\times m}$ in a way that allows us to 
shift the low-rank constraint from the activation
matrix $\vec{A}_\ell$~(as in~\Cref{eq:opt_prob}) to 
$\vec{W}$ providing the following two advantages: The rank constraint is now 

\begin{enumerate}
  \item  on a matrix that is independent
of the batch/dataset size, and
\item  on a parameter as opposed to a
data-dependent intermediate tensor~(like activations), and can thus be updated
directly at each iteration. 
\end{enumerate}
Combining these ideas, our final augmented objective function, with the
regulariser, is:
\begin{align}
  \label{eq:aug_opt_2}
  &\min_{\theta, \phi, \vec{W}, \vb} \mathcal{L}(\vec{X}, \vec{Y}; \theta, \phi) + \lambda_1\mathcal{L}_c(\vec{A}_\ell; \vec{W},\vb) + \lambda_2\mathcal{L}_n(\vec{A}_\ell) \\
  &\text{s.t.,} \vec{W}\in \mathbb{R}^{m\times m}, \rank{\vec{W}} = r,~ \vb\in \mathbb{R}^m,\vec{A}_\ell=f^{-}_\ell(\vec{X}; \phi),\nonumber
\end{align}
where,
\begin{align}
  \label{eq:aug_opt_3}
  \mathcal{L}_c(\vec{A}_\ell; \vec{W}, \vb) = \frac{1}{n}\sum_{i=1}^{n}\norm{\vec{W}^\top(\va_i+\vb) - (\va_i+\vb)}_2^2, 
   &\text{and} \; \; \mathcal{L}_n(\vec{A}_\ell) = \frac{1}{n}\sum_{i=1}^n \Big|1 - \norm{\va_i}\Big|.\nonumber
\end{align}

\noindent Here, $\mathcal{L}_c$ is the projection loss that
ensures that the affine low-rank mappings~($\vec{A}\vec{W}$) of the activations
are close to the original ones i.e. $\vec{AW} \approx \vec{A}$. As the constraint \((\rank{\vec{W}} = r)\) ensures that $\vec{W}$ is
low-rank, $\vec{A}\vec{W}$ is also low-rank and thus implicitly~(due to
$\vec{A}\vec{W}\approx\vec{A}$), $\mathcal{L}_c$ forces the original activations $\vec{A}$ to
be low-rank. The bias $\vb$ allows for the activations to be translated before
projection.%
\footnote{We use the term \emph{projection} loosely as we do not strictly
constrain $\vec{W}$ to be a projection matrix.}

However, note that setting $\vec{A}$ and $\vb$ close to zero trivially minimises
$\mathcal{L}_c$, especially when the activation dimension is large. Also, due to
the positive homogeneity of each layer, the magnitude of the activations can be
minimised in a layer by multiplying the weights of that layer by a small
constant $c$ and then maximised in the next layer by multiplying the weights of
that layer with $\frac{1}{c}$ thereby preserving the final logit magnitudes. We
observed this to happen in practice as it is easier for the network to learn
$\phi$ such that the activations and the bias are very small to
minimise $\mathcal{L}_c$ as compared to learning a low-rank representation space. To prevent this, we use $\mathcal{L}_n$ that acts as a
norm constraint on the activation vector to keep the activations sufficiently
large. Lastly, as the rank constraint is now over $\vec{W}$ and $\vec{W}$ is a
{\em global} parameter independent of the dimension~$n$~(i.e. size of
mini-batch/dataset), we can use mini-batches to optimise~\Cref{eq:aug_opt_2}.
Since $\rank{\vec{AW}} \leq r$ for any $\vec{A}$, optimizing over mini-batches
still ensures that the entire activation matrix is low-rank. Intuitively, this
is because the basis vectors of the low-rank affine subspace are
now captured by the low-rank parameter $\vec{W}$. Thus, as long as $\cL_c$ is
minimised for all mini-batches, $\vec{A} \approx \vec{A}\vec{W}$ holds for
the entire dataset, leading to the low-dimensional support. Thus, the overall
goal of the augmented objective function in~\Cref{eq:aug_opt_2}, is to jointly
penalise the activations~($\vec{A}_\ell$) to make them low-rank and learn the
corresponding low-rank identity map~$\br{\vec{W},b}$. Note, implementation wise,
our regulariser requires adding a  virtual (does not modify the main network)
branch with parameters $\br{\vec{W},b}$ at layer $\ell$. This branch is removed
at the time of inference as the activations learned, by virtue of our objective
function, are already low-rank.

\begin{remark}
  The reason we need to minimise both the reconstruction loss $\cL_c$ and the
  norm constraint loss $\cL_n$ simultaneously is the same as why the spectral
  norm and the stable rank can be independently minimised in
  ~\Cref{chap:stable_rank_main} without affecting each other. To see why, note
  that the rank and the stable rank of the matrix is independent of the scale of
  the matrix i.e. the matrix can be multiplied by any non-zero scalar without
  affecting the rank or the stable rank of the matrix whereas multiplying the
  matrix by a scalar affects the spectral norm proportionally.
\end{remark}

\paragraph{Hyper-parameters}
While our algorithm has three hyper-parameters $\lambda_1,\lambda_2$ and $r$, in
  practice, our approach is insensitive to $\lambda_1,\lambda_2$ and thus, we
  set $\lambda_1=\lambda_2=1$. This is an added advantage given the resources
  needed to do hyper-parameter optimisation. In the classical view of
  regularisation e.g. ridge regression, the regularisation coefficient induces a
  bias-variance trade-off i.e. as $\norm{W}_{\mathrm{F}}\rightarrow 0$, the
  accuracy decreases. In our case, under the assumption that there exist
  low-rank representations achieving zero classification error, even as the
  terms $\cL_n$ and $\cL_c$ go to 0, the original classification loss $\cL$ does
  not (necessarily) increase. One way of thinking about this is that the terms
  $\cL_n$ and $\cL_c$ are guiding the optimisation to specific minimisers of the
  empirical classification loss, of which there will necessarily be several
  because of overparameterisation. To verify this empirically, we ran
  experiments with all combinations of $\lambda_1,\lambda_2$ chosen from
  $\bc{1., 0.1, 0.01}$. We observe that for all these settings, the terms
  $\cL_n$ and $\cL_c$ go to 0 while the original loss does not degrade at all.
\begin{algorithm}[t]  \centering
  \caption{Low-Rank (LR) regulariser}
  \label{alg:lr_layer_main}
  \begin{algorithmic}
  \INPUT Activation Matrix $\vec{A}_l$, gradient input ${\bf g}_l$
  \STATE ${\bf Z} \gets (\vec{A}_l+\vb)^\top \vec{W}$
    \COMMENT {forward propagation towards
      the virtual LR layer}
    \STATE $\mathcal{L}_c \gets \frac{1}{b} \norm{{\bf Z} - (\vec{A}_l +
        \vb)}_2^2$ \COMMENT{the reconstruction loss} 
    \STATE $\mathcal{L}_n \gets
      \frac{1}{b}\sum_{i=1}^{b}\big\vert\mathbf{1} - \norm{\va_i}
      \big\vert$ \COMMENT{norm constraint loss}
    \STATE ${\bf g}_W \gets \frac{\partial \mathcal{L}_c}{\partial
        \vec{W}},\enskip {\bf g}\gets {\bf g}_l +
      \frac{1}{b}\sum_{i=1}^{b}\frac{\partial (\mathcal{L}_c +
        \mathcal{L}_n)}{\partial \va_i}$
    \STATE $\vec{W} \gets \vec{W} - \lambda {\bf
        g}_W$ %
    \STATE $\vec{W} \gets
      \prnk{k}{\vec{W}}$ \label{alg:hard_thresh_step} \COMMENT{hard
      thresholds the rank of $\vec{W}$} 
    \OUTPUT ${\bf g}$
    \COMMENT{the gradient to be passed to the layer before}
  \end{algorithmic}
\end{algorithm}

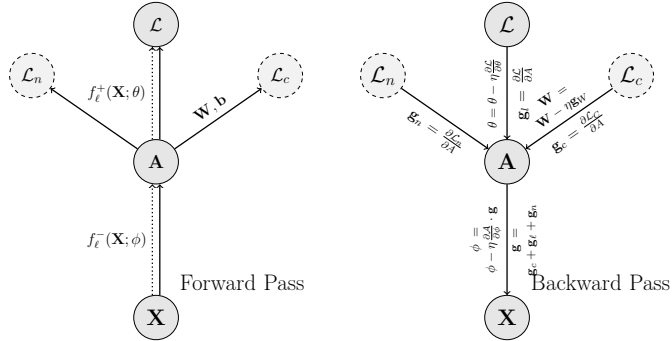
\begin{figure}[t]\centering
  \scalebox{0.49}{\begin{tikzpicture} \node [tensors] (data) {
        \Large$\mathbf{X}$}; \node [tensors, above=3cm of data]
      (activations) { \large$\mathbf{A}$}; \node [rectangle,
      rounded corners, above right=0.8cm of data, yshift=-0.4cm,
      xshift=-0.5cm] (fwd_lbl) { \Large Forward Pass}; \node
      [tensors, above=2.5cm of activations] (output) {
        \large$\mathcal{L}$}; \node [temp_tensors, above right
      =3.5cm of activations, yshift=-1cm] (recons) {
        \large$\mathcal{L}_c$}; \node [temp_tensors, above left
      =3.5cm of activations, yshift=-1cm] (normed) {
        \large$\mathcal{L}_n$};
      % % Draw the links between forces
      \draw[->, thick, color=black, line width=1pt] ([xshift=1 *
      0.1 cm]data.north)-- node[midway, parameters]
      {\normalsize$f_{\ell}^{-}(\mathbf{X};\phi)$} ([xshift=1 *
      0.1 cm]activations.south); \draw[->, dotted, color=black,
      line width=1pt] ([xshift=-1 * 0.1 cm]data.north)--
      ([xshift=-1 * 0.1 cm]activations.south);

      \draw[->, thick, color=black, line width=1pt] ([xshift=1 *
      0.1 cm]activations.north) -- node[parameters, align=center,
      midway] {\normalsize$f_{\ell}^{+}(\mathbf{X};\theta)$}
      ([xshift=1 * 0.1 cm]output.south); \draw[->, dotted,
      color=black, line width=1pt] ([xshift=-1 * 0.1
      cm]activations.north) -- ([xshift=-1 * 0.1 cm]output.south);

      \draw[->, thick, color=black, line width=1pt] (activations)
      -- node[parameters, midway,above, sloped]
      {\normalsize$\mathbf{W},\mathbf{b}$} (recons); \draw[->,
      thick, color=black, line width=1pt] (activations) --
      (normed);
    \end{tikzpicture}}\hspace{15pt}
  \scalebox{0.49}{\begin{tikzpicture} \node [tensors] (data) {
        \Large$\mathbf{X}$}; \node [tensors, above=3cm of data]
      (activations) { \Large$\mathbf{A}$}; \node [tensors,
      above=2.5cm of activations] (output) { \Large$\mathcal{L}$};
      \node [temp_tensors, above right=3.5cm of activations,
      yshift=-1cm] (recons) { \Large$\mathcal{L}_c$}; \node
      [temp_tensors, above left=3.5cm of activations, yshift=-1cm]
      (normed) { \Large$\mathcal{L}_n$}; \node [ rounded corners,
      dashed, above right=0.8cm of data, xshift=-0.5cm,
      yshift=-0.4cm] (fwd_lbl) {\Large Backward
        Pass};% % Draw the links between forces
      \draw[->, thick, color=black, line width=1pt] ([yshift=0 *
      0.1 cm]activations.south) -- node[midway, parameters, below,
      sloped, rotate=180]
      {$\mathbf{g} =\mathbf{g}_c + \mathbf{g}_\ell +\mathbf{g}
        _n$} node[parameters, above, sloped,
      rotate=180]{$\phi=\phi-\eta\frac{\partial
          \mathcal{A}}{\partial \phi}\cdot \mathbf{g}$} ([yshift=0
      * 0.1 cm]data.north);

      \draw[->, thick, color=black, line width=1pt] (output) --
      node[parameters, below, sloped, rotate=180]
      {\normalsize$\mathbf{g}_l=\frac{\partial
          \mathcal{L}}{\partial A}$} node[parameters, above,
      sloped, rotate=180, allow upside
      down]{$\theta=\theta-\eta\frac{\partial
          \mathcal{L}}{\partial \theta}$}(activations) ;

      \draw[->, thick, color=black, line width=1pt] (recons) --
      node[parameters, midway,above, sloped,
      align=center]{$\mathbf{W}=\mathbf{W}-\eta\mathbf{g}_W$ }
      node[parameters, midway,below, sloped, align=center,
      sloped]{\normalsize
        $\mathbf{g}_c = \frac{\partial \mathcal{L}_C}{\partial A}$
      } (activations);

      \draw[->, thick, color=black, line width=1pt] (normed) --
      node[parameters,
      midway,sloped,below,align=center]{\normalsize
        $\mathbf{g}_n = \frac{\partial \mathcal{L}_n}{\partial A}$
      } (activations);

      % \draw[->, dotted, color=black, line width=1pt,
      % yshift=-1cm] ([yshift=-1 * 0.1 cm]activations.west) --
      % node[midway, parameters, below]
      % {\large$\mathbf{g_\ell} + \mathbf{g_C}$} ([yshift=-1 * 0.1
      % cm]data.east); \draw[->, dotted, color=black, line
      % width=1pt] (activations) -- node[parameters, below]
      % {\large$f_{\ell}^{+}(\mathbf{X};\theta)$} (output);
      % \draw[->, dotted, color=black, line width=1pt]
      % (activations) -- node[parameters, midway,below, sloped]
      % {\large$\mathbf{W},\mathbf{b}$} (recons); \draw[->,
      % dotted, color=black, line width=1pt] (activations) --
      % (normed);

    \end{tikzpicture} }
  % \caption{FOSS in Chrome influences industry structure by
  % increasing competition}
  \caption[Illustration of forward and backward propagation of the LR
  Layer]{{\small \bf The LR layer}. The left figure shows the {\em forward
  pass}, {\em solid edges} show the flow of data during training, {\em dashed
  edges}- the flow of data during inference, and {\em dashed nodes} are the {\em
  virtual layer}. The right figure shows the {\em backward pass}.}
  \label{fig:LRtikz}
\end{figure}

\subsection{Algorithm for LR Layer}
\label{sec:alg-lr}

We solve our optimisation problem~(\Cref{eq:aug_opt_2}) by adding a
\emph{virtual} low-rank layer that penalises representations that are far from
their closest low-rank affine representation. Algorithm~\ref{alg:lr_layer_main} %(further details in Appendix~\ref{sec:alg-lr})
describes the operation of the low-rank virtual layer for a mini-batch of size
$b$. We present a flow diagram for the same in Figure~\ref{fig:LRtikz}. This
layer is virtual in the sense that it only includes the parameters $\vec{W}$ and
$\vb$ that are not used  in the NN model itself to make predictions, but
nonetheless, the corresponding loss term $\mathcal{L}_c$ does affect the model
parameters through gradient updates.~\Cref{fig:illus-lr-train} provides an
illustration of how the LR-Layer is added during training as a plug-and-play
layer and then removed during testing, and the generated features are already
low rank. Algorithm~\ref{alg:lr_layer_main} alternately minimises the augmented
loss function in~\Cref{eq:aug_opt_2} and projects the auxiliary parameter
$\vec{W}$ to the space of low-rank matrices.

\begin{figure}
  \centering
  \begin{subfigure}[c]{0.65\linewidth}
    \includegraphics[width=0.99\linewidth]{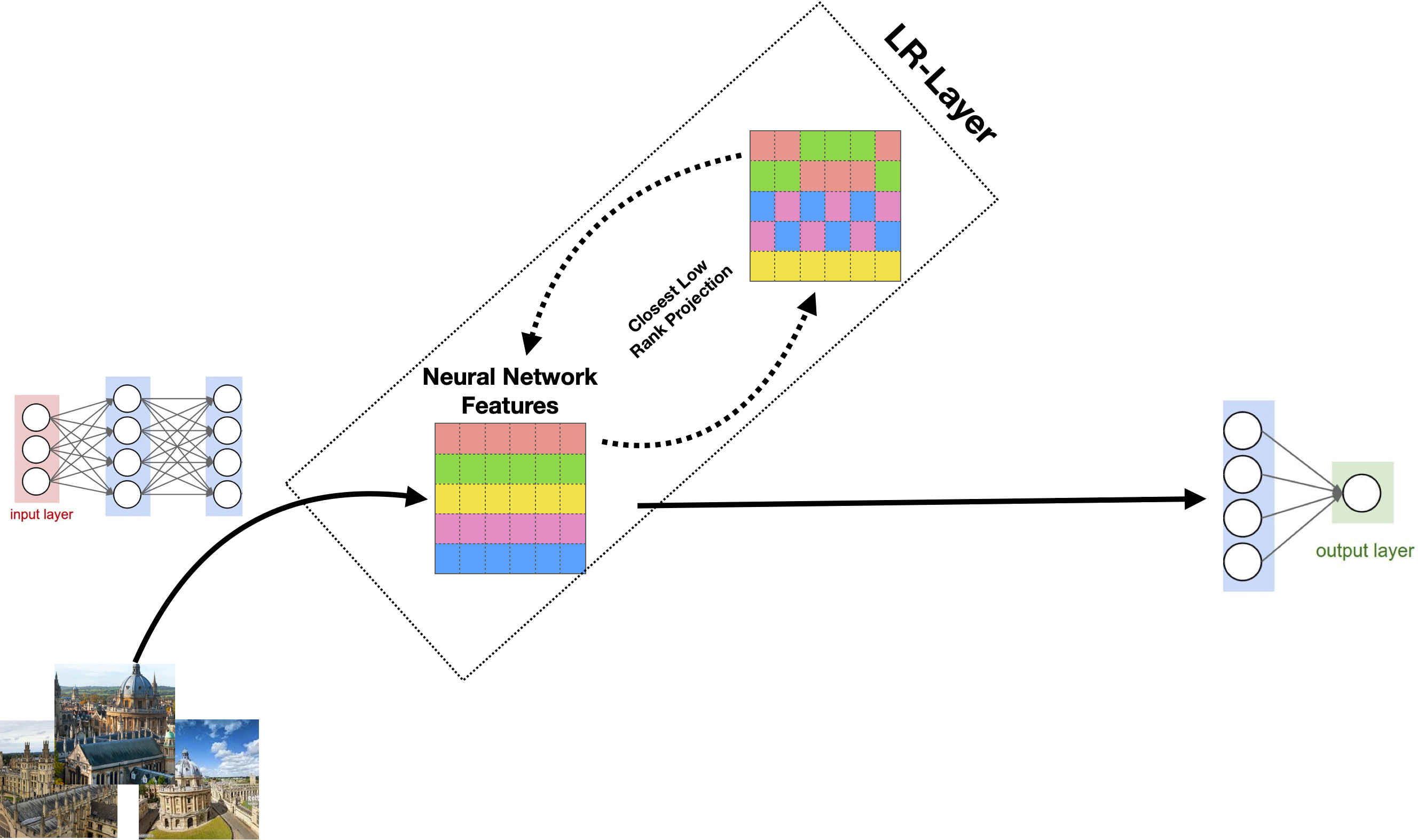}
    \caption{Training}
    \label{fig:illus-train}
  \end{subfigure}\vspace{10pt}
  \begin{subfigure}[c]{0.65\linewidth}
  \includegraphics[width=0.99\linewidth]{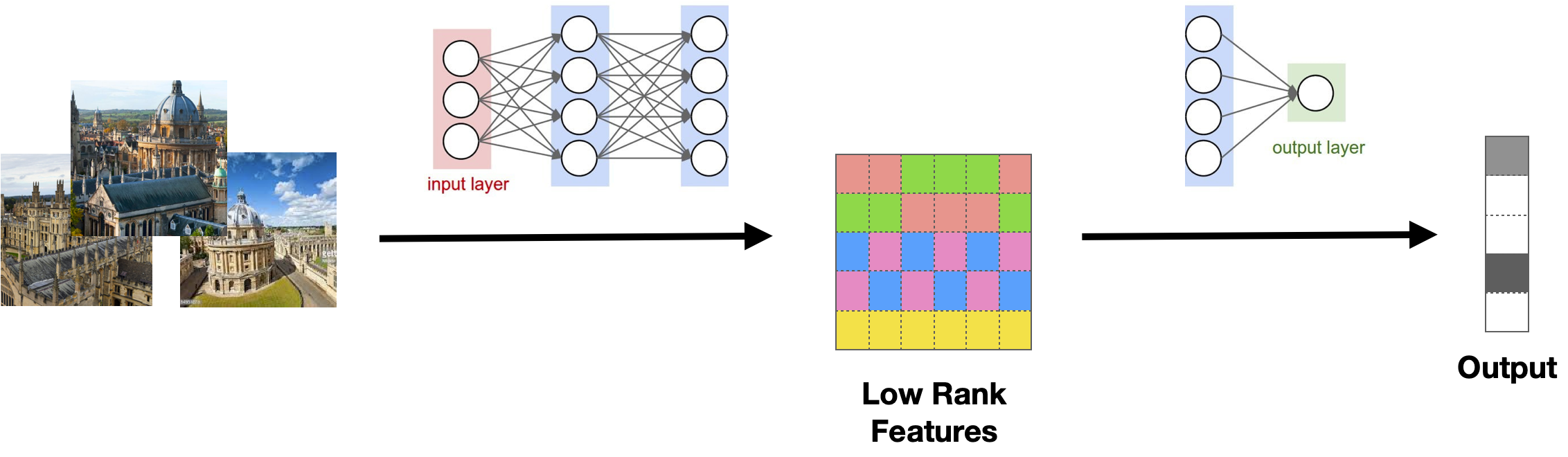}
  \caption{Testing}
  \label{fig:illus-test}
  \end{subfigure}
  \caption{\Cref{fig:illus-train} shows an illustration of how the LR-Layer is added as a plug-and-play layer during training.~\Cref{fig:illus-test} shows an illustration that the LR layer is removed during testing and the generated representations are already low-rank.}
  \label{fig:illus-lr-train}
\end{figure}

The rank projection step in~\Cref{alg:lr_layer_main} is executed by a hard
thresholding operator $\prnk{r}{W}$, which finds the best $r$-rank approximation
of $W$. Essentially, $\prnk{r}{W}$ solves the following optimisation problem,
which can be solved using singular value decomposition (SVD).

\begin{equation}
  \prnk{r}{W} = \argmin_{\rank{Z} = r}\norm{W
                                - Z}_F^2 
 \end{equation}

 However, the projection can be very expensive due to the large dimension of the
representations space~(e.g. $16000$). To get around this, we use the ensembled
Nystr\"om SVD algorithm~\citep{williams2001using, halko2011finding,
kumar2009ensemble}.

\paragraph{Handling large activation
matrices}: %\label{sec:handl-large-activ}
Singular Value Projection~(SVP) introduced in ~\citet{jain2010guaranteed} is an
algorithm for rank minimisation under affine constraints. In each iteration, the
algorithm performs gradient descent on the affine constraints alternated with a
rank-k projection of the parameters and it provides recovery guarantees under
weak isometry conditions. However, the algorithm has a complexity of $O(mnr)$
where $m, n$ are the dimensions of the matrix and $r$ is the desired low rank.
Faster methods for SVD of sparse matrices are not applicable as the matrices in
our case are not necessarily sparse.  We use the ensembled Nystr\"om
method~\citep{williams2001using, halko2011finding, kumar2009ensemble} to boost
our computational speed at the cost of the accuracy of the low-rank
approximation. It is essentially a sampling-based low-rank approximation to a
matrix. The algorithm is described in detail in~\Cref{sec:ensembl-nystr-meth}.
Though the overall complexity for projecting $W$ remains $O(m^2r)$, the
complexity of the hard-to-parallelise SVD step is now $O(r^3)$, while the rest
is due to matrix multiplication, which is fast on modern GPUs.

The theoretical guarantees of the Nystr\"om method hold only when the weight
matrix of the \lr is symmetric and positive semi-definite (SPSD) before each
$\prnk{r}{\cdot}$ operation. Note that the PSD constraint is actually not a
restriction as all projection matrices are PSD by definition.  For example, on
the subspace spanned by columns of a matrix $\vec{X}$, the projection matrix is
$\vec{P} = \vec{X}(\vec{X}^\top \vec{X})^{-1}\vec{X}^\top$, which is always PSD.
We know that a symmetric diagonally dominant real matrix with non-negative
diagonal entries is PSD. With this motivation, the matrix $\vec{W}$ is
smoothened by repeatedly adding $0.01\vec{I}$ until the \textsf{SVD} algorithm
converges where $\vec{I}$ is the identity matrix.\footnote{The computation of
the singular value decomposition sometimes fail to converge if the matrix is
ill-conditioned}. This is a heuristic to make the matrix well-conditioned as
well as diagonally dominant and it helps in the convergence of the algorithm
empirically. 

\paragraph{Symmetric Low-Rank Layer}%\label{sec:symmetric-low-rank}

The symmetricity constraint restricts the projections allowed in our
optimisation, but empirically this restriction does not seem to hurt its
performance. Implementation wise, gradient updates may make the matrix parameter
asymmetric, even if we start with a symmetric matrix. Reparameterising the \lr
fixes this issue; the layer is parameterised using $\vec{W}_s$ (to which
gradient updates are applied), but the layer projects using $\vec{W} =
(\vec{W}_s + \vec{W}_s^\top)/2$, which is by definition a symmetric matrix.
After the rank projection is applied to the (smoothed version of) $\vec{W}$,
$\vec{W}_s$ is set to be $\prnk{r}{\vec{W}}$. By~\Cref{thm:nystrom_sym}, if we
start with an SPSD matrix $\vec{W}_s$, the updated $\vec{W}_s$ is an SPSD
matrix. As a result, the updated $\vec{W}$ is also SPSD.

\begin{restatable}[Column Sampled Nystr\"om approximation preserves SPSD matrices]{lem}{spsdnystrom}
  \label{thm:nystrom_sym} If $\vec{X}\in\reals^{m\times m}$ is an SPSD matrix
and $\vec{X}_r\in\reals^{m\times m}$ is the best Nystr\"om ensembled, column
sampled r-rank approximation of $\vec{X}$, then $\vec{X}_r$ is SPSD as
well.

Proof in~\Cref{sec:lr-spsd-proof}
\end{restatable}

\subsection{Ensembled Nystr\"om method}
\label{sec:ensembl-nystr-meth}

 Let $\vec{W}\in \reals^{m\times m}$ be a symmetric positive semidefinite matrix
(SPSD). We want to generate a matrix $\vec{W}_r$ which is an r-rank approximation
of $\vec{W}$ without performing SVD on the full matrix $\vec{W}$ but only on a
principal submatrix of $\vec{W}$. A principal submatrix of a matrix $\vec{W}$ is
a square matrix formed by removing some columns and the corresponding rows from
$\vec{W}$~\citep{Meyer2000MAA}. Let the principal submatrix be $\vec{Z}\in
\reals^{l\times l}$, where $l\ll m$. We construct $\vec{Z}$ by first sampling
$l$ indices from the set $\{1\ldots m\}$ and selecting the corresponding columns
from $\vec{W}$ to form a matrix $\vec{C}\in \reals^{m\times l}$. Then, we select
the $l$ rows with the same indices from $\vec{C}$ to get $\vec{Z}\in
\mathbb{R}^{l \times l}$. We can rearrange the columns of $\vec{W}$ and
$\vec{C}$ so that \[W =
\begin{bmatrix} \vec{Z} \quad \vec{W}_{21}^T \\
  \vec{W}_{21}\quad \vec{W}_{22} \end{bmatrix} \quad \quad \vec{C} =
  \begin{bmatrix} \vec{Z} \\ \vec{W}_{21}
  \end{bmatrix}
\] According to the Nystr\"om approximation, the low-rank
approximation of $\vec{W}$ can be written as
\begin{equation}
  \label{eq:nystrom_approx} \vec{W}_r = \vec{C} \vec{Z}_r^{+}\vec{C}^T
\end{equation} where $\vec{Z}_r^{+}$ is the pseudo-inverse of the best $r$
rank approximation of $\vec{Z}$. Hence, the entire algorithm is as follows.
\begin{itemize}
\item Compute $\vec{C}$ and $\vec{Z}$ as stated above.
\item Compute the top $r$ singular vectors and values of $\vec{Z}$ : $\vec{U}_r,~\mathbf{\Sigma}_r,~\vec{V}_r$.
\item Invert each element of $\mathbf{\Sigma}_r$ to get the Moore pseudo-inverse of
$\vec{Z}_r$.
\item Compute $\vec{Z}_r^{+} = \vec{U}_r\mathbf{\Sigma}_r^{-1} \vec{V}_r$ and $\vec{W}_r = \vec{C}
\vec{Z}_r^{+}\vec{C}^T$.
\end{itemize} Though by trivial computation, the complexity of the
algorithm seems to be $O(l^2r + ml^2 + m^2l) = O(m^2r)$~(in our
experiments $l = 2r$), however the complexity of the SVD
step is only $O(r^3)$ which is much lesser than $O(m^2r)$ and while
matrix multiplication is easily parallelisable, SVD is not.

To improve the accuracy of the approximation, we use the ensembled Nystr\"om
sampling-based methods~\citep{kumar2009ensemble} by averaging the outputs of $t$
runs of the Nystr\"om method. The $l$ indices for selecting columns and rows are
sampled from a uniform distribution and it has been
shown~\citep{kumar2009sampling} that uniform sampling performs better than most
other sampling methods. Theorem 3 in ~\citet{kumar2009ensemble} provides a
probabilistic bound on the Frobenius norm of the difference between the exact
best r-rank approximation of $\vec{W}$ and the Nystr\"om sampled r-rank
approximation.

\section{Experiments}
\label{sec:lr_experiments}
\todo[color=green]{Remove section on sparsity and Rank, Structure in Linear Transformation, Bottleneck layers}
We perform a wide range of experiments to show the effectiveness of
imposing low-rank constraints on the representations of a dataset
using our proposed LR. %

\paragraph{Architectures and datasets}
We use the standard ResNet~\citep{HZRS:2016} architecture with four residual
blocks. To capture the effect of network depth, we use ResNet-50~(R50) and
ResNet-18~(R18). Please refer to~\Cref{sec:expr-settings} for more details on
datasets and neural network architectures used for the experiments. Since LR can
be applied to any representation layer in the network, we investigate the
following configurations:
\begin{itemize}[leftmargin=0.5cm, itemsep=0em]
\item {\bf 1-LR}, where the LR layer is located just before the last fully-connected (FC) layer that contains 512 and 2048 units in ResNet-18 and ResNet-50, respectively.
\item {\bf 2-LR}, where there are two LR layers, the first one positioned before the fourth ResNet block with $16,384$ incoming units, and the second one just before the FC layer as in ResNet 1-LR.
\item {\bf N-LR}, without any LR layer. This is the standard model without any modification.
\todo[color=blue]{Remove this VGG}
\end{itemize}

\begin{figure}\centering
  \begin{subfigure}[t]{0.24\linewidth}
    \centering
    \def\svgwidth{0.98\columnwidth}
    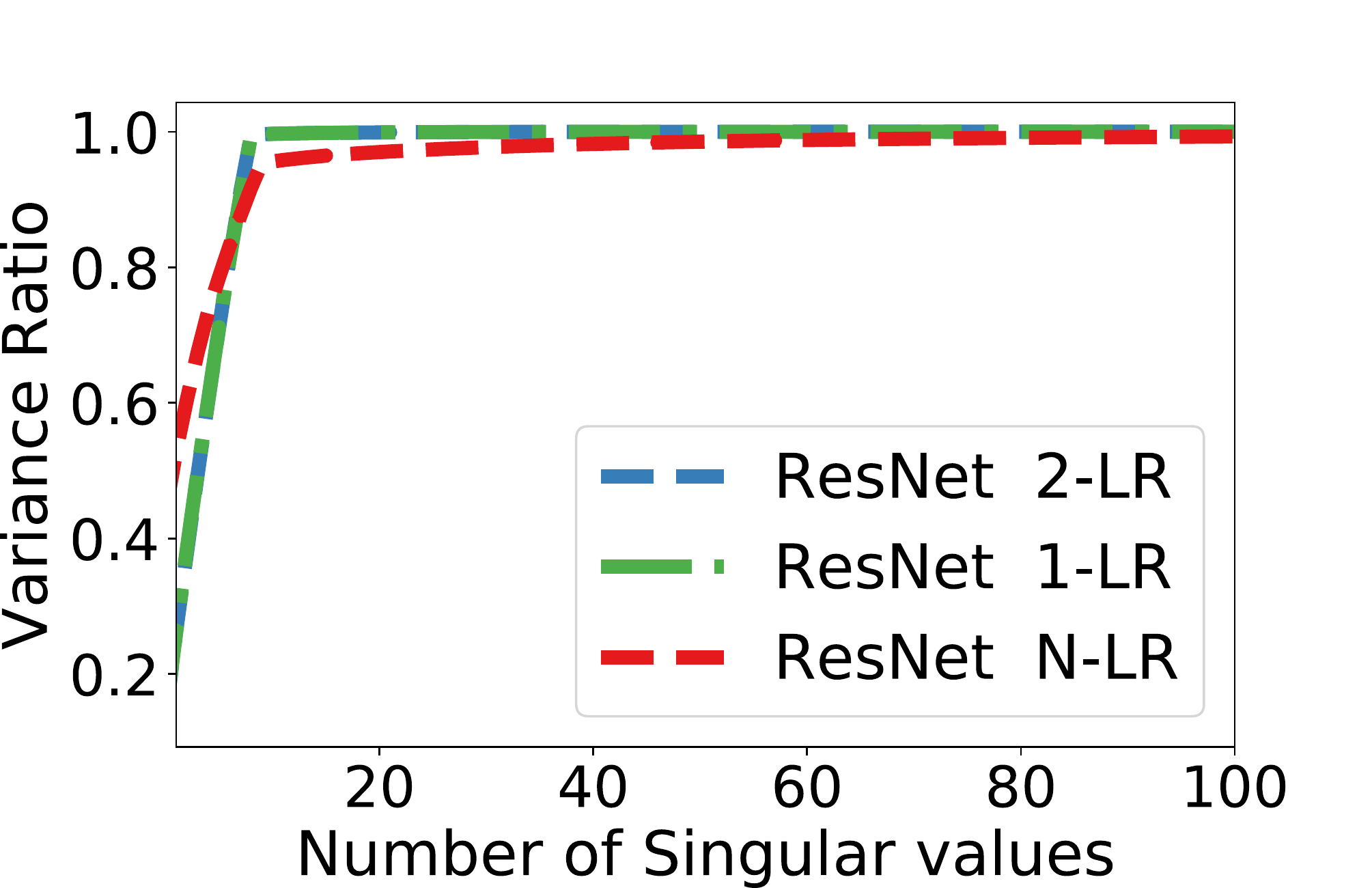
    \caption{After last ResNet block.}
    \label{fig:var_1}
  \end{subfigure}\hfill
  \begin{subfigure}[t]{0.24\linewidth}
    \centering
    \def\svgwidth{0.98\columnwidth}
    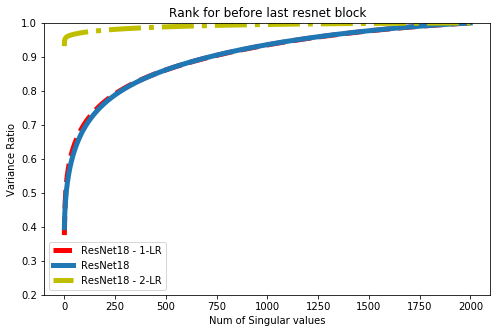
    \caption{Before last ResNet block}
    \label{fig:var_2}
  \end{subfigure}
        \begin{subfigure}[t]{0.24\linewidth}
               \centering
               \def\svgwidth{0.98\columnwidth}
               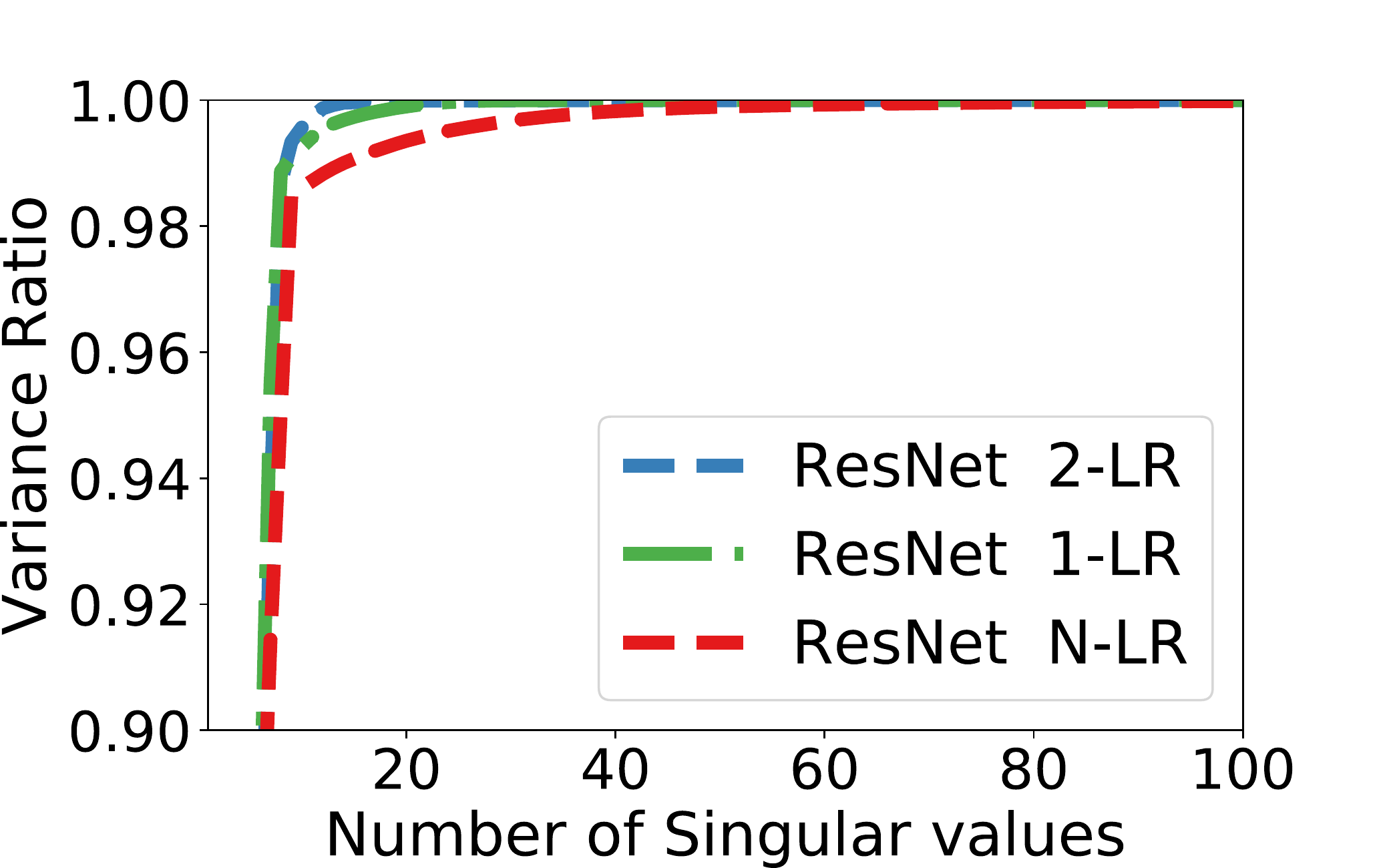
               \caption{After last ResNet block}
               \label{fig:var_3}
             \end{subfigure}\hfill
             \begin{subfigure}[t]{0.24\linewidth}
               \centering
               \def\svgwidth{0.98\columnwidth}
               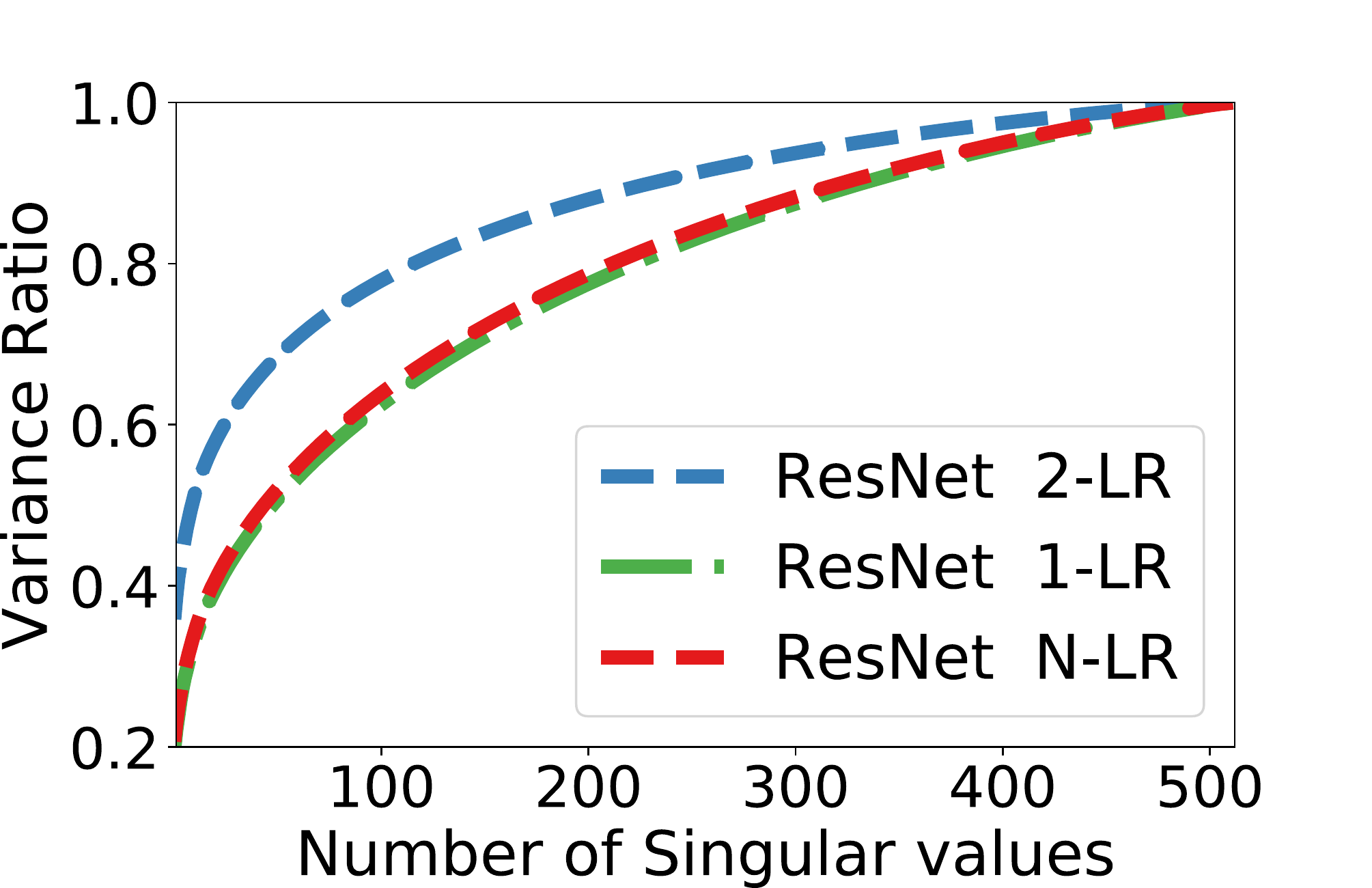
               \caption{Before last ResNet block}
               \label{fig:var_4}
             \end{subfigure}
             \caption[Variance Ratio of LR and NLR Models on CIFAR10 and SVHN]{Variance Ratio
     captured by a varying number of Singular Values in ResNet18 trained on
     CIFAR10~(\Cref{fig:var_1,fig:var_2}) and
     SVHN~(\Cref{fig:var_3,fig:var_4}). Captions indicate where the activations were extracted from.}
             \label{fig:var_ratio_plots}
           \end{figure}

We discuss low-rank weight matrices and nuclear norm penalisation as alternate
regularisations to reduce the rank of representations in~\Cref{sec:alt_algs}.
We use SVHN, CIFAR10 and CIFAR100 datasets and show results using both the
coarse labels (20 classes) and fine labels~(100 classes) of CIFAR100.
Experimentally we observe that the target rank is not a sensitive
hyper-parameter, as the training enforces a much lower rank than what it is set
to. For our experiments, we set a target rank of $100$ for the layer before the
last FC layer and $500$ for the layer before the fourth ResNet block. The
hyper-parameter $l$ in the Nystr\"om method is set to double the target rank.
The model is pre-trained with SGD for the first 50 epochs and then Algorithm
\ref{alg:lr_layer_main} is applied. The rank cutting operation is performed
every 10 iterations.

\subsection{Effective rank of learned representations}
Before we discuss our primary findings, first we
empirically show the effect of LR on the effective rank of activations. We use
the standard  \emph{variance ratio}, defined as
$\sum_{i=1}^r\sigma_i^2/\sum_{i=1}^p\sigma_i^2$, where $\sigma_i$'s are the
ordered singular values of the given activation matrix $\vec{A}$, $p$ is the
rank of the matrix, and $r\leq p$. Given $r$, a higher value of variance ratio
indicates that a larger fraction of the total variance in the data is captured
in the $r$ dimensional subspace.

\paragraph{Effective Rank} 

~\Cref{fig:var_1} shows the variance ratio for the activations before the last
FC layer in ResNet18. Note that even for NLR, the effective rank is as low as
10. A similar low-rank structure was also observed empirically
by~\citet{Oyallon_2017_CVPR}. %
However, the LR-models have almost
negligible variance leakage. Variance leakage for a certain $r$ is defined as $1 - $(variance ratio at $k$).

~\Cref{fig:var_2} shows the variance ratio for the activations before the
4\th ~ResNet block. The activation vector is $16,384$-dimensional and the use of
the Nystr\"om method ensures computational feasibility. ResNet 2-LR is the
only model that has an LR-layer in that position and these figures show that
\emph{2-LR is the only model that shows a (reasonably) low-rank structure on
that layer}. 

In~\Cref{fig:var_3,fig:var_4}, we plot the variance ratio of representations
obtained before and after the last resnet block of ResNet18 models trained on
SVHN. The LR models show a better low-rank structure than N-LR models and this
is consistent with our experiments on CIFAR10.

\begin{table}[t]\centering\small
  \begin{tabular}{llllllllllllc}
  \toprule
  &&&&\multicolumn{8}{c}{Adversarial Test Accuracy($\%)$}           &
                                                                      Test Accuracy ($\%$)    \\ \midrule
  \multicolumn{4}{c}{$L_\infty$ radius}               &   \multicolumn{2}{c}{$8/255$}      &   \multicolumn{2}{c}{$10/255$}               &   \multicolumn{2}{c}{$16/255$}   &\multicolumn{2}{c}{$20/255$}&       \\
  \multicolumn{4}{c}{Attack iterations}            & $7$     & $20$       &$7$    &$20$           &$7$    &$20$      &$7$    &$20$   &       \\\toprule
  \multirow{10}{*}{\rotatebox[origin=c]{90}{White Box}} &
                                                           \multirow{5}{*}{\rotatebox[origin=c]{90}{\footnotesize C10}}&  \multirow{2}{*}{R50}       &N-LR   & 43.1 & 31.0  & 38.5 & 21.8 & 31.2 & 7.8   & 28.9 & 4.5  & $\mathbf{95.4}$ \\
  &&&1-LR & $\mathbf{79.1}$ & $\mathbf{78.5}$   & $\mathbf{78.6}$ & $\mathbf{78.1}$          & $\mathbf{77.9}$ & $\mathbf{77.0}$  & $\mathbf{77.1}$ & $\mathbf{76.6}$ & $\mathbf{95.4}$  \\\addlinespace
  &&  \multirow{3}{*}{R18} &N-LR  & 40.9 & 26.7   & 35.1 & 16.6         & 26.7 & 4.4   &24.3&2.3 & 94.6 \\
  &&&1-LR & 48    & 31.3 & 44.4 & 25.4 & 39.6 & 17.9   & 38.2 & 15.7 & $\mathbf{94.9}$ \\
                        &                   &                &2-LR & $\mathbf{54.7}$ & $\mathbf{37.6}$  & $\mathbf{52.4}$ & $\mathbf{33.1}$   & $\mathbf{48.7}$ & $\mathbf{25.7}$  & $\mathbf{48.0}$ & $\mathbf{23.6}$ & 94.5 \\\cmidrule{4-13}
  & \multirow{5}{*}{\rotatebox[origin=c]{90}{\footnotesize C100}}&  \multirow{2}{*}{R50}       &N-LR  & 37.2 & 29.9  & 34.1 & 24.6& 29.8 & 15.9 &  34.1 & 13.3  & $\mathbf{85.8}$ \\
                        &                   &                &1-LR&   $\mathbf{45.3}$  & $\mathbf{38.7}$   & $\mathbf{43.7}$ & $\mathbf{35.8}$  & $\mathbf{40.9}$ & $\mathbf{31.5}$  & $\mathbf{40.0}$  & $\mathbf{29.8}$ &$\mathbf{85.8}$ \\\addlinespace
                        &                   &  \multirow{2}{*}{R18}       &N-LR&   30.6  & 23.2 & 26.4 & 16.9  & 20.5 & 7.42 & 18.4  & 5.1 & 84.1 \\
                        &                   &                &1-LR& $\mathbf{34.5}$\ & $\mathbf{25.4}$   & $\mathbf{31.3}$ & $\mathbf{20.2}$ & $\mathbf{27.3}$ & $\mathbf{13.1}$  & $\mathbf{25.7}$ & $\mathbf{10.8}$ & $\mathbf{84.2}$ \\
                        &                   &                &2-LR& 33.82&24.37 & 30.9 & 19.1& 26.8 & 11.83  & 25.41& 9.9 & 84    \\\midrule
  \multirow{6}{*}{\rotatebox[origin=c-10]{90}{ Black Box}} &\multirow{4}{*}{\rotatebox[origin=c]{90}{\footnotesize C10}} &  \multirow{1}{*}{R50}       &1-LR&  64.7 & 56.8  & 59.0 & 47.5  & 51.2 & 28.0 &  48.3 & 20.6  &   95.4   \\\addlinespace
                        &            &  \multirow{2}{*}{R18}       &1-LR& 66.6 & 60.8  & 61.1 & 51.0       & 52.2 & 31.52   & 49.8 & 23.6 &  94.9     \\
                        &            &       &2-LR & 68.0 & 62.5   & 62.3 & 53.4  & 53.8 & 33.5 & 50.8 & 25.8 &   94.5    \\\cmidrule{4-13}
                        &\multirow{3}{*}{\rotatebox[origin=c]{90}{\footnotesize C100}} & R50&1-LR&  52.4 & 46.2  & 48.1 & 38.8  & 42.0 & 25.4   & 48.1 & 20.9 &  85.8     \\\addlinespace
                        &     &  \multirow{2}{*}{R18} &1-LR& 53.0 & 48.6  &47.9  & 41.0   & 41.1 & 26.3 &  38.7 & 20.4  &   84.2    \\
                        &     &        &2-LR     &  51.3     &  47.2     & 46.9       & 39.9      & 39.5    & 24.3      &   37.2  &  19.2   &   84    \\\bottomrule
  \end{tabular}\caption[Adversarial Test Accuracy on CIFAR10 and CIFAR100]{Adversarial Test Accuracy against an
    $\ell_\infty$ constrained PGD adversary with the $\ell_\infty$
    radius bounded by $\epsilon$ and the number of attack steps bounded
    by $\tau$. R50 and R18 denotes ResNet50 and ResNet18
    respectively. C10 and C100 refer to CIFAR10 and CIFAR100~(Coarse
    labels) respectively. Black box attacks are generated using the N-LR model.}\label{tab:adv-robust-cifar}
  \end{table}

\subsection{Adversarial robustness}
\label{sec:adversarial-attacks-1}
We now begin our analysis on the impact of low-rank representations on
adversarial robustness. We would like to highlight that in all our experiments,
all the models are trained using a clean dataset. This is important as it shows
whether training on a clean dataset, as opposed to adversarial
training~\citep{madry2018towards}, with well-thought priors or regularisers, can
improve adversarial robustness without compromising on the clean test data
accuracy.

  \begin{table}[t]\centering\footnotesize
    \begin{tabular}{lllllllllllc}
    \toprule
    &&&\multicolumn{8}{c}{Adversarial Test Accuracy($\%)$}           &
                                                                        Test Accuracy ($\%$)    \\ \midrule
    \multicolumn{3}{c}{$L_\infty$ radius}               &   \multicolumn{2}{c}{$8/255$}      &   \multicolumn{2}{c}{$10/255$}               &   \multicolumn{2}{c}{$16/255$}   &\multicolumn{2}{c}{$20/255$}&       \\
    \multicolumn{3}{c}{Attack iterations}            & $7$     & $20$       &$7$    &$20$           &$7$    &$20$      &$7$    &$20$   &       \\\toprule
    \multirow{2}{*}{\rotatebox[origin=c]{0}{White Box}}    & \multirow{2}{*}{R50}       &N-LR& 27.8 & 21.8  & 25.2 & 17.7  & 21.1 & 17.7   & 19.4 & 7.6  & 77.2 \\
                          &                       &1-LR&  $\mathbf{28.8}$ & $\mathbf{23.6}$   & $\mathbf{26.8}$ & $\mathbf{21.4}$   & $\mathbf{24.4}$ & $\mathbf{17.8}$  & $\mathbf{23.5}$ & $\mathbf{16.5}$  & $\mathbf{77.7}$ \\\addlinespace
    \multirow{1}{*}{\rotatebox[origin=c-10]{0}{ Black Box}}     & R50       &1-LR & 38.3 & 32.8   & 34.3 & 26.4   & 28.9 & 15.0   & 27.0 & 11.7 &   -   \\\bottomrule
    \end{tabular}\caption[Adversarial accuracy of LR ResNet50 on CIFAR100]{Adversarial Test Accuracy against a
      $\ell_\infty$ constrained PGD adversary with the $\ell_\infty$
      radius bounded by $\epsilon$ and the number of attack steps bounded
      by $\tau$ on CIFAR100 fine labels for ResNet50.}\label{tab:adv-robust-cifar100-fine-app}
    \end{table}

We recall that adversarial noise are well crafted~(almost imperceptible) input
perturbations that, when added to a clean input, flips the prediction of the
model to an incorrect one~\citep{Dalvi2004,Biggio2018,szegedy2013intriguing}.
Various methods~\citep{szegedy2013intriguing, goodfellow2014explaining,
kurakin2016adversarial,mosaavi2016,Carlini2017,Papernot2016} have been proposed
in recent years for constructing adversarial perturbations. 
Here we use the following three {\bf white-box} 
adversarial attacks to perform our experiments:
\begin{enumerate*}[label=(\roman*)]
\item Iterative Fast Sign Gradient Method (\ifgsm~or
IFGSM)~\citep{kurakin2016,madry2018towards}, 
\item Iterative Least Likely Class
Method~(\ill~or ILL)~\citep{kurakin2016adversarial}, and 
\item
\deepfool~(DFL)~\citep{mosaavi2016}.
\end{enumerate*} 
We discussed these attacks in detail
in~\Cref{sec:adv-attack-bg}. Iter-FGSM is essentially equivalent to the
Projected Gradient Descent (PGD) with $\ell_\infty$ projections on the negative
loss function~\citep{madry2018towards}.

We also consider the {\bf black-box} version of each of the aforementioned
adversarial attacks where the noise is constructed using N-LR. This is
to avoid situations where LR might be at an advantage due to the
low-rank structure that might enforce a form of gradient
masking~\citep{tramer2018ensemble}.

\paragraph{Robustness to Adversarial Attacks} In Table~\ref{tab:adv-robust-cifar}, we measure the adversarial test
accuracy of ResNet18 and ResNet50 models trained on CIFAR10 and
CIFAR100 respectively. %
The adversary used here is an $L_\infty$ PGD adversary~(or IFGSM) that has two
main constraints --- the $\ell_\infty$ radius and the number of attack-steps the
PGD algorithm can take. The $\ell_\infty$ radius is chosen from
$\bc{\nicefrac{8}{255},\nicefrac{10}{255},\nicefrac{16}{255},\nicefrac{20}{255}}$
with either 7 or 20 attack steps of PGD. This represents a wide variety of
severity in the attack model and our LR model performs much better than the N-LR
model in all the settings including the black-box settings. For example, in the
case of a white-box attack with $\ell_\infty = \frac{16}{255}$ and $20$ attack
steps on a ResNet-50 model trained on CIFAR10, the LR ResNet model is nearly
{\bf 10 times more accurate} than the N-LR model. 

In~\Cref{tab:adv-robust-cifar100-fine-app}, we show the adversarial
test accuracy for varying perturbation budgets for ResNet50 trained on
the fine labels of CIFAR100. LR models, not only have the best
adversarial test accuracies but also the best natural test accuracies.

\begin{table}[t]\centering%
  \begin{tabular}{l@{\hspace{20pt}}c@{\hspace{2.5pt}}c@{\hspace{10pt}}c@{\hspace{2.5pt}}c@{\hspace{10pt}}c@{\hspace{2.5pt}}c@{\hspace{6pt}}}
 \toprule
 &\multicolumn{6}{c}{\parbox[c]{2.5in}{Adversarial Test
            Accuracy($\%$)}}\\\midrule%
 $L_\infty$ radius &   \multicolumn{2}{c}{$\nicefrac{8}{255}$}        &   \multicolumn{2}{c}{$\nicefrac{16}{255}$}   &\multicolumn{2}{c}{$\nicefrac{20}{255}$}      \\
 Att. iter  & $7$     & $20$       &$7$ &$20$      &$7$    &$20$          \\\toprule
   N-LR   & 43.1 & 31.0   & 31.2 & 7.8   & 28.9 & 4.5   \\
   SNIP          &   29.4 & 14.5 & 18.5 & 1.3  & 16.2 & 0.4 \\
   SRN    & 47.8& 37.6& 39.8&  21.3& 37.5& 18.4\\\addlinespace
   LR~(Ours) & $\mathbf{79.1}$ & $\mathbf{78.5}$   & $\mathbf{77.9}$ & $\mathbf{77.0}$  & $\mathbf{77.1}$ & $\mathbf{76.6}$ \\\bottomrule
 \end{tabular}\caption[Adversarial accuracy of other regularisation methods]{Robustness of other  regularisation/compression
   approaches to Adversarial attacks. %
 }\label{tab:adv-pert-compre}
 \end{table}

The above results indicate that LR provides low-rank representations
that are robust to adversarial perturbations and also provide modest
improvements on the test accuracy.

In Table~\ref{tab:adv-pert-compre}, we also compare our LR with other methods
that reduce some form of intrinsic dimensions of the parameter space.
SNIP~\citep{lee2018snip}, a pruning technique, increases the sparsity of the
parameter and SRN~\citep{sanyal2020stable} reduces the~(stable) rank and
spectral norm of the parameters.%
~We use the best hyper-parameter
settings suggested for these approaches in their manuscripts. For a
description of these methods and other related approaches please refer to
Appendix~\ref{sec:alt_algs}. Our method performs much better than all
these methods indicating that reducing the dimensionality of the representation
space is much more effective than doing so for the parameter space
when it comes to the robustness of the network.

\paragraph{Accuracy vs the amount of adversarial noise}
Next, we compare the change in accuracy
of adversarial classification with respect to the actual amount of
noise added (as opposed to the perturbation budget as in Table~\ref{tab:adv-robust-cifar}). The amount of noise added can be measured using the
normalized $L_2$ dissimilarity score ($\rho$), defined as:
$$ \rho = \bE\left[\norm{\vx_a -\vx_d}_2/\norm{\vx_d}_2\right],$$ where $\vx_d$
and $\vx_a$ are the clean and adversarially perturbed samples, respectively. The
normalized $L_2$ dissimilarity score $\rho$ measures the magnitude of the
noise~\citep{mosaavi2016} in the input corresponding to a certain adversarial
misclassification rate\footnote{Similar to the setting
in~\citet{kurakin2016adversarial}, the noise is added for a pre-determined
number of steps.}. 

\begin{figure}[t]%
  \centering
  \begin{subfigure}[c]{0.24\linewidth}
    \centering
    \def\svgwidth{0.99\columnwidth}
    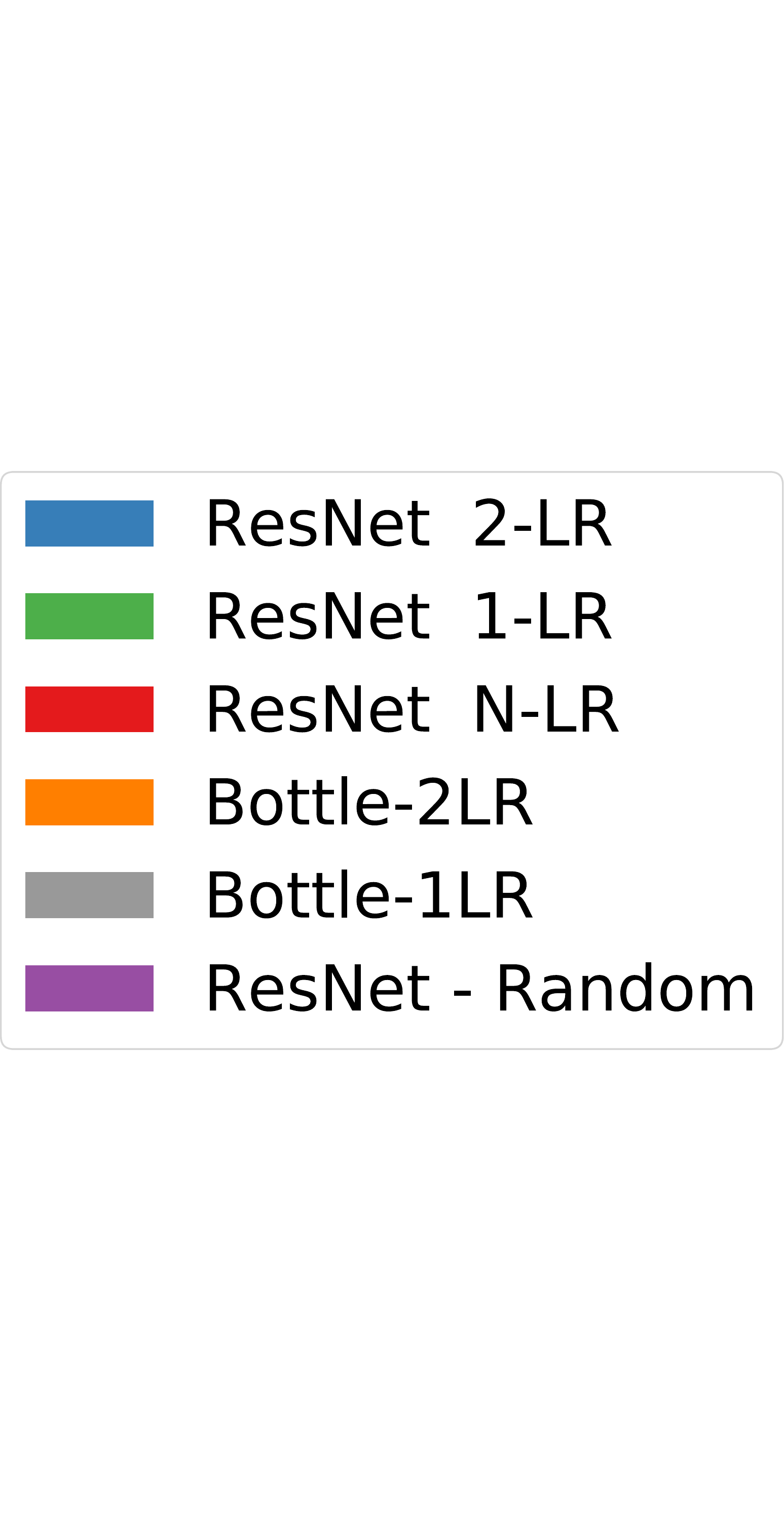
  \end{subfigure}
  \begin{subfigure}[c]{0.35\linewidth}
    \centering
    \def\svgwidth{0.99\columnwidth}
    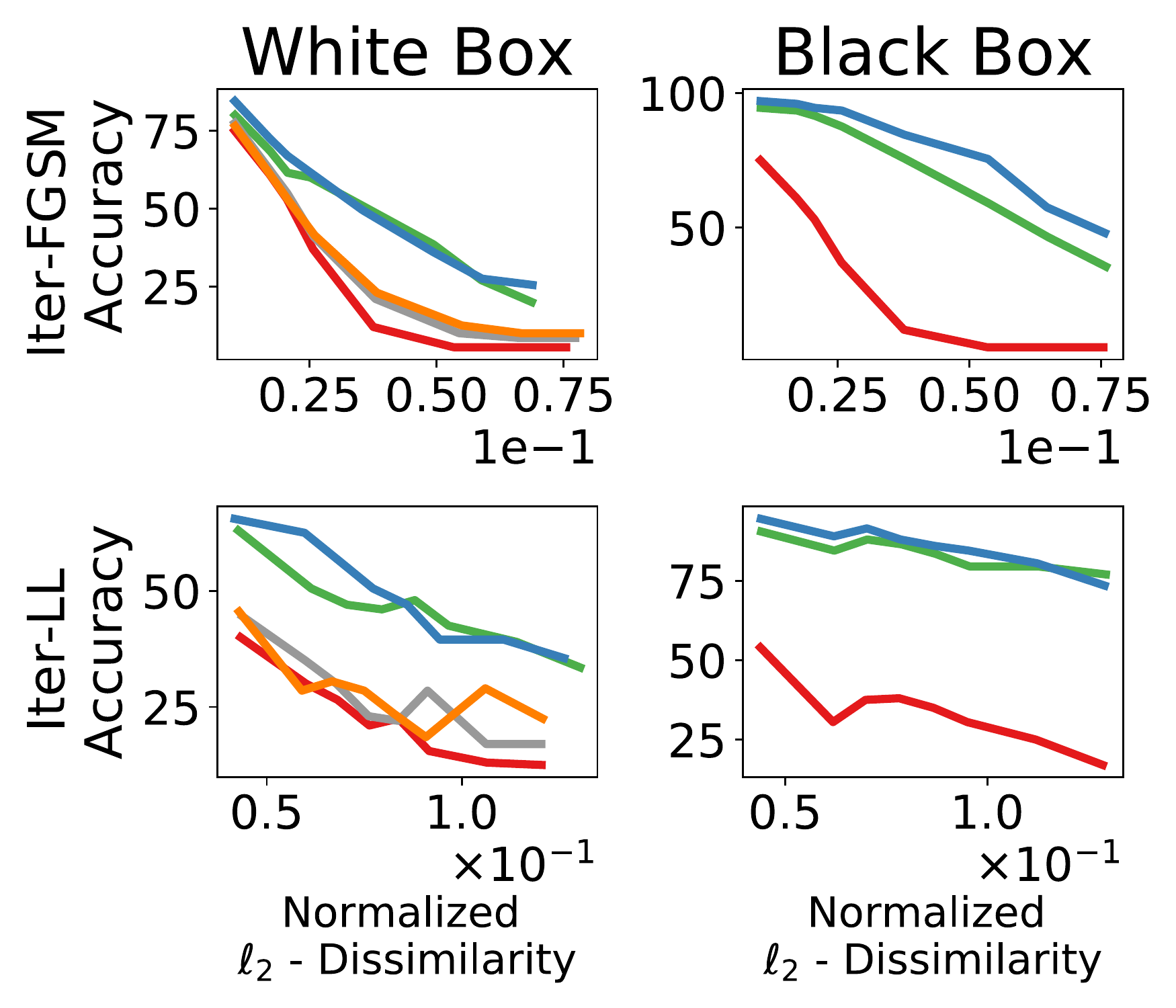 \caption*{CIFAR10}%
  \end{subfigure}
    \begin{subfigure}[c]{0.39\linewidth} \centering
      \def\svgwidth{0.99\columnwidth}
      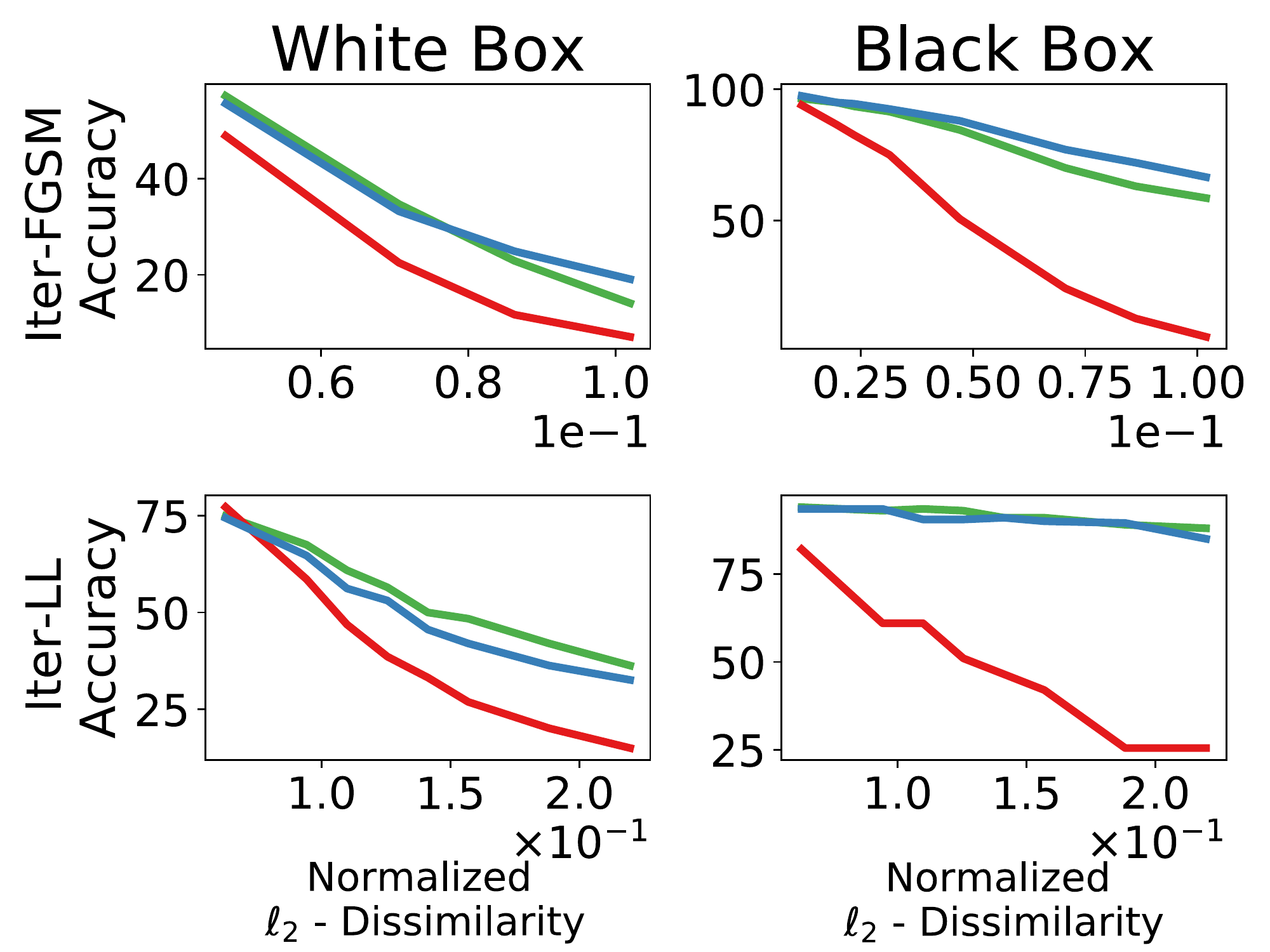\caption*{SVHN}\label{fig:adv-r18-svhn}
      \end{subfigure} \caption[Adversarial accuracy vs perturbation magnitude of ResNet18]{Adversarial accuracy of ResNet18 against the magnitude
      of perturbation~(measured by $\rho$) on CIFAR10 and SVHN.}
      \label{fig:pert}
\end{figure}

Here we also consider another model, for comparison, we call the {\em Bottle-LR}
model. It contains an {\em explicit bottleneck} low-rank layer, rather than \lr,
which is essentially a fully connected layer without any non-linear activation
where the weight matrix $\vec{\bar{W}}\in\reals^{q\times q}$ is parameterised by
$\vec{\bar{W}}_l\in{\reals^{q\times r}}$, where $r\leq q$, so that
$\vec{\bar{W}}=\vec{\bar{W}}_l\vec{\bar{W}}_l^\top$. Note, by design, it can not
have a rank greater than $r$. 

\Cref{fig:pert} shows that as the noise increases, the accuracy of N-LR models decreases much
faster than the LR models.  Specifically, to reach an adversarial
misclassification rate of $50\%$, our models require about twice the noise as
the N-LR or Bottle-LR. 
~\emph{For \emph{all} kinds of attacks we considered, LR models consistently
outperform N-LR and Bottle-LR.} 

Even though the rank constraint of Bottle-LR is the same as LR models, and they
 are placed at the same position as the LR layers, the inferior performance of
 Bottle-RL (sometimes even worst than N-LR) can be explained by the fact that
 the explicit bottleneck has a \emph{multiplicative} effect on the
 back-propagated gradients, whereas, LR's impact on gradients is
 \emph{additive}. With a bottleneck layer, the gradient of any $\vec{W}_{j}$
 before the bottleneck layer is $\nicefrac{\partial \mathcal{L}}{\partial
 \vec{W}_{j}}=(\nicefrac{\partial \mathcal{L}}{\partial \vec{z}_l}) \bar{W}
 (\nicefrac{\partial \vec{a}_{l-1}}{\partial \vec{W}_j})$ where $\vec{a}_l$ and
 $\vec{z}_l$ are the activations and the pre-activations of the $l^{\it{th}}$
 layer respectively. Thus, especially during the early stages of training when
 $\bar{\vec{W}}$ is not yet learned, important directions in
 $(\nicefrac{\partial \vec{a}_{l-1}}{\partial \vec{W}_j})$ can be cancelled out
 due to the low-rank nature of $\bar{\vec{W}}$, thus making  $\nicefrac{\partial
 \mathcal{L}}{\partial \vec{W}_{j}}$  uninformative. In the case of LR, the
 gradients from $\mathcal{L}_c$ (depending on $W$) and $\mathcal{L}$ are
 additive and thus the low-rank $\vec{W}$ only affects the gradients from $\cL$
 additively. The auxiliary loss $\mathcal{L}_c$ has a less direct impact on
 early training (in terms of classification error). We believe this relative
 ``smoothness'' of our approach is the reason why it has a better performance on
 these other tasks.
  
\paragraph{Minimum adversarial perturbation for $99\%$ misclassification}

\begin{table}\centering
  \begin{tabular}[h!]{l@{\quad}l@{\quad}c@{\quad}c@{\quad}c@{\quad}}\toprule
    &Model&$\rho$ [\deepfool]&$\rho$ [\ill]&$\rho$ [\ifgsm]\\\toprule
     \multirow{3}{1.2cm}{White Box}&2-LR&$\mathbf{1.8\times 10^{-1}}$&$9.8\times 10^{-2}$&$\mathbf{7.6\times 10^{-2}}$\\%\cline{2-5}
            &1-LR&$1.7\times 10^{-1}$&$\mathbf{1.1\times 10^{-1}}$&$6.0\times 10^{-2}$\\%\cline{2-5}
            &N-LR&$1.6\times10^{-2}$&$2.4\times10^{-2}$&$2.1\times10^{-2}$\\\midrule
     \multirow{2}{1.2cm}{Black Box}&2-LR&$\mathbf{5.5\times10^{-2}}$&$\mathbf{2.0\times 10^{-1}}$&$\mathbf{7.5\times 10^{-2}}$\\%\cline{2-5}
            &1-LR&$4.7\times10^{-2}$&$1.8\times 10^{-1}$&$5.6\times10^{-2}$\\\bottomrule
   \end{tabular}
   \caption[Minimum perturbation for $99\%$ adversarial
   error in ResNet models]{Minimum perturbation required for $99\%$ Adversarial
   Misclassification by ResNet18 models on
   CIFAR10.~Table~\ref{tab:adv_rob_pert_eps} shows the $\ell_\infty$
   perturbations used. The values of the perturbation budget also show that the
   minimum perturbation required for $99\%$ misclassification is much higher
   for LR models than N-LR models.}
   \label{tab:adv_rob_pert}%
 \end{table}

 \begin{table}[!htb]  \centering
  \begin{tabular}[h!]{l@{\quad}l@{\quad}c@{\quad}c@{\quad}}\toprule
    &Model&$\epsilon$~[\ill]&$\epsilon$~[\ifgsm]\\\toprule 
    \multirow{3}{*}{White Box}&2-LR&\(4\times 10^{-2}\)&\(2\times 10^{-2}\)\\ 
&1-LR&\(6\times 10^{-2}\)&\(1\times 10^{-2}\)\\ 
&N-LR&\(1\times 10^{-2}\)&\(1\times 10^{-2}\)\\\midrule 
\multirow{2}{*}{Black Box}&1-LR&\(8\times 10^{-2}\)&\(1\times 10^{-2}\)\\ 
&2-LR&\(1\times 10^{-1}\)&\(1\times 10^{-2}\)\\\bottomrule
  \end{tabular}
  \caption[Minimum$\epsilon$ for Adversarial
  Misclassification corresponding to Table~\ref{tab:adv_rob_pert}.]{Value for $\epsilon$ required for Adversarial
Misclassification corresponding to Table~\ref{tab:adv_rob_pert}.}
  \label{tab:adv_rob_pert_eps}
\end{table}
Our next experiment is along the lines of that reported in~\citet{mosaavi2016}.
Table~\ref{tab:adv_rob_pert} shows the average minimum perturbation (measured by
$\rho$) required to make the classifier misclassify more than $99\%$ of the
adversarial examples, constructed from a uniformly sampled subset of the test
set. We use the deepfool algorithm described in Algorithm 2
in~\citet{mosaavi2016}~(also discussed in~\Cref{sec:adv-attack-bg}). The
algorithm returns the minimum perturbation $r(\vx)$ required to make the
classifier misclassify the point $\vx$.  The $L_2$ dissimilarity is obtained
by calculating $\rho = \frac{r(\vx)}{\norm{\vx}_2}$. For our experiments, we
used the publicly available
code of \deepfool.~\footnote{\url{https://github.com/LTS4/DeepFool/blob/master/Python/deepfool.py}}.

Even under this scheme of attacks, our models perform better than N-LR as LR
models require 4 to 11 times the amount of noise required by  N-LR models to be
fooled by adversarial attacks~(see~\Cref{tab:adv_rob_pert}). This can also be
visualised in Figure~\ref{fig:adv_images_noticeable}. The adversarial images for
2-LR and 1-LR are noticeably much more perturbed than N-LR.  An interesting
observation is that the values of $L_2$ dissimilarity in
Table~\ref{tab:adv_rob_pert} are lower than those in Figure~\ref{fig:pert}
though the attacks have a higher rate of success. The essential difference
between the attacks in Figure~\ref{fig:pert} and Table~\ref{tab:adv_rob_pert} is
in the number of iterations for which the updates are executed while creating
the attack. In Figure~\ref{fig:pert}, the step is executed for a fixed number of
steps whereas, in Table~\ref{tab:adv_rob_pert}, the updates are executed until
the classifier makes a mistake. This can be explained by our discussion
around~\Cref{fig:adv_fix_step} where we show that attacks that stop adding
adversarial perturbations upon successful misclassification are more powerful
than those that add the perturbation for a fixed number of steps.

\begin{figure}[t]\centering
  \begin{subfigure}{0.6\linewidth}
\parbox[b][22pt][t]{0.17\linewidth}{Input}
\def\svgwidth{0.18\linewidth}
\hspace{-2pt}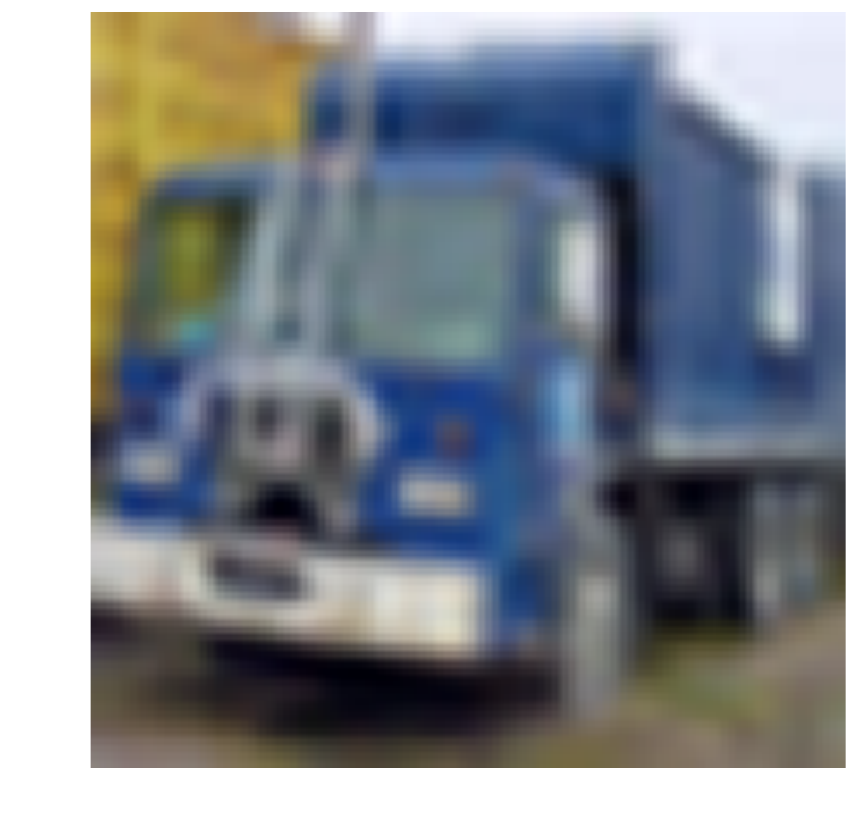\hspace{0pt}
\def\svgwidth{0.18\linewidth}
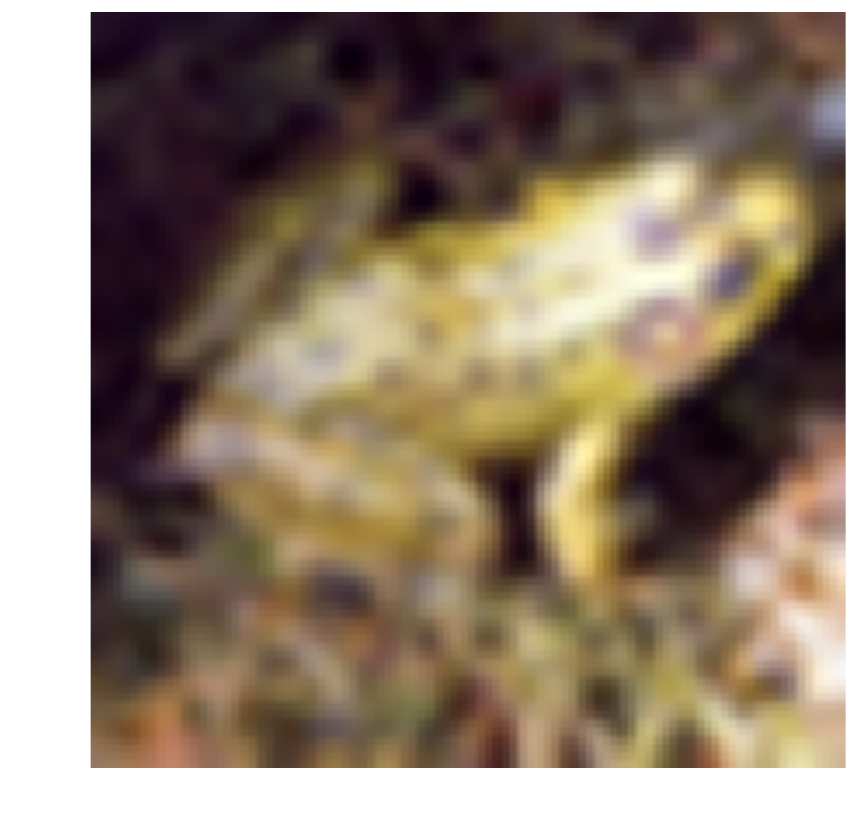%
\hspace{0pt}
\def\svgwidth{0.18\linewidth} 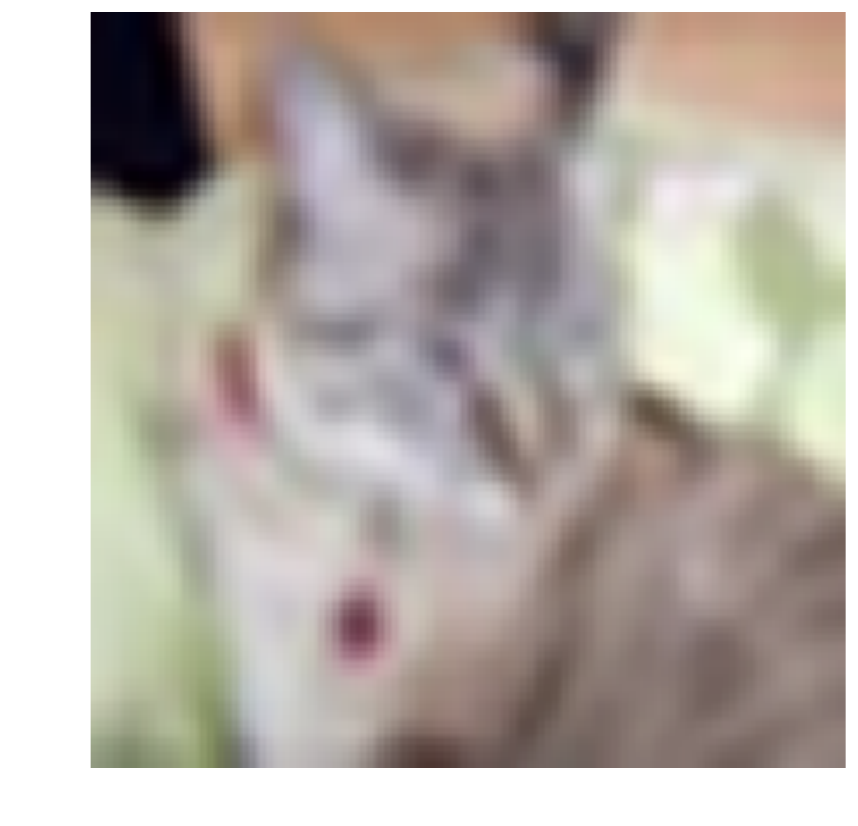\\
\parbox[b][22pt][t]{0.17\linewidth}{{2-LR}}\def\svgwidth{0.18\linewidth}
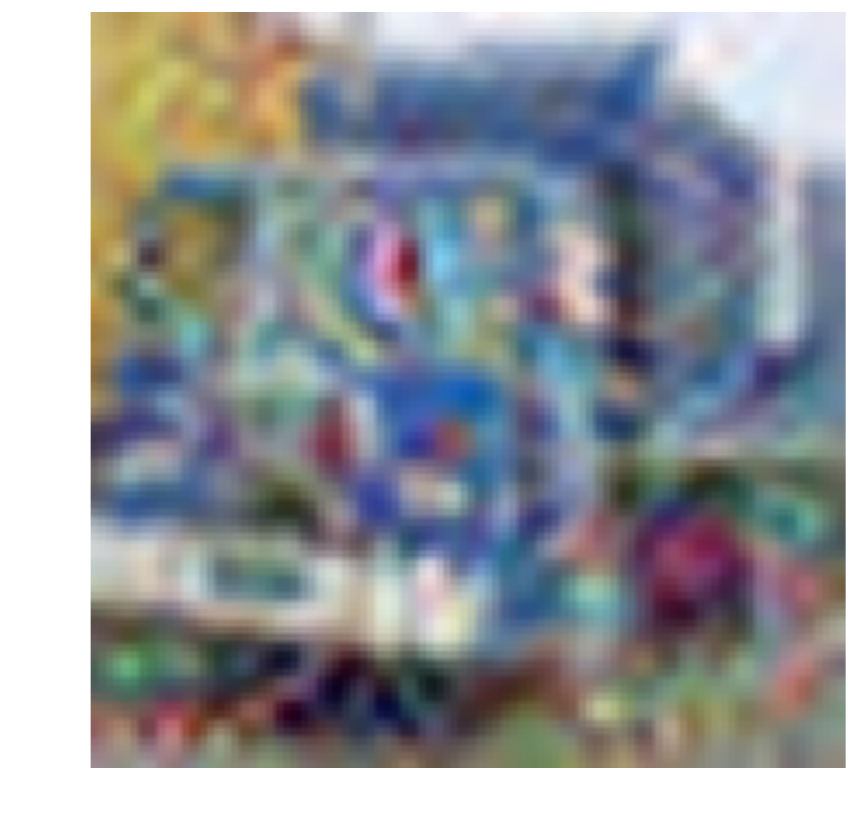
\def\svgwidth{0.18\linewidth} 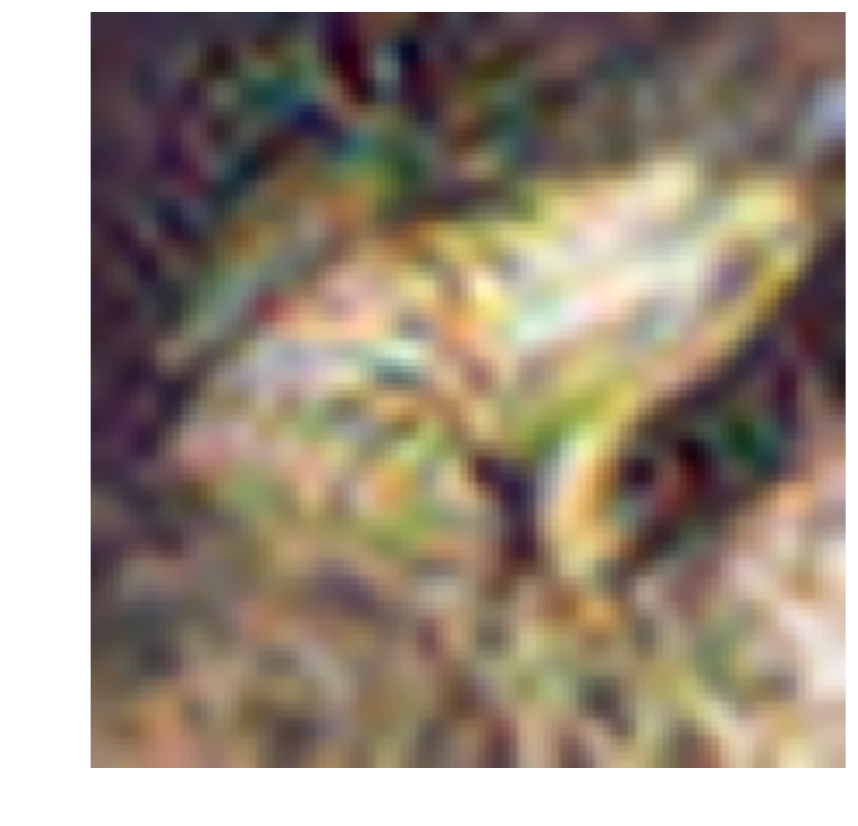
\def\svgwidth{0.18\linewidth} 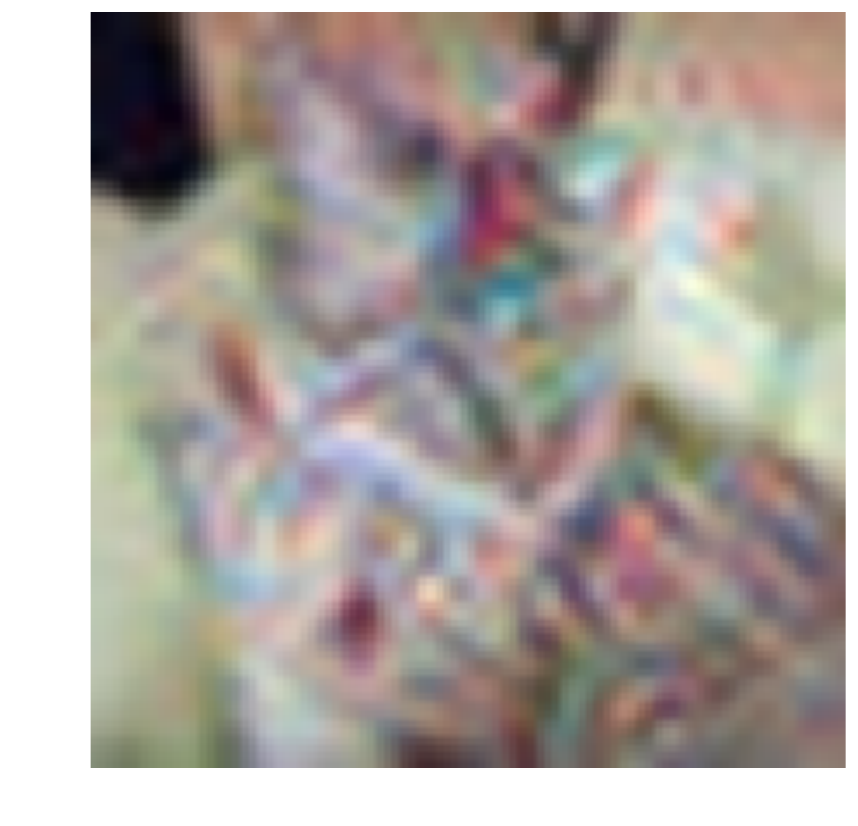\\
\parbox[b][22pt][t]{0.17\linewidth}{{1-LR}}\def\svgwidth{0.18\linewidth}
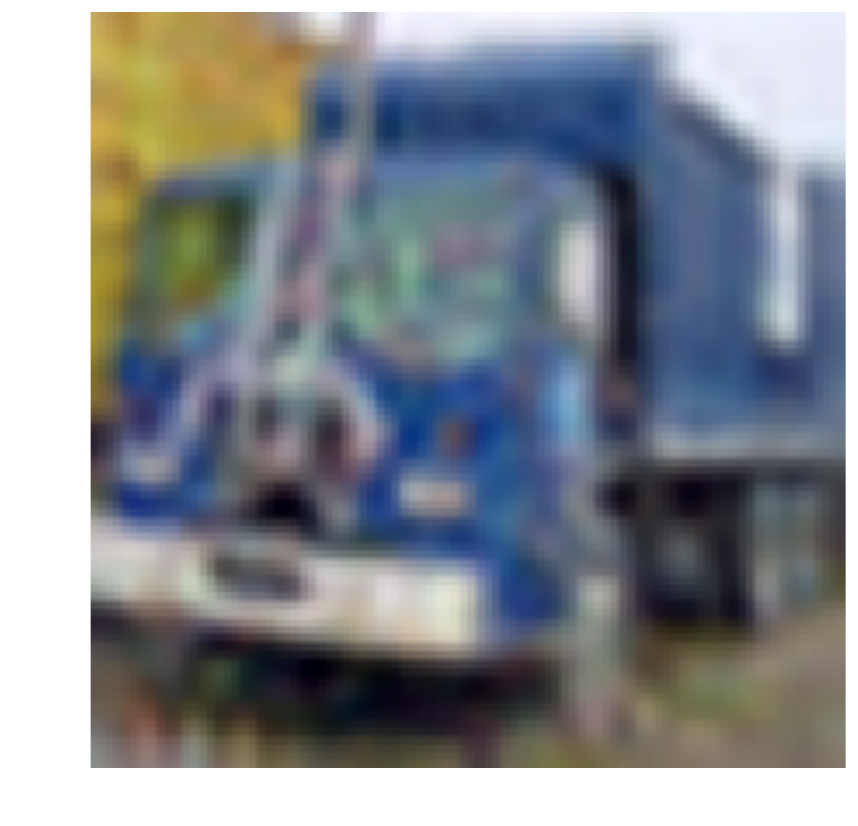
\def\svgwidth{0.18\linewidth} 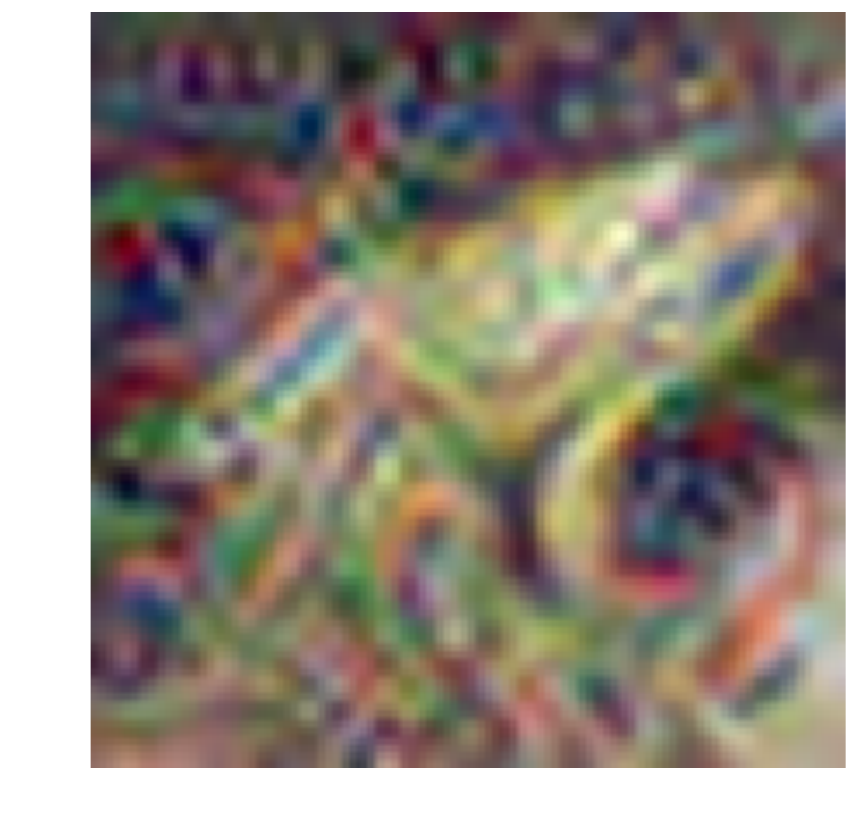
\def\svgwidth{0.18\linewidth} 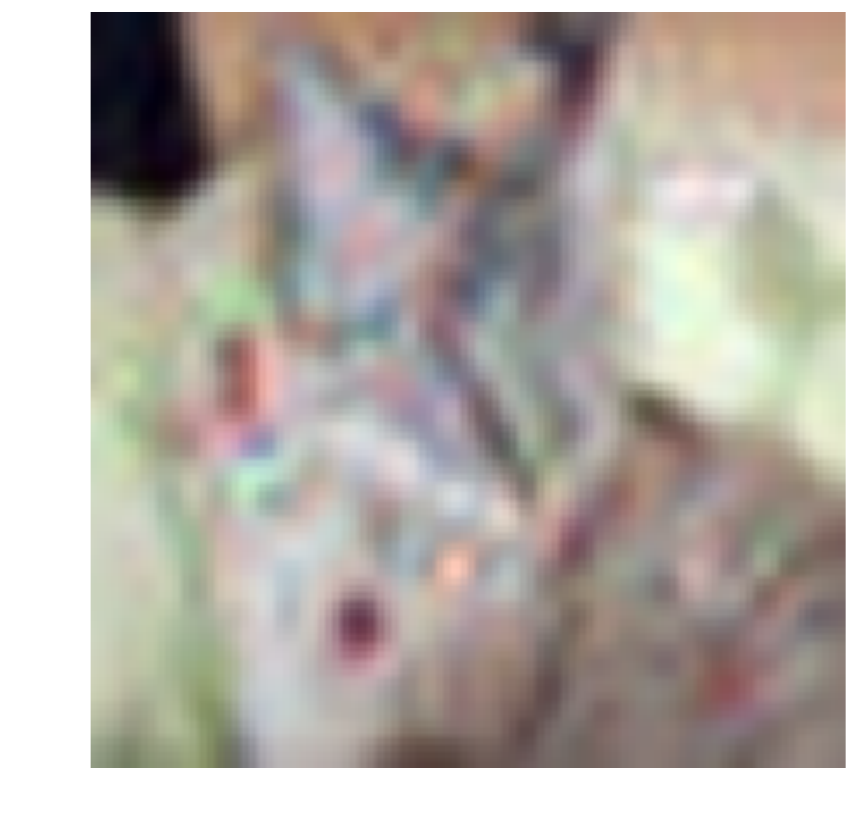\\
\parbox[b][22pt][c]{0.17\linewidth}{ResNet}\def\svgwidth{0.18\linewidth}
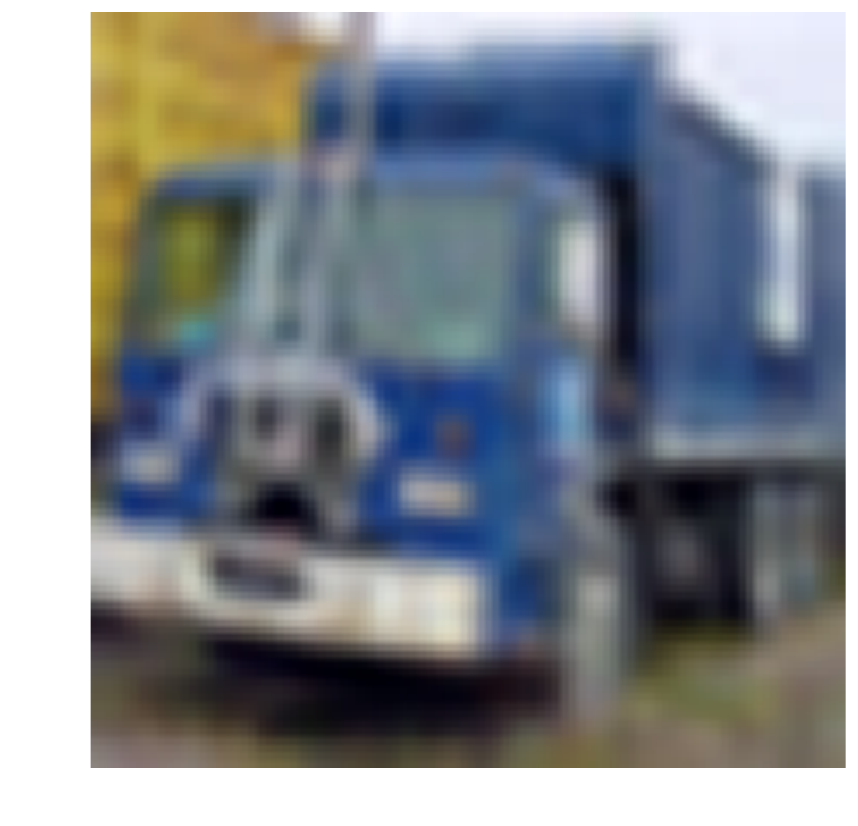
\def\svgwidth{0.18\linewidth} 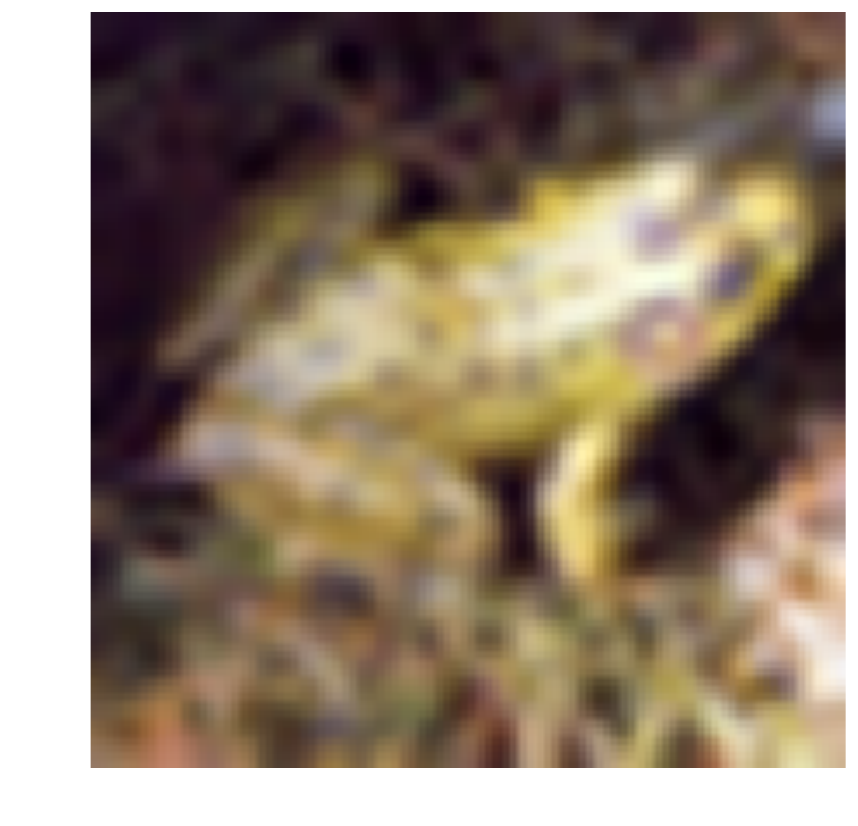
\def\svgwidth{0.18\linewidth} 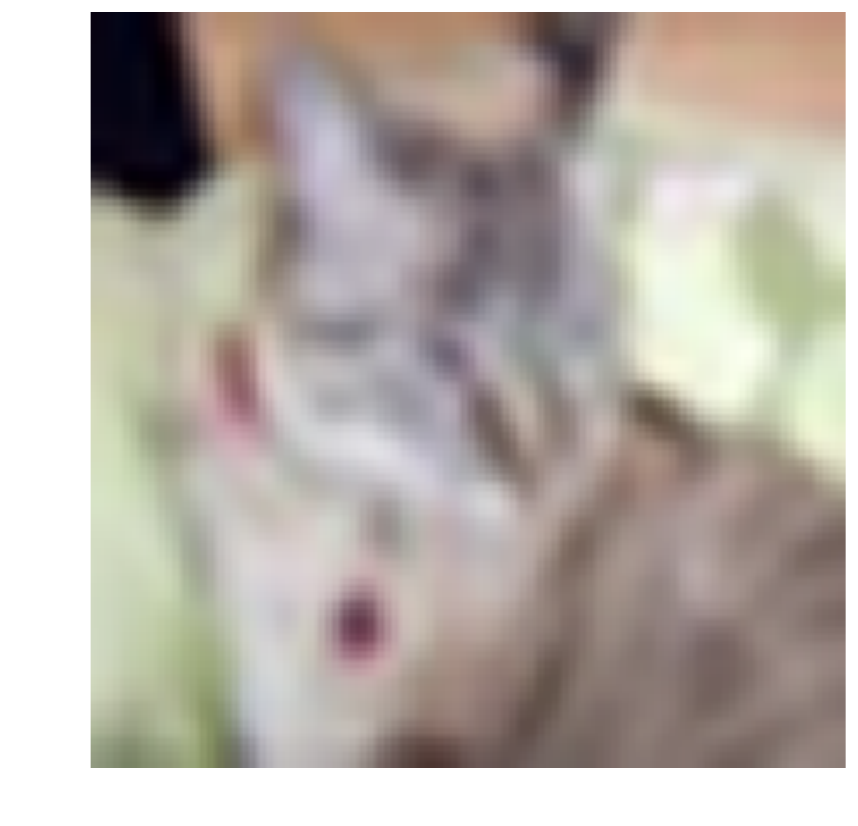
\end{subfigure}
 \caption[Adversarial images using DeepFool against LR and NLR models]{ Adversarial images using DeepFool against different
classifiers. The images against low-rank (LR) classifiers require more
perturbations and are noticeably different.}
\label{fig:adv_images_noticeable} 
\end{figure}

\subsection{Noise stability}

To gain some understanding of this visibly better adversarial robustness of LR
models, in this section, we study the noise stability behaviour of LR models in
detail. Specifically, we show that LR models~(and their representations) are
significantly more stable to input perturbations at test time even when training
is performed using clean data. %

\begin{table}[!htb]
    \centering
    \begin{tabular}{lllllll}\toprule
      \multicolumn{2}{c}{Pert. Prob.~($p$)}& 0.4 & 0.6 & 0.8 & 1.0 \\\toprule
      \multirow{2}{*}{R50}&N-LR& $69.7$ & $26.1$ & $12.6$ & $11.3$\\
                                           &1-LR&$\mathbf{75.1}$&$\mathbf{34.2}$&$\mathbf{15.8}$ 
                          &$\mathbf{13.0}$&\\\addlinespace
      \multirow{3}{*}{R18}&N-LR &$57.7$&$27.3$&$13.0$&$7.2$&\\
                                           &1-LR&$\mathbf{75.1}$&$33.0$&$15.2$&$11.0$\\
                                           &2-LR&$74.1$&$\mathbf{35.5}$&$\mathbf{16.4}$&$\mathbf{11.5}$\\\bottomrule
    \end{tabular}
    \caption[Robustness to random additive pixel perturbations]{Test accuracy of ResNet50~(R50) and ResNet18(R18) to
      Gaussian noise~$\cN\br{0,\nicefrac{128}{255}}$ introduced at
      each pixel with probability $p$. Evaluated on CIFAR10.}\vspace{-2ex}
    \label{tab:rand-noise-robust}
  \end{table}
\textbf{Random Pixel Perturbations} In ~Table~\ref{tab:rand-noise-robust}, we measure the test accuracy
when the input is perturbed with random additive noise.  Specifically, for a given pixel and a given \emph{pixel perturbation
  probability} $p\in\bc{0.4, 0.6, 0.8, 1.0}$, we toss a biased coin~(with
bias $p$)~%
and if heads, add to the pixel a Gaussian noise drawn from~$\cN\br{0,
\nicefrac{128}{255}}$. This is done for all the pixels in the test-set and the
test accuracy is measured over this perturbed dataset. For varying levels of
perturbation,~Table~\ref{tab:rand-noise-robust} shows that LR models are
significantly  more stable to Gaussian noise than N-LR. Our experiments indicate
that learning a model that cancels out irrelevant directions in the
representations suppresses the propagation of the input noise in a way to reduce
its effect on the output of the model. Interestingly, the level of Gaussian
noise seems to not vary the test accuracy as much as the value of $p$ does.

\paragraph{Stability of representations}
In~\Cref{fig:perturbation_spaces,fig:perturbation_spaces_svhn}, we show
how the input adversarial perturbations propagate and impact the feature space
representations. Specifically, for a given adversarial perturbation $\delta$ to
an input $\vec{x}$, the $x$-axis is the normalized $L_2$ dissimilarity score in
the input space i.e. ${\norm{\delta}^2}/{\norm{\vec{x}}^2}$ and the $y$-axis
represents the corresponding quantity in the representation space i.e.
$\norm{f_{\ell}^{-}\br{\vec{x+\delta} } - f_{\ell}^{-}\br{\vec{x}}}^2
/{\norm{f_{\ell}^{-}\br{\vec{x}}}^2}$. The representations $f_{\ell}^{-}(.)$
here are taken from before the last fully connected layer. As our experiments
suggest, the LR model significantly attenuates the adversarial perturbations
thus making it harder to fool the softmax classifier. This observation further
supports the increased robustness of LR.

\begin{figure}[t]
  \begin{subfigure}[c]{0.2\linewidth}
    \centering
    \def\svgwidth{0.99\columnwidth}
    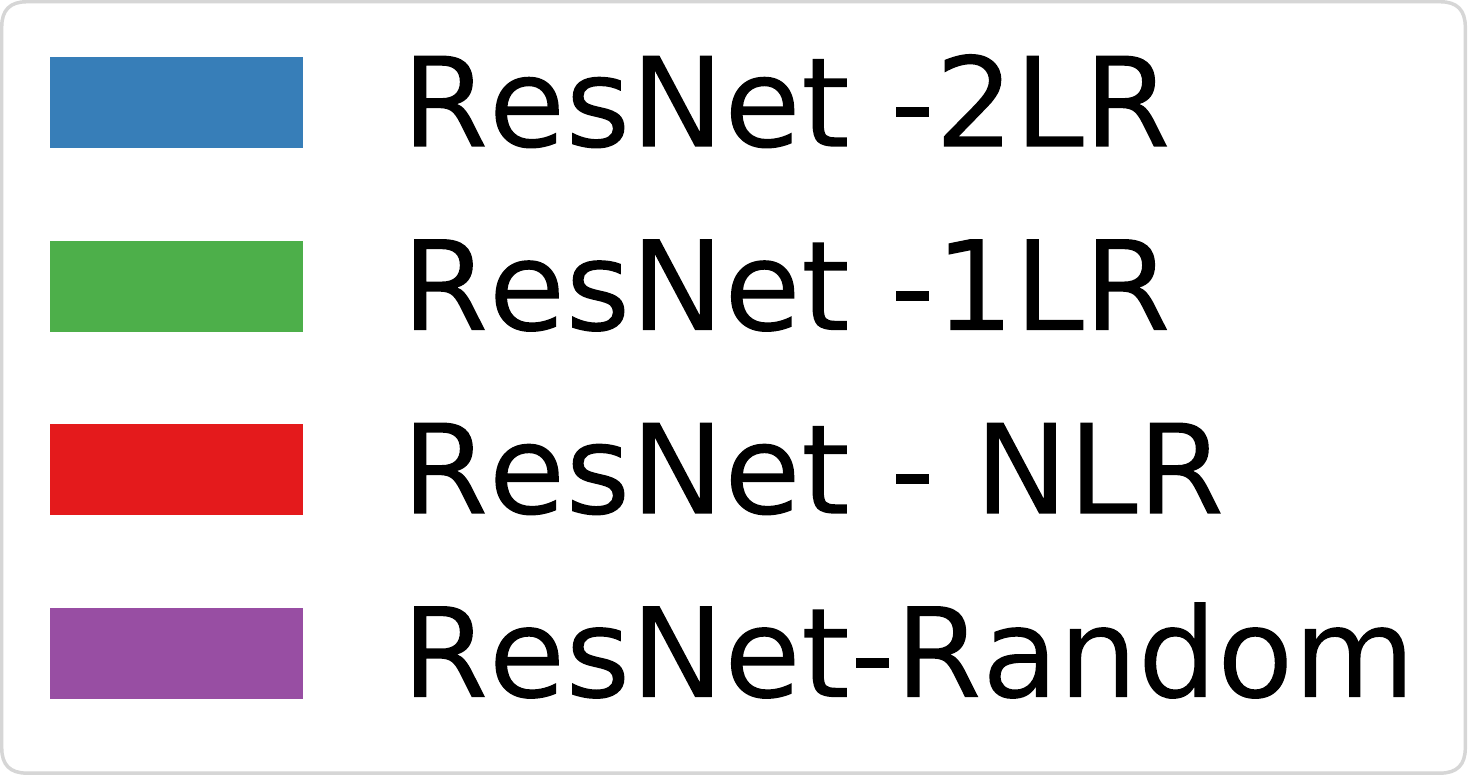
  \end{subfigure}
    \begin{subfigure}[c]{0.8\linewidth}
        \begin{subfigure}[c]{0.3\linewidth}
          \centering
          \def\svgwidth{0.99\columnwidth}
          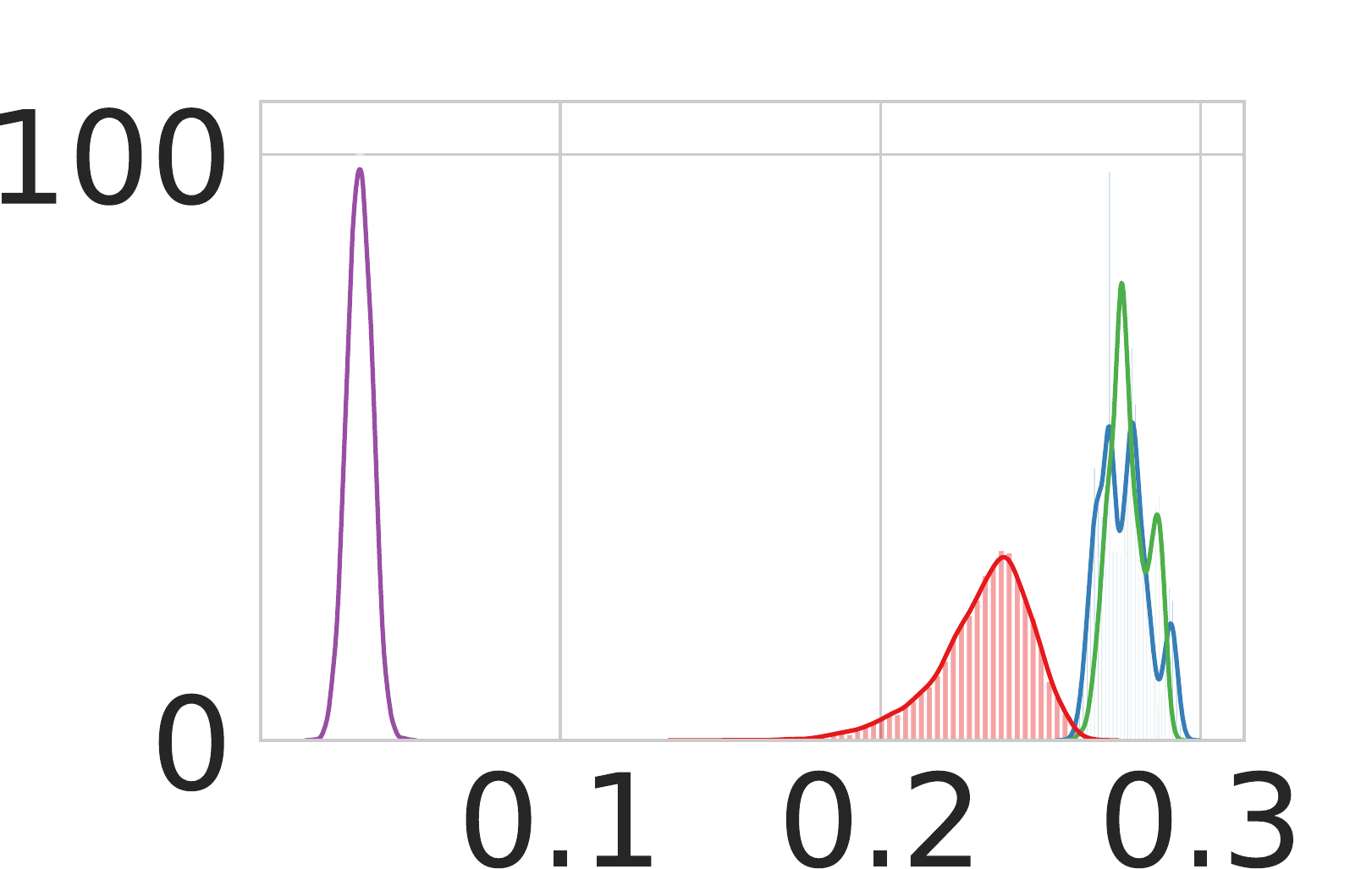\caption*{FC Layer}
        \end{subfigure}
        \begin{subfigure}[c]{0.3\linewidth}
          \centering
          \def\svgwidth{0.99\columnwidth}
          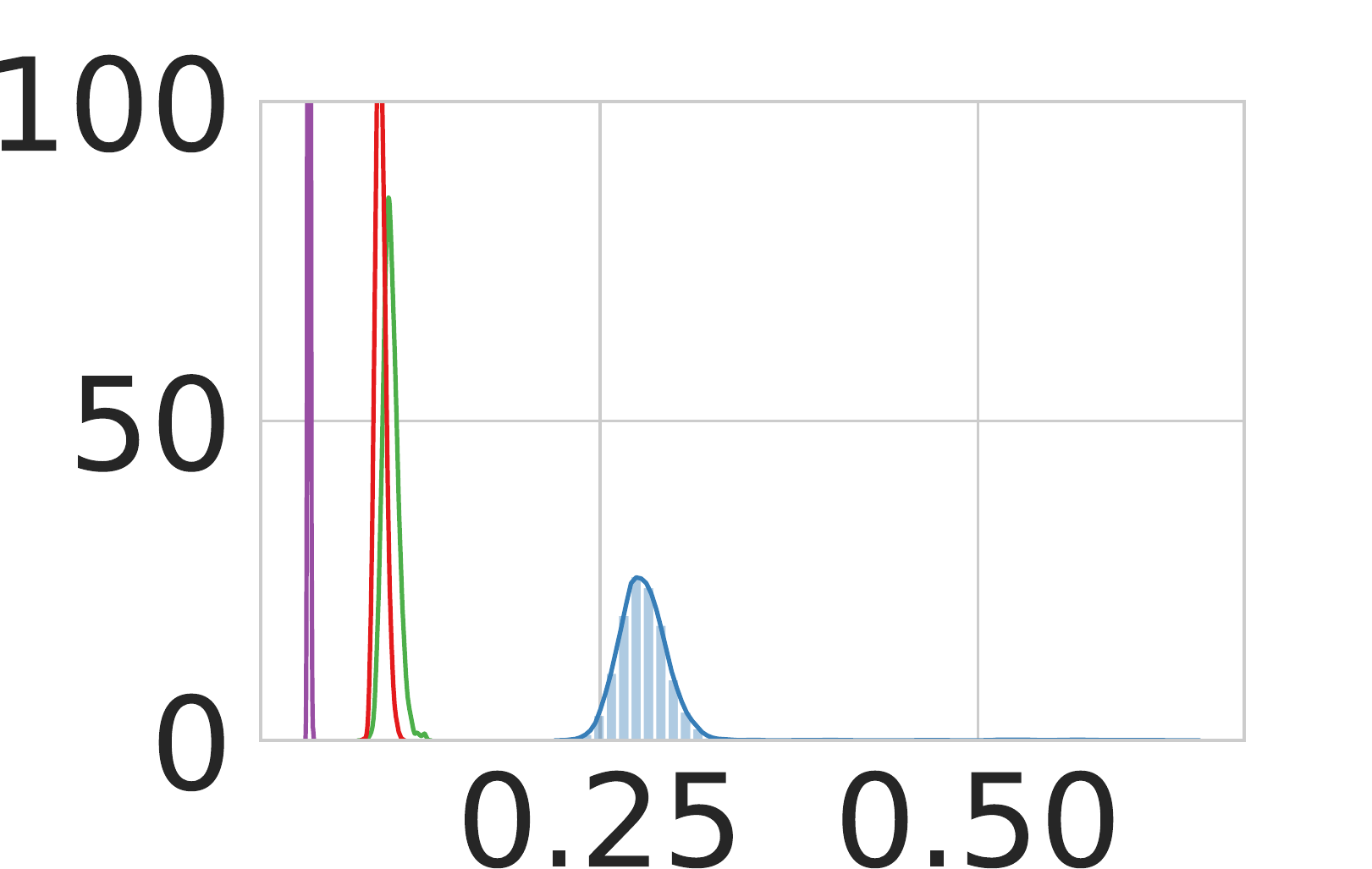\caption*{Third ResNet block}
        \end{subfigure}
        \begin{subfigure}[c]{0.3\linewidth}
          \centering
          \def\svgwidth{0.99\columnwidth}
          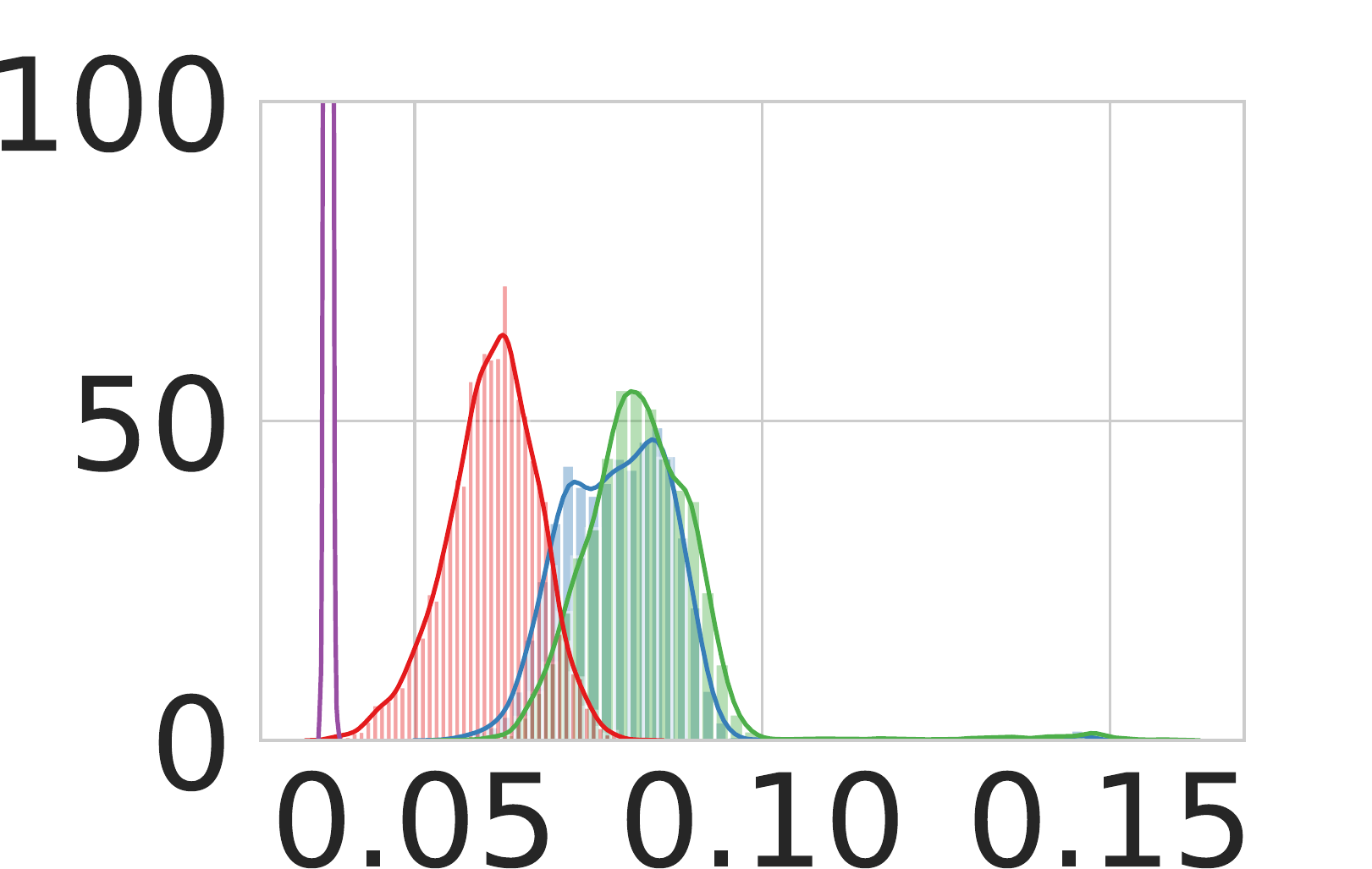\caption*{Fourth ResNet block}
        \end{subfigure}
        \caption{Layer cushion of ResNet layers. (Right is better) }
        \label{fig:Arora}
  \end{subfigure}\vspace{10pt}
  \begin{subfigure}[c]{0.99\linewidth}
    \centering
    \def\svgwidth{0.99\columnwidth}
    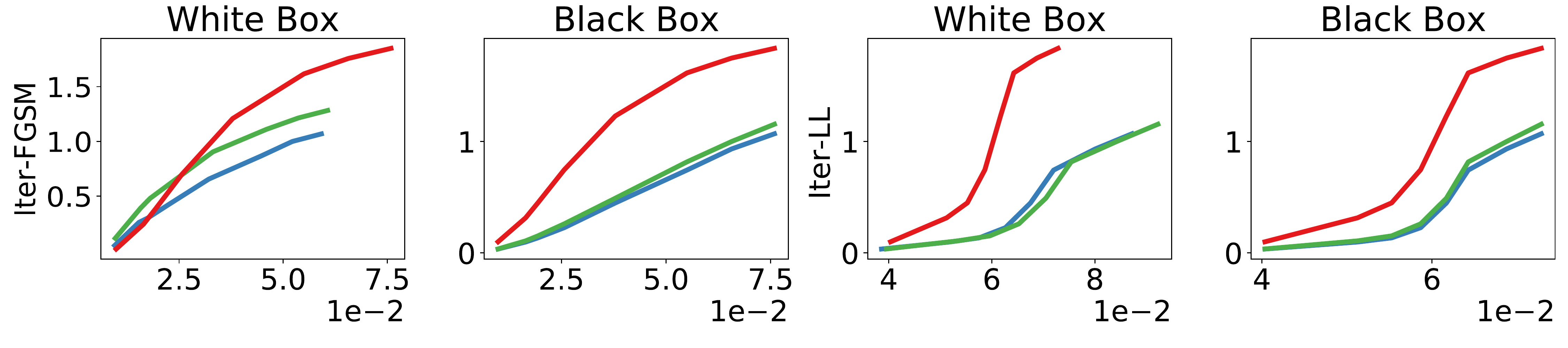
    \caption{Adversarial Perturbation in Input~(x-axis) and Representation~(y-axis) Space for CIFAR10. (Lower is better)}\label{fig:perturbation_spaces}
  \end{subfigure}
  \begin{subfigure}[c]{0.99\linewidth} 
    \centering
  \def\svgwidth{0.99\columnwidth}
  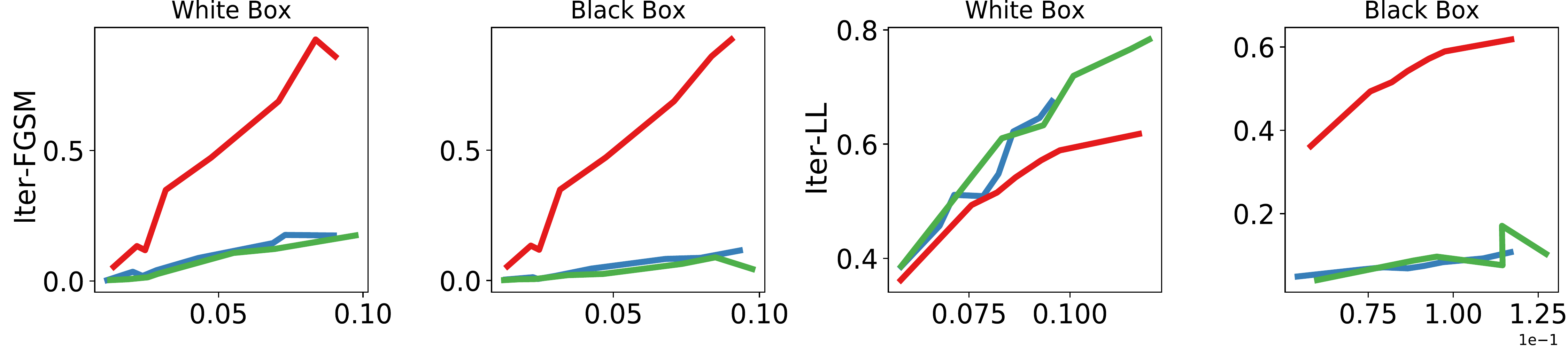
  \caption{Adversarial Perturbation in Input~(x-axis) and Representation~(y-axis) Space for SVHN. (Lower is better)}\label{fig:perturbation_spaces_svhn}
  \end{subfigure}
  \caption[Noise-sensitivity of ResNet-18 on CIFAR10 and SVHN]{ Layer cushion of
  the last fully connected layer, last convolution layer of the third residual
  block, and last convolution layer of the fourth residual blocks are plotted,
  respectively, in~\Cref{fig:Arora}. Noise-sensitivity of ResNet-18 trained on
  is showed via plotting the propagation of perturbation from the input to the
  representation space for CIFAR10 in~(\Cref{fig:perturbation_spaces}) and SVHN in~\Cref{fig:perturbation_spaces_svhn}.}
  \label{fig:noise_sensitivity_cifar}
\end{figure}

\paragraph{Layer cushion}

~\citet{arora18b} gives empirical
evidence that deep networks are stable towards injected Gaussian noise
and use a variation of this noise stability property to derive more
realistic generalisation bounds. They capture the noise-sensitivity through a term called layer cushion, which we define in~\Cref{defn:lyr-cushion}. It is measured for each example in the training dataset.
The higher the value of the layer cushion, the better is the generalisation
ability of the model.~\Cref{fig:Arora} shows histogram plots of the distribution
of this term for the examples in CIFAR10 for four models---2-LR, 1-LR, N-LR and
a {\em random network} (randomly initialised, no training done) with the same
architecture.
As~\Cref{fig:Arora} shows, the histograms of the
LR models are to the right of the N-LR model, which is further to the
right of the randomly initialised network thus indicating that LR
models have the highest layer cushion. %

\subsection{Discriminative properties of embeddings}
Experiments in the previous section showed that our algorithm induces a low-rank
structure in the activation space and that these low-rank activations are
significantly more stable to input perturbations. Here we inspect the
discriminative power of these embeddings.

\paragraph{Hybrid Max-Margin models:} For the experiments in this section, we
will use modified versions of the original models which we will refer to as {\em
hybrid max-margin models}. To convert a base model to a hybrid model, learned
representations are first generated using the original trained base model on the
training dataset. Then a max-margin classifier~(such as SVM) is trained on these
learned representations to classify the original label. In some of the
experiments, before training the max-margin classifier, the learned
representations are projected onto a low dimensional space by performing PCA on
this set of learned representations to classify the original label. This
dimension of projection will be referred to as {\em dim} when the results are
presented. At test time, the original trained model is used to first obtain a
representation, if necessary, and then projected using the learnt PCA projection
matrix, and is then classified using the learnt max-margin linear classifier.
The linear max-margin model is trained using SGD with hinge loss and $0.01$
$L_2$ regularisation coefficient. The learning rate is decreased per iteration
as $\eta_t = \frac{\eta_0}{\br{1+\alpha t}}$ where $\eta_0$ and $\alpha$ are set
by certain
heuristics~\footnote{\url{https://cilvr.cs.nyu.edu/diglib/lsml/bottou-sgd-tricks-2012.pdf}}.

\begin{table}\centering
  \begin{tabular}{l@{\quad}C{1.5cm}C{1.5cm}}
    \toprule
    &\multicolumn{2}{c}{ Accuracy($\%$)}\\ %
    Models& Coarse & Fine \\\hline%
    R50 1-LR & $\mathbf{78.1}$  & $48$ \\%\hline%
    R50 N-LR & $75.6$ & $\mathbf{52}$  \\%\hline%
    R50 Bottle-1LR & $76$ & $38$  \\\hline%
  \end{tabular}
  \caption{Transfer learning on CIFAR-100 using LR and NLR models}
  \label{tab:test-acc-transfer}
  \end{table}
\paragraph{Transfer learning}

As the intrinsic dimensionality of the representations are decreased, it is
plausible that the representations only store the most necessary information to
solve the task at hand and ignore all other information from the input. If that
were true, then representations learned for one task cannot be successfully used
for another related but slightly different task. To test whether the learned
representations can be used in a different task, we conduct a transfer learning
exercise where embeddings generated from a ResNet-50 model~(after the fourth
resnet block), trained on the coarse labels of CIFAR-100, are used to predict
the fine labels of CIFAR-100. A set of ResNet-50 hybrid max-margin classifiers
is trained for this purpose using these 2048 dimensional embeddings.

 The accuracy of the hybrid classifiers is reported in
Table~\ref{tab:test-acc-transfer}. The results of the same experiment, when the
hybrid model is trained on the coarse labels are reported in the second column
of Table~\ref{tab:test-acc-transfer}. Surprisingly, the low-rank model performs
well at all in this experiment as one would expect that all information in the
representations that are not strictly required in the classification of the
original task are discarded from the model.

Table~\ref{tab:test-acc-transfer} shows that the LR model suffers a small loss
of $4\%$ in accuracy as compared to the N-LR model when its embeddings are used
to train a max-margin classifier for predicting fine labels. It should be noted
that the accuracy of the LR model actually increases when the max-margin
classifier is trained to perform the original task, i.e. classifying the coarse
labels. On the other hand, the Bottle-LR model suffers a loss of $14\%$ in
accuracy compared to the N-LR model and shows no significant advantage in the
original task either.

\begin{figure}[t]%
  \begin{center}
\includegraphics[width=0.3\linewidth]{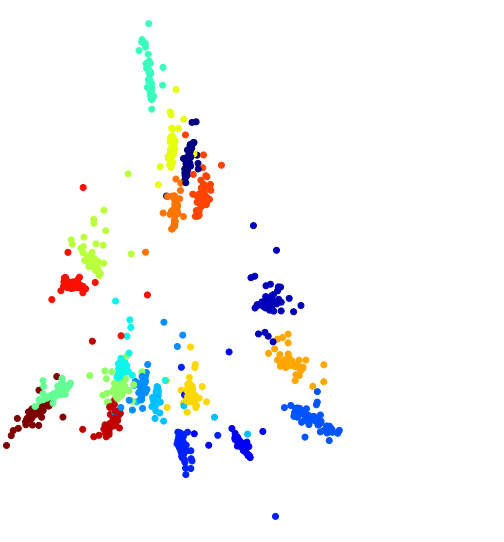}\hspace{20pt}
\includegraphics[width=0.3\linewidth]{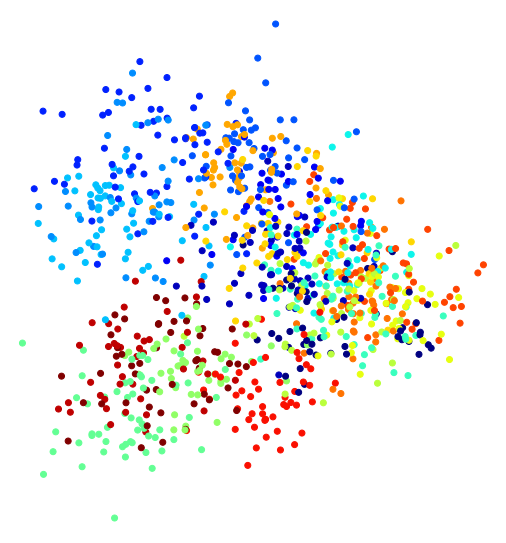}
    \end{center}
  		\caption[PCA Projections of representations for LR and NLR
models]{2-D PCA projection of representations from ResNet50 trained on
coarse labels of CIFAR 100 with~(left) and without~(right) low-rank
constraints, coloured according to the original 20  coarse labels.}
  \label{fig:coarse_lbl_PCA}
\end{figure}

\paragraph{Clusters of low dimensional embeddings}
\label{sec:clust-low-dimens}

Figure \ref{fig:coarse_lbl_PCA} shows the two-dimensional projections of the
2048 dimensional embeddings obtained from ResNet-50-LR and ResNet-50-N-LR. The
colouring is done according to the coarse labels of the input. We can see that
the clusters are more separable in the case of the model with \lr than the model
without, which gives some insight into why a hybrid max-margin classifier might
perform better for the LR model than the N-LR model. Thus, the representations
of the low-rank model are more discriminative in the sense that for the low-rank
representations there are low dimensional linear classifiers that can classify
the dataset with a higher margin than the vanilla models.

\subsection{Compression of model and embeddings}
\label{sec:comp}
 Further, due to the low
dimensionality of the representation space, these learned representations can
be compressed without losing their discriminative power.  Among other things,
we show that low dimensional projections of our embeddings, {\em with a size of
less than $2\%$ of the original embeddings}, can be used for classification
with significantly higher accuracy than similar-sized projections of
embeddings from a model trained without our training
modification~(N-LR) on the CIFAR100 dataset.

\paragraph{Representation compression:} 
In the first experiment, reported in Table~\ref{tbl:low_dim_emb_pred},
we trained two ResNet-50 hybrid max-margin models -with and without the \lr
respectively- on the 20 super-classes of CIFAR-100. As our objective
here is to see if the embeddings and their low dimensional projections
could be effectively used for discriminative tasks, we used PCA, with
standard pre-processing of scaling the input, to project the
embeddings onto a low dimensional space and then trained a linear
maximum margin classifier on it.

Table \ref{tbl:low_dim_emb_pred} shows
that even with sharply decreasing embedding dimension, the hybrid model trained
using the LR preserves the accuracy significantly more so than N-LR. \emph{Even
with a $5$-dimensional embedding~(400x compression), the LR model loses
only $6\%$ in accuracy, but the N-LR model loses
$27\%$}. In~\Cref{tab:max_margin_adv}, we show that compressed
representations are also more robust to adversarial attacks.

\begin{table}[!htb]
  \begin{subtable}[t]{0.46\linewidth}
    \centering
    
  \begin{tabular}{l@{\quad}c@{\quad}c@{}}\toprule
  Model & Dim & Accuracy($\%$) \\ \toprule 
  R50 1-LR & $2k$  & $\mathbf{78.1}$ \\ 
  R50 N-LR & $2k$  & $75.6$ \\ \midrule
  R50 1-LR & \(10\)  & $\mathbf{76.5}$ \\ 
  R50 N-LR & \(10\) &  $68.4$ \\ \midrule
  R50 1-LR & \(5\) & $\mathbf{72}$ \\ 
  R50 N-LR & \(5\) & $48$ \\ \bottomrule
  \end{tabular}
  \caption{Representation before FC layer, trained on
    CIFAR-100. Original dimension is $2k$.} 
  \label{tbl:low_dim_emb_pred}
\end{subtable}\hfill
  \begin{subtable}[t]{0.50\linewidth}
\centering
\begin{tabular}{l@{\quad}c@{\quad}c@{}}
    \toprule
    Model &Dim & Accuracy($\%$)\\ \toprule
    R18 2-LR & $16k$ & $\mathbf{91.14}$ \\
    R18 N-LR & $16k$ & $90.7$ \\\midrule
    R18 2-LR & $20$ &$ \mathbf{88.5}$  \\
    R18 N-LR & $20$ & $76.9$   \\\midrule
    R18 2-LR & $10$ & $\mathbf{75}$ \\
    R18 N-LR & $10$ & $61.7$\\\bottomrule
  \end{tabular}
  \caption{Representation before last ResNet block, trained
    on CIFAR10. Original dimension is $16k$.} 
  \label{tbl:low_dim_lyr_rem}
\end{subtable}\hspace{10pt}
\caption[Accuracy preserved under compression of representations]{\textit{Dim} is the
  size of the \textit{compressed embedding} on which linear
  classifier is trained. }\label{tbl:low_dim_proj_emb}
\end{table}

\paragraph{Model compression:} A consequence of forcing the activations of
the $\ell^{th}$ layer of the model to lie in a low dimensional subspace
with minimal reconstruction error, is that a simpler model can 
replace the latter parts of the original model without significant
reduction in accuracy. Essentially, if we train the hybrid max-margin
classifier on the representations obtained from after the third ResNet
block, we can replace the entire fourth ResNet block and the last FC
layer~(with 8.4M parameters) with a small linear classifier with only $0.02$ times the number of parameters.

In this experiment, we trained two ResNet-18 hybrid max-margin classifiers- with
and without the \lr respectively- on CIFAR-10. The representations were obtained
from before the fourth ResNet block and had a dimension of 16,384. 
 This yields a significant reduction in model size at the cost of a very slight
drop in accuracy~($< 1\%$). The second benefit is that as the low dimensional
embeddings still retain most of the \emph{discriminative} information, the
inputs fed to the linear model also have a small number of features.

\subsection{Adversarial robustness of maximum margin model}
\label{sec:advers-attack-maxim}

\renewcommand{\arraystretch}{1.1}

\begin{table}[!htb]\small\centering
 \begin{tabular}[h!]{l@{\quad}c@{\quad}c@{\quad}c@{\quad}c@{}}
   \toprule
    &R18  &DFL  &ILL & IFGSM \\\hline
  \multirow{3}{1cm}{White Box}&2-LR&$\mathbf{0.43}$&$\mathbf{0.55}$&$\mathbf{0.55}$\\%\cline{2-5}
   & 1-LR&$0.38$&$0.35$&$0.48$\\%\cline{2-5}
   & N-LR&$0.01$&$0.04$&$0.02$\\\hline
   \multirow{2}{1cm}{Black Box}&  2-LR&$\mathbf{0.44}$&$\mathbf{0.50}$&$\mathbf{0.48}$\\%\cline{2-5}
   &1-LR&$0.29$&$0.31$&$0.33$\\\hline
 \end{tabular}
 \caption[Adversarial accuracy of hybrid max-margin classifier]{Accuracy of classification of adversarial examples, constructed by attacking ResNet18, by ResNet18 Max-Margin Classifiers.}
 \label{tab:max_margin_adv}
\end{table}

\paragraph{Robustness of Max-Margin Classifiers} Finally, we show that the
features learned by our models are inherently more linearly discriminative i.e.
there exists a linear classifier that can be used to classify these features
with a wide margin. To this end, in Table~\ref{tab:max_margin_adv}, we show that
for LR models, the max-margin hybrid models are significantly more robust to
adversarial attacks than the corresponding original models. Hybrid max-margin
models with \lrs are particularly more robust than hybrid max-margin models
without \lrs against adversarial attacks. Specifically,  as seen in
Table~\ref{tab:max_margin_adv}, \emph{a hybrid model with an \lr correctly
classifies $50\%$ of the examples that had  fooled the original classifier while
for a similar amount of noise, an N-LR hybrid model has negligible accuracy}.

 We train a variety of hybrid max-margin models  with ResNet18-1-LR,
ResNet18-2-LR, and ResNet18-N-LR along with black box versions of the same. Then
we generate adversarial examples for all three attacks (both black box and white
box) on the trained ResNet models~(not the hybrid models). Then we use these
adversarial examples to attack the corresponding max-margin models

To perform a fair comparison with the hybrid ResNet18-N-LR, it is essential
to add a similar amount of noise to generate the examples for the hybrid
ResNet18-N-LR as is added to hybrid ResNet18-1-LR. The adversarial
examples are hence generated by obtaining the gradient using
ResNet18-N-LR but stopping the iteration only when the adversarial
example could fool ResNet18-1-LR.

\section{Alternative algorithms for low-rank representations}
\label{sec:alt_algs}

In this section, we discuss a few different alternative algorithms
that could be potential alternatives for obtaining low-rank representations.

\subsection*{Low-rank weights} With respect to compression, it is natural to
look at low-rank approximations of network
parameters~\citep{Denton2014,jaderberg2014}. By factorizing the weight matrix
$\vec{W}$ as the product of a wide and a tall matrix, we can get low-rank
\emph{pre-activations}. This however does not lead to low-rank
\emph{activations} as demonstrated both mathematically (by the counter-example
below) and empirically.

\paragraph{Mathematical counter-example}: Consider a rank 1
\emph{pre-activation} matrix $\vec{A}$ and its corresponding
\emph{post-activation}(ReLU) matrix as below. It is easy to see that
the rank of \emph{post-activation} has increased to $2$. \[ \vec{A}
= \begin{bmatrix} &1 &-1 &1\\ &-1 &1
&-1\end{bmatrix}\qquad\text{Relu}(\vec{A}) = \begin{bmatrix} &1 &0 &1\\ &0
&1 &0\end{bmatrix} \]

\textbf{Empirical Result}: To see if techniques for low-rank approximation of
network parameters like ~\citet{Denton2014} would have produced low-rank
activations, we experimented by explicitly making the \emph{pre-activations}
low-rank using SVD. Our experiments showed that despite setting a rank of $100$
to the \emph{pre-activation} matrix, the \emph{post-activation} matrix had full
rank. Though all but the first hundred singular values of the
\emph{pre-activation} matrix were set to zero, the \emph{post-activation}
matrix’s \(101^{\it{st}}\) and $1000^\th$ singular values were $49$ and $7.9$
respectively, and its first $100$ singular values explained only $94\%$ of the
variance.

Theoretically, a bounded activation function lowers the Frobenius norm
of the \emph{pre-activation} matrix i.e. the sum of the squared
singular values. However, it also causes a smoothening of the singular
values by making certain zero singular values non-zero to compensate for
the significant decrease in the larger singular values. This leads to
an increase in rank of the \emph{post-activation} matrix.

In Table~\ref{tab:adv-pert-compre}, we compare against
SRN~\citep{sanyal2020stable} a simple algorithm for reducing the stable rank~(a
softer version of rank) of linear layers in neural networks and observe that the
increase in adversarial robustness is not as high as provided by LR models.

\subsection*{Nuclear norm}

Nuclear norm regularisation is a convex relaxation to the hard rank
regularisation approach that we have adopted. Nuclear norm can be
regularised by adding the nuclear norm of $\vec{A}_\ell$ to the loss
function, i.e. minimizing
$\cL\br{\vec{X},\vec{Y};\theta,\phi}+\lambda\norm{\vec{A}_\ell}_\ast$.

However, there are a few problems with it. First, it runs into the
same problem as the hard rank minimisation - a) unfeasible to optimize for the large whole
activation matrix, b) unclear why batch-wise optimisation should
ensure low-rank for the entire dataset, and c) sensitivity towards the
hyper-parameter. Its sensitivity to hyper-parameter arises for reasons
similar to the hyper-parameter sensitivity of ridge-regression as
discussed at the end of~\Cref{sec:what-low-rank}.

We performed batch-wise nuclear norm minimisation for all \(\lambda\in \bc{1,
0.1, 0.01, 0.001, 0.0001}\) and observed that only 0.001 gave comparable
performance to LR. All other settings gave either a trivial test accuracy or
much worse adversarial robustness thus showing its extreme sensitivity to
hyper-parameters, unlike our method. 

\section{Additional figures and tables in appendix}

~\Cref{fig:cifar10-resnet18-lyr-cushion}
compares the layer cushion of the four different residual blocks of a ResNet18
on CIFAR10.  Note that only 2-LR shows an increased cushion in Layer 3 whereas
both 1-LR and 2-LR have higher cushions in all other layers.
Similarly,~\Cref{fig:svhn-resnet18-lyr-cushion}
plots the layer cushion of a ResNet-18 trained on SVHN.

\chapter{Improving Calibration via Focal Loss}
\label{chap:focal_loss}
In the previous chapters, we looked at vulnerabilities of deep neural networks
arising from poor generalisability~(see~\Cref{chap:stable_rank_main}) and
adversarial robustness~(see~\Cref{chap:causes_vul,chap:low_rank_main}) of neural
networks. Another common issue faced by many state-of-the-art neural networks is
that they are poorly calibrated. Neural networks, used for multi-class
classification, output a distribution over the label classes where each
probability value is supposed to indicate the likelihood of that class label
being the correct one. In many state-of-the-art neural networks, the probability
values they associate with the predicted class labels overestimate the
likelihoods of those class labels being correct in the real world. The
underlying cause is hypothesised to be that the high complexity of these
networks leaves them vulnerable to overfitting on the negative log-likelihood
(NLL) loss they conventionally use during training~\citep{Guo2017}. As discussed
in~\Cref{sec:prelim_calibration}, multiple solutions have been proposed to
tackle this problem including the popularly used {\em Temperature Scaling}.

In this chapter, we propose a technique for improving network calibration that
works by replacing the cross-entropy loss conventionally used when training
classification networks with the focal loss proposed by \cite{Lin2017}. We
observe that unlike cross-entropy, which minimises the KL divergence between the
predicted (softmax) distribution and the target distribution (one-hot encoding
in classification tasks) over classes, focal loss minimises a regularised KL
divergence between these two distributions i.e. focal loss minimises
the KL divergence whilst \emph{increasing the entropy} of the predicted
distribution over the class labels, thereby preventing the model from becoming
overconfident.

The intuition behind using focal loss is to weight the gradient updates during
training more towards samples for which it is currently predicting a low
probability for the correct class. This helps to avoid reducing the NLL on
samples where it is already predicting a high probability for the correct class
as that is liable to lead to NLL overfitting and thereby miscalibration
\citep{Guo2017}. We show in \Cref{sec:focalloss} that focal loss
can be seen as \emph{implicitly} regularising the weights of the network during
training by causing the gradient norms for confident samples to be lower than
they would have been with the cross-entropy loss, which we would expect to
reduce overfitting and improve the network's calibration.

\section{Main contributions}
Overall, we make the following contributions:
\begin{enumerate}
\item We  study the link that~\citet{Guo2017} observed between miscalibration
and NLL overfitting; show that NLL overfitting is associated with the
predicted distributions for misclassified test samples becoming peakier as the
learning algorithm increases the magnitude of the network's weights to reduce
the training NLL.
\item  We propose the use of focal loss for training better-calibrated networks
and provide both theoretical and empirical justifications for our approach. In
addition, we provide a principled method for automatically choosing the
hyper-parameter $\gamma$ for each data point during training.
\item We show, via experiments on a variety of classification datasets and
network architectures, that DNNs trained with focal loss are more calibrated
than those trained with cross-entropy loss (both with and without label
smoothing), MMCE, and Brier loss. 
\end{enumerate}

\section{Understanding the cause of miscalibration}
\label{sec:cause_cali}

\paragraph{Problem Formulation}
Let $\cS_N = \bc{\br{\vec{x}_1,
y_1},\br{\vec{x}_2,y_2},\ldots,\br{\vec{x}_N,y_N}}$ denote a dataset consisting
of $N$ samples from a data distribution $\cD$ over $\cX\times\cY$, where for
each sample $i$, $\mathbf{x}_i \in \mathcal{X}$ is the input and $y_i \in
\mathcal{Y} = \{1, 2, ..., K\}$ is the target class label. Let $\hat{p}_{i,y} =
f_\theta\br{\vec{x}_i}\bs{y}$ be the probability that a neural network $f$ with
model parameters $\theta$ assigns to a class $y$ on a given input $\vec{x}_i$.
The class that $f$ predicts for $\vec{x}_i$ is computed as $\hat{y}_i =
\mathrm{argmax}_{y \in \mathcal{Y}} \; \hat{p}_{i,y}$, and the predicted
confidence is computed as $\hat{p}_i = \mathrm{max}_{y \in \mathcal{Y}} \;
\hat{p}_{i,y}$. The network is said to be \emph{perfectly~(weakly) calibrated}
when, for each sample $\br{\vec{x}_j, y_j} \in \cS_N$,  the confidence
$\hat{p}_j$ is equal to the model accuracy $\bP\bs{\cY = \hat{y}_j~\vert
f_\theta\br{\vec{x}_j}\bs{\hat{y}_j}=\hat{p}_j}$, i.e. the probability that the
predicted class is correct.  For instance, of all the samples to which a
perfectly calibrated neural network assigns a confidence of $0.8$, $80\%$ should
be correctly predicted.

We will use the various metrics described in~\Cref{sec:measuring-calibration} to
measure calibration in our experiments in this chapter. This includes Expected
Calibration Error~(ECE), Classwise Expected Calibration Error, Adaptive
ECE~(AdaECE), and reliability plots.

\begin{figure}[t]
	\centering
		\begin{subfigure}[c]{0.2\linewidth}
			\includegraphics[width=0.99\linewidth]{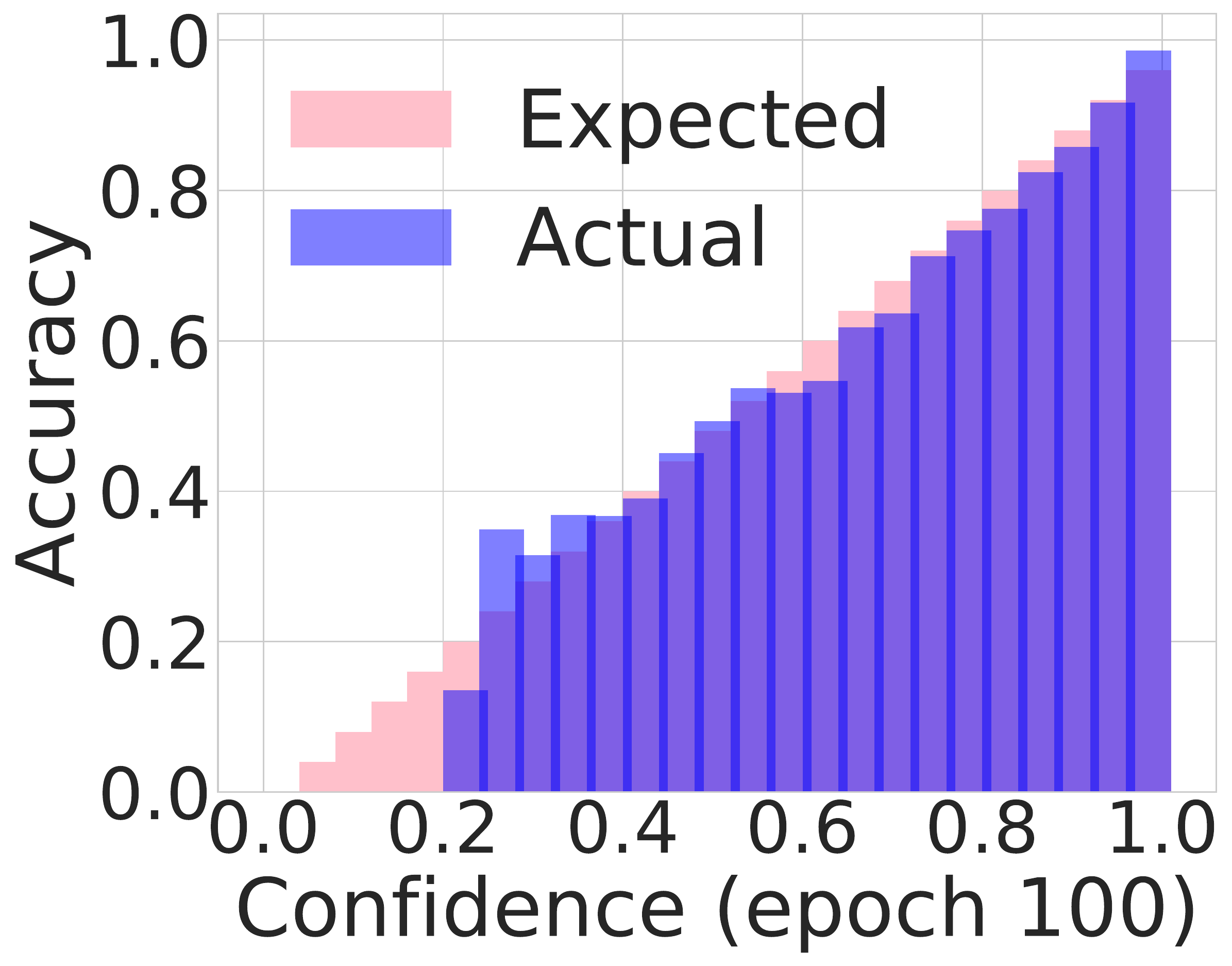}
		\end{subfigure}
		\begin{subfigure}[c]{0.2\linewidth}
			\includegraphics[width=0.99\linewidth]{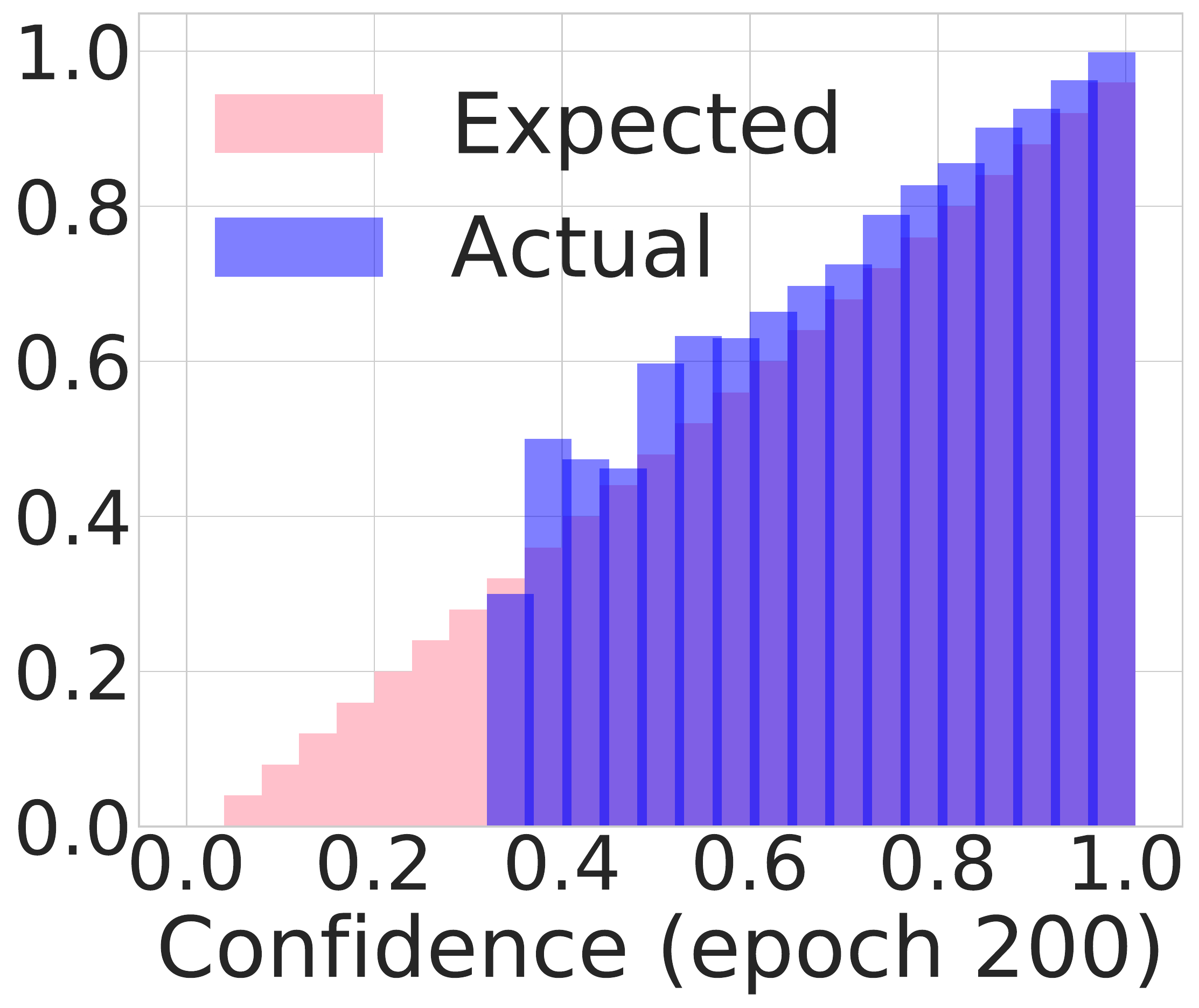}
		\end{subfigure}
		\begin{subfigure}[c]{0.2\linewidth}
			\includegraphics[width=0.99\linewidth]{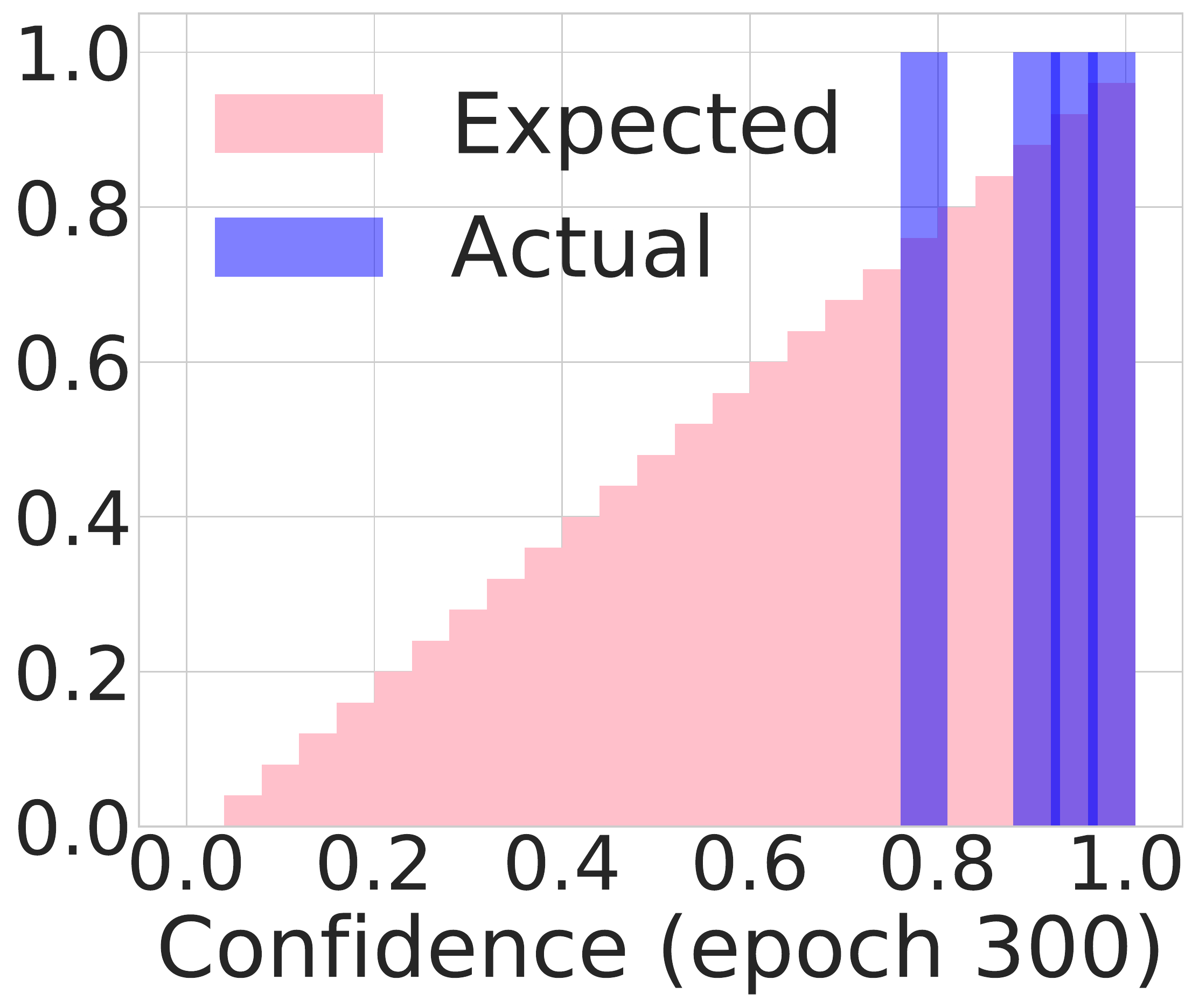}
		\end{subfigure}
		\begin{subfigure}[c]{0.2\linewidth}
			\includegraphics[width=0.99\linewidth]{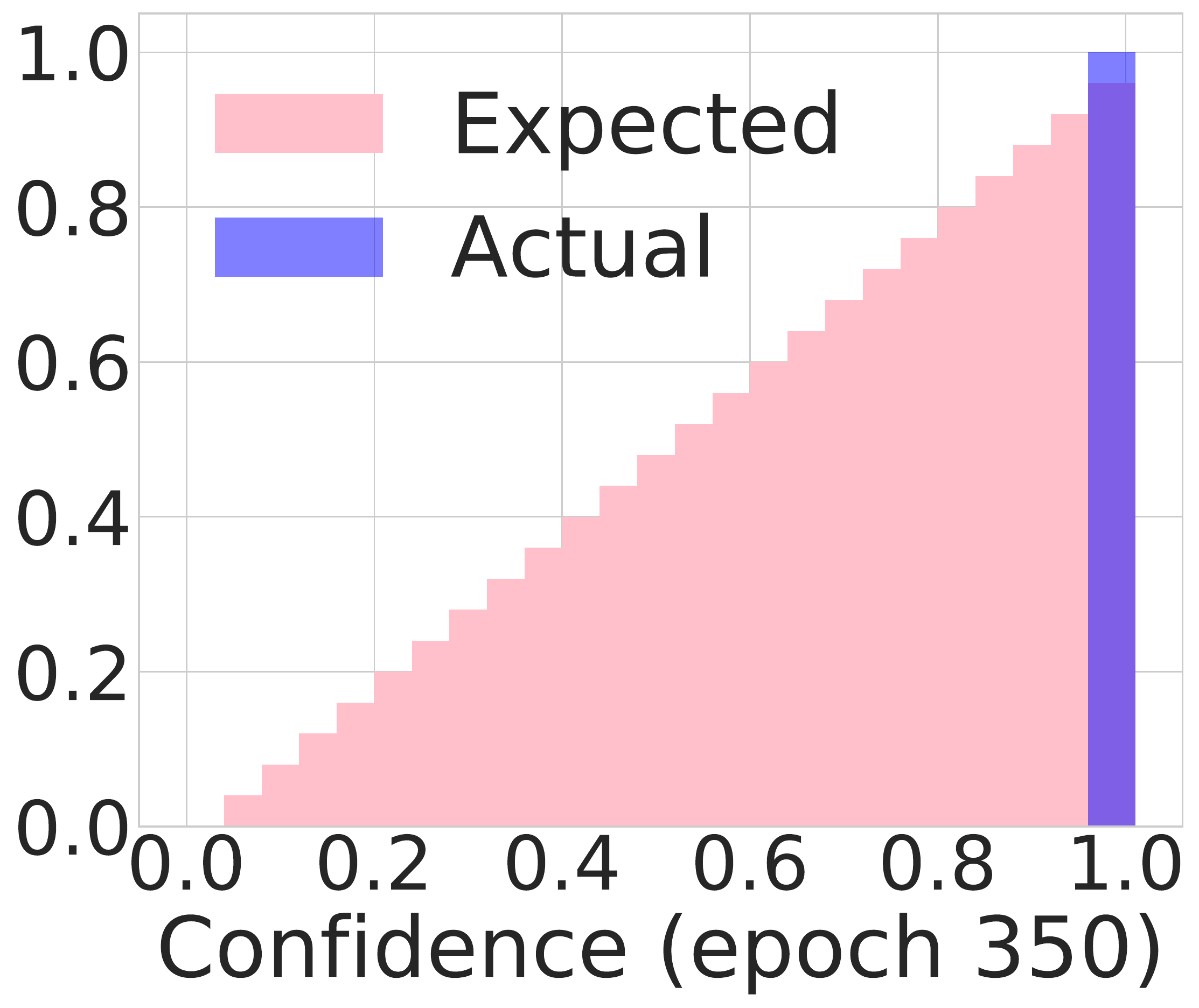}
		\end{subfigure}
		\begin{subfigure}[c]{0.2\linewidth}
			\includegraphics[width=0.99\linewidth]{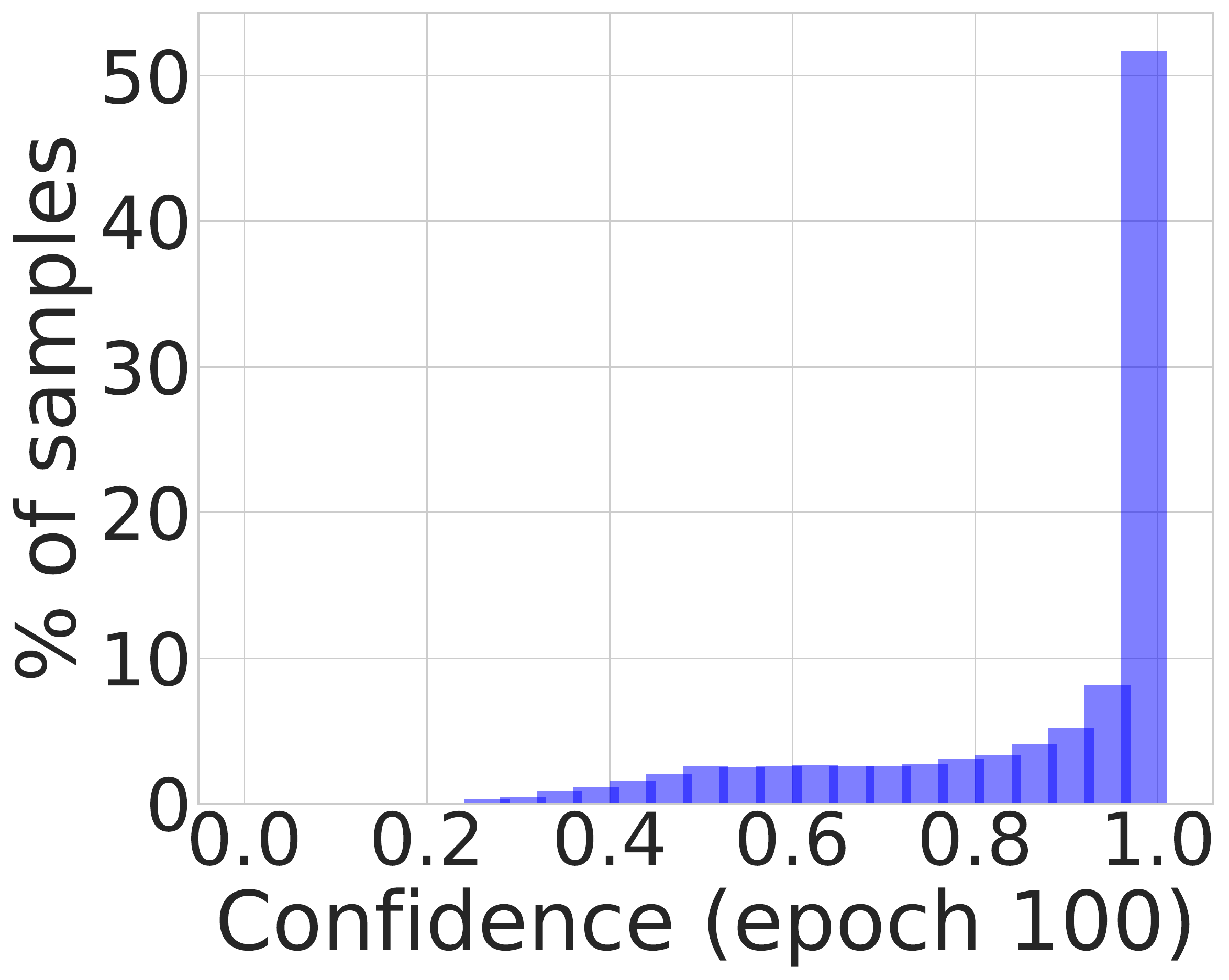}
		\end{subfigure}
		\begin{subfigure}[c]{0.2\linewidth}
			\includegraphics[width=0.99\linewidth]{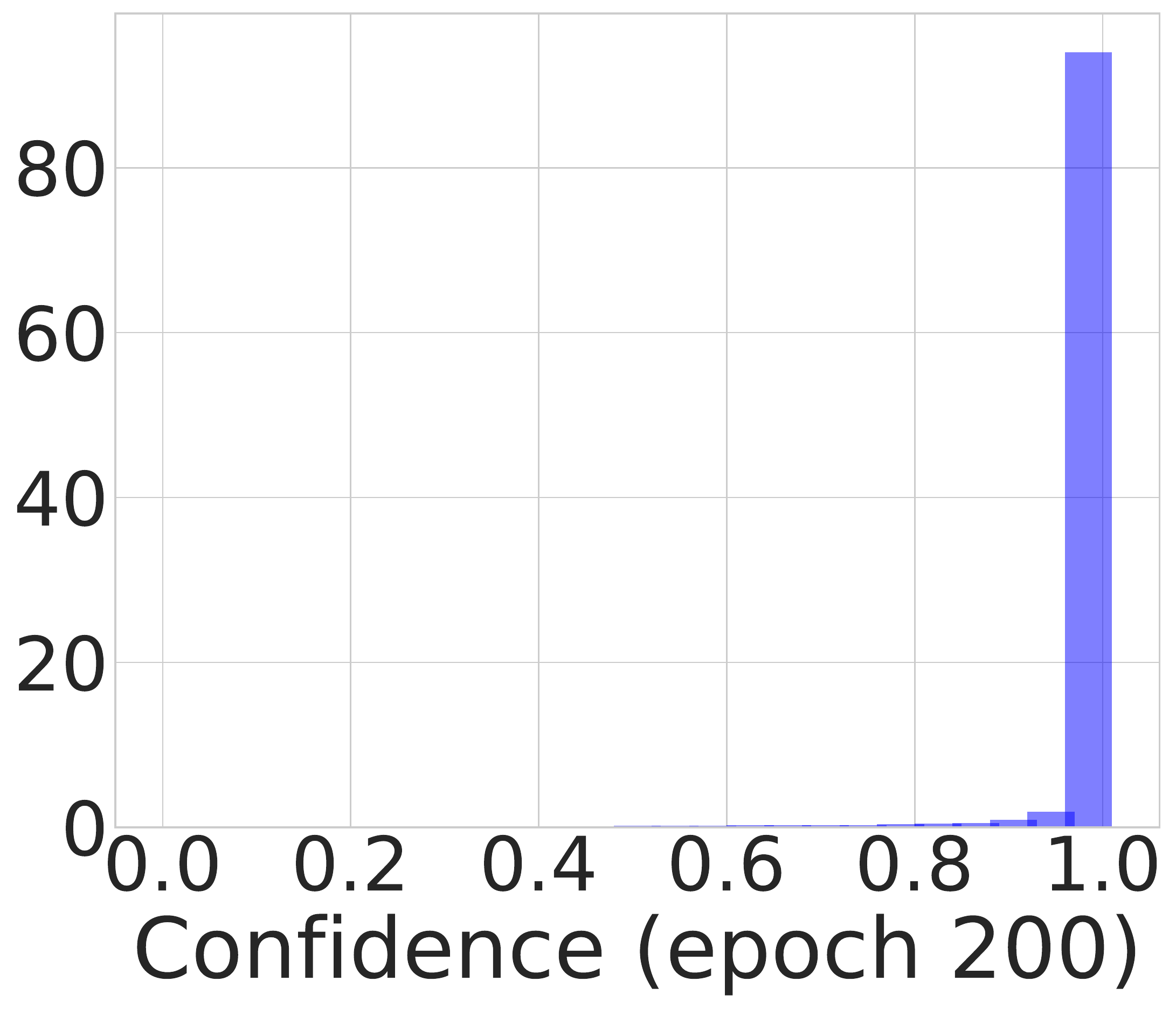}
		\end{subfigure}
		\begin{subfigure}[c]{0.2\linewidth}
			\includegraphics[width=0.99\linewidth]{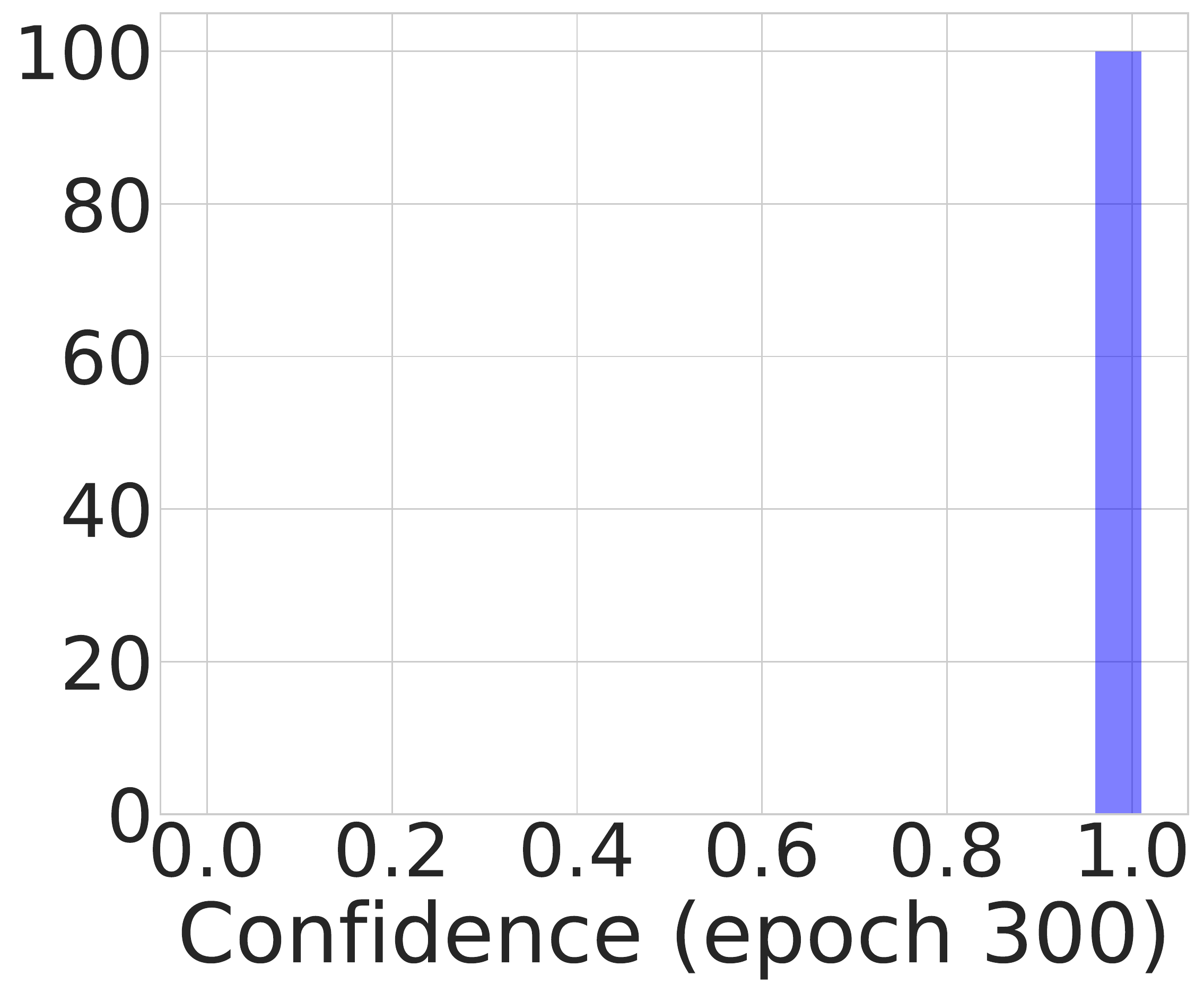}
		\end{subfigure}
		\begin{subfigure}[c]{0.2\linewidth}
			\includegraphics[width=0.99\linewidth]{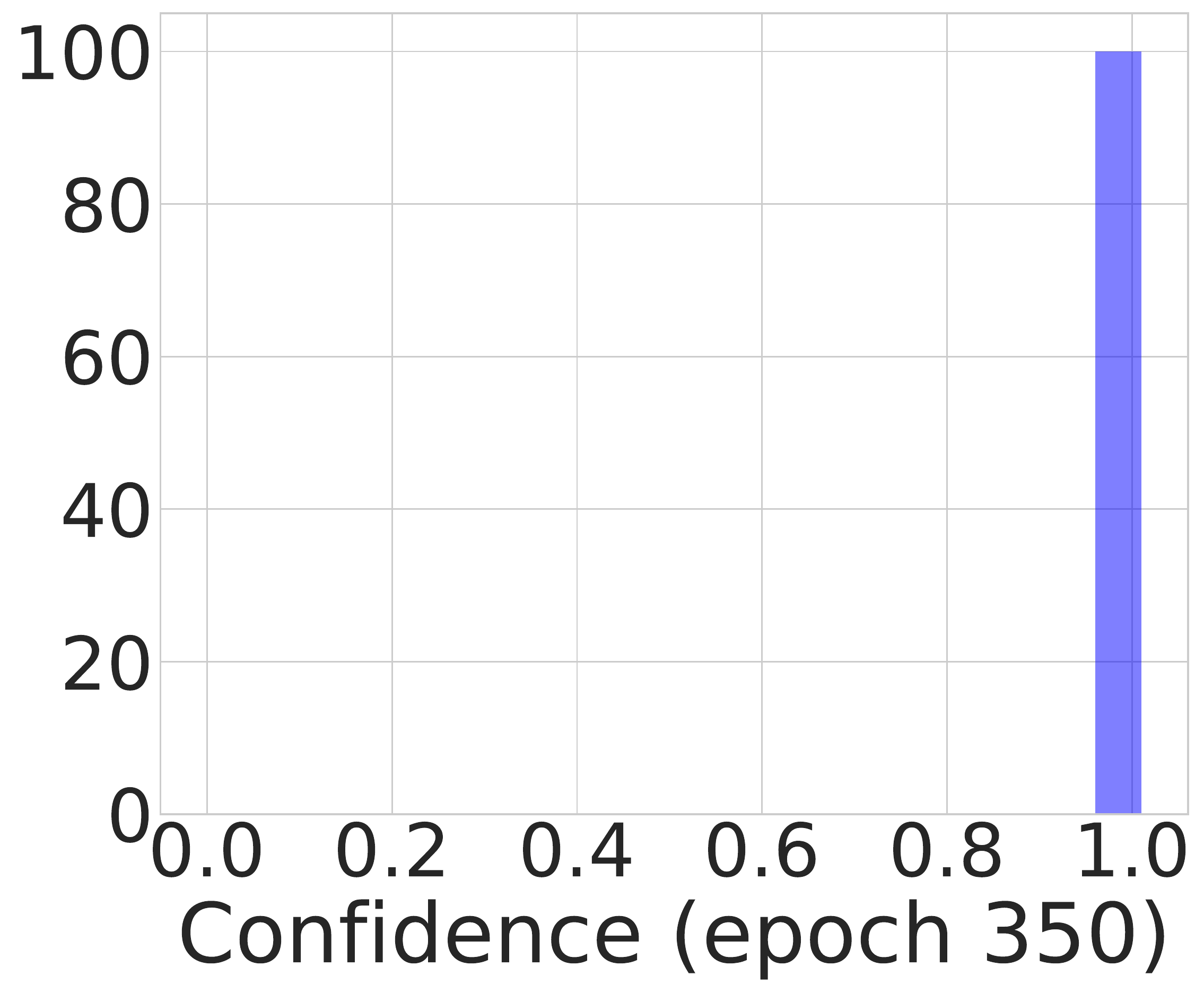}
		\end{subfigure}
	\caption[Confidence values for training samples during NLL training]{The confidence values for training samples at different
	epochs during the NLL training of a ResNet-50 on CIFAR-10 (see
	\Cref{sec:cause_cali}). Top row: reliability plots using $25$
	confidence bins; bottom row: \% of samples in each bin. As training
	progresses, the model gradually shifts all training samples to the
	highest confidence bin. Notably, it continues to do so even after
	achieving 100\% training accuracy by the $300$ epoch point.}
	\label{fig:rel_conf_bin_plot}
	\end{figure}

We now discuss why large networks, despite achieving low classification errors
on well-known datasets, tend to be miscalibrated. A key empirical observation
made by \cite{Guo2017} was that poor calibration of such networks appears to be
linked to overfitting on the negative log-likelihood (NLL) during training. In
this section, we discuss what we mean by NLL overfitting and further inspect how
NLL overfitting impacts calibration.

For the analysis, we train a ResNet-50 network on CIFAR-10 with state-of-the-art
performance settings as discussed in~\Cref{sec:expr-settings}.
We minimise cross-entropy loss (a.k.a.\ NLL) $\mathcal{L}_c$, which, in a
standard classification context, is $-\log \hat{p}_{i,y_i}$, where
$\hat{p}_{i,y_i}$ is the probability assigned by the network to the correct
class $y_i$ for the i$^{th}$ sample. Note that the NLL is minimised when for
each training sample $i$, $\hat{p}_{i,y_i} = 1$, whereas the classification
error is minimised when $\hat{p}_{i,y_i} > \hat{p}_{i,y}$ for all $y \neq y_i$.
This indicates that even when the classification error is $0$, the NLL can be
positive, and the optimisation algorithm can still try to reduce it to $0$ by
further increasing the value of $\hat{p}_{i,y_i}$ for each sample.

To empirically observe this, we divide the confidence range $\bs{0, 1}$ into 25
bins, and present reliability plots computed on the training set at training
epochs $100$, $200$, $300$ and $350$ (see the top row of
Figure~\ref{fig:rel_conf_bin_plot}). In Figure~\ref{fig:rel_conf_bin_plot}, we
also show the percentage of samples in each confidence bin. It is quite clear
from these plots that over time, the network gradually pushes all of the
training samples towards the highest confidence bin. Furthermore, even though
the network has achieved $100\%$ accuracy on the training set by epoch $300$, it
still pushes some of the samples lying in lower confidence bins to the highest
confidence bin by epoch $350$.

To study how miscalibration occurs during training, we plot the
average NLL for the train and test-sets at each training epoch in
Figures~\ref{fig:nll_entropy_ece}(a) and \ref{fig:nll_entropy_ece}(b).
We also plot the average NLL and the entropy of the softmax
distribution produced by the network for the correctly and incorrectly
classified samples. In Figure \ref{fig:nll_entropy_ece}(c), we plot
the classification errors on the train and test datasets, along with the
test-set ECE.

\begin{figure*}[!t]
	\centering
	\includegraphics[width=0.32\linewidth]{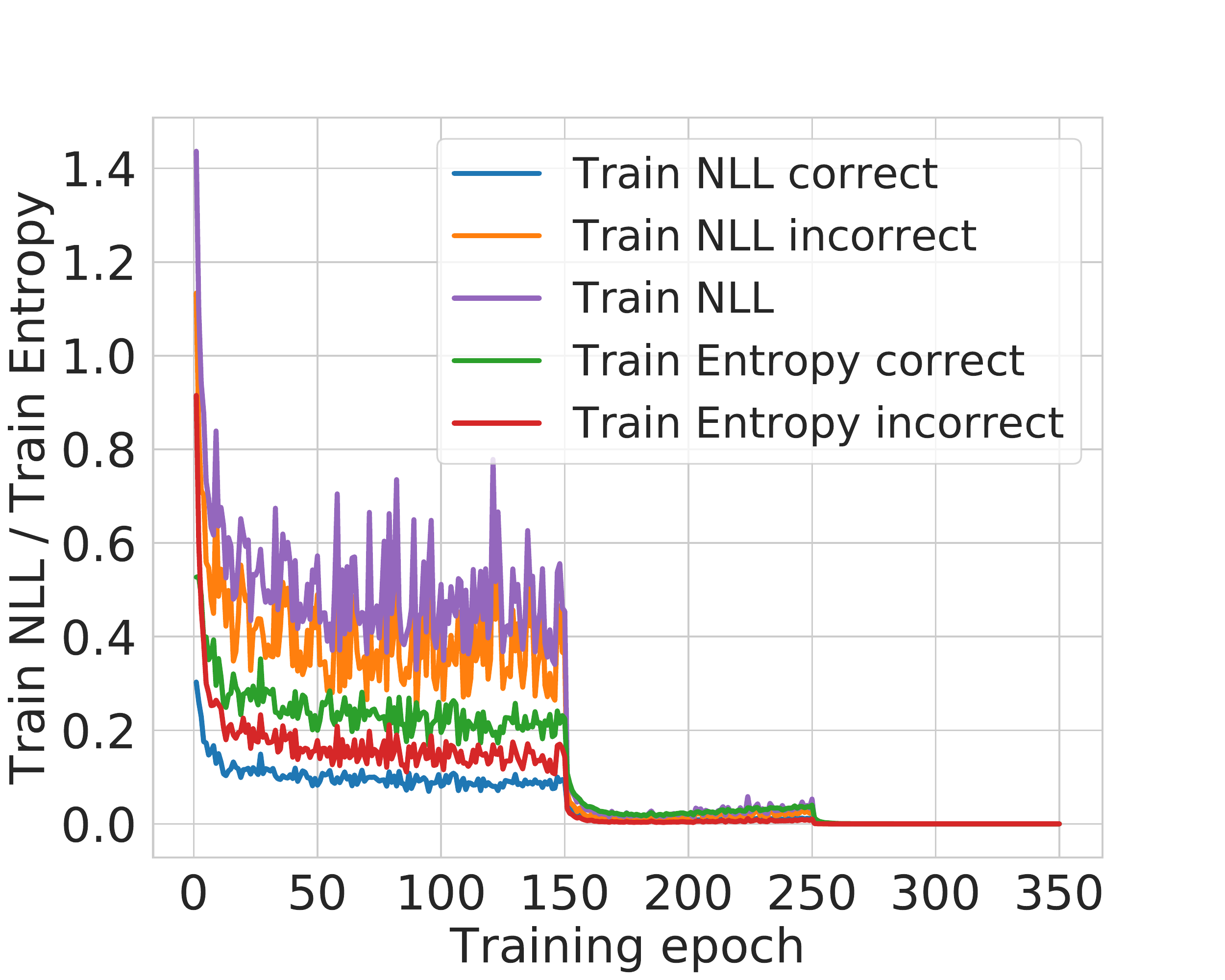}
	\includegraphics[width=0.32\linewidth]{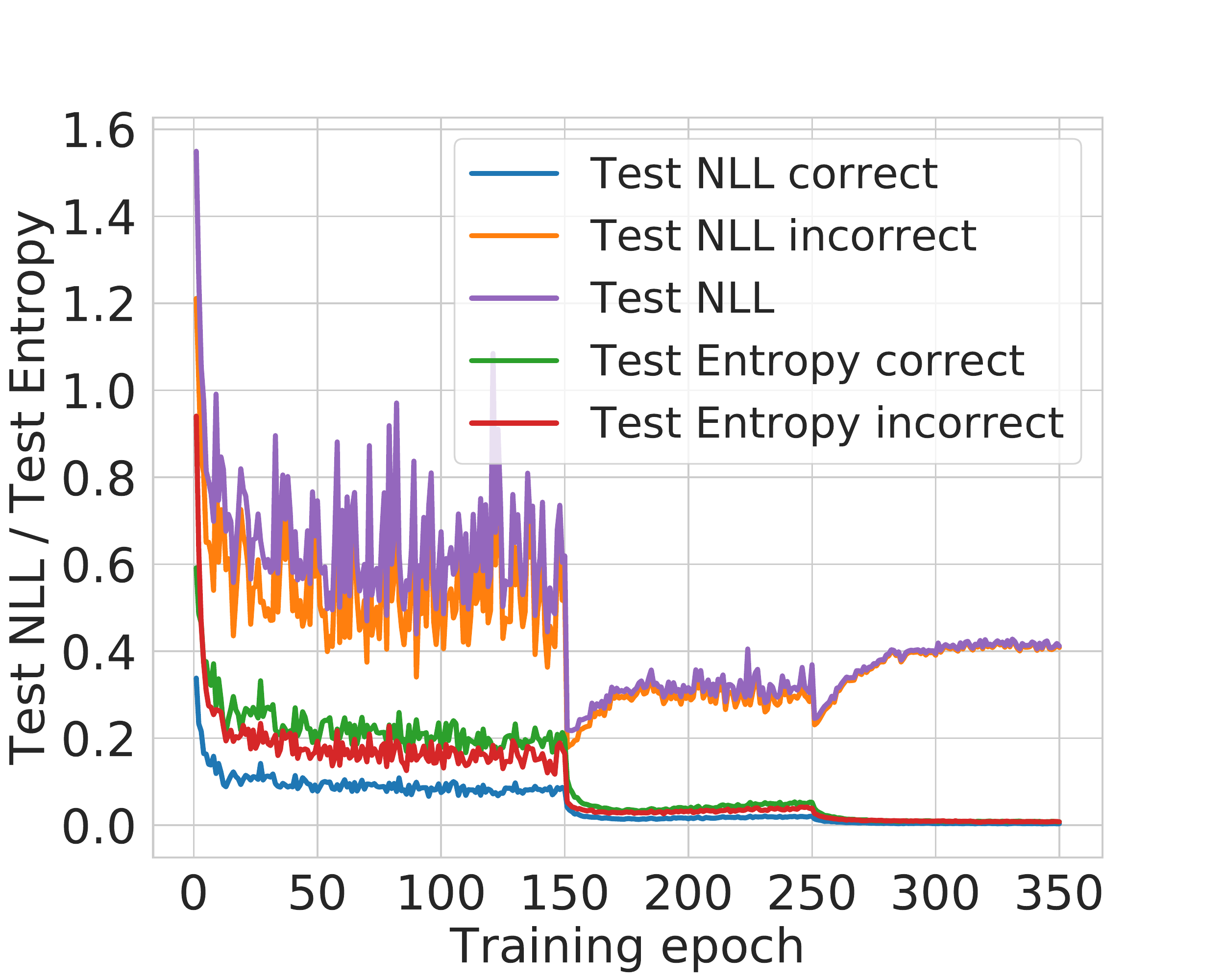}
	\includegraphics[width=0.32\linewidth]{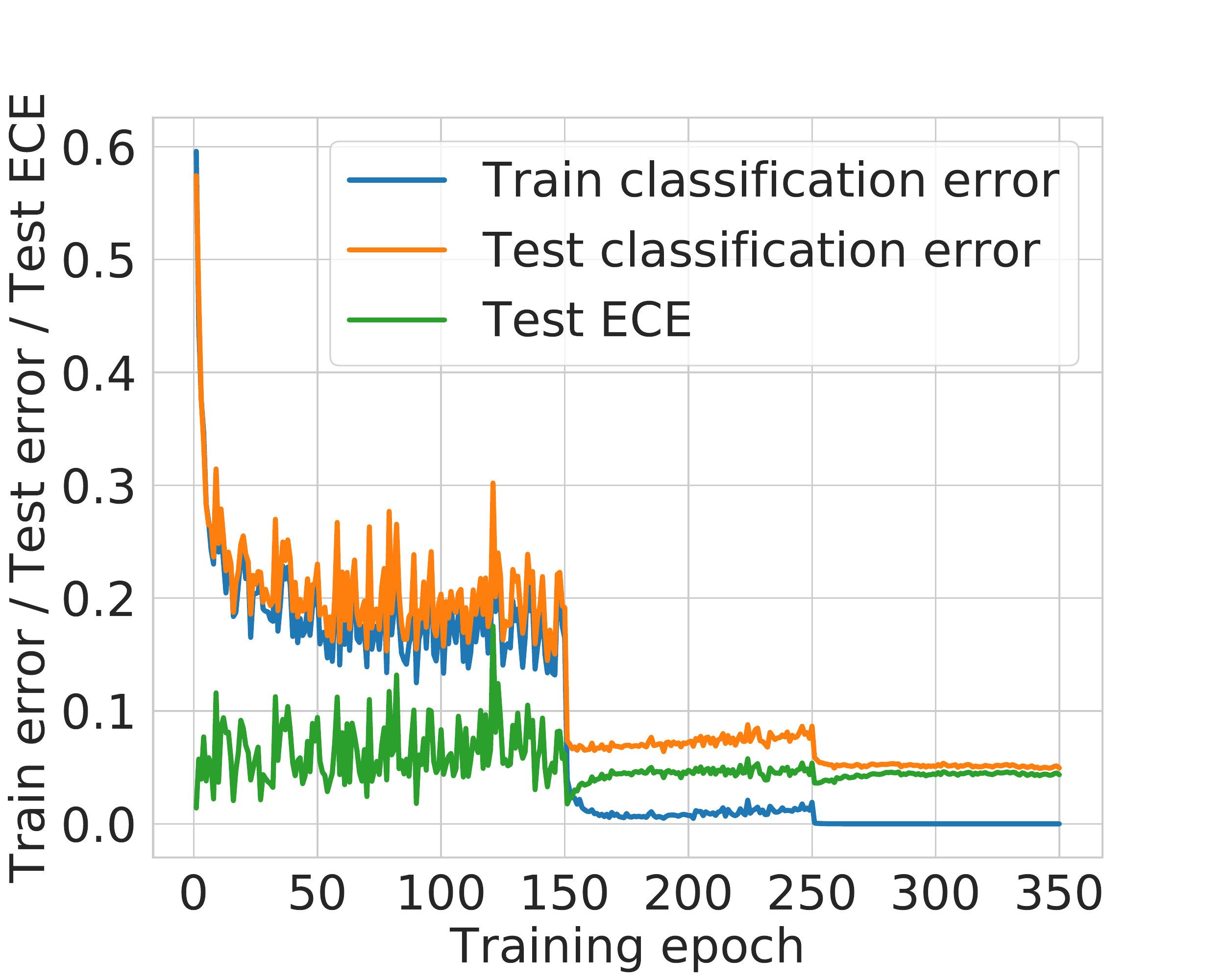}
	\caption[Calibration metrics during training of a ResNet-50 with NLL]{Metrics related to calibration plotted whilst training a ResNet-50 network on CIFAR-10.}
	\label{fig:nll_entropy_ece}
\end{figure*}

\textbf{Curse of misclassified samples:} Figures \ref{fig:nll_entropy_ece}(a)
and \ref{fig:nll_entropy_ece}(b) show that although the average train NLL (for
both correctly and incorrectly classified training samples) broadly decreases
throughout training, after the $150^{th}$ epoch (where the learning rate drops
by a factor of $10$), there is a marked rise in the average test NLL, indicating
that the network starts to overfit on average NLL. This increase in average test
NLL is caused only by the incorrectly classified samples, as the average NLL for
the correctly classified samples continues to decrease even after the $150^{th}$
epoch. This phenomenon is referred to as NLL overfitting. We also observe that
after epoch $150$, the test-set ECE rises, indicating that the network is
becoming miscalibrated. This corroborates the link between NLL overfitting and
miscalibration observation in \cite{Guo2017}.

\textbf{Peak at the wrong place:} We further observe that the
entropies of the softmax distributions for both the correctly and
incorrectly classified {\em test} samples decrease throughout training
(in other words, the distributions get peakier). This observation,
coupled with the one we made above, indicates that {\em for the
wrongly classified test samples, the network gradually becomes more
and more confident about its incorrect predictions}.

\textbf{Weight magnification:} The increase in confidence of the network's
predictions can happen if the network increases the norm of its weights $W$ to
increase the magnitudes of the logits. In fact, the cross-entropy loss is
minimised when for each training sample $i$, $\hat{p}_{i,y_i} = 1$, which is
possible only when $||W|| \to \infty$. Cross-entropy loss thus inherently
induces this tendency of weight magnification in neural network optimisation.
The promising performance of weight decay \citep{Guo2017} (regularising the norm
of weights) on the calibration of neural networks can perhaps be explained using
this. This increase in the network's confidence during training is one of the
key causes of miscalibration.

\section{Improving calibration using focal loss}
\label{sec:focalloss}

As discussed in the previous section, overfitting on NLL, which is observed as
the network grows more confident on all of its predictions irrespective of their
correctness, is strongly related to poor calibration. One cause of this is that
minimising the cross-entropy loss function minimises the difference between the
softmax distribution and the ground-truth one-hot encoding for all samples,
irrespective of how well a network classifies individual samples. In this
chapter, we study an alternative loss function, popularly known as \textit{focal
loss} \citep{Lin2017}, that tackles this by weighting loss components generated
from individual samples by how well the model classifies each of them. For
classification tasks where the target distribution is one-hot encoding, it is
defined as $\mathcal{L}_f = -(1 - \hat{p}_{i,y_i})^\gamma \log \hat{p}_{i,y_i}$,
where $\gamma$ is a user-defined hyperparameter.

\begin{remark}
	We note in passing that, unlike cross-entropy loss, focal loss in its general
	form is not a proper loss function, as minimising it does not always lead to
	the predicted distribution $\hat{p}$ being equal to the target distribution
	$q$. However, when $q$ is a one-hot encoding (as in our case, and for most
	classification tasks), minimising focal loss does lead to $\hat{p}$ being
	equal to $q$.
\end{remark}

\subsection{Focal loss as a regularised Bregman divergence} 

We know that the cross-entropy loss forms an upper bound on the KL-divergence
between the target distribution $q$ and the predicted distribution $\hat{p}$,
i.e.\ $\mathcal{L}_c \geq \mathrm{KL}(q||\hat{p})$, so minimising cross-entropy
minimises $\mathrm{KL}(q||\hat{p})$. Interestingly, a general form of focal loss
can be shown to be an upper bound on the regularised KL-divergence, where the
regulariser is the negative entropy of the predicted distribution $\hat{p}$, and
the regularisation parameter is $\gamma$, the hyperparameter of focal loss.

\begin{restatable}[Focal Loss minimises a regularised Bregman divergence]{thm}{regularisedKL}
	\label{thm:focal-reg-Bregman}
	Let $q$ and $\hat{p}$ denote the target class probabilities and predicted
	posterior class probabilities respectively and $\mathcal{L}_f$ denote the
	focal loss with parameter $\gamma$. Then, 

	\begin{equation}\label{eq:focal-reg-Bregman}
		\mathcal{L}_f \geq \mathrm{KL}(q||\hat{p})+  \underbrace{\mathbb{H}[q]}_{constant} - \gamma \mathbb{H}[\hat{p}].
	\end{equation}\\
	Proof in~\Cref{sec:calibration-proof}
\end{restatable}

~\Cref{thm:focal-reg-Bregman} shows that minimising focal loss minimises the KL
divergence between $\hat{p}$ and $q$, whilst simultaneously increasing the
entropy of the predicted distribution $\hat{p}$. Thus replacing cross-entropy
with focal loss has the effect of adding a maximum-entropy regulariser
\citep{Pereyra2017} to the implicit minimisation of the KL-divergence that was
previously being performed by the cross-entropy loss. Encouraging the predicted
distribution $\hat{p}$ to have higher entropy can help avoid the overconfident
predictions produced by neural networks trained with NLL loss (see the `Peak at
the wrong place' paragraph of \Cref{sec:cause_cali}), and thereby improve
calibration.

In the case of one-hot target labels~(i.e. Dirac delta distribution for $q$),
the entropy of the target label probabilities $\bH\bs{q}$ is equal to $0$ and
the KL-divergence term reduces to $-\log \hat{p}_y$, where $y$ is the ground
truth class. So, focal loss maximises $-\log \hat{p}_y - \bH\bs{\hat{p}}$ and
prefers learning $\hat{p}$ such that $\hat{p}_y$ is assigned a high value
(because of the KL term $-\log \hat{p}_y$), but not too high (because of the
entropy term), and will ultimately avoid preferring overconfident models (by
contrast to cross-entropy loss). We solved the cross-entropy and focal loss
equations numerically, i.e. the value of the predicted probability $\hat{p}$
which minimises the loss, for various values of $q$ in a binary classification
problem and plotted it in Figure~\ref{fig:soln_p}. As expected, focal loss
favours a more entropic solution for $\hat{p}$ that is closer to the uniform
distribution. In other words,~\Cref{fig:soln_p} shows that solutions to
focal loss~(\Cref{eq:fc_loss}) will always have higher entropy than
cross-entropy.
\begin{equation}
\textrm{Solution for Focal Loss:}~\hat{p} = \mathrm{argmin}_x \; -(1-x)^\gamma q \log{x} - x^\gamma (1-q) \log{(1 - x)}\quad 0\le x\le 1\label{eq:fc_loss}
\end{equation}

\begin{figure}[t]
	\begin{center}
	  \includegraphics[width=0.4\linewidth]{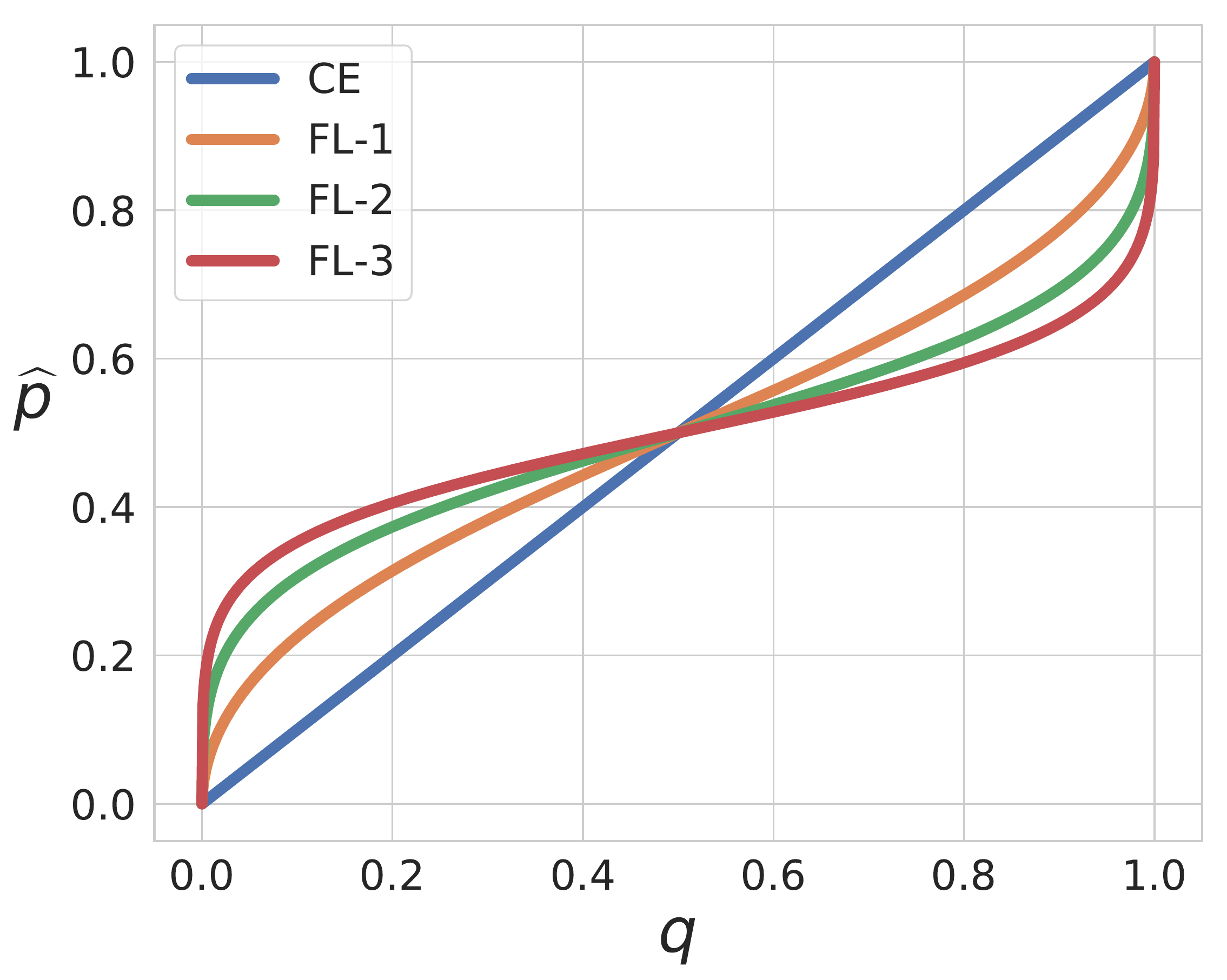}
	\end{center}
	\caption[Numerical solutions to NLL and focal loss]{Optimal $\hat{p}$ for various values of $q$. FL-$1$, FL-$2$, and FL-$3$ indicate Focal Loss with $\gamma=1,2,$ and $3$ respectively.}
	\label{fig:soln_p}
  \end{figure}

\subsection{Empirical observations on training with focal loss} To analyse the behaviour of neural
networks trained on focal loss, we use the same framework as mentioned above and
train four ResNet-50 networks on CIFAR-10: one using cross-entropy loss and
three using focal loss with $\gamma = 1, 2,$ and $3$. We plot various training
statistics related to these four networks in~\Cref{fig:nll_corr_incorr_entropy}.~\Cref{fig:test-nll-all} shows that the test NLL for the
cross-entropy model significantly increases towards the end of training (before
plateauing), whereas the test NLL for the focal loss models remains low. To
better understand this, we analyse the behaviour of these models separately for
correctly and incorrectly classified samples.
~\Cref{fig:test-nll-correct} shows that even though the NLL for
the correctly classified samples~(mostly) decreases over the course of training
for all models, the NLL for the focal loss models remain consistently higher
than that for the cross-entropy model throughout training, implying that the
focal loss models are relatively less confident than the cross-entropy model for
samples that they do predict correctly. This is important, as we have already
discussed that it is overconfidence that normally makes deep neural networks
miscalibrated.~\Cref{fig:test-nll-incorrect} shows that in
contrast to the cross-entropy model, for which the NLL for misclassified test
samples increases significantly after epoch $150$, the rise in this value for
the focal loss models is much less severe and almost absent for $\gamma=3$.
Additionally, in~\Cref{fig:test-nll-entropy}, we notice that the
entropy of the softmax distribution for misclassified test samples is
consistently (if only marginally) higher for focal loss than for cross-entropy.
This is consistent with~\Cref{thm:focal-reg-Bregman}.
\begin{figure}[!t]
	\centering
\begin{subfigure}[t]{0.32\linewidth}
		\includegraphics[width=0.99\linewidth]{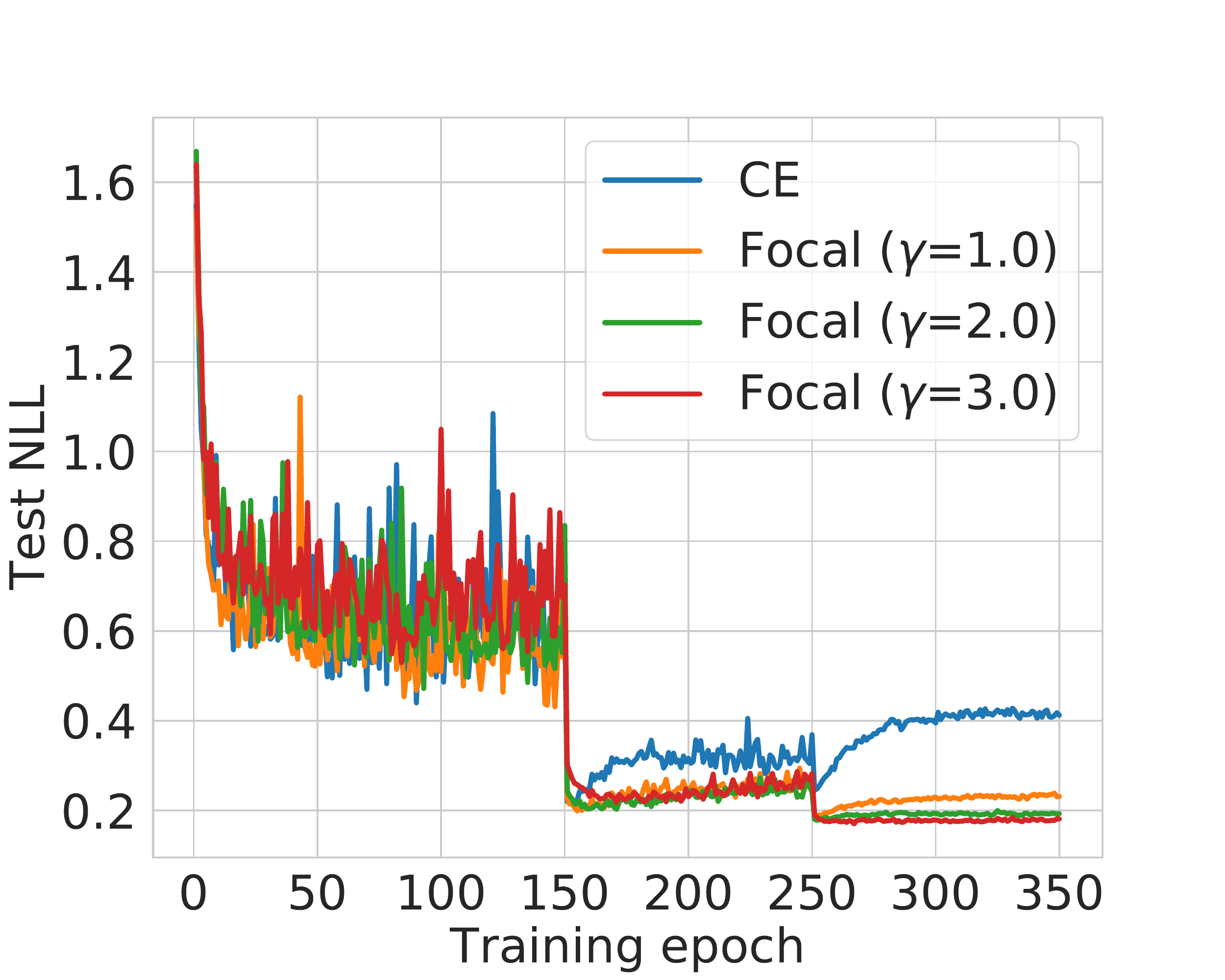}
		\caption{}
		\label{fig:test-nll-all}
\end{subfigure}
\begin{subfigure}[t]{0.32\linewidth}
	\includegraphics[width=0.99\linewidth]{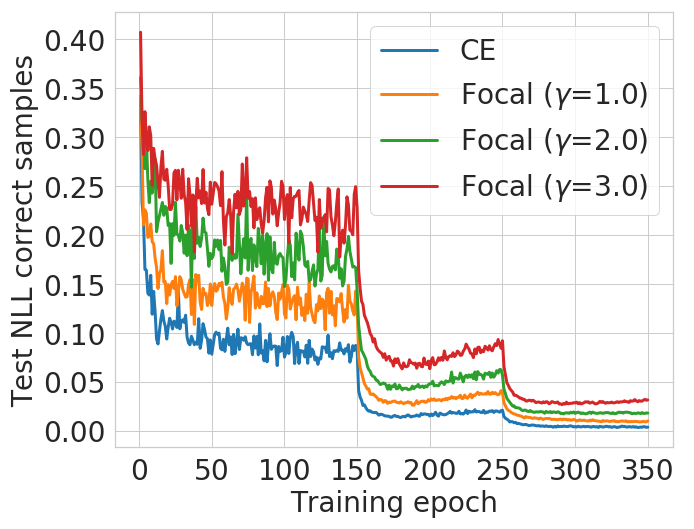}
	\caption{}
	\label{fig:test-nll-correct}
\end{subfigure}
	\begin{subfigure}[t]{0.32\linewidth}
	\includegraphics[width=0.99\linewidth]{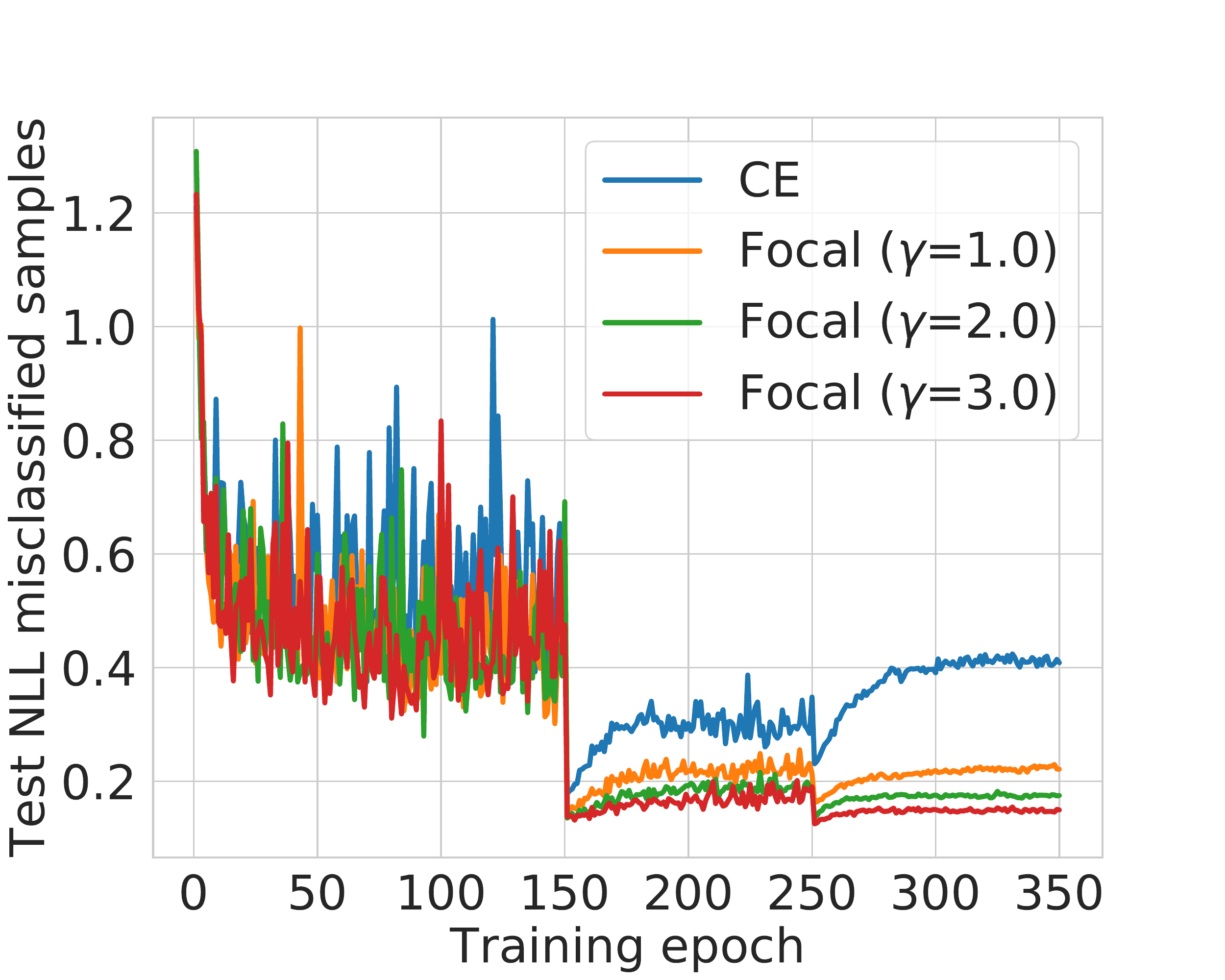}
	\caption{}
	\label{fig:test-nll-incorrect}
\end{subfigure}
	\begin{subfigure}[t]{0.33\linewidth}
	\includegraphics[width=0.99\linewidth]{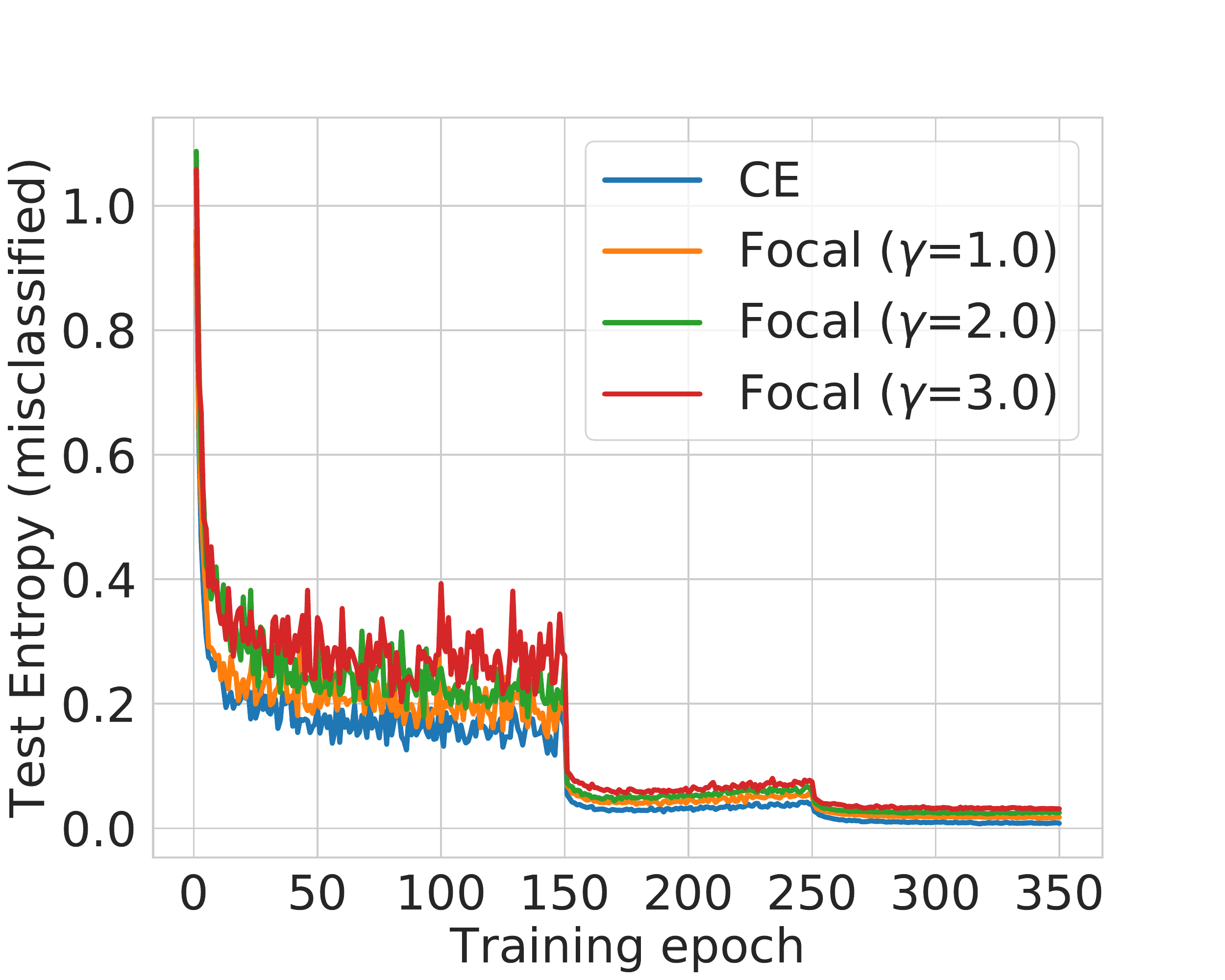}
	\caption{}
	\label{fig:test-nll-entropy}
\end{subfigure}
	\begin{subfigure}[t]{0.33\linewidth}
	\includegraphics[width=0.99\linewidth]{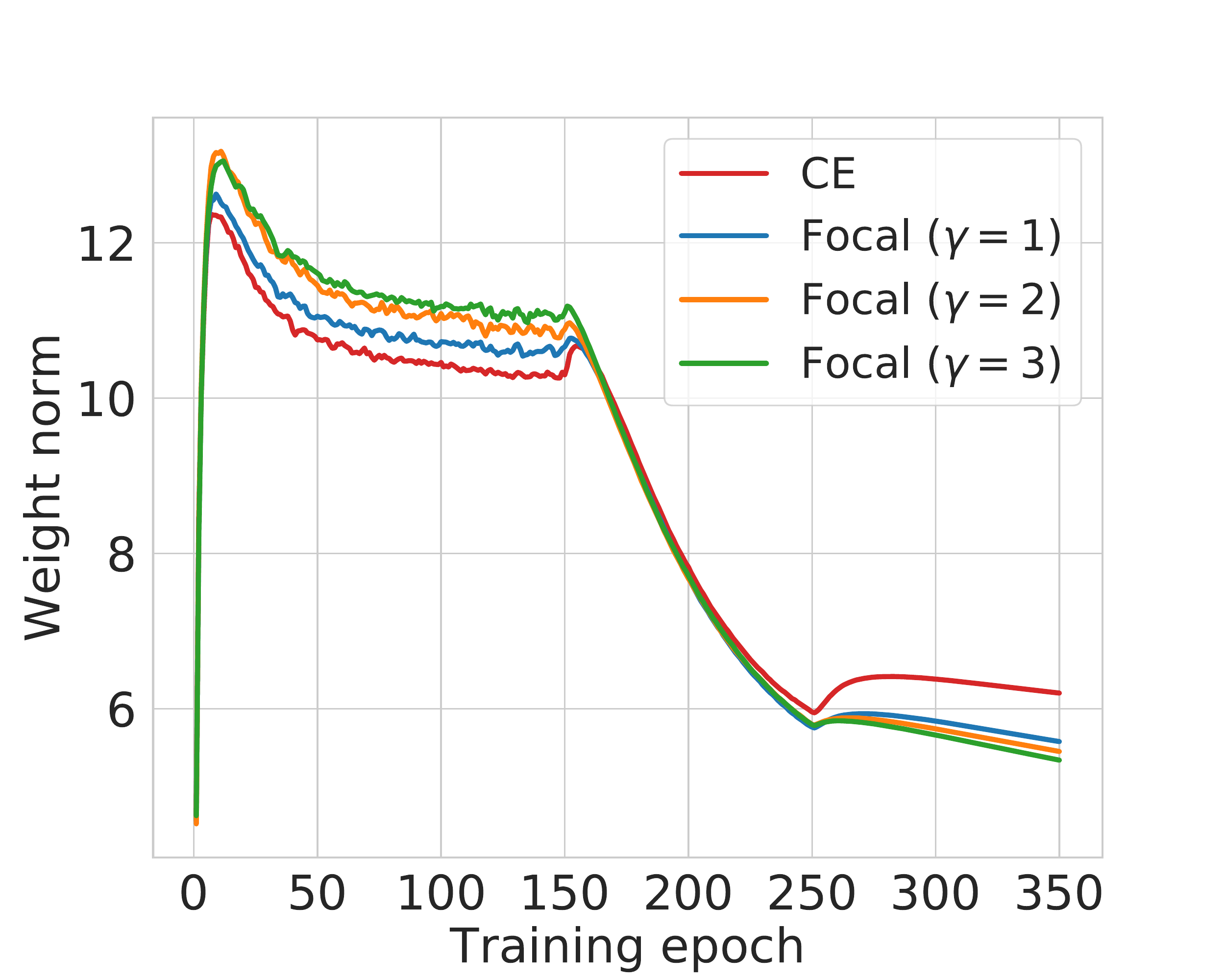}
	\caption{}
	\label{fig:test-nll-weight}
\end{subfigure}
	\caption[Calibration metrics during training of a ResNet-50 with Focal Loss]{How metrics related to model calibration change whilst training several ResNet-50 networks on CIFAR-10, using either cross-entropy loss or focal loss with $\gamma$ set to 1, 2 or 3.}
	\label{fig:nll_corr_incorr_entropy}
\end{figure}

\paragraph{Early Stopping}~\Cref{fig:test-nll-all} suggests that applying early
stopping when training a model on cross-entropy provides better calibration
scores. However, we could not find an ideal way of doing early stopping that
provides both the best calibration error and the best test-set accuracy. For a
fair comparison, we trained ResNet50 networks on CIFAR-10 using both
cross-entropy and focal loss with the best possible (in hindsight) early
stopping. We trained each model for 350 epochs and chose the 3 intermediate
models with the best validation set ECE, NLL, and classification error,
respectively. We present the test-set performance in
Table~\ref{table:early_stopping_table}.

From the table, we can observe that: 
\begin{enumerate}
\item For every early stopping criterion, focal loss outperforms cross-entropy
in both test-set accuracy and ECE, 
\item When using the validation set ECE as a stopping criterion,
the intermediate model for cross-entropy indeed improves its test-set ECE, but
at the cost of a significantly higher test error. 
\item Even without early stopping, focal loss achieves
consistently better error and ECE compared to cross-entropy using any
stopping criterion.
\end{enumerate}

As per \Cref{sec:cause_cali}, an increase in the test NLL and a decrease in the
test entropy for misclassified samples, along with no corresponding increase in
the test NLL for the correctly classified samples, can be interpreted as the
network predicting softmax distributions for the misclassified samples that are
ever more {\em peaky in the wrong place}. Notably, our results
in~\Cref{fig:test-nll-correct,fig:test-nll-incorrect,fig:test-nll-entropy}
clearly show that this effect is significantly reduced when training with focal
loss rather than cross-entropy, leading to a better-calibrated network whose
predictions are less peaky in the wrong place.
\begin{table}[!t]
	\centering
	\scriptsize
	\begin{tabular}{ccccc}
		\toprule
		\textbf{Criterion} & \textbf{Loss} & \textbf{Epoch} & \textbf{Error} & \textbf{ECE \%}\\
		\midrule
		ECE & NLL & 151 & 7.34 & 1.69 \\
		ECE & Focal Loss & 257 & 5.52 & 0.85 \\ 
		NLL & NLL & 153 & 6.69 & 2.28 \\
		NLL & Focal Loss & 266 & 5.34 & 1.33 \\
		Error & NLL & 344 & 5.0 & 4.46 \\
		Error & Focal Loss & 343 & 4.99 & 1.43 \\
		\midrule
		Full & NLL & 350 & 4.95 & 4.35 \\
		Full & Focal Loss & 350 & 4.98 & 1.55 \\
		\bottomrule
	\end{tabular}
	\caption[Test error and ECE with different early stopping
	criteria]{Classification errors and ECE scores obtained from ResNet-50
	models trained using cross-entropy and focal loss with different early
	stopping criteria (best in hindsight ECE, NLL and classification error on
	the validation set) applied during training. In the table, the {\em Full}
	Criterion indicates models where early stopping has not been applied.}
	\label{table:early_stopping_table}
\end{table}
\subsection{Theoretical justifications for focal loss} 
As mentioned previously, once a model trained using cross-entropy reaches high
training accuracy, the learning algorithm tries to further reduce the training
NLL by increasing the confidences of the correctly classified samples. It
achieves this by magnifying the network weights to increase the magnitudes of
the logits. To verify this hypothesis, we plot the $L_2$ norm of the weights of
the last linear layer for all four networks from the previous section as a
function of the training epoch in~\Cref{fig:test-nll-weight}. Notably, although
the norms of the weights for the models trained on focal loss are initially
higher than that for the cross-entropy model, \textit{a complete reversal} in
the ordering of the weight norms occurs between epochs $150$ and $250$. In other
words, as the networks start to become miscalibrated, the weight norm for the
cross-entropy model also starts to become greater than those for the focal loss
models. This is because focal loss, by design, starts to regularise the
network's weights once the model has gained a certain amount of confidence in
its predictions. To better understand this, we consider the following Lemma
\begin{restatable}[Relation between the gradients of cross-entropy and focal loss]{thm}{gradientfocalnll}
\label{pro1}
For focal loss $\cL_f$, with hyper-parameter $\gamma$, and cross-entropy loss
$\cL_c$ as defined before, the gradients with respect to the parameters for the
last linear layer $\vec{w}$ can be related as
\[\frac{\partial \cL_{\mathrm{f}}}{\partial \vec{w}} = \frac{\partial \cL_c}{\partial
\vec{w}} g(\hat{p}_{i,y_i}, \gamma)\] where $g(p, \gamma) = (1-p)^\gamma -
\gamma p (1-p)^{\gamma - 1} \log(p)$ and $\gamma \in \mathbb{R}^+$ is the focal
loss hyperparameter. Thus, 
\[\norm{\frac{\partial  \cL_f}{\partial \vec{w}}} \leq
\norm{\frac{\partial  \cL_c}{\partial \vec{w}}}\] if $g(\hat{p}_{i,y_i}, \gamma)
\in [0, 1]$.\\
Proof in~\Cref{sec:calibration-proof}
\end{restatable}

~\Cref{pro1} shows the relationship between the norms of the gradients of the
last linear layer for focal loss and cross-entropy loss, for the same network
architecture. This relation depends on the function $g(p, \gamma)$, which we
plot in~\Cref{fig:g_pt-vs-p}. It shows that for every $\gamma$, there exists a
unique threshold $p_0$ such that for all $p \in [0,p_0]$, $g(p,\gamma) \ge 1$,
and for all $p \in (p_0, 1]$, $g(p,\gamma) < 1$. (For example, for $\gamma = 1$,
$p_0 \approx 0.4$.) We use this insight to further explain why focal loss
provides implicit weight regularisation.

\begin{figure}[!t]
    \centering
    \includegraphics[width=0.4\linewidth]{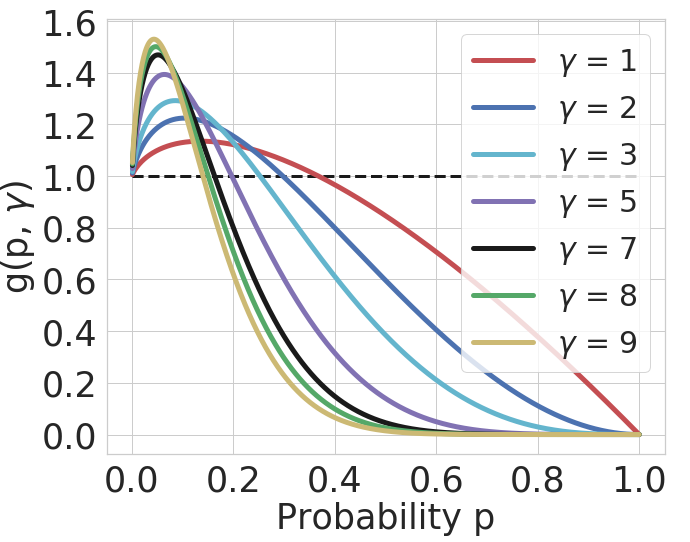}
    \caption{$g(p, \gamma)$ vs.\ $p$ }
	\label{fig:g_pt-vs-p}
\end{figure}

\paragraph{Implicit weight regularisation:} For a network trained using focal
loss with a fixed $\gamma$, during the initial stages of the training, when
$\hat{p}_{i,y_i} \in (0,p_0)$, $g(\hat{p}_{i,y_i}, \gamma) > 1$. Thus, the
confidences of the focal loss model's predictions initially increase faster
than cross-entropy. However, as soon as $\hat{p}_{i,y_i}$ crosses the threshold
$p_0$, $g(\hat{p}_{i,y_i}, \gamma)$ falls below $1$ and reduces the magnitude of
the gradient updates made to the network's weights, thereby having a
regularising effect on the weights. This is why, in~\Cref{fig:test-nll-weight},
we find that the norms of weights of the models trained with focal loss are
initially higher than that for the model trained using cross-entropy. However,
as training progresses, focal loss starts regularising the network weights and
the ordering of the weight norms reverses. We can draw similar insights
from~\Cref{fig:g_pt_grad_norms-10,fig:g_pt_grad_norms-100,fig:g_pt_grad_norms-200},
in which we plot histograms of the gradient norms of the last linear layer (over
all samples in the training set) at epochs $10$, $100$ and $200$, respectively.
At epoch $10$, the gradient norms for cross-entropy and focal loss are similar,
but as training progresses, those for cross-entropy decrease less rapidly than
those for focal loss, indicating that the gradient norms for focal loss are
consistently lower than those for cross-entropy throughout training.

\begin{figure}\centering
    \begin{subfigure}[c]{0.3\textwidth}
        \includegraphics[width=0.99\linewidth]{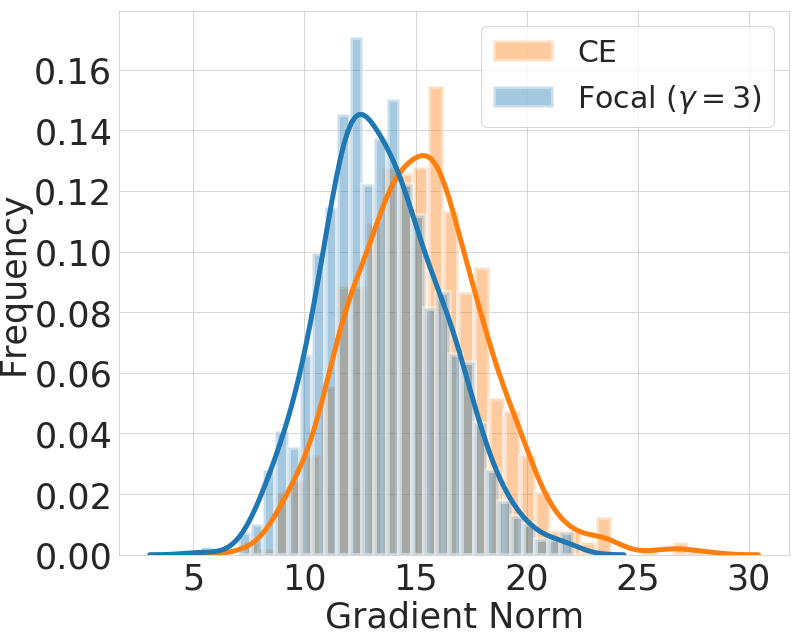}
        \caption{Epoch 10}
		\label{fig:g_pt_grad_norms-10}
    \end{subfigure}
    \begin{subfigure}[c]{0.3\linewidth}
        \includegraphics[width=0.99\linewidth]{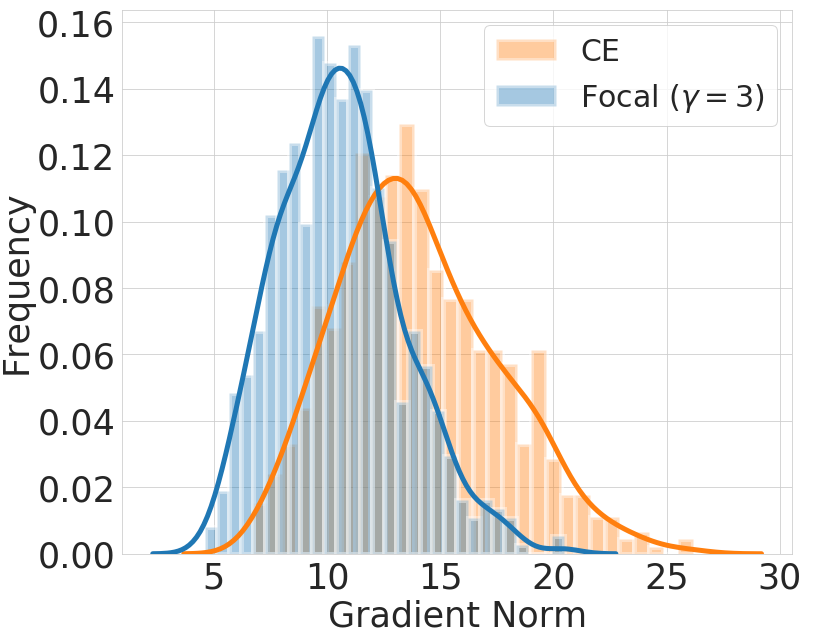}
        \caption{Epoch 100}
		\label{fig:g_pt_grad_norms-100}
    \end{subfigure}
    \begin{subfigure}[c]{0.3\linewidth}
        \includegraphics[width=0.99\linewidth]{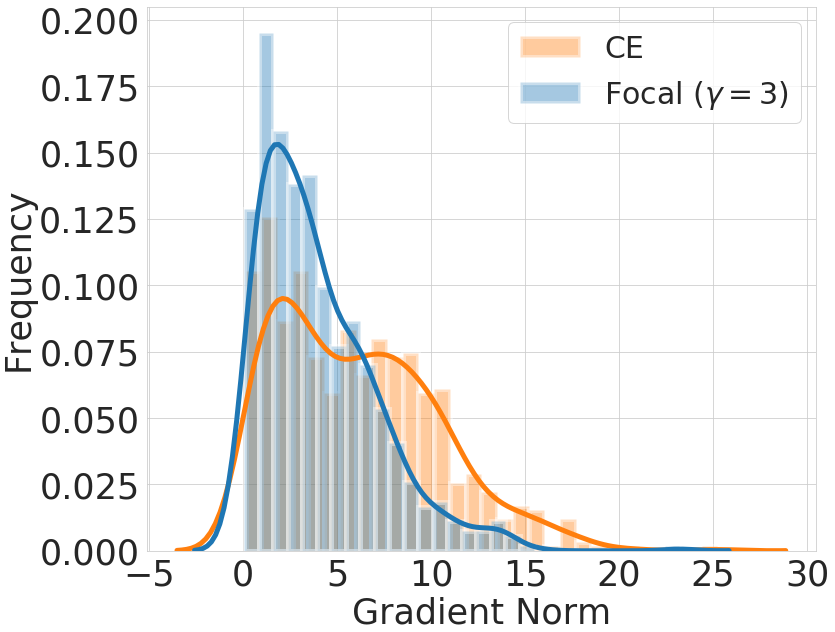}
        \caption{Epoch 200}
		\label{fig:g_pt_grad_norms-200}
    \end{subfigure}
    \caption[Distribution gradient norms for NLL and Focal Loss]{Histograms of the gradient norms of the last linear layer for both cross-entropy and focal loss.}
  \label{fig:g_pt_grad_norms}
\end{figure}

Finally, observe in~\Cref{fig:g_pt-vs-p} that for higher $\gamma$ values, the
fall in $g(p,\gamma)$ is steeper. We would thus expect a greater weight
regularisation effect for models that use higher values of $\gamma$. This
explains why, among the three models we trained using focal loss, the one with
$\gamma = 3$ outperforms (in terms of calibration) the one with $\gamma = 2$,
which in turn outperforms the model with $\gamma = 1$. Based on this
observation, one might think that a higher value of gamma always leads to a more
calibrated model. However, this is not the case, as we notice
from~\Cref{fig:g_pt-vs-p} that for $\gamma \ge 7$, $g(p,\gamma)$ reduces to
nearly $0$ for a relatively low value of $p$ (around $0.5$). As a result, using
values of $\gamma$ that are too high will cause the gradients to vanish~(i.e.\
reduce to nearly $0$) early, at a point at which the network's predictions
remain ambiguous, thereby causing the training process to fail. 

\subsection{How to choose the focal loss hyper-parameter $\gamma$}\label{sec:choose-gamma} We discussed that
 focal loss provides implicit entropy and weight regularisation and $\gamma$
 behaves akin to a regularisation coefficient.~\citet{Lin2017} fixed a $\gamma$,
 chosen by cross-validation, for all samples in the dataset. However, as we saw
 in~\Cref{pro1}, the regularisation effect for a sample $i$ depends on
 $\hat{p}_{i,y_i}$, the predicted probability for the ground truth label for the
 sample. It thus makes sense to choose $\gamma$ for each sample based on the
 value of $\hat{p}_{i,y_i}$. To this end, we provide~\Cref{pro:gamma}.
\begin{restatable}[Choosing the focal loss hyper-parameter]{thm}{choosinggamma}
	\label{pro:gamma}
	For a given $p_0>0$ and for all $1\ge p\ge p_0$ and $\gamma \ge \gamma^* = \frac{a}{b} +
	\frac{1}{\log a}W_{-1} \big(-\frac{a^{(1-a/b)}}{b} \log a \big)$
	where $a = 1-p_0$, $b = p_0 \log p_0$, and $W_{-1}$ is the
	Lambert-W function~\citep{Corless1996}, the following holds
	  \[g(p, \gamma) \leq 1\] 
	
	Moreover, for $p \geq p_0$ and $\gamma \geq \gamma^*$, the equality $g(p,
	\gamma) = 1$ holds only if $p = p_0$ and $\gamma = \gamma^*$.\\
	Proof in ~\Cref{sec:calibration-proof}
\end{restatable}
It is worth noting that for all values of $p\ge p_0$ there exist multiple values of $\gamma$ where $g(p, \gamma) \leq 1$. For a given $p_0$, ~\Cref{pro:gamma} allows us to compute $\gamma$ such that

\[ g(p,\gamma) = \begin{cases} 
	1 & p = p_0 \\
	>1 &  p \in [0,p_0) \\
	<1 & p \in (p_0, 1]
 \end{cases}
\]
This allows us to control the magnitude of the gradients for a particular sample
$i$ based on the current value of $\hat{p}_{i,y_i}$, and gives us a way of
choosing a value of $\gamma$ for each sample. For instance, a reasonable policy
might be to choose $\gamma$ s.t.\ $g(\hat{p}_{i,y_i}, \gamma) > 1$ if
$\hat{p}_{i,y_i}$ is small (say less than $0.25$), and \ $g(\hat{p}_{i,y_i},
\gamma) < 1$ otherwise. Such a policy will have the effect of making the weight
updates larger for samples having a low predicted probability for the correct
class and smaller for samples with a relatively higher predicted probability for
the correct class.

Following the aforementioned arguments, we choose a threshold of $p_0=0.25$ and
use~\Cref{pro:gamma} to obtain a policy for $\gamma$ such that $g(p, \gamma)$ is
observably greater than $1$ for $p \in [0, 0.2)$ and $g(p, \gamma) < 1$ for $p
\in (0.25, 1]$. In particular, we use the following schedule: if
$\hat{p}_{i,y_i} \in [0,0.25)$, then $\gamma = 5$, otherwise $\gamma = 3$ (note
that $g(0.2, 5) \approx 1$ and $g(0.25, 3) \approx 1$:
see~\Cref{fig:g_pt-vs-p}). We find this policy for $\gamma$ to perform
consistently well across multiple classification datasets and network
architectures. Having said that, one can calculate multiple such schedules for
$\gamma$ following Proposition~\ref{pro:gamma}, using the intuition of having a
relatively high $\gamma$ for low values of $\hat{p}_{i, y_i}$ and a relatively
low $\gamma$ for high values of $\hat{p}_{i, y_i}$.

\section{Training a linear model with focal loss and NLL}
\label{linear_model}

The behaviour of deep neural networks is generally quite different from linear
models and the problem of calibration is more pronounced in the case of deep
neural networks. Hence we focus on analysing the calibration of deep networks in
this chapter. However, weight norm analysis for the various layers in a deep
neural network is complex due to components of the training process like
batchnorm and weight decay. Hence, to see the effect of weight magnification on
miscalibration, first, we use a generalised linear model on  a simple data
distribution.

\begin{figure}[t]
    \centering
    \begin{subfigure}[c]{0.34\linewidth}
	\includegraphics[width=0.99\linewidth]{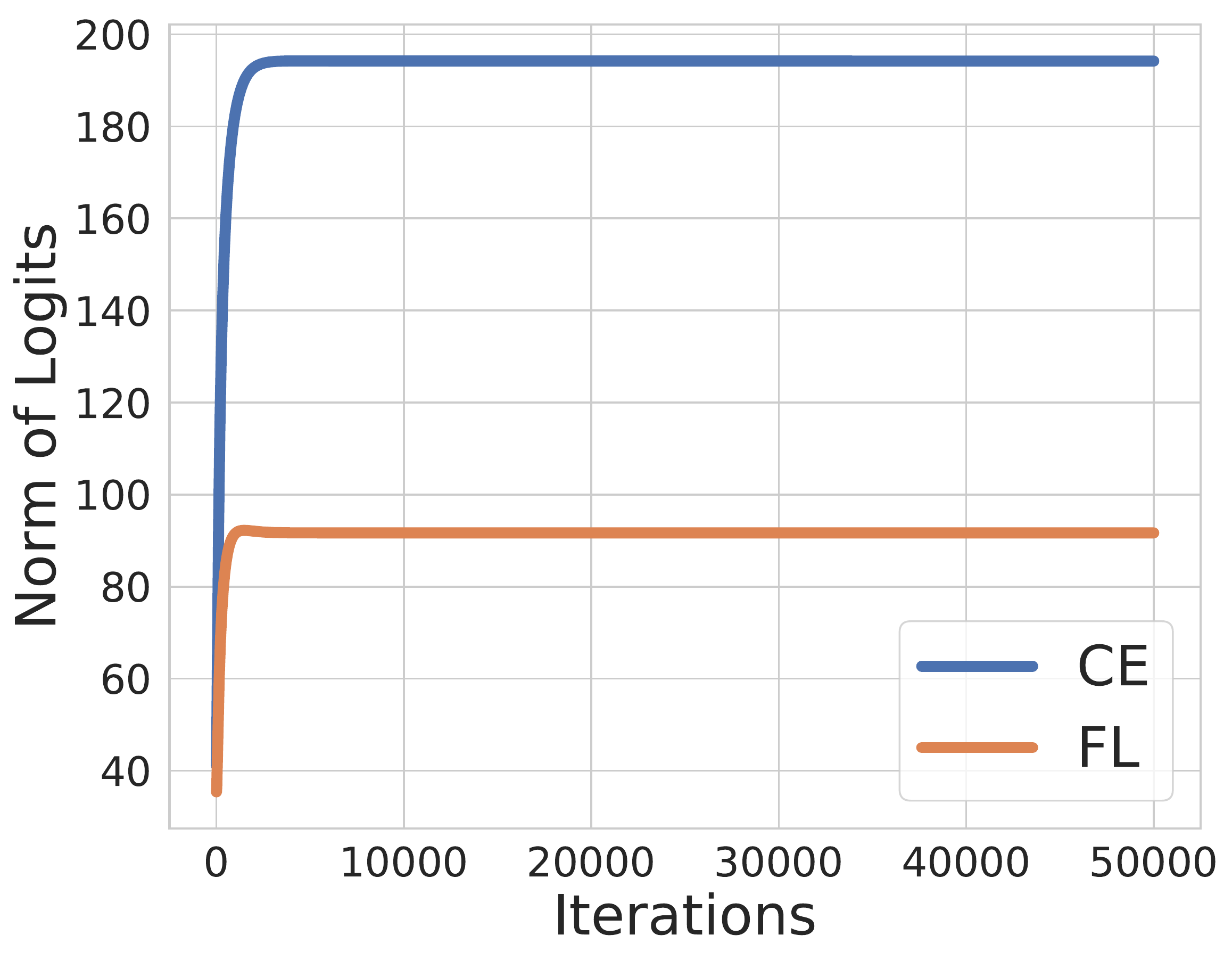}
    \caption{Norm of logits}
    \end{subfigure}
    \begin{subfigure}[c]{0.34\linewidth}
    \includegraphics[width=0.99\linewidth]{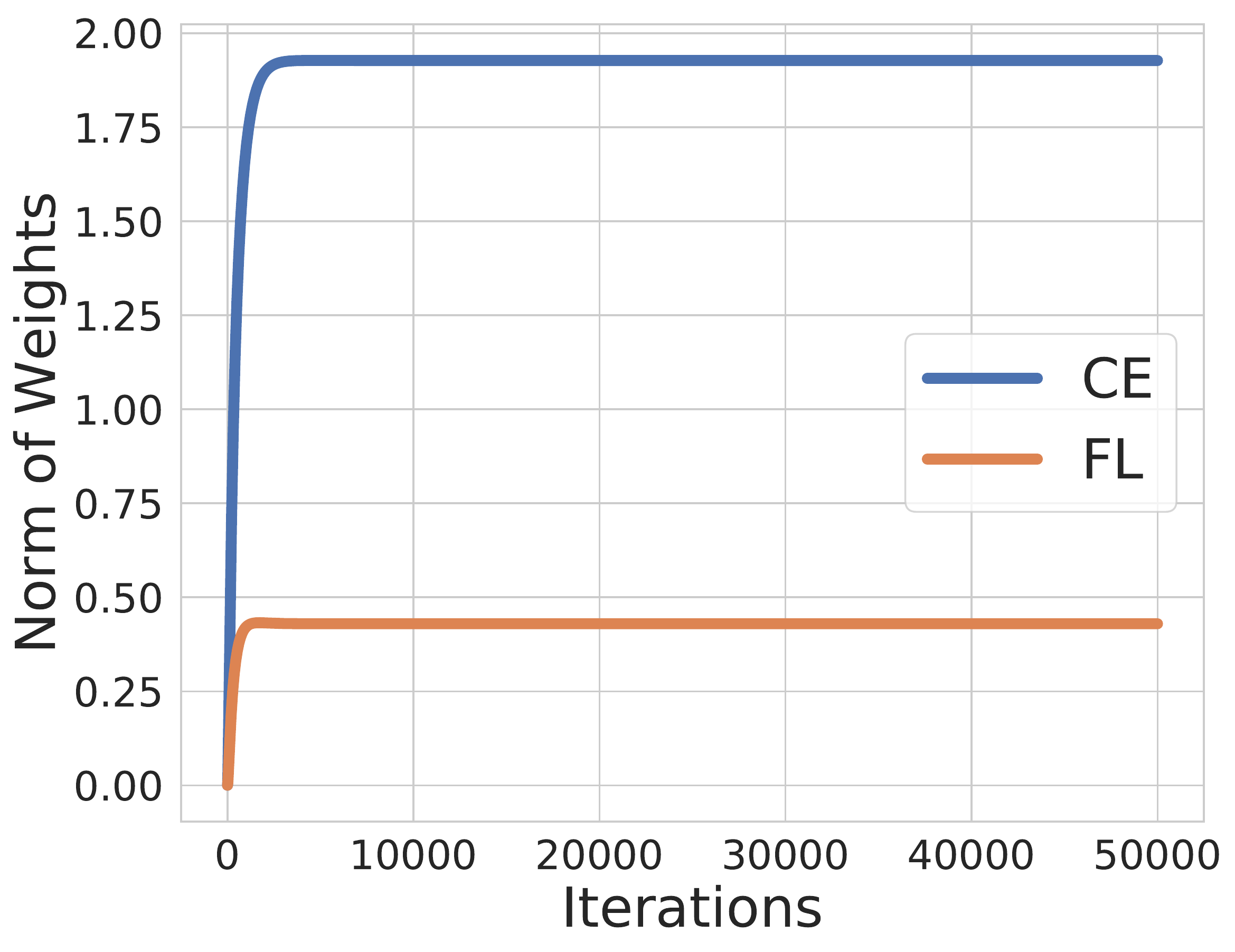}
	\caption{Norm of weights.}
	\end{subfigure}
	\caption[Training a linear model with NLL and focal loss]{Norm of logits and weights of Linear Classifiers trained
	with Focal Loss~(FL) and Cross-Entropy~(CE)}
	\label{fig:norms_linear}
\end{figure}

We consider a binary classification problem. The data matrix
$\vec{X}\in\reals^{2\times N}$ is created by assigning each class, two normally
distributed clusters such that the means of the clusters are linearly separable.
The means of the clusters are situated on the vertices of a two-dimensional
hypercube of side length 4. The standard deviation for each cluster is $1$ and
the samples are randomly linearly combined within each cluster to add
covariance. Further, for $10\%$ of the data points, the labels are flipped.
$4000$ samples are used for training and $1000$ samples are used for testing.
The model consists of a simple 2-parameter logistic regression model. For a
given $\vec{x}=\br{x_1,x_2}$, the model returns $f_{\br{w_1,w_2}}(\vec{x}) =
\sigma(w_1x_1+w_2x_2)$. We train this model using both cross-entropy and focal
loss with $\gamma = 1$.

\paragraph{Weight magnification} We have argued that focal loss implicitly
regularises the weights of the model by providing smaller gradients as compared
to cross-entropy. This helps in calibration as, if all the weights are large,
the logits are large and thus the confidence of the network is large on all test
points, even on the misclassified points. Consequently, when the model
misclassifies, it misclassifies with high confidence.
Figure~\ref{fig:norms_linear}  shows that the norms of the logits and the
weights are much larger for the model trained with the cross-entropy loss as
compared to the model trained with the focal loss.

\paragraph{High confidence for mistakes}
~\Cref{fig:dec-bound-ce,fig:dec-bound-fl} show that gradient descent with
cross-entropy (CE) and focal loss (FL) learns similar decision regions i.e. the
weight vector of the linear classifier points in the same direction. However, as
we have seen that the norm of the weights is much larger for CE as compared to
FL, we would expect the confidence of misclassified test points to be large for
CE as compared to FL. A histogram of the confidence of the misclassified points,
plotted in~~\Cref{fig:conf-linear} shows that CE almost always misclassifies
with greater than $90\%$ confidence whereas the confidence of misclassified
samples is much lower for the FL model.

\begin{figure}[t]
	\centering
	\begin{subfigure}[t]{0.28\linewidth}
	\includegraphics[width=0.99\linewidth]{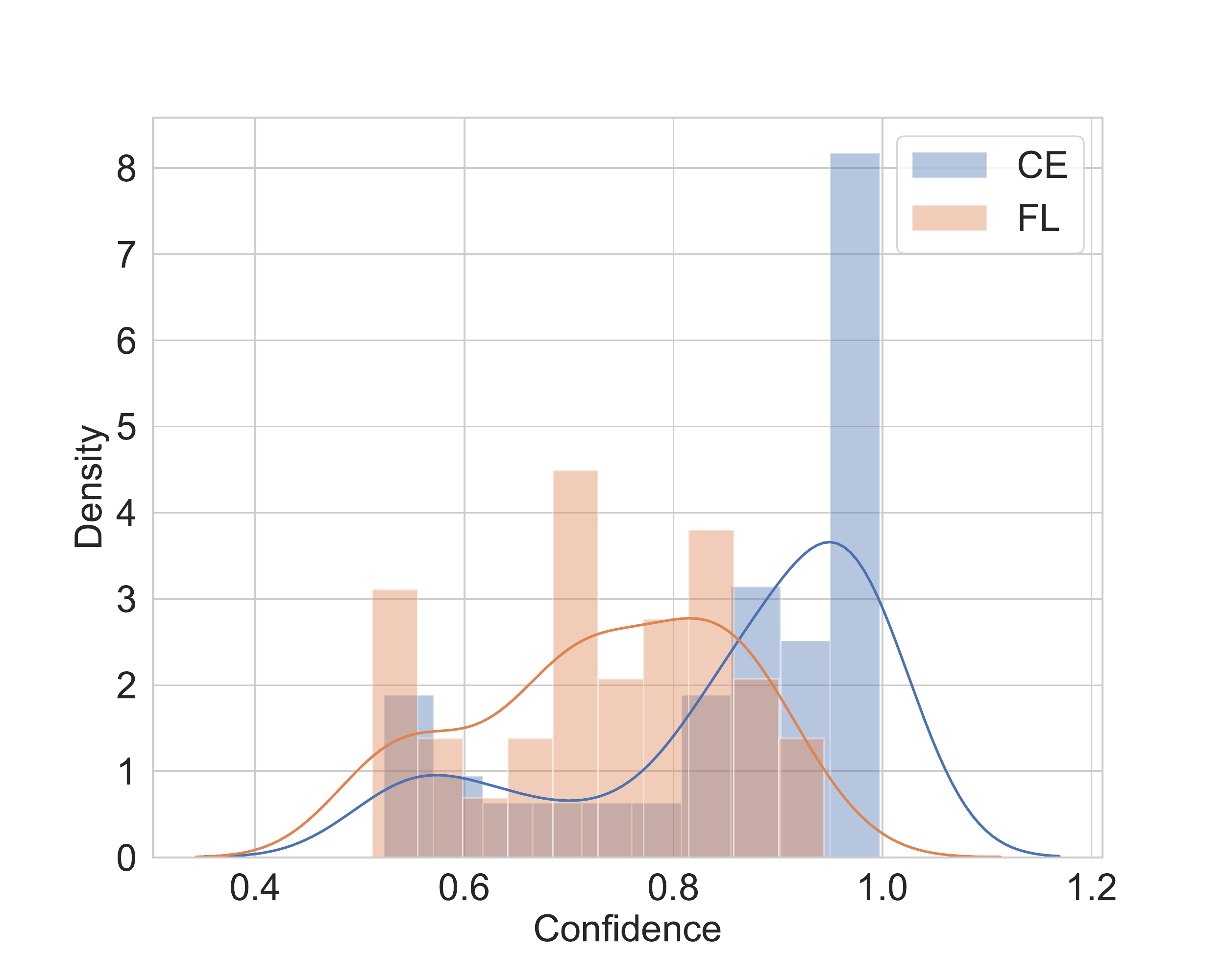}
	\caption{Confidence of mis-classifications}\label{fig:conf-linear}
	\end{subfigure}\hfill
	\begin{subfigure}[t]{0.32\linewidth}
	\includegraphics[width=0.99\linewidth]{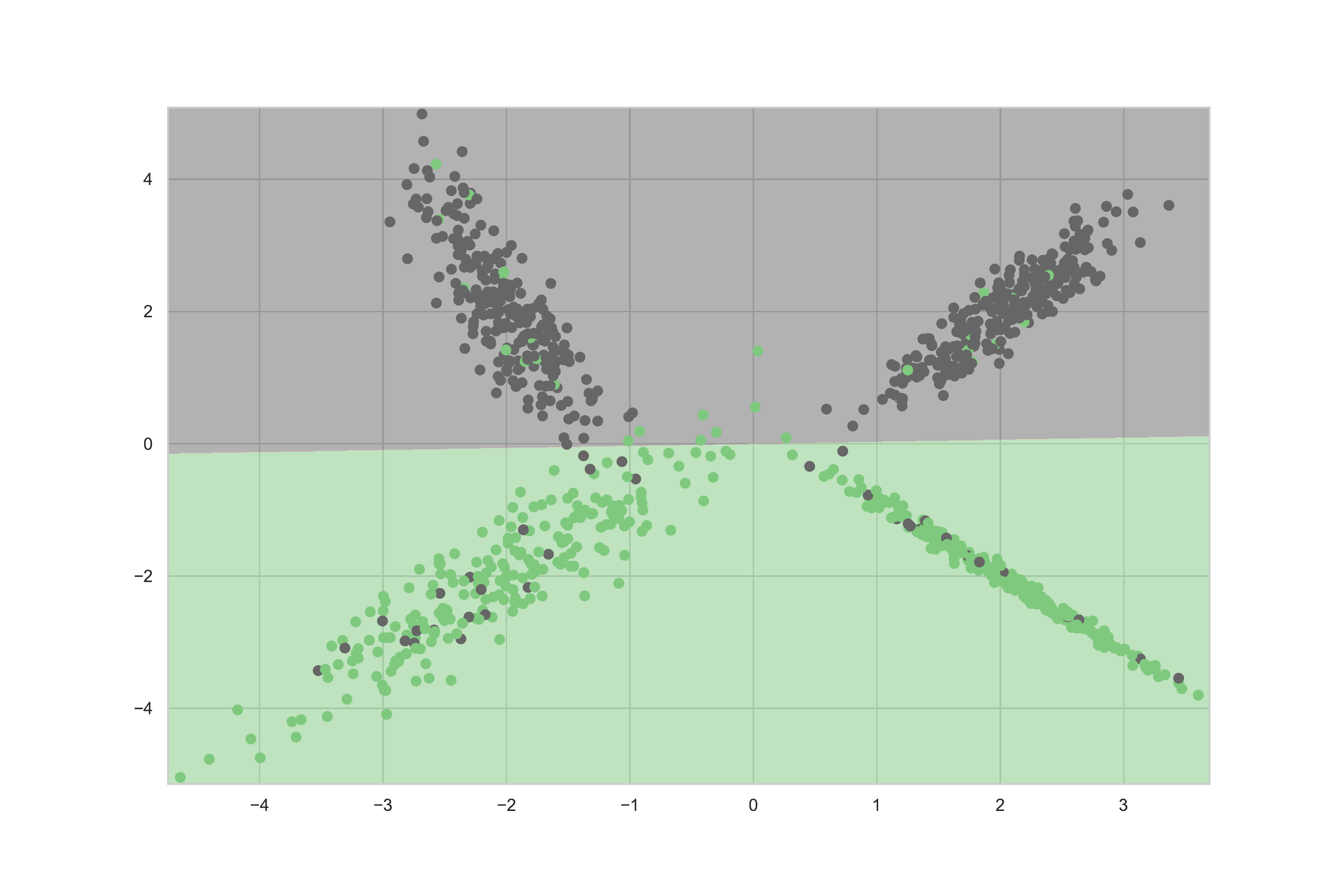}
	\caption{Decision boundary of linear classifier trained using cross-entropy}\label{fig:dec-bound-ce}
	\end{subfigure}
	\begin{subfigure}[t]{0.32\linewidth}
	\includegraphics[width=0.99\linewidth]{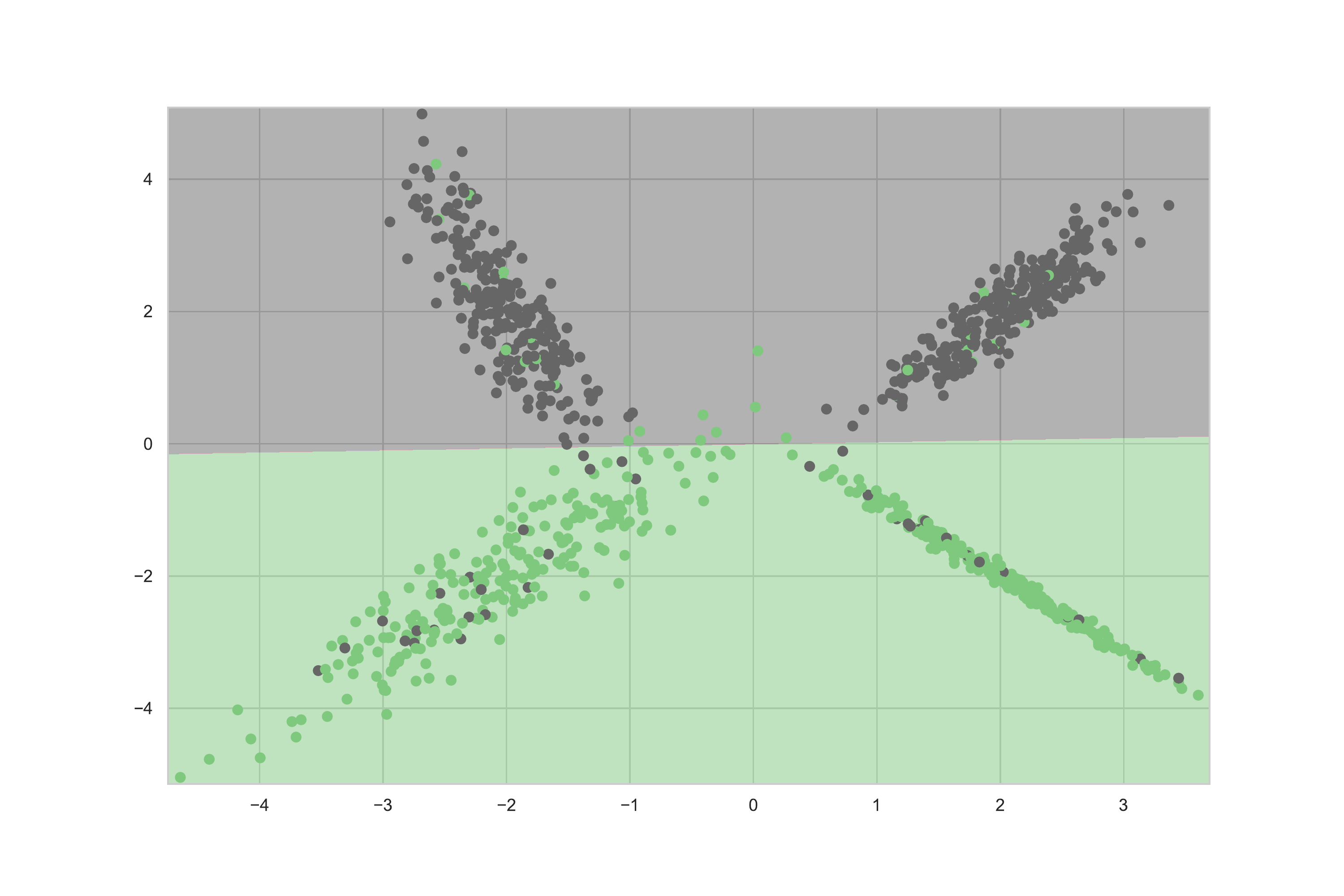}
	\caption[Decision boundary of a linear classifier trained using focal loss]{Decision boundary of linear classifier trained using focal loss}\label{fig:dec-bound-fl}
	\end{subfigure}
	\caption{Decision Boundary and confidences of Linear Classifiers.}
	\label{fig:conf_wrong_dec_bound}
\end{figure}

\todo[color=blue]{Remove 1. conf intervals, 2. OoD, 3. Figure F.5}
\section{Experiments on real-world datasets}
\label{sec:experiments}
We conduct image and document classification experiments to test the performance
of focal loss on more realistic models and datasets. For the former, we use
CIFAR-10/100 \citep{krizhevsky2009learning} and Tiny-ImageNet
\citep{imagenet_cvpr09} and train ResNet-50, ResNet-110 \citep{HZRS:2016},
Wide-ResNet-26-10 \citep{Zagoruyko2016}, and DenseNet-121 \citep{Huang2017}
models. For document classification experiments, we use the 20 Newsgroups
\citep{Lang1995} and Stanford Sentiment Treebank (SST) \citep{Socher2013}
datasets and train Global Pooling CNN \citep{Lin2014} and
Tree-LSTM~\citep{Tai2015} models. Further details on the datasets and training
can be found in~\Cref{sec:expr-settings}.
\begin{table*}[!t]
	\centering
	\scriptsize
	\resizebox{\linewidth}{!}{%
		\begin{tabular}{cccccccccccccc}
			\toprule
			\textbf{Dataset} & \textbf{Model} & \multicolumn{2}{c}{\textbf{Cross-Entropy}} &
			\multicolumn{2}{c}{\textbf{Brier Loss}} & \multicolumn{2}{c}{\textbf{MMCE}} &
			\multicolumn{2}{c}{\textbf{LS-0.05}} & \multicolumn{2}{c}{\textbf{FL-3 (Ours)}} &
			\multicolumn{2}{c}{\textbf{FLSD-53 (Ours)}} \\
			&& Pre T & Post T & Pre T & Post T & Pre T & Post T & Pre T & Post T & Pre T & Post T & Pre T & Post T \\
			\midrule
			
			\multirow{4}{*}{CIFAR-100} & ResNet-50&17.52&3.42(2.1)&6.52&3.64(1.1)&15.32&2.38(1.8)&7.81&4.01(1.1)&\tikzmark{top left}5.13&\textbf{1.97(1.1)}&\textbf{4.5}&2.0(1.1)\\
			& ResNet-110&19.05&4.43(2.3)&\textbf{7.88}&4.65(1.2)&19.14&\textbf{3.86(2.3)}&11.02&5.89(1.1)&8.64&3.95(1.2)&8.56&4.12(1.2)\\
			& Wide-ResNet-26-10&15.33&2.88(2.2)&4.31&2.7(1.1)&13.17&4.37(1.9)&4.84&4.84(1)&\textbf{2.13}&2.13(1)&3.03&\textbf{1.64(1.1)}\\
			& DenseNet-121&20.98&4.27(2.3)&5.17&2.29(1.1)&19.13&3.06(2.1)&12.89&7.52(1.2)&4.15&\textbf{1.25(1.1)}&\textbf{3.73}&1.31(1.1)\\
			\midrule
			\multirow{4}{*}{CIFAR-10} & ResNet-50&4.35&1.35(2.5)&1.82&1.08(1.1)&4.56&1.19(2.6)&2.96&1.67(0.9)&\textbf{1.48}&1.42(1.1)&1.55&\textbf{0.95(1.1)}\\
			& ResNet-110&4.41&1.09(2.8)&2.56&1.25(1.2)&5.08&1.42(2.8)&2.09&2.09(1)&\textbf{1.55}&\textbf{1.02(1.1)}&1.87&1.07(1.1)\\
			& Wide-ResNet-26-10&3.23&0.92(2.2)&\textbf{1.25}&1.25(1)&3.29&0.86(2.2)&4.26&1.84(0.8)&1.69&0.97(0.9)&1.56&\textbf{0.84(0.9)}\\
			& DenseNet-121&4.52&1.31(2.4)&1.53&1.53(1)&5.1&1.61(2.5)&1.88&1.82(0.9)&1.32&1.26(0.9)&\textbf{1.22}&\textbf{1.22(1)}\\
			\midrule
			Tiny-ImageNet & ResNet-50&15.32&5.48(1.4)&4.44&4.13(0.9)&13.01&5.55(1.3)&15.23&6.51(0.7)&1.87&1.87(1)&\textbf{1.76}&\textbf{1.76(1)}\\
			\midrule
			20 Newsgroups & Global Pooling CNN&17.92&2.39(3.4)&13.58&3.22(2.3)&15.48&6.78(2.2)&\textbf{4.79}&2.54(1.1)&8.67&3.51(1.5)&6.92&\textbf{2.19(1.5)}\\
			\midrule
			SST Binary & Tree-LSTM&7.37&2.62(1.8)&9.01&2.79(2.5)&5.03&4.02(1.5)&\textbf{4.84}&4.11(1.2)&16.05&1.78(0.5)&9.19&\textbf{1.83(0.7)}\tikzmark{bottom right}\\
			\bottomrule
		\end{tabular}%
	} \caption[ECE for different approaches]{ECE $(\%)$ computed for different approaches both pre and post
	temperature scaling (cross-validating T on ECE). The optimal temperature for
	each method is indicated in brackets. $T\approx 1$ indicates an innately
	calibrated model. }
	\label{table:ece_tab1}
\end{table*}

\paragraph{Baselines} Along with cross-entropy loss, we compare our method
against the following baselines: \begin{enumerate}
	\item \textbf{MMCE} (Maximum Mean Calibration Error) \citep{Kumar2018}, a
continuous and differentiable proxy for calibration error that is used
as a regulariser alongside cross-entropy, 
	\item \textbf{Brier loss}~\citep{Brier1950verification}, the squared error
between the predicted softmax vector and the one-hot ground truth encoding, and
	\item \textbf{Label smoothing}~\citep{muller2019does} (LS): given a one-hot
target label distribution $\bm{\mathrm{q}}$ and a smoothing factor $\alpha$
(hyperparameter), the smoothed vector $\bm{\mathrm{s}}$ is obtained as
$\bm{\mathrm{s}}_i = (1-\alpha)\bm{\mathrm{q}}_i +
\alpha(1-\bm{\mathrm{q}}_i)/(K-1)$, where $\bm{\mathrm{s}}_i$ and
$\bm{\mathrm{q}}_i$ denote the $i^{th}$ elements of $\bm{\mathrm{s}}$ and
$\bm{\mathrm{q}}$ respectively, and $K$ is the number of classes. Instead of
$\bm{\mathrm{q}}$, $\bm{\mathrm{s}}$ is treated as the target label distribution
during training. We train models using $\alpha = 0.05$ and $\alpha = 0.1$, but
find $\alpha = 0.05$ to perform better. Thus, we report the results obtained
from LS-$0.05$ with $\alpha = 0.05$.
\end{enumerate}

\paragraph{Focal loss}: We looked at different variants of focal loss, which
varies in the way $\gamma$ is assigned a value. When \(\gamma\) is fixed
throughout training, we found \(\gamma=3\) to outperform \(\gamma=1\) and
\(\gamma=2\). Thus, we use \(\gamma=3\) for our experiments and use the
abbreviation FL-3 to report its experimental results.

We also tried multiple variants of the Sample-Dependant
$\gamma$~(c.f.~\Cref{sec:choose-gamma}) and found the variant which uses the
following strategy to be most competitive --- when $\hat{p}_{i,y_i} \in [0, 0.25)$
, the value of \(\gamma\) is set to \(5\) and when $\hat{p}_{i,y_i} \in [0.25,
1)$, \(\gamma\) is set to \(3\). We call this approach Focal Loss
(sample-dependent $\gamma$ 5,3) and use the abbreviation FLSD-53 to report its experimental results.

\begin{table}[!t]
	\centering
	\scriptsize
	\resizebox{\linewidth}{!}{%
	\begin{tabular}{cccccccc}
	\toprule
	\textbf{Dataset} & \textbf{Model} & \textbf{Cross-Entropy} &
	\textbf{Brier Loss} & \textbf{MMCE} & \textbf{LS-0.05} & \textbf{FL-3 (Ours)} & \textbf{FLSD-53 (Ours)} \\
	
	\midrule
	\multirow{4}{*}{CIFAR-100} & ResNet-50&23.3&23.39&23.2&23.43&22.75&23.22\\
	& ResNet-110&22.73&25.1&23.07&23.43&22.92&22.51\\
	& Wide-ResNet-26-10&20.7&20.59&20.73&21.19&19.69&20.11\\
	& DenseNet-121&24.52&23.75&24.0&24.05&23.25&22.67\\
	\midrule
	\multirow{4}{*}{CIFAR-10} & ResNet-50&4.95&5.0&4.99&5.29&5.25&4.98\\
	& ResNet-110&4.89&5.48&5.4&5.52&5.08&5.42\\
	& Wide-ResNet-26-10&3.86&4.08&3.91&4.2&4.13&4.01\\
	& DenseNet-121&5.0&5.11&5.41&5.09&5.33&5.46\\
	\midrule
	Tiny-ImageNet & ResNet-50&49.81&53.2&51.31&47.12&49.69&49.06\\
	\midrule
	20 Newsgroups & Global Pooling CNN&26.68&27.06&27.23&26.03&29.26&27.98\\
	\midrule
	SST Binary & Tree-LSTM&12.85&12.85&11.86&13.23&12.19&12.8\\
	\bottomrule
	\end{tabular}}
	\caption[Test error $(\%)$ for different approaches]{Test set error $(\%)$ computed for different approaches. }
	\label{table:error_tab1}
	\end{table}

\paragraph{Temperature scaling:} To compute the optimal temperature for
temperature scaling, we use two different methods: (a) learning the temperature
by minimising validation set NLL and (b) performing grid search over
temperatures between 0 and 10 with a step size of 0.1, and choosing the one that
minimises validation set ECE. We find the second approach to produce {\em
stronger baselines} and report all our results obtained using this approach.

\subsection{Calibration and test accuracy} We report ECE$\%$ (computed using 15
bins) along with optimal temperatures in~\Cref{table:ece_tab1}, and test-set
error in~\Cref{table:error_tab1}. Firstly, for all dataset-network pairs, we
obtain very competitive classification accuracies (shown
in~\Cref{table:error_tab1}). This is important as it is easy to obtain a highly
calibrated model while incurring a large test error by simply predicting a
random class label with a uniform distribution over the classes. Secondly, {\em
it is clear from~\Cref{table:ece_tab1} that focal loss with sample-dependent
$\gamma$ and with $\gamma = 3$ outperforms all the baselines: cross-entropy,
label smoothing, Brier loss, and MMCE.} They produce the lowest calibration
errors {\em both before and after temperature scaling}. This observation is
particularly encouraging as it also indicates that a principled method for
obtaining values of $\gamma$ for focal loss can produce a very calibrated model
with no need to use a validation set for tuning $\gamma$.

	In~\Cref{table:ada_ece_tab1,table:sce_tab1,table:mce}, we present the
	AdaECE, Classwise-ECE, and MCE scores for our models and compare it with all the baselines discussed above.  The optimal temperature for each model is obtained by cross-validating it on ECE.

\begin{table*}[t]
	\centering
	\scriptsize
	\resizebox{\linewidth}{!}{%
		\begin{tabular}{cccccccccccccc}
			\toprule
			\textbf{Dataset} & \textbf{Model} & \multicolumn{2}{c}{\textbf{Cross-Entropy}} &
			\multicolumn{2}{c}{\textbf{Brier Loss}} & \multicolumn{2}{c}{\textbf{MMCE}} &
			\multicolumn{2}{c}{\textbf{LS-0.05}} & \multicolumn{2}{c}{\textbf{FL-3 (Ours)}} &
			\multicolumn{2}{c}{\textbf{FLSD-53 (Ours)}} \\
			&& Pre T & Post T & Pre T & Post T & Pre T & Post T & Pre T & Post T & Pre T & Post T & Pre T & Post T \\
			\midrule
			
			\multirow{4}{*}{CIFAR-100} & ResNet-50&17.52&3.42(2.1)&6.52&3.64(1.1)&15.32&2.38(1.8)&7.81&4.01(1.1)&\tikzmark{top left}5.08&2.02(1.1)&\textbf{4.5}&\textbf{2.0(1.1)}\\
			& ResNet-110&19.05&5.86(2.3)&\textbf{7.73}&4.53(1.2)&19.14&4.85(2.3)&11.12&8.59(1.1)&8.64&4.14(1.2)&8.55&\textbf{3.96(1.2)}\\
			& Wide-ResNet-26-10&15.33&2.89(2.2)&4.22&2.81(1.1)&13.16&4.25(1.9)&5.1&5.1(1)&\textbf{2.08}&2.08(1)&2.75&\textbf{1.63(1.1)}\\
			& DenseNet-121&20.98&5.09(2.3)&5.04&2.56(1.1)&19.13&3.07(2.1)&12.83&8.92(1.2)&4.15&\textbf{1.23(1.1)}&\textbf{3.55}&\textbf{1.24(1.1)}\\
			\midrule
			\multirow{4}{*}{CIFAR-10} & ResNet-50&4.33&2.14(2.5)&1.74&\textbf{1.23(1.1)}&4.55&2.16(2.6)&3.89&2.92(0.9)&1.95&1.83(1.1)&\textbf{1.56}&1.26(1.1)\\
			& ResNet-110&4.4&1.99(2.8)&2.6&1.7(1.2)&5.06&2.52(2.8)&4.44&4.44(1)&\textbf{1.62}&\textbf{1.44(1.1)}&2.07&1.67(1.1)\\
			& Wide-ResNet-26-10&3.23&1.69(2.2)&1.7&1.7(1)&3.29&1.6(2.2)&4.27&2.44(0.8)&1.84&1.54(0.9)&\textbf{1.52}&\textbf{1.38(0.9)}\\
			& DenseNet-121&4.51&2.13(2.4)&2.03&2.03(1)&5.1&2.29(2.5)&4.42&3.33(0.9)&\textbf{1.22}&1.48(0.9)&1.42&\textbf{1.42(1)}\\
			\midrule
			Tiny-ImageNet & ResNet-50&15.23&5.41(1.4)&4.37&4.07(0.9)&13.0&5.56(1.3)&15.28&6.29(0.7)&1.88&1.88(1)&\textbf{1.42}&\textbf{1.42(1)}\\
			
			\midrule
			20 Newsgroups & Global Pooling CNN&17.91&\textbf{2.23(3.4)}&13.57&3.11(2.3)&15.21&6.47(2.2)&\textbf{4.39}&2.63(1.1)&8.65&3.78(1.5)&6.92&2.35(1.5)\\
			\midrule
			SST Binary & Tree-LSTM&7.27&3.39(1.8)&8.12&2.84(2.5)&\textbf{5.01}&4.32(1.5)&5.14&4.23(1.2)&16.01&2.16(0.5)&9.15&\textbf{1.92(0.7)}\tikzmark{bottom right}\\
			\bottomrule
		\end{tabular}%
	}

	\caption[AdaECE for different approaches]{Adaptive ECE $(\%)$ computed for
	different approaches both pre and post temperature scaling
	(cross-validating T on ECE). Optimal temperature for each method
	is indicated in brackets. }
	\label{table:ada_ece_tab1}

\end{table*}

\begin{table*}[t]
	\centering
	\scriptsize
	\resizebox{\linewidth}{!}{%
		\begin{tabular}{cccccccccccccc}
			\toprule
			\textbf{Dataset} & \textbf{Model} & \multicolumn{2}{c}{\textbf{Cross-Entropy}} &
			\multicolumn{2}{c}{\textbf{Brier Loss}} & \multicolumn{2}{c}{\textbf{MMCE}} &
			\multicolumn{2}{c}{\textbf{LS-0.05}} & \multicolumn{2}{c}{\textbf{FL-3 (Ours)}} &
			\multicolumn{2}{c}{\textbf{FLSD-53 (Ours)}} \\
			&& Pre T & Post T & Pre T & Post T & Pre T & Post T & Pre T & Post T & Pre T & Post T & Pre T & Post T \\
			\midrule
			
			\multirow{4}{*}{CIFAR-100} & ResNet-50 & 0.38 & 0.22(2.1)&0.22&0.20(1.1)&0.34&0.21(1.8)&0.23&0.21(1.1)&\tikzmark{top left}\textbf{0.20}&\textbf{0.20(1.1)}&\textbf{0.20}&\textbf{0.20(1.1)}\\
			& ResNet-110&0.41&0.21(2.3)&0.24&0.23(1.2)&0.42&0.22(2.3)&0.26&0.22(1.1)&\textbf{0.24}&0.22(1.2)&\textbf{0.24}&\textbf{0.21(1.2)}\\
			& Wide-ResNet-26-10&0.34&0.20(2.2)&0.19&0.19(1.1)&0.31&0.20(1.9)&0.21&0.21(1)&\textbf{0.18}&\textbf{0.18(1)}&\textbf{0.18}&0.19(1.1)\\
			& DenseNet-121&0.45&0.23(2.3)&0.20&0.21(1.1)&0.42&0.24(2.1)&0.29&0.24(1.2)&0.20&0.20(1.1)&\textbf{0.19}&\textbf{0.20(1.1)}\\
			\midrule
			\multirow{4}{*}{CIFAR-10} & ResNet-50&0.91&0.45(2.5)&0.46&0.42(1.1)&0.94&0.52(2.6)&0.71&0.51(0.9)&0.43&0.48(1.1)&\textbf{0.42}&\textbf{0.42(1.1)}\\
			& ResNet-110&0.91&0.50(2.8)&0.59&0.50(1.2)&1.04&0.55(2.8)&0.66&0.66(1)&\textbf{0.44}&\textbf{0.41(1.1)}&0.48&0.44(1.1)\\
			& Wide-ResNet-26-10&0.68&0.37(2.2)&0.44&0.44(1)&0.70&0.35(2.2)&0.80&0.45(0.8)&0.44&0.36(0.9)&\textbf{0.41}&\textbf{0.31(0.9)}\\
			& DenseNet-121&0.92&0.47(2.4)&0.46&0.46(1)&1.04&0.57(2.5)&0.60&0.50(0.9)&0.43&0.41(0.9)&\textbf{0.41}&\textbf{0.41(1)}\\
			\midrule
			Tiny-ImageNet & ResNet-50&0.22&0.16(1.4)&0.16&0.16(0.9)&0.21&0.16(1.3)&0.21&0.17(0.7)&0.16&0.16(1)&\textbf{0.16}&\textbf{0.16(1)}\\
			
			\midrule
			20 Newsgroups & Global Pooling CNN&1.95&0.83(3.4)&1.56&\textbf{0.82(2.3)}&1.77&1.10(2.2)&\textbf{0.93}&0.91(1.1)&1.31&1.05(1.5)&1.40&1.19(1.5)\\
			\midrule
			SST Binary & Tree-LSTM&5.81&3.76(1.8)&6.38&2.48(2.5)&\textbf{3.82}&\textbf{2.70(1.5)}&3.99&3.20(1.2)&6.35&2.81(0.5)&4.84&3.24(0.7)\tikzmark{bottom right}\\
			\bottomrule
		\end{tabular}%
	}

	\caption[Classwise-ECE for different approaches]{Classwise-ECE $(\%)$ computed for
	different approaches both pre and post temperature scaling
	(cross-validating T on ECE). Optimal temperature for each method
	is indicated in brackets. }
	\label{table:sce_tab1}
\end{table*}

	\begin{table*}[t]
		\centering
		\footnotesize
		\resizebox{\linewidth}{!}{%
			\begin{tabular}{cccccccccccccc}
				\toprule
				\textbf{Dataset} & \textbf{Model} & \multicolumn{2}{c}{\textbf{Cross-Entropy}} &
				\multicolumn{2}{c}{\textbf{Brier Loss}} & \multicolumn{2}{c}{\textbf{MMCE}} &
				\multicolumn{2}{c}{\textbf{LS-0.05}} & \multicolumn{2}{c}{\textbf{FL-3 (Ours)}} &
				\multicolumn{2}{c}{\textbf{FLSD-53 (Ours)}} \\
				&& Pre T & Post T & Pre T & Post T & Pre T & Post T & Pre T & Post T & Pre T & Post T & Pre T & Post T \\
				\midrule
				\multirow{4}{*}{CIFAR-100} & ResNet-50&44.34&12.75(2.1)&36.75&21.61(1.1)&39.53&11.99(1.8)&26.11&18.58(1.1)&\textbf{13.02}&\textbf{6.76(1.1)}&16.12&27.18(1.1)\tikzmark{top right}\\
				& ResNet-110&55.92&22.65(2.3)&24.85&12.56(1.2)&50.69&19.23(2.3)&36.23&30.46(1.1)&26&13.06(1.2)&\textbf{22.57}&\textbf{10.94(1.2)}\\
				& Wide-ResNet-26-10&49.36&14.18(2.2)&14.68&13.42(1.1)&40.13&16.5(1.9)&23.79&23.79.1(1)&\textbf{9.96}&9.96(1)&10.17&\textbf{9.73(1.1)}\\
				& DenseNet-121&56.28&21.63(2.3)&15.47&8.55(1.1)&49.97&13.02(2.1)&43.59&29.95(1.2)&11.61&6.17(1.1)&\textbf{9.68}&\textbf{5.68(1.1)}\\
				\midrule
				\multirow{4}{*}{CIFAR-10} & ResNet-50&38.65&20.6(2.5)&31.54&22.46(1.1)&60.06&23.6(2.6)&35.61&40.51(0.9)&21.83&\textbf{15.76(1.1)}&\textbf{14.89}&26.37(1.1)\\
				& ResNet-110&44.25&29.98(2.8)&25.18&22.73(1.2)&67.52&31.87(2.8)&45.72&45.72(1)&25.15&37.610(1.1)&\textbf{18.95}&\textbf{17.35(1.1)}\\
				& Wide-ResNet-26-10&48.17&26.63(2.2)&77.15&77.15(1)&36.82&32.33(2.2)&24.89&37.53(0.8)&\textbf{23.86}&\textbf{25.64(0.9)}&74.07&36.56(0.9)\\
				& DenseNet-121&45.19&32.52(2.4)&19.39&19.39(1)&43.92&27.03(2.5)&45.5&53.57(0.9)&77.08&76.27(0.9)&\textbf{13.36}&\textbf{13.36(1)}\\
				\midrule
				Tiny-ImageNet & ResNet-50&30.83&13.33(1.4)&8.41&12.82(0.9)&26.48&12.52(1.3)&25.48&17.2(0.7)&6.11&6.11(1)&\textbf{3.76}&\textbf{3.76(1)}\\
				\midrule
				20 Newsgroups & Global Pooling CNN&36.91&36.91(3.4)&31.35&31.35(2.3)&34.72&34.72(2.2)&\textbf{8.93}&\textbf{8.93(1.1)}&18.85&18.85(1.5)&17.44&17.44(1.5)\\
				\midrule
				SST Binary & Tree-LSTM&71.08&88.48(1.8)&92.62&91.86(2.5)&68.43&32.92(1.5)&39.39&\textbf{35.72(1.2)}&\textbf{22.32}&74.52(0.5)&73.7&76.71(0.7)\tikzmark{bottom right}\\
				\bottomrule
			\end{tabular}%
		}
	
		\caption[MCE $(\%)$ for different approaches]{MCE $(\%)$ computed for
		different approaches both pre and post temperature scaling
		(cross-validating T on ECE). Optimal temperature for each method
		is indicated in brackets. }
		\label{table:mce}
	
	\end{table*}

Finally, calibrated models should have a higher logit score (or softmax
probability) on the correct class even when they misclassify, as compared to
models which are less calibrated. Thus, intuitively, such models should have a
higher Top-5 accuracy. Top-5 accuracy is the probability that one of the five
classes with the top five conditional likelihoods is the correct class. In
Table~\ref{table:top5}, we report the Top-5 accuracies for all our models on
datasets where the number of classes is relatively high (i.e., on CIFAR-100 with
100 classes and Tiny-ImageNet with 200 classes). We observe focal loss with
sample-dependent $\gamma$ to produce the highest top-5 accuracies on all models
trained on CIFAR-100 and the second-best top-5 accuracy (only marginally below
the highest accuracy) on Tiny-ImageNet.

\begin{table}[t]
	\renewcommand{\arraystretch}{1.3}
	\centering
	\footnotesize
	\resizebox{\linewidth}{!}{%
	\begin{tabular}{cccccccccccc}
	\toprule
	\textbf{Dataset} & \textbf{Model} & \multicolumn{2}{c}{\textbf{Cross-Entropy}} & \multicolumn{2}{c}{\textbf{Brier Loss}} & \multicolumn{2}{c}{\textbf{MMCE}} & \multicolumn{2}{c}{\textbf{LS-0.05}} & \multicolumn{2}{c}{\textbf{FLSD-53 (Ours)}}\\
	&& Top-1 & Top-5 & Top-1 & Top-5 & Top-1 & Top-5 & Top-1 & Top-5 & Top-1 & Top-5 \\
	
	\midrule
	\multirow{4}{*}{CIFAR-100} & ResNet-50&76.7&93.77&76.61&93.24&76.8&93.69&76.57&92.86&76.78&\textbf{94.44}\\
	& ResNet-110&77.27&93.79&74.9&92.44&76.93&93.78&76.57&92.27&77.49&\textbf{94.78}\\
	& Wide-ResNet-26-10&79.3&93.96&79.41&94.56&79.27&94.11&78.81&93.18&79.89&\textbf{95.2}\\
	& DenseNet-121&75.48&91.33&76.25&92.76&76&91.96&75.95&89.51&77.33&\textbf{94.49}\\
	\midrule
	Tiny-ImageNet & ResNet-50&50.19&74.24&46.8&70.34&48.69&73.52&52.88&\textbf{76.15}&50.94&76.07\\
	\bottomrule
	\end{tabular}}
	\caption{Top-1 and Top-5 accuracies for different approaches.}
	\label{table:top5}
	\end{table}

\subsection{Confident and calibrated models} It is worth noting that focal loss with sample-dependent $\gamma$ has optimal temperatures that are very close to 1, mostly lying between 0.9 and 1.1 (see Table~\ref{table:ece_tab1}). This property is shown by the Brier loss and label smoothing models as well, albeit with worse calibration errors. By contrast, the temperatures for cross-entropy and MMCE models are significantly higher, with values lying between 2.0 and 2.8. An optimal temperature close to 1 indicates that the model is innately calibrated and cannot be made significantly more calibrated by temperature scaling. In fact, a temperature much greater than 1 can make a model underconfident in general as it is applied to all predictions irrespective of the correctness of the model's outputs. 

We follow the approach adopted in~\citet{Kumar2018} and measure the percentage
of test samples that are predicted with a confidence of 0.99 or more (we call
this set of test samples $S99$). In~\Cref{table:side_table}, we report $|S99|$
as a percentage of the total number of test samples, along with the accuracy of
the samples in $S99$ for ResNet-50 and ResNet-110 trained on CIFAR-10, using
cross-entropy loss, MMCE loss, and focal loss. We observe that $|S99|$ for the
focal loss model is much lower than for the cross-entropy or MMCE models before
temperature scaling. However, after temperature scaling, $|S99|$ for focal loss
is significantly higher than for both MMCE and cross-entropy. The reason is that
with an optimal temperature of 1.1, the confidence of the temperature-scaled
model for focal loss does not reduce as much as it does for models trained with
cross-entropy and MMCE, for which the optimal temperatures lie between 2.5 and
2.8. We thus conclude that models trained on focal loss are not only more
calibrated, but also preserve their confidence on predictions, even after being
post-processed with temperature scaling.

\begin{table}[!t]
	\centering
	\scriptsize
	\resizebox{\linewidth}{!}{%
	\begin{tabular}{cccccccccccccc}
	\toprule
	\textbf{Dataset} & \textbf{Model} & \multicolumn{2}{c}{\textbf{Cross-Entropy (Pre T)}} & \multicolumn{2}{c}{\textbf{Cross-Entropy (Post T)}} & \multicolumn{2}{c}{\textbf{MMCE (Pre T)}} & \multicolumn{2}{c}{\textbf{MMCE (Post T)}} & \multicolumn{2}{c}{\textbf{Focal Loss (Pre T)}} & \multicolumn{2}{c}{\textbf{Focal Loss (Post T)}} \\
	&& |S99|$\%$ & Accuracy & |S99|$\%$ & Accuracy & |S99|$\%$ & Accuracy & |S99|$\%$ & Accuracy & |S99|$\%$ & Accuracy & |S99|$\%$ & Accuracy \\
	\midrule
	CIFAR-10 & ResNet-110 & $97.11$ & $96.33$ & $11.5$ & $97.39$ & $97.65$ & $96.72$ & $10.62$ & $99.83$ & $61.41$ & $99.51$ & $31.10$ & $99.68$ \\ 
	CIFAR-10 & ResNet-50 & $95.93$ & $96.72$ & $7.33$ & $99.73$ & $92.33$ & $98.24$ & $4.21$ & $100$ & $46.31$ & $99.57$ & $14.27$ & $99.93$ \\ 
	\bottomrule
	\end{tabular}} 
	\caption[Percentage of test samples predicted with
	confidence higher than $99\%$]{Percentage of test samples predicted with
	confidence higher than $99\%$ and the corresponding accuracy for Cross
	Entropy, MMCE and Focal loss computed both pre and post temperature
	scaling (represented in the table as pre T and post T respectively).}
	\label{table:side_table}
	\end{table}

\chapter{Accelerating Encrypted Prediction via Binary Neural Networks}
\label{chap:TAPAS}

In this chapter, we will look at an issue that is commonly faced when neural
networks are deployed in the real world in a {\em Prediction As a Service}
framework~(see~\Cref{sec:privacy} for a discussion of this framework). In
particular, we look at a setting where a service provider has trained a machine
learning model and users can use that model by sending their private data to the
service provider.  However, neither the user wants the service provider to be
able to read their private data or the output of the model on that data nor does
the service provider want the user to learn anything about their machine
learning model. We propose some strict computational and privacy requirements
that encapsulate this setting and refer to this paradigm as {\em Encrypted
Prediction As A Service}~(EPAAS).

One way of implementing this is through the technique of Fully Homomorphic
Encryption~(FHE), which allows for arbitrary computations on encrypted data
without decrypting it. However, this can be incredibly slow in practice
rendering the technique useless in the real world. In this chapter, along with
proposing the EPAAS framework, we also discuss ways to make the implementation
of FHE for EPAAS faster in practice so that it can be applied on modern complex
neural networks.
\section{Main contributions}
\label{ssec:contrib}
In this chapter, our focus is on achieving speed-ups when using complex models
with fully homomorphically encrypted data. To achieve these speed-ups,
we propose several methods to modify the training and design of neural networks,
as well as algorithmic tricks to parallelise and accelerate computation on
encrypted data. In particular, we propose the following techniques
\begin{itemize}
	\item We propose the use of binary neural networks to make EPAAS faster
	without sacrificing much on its test accuracy.
	\item We propose two types of circuits for performing inner products between
		unencrypted and encrypted data: reduce tree circuits and sorting
		networks. We give a runtime comparison of each method.
	\item We introduce an easy trick, which we call the \emph{+1 trick} to
		sparsify encrypted computations.
	\item We demonstrate that our techniques are easily parallelisable and we report 
		timing for a variety of computation settings on real-world datasets,
		alongside classification accuracies.
\end{itemize}

Most similar to our work is \citet{BMM+:2017} who use neural networks with
signed integer weights and binary activations to perform encrypted predictions.
However, their method is only evaluated on MNIST and achieves accuracy
comparable to a linear classifier ($92\%$)~\citep{LBBH:1998}, and the encryption
scheme parameters depend on the structure of the model, potentially requiring
clients to re-encrypt their data if the service provider updates their model.
Our framework allows the service provider to update their model at any time and
allows one to use binary neural networks of~\citet{CHSEB:2016} which, in
particular, achieve high accuracy on MNIST ($99.04\%$). Another closely related
work is \citet{MLH:2018} who design encrypted adder and multiplier circuits so
that they can implement machine learning models on integers. This can be seen as
complementary to our work on binary networks: while they achieve improved
accuracy because of greater precision, they are less efficient than our methods
(on MNIST we achieve the same accuracy with a $29\times$ speedup, via
our sparsification and parallelisation tricks).

\paragraph{Private training.}
In this thesis, we do not address the question of training machine learning
models with encrypted data. There has been some recent work in this area
\citep{HHI-LNPST:2017,AHWM:2017}. However, as of now, it appears possible only
to train very small models using fully homomorphic encryption. We leave this for
future work. %

\section{Encrypted prediction as a Service}
\label{sec:problem-definition}
In this section, we describe our \emph{Encrypted Prediction as a Service (EPAAS)} paradigm. We then detail our privacy and computational guarantees. Finally, we discuss how different related work is suited to this paradigm and propose a solution.

\begin{figure}[t!]
    \centerline{\includegraphics[width=0.5\columnwidth]{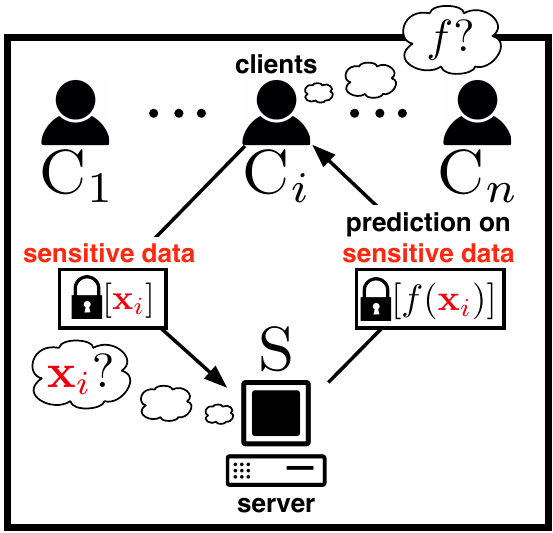}}
  \caption[EPAAS framework]{Encrypted prediction as a service. }
  \label{figure.service}
\end{figure}

In the EPAAS setting we have any number of clients, say $C_1, \ldots, C_n$ that have data $\x_1,\ldots,\x_n$. The clients would like to use a highly accurate model $f$ provided by a server $S$ to predict some outcome. In cases where data $\x$ is not sensitive, there are already many solutions for this such as BigML\footnotemark[1], Wise.io\footnotemark[2], Google Cloud AI\footnotemark[3], Amazon Machine Learning\footnotemark[4], among others. However, if the data is sensitive so that the clients would be uncomfortable giving the raw data to the server, none of these systems can offer the client a prediction. 

\footnotetext[1]{https://wise.io/}
\footnotetext[2]{https://cloud.google.com/products/machine-learning/}
\footnotetext[3]{https://bigml.com/}
\footnotetext[4]{https://aws.amazon.com/machine-learning/}
\begin{table}
\centering\small
\begin{tabular}{c|cc|cccc} 
\hline
 &  \multicolumn{2}{c}{\bf Privacy} & \multicolumn{4}{|c}{\bf Computation} \\
Prior Work  &  {P1} & {P2} & {C1} & {C2} & {C3(i)} & {C3(ii)} \\
\hline
CryptoNets~\cite{G-BDL+:2016} & \checkmark & - & \checkmark & \checkmark & - & \checkmark \\
\hline
\cite{CWMMP:2017} & \checkmark & - & \checkmark & \checkmark & - & \checkmark \\
\hline
\cite{BMM+:2017} & \checkmark & \checkmark & \checkmark & \checkmark & - & \checkmark \\
\hline
\begin{tabular}{@{}c@{}}MPC~ \cite{MZ:2017,LJL+:2017} \\ \cite{RWT+:2017,CG-BLLR:2017,JVC:2018}\end{tabular} & \checkmark & - & \checkmark & - & - & - \\
\hline
\cite{MLH:2018}, Ours & \checkmark & \checkmark & \checkmark & \checkmark & \checkmark & \checkmark \\
\hline
\end{tabular}
\caption[Privacy and computational guarantees of existing methods.]{Privacy and computational
	guarantees of existing methods for sensitive data classification.}
	\label{table.existing}
\end{table}

\subsection{Privacy and computational guarantees}
If data $\x$ is sensitive (e.g., $\x$ may be the health record of client $C$, and $f(\x)$ may be the likelihood of heart disease), then we would like to have the following privacy guarantees:
\begin{enumerate}
\item[P1.] Neither the server $S$ nor any other party, learn anything about
client data $\x$, other than its size {\em (privacy of the data)}.
\item[P2.] Neither the client $C$ nor any other party, learn anything about
  model $f$, other than the prediction $f(x)$ given client data $\x$ (and
  whatever can be deduced from it) {\em (privacy of the model)}. 
\end{enumerate}
Further, the main attraction of EPAAS is that the client is involved as little as possible. More concretely, we wish to have the following computational guarantees:
\begin{enumerate}
\item[C1.] No external party is involved in the computation.
\item[C2.] The rounds of communication between client and server should be limited to $2$ (send data \& receive prediction).
\item[C3.] Communication and computation at the client-side
  should be independent of model $f$. In particular,
  (i) the server should be able to update $f$ without communicating with any client,
  and (ii) clients should not need to be online during the computation of $f(\x)$.
\end{enumerate}
Note that these requirements rule out protocols with preprocessing stages or
that involve third parties.
Generally speaking, a satisfactory solution based on FHE would proceed as follows:
(1) a client generates encryption parameters, encrypts their data $\x$ using
the private key, and sends the resulting encryption $\tilde{\x}$, as well as the
public key to the server. (2) The server evaluates $f$ on $\tilde{\x}$ leveraging the homomorphic
properties of the encryption, to obtain an encryption $\tilde{f}(\x)$
without learning anything whatsoever about $\x$, and sends $\tilde{f}(\x)$
to the client. (3) Finally, the client decrypts and recovers the prediction
$f(\x)$ in the clear. A high-level depiction of these steps is shown in Figure~\ref{figure.service}.

\subsection{Existing approaches}
Table~\ref{table.existing} describes whether prior work satisfies the above
privacy and computational guarantees. First, note that
Cryptonets~\cite{G-BDL+:2016} violates C3(i) and P2. This is because the clients
would have to generate parameters for the encryption according to the structure
of $f$, so the client can make inferences about the model (violating P2)
and the client is not allowed to change the model $f$ without telling the client
(violating C3(i)). The same holds for the work of \citet{CWMMP:2017}. The
approach of \citet{BMM+:2017} requires the server to calibrate the parameters of
the encryption scheme according to the magnitude of intermediate values, thus
C3(i) is not necessarily satisfied. Closely related to our work is that of
\citet{MLH:2018} which satisfies our privacy and computational requirements. We
will show that our method is significantly faster than this method, with very
little sacrifice in accuracy.

\paragraph{Multi-Party Computation (MPC).}
It is important to distinguish between approaches based purely on homomorphic
encryption (described above), and those involving Multi-Party Computation (MPC)
techniques, such
as~\cite{MZ:2017,LJL+:2017,RRK:2017,RWT+:2017,CG-BLLR:2017,JVC:2018}. While
generally, MPC approaches are faster, they crucially rely on all parties being
involved in the whole computation, which  conflicts with requirement C3(ii).
Additionally, in MPC the structure of the computation is public to both parties,
which means that the server would have to communicate basic information such as
the number of layers of $f$. This conflicts with requirements P2, C2, and C3(i).

In this work, we propose to use a very tailored homomorphic encryption technique
to guarantee all privacy and computational requirements. In the next section, we
give background on homomorphic encryption. Further, we motivate the encryption
protocol and the machine learning model class we use to satisfy all guarantees.

\section{Background}
\label{sec:prelim}
All cryptosystems define two functions: 1. an encryption function
$\mathcal{E}(\cdot)$ that maps data (often called \emph{plaintexts})
to encrypted data (\emph{ciphertexts}); 2. a decryption function
$\mathcal{D}(\cdot)$ that maps ciphertexts back to plaintexts. In
public-key cryptosystems, to evaluate the encryption function
$\mathcal{E}$, one needs to hold a public key $k_\textsc{pub}$, so the
encryption of data $x$ is $\mathcal{E}(x, k_\textsc{pub})$. Similarly,
to compute the decryption function $\mathcal{D}(\cdot)$ one needs to
hold a secret key $k_\textsc{sec}$ which allows us to recover:
$\mathcal{D}(\mathcal{E}(x, k_\textsc{pub}),k_\textsc{sec}) = x$.

A cryptosystem is \emph{homomorphic} in some operation $\blacksquare$
if it is possible to perform another (possibly different) operation
$\square$ such that: $\mathcal{E}(x, k_\textsc{pub}) \; \square \;
\mathcal{E}(x, k_\textsc{pub}) = \mathcal{E}(x \; \blacksquare \; y,
k_\textsc{pub})$. Finally, in this work we assume all data to be
binary $\in \{0,1\}$. For more detailed background on FHE beyond what
is described below, see the excellent tutorial of \citet{H:2017}.

\subsection{Fully Homomorphic Encryption}
\label{ssec:fhe}
In 1978, cryptographers posed the question: \emph{Does an encryption
scheme exist that allows one to perform arbitrary computations on
encrypted data?} The implications of this, called a \emph{Fully
Homomorphic Encryption} (FHE) scheme, would enable clients to send
computations to the cloud while retaining control over the secrecy of
their data.
This was still an open problem however 30 years later. Then, in 2009,
a cryptosystem \citep{Gen:2009} was devised that could, in principle,
perform such computations on encrypted data. Similar to previous
approaches, in each computation, noise is introduced into the
encrypted data. And after a certain number of computations, the noise
grows too large so that the encryptions can no longer be decrypted.
The key innovation was a technique called \emph{bootstrapping}, which
allows one to reduce the noise to its original level without
decrypting. 

At a high level, the idea is as follows. Assume the cryptosystem can evaluate a version of its decryption function where the secret key is also encrypted: $\mathcal{D}(\mathcal{E}(x, k_\textsc{pub}),\mathcal{E}(k_\textsc{sec},k_\textsc{pub}))$. \citet{Gen:2009} showed that if the secret key was newly encrypted then the output of this function would be an encrypted version of data $x$ \emph{with all of the noise removed}. Finally, also assume this cryptosystem can homomorphically evaluate a \texttt{NAND} gate on encrypted data. %
Then because via \texttt{NAND} one can express all possible logical operations, such a cryptosystem would be fully homomorphic. \citet{Gen:2009} subsequently derived such a scheme, the first FHE scheme.
That result constituted a massive breakthrough, as it established, for
the first time, a fully homomorphic encryption scheme
\citep{Gen:2009}. 
Unfortunately, the original bootstrapping procedure was highly
impractical, %
as massive noise was introduced in each homomorphic operation. 
\begin{figure*}[t!]
    \centerline{\includegraphics[width=0.75\textwidth]{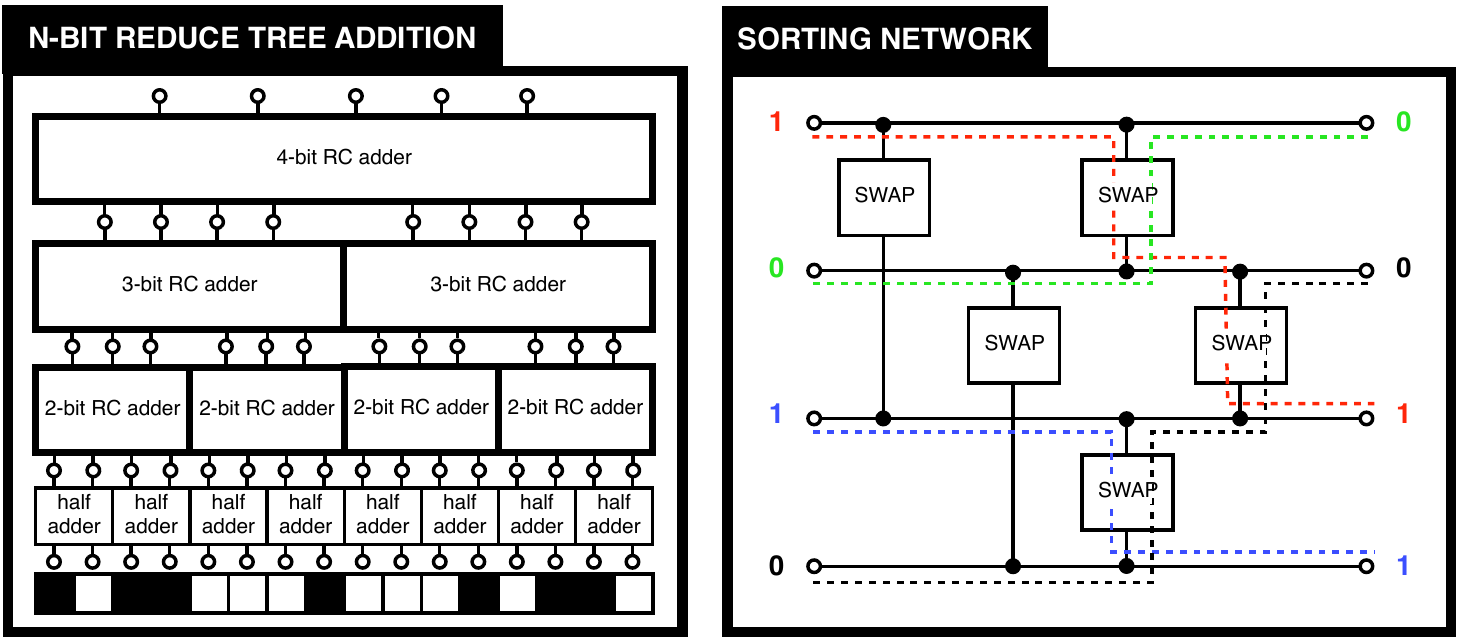}}
  \caption[Binary Circuits for reduce-tree and sorting networks]{Binary circuits used for inner product: reduce tree
  (\emph{Left}) and sorting network (\emph{Right}). RC is short for
  ripple-carry.}
  \label{figure.binary_ops}
\end{figure*}
Consequently, much of the research since the first FHE scheme has been
devoted to reducing the growth of noise so that the scheme never has
to perform bootstrapping. Indeed, even in recent FHE schemes
bootstrapping is slow 
(roughly six minutes in a highly optimised implementation of a recent popular scheme \citep{HS:2015}) and bootstrapping many times increases the memory requirements of encrypted data. 

\subsubsection{Encrypted prediction with levelled HE}
Thus, one common technique to implement encrypted prediction was to take an
existing ML algorithm and approximate it with as few operations as possible, to
never have to bootstrap. This involved careful parameter tuning to ensure that
the security of the encryption scheme was sufficient, that it didn't require too
much memory, and that it ran in a reasonable amount of time. One prominent
example of this is Cryptonets \citep{G-BDL+:2016}. If a practitioner wanted to
add a few layers to the neural network model, they would need to ensure that all
of the operations could still be performed without bootstrapping. Otherwise,
security parameters or previous layers would need to be adjusted to account for
the added layers.

\subsubsection{Encrypted prediction with FHE}
Recent developments in cryptography call for rethinking this approach.
\citet{DM:2015} devised a scheme that that could bootstrap a single
Boolean gate in under one second with reduced memory. Recently,
\citet{CGGI:2016} introduced optimisations implemented in the TFHE
library, which further reduced bootstrapping to under 0.1 seconds.
In this chapter, we demonstrate that this change has a huge impact on
designing encrypted machine learning algorithms. Specifically,
encrypted computation is now modular: the cost of adding a few layers
to an encrypted neural network is simply the added cost of each layer
in isolation. This is particularly important as recent developments in
deep learning such as Residual Networks \citep{HZRS:2016} and Dense
Networks \citep{HZWV:2017} have shown that networks with many layers
are crucial to achieving state-of-the-art accuracy.

\subsection{Binary Neural Networks}
\label{ssec:binary}
The cryptosystem that we will use in this chapter, TFHE, is however restricted
to computing binary operations. We note that, concurrent to the work that led to
TFHE, was the development of neural network models that perform binary
operations between binary weights and binary activations. These models, called
Binary Neural Networks (BNNs), were first devised by~\citet{KS:2015}
and~\citet{CHSEB:2016}, and were motivated by the prospect of training and
testing deep models on limited memory and limited compute devices, such as
mobile phones.

\paragraph{Technical details.}
We now describe the technical details of binary networks that we will
aim to replicate on encrypted data. In a \emph{Binary Neural Network}
(BNN) every layer maps a binary input $\mathbf{x} \in \{-1,1\}^{d}$ to
a binary output $\mathbf{z} \in \{-1,1\}^p$ using a set of binary
weights $\mathbf{W} \in \{-1,1\}^{(p,d)}$ and a binary activation
function $\texttt{sign}(\cdot)$ that is $1$ if $x \geq 0$ and $-1$
otherwise. 
 Although binary nets do not typically use a bias term, applying
 batch-normalization~\citep{IS:2015} when evaluating the model means that a
 bias term $\vec{b} \in \bZ^p$ may need to be added before applying the
 activation function (cf. Sec.~\ref{subsubsec:bn}). Thus, when evaluating the
 model, a fully connected layer in a BNN implements the following transformation
 $\mathbf{z} := \mathrm{sign}(\mathbf{W}\mathbf{x} + \vec{b})$. From now on we
 will call all data represented as $\{-1,1\}$ \emph{non-standard binary} and
 data represented as $\{0,1\}$ as \emph{binary}. \citet{KS:2015,CHSEB:2016} were
 the first to note that the above inner product nonlinearity in BNNs could be
 implemented using the following steps:

\begin{enumerate}
	\item Transform data and weights from non-standard binary to
		binary: $\mathbf{w},\mathbf{x} \rightarrow
		\overline{\mathbf{w}},\overline{\mathbf{x}}$ by replacing $-1$
		with $0$.  Now all data is in $\{0,1\}$.
	\item Apply element-wise multiplication by using the logical \texttt{XNOR}
		operator$(\overline{\mathbf{w}},\overline{\mathbf{x}})$ for each element
		of $\overline{\mathbf{w}}$ and $\overline{\mathbf{x}}$.
		
	\item Sum the result of the  previous step by using \texttt{popcount}
		operation (which counts the number of 1s), call this $S$.
	\item If the bias term is $b$, check if $2 S \geq d - b$, if so
		the activation is positive and return $1$, otherwise return
		$-1$.
\end{enumerate}

Thus we have that,
\begin{align*}
	z_i = \sgn{2 \cdot \mathrm{popcount}(\mathrm{XNOR}(\overline{\mathbf{w}}_{i},\overline{\mathbf{x}})) - d + b}
\end{align*}

\paragraph{Related binary models.}
Since the initial work on BNNs there has been a wealth of work on
binarising, ternarising, and quantizing neural networks
\citet{CWTWC:2015,CBD:2015,HMD:2016,HCSEB:2016,ZHMD:2016,CWMMP:2017,CHZX:2017}.
Our approach is currently tailored to methods that have binary
activations and we leave the implementation of these methods on
encrypted data for future work.

\section{Methods}
\label{sec:methods}
In this work, we observe that BNNs can be run on encrypted data by designing
circuits in TFHE for computing their operations. We consider boolean circuits
that operate on encrypted data and unencrypted weights and biases. We show how
these circuits allow us to efficiently implement the three main layers of binary
neural networks: fully connected, convolutional, and batch-normalization. We
then show how a simple trick that allows us to sparsify our computations.  Our
techniques can be easily parallelised. During the evaluation of a circuit, gates
at the same level in the tree representation of the circuit can be evaluated in
parallel. Hence, when implementing a function, ``shallow'' circuits are
preferred in terms of parallelisation. While parallel computation was often used
to justify employing the second generation FHE techniques---where
parallelisation comes from ciphertext packing---we show in the following section
that our techniques create dramatic speedups for a state-of-the-art FHE
technique. We emphasise that a key challenge is that we need to use \emph{data
oblivious} algorithms (circuits) when dealing with encrypted data as the
algorithm never discovers the actual value of any query made on the data.

\subsection{Binary OPs}
\label{ssec:binaryops}
The three primary circuits we need are for the following tasks: 1.
computing the inner product; 2. computing the binary activation
function (described in the previous section) and; 3. dealing with the
bias.

\subsubsection{Encrypted inner product}
As described in the previous section, BNNs can speed up an inner
product by computing \textsc{XNOR}s (for element-wise multiplication)
followed by a \textsc{popcount} (for summing). In our case, we compute
an inner product of size $d$ by computing \textsc{XNOR}s element-wise
between $d$ bits of encrypted data and $d$ bits of unencrypted data,
which results in an encrypted $d$ bit output. To sum this output, the
\textsc{popcount} operation is useful when weights and data are
unencrypted because \textsc{popcount} is implemented in the
instruction set of Intel and AMD processors, but when dealing with
encrypted data we simply resort to using shallow circuits. We consider
two circuits for summation, both with sublinear depth: a reduce tree
adder and a sorting network.

\paragraph{Reduce tree adder.}
We implement the sum using a binary tree of half and ripple-carry (RC)
adders organised into a reduction tree, as shown in
Figure~\ref{figure.binary_ops} (\emph{Left}). All these structures can
be implemented to run on encrypted data because TFHE allows us to
compute \textsc{XNOR}, \textsc{AND}, and \textsc{OR} on encrypted
data. The final number returned by the reduction tree $\tilde{S}$ is
the binary representation of the number of $1$s resulting from the
$\textsc{XNOR}$, just like \textsc{popcount}. Thus, to compute the BNN
activation function $\sgn{\cdot}$ we need to check whether
$2\tilde{S} \geq d - b$, where $d$ is the number of bits in
$\tilde{S}$ and $b$ is the bias. Note that if the bias is zero we
simply need to check if $\tilde{S} \geq d/2$. To do so we can simply
return the second-to-last bit of $\tilde{S}$. If it is $1$ then
$\tilde{S}$ is at least $d/2$. If the bias $b$ is non-zero (because of
batch-normalization, described in Section~\ref{subsubsec:bn}), we can
implement a circuit to perform the check $2\tilde{S} \geq d - b$. The
bias $b$ (which is available in the clear) may be an integer as large
as $\tilde{S}$. Let $\mathbb{B}[(d-b)/2]$ and $\mathbb{B}[\tilde{S}]$ be
the binary representations of $\br{d-b}/2$ and $\tilde{S}$ respectively.
Algorithm~\ref{alg:compare} describes a comparator circuit that
returns an encrypted value of $1$ if the above condition holds and
(encrypted) $0$ otherwise (where $\textsc{MUX}(s,a,b)$ returns $a$ if
$s=1$ and $b$ otherwise). As encrypted operations dominate the running
time of our computation, in practice this computation essentially
corresponds to evaluating $d$ MUX gates. This gate has a dedicated
implementation in TFHE, which results in a very efficient comparator
in our setting.

\begin{algorithm}[htb]
  \caption[Comparator Circuit]{Comparator}
  {\bf Inputs:}~~Encrypted $\mathbb{B}[\tilde{S}]$, unencrypted
  $\mathbb{B}[(d-b)/2]$, size $d$ of
  $\mathbb{B}[(d-b)/2]$,$\mathbb{B}[\tilde{S}]$ \\
  {\bf Output:}~~Result of $2\tilde{S} \geq d - b$
  \begin{algorithmic}[1]
  \label{alg:compare}
  	\STATE $o = 0$ \FOR{$i = 1,\ldots,d$} \IF{$\mathbb{B}[(d-b)/2]_i =
    0$} 
    \STATE $o = \textsc{MUX}(\mathbb{B}[\tilde{S}]_i, \tilde{1},
    o)$ 
    \ELSE 
    \STATE $o = \textsc{MUX}(\mathbb{B}[\tilde{S}]_i, o,
    \tilde{0})$ 
    \ENDIF 
    \ENDFOR 
    \STATE {\bf Return:} $o$
  \end{algorithmic}
\end{algorithm}

\paragraph{Sorting network.}
We do not technically care about the sum of the result of the element-wise
$\textsc{XNOR}$ between $\bar{\mathbf{w}}$ and $\bar{\mathbf{x}}$. In fact, all
we care about is if the result of the comparison: $2\tilde{S} \geq d - b$. Thus,
another idea is to take the output of the (bitwise) $\textsc{XNOR}$ and sort it.
Although this sorting needs to be performed over encrypted data, the rest of the
computation does not require any homomorphic operations; after sorting we hold a
sequence of encrypted $1$s, followed by encrypted $0$s. To output the correct
value, we only need to select one of the (encrypted) bit in the correct position
and return it. If $b=0$ we can simply return the encryption of the central bit
in the sequence; indeed, if the central bit is $1$, then there are more $1$s
than $0$s and thus $2\tilde{S} \geq d$ and we return $1$. If $b\neq0$ we need to
offset the returned index by $b$ in the correct direction depending on the sign
of $b$. To sort the initial array we implement a sorting network, shown
in Figure~\ref{figure.binary_ops} (\emph{Right}). The sorting network is a
sequence of swap gates between individuals bits, where $\textsc{SWAP}(a,b) =
(\textsc{OR}(a,b), \textsc{AND}(a,b))$. Note that if $a \geq b$ then
$\textsc{SWAP}(a,b) = (a,b)$, and otherwise is $(b,a)$. More specifically, we
implement Batcher's sorting network~\cite{batcher_sorting_1968}, which consists
of $O(n \log^2(n))$ swap gates, and has depth $O(\log^2(n))$.

\subsubsection{Batch normalization}
\label{subsubsec:bn}

Batch normalization is mainly used during training; however, during the
evaluation of a model, this requires us to scale and translate and scale the input
(which is the output of the previous layer). In practice, when our activation
function is the $\sgn{\cdot}$ function, this only means that we need to update
the bias term (the actual change to the bias term is an elementary calculation).
As our circuits are designed to work with a bias term, and the scaling and
translation factors are available as plaintext (as they are part of the model),
this operation is easily implemented during test time.

\begin{figure}[t!]
   \centering
   \includegraphics[width=0.5\linewidth]{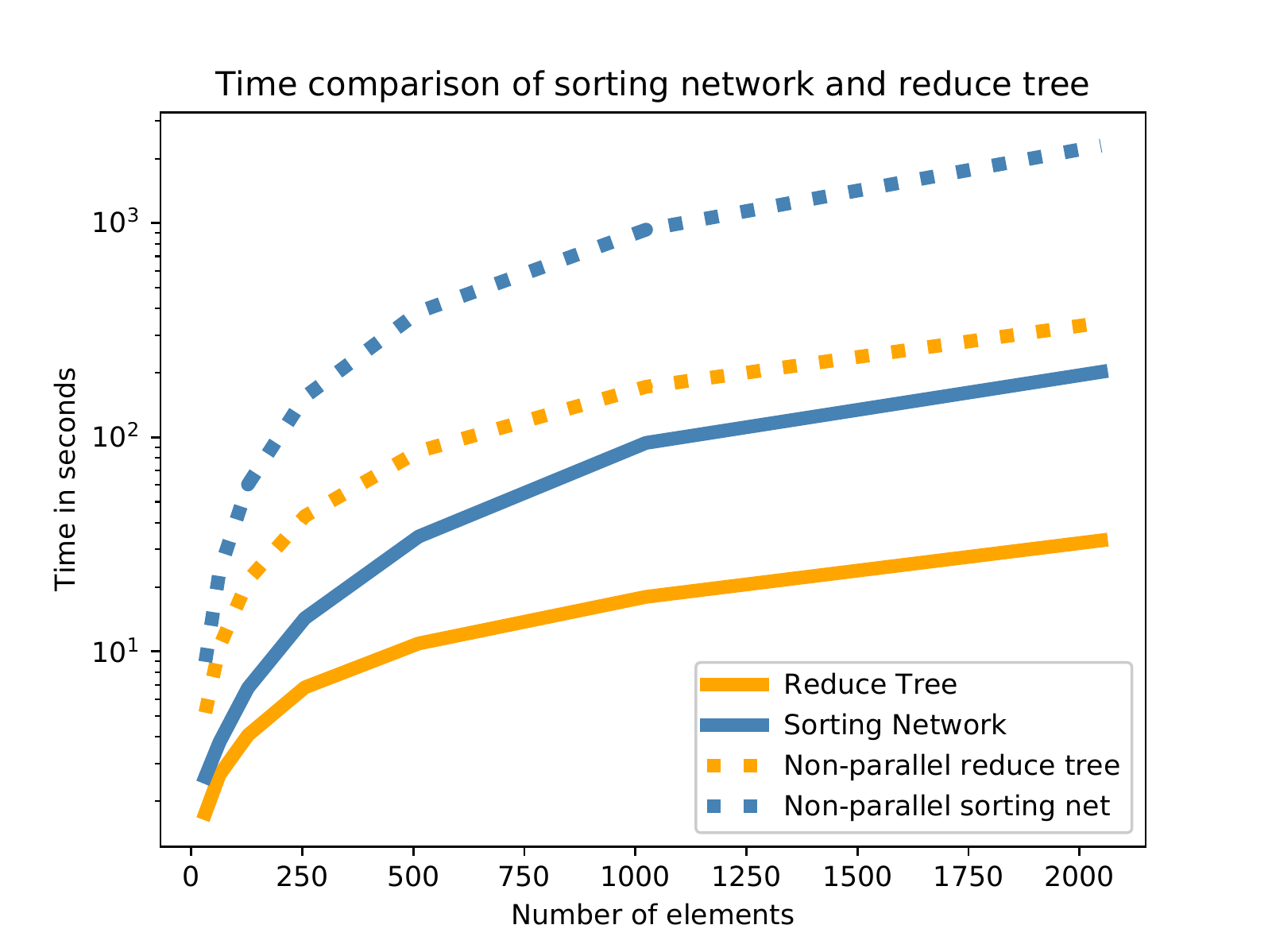}
   \caption[Timing of sorting network and reduce tree addition]{Timing of sorting network and reduce tree addition for different sized vectors, with and without parallelisation.}
   \label{Figure.timings}
 \end{figure}

\subsection{Sparsification via ``+1''-trick}
\label{ssec:methods-sparsification}

Since we have access to the weight matrix $\mathbf{W} \in \{-1, 1\}^{p \times d}$ and
the bias term $\vec{b} \in \bZ^p$ in the clear (only data $\x$ and
subsequent activations are encrypted), we can exploit the fact that
$\mathbf{W}$ always has values $\pm 1$ to, roughly, halve the cost
computation. We consider $\w \in \{-1, 1\}^d$ which is a single row of
$\mathbf{W}$ and observe that:
\begin{align}
	\w^\top\x = (\vec{1} + \w)^\top (\vec{1} + \x) - \sum_{i} w_i - (\vec{1} + \x)^\top \vec{1}, \nonumber
\end{align}
where $\vec{1}$ denotes the vector in which every entry is $1$. Further note
that $(\vec{1} + \w) \in \{0,2\}^{d}$ which means that the product $(\vec{1} +
\w)^\top (\vec{1} + \x)$ is simply the quantity $4 \sum_{i: w_i = 1} \bar{x}_i$,
where $\bar{\x}$ refers to the standard binary representation of the
non-standard binary $\x$. Assuming at most half of the $w_i$s were originally
$+1$, if $w \in\{-1, 1\}^d$, only $d/2$ encrypted values need to be added. We
also need to compute the encrypted sum $\sum_{i} x_i$; however, this latter sum
need only be computed once, no matter how many output units the layer has. Thus,
this small bit of extra overhead roughly {\em halves} the amount of computation
required. We note that if $\w$ has more $-1$s than $+1$s, $\w^\top \x$ can be
computed using $(\vec{1} - \w)$ and $(\vec{1} - \x)$ instead. This guarantees
that we never need to sum more than half the inputs for any output unit. The
sums of encrypted binary values can be calculated as described in
Sec.~\ref{ssec:binaryops}. The overheads are two additions required to compute
$(\vec{1} + \x)^\top \vec{1}$ and $(\vec{1} - \x)^\top \vec{1}$, and then a
subtraction of two $\log(d)$-bit long encrypted numbers.  (The multiplication by
$2$ or $4$ as may be sometimes required is essentially free, as bit shifts
correspond to dropping bits, and hence do not require homomorphic operations).
As our experimental results show this simple trick roughly halves the
computation time of one layer; the actual savings appear to be even more than
half as in many instances the number of elements we need to sum over is
significantly smaller than half.

It is worth emphasizing the advantage of binarising and then using
the above approach to making the sums sparse. By default, units in a
neural network compute an affine function to which an activation
function is subsequently applied. The affine map involves an inner
product that involves $d$ multiplications. Multiplication under fully
homomorphic encryption schemes is, however, significantly more expensive
than addition. By binarising and applying the above calculation, we've
replaced the inner product operation by selection (which is done in
the clear as $\mathbf{W}$ is available in plaintext) and (encrypted)
addition.

\subsection{Ternarisation (Weight Dropping)}
Ternary neural networks use weights in $\{-1, 0, 1\}$ rather than
$\{-1, 1\}$; this can alternatively be viewed as dropping connections
from a BNN. Using ternary neural networks rather than binary reduces
the computation time as encrypted inputs for which the corresponding
$w_i$ is $0$ can be safely dropped from the computation, before the
method explained in section~\ref{ssec:methods-sparsification} is
applied to the remaining elements. Our experimental results show that
a binary network can be ternarised to maintain the same level of test
accuracy with roughly a quarter of the weights being $0$ (cf.
Sec.~\ref{ssec:accuracy}).

\section{Experimental Results}
\label{sec:results}
In this section, we report encrypted binary neural network prediction
experiments on several real-world datasets. We begin by comparing the efficiency
of the two circuits used for inner product, the reduce tree and the sorting
network. We then describe the datasets and the architecture of the BNNs used for
classification. We report the classification timings of these BNNs for each
dataset, for different computational settings. Finally, we give accuracies of
the BNNs compared to floating-point networks. Our code is freely available
at~\cite{tapas}.

\subsection{Reduce tree vs. sorting network}
We show timings of reduce tree and sorting network for a different
number of input bits, with and without parallelisation in
Figure~\ref{Figure.timings} (parallelisation is over 16 CPUs). We
notice that the reduce tree is strictly better when comparing parallel
or non-parallel timings of the circuits. As such, from now on we use
the reduce tree circuit for inner product.

It should be mentioned that at the outset this result was not obvious because
while sorting networks have more levels of computation, they have fewer gates.
Specifically, the sorting network used for encrypted sorting is the bitonic
sorting network which for $n$ bits has $O(\log^2 n)$ levels of computation
whereas the reduce tree only has $O(\log n)$ levels. On the other hand, the
reduce tree requires $2$ gates for each half adder and $5k$ gates for each
$k$-bit RC adder, whereas a sorting network only requires $2$ gates per
\textsc{SWAP} operation. Another factor that may slow down sorting networks is
that our implementation of sorting networks is recursive, whereas the reduce
tree is iterative.

\paragraph{Datasets}
We evaluate on four datasets, three of which have privacy implications
due to health care information (datasets Cancer and Diabetes) or
applications in surveillance (dataset Faces). We also evaluate on the
standard benchmark MNIST dataset. All of the datasets and the corresponding model architectures are described in~\Cref{sec:expr-settings}.

\subsection{Timing}
\label{sec:parall-strat}
We give timing results for the classification of an instance in different
computational settings. 
All of the strategies use the parallel implementations of the reduce tree
circuit computed across 16 CPUs (the solid line orange line in
Figure~\ref{Figure.timings}). The \textit{Out Seq} strategy computes each
operation of a BNN sequentially (using the parallel reduce tree circuit). Notice
that for any layer of a BNN mapping $d$ inputs to $p$ output nodes, the
computation over each of the $p$ output nodes can be parallelised. From this, we
can easily estimate the effect of further parallelisation over the outputs of
BNN layers, as the encrypted computation will remain identical. The \textit{Out
16-P} strategy estimates parallelizing the computation of the $p$ output nodes
across a cluster of $16$ machines (each with $16$ CPUs). The \textit{Out Full-P}
strategy estimates complete parallelisation, in which each output node can be
computed independently on a separate machine. We note that for companies that
already offer prediction as a service, both of these parallel strategies are not
unreasonable requirements. Indeed it is not uncommon for such companies to run
hundreds of CPUs/GPUs over multiple days to tune hyperparameters for deep
learning models\footnote{https://tinyurl.com/yc8d79oe}. Additionally, we report
how timings change with the introduction of the \textit{+1-trick} is described
in Section \ref{ssec:methods-sparsification}. 
\begin{table}[h!]\centering
  \begin{tabular}[h!]{lrrrr}
  \toprule
    {\bf Parallelism} & {\bf Cancer} & {\bf Diabetes} & {\bf Faces} & {\bf MNIST} \\\midrule
    Out Seq & 3.5s & 283s & 763.5h & 65.1h\\%\hline
    +1-trick & 3.5s & 250s & 564h & 37.22 h \\%\hline
    \begin{tabular}{@{}c@{}} Out 16-P.\\ +1 trick\end{tabular}& 3.5s & 31.5 s & 33.1h&  2.41 h \\%\hline
    Out Full-P&3.5s&29s&1.3h& 147s\\%\hline
    \bottomrule
  \end{tabular}
  \caption[Neural Network timings for different techniques of parallelism]{Neural Network timings on various datasets using different forms of parallelism. \label{tbl:EXP.TIME}}
\end{table}
These timings are given in Table~\ref{tbl:EXP.TIME} (computed with Intel Xeon
CPUs @ 2.40GHz, processor number E5-2673V3). We notice that without
parallelisation over BNN outputs, the predictions on datasets that use fully
connected layers: Cancer and Diabetes, finish within seconds or minutes. While
for the datasets that use convolutional layers: Faces and MNIST, predictions
require multiple days. The \textit{+1-trick} cuts the time of MNIST prediction
by half and reduces the time of Faces prediction by $200$ hours. With only a bit
of parallelism over outputs (\textit{Out 16-Parallel}) prediction on the Faces
dataset now requires less than 1.5 days and MNIST can be done in $2$ hours. With
complete parallelism (\textit{Out N-Parallel}) all methods reduce to under $2$
hours.

\subsection{Accuracy}
\label{ssec:accuracy}
We wanted to ensure that BNNs can still achieve similar test-set accuracies to
floating-point networks. To do so, for each dataset we construct similar
floating-point networks. For the Cancer dataset, we use the same network except
we use the original $30$ real-valued features, so the fully connected layer is
$30 \rightarrow 1$, as was used in \citet{MLH:2018}. For Diabetes and Faces,
just like for our BNNs we cross-validate to find the best networks (for Faces:
$4$ convolutional layers, with filter sizes of $5 \times 5$ and $64$ output
channels; for Diabetes the best network is the same as used in the BNN). For
MNIST we report the accuracy of the best performing method as reported
in~\citet{WZZ+:2013}\footnote{https://tinyurl.com/knn2434}. Additionally, we
report the accuracy of the weight-dropping method described in
Section~\ref{sec:methods}.
\begin{table}[h!]\centering
  \begin{tabular}[h!]{lrrrr}
  \toprule
     & {\bf Cancer} & {\bf Diabetes} & {\bf Faces} & {\bf MNIST} \\
     \midrule
    Floating & 0.977 & 0.556 & 0.942 & 0.998  \\%\hline
    BNN & 0.971 & 0.549 & 0.891  & 0.986  \\%\hline
    \begin{tabular}{@{}l@{}} BNN \\ drop $10\%$ \end{tabular}& 0.976 & 0.549 & 0.879 &  0.976 \\%\hline
    \begin{tabular}{@{}l@{}} BNN \\ drop $20\%$ \end{tabular}&0.912 & 0.541 & 0.878 & 0.973\\
    \bottomrule
  \end{tabular}
	\caption[Accuracy of floating-point networks compared with BNNs]{The accuracy of floating-point networks compared with BNNs, with and without weight dropping. The Cancer dataset floating-point accuracy is given by \cite{MLH:2018}, the MNIST floating-point accuracy is given by \cite{WZZ+:2013}, and the MNIST BNN accuracy (without dropping) is given by \cite{CHSEB:2016}.\label{tbl:ACC}}
\end{table}
The results are shown in Table~\ref{tbl:ACC}. We notice that apart from the
Faces dataset, the difference in accuracies between the floating-point networks
and BNNs are at most $1.2\%$ (on MNIST). The face dataset uses a different
network in floating-point which seems to be able to exploit the increased
precision to increase accuracy by $5.1\%$. We also observe that weight dropping
by $10\%$ reduces the accuracy by at most $1.2\%$ (on Faces). Dropping $20\%$ of
the weights seem to have a small effect on all datasets except Cancer, which has
only a single layer and so likely relies more on every individual weight.

\newpage
\phantomsection
\addcontentsline{toc}{chapter}{\normalfont\bfseries\numberline{}Epilogue\protect}%
\chapter*{Epilogue}
This thesis looks at four different concepts of reliability in the context of
deep neural networks --- generalisation, adversarial robustness, calibration,
and privacy. As various existing
works~\citep{Zhang2016,SzegedyVISW16,Guo2017,G-BDL+:2016}, as well as this
thesis, point out, these problems seem to have arisen due to deep neural
networks getting larger, more {\em complex}, and also, more accurate. The
ability of modern deep neural networks to memorise noise makes it difficult to
certify their generalisability~(\Cref{chap:stable_rank_main}) and it provably
makes them more vulnerable to adversarial attacks~(\Cref{chap:causes_vul}). The
large capacity of these networks, which allows them to memorise label noise,
also makes them prone to NLL-overfitting~(\Cref{chap:focal_loss}) and to
miscalibration. Further, as we see in~\Cref{chap:TAPAS}, the network
architectures of these deep neural networks are ill-suited to satisfy the EPAAS
constraints. Thus, the same large capacity which allows them to obtain high test
accuracy also results in these side-effects, hurting their reliability.

This poses a deeper question as to whether these quantities of reliability are
fundamentally at odds with accuracy. This question has been addressed
theoretically in several works, some of which we discuss in~\Cref{sec:trml},
especially for adversarial robustness. Some works have suggested that to get a
low adversarial error, we need more
data~\citep{schmidt2018adversarially,montasser19a} or more
computation~\citep{degwekar19a,blpr18}. In~\Cref{chap:causes_vul}, we show
theoretical examples where we do not need either of these; choosing the
inductive biases carefully suffices for robust
generalisation.~\Cref{chap:low_rank_main} shows that a low rank prior over the
representation space~(using the \lr) achieves a small adversarial error
without suffering in test accuracy. 

We show similar results for the other notions of reliability. In particular, we
show that by using proper regularisations and network architectures, we can
obtain less memorisation of label noise~(SRN in~\Cref{chap:stable_rank_main}),
more calibrated models~(Focal loss in~\Cref{chap:focal_loss}), and privacy
guarantees for user data~(BNNs in~\Cref{chap:TAPAS}) without sacrificing
accuracy or needing more data or computation. Interestingly, both SRN and \lr
regularises variants of the rank operator. While SRN controls a soft version of
the rank of the parameters, the \lr controls the hard rank of the
representations. While existing works mostly look at norm-based regularisations,
our work suggests that a rank-like measure, which controls the effective degree
of freedom, is more effective.

All of these works show that choosing the right inductive biases, by way of
regularisations and network architectures, has a significant impact on obtaining
a simultaneously reliable and accurate model, with limited data and computation.
However, we have not completely minimised the apparent tradeoff yet. For
example, the adversarial robustness of our models in~\Cref{chap:low_rank_main}
is not as high as the models trained with \AT~(which albeit has a lower
accuracy). Assuming that this tradeoff should not exist as we, humans, do not
suffer from it, it is an open question to determine whether we need more data,
more computation, better architectures, or perhaps all of these to completely
mitigate the tradeoff. 

Requiring more data or computation is often considered undesirable, but we
should perhaps recall the progress that has been made, along this direction, in
the last decade, and be hopeful that this will be a much lesser constraint in
the coming years with the rapid advances in computer architecture and
crowdsourcing services. Finally, I would like to end by remarking that as we
develop new and more powerful neural network architectures and learning
algorithms, we should be benchmarking them not just on traditional metrics of
performance like accuracy, but also on metrics of reliability such as those discussed in this thesis.

\appendix
\chapter{Mathematical prerequisites}
\label{sec:math-prer-notat}
This Appendix will define some mathematical preliminaries that are used in the
thesis and recall some things we already know in Linear Algebra, Statistics and
Probability. While most of the things in this appendix is fairly basic, it has
been included for the sake of completeness and to avoid looking up other
resources.

\section*{Basic Linear Algebra}
\label{sec:basic-linear-algebra}

We start by assuming that the reader has a basic knowledge of vector spaces and
linear operators on the elements of  vector spaces. As this is not the main
topic of discussion, we only skim over topics that are indispensable to the
understanding of this document. For a broader introduction to the basics of
Linear Algebra, the reader is directed to~\citet{strang1993introduction}.

Specifically, a vector is an element of a vector space and a matrix is a linear
operator on vectors\footnote{Only true for finite vector spaces but we will
avoid the mathematical complexities for simplicity.} i.e. it transfers vectors
from one vector space to another. In most cases, we will deal with the vector
space $\reals^d$ where $d$. 
Wherever the dimension of the matrix~(or the domain
of the linear transformation) is not mentioned, it should be clear
from the context.

\paragraph{Eigenvalues and Eigenvectors}

If $\vec{W}$ is a square matrix and $\vec{x}$ is  a non-zero vector, such that
the vector $\vec{W}\vec{x}$ is a scalar multiple of $\vec{x}$, i.e.
$\vec{W}\vec{x} = \lambda\vec{x}$, then $\vec{x}$ is an eigenvector of $\vec{A}$
and $\lambda$ is the corresponding eigenvalue. However, eigenvectors and
eigenvalues only exist for square matrices. Singular Value Decomposition or SVD
is a decomposition for any general rectangular matrix $\vec{A}$,
$\mathrm{SVD}\br{\vec{A}}=\vec{U}\vec{\Sigma} \vec{V}^\top$ where $\vec{U}$ and
$\vec{V}$ are unitary matrices~(i.e. $\vec{U}^\top\vec{U} = I$) and
$\vec{\Sigma}$ is a non-negative diagonal matrix. Its diagonal entries are
usually arranged in descending order and are known as singular values. We use
$\lambda_i(\vec{A})$ and $\sigma_i(\vec{A})$ to denote the $i^{\it{th}}$ largest
eigenvalue and singular value of $\vec{A}$ respectively. Specifically,
\(\lambda_{\mathrm{max}}\br{\vec{A}}\) and
\(\lambda_{\mathrm{min}}\br{\vec{A}}\) represents the largest and smallest
eigenvalues of $\vec{A}$, and \(\sigma_{\mathrm{max}}\br{\vec{A}}\) and
\(\sigma_{\mathrm{min}}\br{\vec{A}}\) denotes the largest and smallest singular
values of $\vec{A}$.  

\paragraph{Vector norms}

For a vector \(\vec{x}\in\reals^d\), $\norm{\vec{x}}_p$ represents its $\ell_p$
norm, defined as $\norm{\vec{x}}_p = \sqrt[p]{\sum_{i=1}^{d}x_i^p}$ where
\(\vec{x}_i\) indexes the \(i^{\it th}\) element of \(\vec{x}\).  By default,
$\norm{\vec{x}}$ represents the $\ell_2$ norm of $\vec{x}$. \(\norm{\cdot}_0\)
is not strictly a norm but it often referred to as the \(\ell_0\) norm and
measures the sparsity of its argument vector.

\paragraph{Matrix operator norms}

The most common matrix norms are the induced operator norms. If
$\norm{\cdot}_{\alpha}$ is a vector norm  then the induced operator norm is
defined as
\begin{equation}\label{defn:induced_norm} \norm{\vec{A}}_{\alpha} =
\sup_{x\neq\vec{0}}
\dfrac{\norm{\vec{A}\vec{x}}_{\alpha}}{\norm{\vec{x}}_{\alpha}}
\end{equation} Perhaps, the most common operator norm is the \(\ell_2\) operator norm, also known as the spectral norm and is induced by \(\norm{\cdot}_2\). In~\Cref{chap:stable_rank_main}, we discuss~\gls{lip}ness using the \(\alpha,\beta\)-operator norm which is defined as 

\begin{equation}\label{defn:p_q-induced_norm} \norm{\vec{A}}_{\alpha,\beta}^{\mathrm{op}} =
  \sup_{x\neq\vec{0}}
  \dfrac{\norm{\vec{A}\vec{x}}_{\beta}}{\norm{\vec{x}}_{\alpha}}
  \end{equation}
The notation with \(\mathrm{op}\) in the super-script is necessary to differentiate it from the entry-wise \(p,q\) matrix norm defined below.

\paragraph{\(\ell_p\)-schatten norms}
The $\ell_p$ schatten norm is applied on a matrix by applying the corresponding
$\ell_p$ norm on the vector of singular values of the matrix. We do not use this
norm in this thesis to avoid confusion with other matrix norms. However, it is
interesting to note that the \(\ell_2\) operator norm defined above is equivalent to the \(\ell_\infty\)-schatten norm. The \(\ell_2\)-schatten norm is the \(\ell_2\) norm of the singular values and is also known as the Frobenius norm~(\(\norm{\cdot}_\forb\) of the matrix. The rank of a matrix, which is equal to the number of non-zero singular values can be thought of as the \(\ell_0\)-schatten norm though neither the rank nor the \(\ell_0\) {\em norm} is a true norm.

\paragraph{Entry-wise matrix norms}
Finally, the last kind of matrix norms we look at are the entry-wise norms. The
$\ell_p$ entry-wise norms are applied on a matrix by unrolling the matrix as one
long vector and applying the norm on that vector.  The most common entry-wise norms for matrices is the \(\ell_2\) entry-wise norm, also known as the Frobenius norm, and is denoted as \(\norm{\cdot}_\forb\). The Frobenius norm of a matrix \(\vec{A}\) is 
\begin{equation}\label{eq:frob-norm}
  \norm{\vec{A}}_\forb = \sqrt{\sum_{i}\sum_j \vec{A}_{i,j}^2}
\end{equation}
where  \(\vec{A}_{i,j}\) represents the $\br{i,j}^{\it{th}}$ entry of the matrix
\(\vec{A}\). \Cref{{eq:defn-ins-spec-compl}}, uses a slightly different variant of the entry-wise norm known as the \(p,q\) entry-wise norm

\begin{equation}\label{eq:entry-wise-pq-norm} \norm{\vec{A}}_{p,q} =
\sqrt[q]{\sum_{j=1}^n\br{\sqrt[p]{\sum_{i=1}^m\br{\vec{A}_{i,j}}^p}}^q}
\end{equation}

To summarize, we use the following norms frequently in this thesis. The spectral
norm of a matrix \(\vec{A}\) is represented by both $\norm{\vec{A}}$ and
$\norm{\vec{A}}_2$~(its operator norm notation). $\norm{\vec{A}}_F$ represents
the Frobenius norm of a matrix $\vec{A}$. $\rank{\vec{A}}$ denotes the rank of
$\vec{A}$ i.e. the number of non-zero singular values of $\vec{A}$ or the
$\ell_0$ schatten norm of $A$\footnote{$\ell_0$ norms are not true norms.
Mathematically, this is a misuse of notation.}. The \(p,q\) operator norm is
denoted as \(\norm{\cdot}_{p,q}^{\mathrm{op}}\) and the \(p,q\) entry-wise norm
is denoted as \(\norm{\cdot}_{p,q}\).

\paragraph{Positive semi-definiteness of matrices}

 An important kind of matrix that often appears in learning are
 matrices $\vec{M}$ such that $\forall x,~ x^\top Mx ~\& 0$ where
 $\&\in\bc{<, \le, >, \ge }$, Such a matrix is called negative
 definite, negative semi-definite, positive definite and positive
 semi-definite respectively. Clearly, a matrix with all positive
 eigenvalues is positive definite~(PD), a matrix with all non-negative
 eigenvalues is positive semi-definite~(PSD) and similarly for negative
 definite~(ND) and negative semi-definite matrix~(NSD).

A commonly used PSD matrix is the outer-product matrix. For any vector
\(\vec{z}\in\reals^{d\times 1}\), the outer product matrix \(\vec{z} \vec{z}^\top\) is PSD. This can be shown as follows 
 \begin{equation}\label{lem:outer-prod-psd}
\forall \vec{x}\in\reals^{d\times 1},\quad \vec{x}^\top\vec{z}\vec{z}^\top\vec{x} = \ip{\vec{z}^\top\vec{x}}{\vec{z}^\top,\vec{x}} = \norm{\vec{z}^\top\vec{x}}_2^2\ge 0
 \end{equation}

\todo[color=red]{Maybe worth writing a section on Lagrangian}

\section*{Common Inequalities}
\label{sec:basic-inequalities-1}
The following inequalities are used in various parts of the thesis and are listed here for ease of reference.

\begin{ineq}\label{lem:exp_ineq} For any \(x\in\reals,\)
  \[ 1 + x\le e^x\]
\end{ineq}
While this can be easily proved through differentiation, a proof
without using differentiation is presented below.

\begin{proof}
  We know that $e^x$ can be expanded as follows \[e^x = 1 + x + \sum_{k=2}^{\infty}\frac{x^k}{k!}\] We will look at the following three cases.
  \begin{itemize}
  \item $x\le - 1$ We know $e^x$ is always positive and in this case $1+x$ is non-positive. Hence, the inequality trivially holds.
  \item $x\ge 0 $ By the expansion above, we can write $e^x = 1 + x + p$, where $p$ is some non-negative term. Hence, the inequality trivially holds.
  \item $x\in\br{-1, 0}$ Rewriting the expansion as follows completes the proof. \[e^x - (1 + x) = \sum_{k=1}^\infty \frac{x^{2k}}{2k!} + \frac{x^{2k+1}}{\br{2k+1}!}  = \sum_{k=1}^\infty \frac{x^{2k}}{\br{2k+1}!} \br{2k+1 - |x| } \ge 0  \]
  \end{itemize}
\end{proof}

\begin{ineq}[Union bound]\label{ineq:union-bound}
  Consider any finite or countable collection of events denoted as \(\br{A_1,A_2,A_3\ldots,}\), then the probability of occurring of at least one of the events from that collection is no more than the sum of the individual probabilities of all the events in that collection i.e.
  \[\bP\bs{\bigcup_i A_i }\le \sum_i \bP\bs{A_i}\]
\end{ineq} 
\todo[color=green]{Make the notation of probability consistent \(\bP\) and not \(P\)}
\begin{ineq}[Jensen's Inequality]\label{ineq:ineq-jensen} For a convex
  real-valued function $f:\reals^n\rightarrow \reals$ and an integrable function
  \(g\) defined on \(\bs{a,b}\), such that \(f\) is defined on, at least, the
  co-domain of \(g\), the following holds ---
  \[f\br{\frac{1}{b-a}\int_{a}^b g}\le\frac{1}{b-a}\int_{a}^{b} g\br{f} \] 
\end{ineq}

\begin{ineq}[Holder's Inequality and the Cauchy Schwartz
  Inequality]\label{lem:holders_inequality} Let \(\vec{f},\vec{g}\) be
  real-valued vectors and consider any $p,q \in [1,\infty)$ such that
  $\frac{1}{p} + \frac{1}{q} = 1$, then
  \[ \abs{\ip{\vec{f}}{\vec{g}}} \le \norm{\vec{f}}_p\norm{\vec{g}}_q\] For
  $p=q=2$, this reduces to the Cauchy-Schwartz inequality
  \[ \abs{\ip{\vec{f}}{\vec{g}}}\le \norm{\vec{f}}\norm{\vec{g}}|\]
\end{ineq}

\begin{ineq}[Markov's Inequality]\label{ineq:markov} For a non-negative
  random variable $X$  with its mean $\mu = \bE\bs{X}$ and a positive
  number $a$,
  \[\bP\bs{X\ge a}\le \dfrac{\mu}{a}\]
\end{ineq}

\begin{ineq}[Chernoff Bound]\label{ineq:hoeffding} Suppose
  $X_1\cdots X_m$ are $m$ $\bc{0,1}$-valued independent random with $\bP\bs{X_i
  = 1} = p_i$ where $0\le p_i \le 1$. Denote $X = \sum_{i=1}^m X_i  $
  and $\mu = \bE\bs{X} = \sum_{i=1}^mp_i$.  Then
  \(\forall\delta,~0\le\delta\le 1\) the Chernoff bound states that
  
  \[ \bP \bs{X  \ge \br{1+\delta}\mu }\le
  \exp{\br{-\frac{\delta^2\mu }{\br{3}}}}\]
  \[ \bP\bs{X  \le  \br{1- \delta}\mu} \le
  \exp{\bs{-\frac{\delta^2\mu }{2}}}\] 
  
  Combining them together, we can write
  
  \[ \bP\bs{\abs{X -\mu} \ge   \delta\mu} \le
  2\exp{\br{-\frac{\delta^2\mu }{3}}}\] 
  
\end{ineq}

\chapter{Common datasets and neural network architectures}
\label{sec:expr-settings}
\section{Datasets}
We use the following image and document classification datasets in our experiments:
\begin{enumerate}[leftmargin=*]
\item \textbf{MNIST} ~\citep{LBBH:1998}: This dataset has $60,000$ training and
$10,000$ test images. Each image has a dimension of 28x28. This dataset is used
in~\Cref{chap:causes_vul,chap:TAPAS}. For experiments in~\Cref{chap:TAPAS}, we
use the model (torch7) described in \cite{CHSEB:2016} whereas the model used
in~\Cref{chap:causes_vul} is described in the chapter itself. 

\item \textbf{CIFAR-10} \citep{krizhevsky2009learning}: This dataset
has 60,000 colour images of size $32 \times 32$, divided equally into
10 classes. For all experiments other than the ones in~\Cref{chap:focal_loss}, we use the standard train/test split of $50,000/10,000$ images. In~\Cref{chap:focal_loss}, we use a train/validation/test split of
$45,000/5,000/10,000$ images. Furthermore, we augment the training images by applying
random crops and random horizontal flips. This dataset is used in the experiments for~\Cref{chap:stable_rank_main,chap:causes_vul,chap:low_rank_main,chap:focal_loss}

\item \textbf{CIFAR-100} \citep{krizhevsky2009learning}: This dataset has 60,000
colour images of size $32 \times 32$, divided equally into 100 fine classes.
There is also a fixed partitioning of the $100$ classes into $20$ coarse classes.
Note that the images in this dataset are not the same images as in CIFAR-10. We
also augment the training images by applying random crops and random horizontal
flips.  We again use a train/validation/test split of 45,000/5,000/10,000
images. For all experiments other than the ones in~\Cref{chap:focal_loss}, we
again use the standard train/test split of $50,000/10,000$ images.
In~\Cref{chap:focal_loss}, we use a train/validation/test split of
$45,000/5,000/10,000$ images. This dataset is used in the experiments
for~\Cref{chap:stable_rank_main,chap:causes_vul,chap:low_rank_main,chap:focal_loss}

\item \textbf{SVHN}~\citep{Netzer2011}: The Street View House Number or SVHN dataset has $73,257$ digits for training, $26,032$ digits for testing, and $531,131$ additional easier samples, to be used as extra training data. There are $10$ classes like MNIST and CIFAR10. We use this dataset in~\Cref{chap:low_rank_main}.

\item \textbf{Restricted Imagenet Settings}~\citep{tsipras2018robustness}
Restricted-Imagenet is a subset of ImageNet with 64 x 64 dimensional
images. There are $60$ fine classes and $10$ coarse classes with each coarse
class having $6$ distinct fine classes in them. The train set size is 77237 and
the test-set size is 3000. The fine classes within each coarse are balanced i.e.
given a coarse class all the fine classes in it are equally represented in this
dataset. This dataset is used in the experiments for~\Cref{chap:causes_vul}.

\begin{table}[!htb]\renewcommand{\arraystretch}{1.2}
    \begin{tabular}[!htb]{|l@{\quad}l@{}|}\toprule
        Coarse Class& Fine Classes\\\midrule
    Dog&Chihuahua, Japanese spaniel, Maltese dog, Pekinese,\\ & Shih-Tzu,Blenheim spaniel\\ 
    Bird&cock, hen, ostrich, brambling, goldfinch, house finch\\ 
    Insect&tiger beetle,ladybug,ground beetle, long-horned beetle, \\ &leaf
    beetle, dung beetle\\ 
    Monkey&guenon, patas, baboon, macaque, langur, colobus\\ 
    Car&jeep, limousine,cab, beach wagon, ambulance, convertible\\ 
    Feline&leopard, snow leopard, jaguar, lion, cougar, lynx\\ 
    Truck&tow truck, moving van, fire engine,  garbage truck,\\& pickup,police van\\ 
    Fruit&Granny Smith, rapeseed, corn, acorn, hip, buckeye\\ 
    Fungus&gyromitra,  hen-of-the-woods, coral fungus,stinkhorn,\\ & agaric, earthstar\\ 
    Boat&gondola, fireboat, speedboat, lifeboat, yawl, canoe\\ 
    \bottomrule
    \end{tabular}
    \caption{Fine-grained classes in Restricted Imagenet}
    \label{tab:fine-grained-classs}
    \end{table}

\item \textbf{Tiny-ImageNet} \citep{imagenet_cvpr09}: Tiny-ImageNet is
a subset of ImageNet with 64 x 64 dimensional images, 200 classes and
500 images per class in the training set and 50 images per class in
the validation set. The image dimensions of Tiny-ImageNet are twice
that of CIFAR-10/100 images. 

 For Tiny-ImageNet, we train for 100 epochs with a learning rate of 0.1 for the
first 40 epochs, 0.01 for the next 20 epochs and 0.001 for the last 40 epochs.
We use a training batch size of 64. It should be noted that for Tiny-ImageNet,
we saved 50 samples per class (i.e., a total of 10000 samples) from the training
set as our own validation set to fine-tune the temperature parameter (hence, we
trained on 90000 images) and we use the Tiny-ImageNet validation set as our test
set. This dataset is used in~\Cref{chap:focal_loss}.

\item \textbf{CelebA}~\citep{liu2015faceattributes} The CelebA dataset contains $202,599$ coloured images scaled to a size of $64\times64$. We use this dataset for the experiments on image generation in~\Cref{chap:stable_rank_main}.

\item \textbf{20 Newsgroups} \citep{Lang1995}: This dataset contains
20,000 news articles, categorised evenly into 20 different newsgroups
based on their content. It is a popular dataset for text
classification. Whilst some of the newsgroups are very related (e.g.\
rec.motorcycles and rec.autos), others are quite unrelated (e.g.\
sci.space and misc.forsale). We use a train/validation/test split of
15,098/900/3,999 documents.

On this dataset, we train the Global Pooling Convolutional Network
\citep{Lin2014} using the Adam optimiser, with learning rate $0.001$, and betas
$0.9$ and $0.999$.\footnote{The code is a PyTorch adaptation of~\url{https://github.com/aviralkumar2907/MMCE}}. We
used Glove word embeddings \citep{Pennington2014} to train the network. We
trained all the models for 50 epochs and used the models with the best
validation accuracy. This dataset is used in~\Cref{chap:focal_loss}.

\item \textbf{Stanford Sentiment Treebank (SST)} \citep{Socher2013}: This dataset contains movie reviews in the form of sentence parse trees, where each node is annotated by sentiment. We use the dataset version with binary labels, for which 6,920/872/1,821 documents are used as the training/validation/test split. In the training set, each node of a parse tree is annotated as positive, neutral or negative. At test time, the evaluation is done based on the model classification at the root node, i.e.\ considering the whole sentence, which contains only positive or negative sentiment. 

On this dataset, we train the Tree-LSTM~\citep{Tai2015} using the AdaGrad
optimiser with a learning rate of $0.05$ and a weight decay of $10^{-4}$, as
suggested by the authors. We used the constituency model, which considers binary
parse trees of the data and trains a binary Tree-LSTM on them. The Glove word
embeddings \citep{Pennington2014} were also tuned for best results. The code
framework we used is inspired by \cite{TreeLSTM}. We trained these models for 25
epochs and used the models with the best validation accuracy. This dataset is
used in~\Cref{chap:focal_loss}.

\item \textbf{Cancer}~\citep{Dua:2019}%
The Cancer dataset\footnote{https://tinyurl.com/gl3yhzb} contains $569$ data
points where each point has $30$ real-valued features. The task is to predict
whether a tumor is malignant (cancerous) or benign. We use this dataset
in~\Cref{chap:TAPAS}. Similar to \citet{MLH:2018} we divide the dataset into a
training set and a test in a $70:30$ ratio. For every real-valued feature, we
divide the range of the feature into three equal-spaced bins and one-hot encode
each feature by its bin-membership. 
This creates a $90$-dimensional binary vector for each example. We use
a single fully connected layer $90 \rightarrow 1$ followed by a batch
normalization layer, as is common practice for BNNs \cite{CHSEB:2016}.

\textbf{Diabetes}~\citep{Strack2014}%
This dataset\footnote{https://tinyurl.com/m6upj7y} contains data on
$100,000$ patients with diabetes.
The task is to predict one of three possible labels regarding hospital
readmission after release: 1. a patient is readmitted to the hospital after
release in less than or equal to $30$ days; 2. readmission happens after $30$
days;3. a patient is not readmitted. We use this dataset in~\Cref{chap:TAPAS}.
We divide patients into a $80/20$ train/test split. We ensure that the same
patient does not appear in both the training and test dataset by splitting patients
who appear more than once by putting them in either the training or the test dataset
in a $80:20$ split ratio. As this dataset contains real and categorical
features, we bin them as in the Cancer dataset. We obtain a $1,704$ dimensional
binary data point for each entry. Our network (selected by cross validation)
consists of a fully connected layer $1704 \rightarrow 10$, a batch normalization
layer, a \textsc{sign} activation function, followed by another fully connected
layer $10 \rightarrow 3$, and a batch normalization layer.

\item \textbf{Labeled Faces in the Wild-a}~\citep{LFWTech,Wolf2011}
The \textit{Labeled Faces in the Wild-a} dataset contains $13,233$ gray-scale
face images. We use the binary classification task of gender identification from
the images. We use this dataset in~\Cref{chap:TAPAS}. We use the \textit{LFW-a}
version~\footnote{https://www.openu.ac.il/home/hassner/data/lfwa/} of the
dataset where the images are aligned using a commercial alignment software and
are grayscale. We resize the images to size $50\times 50$. Our network
architecture (selected by cross-validation) contains $5$ convolutional layers,
each of which is followed by a batch normalization layer and a $\textsc{sign}$
activation function (except the last which has no activation). All convolutional
layers have unit stride and filter dimensions $10\times 10$. All layers except
the last layer have $32$ output channels (the last has a single output channel).
The output is flattened and passed through a fully connected layer $25
\rightarrow 1$ and a batch normalization layer.

\end{enumerate}

For all our experiments, we use the PyTorch framework, setting any
hyperparameters not explicitly mentioned to the default values used in the
standard models.

\section{Neural network architectures}

We use the following neural network architectures throughout the thesis.
\begin{enumerate}
    \item \textbf{ResNet-18/50/110} The ResNet-18/50/100 are standard $18,50,$
    and $110$ layered ResNets respectively with batch Norm and ReLU.  Each of
    these networks have a convolution layer followed by four residual blocks and
    a final fully connected layer. The depth of these blocks vary depending on
    the total number of layers. 
     
    When using these networks to learn CIFAR10 and CIFAR100, we train with
    stochastic gradient descent for a total of $350$ epochs with an initial
    learning rate of $0.1$, which is multiplied $0.1$ after $150$ and $250$
    epochs respectively, a weight decay of $5e-4$ and a momentum of $0.9$. We use ResNet-18 in~\Cref{chap:causes_vul,chap:low_rank_main}, ResNet-50 in~\Cref{chap:causes_vul,chap:low_rank_main,chap:focal_loss}, and ResNet-100 in~\Cref{chap:stable_rank_main,chap:focal_loss}

    \item \textbf{WideResNet-28/26-10} We use a standard WideResNet with $28$
    and $26$ layers and a growth factor of $10$. In total, the network has
    36,539,124 trainable parameters.  The network is the standard configuration
    with batchnorm and ReLU activations and is trained with a weight decay of
    $1e-4$. The learning rate was multiplied by $0.2$ after $60,120$, and $160$
    epochs respectively. We use WideResNet-28-10 in~\Cref{chap:stable_rank_main}
    and WideResNet-26-10 in~\Cref{chap:focal_loss}.

    \item \textbf{Densenet-100/121} The DenseNet-100 is a standard $100$-layered
    densenet with Batchnorm and ReLU and has a total of $800,032$ trainable
    parameters. We train the network on CIFAR10/100 for a total of $350$ epochs
    with SGD using an initial learning rate of $0.1$, which is multiplied by
    $0.1$ after $150$ and $250$ epochs respectively, a weight decay of $1e-4$,
    and a momentum of $0.9$. We use Densenet-100 in~\Cref{chap:stable_rank_main} and Densenet-121 in~\Cref{chap:causes_vul,chap:focal_loss}.
    
    \item \textbf{VGG19} The VGG19 model is the standard 19-layered VGG model
    with Batchnorm and ReLU.  It has three fully connected~(FC)
    layers after sixteen convolution layers. It has a total of $20,548,392$
    trainable parameters and is trained with SGD with a momentum of $0.9$ and a
    weight decay of $5e-4$. The initial learning rate is $0.1$ and is multiplied
    by $0.1$ after $150$ and $250$ epochs respectively. We use this network for
    experiments
    in~\Cref{chap:stable_rank_main,chap:causes_vul}. For the
    shattering experiments in~\Cref{chap:stable_rank_main}, we used the same
    architecture and the same training recipe except the initial learning rate,
    which was deceased to $0.01$ as the model failed to learn the random labels
    with a large learning rate.
    
    \item \textbf{AlexNet} The Alexnet model is the standard ALexNet model with
    $4,965,092$ trainable parameters. It was trained with SGD, with a
    momentum of $0.9$, with an initial learning rate is $0.01$, which is
    multiplied by $0.1$ after $150$ and $250$ epochs respectively. The
    optimiser was further augmented with a weight decay rate of $5e-4$. We use this network for the experiments in~\Cref{chap:causes_vul}.

    \item \textbf{GAN Architecture} The model architecture for  both the
    generator and the discriminator was chosen to be a 32 layered
    ResNet~\citep{HZRS:2016}~(similar to ResNet-18 above) due to its previous
    superior performance in other works~\citep{miyato2018spectral}.  We use Adam
    optimiser~\citep{kingma2014adam} which depends on three main
    hyper-parameters $\alpha$- the initial learning rate, $\beta_1$- the first
    order moment decay rate and $\beta_2$- the second order moment decay rate.
    We cross-validate these parameters in the set $\alpha\in\{0.0002,
    0.0005\},~\beta_1\in\{0, 0.5\},~\beta_2\in\{0.9, 0.999\}$ and chose
    $\alpha=0.0002$, $\beta_1=0.0$ and $\beta_2=0.999$ which performed
    consistently well in all of the experiments.
    
    \item \textbf{Neural Divergence Setup} We train a new classifier inline with
    the architecture in~\citet{gulrajani2018towards}. It includes three
    convolution layers with 16, 32 and 64 channels, a kernel size of $5\times~5$
    and a stride of $2$. Each of these layers are followed by a Swish
    activation~\citep{ramachandran2018searching} and then finally a linear layer
    that gives a single output. The network is initialised using normal
    distribution with zero mean and the standard deviation of $0.02$, and
    trained using Adam optimiser with $\alpha=0.0002,~\beta_1=0.,~\beta_2=0.9$
    for a total of $100,000$ iterations with mini-batch of $128$ generated
    samples and $128$ samples from the test-set\footnote{For CelebA, we used the
    training set.}. We use the standard WGAN-GP loss function, $\log\br{1 +
    \exp\br{f\br{\vec{x}_{\mathrm{fake}}}}} + \log{\br{1 +
    \exp\br{-\vec{x}_{\mathrm{real}}}}}$, where $f$ represents the network
    described above. Finally, we generate $1~\mathrm{Million}$ samples from the
    generator and report the average $\log\br{1 +
    \exp\br{f\br{\vec{x}_{\mathrm{fake}}}}}$ over these samples. Higher average
    value implies better generation as the network in this case is unable to
    distinguish the generated and the real samples.

\end{enumerate}
\clearpage
\onehalfspacing
\chapter{Appendix for~\Cref{chap:stable_rank_main}}
\label{app:stable_rank}
\section{Proofs for~\Cref{sec:lipschitz}}
\label{sec:lipsch-proof}
\begin{proof}[Proof of~\Cref{lem:upper-bound-nn-lip}]
  As mentioned in~\cref{eq:lipLocal}, local lipschitzness of a function $f$ at
  $\vec{x}$ is equal to the jacobian at that point i.e. $L_l(\vec{x}) =
  \norm{J_f(\vec{x})}_{p,q}^{\mathrm{op}}$

  \begin{align}
      \label{eq:jacboianNN}
      J_f(\vec{x}) = \frac{\partial f\br{\vec{x}}}{\partial \vec{x}} := \frac{\partial \vec{z}_1}{\partial \vec{x}}\frac{\partial \phi_1 (\vec{z}_1)}{\partial \vec{z}_1} \cdots  \frac{\partial \vec{z}_L}{\partial \vec{a}_{L-1}}\frac{\partial \phi_L (\vec{z}_L)}{\partial \vec{z}_L}.
  \end{align}
  Using \( \frac{\partial \vec{z}_i}{\partial \vec{a}_{i-1}} = \vec{W}_i \) where
  \( \vec{z}_i \) is the pre-activation vector of the \(i^{\it th}\) layer,
  \(\vec{W}_i\) represents the linear operator of the \(i^{\it th}\) layer, and
  \(\vec{a}_i=\phi_i\br{\vec{z}_i}\) is the post-activation vector of the \(i^{\it
  th}\) layer. By sub-multiplicativity of the matrix norms:
  \begin{align}
      \label{eq:lipboundNN}
      \norm{J_f(\vec{x})}_{p,q}^{\mathrm{op}} \leq \norm{\vec{W}_1}_{p,q}^{\mathrm{op}} \norm{\frac{\partial \phi_1 (\vec{z}_1)}{\partial \vec{z}_1}} \cdots \norm{\vec{W}_L}_{p,q}^{\mathrm{op}} \norm{\frac{\partial \phi_L (\vec{z}_l)}{\partial \vec{z}_L}}.
  \end{align}
  Note, most commonly used activation functions $\phi(.)$ such as ReLU, sigmoid, tanh and maxout are known to have \Gls{lip} constant less than or equal to $1$
  Thus, the upper bound can further be written only using the operator norms of the intermediate matrices as 
  \begin{align}\label{eq:lip_upp_bnd_app}
      L_l(\vec{x}) \leq \norm{J_f(\vec{x})}_{p,q}^{\mathrm{op}} \leq \norm{\vec{W}_L}_{p,q}^{\mathrm{op}} \cdots \norm{\vec{W}_1}_{p,q}^{\mathrm{op}}
  \end{align}%
  As this upper bound is independent of the data point \(\vec{x}\), this is also equal to the local lipschitz constant of $f$.
  \begin{align*}
    L_l = \mathrm{max}_{\vec{x}} L_l\br{\vec{x}} \le  \norm{\vec{W}_L}_{p,q}^{\mathrm{op}} \cdots \norm{\vec{W}_1}_{p,q}^{\mathrm{op}}  
  \end{align*}
  \end{proof}

  \begin{proof}[Proof of~\Cref{prop:empiricalLocalLip}]
    Let $f:\reals^d\rightarrow \reals$ be a differentiable function on an
    open set containing $\vec{x}_i$ and $\vec{x}_j$ such that
    $\vec{x}_i\neq\vec{x}_j$. By applying fundamental theorem of calculus
    \begin{align}
      \abs{f\br{\vec{x}_i} - f\br{\vec{x}_j}} &= \abs{\int_{0}^1\nabla f\br{\vec{x}_i+\theta\br{\vec{x}_j-\vec{x}_i}}^\top\br{\vec{x}_j-\vec{x}_i}\partial\theta}  \nonumber \\
                                        &\le\overset{(a)}{\le} \int_{0}^1\abs{\nabla f\br{\vec{x}_i+\theta\br{\vec{x}_j-\vec{x}_i}}^\top\br{\vec{x}_j-\vec{x}_i}}\partial\theta  \nonumber\\
                                        &\overset{(b)}{\le} \int_{0}^1\norm{\nabla f\br{\vec{x}_i+\theta\br{\vec{x}_j-\vec{x}_i}}}_q\norm{\br{\vec{x}_j-\vec{x}_i}}_p\partial\theta  \nonumber\\
                                        &\le \int_{0}^1\max_{\theta\in\br{0, 1}}\norm{\nabla f\br{\vec{x}_i+\theta\br{\vec{x}_j-\vec{x}_i}}}_q\norm{\br{\vec{x}_j-\vec{x}_i}}_p\partial\theta  \nonumber \\
                                        &=\max_{\theta\in\br{0, 1}}\norm{\nabla f\br{\vec{x}_i+\theta\br{\vec{x}_j-\vec{x}_i}}}_q\norm{\br{\vec{x}_j-\vec{x}_i}}_p \int_{0}^1\partial\theta  \nonumber \\
    \therefore \dfrac{\abs{f\br{\vec{x}_i} -   f\br{\vec{x}_j}}}{\norm{\br{\vec{x}_j-\vec{x}_i}}_p}&\le\max_{\theta\in\br{0, 1}}\norm{\nabla f\br{\vec{x}_i+\theta\br{\vec{x}_j-\vec{x}_i}}}_q = \max_{\vec{x}\in \textit{Conv}\;(\vec{x}_i, \vec{x}_j)}\norm{\nabla f\br{\vec{x}}}_q. \nonumber
    \end{align}
    The inequality (a) is due to  Jensen's Inequality~(\Cref{ineq:ineq-jensen})
    and inequality (b) is due to H\"{o}lder's
    inequality~(\Cref{lem:holders_inequality}).
    \end{proof}

\section{Proofs for~\Cref{sec:optimalSrank}}
\label{sec:srankProof}
\subsection*{Proof for Optimal Stable Rank Normalization.~(Main Theorem)}
\begin{proof} 
Here we provide the proof of~\Cref{thm:srankOptimal} for all the three cases with optimality and uniqueness
   guarantees. Let $\widehat{\vec{W}}_k$ be the optimal solution to
   the problem for any of the two cases.
   From~\Cref{lem:opt-frobenius}, the $\mathrm{SVD}$ of $\vec{W}$ and
   $\widehat{\vec{W}}_k$ can be written as
   $\vec{W}=\vec{U}\Sigma\vec{V}^\top$ and
   $\widehat{\vec{W}}_k=\vec{U}\Lambda\vec{V}^\top$, respectively.
   Then, \( L = \norm{\vec{W} - \widehat{\vec{W}}_k}_{\forb}^2 =
   \ip{\Sigma - \Lambda}{\Sigma - \Lambda}_{\forb} \). From now
   onwards, we denote $\Sigma$ and $\Lambda$ as vectors consisting of
   the diagonal entries, and $\ip{.}{.}$ as the vector inner product
   \footnote{$\ip{.}{.}_\forb$ represents the Frobenius inner product
   of two matrices, which in the case of diagonal matrices is the same
   as the  inner product of the diagonal vectors.}. 
\paragraph{Proof for Case (a):}
In this case, there is no constraint enforced to preserve any of the singular values of the given matrix while obtaining the new one. The only constraint is that the new matrix should have the stable rank of $r$. Let us assume $\Sigma = \br{\sigma_1, \cdots, \sigma_p}$, $\Sigma_2 = \br{\sigma_2, \cdots, \sigma_p}$, $\Lambda = \br{\lambda_1, \cdots, \lambda_p}$ and $\Lambda_2 = \br{\lambda_2, \cdots, \lambda_p}$. Using these notations, we can write $L$ as:
\begin{align}
    L &=\ip{\Sigma}{\Sigma} +  \ip{\Lambda}{\Lambda} - 2\ip{\Sigma}{\Lambda}\nonumber\\
    \label{eq:l1}
    &= \ip{\Sigma}{\Sigma} +  \lambda_1^2 + \ip{\Lambda_2}{\Lambda_2} - 2\sigma_1\lambda_1 - 2\ip{\Sigma_2}{\Lambda_2}
  \end{align}

\noindent Using the stable rank constraint
$\srank{\widehat{\vec{W}}_k} = r$, which is \(r = 1 +
\dfrac{\sum_{j=2}^p\lambda_j^2}{\lambda_1^2} \).

\paragraph{Case for $\mathbf{r>1}$} If $r>1$ we obtain the following equality constraint, making the
problem non-convex.

\begin{align}
\label{eq:lambda1}
\lambda_1^2 = \dfrac{\ip{\Lambda_2}{\Lambda_2}}{r - 1}
\end{align} 
However, we will show that the solution we obtain is optimal and
unique. Substituting~\cref{eq:lambda1} into~\cref{eq:l1} we get
  \begin{align}
  \label{eq:l2}
    L = \ip{\Sigma}{\Sigma} + \dfrac{\ip{\Lambda_2}{\Lambda_2}}{r - 1}  + \ip{\Lambda_2}{\Lambda_2} -  2\sigma_1\sqrt{\dfrac{\ip{\Lambda_2}{\Lambda_2}}{r - 1}} -  2\ip{\Sigma_2}{\Lambda_2}
  \end{align}
   
Setting $\dfrac{\partial L}{\partial \Lambda_2} = 0$ to get the family of critical points
  \begin{align}
    & \dfrac{2\Lambda_2}{r - 1}  + 2\Lambda_2 - \dfrac{ 4\sigma_1\Lambda_2}{2\sqrt{\br{r - 1}\ip{\Lambda_2}{\Lambda_2}}} -  2\Sigma_2\nonumber = 0\\
    & \implies \Sigma_2 = \Lambda_2\br{\dfrac{1}{r - 1} + 1 - \dfrac{
      \sigma_1}{ 1\sqrt{\br{r - 1}\ip{\Lambda_2}{\Lambda_2}}}
      }\label{eq:scalar_mult}\\
      &\implies \dfrac{\Sigma_2\bs{i}}{\lambda_2\bs{i}} = \br{\dfrac{1}{r - 1} + 1 - \dfrac{ \sigma_1}{ 1\sqrt{\br{r -
        1}\ip{\Lambda_2}{\Lambda_2}}}} = \dfrac{1}{\gamma_2} \quad \forall~1\le i\le p\label{eq:Scalar_mult}
  \end{align}
As the R.H.S. of~\ref{eq:Scalar_mult} is independent of $i$, the above
equality implies that all the critical points of~\cref{eq:l2} are a
scalar multiple of $\Sigma_2$, implying,  $\Lambda_2 = \gamma_2
\Sigma_2$. Note that the domain of $\Lambda_2$ are all strictly
positive vectors and thus, we can ignore the critical point at
$\Lambda_2 = \vec{0}$. Substituting this into~\cref{eq:scalar_mult} we
obtain
  \begin{align*}
    & \Sigma_2 = \gamma_2\Sigma_2\br{\dfrac{1}{r - 1} + 1 - \dfrac{ \sigma_1}{ \gamma_2\sqrt{\br{r - 1}\ip{\Sigma_2}{\Sigma_2}}} }
  \end{align*}
\noindent Using the fact that $\ip{\Sigma_2}{\Sigma_2} =
\norm{\vec{S}_2}_{\forb}^2$ in the above equality and with some
algebraic manipulations, we obtain \( \gamma_2  = \frac{\gamma + r -
1}{r} \) where, $\gamma = \frac{\sqrt{r-1}
\sigma_1}{\norm{\vec{S}_2}_\forb}$. Note, $r \geq 1$, $\gamma \geq 0$,
and $\Sigma \geq 0$, implying, $\Lambda_2 = \gamma_2 \Sigma_2 \geq 0$.

\paragraph{Local minima:} Now, we will show that $\Lambda_2$ is indeed
a minima of~\cref{eq:l2}. To show this, we  compute the hessian of
$L$. Recall that
\begin{align*}
  \dfrac{\partial L}{\partial \Lambda} &=    \dfrac{2r}{r - 1}\Lambda
                                         -\dfrac{
                                         2\sigma_1\Lambda}{\sqrt{\br{r
                                         - 1}\norm{\Lambda}_2^2}} -
                                         2\Sigma_2\\ 
  \vec{H}=\dfrac{\partial^2 L}{\partial^2 \Lambda} &=    \dfrac{2r}{r -
                                            1}\vec{I}  -\dfrac{
                                            2\sigma_1}{\sqrt{\br{r -
                                            1}}\norm{\Lambda}_2^2}\br{\norm{\Lambda}_2
                                            \vec{I} - \dfrac{1}{\norm{\Lambda}_2}\Lambda\Lambda^\top}\\
                &=  2\br{\dfrac{r}{r-1} - \dfrac{
                                            \sigma_1\norm{\Lambda}_2}{\sqrt{\br{r -
                                            1}}\norm{\Lambda}_2^2}
                  }\vec{I}+
                  \dfrac{2\sigma_1}{\sqrt{r-1}\norm{\Lambda}_2^3}\br{\Lambda\Lambda^\top}\\
\end{align*}
Now we need to show that $\vec{H}$ at the solution $\Lambda_2$ is PSD i.e.
$\forall~\vec{x}\in\reals^{p-1},\enskip  \vec{x}^\top\vec{H\br{\Lambda_2}} \vec{x}\ge
0$
\begin{align*}
  \vec{x}^\top\vec{H} \vec{x} &=  2\br{\dfrac{r}{r-1} - \dfrac{
                                            \sigma_1\norm{\Lambda}_2}{\sqrt{\br{r -
                                            1}}\norm{\Lambda}_2^2}
  }\norm{\vec{x}_2^2}+
  \dfrac{2\sigma_1}{\sqrt{r-1}\norm{\Lambda}_2^3}\vec{x}^\top\br{\Lambda\Lambda^\top}\vec{x}\\
                              &\stackrel{(a)}{\ge} 2\br{\dfrac{r}{r-1} - \dfrac{
                                            \sigma_1\norm{\Lambda}_2}{\sqrt{\br{r -
                                            1}}\norm{\Lambda}_2^2}
     }\norm{\vec{x}}_2^2 \\
                              &\stackrel{(b)}{\ge} 2\br{\dfrac{r}{r-1} - \dfrac{
                                            \sigma_1}{\br{r -
                                            1}\lambda_1}
     }\norm{\vec{x}}_2^2
                              \stackrel{(c)}{=}  \dfrac{2r}{r-1}\br{ 1 - \dfrac{
                                            \gamma }{\br{\gamma+r - 1}}
                                }\norm{\vec{x}}_2^2   \\
  &\stackrel{(d)}{\ge}  \dfrac{2r}{r-1}\br{ 1 - \dfrac{
                                            1 }{\br{1+r - 1}}
     }\norm{\vec{x}}_2^2 = 2\norm{\vec{x}}_2^2\ge 0
\end{align*}
Here $(a)$ is due to the fact that the matrix $\Lambda\Lambda^\top$ is an outer
product matrix and is hence PSD~(see~\Cref{lem:outer-prod-psd}). $(b)$ follows
due to~\cref{eq:lambda1} and $(c)$ follows by substituting $\lambda_1 =
\gamma_1\sigma_1$ and then the value of $\gamma_1$. Finally $(d)$ follows as
$\br{ 1 - \dfrac{\gamma }{\br{\gamma+r - 1}}}$ is decreasing with respect to
$\gamma$ and we know that $\gamma< 1$ due to the assumption that
$\srank{\vec{W}}< r$. Thus, we can substitute $\gamma = 1$ to find the minimum
value of the expression. This concludes our proof that $\Lambda_2$ is indeed a
local minima of $L$.

  \paragraph{Uniqueness:}   The uniqueness of $\Lambda_2$  as a
  solution to ~\cref{eq:l2} is shown
in~\Cref{lem:uniqueness} and is also guaranteed by the fact that
$\gamma_2$ has a unique value. Using $\Lambda_2 = \gamma_2 \Sigma_2$ and
$\lambda_1 = \gamma_1\sigma_1$ in~\cref{eq:lambda1}, we obtain a
unique solution $\gamma_1 = \frac{\gamma_2}{\gamma}$.

Now, we need to show that it is also an unique solution
to~\Cref{thm:srankOptimal}.

For all solutions to~\Cref{thm:srankOptimal} that have singular
vectors which are different than that of $\vec{W}$,
by~\Cref{lem:opt-frobenius}, the matrix formed by replacing the
singular vectors of the solution with that of $\vec{W}$ is also a
solution. Thus, if there were a solution with different singular
values than $\widehat{\vec{W}}_k$, it should have appeared as a solution
to~\cref{eq:l2}. However, we have shown that~\cref{eq:l2} has a unique
solution.

 Now, we need to show that among all matrices with the same singular
 values as that of $\widehat{\vec{W}}_k$,  $\widehat{\vec{W}}_k$ is
 strictly better  in terms of   $\norm{\vec{W} - \widehat{\vec{W}}_k}$
 . This requires a further assumption that every non-zero singular
 value of $\Lambda_2$ has a multiplicity   of $1$ i.e. they are all distinctly unique. Intuitively, this
  doesn't allow to create a different matrix by simply interchanging the
  singular vectors associated with the equal singular values. As the elements of $\Sigma_2$ are
  distinct, the elements of  $\Lambda_2 = \gamma_2\Sigma_2$ are also
  distinct and thus by the second part of~\Cref{lem:opt-frobenius},
  $\widehat{\vec{W}}_k$ is strictly better, in terms of
  $\norm{\vec{W} - \widehat{\vec{W}}_k}$, than all matrices which have the same
  singular values as that of $\widehat{\vec{W}_k}$. This concludes our
  discussion on the uniqueness of the solution.

\paragraph{Case for $\mathbf{r=1}$:}
 Substituting $r=1$  in the
constraint  \(r = 1 + \dfrac{\sum_{j=2}^p\lambda_j^2}{\lambda_1^2}
\) we get \[ r  - 1  =
\dfrac{\sum_{j=2}^p\lambda_j^2}{\lambda_1^2}  = 0 \implies
\sum_{j=2}^p\lambda_j^2=0\] As it is a sum of squares, each of the
individual elements is also zero i.e. $\lambda_j=0~\forall 2\le j\le
p$. Substituting this into ~\cref{eq:l1}, we get the following quadratic
equation in $\lambda_1$ 
\begin{equation}
    L  = \ip{\Sigma}{\Sigma} +  \lambda_1^2 - 2\sigma_1\lambda_1\label{eq:ll0}
\end{equation}
which is minimised at
$\lambda_1 = \sigma_1$, thus proving that $\gamma_1 = 1$ and $\gamma_2
= 0$.
\paragraph{Proof for Case (b):} In this case, the constraints are meant to preserve the top $k$ singular values of the given matrix while obtaining the new one. Let $\Sigma_1 = \br{\sigma_1, \cdots, \sigma_k},~\Sigma_2 = \br{\sigma_{k+1}, \cdots, \sigma_p},~\Lambda_1 = \br{\lambda_1, \cdots, \lambda_{k}},~\Lambda_2 = \br{\lambda_{k+1}, \cdots, \lambda_p}$. Since satisfying all the constraints imply $\Sigma_1 = \Lambda_1$, thus, \( L := \norm{\vec{W} - \widehat{\vec{W}}_k}_{\forb}^2 = \ip{\Sigma_2 - \Lambda_2}{\Sigma_2 - \Lambda_2}\). From the stable rank constraint $\srank{\widehat{\vec{W}}_k} = r$, we have
  \begin{align}
    r &= \dfrac{\ip{\Lambda_1}{\Lambda_1}+\ip{\Lambda_2}{\Lambda_2}}{\lambda_1^2}\nonumber\\
    \therefore \; \; \ip{\Lambda_2}{\Lambda_2} &= r\lambda_1^2 - \ip{\Lambda_1}{\Lambda_1} = r\sigma_1^2 - \ip{\Sigma_1}{\Sigma_1}\label{eq:stable_const}
  \end{align}
The above equality constraint makes the problem non-convex. Thus, we relax it to \( \srank{\widehat{\vec{W}}_k} \leq r \) to make it a convex problem and show that the optimality is achieved with equality. Let \( r\sigma_1^2 - \ip{\Sigma_1}{\Sigma_1} = \eta \). Then, the relaxed problem can be written as
  \begin{align*}
&\min_{\Lambda_2\in\reals^{p-k}} L:=\ip{\Sigma_2 - \Lambda_2}{\Sigma_2 - \Lambda_2}  \\
& \mathrm{s.t.} \quad \Lambda_2 \geq 0, \ip{\Lambda_2}{\Lambda_2} \leq \eta. 
\end{align*}
We introduce the Lagrangian dual variables $\Gamma \in \reals^{p - k}$ and $\mu$ corresponding to the positivity and the stable rank constraints, respectively. The Lagrangian can then be written as 
\begin{align}
\cL\br{\Lambda_2, \Gamma, \mu}_{\Gamma\ge \vec{0}, \mu \geq 0} =  \ip{\Sigma_2 - \Lambda_2}{\Sigma_2 - \Lambda_2} + \mu\br{\ip{\Lambda_2}{\Lambda_2} -  \eta} - \ip{\Gamma}{\Lambda_2}
\end{align}
Using the primal optimality condition \( \dfrac{\partial \cL}{\partial \Lambda_2} =  \vec{0} \), we obtain
\begin{align}
	& 2\Lambda_2 - 2\Sigma_2 + 2\mu\Lambda_2 - \Gamma = \vec{0} \nonumber\\
  & \implies \Lambda_2 = \dfrac{\Gamma + 2\Sigma_2}{2\br{1 + \mu}}\label{eq:subst_lambda2}
\end{align}
Using the above condition on $\Lambda_2$ with the constraint $\ip{\Lambda_2}{\Lambda_2} \leq \eta$, combined with the stable rank constraint of the given matrix $\vec{W}$ that comes with the problem definition, $\srank{\vec{W}} > r$ (which implies $\ip{\Sigma_2}{\Sigma_2} > \eta$), the following inequality must be satisfied for any $\Gamma \geq 0$
\begin{align}
1 <  \frac{\ip{\Sigma_2}{\Sigma_2}}{\eta} \leq  \frac{\ip{\Gamma + \Sigma_2}{\Gamma + \Sigma_2}}{\eta} \leq (1+\mu)^2
\end{align}
For the above inequality to satisfy, the dual variable $\mu$ must be greater than zero, implying, $\ip{\Lambda_2}{\Lambda_2} - \eta$ must be zero for the complementary slackness to satisfy. Using this with the optimality condition~\cref{eq:subst_lambda2} we obtain
\begin{align}
   \br{1 + \mu}^2 &=  \dfrac{\ip{\Gamma + 2\Sigma_2}{\Gamma + 2\Sigma_2}}{4\eta} \nonumber
\end{align}
Substituting the above solution back into the primal optimality condition we get
\begin{equation}
  \label{eq:lambda_gamma}
  \Lambda_2 = \br{\Gamma + 2\Sigma_2} \dfrac{\sqrt{\eta}}{\sqrt{\ip{\Gamma + 2\Sigma_2}{\Gamma + 2\Sigma_2}}}
\end{equation}
Finally, we use the complimentary slackness condition $\Gamma\odot\Lambda_2 = \vec{0}$\footnote{$\odot$ is the hadamard product} to get rid of the dual variable $\Gamma$ as follows
\begin{align*}
  \Gamma \odot\br{\Gamma + 2\Sigma_2}\dfrac{\sqrt{\eta}}{\sqrt{\ip{\Gamma + 2\Sigma_2}{\Gamma + 2\Sigma_2}}} &= \vec{0}
\end{align*}
It is easy to see that the above condition is satisfied only when $\Gamma = \vec{0}$ as $\Sigma_2\ge\vec{0}$ and $\eta > 0$. Therefore, using $\Gamma = \vec{0}$ in~\cref{eq:lambda_gamma} we obtain the optimal solution of $\Lambda_2$ as
\begin{align}
\Lambda_2 = \dfrac{\sqrt{\eta}}{\sqrt{\ip{\Sigma_2}{\Sigma_2}}} \Sigma_2 = \frac{\sqrt{r\sigma_1^2 - \norm{\vec{S}_1}_\forb^2}}{\norm{\vec{S}_2}_\forb^2}\Sigma_2 = \gamma \Sigma_2
\end{align}
\paragraph{Proof for Case (c):} The monotonicity of \(\norm{\widehat{\vec{W}}w_k - \vec{W}}_\forb \) for $k\geq 1$ is shown in~\Cref{lem:monotonicityK}. 
\end{proof}
Note that by the assumption that $\srank{\vec{W}}<r$, we can say that $\gamma <1$. Therefore in all the cases $\gamma_2<1$. Let us look at the required conditions for $\gamma_1\ge 1$ to hold. When $k\geq 1$, $\gamma_1 = 1$ holds. When $k=0$, for $\gamma_1> 1$ to be true, $\gamma_2< \gamma$ should hold, implying, $\br{\gamma - 1} < r\br{\gamma -1}$, which is always true as $r >1$ (by the definition of stable rank).

\begin{lem}[Monotoncity of Solutions \wrt the partitioning index]
\label{lem:monotonicityK}
For $k\geq 1$, the solution to the optimisation
problem~\cref{eq:srankProblem} obtained using
Theorem~\ref{thm:srankOptimal} is closest to the original matrix
$\vec{W}$ in terms of Frobenius norm when only the spectral norm is
preserved, implying, $k=1$.
\begin{proof}
For a given matrix $\vec{W}$ and a partitioning index $k \in \{1, \cdots, p\}$, let $\widehat{\vec{W}}_k = \vec{S}_1^k + \gamma \vec{S}_2^k$ be the matrix obtained using Theorem~\ref{thm:srankOptimal}. We use the superscript $k$ along with $\vec{S}_1$ and $\vec{S}_2$ to denote that this refers to the particular solution of $\widehat{\vec{W}}_k$. Plugging the value of $\gamma$ and using the fact that $\norm{\vec{S}_2^k}_\forb \neq 0$, we can write
  \begin{align*} \norm{\vec{W} - \widehat{\vec{W}}_k}_\forb &= \br{1 - \gamma}\norm{\vec{S}_2^k}_\forb\\
                                                  &= \norm{\vec{S}_2^k}_\forb - \sqrt{r\sigma_1^2 - \norm{\vec{S}_1^k}_\forb^2}\\
                                                  &= \norm{\vec{S}_2^k}_\forb - \sqrt{r\sigma_1^2 - \norm{\vec{W}}_\forb^2+ \norm{\vec{S}_2^k}_\forb^2}.
  \end{align*}
 Thus, $\norm{\vec{W} - \widehat{\vec{W}}_k}_\forb$ can be written in a simplified form as $f(x) = x - \sqrt{a + x^2}$, where $x = \norm{\vec{S}_2^k}_\forb$ and $a = r\sigma_1^2 - \norm{\vec{W}}_\forb^2$. Note, $a \leq 0$ as $ 1 \leq r \leq  \srank{\vec{W}}$, and $a + x^2 \geq 0$ because of the condition in \Cref{thm:srankOptimal}. Under these settings, it is trivial to verify that $f$ is a monotonically decreasing function of $x$. Using the fact that as the partition index $k$ increases, $x$ decreases, it is straightforward to conclude that the minimum of $f(x)$ is obtained at $k=1$.
\end{proof}
\end{lem}

\subsection*{Auxiliary Lemmas}
\label{sec:auxLemmas}
\begin{lemL}[Reproduced from Theorem~5 in~\citet{mirsky1960symmetric}]\label{lem:ineq:frob_sing}
  For any two matrices $\vec{A},{\vec{B}}\in\reals^{m\times n}$ with singular values as $\sigma_1 \geq \cdots \geq \sigma_n$ and $\rho_1 \geq \cdots \geq \rho_n$, respectively 
  \[\norm{\vec{A}-\vec{B}}_{\forb}^2 \ge \sum_{i=1}^n\br{\sigma_i - \rho_i}^2\]
\end{lemL}
\begin{proof}
  Consider the following symmetric matrices \[
\vec{X}=
  \begin{bmatrix}
    \vec{0} & \vec{A}\\
    \vec{A}^\top & \vec{0}
  \end{bmatrix},
\vec{Y}=
  \begin{bmatrix}
    \vec{0} & \vec{B}\\
    \vec{B}^\top & \vec{0}
  \end{bmatrix},
\vec{Z}=
  \begin{bmatrix}
    \vec{0} & \vec{A-B}\\
    \vec{(A-B)}^\top & \vec{0}
  \end{bmatrix}
\]
Let $\tau_1 \geq \cdots \geq \tau_n$ be the singular values of  $\vec{Z}$. Then the set of characteristic roots of $\vec{X},\vec{Y}$ and $\vec{Z}$ in descending order are \(\bc{\rho_1,\cdots,\rho_n,-\rho_n,\cdots,-\rho_1}\), \(\bc{\sigma_1,\cdots,\sigma_n,-\sigma_n,\cdots,-\sigma_1} \), and \(\bc{\tau_1,\cdots,\tau_n,-\tau_n,\cdots,-\tau_1}\), respectively.
By Lemma~2 in~\citet{Wielandt1955} 
\[\bs{\sigma_1 - \rho_1, \cdots, \sigma_n - \rho_n, \rho_n - \sigma_n, \cdots, \rho_1 - \sigma_1}\preceq \bs{\tau_1,\cdots\tau_n,-\tau_n, -\tau_1},\] 
which implies that \begin{equation}\label{ineq:wielandt}
  \sum_{i=1}^n\br{\sigma_i - \rho_i}^2 \le \sum_{i=1}^n \tau_i^2 =
  \norm{\vec{A}-\vec{B}}_{\forb}^2
\end{equation}
\end{proof}

\begin{lem}[Minimisation wrt Frobenius Norm requires singular values only]
\label{lem:opt-frobenius}
Let $\vec{A},\vec{B} \in \reals^{m\times n}$ where $ \mathrm{SVD}(\vec{A})=\vec{U}\Sigma\vec{V}^\top$ and $\vec{B}$ is the solution to the following problem 
\begin{equation}
\label{eq:stable_rank_opt_copy}
\vec{B} = \argmin_{\srank{\vec{W}}= r}\norm{\vec{W}-\vec{A}}^2_\forb.
\end{equation}
Then, $\mathrm{SVD}\br{\vec{B}} = \vec{U}\Lambda\vec{V}^\top$ where $\Lambda$ is a diagonal matrix with non-negative entries. Implying, $\vec{A}$ and $\vec{B}$ will have the same singular vectors.

\begin{proof}
Let us assume that $\vec{Z} = \vec{S}\Lambda\vec{T}^\top$ is
a solution to the problem~\ref{eq:stable_rank_opt_copy} where
$\vec{S}\neq\vec{U}$ and $\vec{T}\neq\vec{V}$. Trivially,
$\vec{X} = \vec{U}\Lambda\vec{V}^\top$ also lies in the feasible set
as it satisfies $\srank{\vec{X}}= r$~(note stable rank only depends on
the singular values). Using the fact that the Frobenius norm is invariant to 
unitary transformations, we can write $\norm{\vec{A}-\vec{X}}_{\forb}^2 = \norm{\vec{U}\br{\Sigma
    - \Lambda}\vec{V}^\top}_{\forb}^2  = \norm{\br{\Sigma -
    \Lambda}}_{\forb}^2$. Combining this with ~\Cref{lem:ineq:frob_sing}, we obtain
    \(\norm{\vec{A}-\vec{X}}_{\forb}^2 = \norm{\br{\Sigma -
    \Lambda}}_{\forb}^2 \le \norm{\vec{A}-\vec{Z}}_{\forb}^2
\).
Since,~$\vec{S}\neq\vec{U}$ and $\vec{T}\neq\vec{V}$, we can further 
change $\le$ to a strict inequality $<$. This completes the proof.

Generally speaking, the optimal solution to 
problem~\ref{eq:stable_rank_opt_copy} with constraints depending only
on the singular values (\eg stable rank in this case) will
have the same singular vectors as that of the original matrix.
Further the inequality in~\cref{ineq:wielandt} can be converted into a
strict inequality if neither of  $\vec{A}$ and $\vec{B}$ have repeated
singular values. 
\end{proof}
\end{lem}

\begin{proposition}
\label{lem:uniqueness}
Let $\vec{y}_1 = a \vec{x}_1 + b \hat{\vec{x}}_1$ and $\vec{y}_2 = a \vec{x}_2 + b \hat{\vec{x}}_2$, where $\hat{\vec{x}}_1$ and $\hat{\vec{x}}_2$ denotes the unit vectors. Then, $\vec{y}_1 = \vec{y}_2$ if $\vec{x}_1 = \vec{x}_2$.
\end{proposition}

\clearpage
\chapter{Appendix for~\Cref{chap:causes_vul}}
\label{sec:appendix_adv_causes}
\section{Proofs for~\Cref{sec:repr-learn-no-lbl-noise}}
\label{sec:proof-22}

\begin{proof}[Proof of~\Cref{thm:parity_robust_repre_all}]
  We define a family of distribution $\cD$, such that each
  distribution in $\cD$ is supported on balls of radius $r$ around
  $\br{i,i}$ and  $\br{i+1,i}$ for positive integers $i$. Either all the balls around
  $\br{i,i}$ have the labels $1$ and the balls around $\br{i+1,i}$ have
  the label $0$ or vice versa. ~\cref{fig:complex_simple} shows an
  example where the colours indicate the label.

  Formally, for $r>0$,
  $k\in\bZ_+$, the $\br{r,k}$-1 bit parity class 
    conditional model is defined over
    $\br{x,y}\in\reals^2\times\bc{0,1}$ as follows. First, a label $y$
    is sampled uniformly from $\bc{0,1}$, then and integer $i$ is
    sampled uniformly from the set $\bc{1,\cdots,k}$ and finally
    $\vec{x}$ is generated by sampling uniformly from the $\ell_2$
    ball of radius $r$ around $\br{i+y,i}$.

     In~\Cref{thm:linear_parity_all} we first show that a set of $m$
     points sampled iid from any distribution as defined above for
     $r<\frac{1}{2\sqrt{2}}$ is with probability $1$ linear separable
     for any $m$. In   addition, standard VC bounds show that any linear classifier that
     separates $S_{m}$ for large enough $m$ will have
     small test error.~\Cref{thm:linear_parity_all} also proves that
     there exists a range of $\gamma,r$ such that for any distribution
     defined with $r$ in that range, though it is possible to obtain a
     linear classifier with $0$ training and test error, the  minimum
     adversarial risk will be bounded away from $0$.

      However, while it is possible to obtain a linear classifier with $0$ test
      error,  all such linear classifiers has a large adversarial vulnerability.
      In~\Cref{thm:parity_robust}, we show that there exists a different
      representation for this problem, which also achieves zero training and
      test error, and in addition has zero adversarial risk for a range of \(r\)
      and \(\gamma\) where the linear classifier's adversarial error was at
      least a non-zero positive constant.

\end{proof}

\begin{lem}[Robustness of Linear Classifier]
  \label{thm:linear_parity_all}
   There exists universal  constants $\gamma_0,\rho$, such
   that for any perturbation $\gamma>\gamma_0$,
     radius $r\ge\rho$, and $k\in\bZ_+$, the following holds. Let $\cD$ be the family of $\br{r,k}$-
  1-bit parity class conditional model, $\cP\in\cD$ and
  $\cS_n=\bc{\br{\vec{x}_1,y_1},\cdots,\br{\vec{x}_n,y_1}}$ be a set
  of $n$ points sampled i.i.d. from
  $\cP$. 
\begin{enumerate}
\item[1)]  For any $n>0$, $S_n$ is linearly separable with probability $1$
  i.e. there exists a $h:\br{\vec{w},w_0}$,
  $\vec{w}\in\reals^2,w_0\in\reals$ such that the linear hyperplane
  $\vec{x}\rightarrow\vec{w}^\top\vec{x}+w_0$ separates $\cS_n$ with
  probability $1$:
  \[\forall \br{\vec{x},y}\in\cS_n\quad
    z\br{\vec{w}^\top\vec{x}+w_0}>0\quad\text{where}~ z=2y-1\]

\item[2)] Further there exists an universal constant $c$ such that for
  any $\epsilon,\delta>0$ with
  probability $1-\delta$ for any $\cS_n$ with
  $n=c\frac{1}{\epsilon^2}\log\frac{1}{\delta}$, any linear classifier
  $\tilde{h}$ that separates $\cS_n$ has
  $\risk{\cP}{\tilde{h}}\le\epsilon$.
 \item [3)] Let $h:\br{\vec{w},w_0}$ be any linear classifier that has
   $\risk{\cP_P}{h}=0$. Then, $\radv{\gamma}{h;\cP}>0.0005$.
\end{enumerate}
\end{lem}

 We will prove the first part for any $r<\frac{1}{2\sqrt{2}}$ by
 constructing a $\vec{w},w_0$ such that it 
  satisfies the constraints of linear separability. Let
  $\vec{w}=\br{1,-1},~w_0=-0.5$. Consider any point
  $\br{\vec{x},y}\in\cS_n$ and $z=2y-1$. Converting to the polar coordinate system
  there exists a $\theta\in\bs{0,2\pi},j\in\bs{0,\cdots,k}$ such that
  $\vec{x}=\br{j+\frac{z+1}{2}+r\mathrm{cos}\br{\theta},j+r\mathrm{sin}\br{\theta}}$ 
  \begin{align*}
    z\br{\vec{w}^\top\vec{x}+w_0}&=z\br{j+\frac{z+1}{2}+r\mathrm{cos}\br{\theta}-j
                                   - r\mathrm{sin}\br{\theta}-
                                   0.5}&&\vec{w}=\br{1,-1}^\top\\
                                 &=z\br{\frac{z}{2}+0.5+r\mathrm{cos}\br{\theta} -
                                   r\mathrm{sin}\br{\theta}-0.5}\\
                                 &=\frac{1}{2} +
                                   zr\br{\mathrm{cos}\br{\theta}-\mathrm{sin}\br{\theta}}&&\abs{\mathrm{cos}\br{\theta}-\mathrm{sin}\br{\theta}}<\sqrt{2}\\
                                 &>\frac{1}{2} - r\sqrt{2}\\
                                 &>0 &&r<\frac{1}{2\sqrt{2}}
  \end{align*}

  Part 2 follows from a simple application of VC bounds for linear
  classifiers.
  
  Let the universal constants $\gamma_0,\rho$ be $0.02$ and
  $\frac{1}{2\sqrt{2}}-0.008$ respectively. Note that there is nothing special
  about this constants except that \emph{some} constant is required to bound the
  adversarial risk away from $0$.  Now, consider a distribution $\cP$ 1-bit
  parity model where the radius of each ball is at least $\rho$. This is smaller
  than $\frac{1}{2\sqrt{2}}$ and thus satisfies the linear separability
  criterion.

  Consider $h$ to be a hyper-plane that has $0$ test error.  Let the
  $\ell_2$ radius of adversarial perturbation be
  $\gamma>\gamma_0$. The region of each circle that will be vulnerable
  to the attack will be a circular segment with the chord of the
  segment parallel to the hyper-plane. Let the minimum height of
  all such circular segments be $r_0$. Thus,
  $\radv{\gamma}{h;\cP}$ is greater than the mass of the circular
  segment of radius $r_0$. Let the radius of each ball in the support
  of $\cP$ be $r$.

  Using the fact that $h$ has zero test error; and thus classifies the
  balls in the support of $\cP$ correctly and simple geometry

  \begin{align}
    \frac{1}{\sqrt{2}}&\ge r +\br{\gamma-r_0}+r\nonumber\\
    r_0&\ge 2r + \gamma- \frac{1}{\sqrt{2}}\label{eq:radii}
  \end{align}
  To compute
  $\radv{\gamma}{h;\cP}$ we need to compute the ratio of the area of a circular
  segment of height $r_0$ of a circle of radius $r$ to the area of the
  circle. The ratio  can be written 

  \begin{align}\label{eq:circ-seg}
    A\br{\frac{r_0}{r}} =\frac{{cos}^{-1}\br{1-\frac{r_0}{r}} - \br{1 -
        \frac{r_0}{r}}\sqrt{2\frac{r_0}{r} - \frac{r_0^2}{r^2}}}{\pi}
  \end{align}

  As~\Cref{eq:circ-seg} is increasing with $\frac{r_0}{r}$, we can evaluate 
  \begin{align*}\label{eq:c_1_eq}
    \frac{r_0}{r}&\ge\frac{2r - \frac{1}{\sqrt{2}}+\gamma}{r}&&\text{Using}~\Cref{eq:radii}\\
                 &\ge 2 - \frac{\frac{1}{\sqrt{2}}-0.02}{r}&&\gamma
                                                              >\gamma_0
    = 0.02\\
                 &\ge 2 -
                   \frac{\frac{1}{\sqrt{2}}-0.02}{\frac{1}{\sqrt{2}}-0.008}>0.01&&r>\rho=\frac{1}{2\sqrt{2}}-0.008
  \end{align*}
  Substituting $\frac{r_0}{r}>0.01$ into Eq.~\Cref{eq:circ-seg}, we
  get that $A\br{\frac{r_0}{r}}>0.0005$. Thus, for all
  $\gamma>0.02$, we have $\radv{\gamma}{h;\cP}>0.0005$.

\begin{lem}[Robustness of parity
  classifier]
\label{thm:parity_robust}
  There exists a concept class $\cH$  such that for any
  $\gamma\in\bs{\gamma_0,\gamma_0+\frac{1}{8}}$,
  $k\in\bZ_+$, $\cP$ being the 
  corresponding $\br{\rho,k}$ 1-bit parity class distribution where
  $\rho,\gamma_0$ are the same as in~\Cref{thm:linear_parity_all} there
  exists $g\in\cH$ such that
  \[\risk{\cP}{g} =  0\qquad\radv{\gamma}{g;\cP}=0\]
\end{lem}

\begin{proof}[Proof of~\Cref{thm:parity_robust}]
  We will again provide a proof by construction.  Consider the
  following class of concepts $\cH$ such that $g_b\in\cH$ is defined
  as \begin{equation}
  \label{eq:2}
  g\br{\br{x_1,x_2}^\top}=\begin{cases}
      1&\text{if} \bs{x_1}+\bs{x_2}  =b \br{\text{mod 2}}\\
      1-b &\text{o.w.}
  \end{cases}
\end{equation} where $\bs{x}$ rounds $x$ to the nearest integer and
$b\in\bc{0,1}$. In~\Cref{fig:complex_simple}, the
green staircase-like classifier belongs to this class. Consider the
classifier $g_1$. Note that by construction $\risk{\cP}{g_1}=0$. The
decision boundary of $g_1$ that are closest to a ball in the support
of $\cP$ centred at $\br{a,b}$ are the lines $x=a\pm 0.5$ and
$y=b\pm 0.5$.

As $\gamma<\gamma_0 + \frac{1}{8}$, the adversarial perturbation is
upper bounded by $\frac{1}{50} + \frac{1}{8}$. The radius of
the ball is upper bounded by $\frac{1}{2\sqrt{2}}$, and as we noted
the center of the ball is at a distance of $0.5$ from the decision
boundary. If the sum of the maximum adversarial perturbation and the
maximum radius of the ball is less than the minimum distance of the
center of the ball from the decision boundary, then the adversarial
error is $0$. Substituting the values, \[\frac{1}{50} +
  \frac{1}{8} + \frac{1}{2\sqrt{2}}
  < 0.499 <\frac{1}{2} \]
This completes the proof.
\end{proof}

\section{Proofs for~\Cref{sec:repr-theorey-lbl-noise}}
\begin{proof}[Proof of~\Cref{thm:repre-par-inter}]
  We will provide a constructive proof to this theorem by
  constructing a distribution $\cD$, two concept classes~$\cC$ and $\cH$
  and provide the ERM algorithms to learn the concepts and then
  use~\Cref{lem:parity_repre,lem:uni_int_repre} to complete the proof.

  \textbf{Distribution:} Consider the family of distribution $\cD^n$
  such that $\cD_{S,\zeta}\in\cD^n$ is defined on $\cX_\zeta\times\bc{0,1}$ for
  $S\subseteq\bc{1,\cdots,n},\zeta\subseteq\bc{1,\cdots,2^n-1}$  such that the support of $\cX_\zeta$ is a union of
  intervals. 
  \begin{equation}
    \label{eq:dist_union_int}
    \mathrm{supp}\br{\cX}_\zeta=\bigcup_{j\in\zeta}I_j\text{ where }
    I_j:=\br{j-\frac{1}{4}, j+\frac{1}{4}}
  \end{equation}
  We consider distributions with a relatively small
  support i.e. where $\abs{\zeta}=\bigO{n}$. Each sample $\br{\vec{x},y}~\sim\cD_{S,\zeta}$ is created by sampling
  $\vec{x}$ uniformly from $\cX_\zeta$ and assigning $y=c_S\br{\vec{x}}$ where
  $c_S\in\cC$ is defined below~(\cref{eq:parity_concept}). We define the
  family of distributions $\cD =
  \bigcup_{n\in\bZ_+}\cD^n$.  Finally, we create
  $\cD_{S,\zeta}^\eta$ -a noisy version of $\cD_{S,\zeta}$, by flipping $y$ in each sample
  $\br{x,y}$ with probability $\eta<\frac{1}{2}$. Samples from
  $\cD_{S,\zeta}$ can be obtained using the example oracle
  $\mathrm{EX}\br{\cD_{S,\zeta}}$ and samples from the noisy
  distribution can be obtained through the noisy oracle $\mathrm{EX}^\eta\br{\cD_{S,\zeta}}$
  
  \textbf{Concept Class $\cC$:} We define the  concept class $\cC^n$ of concepts
  $c_S:\bs{0,2^n}\rightarrow
  \bc{0,1}$ such that
  \begin{equation}
    \label{eq:parity_concept}
    c_S\br{\vec{x}}=\begin{cases}
      1,
      &\text{if}\br{\langle\bs{\vec{x}}\rangle_b~\mathrm{XOR}~S}~\text{
        is odd.}\\
      0 &~\text{o.w.}
    \end{cases}
  \end{equation}
  where $\bs{\cdot}:\reals\rightarrow\bZ$ rounds a decimal
  to its nearest
  integer,~$\langle\cdot\rangle_b:\bc{0,\cdots,2^n}\rightarrow\bc{0,1}^n$
  returns the binary encoding of the integer,~and
  $\br{\langle\bs{\vec{x}}\rangle_b~\textrm{XOR}~S} = \sum_{j\in S}
  \langle\bs{x}\rangle_b\bs{j}~\textrm{mod}~2$. $\langle\bs{x}\rangle_b\bs{j}$
  is the $j^{\it th}$ least significant bit in the binary encoding of
  the nearest integer to $\vec{x}$. It is essentially the
  class of parity functions  defined on the bits corresponding to the
  indices in $S$ for the binary
  encoding of the nearest  integer to $\vec{x}$. For example, as
in~\Cref{fig:thm-3} if $S = \{0, 2\}$, then only the least significant and
the third least significant bit of $i$ are examined and the class label is
$1$ if an odd number  of them are $1$ and $0$ otherwise.

  \textbf{Concept Class $\cH$:} Finally, we define the concept class
  $\cH=\bigcup_{k=1}^\infty\cH_k$ where $\cH_k$ is the class of union of
$k$ intervals on the real line  $\cH^k$. Each concept $h_I\in\cH^k$
can be written as a set of $k$ disjoint intervals
$I=\bc{I_1,\cdots,I_k}$ on the real line i.e. for $1\le j\le k$,
$I_j=\bs{a,b}$ where $0\le a\le b$ and
\begin{equation}
  \label{eq:union_int-defn_class}
  h_I\br{\vec{x}} =\begin{cases}
    1&\text{if}~\vec{x}\in\bigcup_j I_j\\
    0&\text{o.w.}
  \end{cases}
\end{equation}

Now, we look at the algorithms to learn the concepts from $\cC$ and
$\cH$ that minimise the train error. Both of the algorithms will use a
majority vote to determine the correct~(de-noised) label for each interval, which
will be necessary to minimise the test error. The intuition is that if
we draw a sufficiently large number of samples, then the majority of
samples on each interval will have the correct label with a high
probability. 

~\Cref{lem:parity_repre} proves that there exists an algorithm $\cA$
such that $\cA$ draws
$m=\bigO{\abs{\zeta}^2\frac{\br{1-\eta}}{\br{1-2\eta}^2}\log{\frac{\abs{\zeta}}{\delta}}}$
samples from the noisy oracle $\mathrm{EX}^\eta\br{\cD_{s,\zeta}}$ and with probability $1-\delta$
where the probability is over the randomisation in the oracle, returns
$f\in\cC$ such that $\risk{\cD_{S,\zeta}}{f}=0$ and 
$\radv{\gamma}{f;\cD_{S,\zeta}}=0$ for all
$\gamma<\frac{1}{4}$. As~\Cref{lem:parity_repre} states, the algorithm
involves gaussian elimination over $\abs{\zeta}$ variables and
$\abs{\zeta}$ majority votes~(one in each interval) involving a total
of $m$ samples. Thus the
algorithm runs in $\bigO{\poly{m}+\poly{\abs{\zeta}}}$ time. Replacing
the complexity of $m$ and the fact that $\abs{\zeta}=\bigO{n}$, the
complexity of the algorithm is
$\bigO{\poly{n,\frac{1}{1-2\eta},
\frac{1}{\delta}}}$.  

~\Cref{lem:uni_int_repre} proves that there
 exists an algorithm $\widetilde{A}$ such that $\widetilde{A}$ draws \[m>\mathrm{max}\bc{
      2\abs{\zeta}^2\log{\frac{2\abs{\zeta}}{\delta}} 
   \br{8\frac{\br{1-\eta}}{\br{1-2\eta}^2}+1},
   \frac{0.1\abs{\zeta}}{\eta\gamma^2} 
  \log\br{\frac{0.1\abs{\zeta}}{\gamma\delta}}}\] samples and returns
$h\in\cH$ such that $h$ has $0$ training error, $0$ test error and an
adversarial test error of at least $0.1$. We can replace $\abs{\zeta} =
\bigO{n}$ to get the required bound on $m$ in the theorem. The
algorithm to construct $h$ visits every point at most twice - once
during the construction of the intervals using majority voting, and
once while accommodating for the mislabelled points.  Replacing
the complexity of $m$, the
complexity of the algorithm is  $\bigO{\poly{n,\frac{1}{1-2\eta},\frac{1}{\gamma},\frac{1}{\delta}}}$. This completes the proof.
\end{proof}

\begin{lem}[Parity Concept Class]\label{lem:parity_repre}
  There exists a  learning algorithm $\cA$ such that given
  access to the noisy example oracle
  $\mathrm{EX}^\eta\br{\cD_{S,\zeta}}$, $\cA$ makes
  $m=\bigO{\abs{\zeta}^2\frac{\br{1-\eta}}{\br{1-2\eta}^2}\log{\frac{\abs{\zeta}}{\delta}}}$
  calls to the oracle and returns a
  hypothesis $f\in\cC$  such that with probability
  $1-\delta$, we have that $\risk{\cD_{S,\zeta}}{f}=0$ and
  $\radv{\gamma}{f;\cD_{S,\zeta}}=0$ for all $\gamma<\frac{1}{4}$.
\end{lem}

\begin{proof}
  The algorithm $\cA$ works as follows. It  makes $m$ calls to the oracle
  $\mathrm{EX}\br{\cD_s^m}$ to obtain a set of
  points~$\bc{\br{x_1,y_1},\cdots,\br{x_m,y_m}}$ where
  $m\ge 2\abs{\zeta}^2\log{\frac{2\abs{\zeta}}{\delta}}\br{8\frac{\br{1-\eta}}{\br{1-2\eta}^2}+1}$
  . Then, it replaces each $x_i$ with $\bs{x_i}$~($\bs{\cdot}$ rounds a
  decimal to the nearest integer) and then removes duplicate
  $x_i$s by preserving the most frequent label $y_i$ associated with each
  $x_i$.
  For example, if $\cS_5 = \bc{\br{2.8,1}, \br{2.9, 0}, \br{3.1,
      1},\br{3.2, 1}, \br{3.9, 0}}$ then after this operation, we will
  have $\bc{\br{3,1}, \br{4,0}}$.

   As   $m\ge 2\abs{\zeta}^2\log{\frac{2\abs{\zeta}}{\delta}}
   \br{8\frac{\br{1-\eta}}{\br{1-2\eta}^2}+1}$, using
   $\delta_2=\frac{\delta}{2}$ and
   $k=\frac{8\br{1-\eta}}{\br{1-2\eta}^2}\log\frac{2\abs{\zeta}}{\delta}$
   in
   ~\Cref{lem:min-wt} guarantees that with probability $1-\frac{\delta}{2}$, each
  interval will have at least
  $\frac{8\br{1-\eta}}{\br{1-2\eta}^2}\log\frac{2\abs{\zeta}}{\delta}$
  samples.

  Then for any specific interval, using
  $\delta_1=\frac{2\abs{\zeta}}{\delta}$ in ~\Cref{lem:majority_lem} guarantees that with
  probability at least $1-\frac{2\abs{\zeta}}{\delta}$, the majority
  vote for the label in that interval will succeed in returning
  the de-noised label. Applying a union bound~(\Cref{ineq:union-bound}) over all $\abs{\zeta}$ intervals, will
  guarantee that with probability at least $1-\delta$, the majority
  label of every interval will be the denoised label.

  Now, the problem reduces to solving a parity problem on this reduced
  dataset of $\abs{\zeta}$ points~(after denoising, all points in that
  interval can be reduced to the integer in the interval and the
  denoised label). We know that there exists a polynomial
  algorithm using Gaussian Elimination that finds a consistent
  hypothesis for this problem. We have already guaranteed that there is a
  point in $\cS_m$ from every interval in the 
  support of $\cD_{S,\zeta}$. Further, $f$ is consistent on $\cS_m$ and $f$ is
  constant in each of these intervals by design. Thus, with
  probability at least  $1-\delta$ we have that  $\risk{\cD_{S,\zeta}}{f}=0$.

  By construction, $f$  makes a constant
  prediction on each interval $\br{j-\frac{1}{2},j+\frac{1}{2}}$ for
  all $j\in\zeta$. Thus, for any perturbation radius
  $\gamma<\frac{1}{4}$ the adversarial risk
  $\radv{\cD_{S,\prime{\zeta}}}{f}=0$. Combining everything, we have shown that there is an algorithm
  that makes $2\abs{\zeta}^2\log{\frac{2\abs{\zeta}}{\delta}}\br{8\frac{\br{1-\eta}}{\br{1-2\eta}^2}+1}$ calls to the
  $\mathrm{EX}\br{\cD_{S,\zeta}^\eta}$ oracle,  runs in time polynomial in $\abs{\zeta},\frac{1}{1-2\eta},\frac{1}{\delta}$ to return
  $f\in\cC$ such that $\risk{\cD_{S,\zeta}}{f}=0$ and
  $\radv{\gamma}{f;\cD_{S,\zeta}}=0$ for $\gamma<\frac{1}{4}$.
\end{proof}

\begin{lem}[Union of Interval Concept Class]\label{lem:uni_int_repre}
   There exists a  learning algorithm $\widetilde{\cA}$ such that given
  access to a noisy example oracle makes
  $m=\bigO{\abs{\zeta}^2\frac{\br{1-\eta}}{\br{1-2\eta}^2}
    \log{\frac{\abs{\zeta}}{\delta}}}$  calls to the oracle and
  returns a hypothesis $h\in\cH$ 
  such that training error is $0$ and with probability
  $1-\delta$, $\risk{\cD_{S,\zeta}}{f}=0$.

  Further for any $h\in\cH$ that has zero training error on
  $m^\prime$ samples drawn from $\mathrm{EX}^\eta\br{\cD_{S,\zeta}}$
  for $m^\prime >  \frac{\abs{\zeta}}{10\eta\gamma^2}
  \log\frac{\abs{\zeta}}{10\gamma\delta}$   and
  $\eta\in\br{0,\frac{1}{2}}$ then 
  $\radv{\gamma}{f;\cD_{S,\zeta}}\ge 0.1$ for all $\gamma>0$.
\end{lem}

\begin{proof}[Proof of~\Cref{lem:uni_int_repre}]
  The first part of the algorithm works similarly
  to~\Cref{lem:parity_repre}. The algorithm $\widetilde{\cA}$ makes
  $m$ calls to the oracle  $\mathrm{EX}\br{\cD_s^m}$ to obtain a set of
  points~$\cS_m = \bc{\br{x_1,y_1},\cdots,\br{x_m,y_m}}$ where
  $m\ge 2\abs{\zeta}^2\log{\frac{2\abs{\zeta}}{\delta}}
  \br{8\frac{\br{1-\eta}}{\br{1-2\eta}^2}+1}$. $\widetilde{\cA}$
  computes $h\in\cH$ as follows. To begin, let the list of
  intervals in $h$ be $I$ and $\cM_z=\bc{}$ Then do the following for every
  $\br{x,y}\in\cS_m$.
  \begin{enumerate}[leftmargin=0.5cm,itemsep=0ex]
  \item let $z := \bs{x}$, 
  \item Let $\cN_z\subseteq\cS_m$ be the set of all $\br{x,y}\in\cS_m$ such that
    $\abs{x-z}<0.5$.
  \item Compute the majority label $\tilde{y}$ of $\cN_z$.
  \item Add all $\br{x,y}\in\cN_z$ such that $y\neq \tilde{y}$ to $\cM_z$
  \item If $\tilde{y}=1$, then add the interval $(z-0.5,z+0.5)$ to $I$.
  \item Remove all elements of $\cN_z$ from $\cS_m$ i.e. $\cS_m:=\cS_m\setminus\cN_z$.
  \end{enumerate}
For reasons similar to~\Cref{lem:parity_repre}, as   $m\ge
   2\abs{\zeta}^2\log{\frac{2\abs{\zeta}}{\delta}}
   \br{8\frac{\br{1-\eta}}{\br{1-2\eta}^2}+1}$, ~\Cref{lem:min-wt} guarantees
   that with probability $1-\frac{\delta}{2}$, each interval will have at least
   $\frac{8\br{1-\eta}}{\br{1-2\eta}^2}\log\frac{2\abs{\zeta}}{\delta}$ samples.
   Then for any specific interval, ~\Cref{lem:majority_lem} guarantees that with
   probability at least $1-\frac{2\abs{\zeta}}{\delta}$, the majority vote for
   the label in that interval will succeed in returning the de-noised label.
   Applying a union bound~(\Cref{ineq:union-bound}) over all intervals, will
   guarantee that with probability at least $1-\delta$, the majority label of
   every interval will be the denoised label. As each interval in$\zeta$ has at
   least one point, all the intervals in $\zeta$ with label $1$ will  be
   included in $I$ with probability $1-\delta$. Thus,
   $\risk{\cD_{S,\zeta}}{h}=0$.

  Now, for all $\br{x,y}\in\cM_z$, add the interval $\bs{x}$ to $I$ if
  $y=1$. If $y=0$ then $x$ must lie a interval $(a,b)\in
  I$. Replace that interval as follows $I:= I\setminus(a,b)\cup
  \bc{(a,x),(x,b)}$. As only a finite number of sets with Lebesgue
  measure of $0$ were added or deleted
  from $I$, the net test error of $h$ doesn't change and is still
  $0$ i.e.  $\risk{\cD_{S,\zeta}}{h}=0$

  For the second part, we will invoke~\Cref{thm:inf-label}. To avoid
  confusion in notation, we will use $\Gamma$ instead of $\zeta$ to
  refer to the sets in~\Cref{thm:inf-label} and reserve $\zeta$ for
  the support of interval of $\cD_{S,\zeta}$. Let $\Gamma$ be any set of
  disjoint intervals of width $\frac{\gamma}{2}$ such that $\abs{\Gamma}=
  \frac{0.1\abs{\zeta}}{\gamma}$. This is always possible as the total
  width of all intervals in $\Gamma$ is $
  \frac{0.1\abs{\zeta}}{\gamma}\frac{\gamma}{2} =
  0.1\frac{\abs{\zeta}}{2}$ which is less than the total width of the
  support $\frac{\abs{\zeta}}{2}$. $c_1,c_2$ from
  Eq.~\Cref{eq:balls_density} is \[c_1 =
    \bP_{\cD_{S,\zeta}}\bs{\Gamma} =
    \frac{2*0.1\abs{\zeta}}{2\abs{\zeta}} = 0.1,\quad
    c_2=\frac{2\gamma}{2\abs{\zeta}}\abs{\zeta}=\gamma\]

  Thus, if $h$ has an error of zero on a set of $m^\prime$ examples
  drawn from $\mathrm{EX}^{\eta}\br{\cD_{S,\zeta}}$ where $m^\prime>
  \frac{0.1\abs{\zeta}}{\eta\gamma^2}
  \log\br{\frac{0.1\abs{\zeta}}{\gamma\delta}}$, then
  by~\Cref{thm:inf-label}, $\radv{\gamma}{h;\cD_{S,\zeta}}>0.1$.

  Combining the two parts for \[m>\mathrm{max}\bc{
      2\abs{\zeta}^2\log{\frac{2\abs{\zeta}}{\delta}} 
   \br{8\frac{\br{1-\eta}}{\br{1-2\eta}^2}+1},
   \frac{0.1\abs{\zeta}}{\eta\gamma^2} 
  \log\br{\frac{0.1\abs{\zeta}}{\gamma\delta}}}\] it is possible to
obtain $h\in\cH$ such that $h$ has zero training error,
$\risk{h}{\cD_{S,\zeta}}=0$ and $\radv{\gamma}{h;\cD_{S,\zeta}}>0.1$
for any $\gamma>0$.

\end{proof}

\begin{lem}\label{lem:min-wt}
  Given $k\in\bZ_+$ and a distribution $\cD_{S,\zeta}$, for any
  $\delta_2 > 0$ if
  $m>2\abs{\zeta}^2k + 2\abs{\zeta}^2\log{\frac{\abs{\zeta}}{\delta_2}}$ samples are
  drawn from $\mathrm{EX}\br{\cD_{S,\zeta}}$ then with probability
  at least $1 - \delta_2$ there are at least $k$ samples in each
  interval $\br{j-\frac{1}{4},j+\frac{1}{4}}$ for all $j\in\zeta$.
\end{lem}
\begin{proof}[Proof of~\Cref{lem:min-wt}]
   We will repeat the following procedure $\abs{\zeta}$ times once for
   each interval in $\zeta$ and show that with probability
   $\frac{\delta}{\abs{\zeta}}$ the $j^{\it{th}}$ run will result in
   at least $k$ samples in the $j^{\it th}$ interval.

   Corresponding to each interval in $\zeta$, we will sample at least  $m^\prime$ samples where $m^\prime=2\abs{\zeta}k +
  2\abs{\zeta}\log{\frac{\abs{\zeta}}{\delta_2}}$.  If $z_i^j$ is the 
  random variable that is $1$ when the $i^{\it th}$ 
  sample belongs to the $j^{\it th}$ interval, then $j^{\it th}$
  interval has at least $k$ points out of the $m^\prime$ points sampled
  for that interval with probability  less
  than $\frac{\delta_2}{\abs{\zeta}}$.
  \begin{align*}
    \bP\bs{\sum_i z_{i}^j \le k} &= \bP\bs{\sum_i z_{i}^j \le \br{1 -
    \delta}\mu} &&\delta = 1 - \frac{k}{\mu}, \mu = \bE\bs{\sum_i
                   z_i^j}\\
    &\le \exp\br{-\br{1-\frac{k}{\mu}}^2\frac{\mu}{2}} &&\text{By
                                                       Chernoff's
                                                       inequality~(\Cref{ineq:hoeffding})}\\
    &\le
      \exp\br{-\br{\frac{m^\prime}{2\abs{\zeta}}-k+\frac{k^2\abs{\zeta}}{2m^\prime}}}
                && \mu=\frac{m^\prime}{\abs{\zeta}}\\
    &\le
      \exp\br{k-\frac{m^\prime}{2\abs{\zeta}}}\le \frac{\delta_2}{\abs{\zeta}}
  \end{align*}\todo[color=green]{Refer to chernoff in app} where the last step follows from $m^\prime>2\abs{\zeta}k +
  2\abs{\zeta}\log{\frac{\abs{\zeta}}{\delta_2}}$. With probability
  at least $\frac{\delta}{\abs{\zeta}}$, every interval will have
  at least $k$ samples. Finally, an union
  bound~(\Cref{ineq:union-bound}) over each interval gives the desired result. As we repeat the
  process for all $\abs{\zeta}$ intervals, the total
  number of samples drawn will be at least $\abs{\zeta}m^\prime =
  2\abs{\zeta}^2k +  2\abs{\zeta}^2\log{\frac{\abs{\zeta}}{\delta_2}}$.
\end{proof}

\begin{lem}[Majority Vote]\label{lem:majority_lem}
For a given $y\in\bc{0,1}$, let $S=\bc{s_1,\cdots,s_m}$ be a set of size $m$ where each element is $y$ with
probability $1-\eta$ and $1-y$ otherwise. If
$m>\frac{8\br{1-\eta}}{\br{1-2\eta}^2}\log\frac{1}{\delta_1}$ then with
probability at least $1-\delta_1$ the majority of $S$ is $y$.
\end{lem}
\begin{proof}[Proof of~\Cref{lem:majority_lem}]
  Without loss of generality let $y=1$. For the majority to be $1$ we
  need to show that there are more than $\frac{m}{2}$ ``$1$''s in $S$
  i.e. we need to show that the following probability is less than $\delta_1$.
  \begin{align*}
    \bP\bs{\sum s_i< \frac{m}{2}} &= \bP\bs{\sum s_i <
                                    \frac{m}{2\mu}*\mu +\mu -
                                      \mu}&&\mu = \bE\bs{\sum s_i}\\
                                    &= \bP\bs{\sum s_i < \br{1 - \br{1
                                      - \frac{m}{2\mu}}}\mu}\\
                                    &\le
                                      \exp{\br{-\frac{\br{1-2\eta}^2}{8\br{1-\eta}^2}\mu}}
                                          &&\text{By Chernoff's
                                             Inequality~(\Cref{ineq:hoeffding})}\\
                                    &=\exp{\br{-\frac{\br{1-2\eta}^2}{8\br{1-\eta}}m}}
                                          &&\because \mu=\br{1-\eta}m\\
                                    &\le \delta_1
                                          &&\because m>\frac{8\br{1-\eta}}{\br{1-2\eta}^2}\log{\frac{1}{\delta_1}}
  \end{align*}
\end{proof}

\chapter{Appendix for~\Cref{chap:low_rank_main}}
\label{app:low_rank}
\section{Proofs for~\Cref{sec:alg-lr}}\label{sec:lr-spsd-proof}

\begin{proof}[Proof of~\Cref{thm:nystrom_sym}] By the Construction of the Nystr\"om SVD algorithm, we know that
  $vec{X}_r = \vec{C}\vec{W}_r^{+}\vec{C}^T$. We will first show that
  $\vec{W}_r^{+}$ is a symmetric matrix.
  
    We know that $\vec{X}$ is SPSD. Let $\vec{I}$ be a sorted list of distinct
  indices of length $l$. Then by construction of $\vec{W}$, \[\vec{W}_{i,j} =
  \vec{X}_{I[i],\vec{I}[j]}\] As $\vec{X}_{\vec{I}[i],\vec{I}[j]} =
  \vec{X}_{\vec{I}[j], \vec{I}[i]}$, $\vec{W}$ is symmetric. At this step, our
  algorithm adds $\delta\cdot \vec{I}$ to $\vec{W}$ where $\delta\ge 0$. We show
  that $W + \delta\cdot\mathcal{I}$ is positive semidefinite. Consider a vector
  $\vec{a}\in \reals^{|X|}$.  Create a vector $\bar{\vec{a}}\in\reals^m $ where \[\bar{\vec{a}}_i =
      \begin{cases} 0 &\text{if } i\not\in I\\ \vec{a}_i & \text{o.w.}
      \end{cases}
    \]
    
    \begin{equation} \vec{a}^\top \br{\vec{W} + \delta\cdot \cI}\vec{a}= \bar{\vec{a}}^\top
  X\bar{\vec{a}} + \delta\cdot \vec{a}^\top\vec{I} \vec{a} \ge 0 + \delta\norm{\vec{a}}^2\ge 0 \label{eq:spsd_proof}
  \end{equation}
  
  Let $\vec{W} + \delta\vec{I}$ be the new $\vec{W}$;~\Cref{eq:spsd_proof} shows
  that the updated $\vec{W}$ is also positive semidefinite.
  
    Now we will show that $\vec{X}_r$ is symmetric as well.  As $\vec{W}$ is
  symmetric, there exists an orthogonal matrix $\vec{Q}$ and a non-negative
  diagonal matrix $\Lambda$ such that  \[W = Q\Lambda Q^T\] 
  We know that $\vec{W}_r = \vec{Q}_{[1:r]}\Lambda_{[1:r]}
    \vec{Q}_{[1:r]}^T$ and  $\vec{W}_r^{+} = \vec{Q}_{[1:r]}\Lambda_{[1:r]}^{-1}
    \vec{Q}_{[1:r]}^T$.\\ Hence,
    \begin{align*} 
      \vec{X}_r &= \vec{C} \vec{W}_r^{+}\vec{C}^T\\ 
      &=\vec{C} \vec{Q}_{[1:r]}\Lambda_{[1:r]}^{-1} \vec{Q}_{[1:r]}^T \vec{C}^T\\ \vec{X}_r^T &= (\vec{C}\vec{Q}_{[1:r]}\Lambda_{[1:r]}^{-1} \vec{Q}_{[1:r]}^T \vec{C}^T)^T\\ 
      &= \vec{C}\vec{Q}_{[1:r]}\Lambda_{[1:r]}^{-1} \vec{Q}_{[1:r]}^T \vec{C}^T\\ &= \vec{X}_r\\
    \end{align*} $\therefore \vec{X}_r$ is symmetric. We can also see that the
  $\vec{X}_r^T$ is positive semi definite by pre-multiplying and post
  multiplying it with a non-zero vector and using the fact that
  $\vec{W}_r^{+}$ is positive semi-definite.
  \end{proof}

\section{Additional Results}

  \begin{figure}[H]\centering
    \begin{subfigure}[c]{1.0\linewidth} 
  \begin{subfigure}[c]{0.24\linewidth} \centering
\def\svgwidth{0.99\columnwidth} 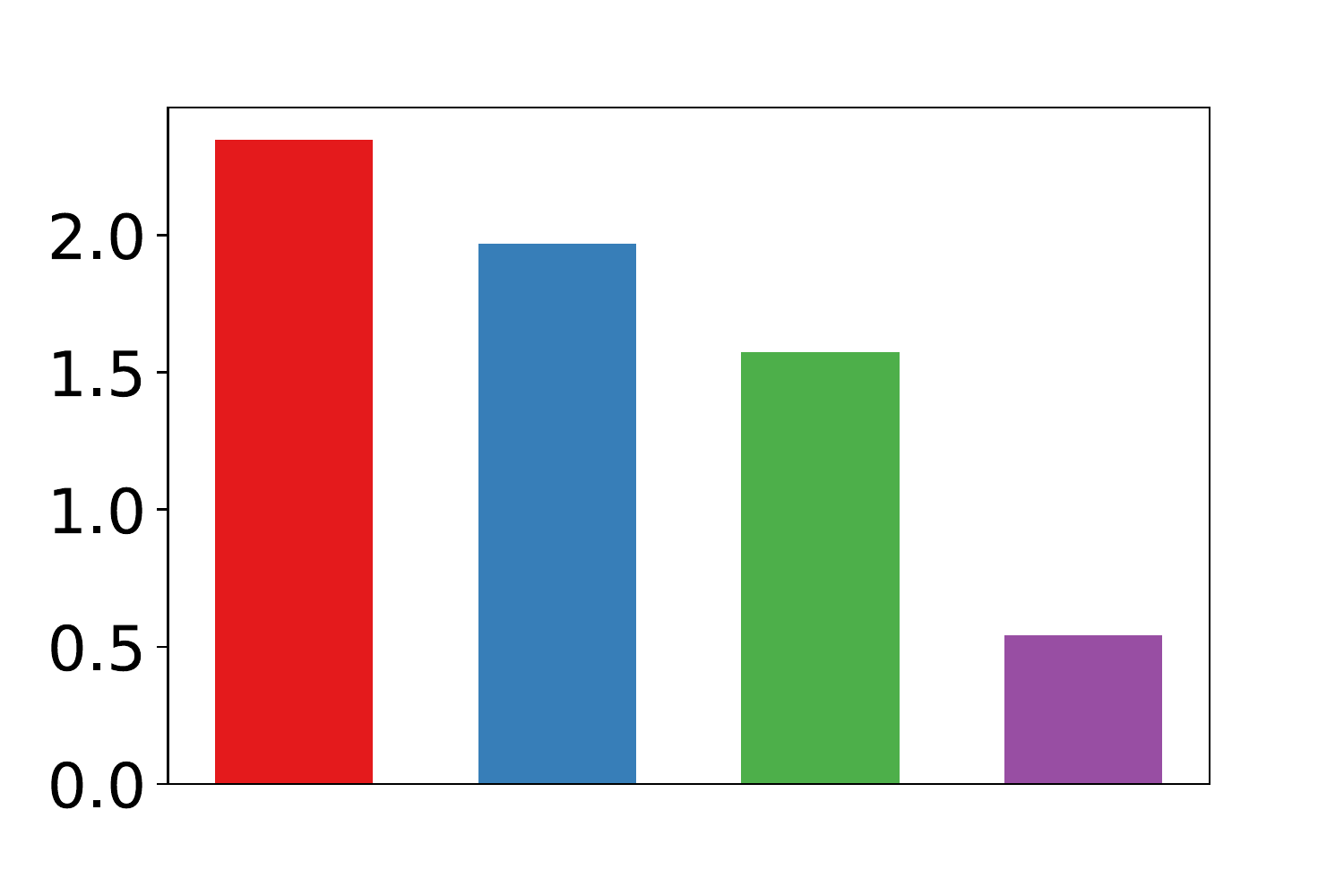
\caption*{Conv Layer 1}
  \end{subfigure}
  \begin{subfigure}[c]{0.24\linewidth} \centering
\def\svgwidth{0.99\columnwidth} 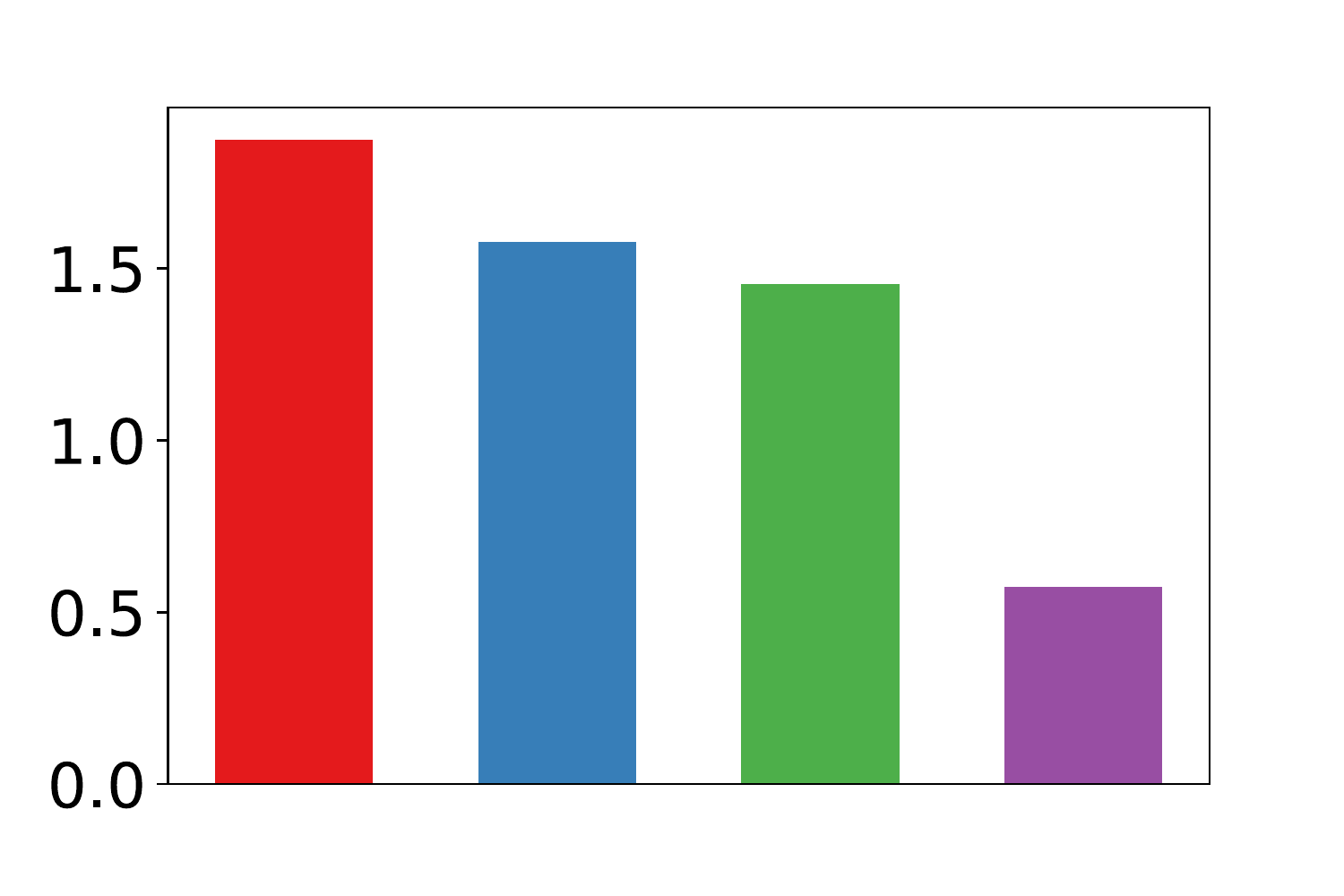
\caption*{Conv Layer 2}
  \end{subfigure}
  \begin{subfigure}[c]{0.24\linewidth} \centering
\def\svgwidth{0.99\columnwidth} \input{./low_rank_folder//figs/spec_lyr1_b1_c1.pdf_tex}
\caption*{Conv Layer 3}
  \end{subfigure}
  \begin{subfigure}[c]{0.24\linewidth} \centering
\def\svgwidth{0.99\columnwidth} 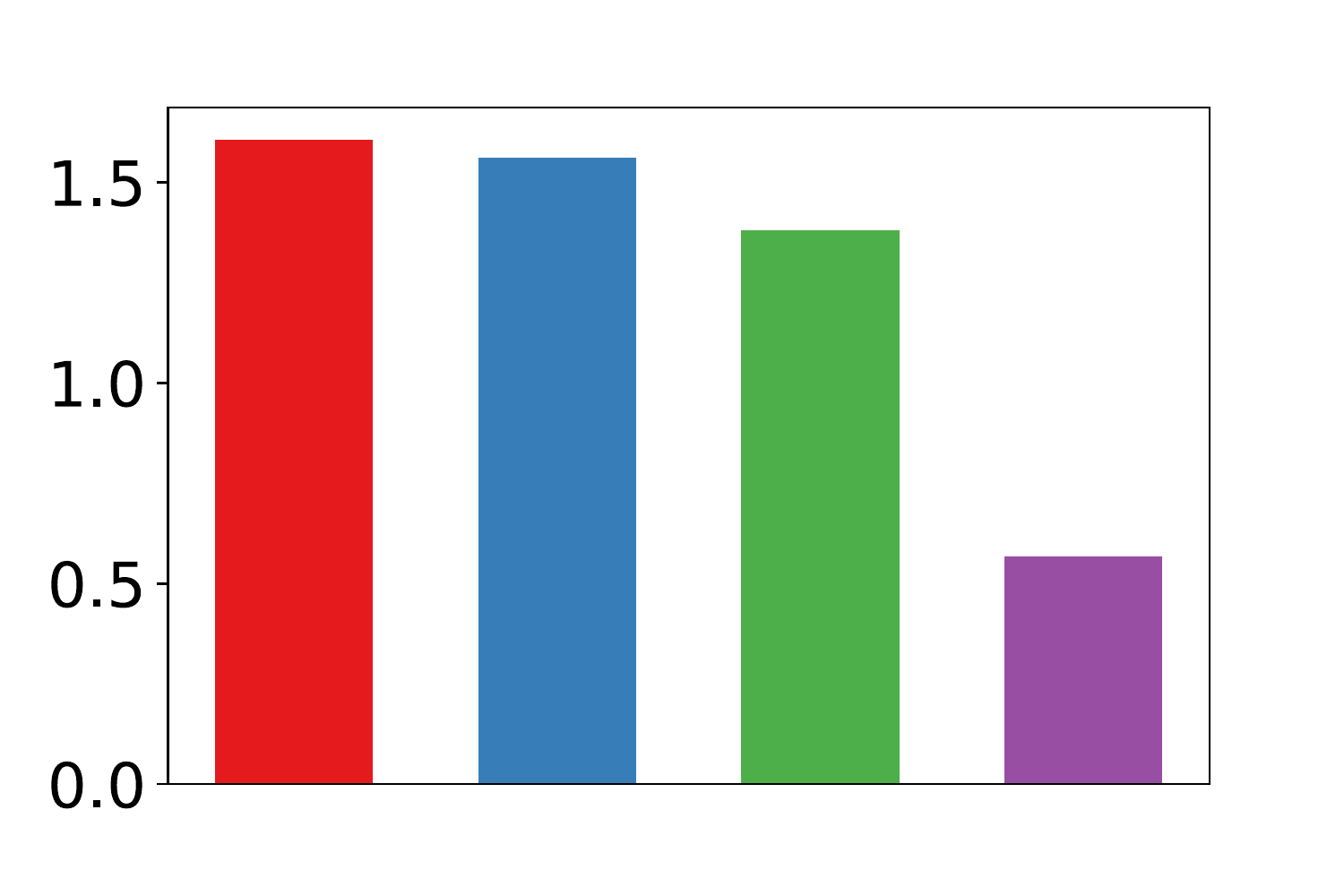
\caption*{Conv Layer 4}
  \end{subfigure}
  \caption{Cushion of ResNet-18 Block 1 on CIFAR10}
  \label{fig:int_spec_lyr_cush}
\end{subfigure}
\begin{subfigure}[c]{1.0\linewidth} 
  \begin{subfigure}[c]{0.24\linewidth} \centering
\def\svgwidth{0.99\columnwidth} 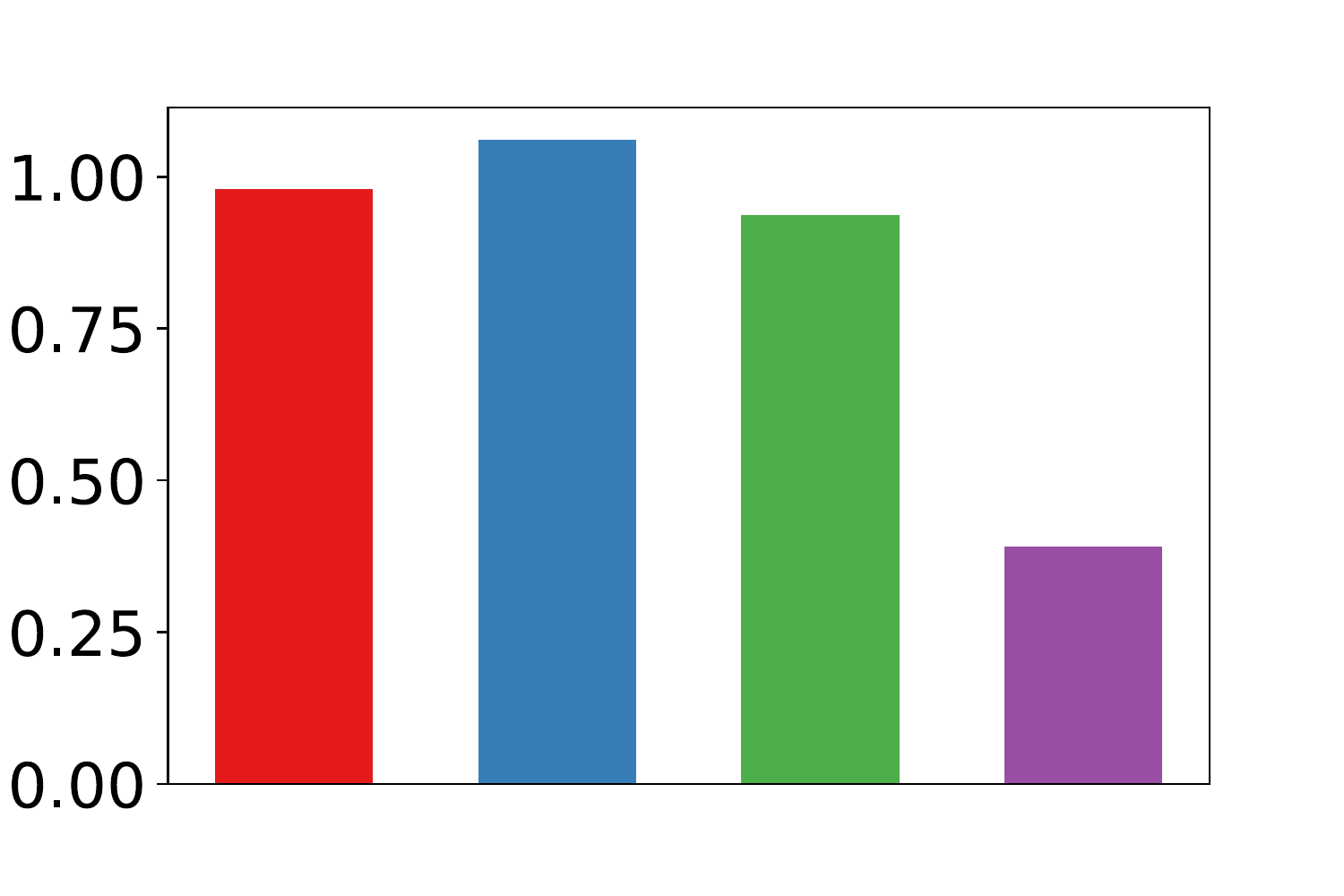
\caption*{Conv Layer 1}
  \end{subfigure}
  \begin{subfigure}[c]{0.24\linewidth} \centering
\def\svgwidth{0.99\columnwidth} 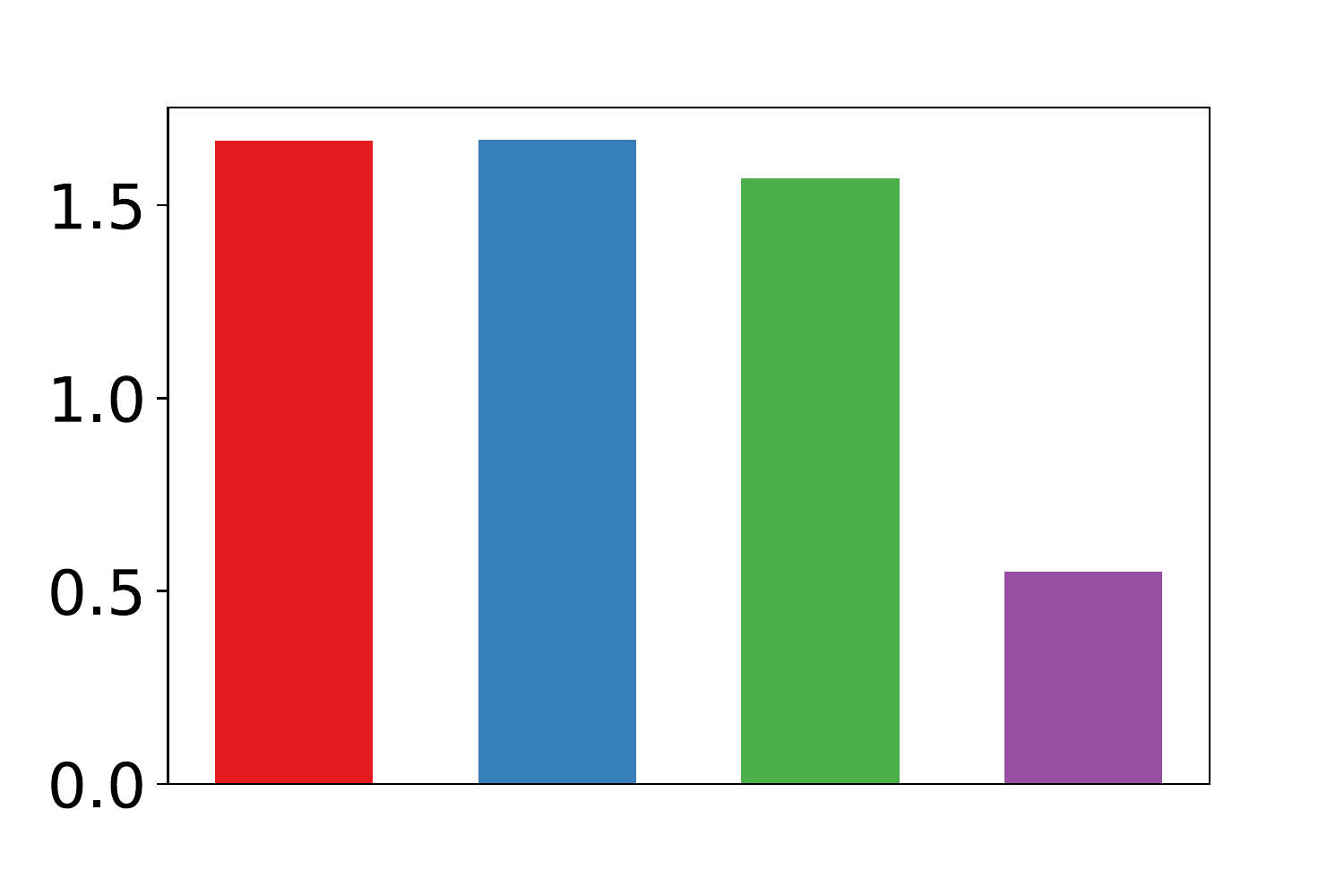
\caption*{Conv Layer 2}
  \end{subfigure}
  \begin{subfigure}[c]{0.24\linewidth} \centering
\def\svgwidth{0.99\columnwidth} 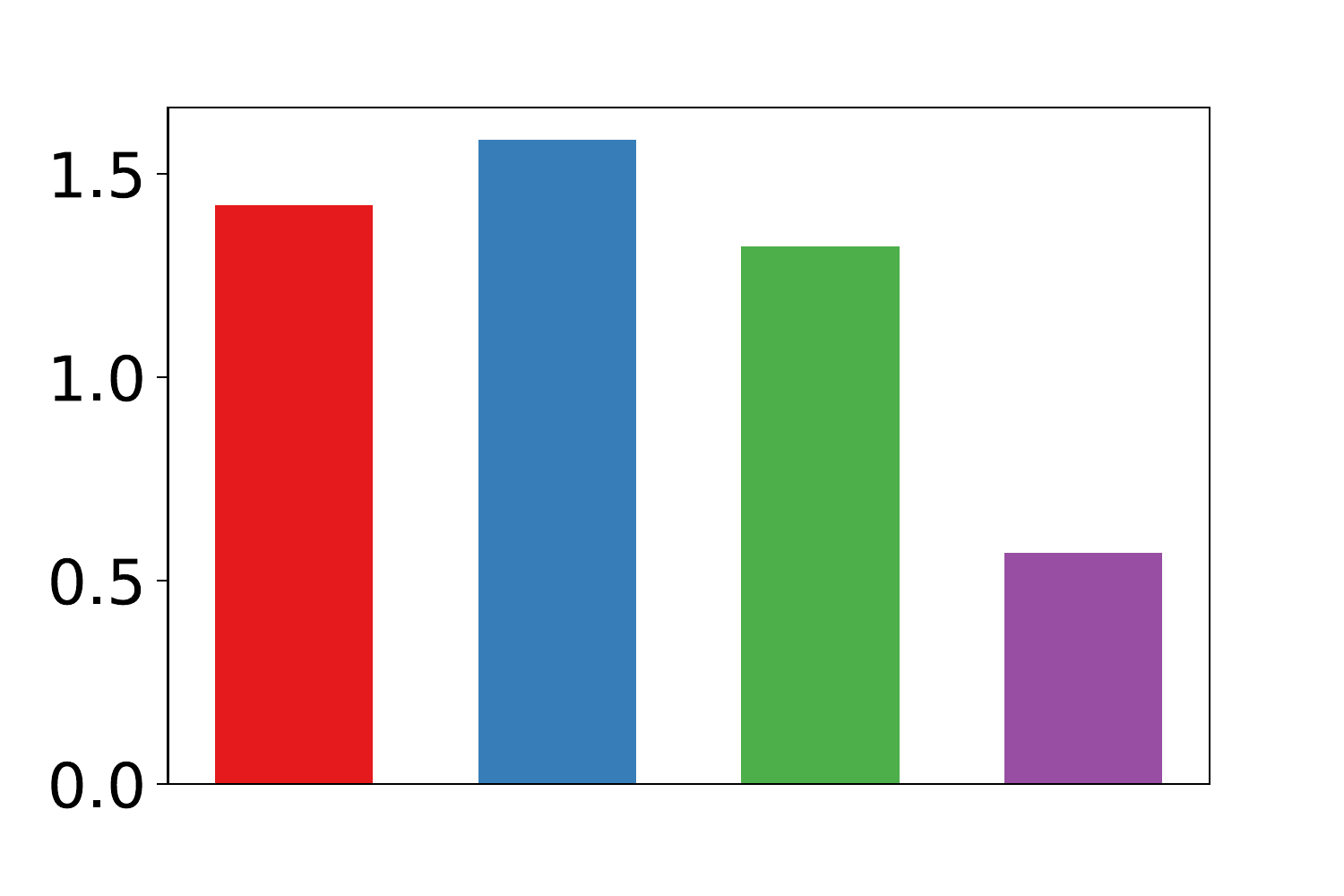
\caption*{Conv Layer 3}
\end{subfigure}
\begin{subfigure}[c]{0.24\linewidth} \centering
\def\svgwidth{0.99\columnwidth} 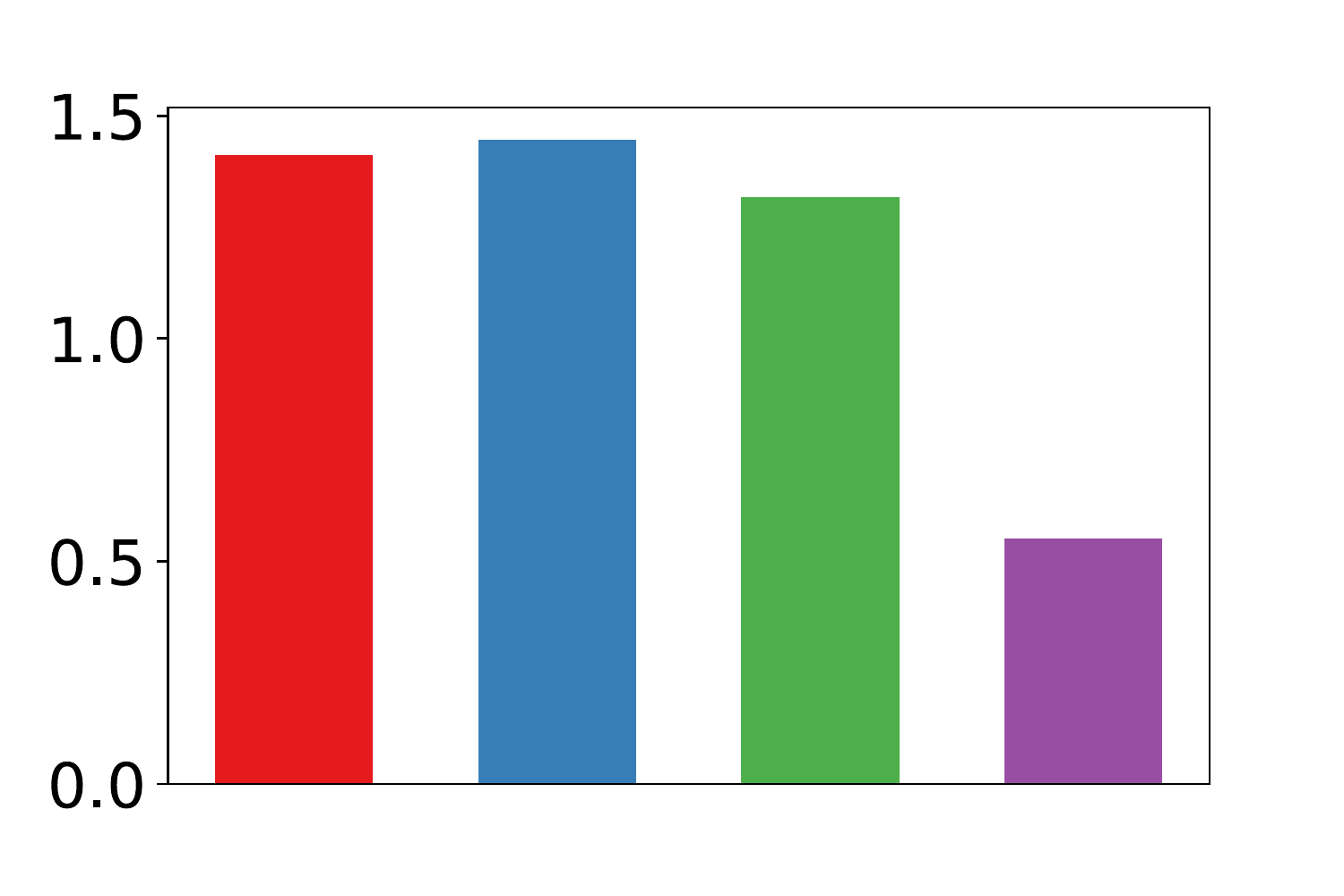
\caption*{Conv Layer 4}
  \end{subfigure}
  \caption{Cushion of ResNet-18 Block 2 on CIFAR10}
  \label{fig:int_spec_lyr2_cush}
\end{subfigure}

\begin{subfigure}[c]{1.0\linewidth} 
  \begin{subfigure}[c]{0.24\linewidth} \centering
\def\svgwidth{0.99\columnwidth} 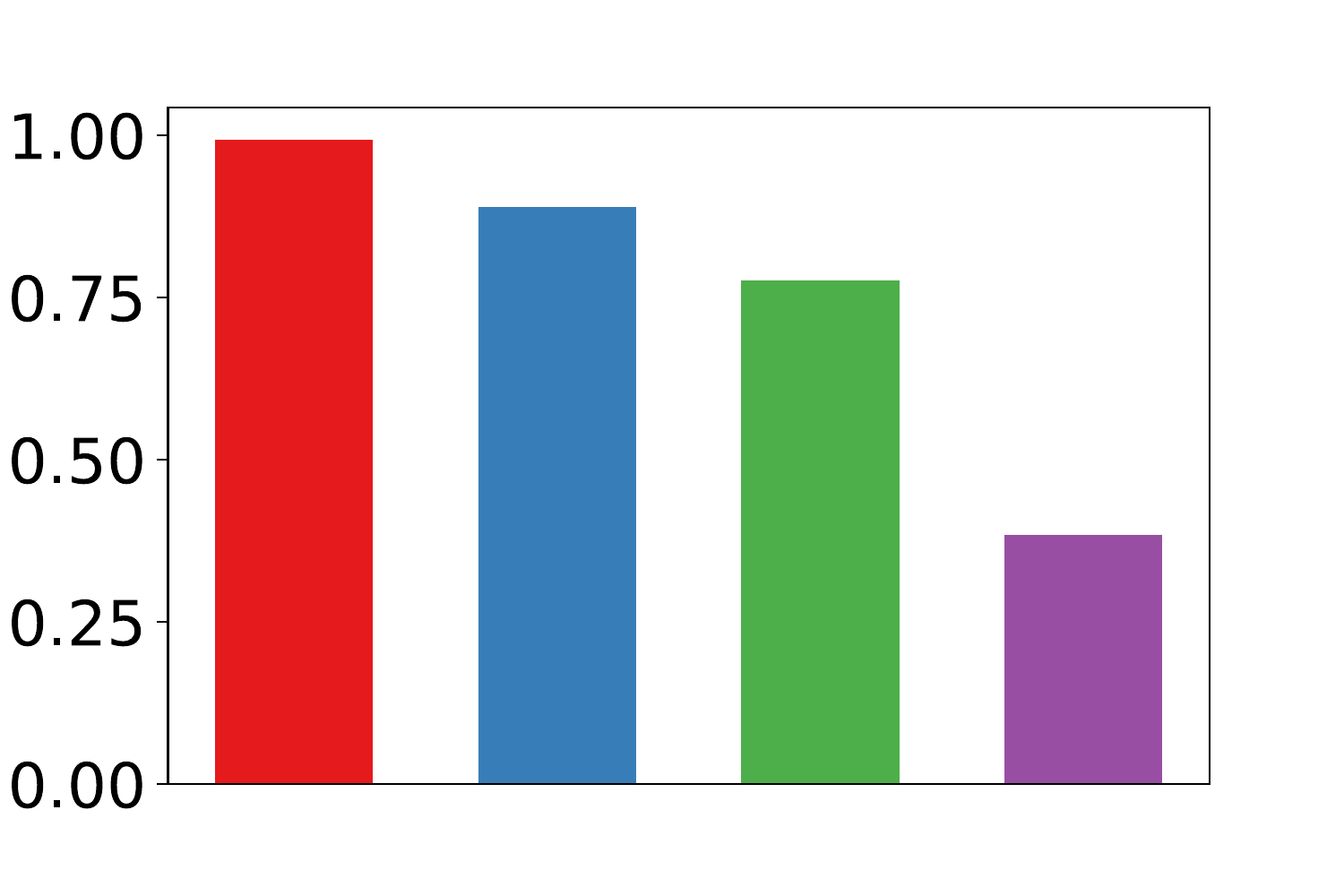
\caption*{Conv Layer 1}
  \end{subfigure}
  \begin{subfigure}[c]{0.24\linewidth} \centering
\def\svgwidth{0.99\columnwidth} 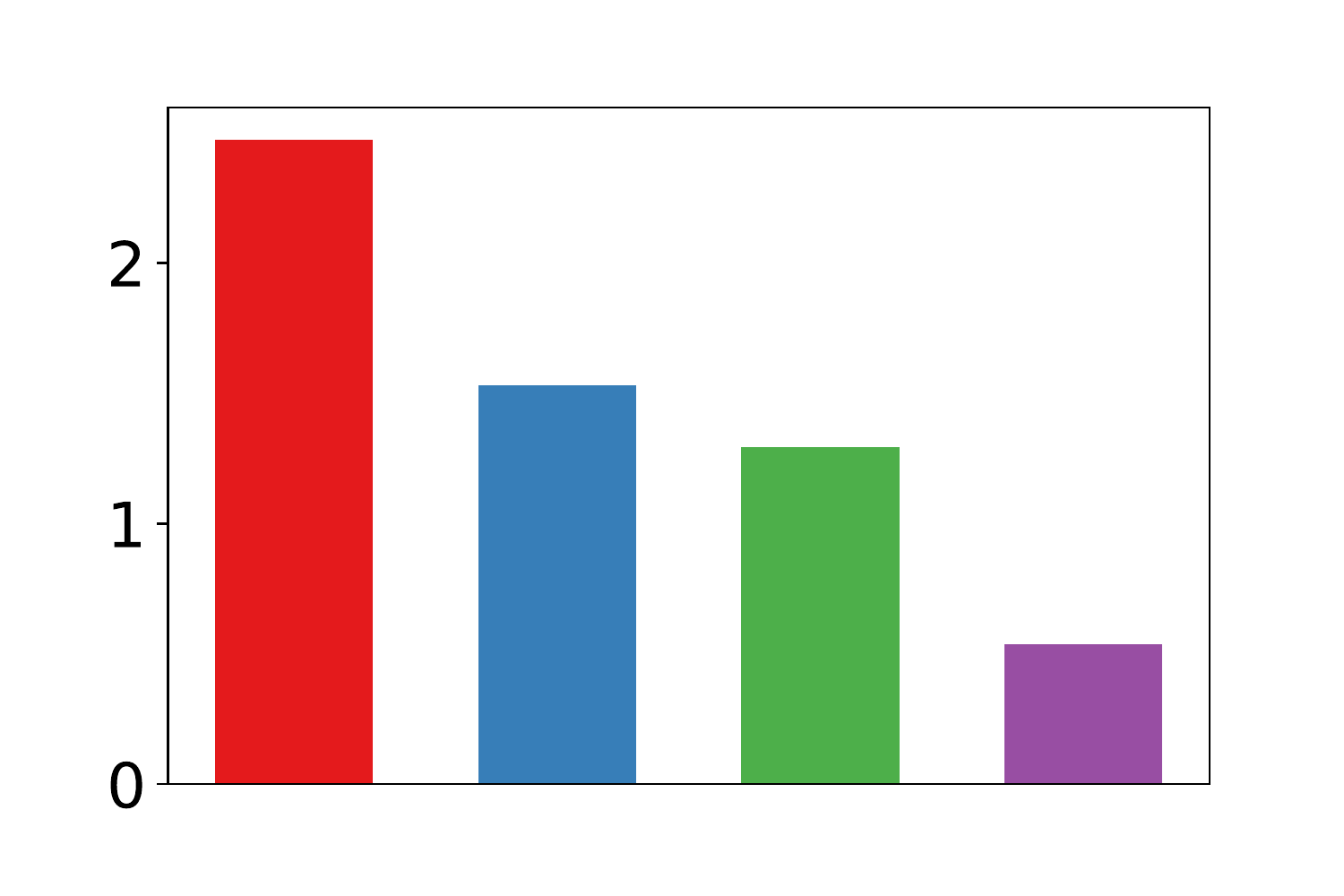
\caption*{Conv Layer 2}
  \end{subfigure}
  \begin{subfigure}[c]{0.24\linewidth} \centering
\def\svgwidth{0.99\columnwidth} 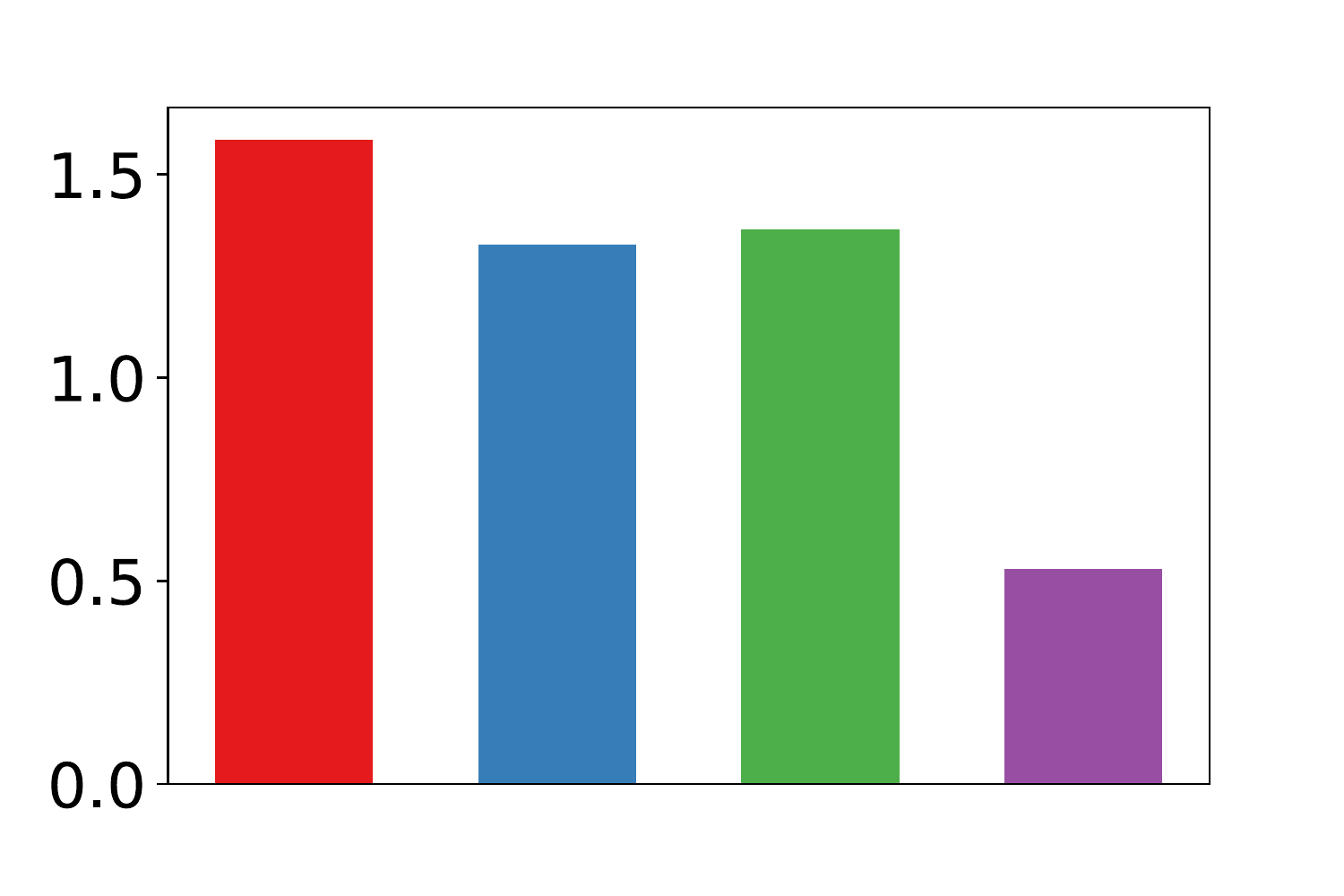
\caption*{Conv Layer 3}
  \end{subfigure}
  \begin{subfigure}[c]{0.24\linewidth} \centering
\def\svgwidth{0.99\columnwidth} 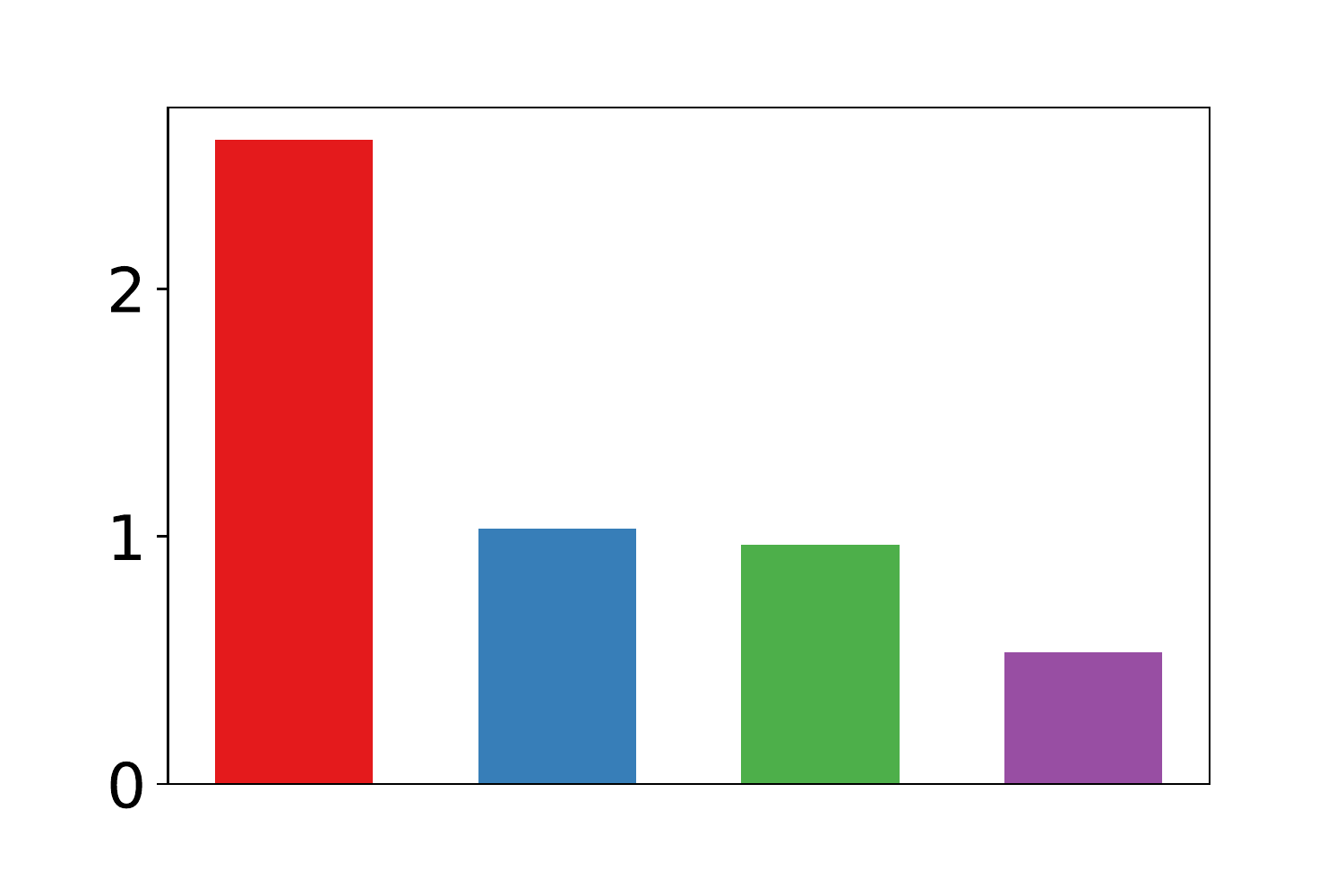
\caption*{Conv Layer 4}
  \end{subfigure}
  \caption{Cushion of ResNet-18 Block 3 on CIFAR10}
  \label{fig:int_spec_lyr3_cush}
\end{subfigure}
    \begin{subfigure}[c]{1.0\linewidth} 
  \begin{subfigure}[c]{0.24\linewidth} \centering
\def\svgwidth{0.99\columnwidth} 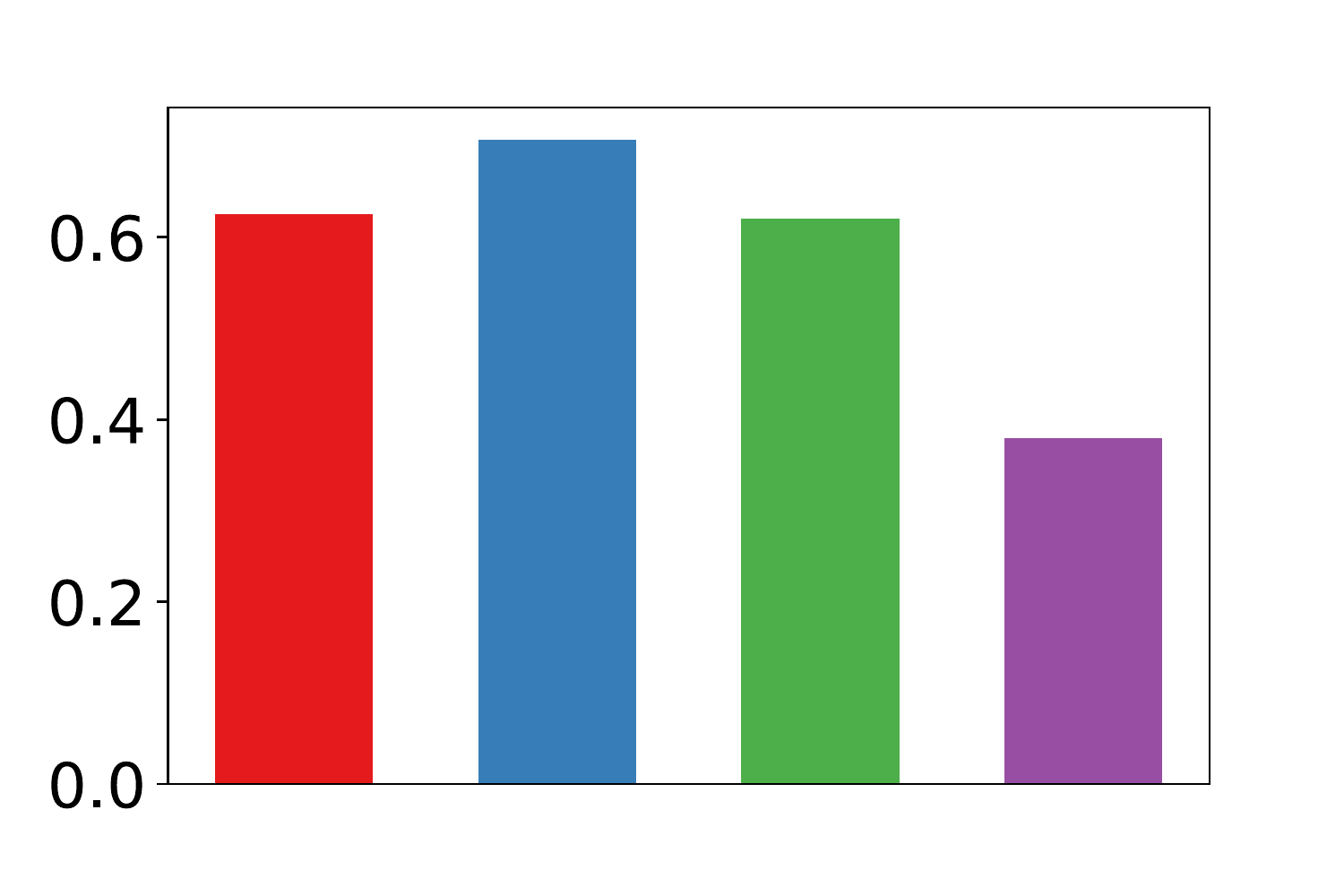
\caption*{Conv Layer 1}
  \end{subfigure}
  \begin{subfigure}[c]{0.24\linewidth} \centering
\def\svgwidth{0.99\columnwidth} 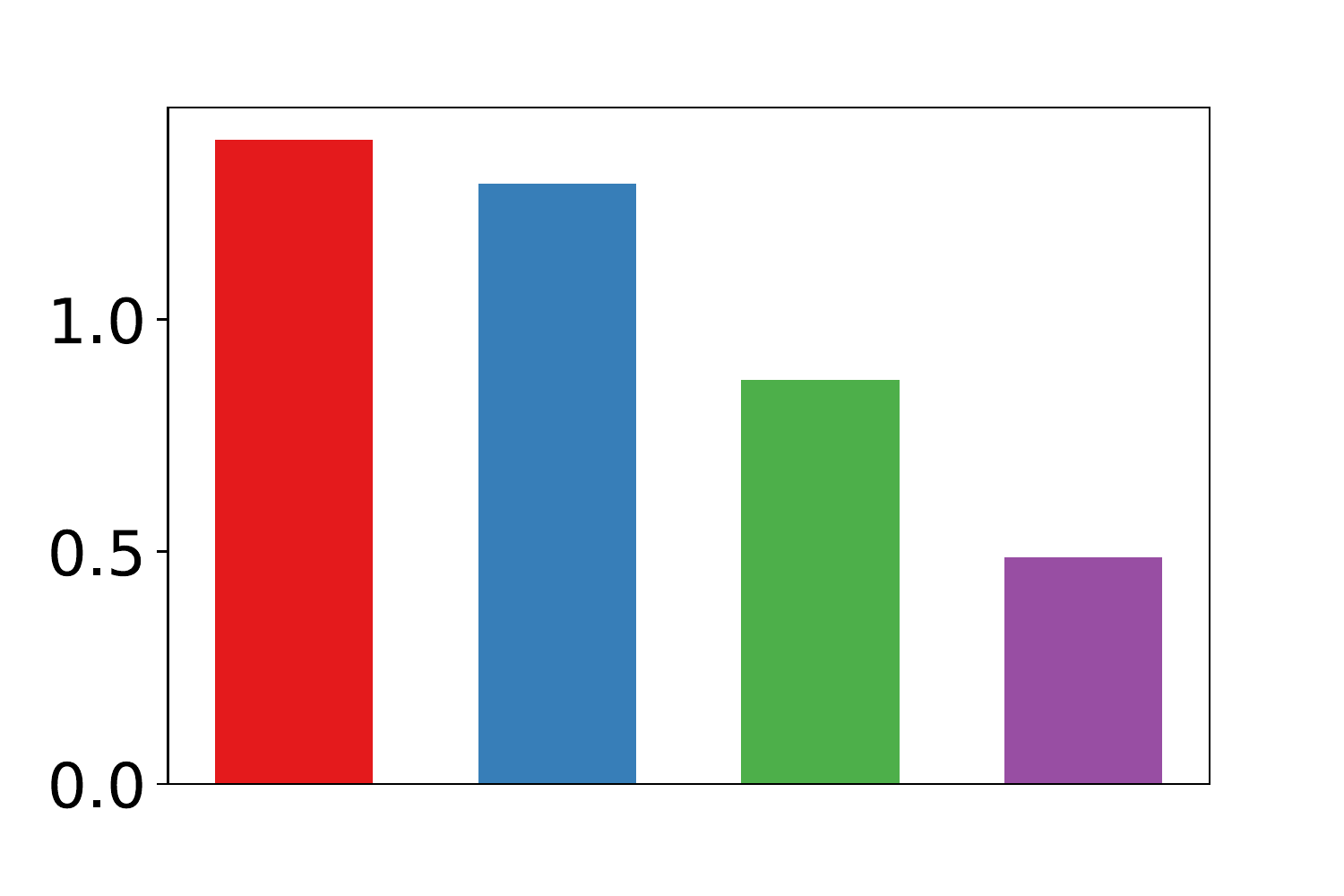
\caption*{Conv Layer 2}
  \end{subfigure}
  \begin{subfigure}[c]{0.24\linewidth} \centering
\def\svgwidth{0.99\columnwidth} 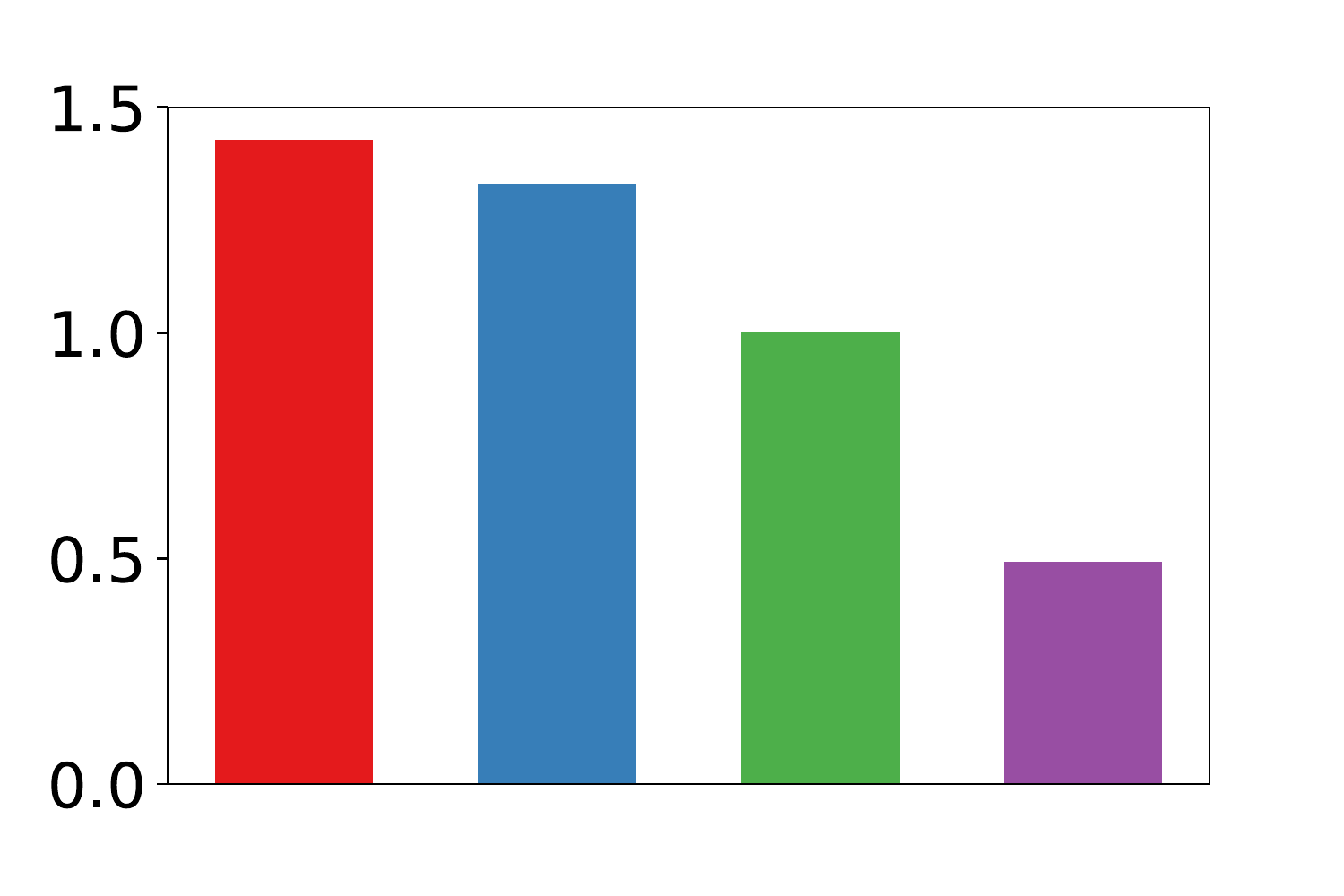
\caption*{Conv Layer 3}
\end{subfigure}
 \begin{subfigure}[c]{0.24\linewidth} \centering
\def\svgwidth{0.99\columnwidth} 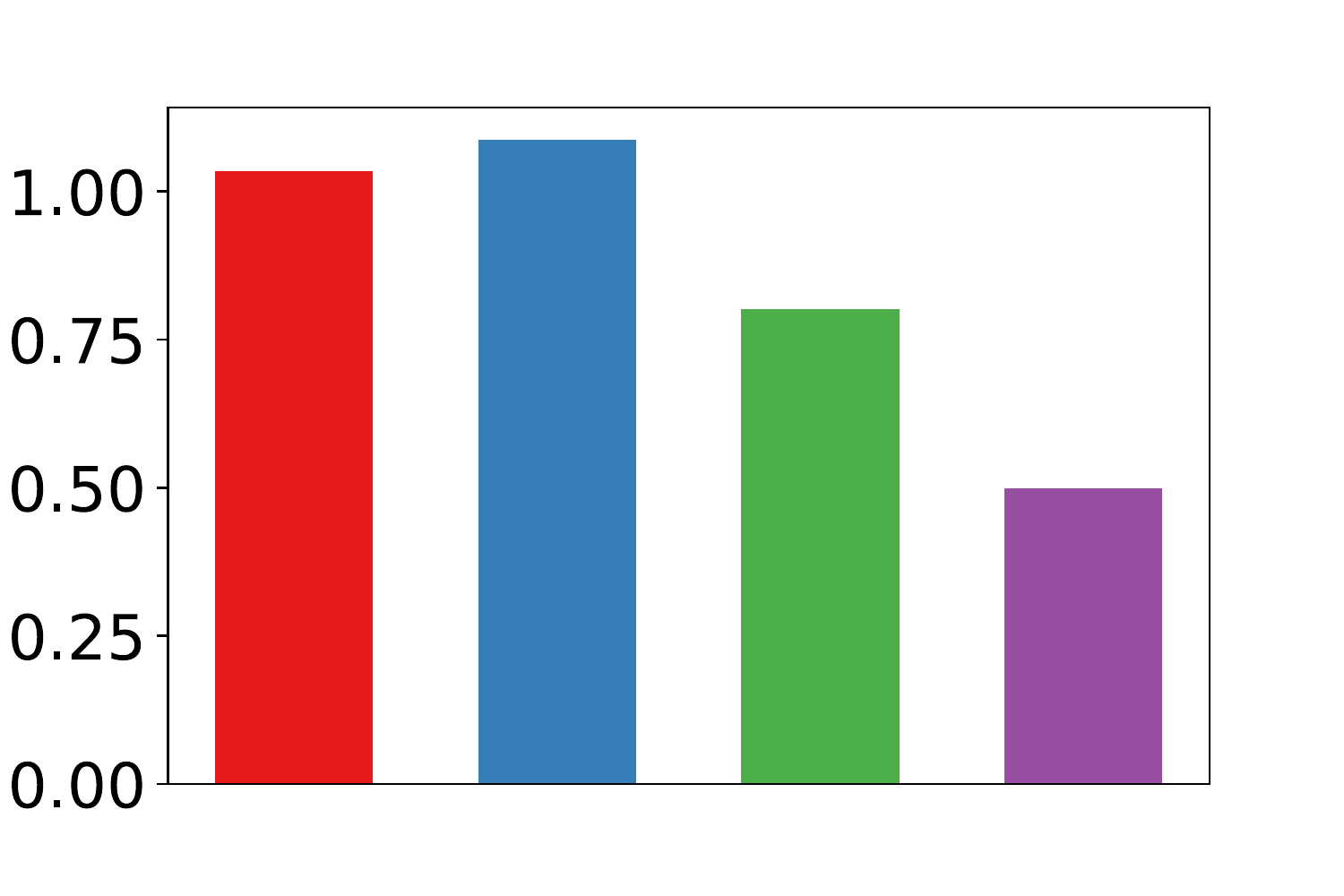
\caption*{Conv Layer 4}
  \end{subfigure}
  \caption{Cushion of ResNet-18 Block 4 on CIFAR10.}
  \label{fig:int_spec_lyr4_cush}
\end{subfigure}
\caption[Layer Cushion of ResNet-18 on CIFAR10]{Layer Cushion of various layers of a ResNet-18 trained on CIFAR10. Each ResNet-18 has
four blocks with each block containing four convolutional layers. }
\label{fig:cifar10-resnet18-lyr-cushion}
\end{figure}

  \begin{figure}[H]

    \begin{subfigure}[c]{1.0\linewidth} 
  \begin{subfigure}[c]{0.24\linewidth} \centering
\def\svgwidth{0.99\columnwidth}
\input{./low_rank_folder//figs/svhn_spec_lyr1_b1_c1.pdf_tex}
\caption*{Conv Layer 1}
  \end{subfigure}
  \begin{subfigure}[c]{0.24\linewidth} \centering
\def\svgwidth{0.99\columnwidth}
\input{./low_rank_folder//figs/svhn_spec_lyr1_b1_c2.pdf_tex}
\caption*{Conv Layer 2}
  \end{subfigure}
  \begin{subfigure}[c]{0.24\linewidth} \centering
\def\svgwidth{0.99\columnwidth}
\input{./low_rank_folder//figs/svhn_spec_lyr1_b2_c1.pdf_tex}
\caption*{Conv Layer 3}
  \end{subfigure}
  \begin{subfigure}[c]{0.24\linewidth} \centering
\def\svgwidth{0.99\columnwidth}
\input{./low_rank_folder//figs/svhn_spec_lyr1_b2_c2.pdf_tex}
\caption*{Conv Layer 4}
  \end{subfigure}
  \caption{Cushion of ResNet-18 Block 1 on SVHN.}
  \label{fig:int_svhn_spec_lyr_cush}
\end{subfigure}

    \begin{subfigure}[c]{1.0\linewidth} 
  \begin{subfigure}[c]{0.24\linewidth} \centering
\def\svgwidth{0.99\columnwidth}
\input{./low_rank_folder//figs/svhn_spec_lyr2_b1_c1.pdf_tex}
\caption*{Conv Layer 1}
  \end{subfigure}
  \begin{subfigure}[c]{0.24\linewidth} \centering
\def\svgwidth{0.99\columnwidth}
\input{./low_rank_folder//figs/svhn_spec_lyr2_b1_c2.pdf_tex}
\caption*{Conv Layer 2}
  \end{subfigure}
  \begin{subfigure}[c]{0.24\linewidth} \centering
\def\svgwidth{0.99\columnwidth}
\input{./low_rank_folder//figs/svhn_spec_lyr2_b2_c1.pdf_tex}
\caption*{Conv Layer 3}
  \end{subfigure}
  \begin{subfigure}[c]{0.24\linewidth} \centering
\def\svgwidth{0.99\columnwidth}
\input{./low_rank_folder//figs/svhn_spec_lyr2_b2_c2.pdf_tex}
\caption*{Conv Layer 4}
  \end{subfigure}
  \caption{Cushion of ResNet-18 Block 2 on SVHN.}
  \label{fig:int_svhn_spec_lyr2_cush}
\end{subfigure}

    \begin{subfigure}[c]{1.0\linewidth} 
      \begin{subfigure}[c]{0.24\linewidth} \centering
\def\svgwidth{0.99\columnwidth}
\input{./low_rank_folder//figs/svhn_spec_lyr3_b1_c1.pdf_tex}
\caption*{Conv Layer 1}
  \end{subfigure}
  \begin{subfigure}[c]{0.24\linewidth} \centering
\def\svgwidth{0.99\columnwidth}
\input{./low_rank_folder//figs/svhn_spec_lyr3_b1_c2.pdf_tex}
\caption*{Conv Layer 2}
  \end{subfigure}
  \begin{subfigure}[c]{0.24\linewidth} \centering
\def\svgwidth{0.99\columnwidth}
\input{./low_rank_folder//figs/svhn_spec_lyr3_b2_c1.pdf_tex}
\caption*{Conv Layer 3}
  \end{subfigure}
  \begin{subfigure}[c]{0.24\linewidth} \centering
\def\svgwidth{0.99\columnwidth}
\input{./low_rank_folder//figs/svhn_spec_lyr3_b2_c2.pdf_tex}
\caption*{Conv Layer 4}
  \end{subfigure}
  \caption{Cushion of ResNet-18 Block 3 on SVHN.}
  \label{fig:int_svhn_spec_lyr3_cush}
\end{subfigure}
  \begin{subfigure}[c]{1.0\linewidth} 
  \begin{subfigure}[c]{0.24\linewidth} \centering
\def\svgwidth{0.99\columnwidth}
\input{./low_rank_folder//figs/svhn_spec_lyr4_b1_c1.pdf_tex}
\caption*{Conv Layer 1}
  \end{subfigure}
  \begin{subfigure}[c]{0.24\linewidth} \centering
\def\svgwidth{0.99\columnwidth}
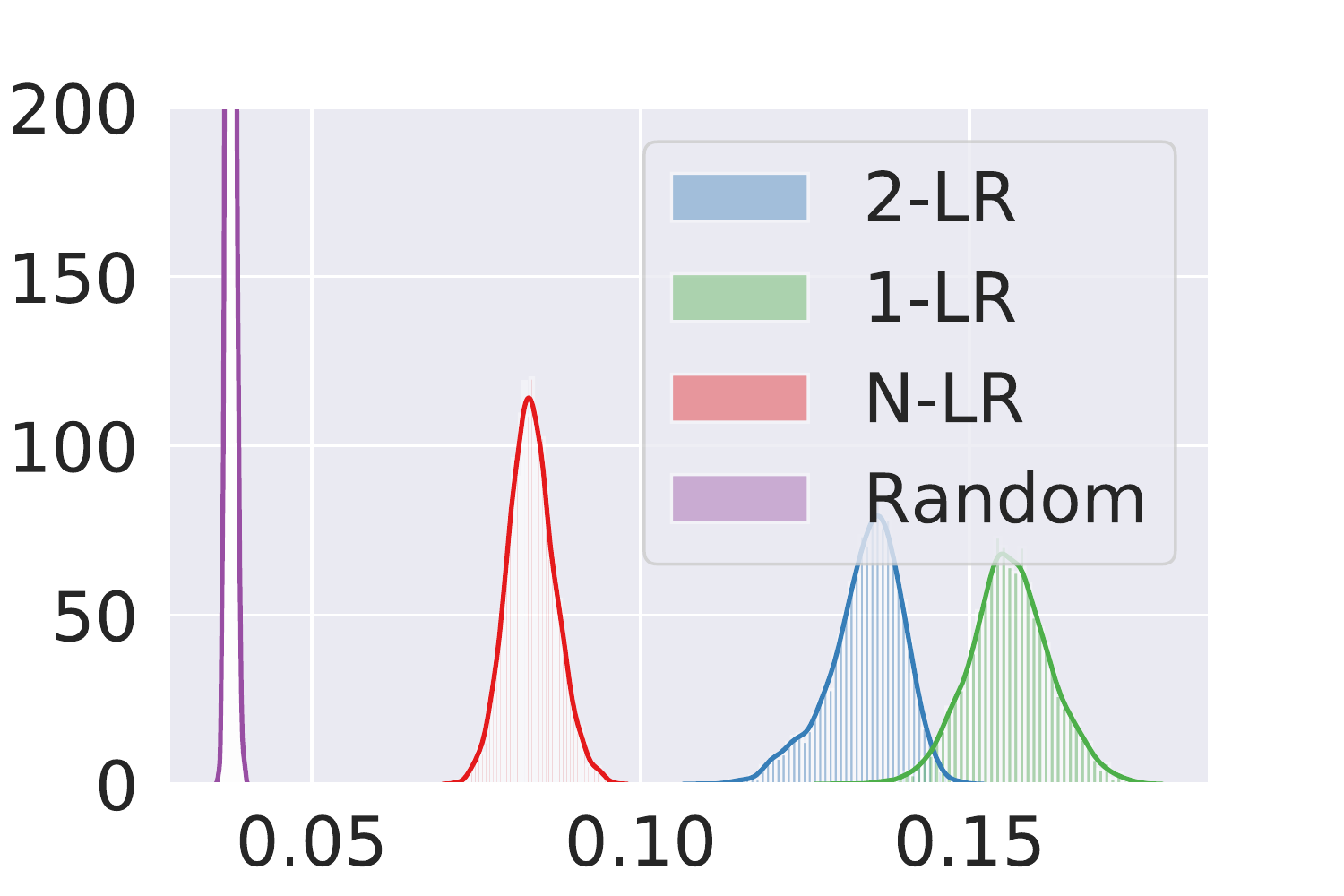
\caption*{Conv Layer 2}
  \end{subfigure}
  \begin{subfigure}[c]{0.24\linewidth} \centering
\def\svgwidth{0.99\columnwidth}
\input{./low_rank_folder//figs/svhn_spec_lyr4_b2_c1.pdf_tex}
\caption*{Conv Layer 3}
  \end{subfigure}
  \begin{subfigure}[c]{0.24\linewidth} \centering
\def\svgwidth{0.99\columnwidth}
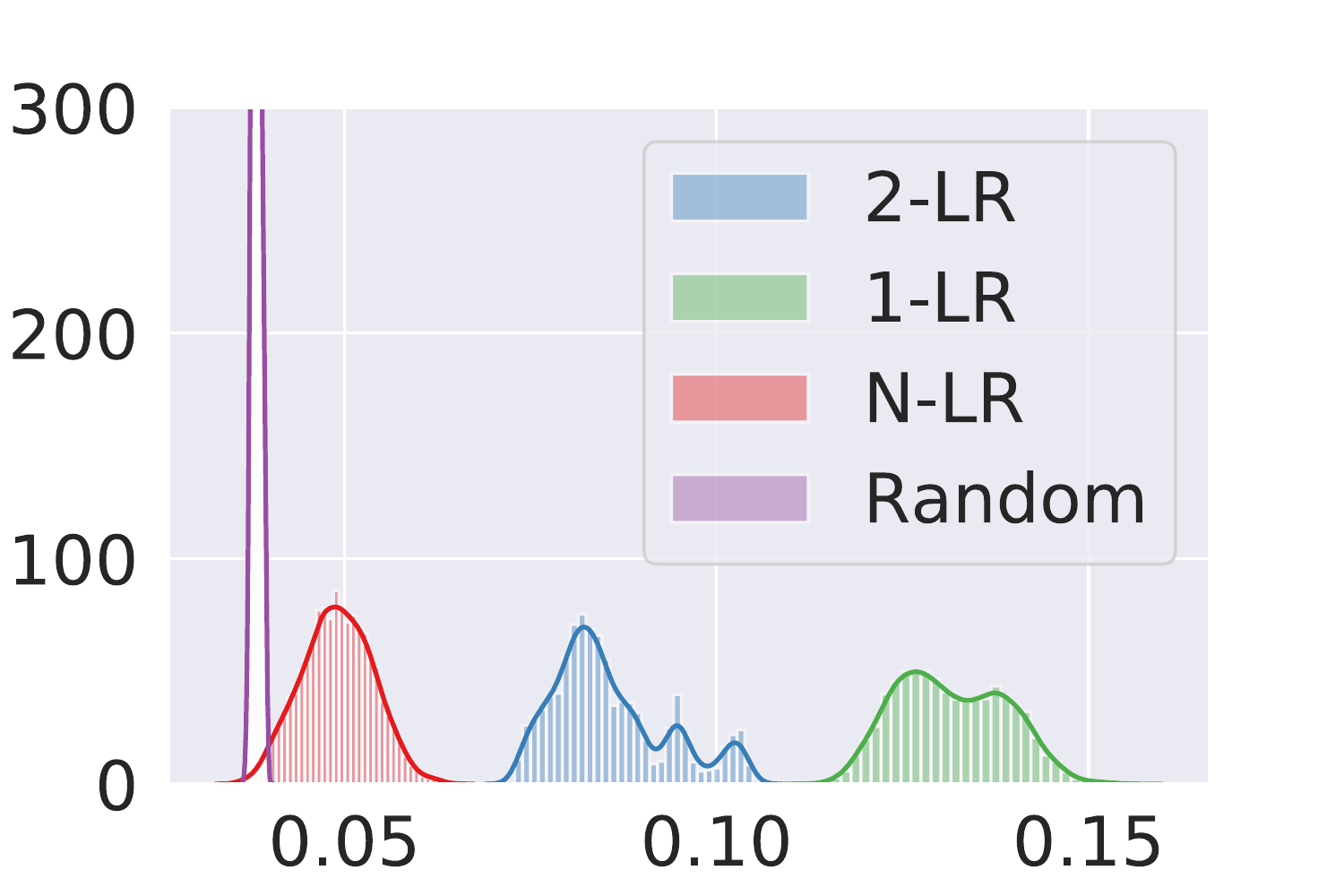
\caption*{Conv Layer 4}
  \end{subfigure}
  \caption{Cushion of ResNet-18 Block 4 on SVHN.}
  \label{fig:int_svhn_spec_lyr4_cush}
\end{subfigure}
\caption[Layer Cushion of ResNet-18 on SVHN]{Layer Cushion of various layers of a ResNet-19 trained on SVHN. Each ResNet-18 has
four blocks with each block containing four convolutional layers. }
\label{fig:svhn-resnet18-lyr-cushion}
\end{figure}

\chapter{Appendix for~\Cref{chap:focal_loss}}
\label{app:focal_loss}

\section{Proofs for~\Cref{sec:focalloss}}
\label{sec:calibration-proof}

\begin{proof}[Proof of~\Cref{thm:focal-reg-Bregman}]
	let $K$ denotes the number of classes, $\mathcal{L}_f$ denote the
	focal loss with parameter $\gamma$, $\cL_c$ denote the cross entropy
	between $\hat{p}$ and $q$, and let $q_y$ denotes the ground-truth
	probability assigned to the $y^{th}$ class (similarly for
	$\hat{p}_y$). We consider the following simple extension of focal
	loss: 
	
	\begin{align*}
	\mathcal{L}_f &= -\sum_{y=1}^K (1 -  \hat{p}_{y})^\gamma q_{y} \log{\hat{p}_{y}}\\
			&\ge -\sum_{y=1}^K(1 - \gamma \hat{p}_{y})q_{y} \log{\hat{p}_{y}} &&\text{By Bernoulli's inequality $\forall \gamma \ge 1$, since $\hat{p}_{y} \in [0,1]$}\\
			&=- \sum_{y=1}^K q_{y} \log{\hat{p}_{y}} - \gamma \left|\sum_{y=1}^K q_{y} \hat{p}_{y} \log{\hat{p}_{y}}\right|&& \text{$\forall y$, $\log{\hat{p}_{y}}\le 0$}\\
			&\ge-\sum_{y=1}^K q_{y} \log{\hat{p}_{y}}  - \gamma\max_j q_{j} \sum_{y=1}^K |\hat{p}_{y} \log{\hat{p}_{y}}|&&\text{By H\"older's inequality $||fg||_1 \leq ||f||_{\infty}||g||_1$}\\ 
			&\ge-\sum_{y=1}^K q_{y} \log{\hat{p}_{y}}  + \gamma\sum_{y=1}^K \hat{p}_{y} \log{\hat{p}_{y}}&&\forall j, q_j \in [0,1]\\
			&= \mathcal{L}_c - \gamma\mathbb{H}[\hat{p}].
	\end{align*}
	
	We know that $\mathcal{L}_c = \mathrm{KL}(q||\hat{p})+
	\mathbb{H}[q]$. Combining this equality with the above inequality
	leads to:
	\begin{align*}
		\mathcal{L}_f \geq \mathrm{KL}(q||\hat{p})+  \underbrace{\mathbb{H}[q]}_{constant} - \gamma \mathbb{H}[\hat{p}].
		\end{align*}
	\end{proof}

Here we provide the proofs of both the Lemmas presented in the main
text. While~\Cref{pro1} helps us understand the regularisation effect
of focal loss,~\Cref{pro:gamma} provides the $\gamma$ values in a principled
way such that it is sample-dependent. Implementing the
sample-dependent $\gamma$ is very easy as implementation of the
Lambert-W function~\citep{Corless1996} is available in
standard libraries (e.g.\ python scipy).

\begin{proof}[Proof of~\Cref{pro1}]
Let $\vec{w}$ be the linear layer parameters connecting the feature
map to the logit $s$. Then, using the chain rule, $\frac{\partial
\cL_f}{\partial \vec{w}} = \Big( \frac{\partial s}{\partial \vec{w}}
\Big) \Big( \frac{\partial \hat{p}_{i,y_i}}{\partial s} \Big) \Big(
\frac{\partial \cL_f}{\partial \hat{p}_{i,y_i}} \Big)$. Similarly,
$\frac{\partial \cL_c}{\partial \vec{w}} = \Big( \frac{\partial
s}{\partial \vec{w}} \Big) \Big( \frac{\partial
\hat{p}_{i,y_i}}{\partial s} \Big) \Big( \frac{\partial
\cL_c}{\partial \hat{p}_{i,y_i}} \Big)$. The derivative of the focal
loss with respect to $\hat{p}_{i,y_i}$, the softmax output of the
network for the true class $y_i$, takes the form
\begin{equation}
\begin{split}
        \frac{\partial \cL_f}{\partial \hat{p}_{i,y_i}} & = - \frac{1}{\hat{p}_{i,y_i}} \Big( (1-\hat{p}_{i,y_i})^\gamma - \gamma \hat{p}_{i,y_i} (1-\hat{p}_{i,y_i})^{\gamma -1} \log(\hat{p}_{i,y_i}) \Big) \nonumber \\
        & = \frac{\partial \cL_c}{\partial \hat{p}_{i,y_i}} g(\hat{p}_{i,y_i}, \gamma),
\end{split}
\end{equation}
in which $g(\hat{p}_{i,y_i}, \gamma) = (1-\hat{p}_{i,y_i})^\gamma -
\gamma \hat{p}_{i,y_i} (1-\hat{p}_{i,y_i})^{\gamma -1}
\log(\hat{p}_{i,y_i})$ and $ \frac{\partial \cL_c}{\partial
\hat{p}_{i,y_i}}  = -\frac{1}{\hat{p}_{i,y_i}}$. It is thus
straightforward to verify that if $g(\hat{p}_{i,y_i}, \gamma) \in
[0,1]$, then $\norm{\frac{\partial  \cL_f}{\partial \hat{p}_{i,y_i}}}
\leq \norm{\frac{\partial  \cL_c}{\partial \hat{p}_{i,y_i}}}$, which
itself implies that $\norm{\frac{\partial  \cL_f}{\partial \vec{w}}}
\leq \norm{\frac{\partial  \cL_c}{\partial \vec{w}}}$.
\end{proof}

\begin{proof}[Proof of~\Cref{pro:gamma}]
We derive the value of $\gamma > 0$ for which $g(p_0, \gamma)=1$ for a given $p_0 \in [0, 1]$. From Proposition 4.1, we already know that
\begin{equation}
    \frac{\partial \cL_f}{\partial \hat{p}_{i,y_i}} = \frac{\partial \cL_c}{\partial \hat{p}_{i,y_i}} g(\hat{p}_{i,y_i}, \gamma),
\end{equation}
where $\cL_f$ is focal loss, $\cL_c$ is cross entropy loss, $\hat{p}_{i,y_i}$ is the probability assigned by the model to the ground-truth correct class for the $i^{th}$ sample, and
\begin{equation}
    g(\hat{p}_{i,y_i}, \gamma) = (1-\hat{p}_{i,y_i})^\gamma - \gamma \hat{p}_{i,y_i} (1-\hat{p}_{i,y_i})^{\gamma -1} \log(\hat{p}_{i,y_i}).
\end{equation}
For $p \in [0, 1]$, if we look at the function $g(p, \gamma)$, then we can clearly see that $g(p, \gamma) \rightarrow 1$ as $p \rightarrow 0$, and that $g(p, \gamma) = 0$ when $p = 1$. To observe the behaviour of $g(p, \gamma)$ for intermediate values of $p$, we first take its derivative with respect to $p$:
\begin{equation}
    \label{eq:11}
        \frac{\partial g(p, \gamma)}{\partial p} = \gamma (1-p)^{\gamma-2} \big[-2(1-p)-(1-p)\log p + (\gamma-1) p \log p\big]
\end{equation}
In Equation \ref{eq:11}, $\gamma(1-p)^{\gamma-2} > 0$ except when $p = 1$ (in which case $\gamma(1-p)^{\gamma-2} = 0$). Thus, to observe the sign of the gradient $\frac{\partial g(p, \gamma)}{\partial p}$, we focus on the term
\begin{equation}
\label{eq:12}
    -2(1-p)-(1-p)\log p + (\gamma-1)p \log p.
\end{equation}
Dividing Equation \ref{eq:12} by $(-\log p)$, the sign remains unchanged and we get
\begin{equation}
    k(p, \gamma) = \frac{2(1-p)}{\log p} + 1 - \gamma p.
\end{equation}
We can see that $k(p,\gamma) \rightarrow 1$ as $p \rightarrow 0$ and $k(p,\gamma) \rightarrow -(1+\gamma)$ as $p \rightarrow 1$ (using l'H{\^o}pital's rule). Furthermore, $k(p, \gamma)$ is monotonically decreasing for $p \in [0, 1]$. Thus, as the gradient $\frac{\partial g(p, \gamma)}{\partial p}$ is positive initially starting from $p = 0$ and negative later till $p = 1$, we can say that $g(p, \gamma)$ first monotonically increases starting from $1$ (as $p\rightarrow 0$) and then monotonically decreases down to $0$ (at $p = 1$). Thus, if for some threshold $p_0 > 0$ and for some $\gamma > 0$, $g(p, \gamma) = 1$, then $\forall p > p_0$, $g(p, \gamma) < 1$. We now want to find a $\gamma$ such that $\forall p \geq p_0$, $g(p, \gamma) \le 1$. First, let $a=(1-p_0)$ and $b=p_0\log p_0$. Then:
\begin{equation}
\begin{split}
\label{eq:exp_gamma}
    &g(p_0, \gamma) = (1-p_0)^\gamma-\gamma p_0 (1-p_0)^{\gamma-1}\log p_0 \le 1\\
    \implies &(1-p_0)^{\gamma-1}[(1-p_0)-\gamma p_0\log p_0] \le 1\\
    \implies &a^{\gamma -1}(a-\gamma b) \le 1\\
    \implies &(\gamma -1)\log a + \log (a-\gamma b) \le 0\\
    \implies &\Big( \gamma-\frac{a}{b} \Big) \log a+\log(a-\gamma b) \le \Big( 1-\frac{a}{b} \Big)\log a\\
    \implies &(a-\gamma b)e^{(\gamma -a/b)\log a} \le a^{(1-a/b)}\\
    \implies &\Big( \gamma-\frac{a}{b} \Big) e^{(\gamma -a/b)\log a} \le -\frac{a^{(1-a/b)}}{b}\\
    \implies &\Big( \Big( \gamma-\frac{a}{b} \Big) \log a\Big) e^{(\gamma -a/b)\log a} \ge -\frac{a^{(1-a/b)}}{b}\log a
\end{split}
\end{equation}
where $a=(1-p_0)$ and $b=p_0\log p_0$. We know that the inverse of $y=x e^x$ is defined as $x=W(y)$, where $W$ is the Lambert-W function~\citep{Corless1996}. Furthermore, the r.h.s. of the inequality in Equation \ref{eq:exp_gamma} is always negative, with a minimum possible value of $-1/e$ that occurs at $p_0=0.5$. Therefore, applying the Lambert-W function to the r.h.s.\ will yield two real solutions (corresponding to a principal branch denoted by $W_0$ and a negative branch denoted by $W_{-1}$). We first consider the solution corresponding to the negative branch (which is the smaller of the two solutions):
\begin{equation}
\begin{split}
\label{eq:exp_gamma_cont}
    &\Big((\gamma-\frac{a}{b})\log a\Big) \le W_{-1}\Big(-\frac{a^{(1-a/b)}}{b}\log a\Big)\\
    \implies &\gamma \ge \frac{a}{b}+\frac{1}{\log a}W_{-1}\Big(-\frac{a^{(1-a/b)}}{b}\log a\Big)\\
\end{split}
\end{equation}
If we consider the principal branch, the solution is
\begin{equation}
    \gamma \le \frac{a}{b}+\frac{1}{\log a}W_{0}\Big(-\frac{a^{(1-a/b)}}{b}\log a\Big),
\end{equation}
which yields a negative value for $\gamma$ that we discard. Thus Equation \ref{eq:exp_gamma_cont} gives the values of $\gamma$ for which if $p>p_0$, then $g(p,\gamma) < 1$. In other words, $g(p_0, \gamma) = 1$, and for any $p < p_0$, $g(p, \gamma) > 1$.
\end{proof}

\clearpage
\addcontentsline{toc}{chapter}{\numberline{}Bibliography}%
\bibliography{thesis_ref}
\bibliographystyle{plainnat}
\clearpage

\end{document}